\documentclass[11pt,openany]{book}

\usepackage[T1]{fontenc}
\usepackage{lmodern}
\usepackage{microtype}
\usepackage[margin=1.05in,headheight=15pt]{geometry}
\usepackage{setspace}
\usepackage{amsmath,amssymb,amsthm,mathtools,bm}
\usepackage{physics}
\usepackage{booktabs,array}

\usepackage{graphicx}
\usepackage{tikz}
\usetikzlibrary{arrows.meta,calc,positioning}

\usepackage[dvipsnames]{xcolor}
\definecolor{prismblue}{HTML}{16324F}
\definecolor{prismteal}{HTML}{2A7F7F}
\definecolor{prismgold}{HTML}{C8963E}
\definecolor{softgray}{HTML}{F3F5F6}
\usepackage{titlesec}
\usepackage{fancyhdr}
\usepackage{enumitem}
\usepackage{natbib}
\usepackage{hyperref}
\usepackage[nameinlink,noabbrev]{cleveref}

\setlist{nosep,leftmargin=*}
\hypersetup{
  pdftitle={Modern Optimal Control and Reinforcement Learning for Robotics},
  pdfauthor={Yonghyeon Lee},
  colorlinks=true,
  linkcolor=prismblue,
  citecolor=prismteal,
  urlcolor=prismteal
}

\titleformat{\chapter}[display]
  {\normalfont\bfseries\color{prismblue}}
  {\filleft\fontsize{54}{54}\selectfont\thechapter}
  {-1ex}
  {\titlerule[1.2pt]\vspace{1.2ex}\fontsize{22}{27}\selectfont\raggedright}
\titleformat{name=\chapter,numberless}[display]
  {\normalfont\bfseries\color{prismblue}}
  {}
  {0pt}
  {\Huge}
\titleformat{\section}
  {\Large\bfseries\color{prismblue}}
  {\thesection}{0.7em}{}
\titleformat{\subsection}
  {\large\bfseries\color{prismteal}}
  {\thesubsection}{0.7em}{}

\renewcommand{\sectionmark}[1]{\markright{Section~\thesection}}

\fancypagestyle{plain}{\fancyhf{}\fancyfoot[C]{\color{prismblue}\thepage}}

\newtheoremstyle{prismtheorem}
  {0.8em}{0.8em}{\itshape}{}
  {\color{prismblue}\bfseries}{.}{0.55em}{}
\theoremstyle{prismtheorem}
\newtheorem{theorem}{Theorem}[chapter]
\newtheorem{proposition}[theorem]{Proposition}

\theoremstyle{definition}
\newtheorem{definition}[theorem]{Definition}

\newtheorem{example}[theorem]{Example}
\newtheorem{exercise}[theorem]{Exercise}

\theoremstyle{remark}
\newtheorem*{remark}{Remark}

\newcommand{\R}{\mathbb{R}}

\newcommand{\E}{\mathbb{E}}
\newcommand{\Prob}{\mathbb{P}}

\newcommand{\trans}{^{\mathsf{T}}}

\DeclareMathOperator*{\argmin}{arg\,min}
\DeclareMathOperator*{\argmax}{arg\,max}
\DeclareMathOperator*{\minimize}{minimize}
\DeclareMathOperator*{\maximize}{maximize}
\newcommand{\subjectto}{\text{subject to}}

\newcommand{\chapteroverview}[1]{%
  \begin{quote}
    \color{prismteal}\small\textbf{Chapter map.} #1
  \end{quote}
  \vspace{0.5\baselineskip}
}

\begin{document}

\frontmatter
\begin{titlepage}
\thispagestyle{empty}
\begin{tikzpicture}[remember picture,overlay]
  \fill[prismblue] (current page.north west)
    rectangle ([yshift=-1.55cm]current page.north east);
  \fill[prismgold] ([yshift=-1.55cm]current page.north west)
    rectangle ([yshift=-1.62cm]current page.north east);
  \draw[prismteal,line width=1.2pt]
    ([xshift=1.25in,yshift=1.65in]current page.south west) --
    ([xshift=-1.25in,yshift=1.65in]current page.south east);
\end{tikzpicture}

\vspace*{2.2cm}

{\color{prismteal}\large LECTURE NOTES}\par
\vspace{0.8cm}
{\color{prismblue}\fontsize{34}{40}\selectfont\bfseries
Optimization,\par 
Optimal Control,\par
and Reinforcement Learning\par
for Robotics\par}



\begin{flushright}
{\Large\bfseries Yonghyeon Lee}\par
\vspace{0.25cm}
{\large Department of Artificial Intelligence}\par
{\large Yonsei University}\par
\vspace{0.6cm}
{\color{prismteal}\large Fall 2026}
\end{flushright}

\vspace*{1.9cm}
\end{titlepage}

\cleardoublepage
\chapter*{Preface}
\addcontentsline{toc}{chapter}{Preface}

\begingroup
\small
\setstretch{1.0}
\setlength{\parskip}{0.25\baselineskip}

These notes introduce \textbf{optimal control and reinforcement learning
as two complementary perspectives on sequential decision-making}, with a
particular emphasis on robotics. Optimal control traditionally begins with
mathematical models, objectives, and the design of feedback laws, whereas
reinforcement learning emphasizes learning effective decision-making
strategies through interaction and experience. Despite their different
starting points, the two fields share a common mathematical foundation in
optimization, dynamical systems, value functions, and Bellman recursions.

The intended audience is \textbf{graduate students, researchers, and
engineers working in robotics and related fields}. The assumed
prerequisites are undergraduate-level linear algebra, multivariable
calculus, and probability, along with basic familiarity with robot
kinematics and dynamics. Part~I establishes the mathematical foundations
of optimization, while subsequent chapters introduce additional
concepts as needed.

The goal is to provide a \textbf{concise, self-contained introduction},
rather than an exhaustive survey of either field. Mathematical
derivations are included when they clarify underlying principles,
algorithms, or modeling choices, while technical regularity conditions
and specialized proofs are generally omitted or treated as optional
material. Throughout the text, emphasis is placed on physical and
geometric intuition, illustrative examples, and practical implementation
considerations. Examples are drawn primarily from robotics, including
robot motion and control, contact dynamics, trajectory optimization,
and learning-based control, to illustrate how the theoretical concepts
arise in real robotic systems.

These notes are developed at the \textbf{Robotics and Physical AI
Laboratory (RoPhi Lab)} at Yonsei University. Further information
about the laboratory and its research can be found at
\url{https://rophi-lab.github.io}.

\textbf{Acknowledgment of AI Assistance.} This manuscript was developed through a human--AI collaborative writing
process. Yonghyeon Lee established the intellectual direction, selected
and organized the topics, guided the mathematical and pedagogical
development, and reviewed and revised the manuscript. Under his guidance,
GPT contributed substantially to the drafting, rewriting, and stylistic
refinement of the text. All final decisions regarding the content,
mathematical interpretation, and presentation, as well as responsibility
for the resulting manuscript, remain with Yonghyeon Lee.

\vfill
\begin{flushright}
Yonghyeon Lee\\
Department of Artificial Intelligence\\
Yonsei University\\
September 2026
\end{flushright}
\endgroup

\begingroup
\hypersetup{linkcolor=black}
\tableofcontents
\endgroup




\mainmatter

\part{Optimization Foundations}
\chapter{Optimization in Vector Spaces}
\chapteroverview{The geometry, differential calculus, and optimality conditions underlying optimization with finitely many variables.}

\section{Vector Spaces and Geometry}

Optimization begins with a choice of variables and a description of the space in
which those variables live. An optimization variable may be a vector in $\R^m$,
or a sequence of such vectors stacked into one longer vector. Although these
objects may have different interpretations, the same geometric
ideas---linearity, length, angle, and projection---organize all of them.

Throughout this chapter, vector spaces are defined over the real numbers unless
stated otherwise. Vectors are written as column vectors. In particular, for
$x\in\R^m$,
\begin{equation}
  x=
  \begin{bmatrix}
    x_1 \\
    x_2 \\
    \vdots \\
    x_m
  \end{bmatrix}.
\end{equation}
The vectors $e_1,\ldots,e_m$, where $e_i$ has a $1$ in position $i$ and zeros
elsewhere, form the \emph{standard basis} of $\R^m$. Every vector can be written
as
\begin{equation}
  x=x_1e_1+\cdots+x_me_m.
\end{equation}
The coefficients $x_1,\ldots,x_m$ are called the \emph{standard coordinates}
of $x$. They are the entries of the displayed column vector.

For $x,y\in\R^m$, their dot product is
\begin{equation}
  x\trans y=\sum_{i=1}^m x_i y_i.
\end{equation}
Its geometric interpretation as an inner product is developed below.

\subsection{Vectors, Matrices, and Linear Maps}

The word \emph{vector} does not necessarily mean a list of numbers. It means an
element of a set on which addition and scalar multiplication behave linearly.

\begin{definition}[Vector space]
A real vector space $\mathcal V$ is a set equipped with vector addition and
scalar multiplication. It is closed under linear combinations, and for every
$x,y,z\in\mathcal V$ and $a,b\in\R$,
\begin{align}
  x+y&=y+x, & (x+y)+z&=x+(y+z),\\
  x+0&=x,   & x+(-x)&=0,\\
  a(x+y)&=ax+ay, & (a+b)x&=ax+bx,\\
  (ab)x&=a(bx), & 1x&=x.
\end{align}
\end{definition}

The most familiar example is $\R^n$. Other examples include matrices
$\R^{m\times n}$ and polynomials of bounded degree. For now, we focus on these
spaces with finitely many scalar coordinates.

Given vectors $v_1,\ldots,v_k\in\mathcal V$, a vector of the form
\begin{equation}
  a_1v_1+\cdots+a_kv_k
\end{equation}
is a \emph{linear combination}. The set of all such combinations is their span,
\begin{equation}
  \operatorname{span}\{v_1,\ldots,v_k\}
  =\left\{\sum_{i=1}^k a_i v_i : a_i\in\R\right\}.
\end{equation}
The vectors are linearly independent if
$\sum_{i=1}^k a_i v_i=0$ implies $a_i=0$ for every $i$. Thus, independence
means that none of the vectors can be reconstructed from the others.

A vector space is finite-dimensional if it is spanned by a finite collection
of vectors. All vector spaces in Part~I are finite-dimensional.

\begin{example}[A stacked sequence]
Suppose an optimization problem chooses vectors $u_t\in\R^m$ indexed by
$t=0,\ldots,T-1$. The entire sequence can be stacked into one vector,
\begin{equation}
  \mathbf u=\begin{bmatrix}
    u_0\trans & u_1\trans & \cdots & u_{T-1}\trans
  \end{bmatrix}\trans
  \in\R^{mT}.
\end{equation}
The problem can therefore treat a sequence as a single vector in $\R^{mT}$.
The interpretation of the blocks does not change the vector-space algebra.
\end{example}

A map between vector spaces is linear when it preserves linear combinations.

\begin{definition}[Linear map]
A map $L:\mathcal V\to\mathcal W$ is linear if
\begin{equation}
  L(ax+by)=aL(x)+bL(y)
\end{equation}
for all $x,y\in\mathcal V$ and $a,b\in\R$.
\end{definition}

Using the standard coordinates of $\R^n$ and $\R^m$, every linear map
$L:\R^n\to\R^m$ has a matrix representation: there is a matrix
$A\in\R^{m\times n}$ such that $L(x)=Ax$. Matrix multiplication then represents
composition: if $M(y)=By$, then $(M\circ L)(x)=BAx$.

Two sets associated with a linear map describe what it removes and what it can
produce:
\begin{align}
  \ker(A) &= \{x\in\R^n:Ax=0\}, \\
  \operatorname{range}(A) &= \{Ax:x\in\R^n\}.
\end{align}
The kernel is also called the null space, and the range of a matrix is its
column space. The kernel contains directions removed by the map, while the
range contains every output the map can produce.

An \emph{affine} map has the form $x\mapsto Ax+b$. It need not be linear because
it may not map the origin to the origin. Affine maps are nevertheless central to
optimization because many constraints and approximations have this form.

For $A\in\R^{m\times n}$, the transpose $A\trans\in\R^{n\times m}$ is defined
by $(A\trans)_{ij}=A_{ji}$: it exchanges rows and columns. Its geometric role is
explained after inner products are defined.

\subsection{Norms, Inner Products, and Distances}

A vector space describes which algebraic operations are possible, but it does
not by itself specify the size of a vector or whether two vectors are close.
These geometric notions are introduced by norms and inner products.

\begin{definition}[Norm]
A norm on $\mathcal V$ is a function $\norm{\cdot}:\mathcal V\to\R$ satisfying,
for all $x,y\in\mathcal V$ and $a\in\R$,
\begin{align}
  \norm{x} &\geq 0, \qquad \norm{x}=0 \iff x=0, \\
  \norm{ax} &= |a|\norm{x}, \\
  \norm{x+y} &\leq \norm{x}+\norm{y}.
\end{align}
The last property is the triangle inequality.
\end{definition}

The most common norms on $\R^n$ are
\begin{align}
  \norm{x}_1 &= \sum_{i=1}^n |x_i|, \\
  \norm{x}_2 &= \left(\sum_{i=1}^n x_i^2\right)^{1/2}, \\
  \norm{x}_\infty &= \max_{1\leq i\leq n}|x_i|.
\end{align}
More generally, $\norm{x}_p=(\sum_i |x_i|^p)^{1/p}$ for $p\geq1$. Different
norms encode different notions of size. Their distinct geometries will later
lead to different optimization models and solutions.

A norm induces a distance
\begin{equation}
  d(x,y)=\norm{x-y}
\end{equation}
and therefore defines neighborhoods and convergence: $x_k\to x$ means
$\norm{x_k-x}\to0$. The open ball
of radius $r$ centered at $x$ is
\begin{equation}
  B_r(x)=\{y\in\mathcal V:\norm{y-x}<r\}.
\end{equation}
A set is open if every point in it is the center of some open ball contained in
the set. A function $f$ is continuous at $x$ if $f(x_k)\to f(x)$ whenever
$x_k\to x$. These definitions make the words ``nearby'' and ``small change''
precise. On $\R^n$, all norms define the same convergent sequences and open
sets, although the shapes of their balls differ.

\begin{definition}[Inner product]
An inner product on a real vector space $\mathcal V$ is a function
$\langle\cdot,\cdot\rangle:\mathcal V\times\mathcal V\to\R$ that is symmetric,
linear in each argument, and positive definite:
\begin{equation}
  \langle x,y\rangle=\langle y,x\rangle,
  \qquad
  \langle x,x\rangle>0 \quad\text{for }x\neq0.
\end{equation}
\end{definition}

Every inner product induces a norm,
\begin{equation}
  \norm{x}=\sqrt{\langle x,x\rangle}.
\end{equation}
The \emph{standard inner product} on $\R^n$ is the dot product
\begin{equation}
  \langle x,y\rangle=x\trans y=\sum_{i=1}^n x_i y_i.
\end{equation}
It induces the Euclidean norm
$\norm{x}_2=\sqrt{x\trans x}$. With the standard inner products on
$\R^n$ and $\R^m$, the transpose satisfies
\begin{equation}
  \langle Ax,y\rangle=\langle x,A\trans y\rangle,
  \qquad x\in\R^n,\ y\in\R^m.
  \label{eq:transpose-adjoint}
\end{equation}
The linear map that moves an operation between the two arguments in this way is
called its \emph{adjoint}. Thus $A\trans$ is the adjoint of $x\mapsto Ax$ under
the standard Euclidean inner products.

An \emph{eigenvector} of a square matrix $W$ is a nonzero vector $v$ satisfying
\begin{equation}
  Wv=\lambda v,
\end{equation}
where the scalar $\lambda$ is the corresponding \emph{eigenvalue}. Multiplying
an eigenvector by $W$ therefore changes only its magnitude, not its direction.
A real symmetric matrix has mutually orthogonal unit eigenvectors and can be
written
\begin{equation}
  W=U\Lambda U\trans,
\end{equation}
where the columns of $U$ are eigenvectors and the diagonal entries of
$\Lambda$ are their eigenvalues.

A symmetric matrix $W$ is called \emph{positive definite}, written $W\succ0$,
if $x\trans Wx>0$ for every nonzero $x$. For a symmetric matrix, this is
equivalent to requiring every eigenvalue to be positive. Indeed, with
$z=U\trans x$,
\begin{equation}
  x\trans Wx=z\trans\Lambda z=\sum_{i=1}^n\lambda_i z_i^2.
\end{equation}
Thus the quadratic form is positive in every nonzero direction precisely when
each $\lambda_i>0$. For such a matrix,
\begin{equation}
  \langle x,y\rangle_W=x\trans Wy,
  \qquad
  \norm{x}_W=\sqrt{x\trans Wx}
\end{equation}
define a weighted inner product and norm. Along a unit eigenvector $v_i$,
$\norm{v_i}_W=\sqrt{\lambda_i}$. Large eigenvalues therefore make their
directions more expensive and make the weighted unit ball narrower along those
directions.

\begin{proposition}[Cauchy--Schwarz inequality]
For any inner-product space and any $x,y\in\mathcal V$,
\begin{equation}
  |\langle x,y\rangle|\leq \norm{x}\norm{y}.
\end{equation}
Equality holds if and only if $x$ and $y$ are linearly dependent.
\end{proposition}

For nonzero vectors, the angle $\theta\in[0,\pi]$ between them is consequently
well defined by
\begin{equation}
  \cos\theta=\frac{\langle x,y\rangle}{\norm{x}\norm{y}}.
\end{equation}
Vectors are orthogonal when $\langle x,y\rangle=0$. Orthogonality is more than a
visual notion: it identifies directions whose effects decouple under the chosen
inner product.

\begin{example}[Geometry of a quadratic function]
Let $Q\succ0$ and consider $J(x)=x\trans Qx$. For $c>0$, the set
$\{x:J(x)=c\}$, called a level set, is an ellipsoid. In the decomposition
$Q=U\Lambda U\trans$, the eigenvectors give the principal axes and the
eigenvalues determine the scaling along those axes. A large eigenvalue produces
a narrow direction in the level set and a rapidly changing direction in the
function.
\end{example}

\subsection{Subspaces, Bases, and Orthogonal Projections}

A subset of a vector space is a subspace when it retains the vector-space
structure.

\begin{definition}[Subspace]
A nonempty subset $\mathcal S\subseteq\mathcal V$ is a linear subspace if
\begin{equation}
  ax+by\in\mathcal S
\end{equation}
for every $x,y\in\mathcal S$ and $a,b\in\R$.
\end{definition}

The span of any collection of vectors is a subspace. Kernels and ranges of
linear maps are also subspaces. By contrast, an affine set
\begin{equation}
  x_0+\mathcal S=\{x_0+s:s\in\mathcal S\}
\end{equation}
is a translated subspace and need not contain the origin. For example, the
solution set of $Ax=b$, when nonempty, is affine:
\begin{equation}
  \{x:Ax=b\}=x_p+\ker(A),
\end{equation}
where $x_p$ is any particular solution. The null space therefore describes all
feasible directions along which an equality-constrained variable can move.

\begin{definition}[Basis and dimension]
A basis of a vector space $\mathcal V$ is a linearly independent collection
$\mathcal B=\{v_1,\ldots,v_k\}$ that spans $\mathcal V$. Given this basis,
every $x\in\mathcal V$ has a unique representation
\begin{equation}
  x=\sum_{i=1}^k x_i v_i.
\end{equation}
The scalars $x_1,\ldots,x_k$ are the coordinates of $x$ with respect to
$\mathcal B$. The number $k$ of basis vectors is the dimension of
$\mathcal V$.
\end{definition}

\paragraph{Change of basis.}
Coordinates describe a vector relative to a chosen basis, so the coordinate
column changes when the basis changes even though the underlying vector does
not. In $\R^n$, let
\begin{equation}
  B=\begin{bmatrix}v_1&\cdots&v_n\end{bmatrix},
  \qquad
  C=\begin{bmatrix}w_1&\cdots&w_n\end{bmatrix}
\end{equation}
contain two bases as columns. Write $x_B$ and $x_C$ for the coordinate columns
of the same vector $x$ in these bases. Then
\begin{equation}
  x=Bx_B=Cx_C.
\end{equation}
Define the change matrix
\begin{equation}
  S=C^{-1}B.
\end{equation}
It converts coordinates in basis $B$ to coordinates in basis $C$:
\begin{equation}
  x_C=Sx_B,
  \qquad
  x_B=S^{-1}x_C.
  \label{eq:change-of-basis-vector}
\end{equation}

\paragraph{Linear maps under a change of basis.}
Let $L:\R^n\to\R^n$ be a linear operator. Denote its matrix in basis $B$ by
$A_B$ and its matrix in basis $C$ by $A_C$, so that
\begin{equation}
  (Lx)_B=A_Bx_B,
  \qquad
  (Lx)_C=A_Cx_C.
\end{equation}
Using $x_C=Sx_B$ gives
\begin{equation}
  A_C=SA_BS^{-1}.
  \label{eq:change-of-basis-linear-map}
\end{equation}
This is a similarity transformation. The entries of the matrix change, but the
linear operator and its eigenvalues do not. More generally, if input
coordinates change by $x_{\mathrm{new}}=Sx_{\mathrm{old}}$ and output
coordinates by $y_{\mathrm{new}}=Ty_{\mathrm{old}}$, then
\begin{equation}
  A_{\mathrm{new}}=TA_{\mathrm{old}}S^{-1}.
\end{equation}

\paragraph{Inner products under a change of basis.}
In basis $B$, an inner product is represented by a symmetric positive-definite
Gram matrix $G_B$:
\begin{equation}
  \langle x,y\rangle
  =x_B\trans G_B y_B.
\end{equation}
Because $x_C=Sx_B$, the matrix in basis $C$ is
\begin{equation}
  G_C=(S^{-1})\trans G_BS^{-1}.
  \label{eq:change-of-basis-inner-product}
\end{equation}
The coordinates and the Gram matrix change together, so lengths and angles do
not change. In an orthonormal basis, $G=I$. For the standard Euclidean inner
product in a general basis matrix $C$, the Gram matrix is $G_C=C\trans C$.

The rank of a matrix is the dimension of its range. For
$A\in\R^{m\times n}$, the rank--nullity relation always states that
\begin{equation}
  \dim\ker(A)+\dim\operatorname{range}(A)=n.
\end{equation}
The right-hand side is $n$ because the domain of $A$ is $\R^n$. Since
$\operatorname{rank}(A)\leq\min\{m,n\}$, if $m<n$ then
\begin{equation}
  \dim\ker(A)=n-\operatorname{rank}(A)\geq n-m>0.
\end{equation}
Thus a map from a higher-dimensional input space to a lower-dimensional output
space must remove some nonzero input directions. If $A$ has full row rank
$m$, its null space has dimension exactly $n-m$; if its rank is smaller, the
null space is even larger.

A basis $\{v_1,\ldots,v_n\}$ is \emph{orthonormal} if
\begin{equation}
  \langle v_i,v_j\rangle=
  \begin{cases}
    1, & i=j,\\
    0, & i\neq j.
  \end{cases}
\end{equation}
If $x=\sum_{i=1}^n x_i v_i$, taking the inner product with $v_j$ gives
\begin{equation}
  \langle v_j,x\rangle
  =\sum_{i=1}^n x_i\langle v_j,v_i\rangle
  =x_j.
\end{equation}
Thus each coordinate is obtained by a projection:
$x_j=\langle v_j,x\rangle$. For
$x=\sum_i x_i v_i$ and $y=\sum_i y_i v_i$, orthonormality also gives
\begin{equation}
  \langle x,y\rangle=\sum_{i=1}^n x_i y_i,
  \qquad
  \norm{x}^2=\sum_{i=1}^n x_i^2.
  \label{eq:orthonormal-coordinate-identities}
\end{equation}
Hence coordinates in an orthonormal basis preserve inner products, lengths,
and angles.

In $\R^n$ with the standard inner product, let
$V=[v_1\ \cdots\ v_n]$. Orthonormality is equivalent to
\begin{equation}
  V\trans V=VV\trans=I,
  \qquad
  V^{-1}=V\trans.
\end{equation}
The coordinate conversion is therefore $x_V=V\trans x$, and
$\norm{x_V}_2=\norm{x}_2$. Its $2$-norm condition number is $1$, so changing to
an orthonormal basis does not amplify vector lengths or small numerical errors.

For a subspace $\mathcal S$, its orthogonal complement is
\begin{equation}
  \mathcal S^\perp
  =\{z\in\mathcal V:\langle z,s\rangle=0\text{ for all }s\in\mathcal S\}.
\end{equation}
In finite dimensions, every vector can be decomposed uniquely into a component
in $\mathcal S$ and a component in $\mathcal S^\perp$.

\begin{theorem}[Orthogonal projection]
Let $\mathcal S$ be a linear subspace of a finite-dimensional inner-product
space $\mathcal V$. For every $x\in\mathcal V$, there is a unique vector
$s^\star\in\mathcal S$ satisfying
\begin{equation}
  s^\star=\argmin_{s\in\mathcal S}\norm{x-s}.
\end{equation}
It is characterized by the orthogonality condition
\begin{equation}
  x-s^\star\in\mathcal S^\perp.
\end{equation}
The vector $s^\star$ is the orthogonal projection of $x$ onto $\mathcal S$.
\end{theorem}

Suppose the columns of $B\in\R^{n\times k}$ form a basis for $\mathcal S$.
Every $s\in\mathcal S$ can be written as $s=Bz$. Projecting $x$ therefore means
solving
\begin{equation}
  z^\star=\argmin_z\norm{x-Bz}_2^2.
\end{equation}
The residual must be orthogonal to every column of $B$, giving the normal
equations
\begin{equation}
  B\trans(Bz^\star-x)=0.
\end{equation}
If the columns of $B$ are linearly independent, then
\begin{equation}
  z^\star=(B\trans B)^{-1}B\trans x,
  \qquad
  s^\star=\underbrace{B(B\trans B)^{-1}B\trans}_{P_{\mathcal S}}x.
\end{equation}
The matrix $P_{\mathcal S}$ is symmetric and idempotent:
$P_{\mathcal S}\trans=P_{\mathcal S}$ and $P_{\mathcal S}^2=P_{\mathcal S}$.
If $B$ has orthonormal columns, the formula simplifies to
$P_{\mathcal S}=BB\trans$.

\begin{example}[Projection as least squares]
If the equation $Bz=x$ has no exact solution, $x$ does not belong to the column
space of $B$. The least-squares solution chooses $Bz^\star$ as the point in that
column space nearest to $x$. Its residual $x-Bz^\star$ is orthogonal to the
column space. Thus least squares is fundamentally a geometric projection
problem, not merely an algebraic recipe.
\end{example}

\paragraph{Section summary.}
Vector spaces provide the algebra of optimization variables; norms determine
size and proximity; inner products add angles and orthogonality; and projections
find the closest point in a subspace. The remaining sections build differential
calculus and optimality conditions on top of this geometry.

\begin{exercise}
Show that
\begin{equation}
  \mathcal S=\{x\in\R^3:x_1+x_2+x_3=0\}
\end{equation}
is a subspace. Find a basis and its dimension.
\end{exercise}
\begin{proof}[Short solution]
The set is the kernel of the linear map $x\mapsto x_1+x_2+x_3$, so it is a
subspace. Every element has the form
\begin{equation}
  (a,b,-a-b)=a(1,0,-1)+b(0,1,-1).
\end{equation}
The two displayed vectors are independent, so they form a basis and
$\dim\mathcal S=2$.
\end{proof}

\begin{exercise}
Let $v_1=(1,1)\trans$ and $v_2=(1,-1)\trans$. Find the coordinates of
$x=(3,1)\trans$ in the basis $\{v_1,v_2\}$.
\end{exercise}
\begin{proof}[Short solution]
Solving $x=a v_1+b v_2$ gives $a+b=3$ and $a-b=1$. Hence $a=2$, $b=1$, and
the coordinate column is $(2,1)\trans$.
\end{proof}

\begin{exercise}
Let $B=I$ be the standard basis matrix and let
\begin{equation}
  C=\begin{bmatrix}1&1\\1&-1\end{bmatrix}.
\end{equation}
Find the change matrix $S=C^{-1}B$ and use it to convert the standard
coordinates $x_B=(3,1)\trans$ to $C$-coordinates.
\end{exercise}
\begin{proof}[Short solution]
Since
\begin{equation}
  S=C^{-1}=\frac12\begin{bmatrix}1&1\\1&-1\end{bmatrix},
\end{equation}
we obtain $x_C=Sx_B=(2,1)\trans$, agreeing with the previous exercise.
\end{proof}

\begin{exercise}
In standard coordinates, let a linear operator have matrix
$A_B=\operatorname{diag}(2,1)$. Using the basis matrix $C$ from the previous
exercise, compute $A_C=SA_BS^{-1}$ and compare its eigenvalues with those of
$A_B$.
\end{exercise}
\begin{proof}[Short solution]
Direct multiplication gives
\begin{equation}
  A_C=\frac12\begin{bmatrix}3&1\\1&3\end{bmatrix}.
\end{equation}
Its eigenvalues are $2$ and $1$, the same as those of $A_B$. A change of basis
changes the matrix entries, not the underlying operator or its eigenvalues.
\end{proof}

\begin{exercise}
For the standard Euclidean inner product and the basis matrix
$C=\begin{bmatrix}1&1\\1&-1\end{bmatrix}$, find the Gram matrix $G_C$. Verify
the length of $x=(3,1)\trans$ using its coordinate column $x_C=(2,1)\trans$.
\end{exercise}
\begin{proof}[Short solution]
The Gram matrix is
\begin{equation}
  G_C=C\trans C=2I.
\end{equation}
Therefore $x_C\trans G_Cx_C=2(2^2+1^2)=10$, which equals
$x\trans x=3^2+1^2=10$.
\end{proof}

\begin{exercise}
Let
\begin{equation}
  A=\begin{bmatrix}1&1&0\\0&1&1\end{bmatrix}.
\end{equation}
Find bases for $\ker(A)$ and $\operatorname{range}(A)$, and verify the
rank--nullity relation.
\end{exercise}
\begin{proof}[Short solution]
The equations $Ax=0$ give $x=(-t,t,-t)$, so
$\ker(A)=\operatorname{span}\{(-1,1,-1)\trans\}$. The first two columns are
independent and span $\R^2$, so they form a basis for the range. Thus the
nullity is $1$, the rank is $2$, and $1+2=3$, the dimension of the domain.
\end{proof}

\begin{exercise}
Prove that every matrix $A\in\R^{m\times n}$ with $m<n$ has a nonzero vector
in its kernel.
\end{exercise}
\begin{proof}[Short solution]
Since $\operatorname{rank}(A)\leq m$, rank--nullity gives
\begin{equation}
  \dim\ker(A)=n-\operatorname{rank}(A)\geq n-m>0.
\end{equation}
Hence the kernel contains a nonzero vector, and $A$ cannot be one-to-one.
\end{proof}

\begin{exercise}
Let $W\succ0$. Show that $\langle x,y\rangle_W=x\trans Wy$ is an inner product.
Describe the unit ball $\{x:x\trans Wx\leq1\}$ when
$W=\operatorname{diag}(1,4)$.
\end{exercise}
\begin{proof}[Short solution]
Bilinearity follows from matrix multiplication, symmetry from $W=W\trans$,
and positive definiteness from $x\trans Wx>0$ for $x\neq0$. For the stated
matrix, the unit ball is
\begin{equation}
  x_1^2+4x_2^2\leq1,
\end{equation}
an ellipse with semiaxes $1$ and $1/2$.
\end{proof}

\begin{exercise}
Let
\begin{equation}
  q_1=\frac{1}{\sqrt2}(1,1)\trans,
  \qquad
  q_2=\frac{1}{\sqrt2}(1,-1)\trans.
\end{equation}
Verify that $\{q_1,q_2\}$ is orthonormal, find the coordinates of
$x=(3,1)\trans$, and verify that the coordinates preserve its norm.
\end{exercise}
\begin{proof}[Short solution]
The vectors have unit norm and $q_1\trans q_2=0$. Their coordinates are
\begin{equation}
  q_1\trans x=2\sqrt2,
  \qquad
  q_2\trans x=\sqrt2.
\end{equation}
Their squared sum is $8+2=10=\norm{x}_2^2$.
\end{proof}

\begin{exercise}
Let $q\in\R^n$ satisfy $\norm{q}_2=1$. Show that $P=qq\trans$ is the
orthogonal projection onto $\operatorname{span}\{q\}$. Then compute the
projection of $x=(3,1)\trans$ when $q=(1,2)\trans/\sqrt5$.
\end{exercise}
\begin{proof}[Short solution]
We have $P\trans=P$ and
$P^2=q(q\trans q)q\trans=qq\trans=P$. Moreover,
$Px=q(q\trans x)$ lies in $\operatorname{span}\{q\}$ and
$q\trans(x-Px)=0$. For the given vectors,
\begin{equation}
  P=\frac15\begin{bmatrix}1&2\\2&4\end{bmatrix},
  \qquad
  Px=(1,2)\trans.
\end{equation}
The residual $(2,-1)\trans$ is orthogonal to $(1,2)\trans$.
\end{proof}

\section{Differential Calculus in Vector Spaces}

Optimization asks how a function changes when its argument changes. Derivatives
provide a local linear description of that change. We begin with scalar-valued
functions and then extend the definitions to vector-valued maps.

\subsection{Directional and Total Derivatives}

Let $f:\R^n\to\R$ and fix a point $x\in\R^n$. A direction
$d\in\R^n$ defines the one-dimensional curve $t\mapsto x+td$. The derivative
of $f$ along this curve measures the instantaneous change of $f$ in direction
$d$.

\begin{definition}[Directional derivative]
The directional derivative of $f$ at $x$ in direction $d$ is
\begin{equation}
  Df(x)[d]
  =\lim_{t\to0}\frac{f(x+td)-f(x)}{t},
  \label{eq:directional-derivative}
\end{equation}
provided the limit exists.
\end{definition}

For a coordinate vector $e_i$, the directional derivative $Df(x)[e_i]$ is the
partial derivative $\partial f(x)/\partial x_i$. Directional derivatives are
therefore a coordinate-free extension of partial derivatives. Their existence
in every direction, however, does not by itself guarantee that one linear map
describes all small perturbations consistently.

The reason is that a directional derivative examines one fixed line at a time.
A genuine derivative must give one linear approximation that remains accurate
for perturbations approaching the point from \emph{any} direction. In
particular, if $f$ is differentiable, the map
$d\mapsto Df(x)[d]$ must be linear in $d$.

For example, define $f:\R^2\to\R$ by
\begin{equation}
  f(x_1,x_2)=
  \begin{cases}
    \displaystyle\frac{x_1^3}{x_1^2+x_2^2},
      &(x_1,x_2)\neq(0,0),\\[1ex]
    0, &(x_1,x_2)=(0,0).
  \end{cases}
\end{equation}
For a nonzero direction $d=(d_1,d_2)\trans$,
\begin{equation}
  Df(0)[d]
  =\lim_{t\to0}\frac{f(td)-f(0)}{t}
  =\frac{d_1^3}{d_1^2+d_2^2}.
\end{equation}
Thus every directional derivative exists. Nevertheless,
\begin{equation}
  Df(0)[e_1]=1,
  \qquad
  Df(0)[e_2]=0,
  \qquad
  Df(0)[e_1+e_2]=\frac12.
\end{equation}
The last value is not the sum of the first two, so the directional derivatives
do not form a linear map. Consequently, $f$ is not differentiable at the
origin, even though it is continuous there and has a derivative along every
line.

\begin{definition}[Total derivative]
A function $f:\R^n\to\R^m$ is differentiable at $x$ if there exists a linear
map $L:\R^n\to\R^m$ such that
\begin{equation}
  f(x+h)=f(x)+Lh+r(h),
  \qquad
  \lim_{\norm{h}_2\to0}\frac{\norm{r(h)}_2}{\norm{h}_2}=0.
  \label{eq:total-derivative}
\end{equation}
The unique map $L$ is the total derivative of $f$ at $x$, denoted $Df(x)$.
The remainder condition is also written $r(h)=o(\norm{h}_2)$.
\end{definition}

Equation~\eqref{eq:total-derivative} says that the error of the linear
approximation becomes negligible compared with the size of the perturbation.
Differentiability implies continuity. It also implies that every directional
derivative exists and satisfies
\begin{equation}
  Df(x)[d]=Df(x)d.
\end{equation}
The expression on the left emphasizes a directional rate of change; the one on
the right emphasizes application of a linear map.

\begin{example}[A linear and an affine map]
For $f(x)=Ax$, the derivative is $Df(x)=A$ at every $x$, and the remainder is
zero. The same derivative is obtained for the affine map $f(x)=Ax+b$ because
the constant vector $b$ disappears under subtraction.
\end{example}

Differentiable maps compose according to the chain rule.

\begin{theorem}[Chain rule]
Let $f:\R^n\to\R^m$ be differentiable at $x$, and let
$g:\R^m\to\R^p$ be differentiable at $f(x)$. Then $g\circ f$ is
differentiable at $x$ and
\begin{equation}
  D(g\circ f)(x)=Dg(f(x))Df(x).
  \label{eq:chain-rule-linear-map}
\end{equation}
\end{theorem}

The order in \eqref{eq:chain-rule-linear-map} follows the order of function
composition: a perturbation first passes through $Df(x)$ and then through
$Dg(f(x))$.

\subsection{Gradients, Jacobians, and Hessians}

In Euclidean coordinates, derivatives are represented by familiar vectors and
matrices.

\begin{definition}[Gradient]
For a differentiable scalar function $f:\R^n\to\R$, the gradient
$\nabla f(x)\in\R^n$ is the unique vector satisfying
\begin{equation}
  Df(x)[d]=\langle\nabla f(x),d\rangle
  =\nabla f(x)\trans d
  \qquad\text{for every }d\in\R^n.
  \label{eq:gradient-definition}
\end{equation}
In standard coordinates,
\begin{equation}
  \nabla f(x)=
  \begin{bmatrix}
    \frac{\partial f}{\partial x_1}(x) & \cdots &
    \frac{\partial f}{\partial x_n}(x)
  \end{bmatrix}\trans.
\end{equation}
\end{definition}

The gradient points in the direction of greatest instantaneous increase under
the Euclidean norm. Indeed, Cauchy--Schwarz gives
\begin{equation}
  Df(x)[d]\leq\norm{\nabla f(x)}_2\norm{d}_2,
\end{equation}
with equality among unit directions when
$d=\nabla f(x)/\norm{\nabla f(x)}_2$. Consequently,
$-\nabla f(x)$ is the direction of steepest instantaneous decrease.

\begin{definition}[Jacobian]
For $f:\R^n\to\R^m$ with component functions
$f_1,\ldots,f_m$, the Jacobian is the matrix
\begin{equation}
  J_f(x)=
  \begin{bmatrix}
    \frac{\partial f_1}{\partial x_1} & \cdots &
    \frac{\partial f_1}{\partial x_n}\\
    \vdots & \ddots & \vdots\\
    \frac{\partial f_m}{\partial x_1} & \cdots &
    \frac{\partial f_m}{\partial x_n}
  \end{bmatrix}.
\end{equation}
It is the matrix representation of $Df(x)$, so
$Df(x)[d]=J_f(x)d$.
\end{definition}

With this convention, the chain rule becomes
\begin{equation}
  J_{g\circ f}(x)=J_g(f(x))J_f(x).
\end{equation}
If $g:\R^m\to\R$ is scalar-valued, then
\begin{equation}
  \nabla(g\circ f)(x)=J_f(x)\trans\nabla g(f(x)).
  \label{eq:gradient-chain-rule}
\end{equation}
The transpose in \eqref{eq:gradient-chain-rule} follows directly from the
adjoint identity introduced in \cref{eq:transpose-adjoint}.

\begin{definition}[Hessian]
If $f:\R^n\to\R$ has differentiable first partial derivatives, its Hessian is
the matrix
\begin{equation}
  \nabla^2 f(x)=
  \begin{bmatrix}
    \frac{\partial^2 f}{\partial x_1^2} & \cdots &
    \frac{\partial^2 f}{\partial x_1\partial x_n}\\
    \vdots & \ddots & \vdots\\
    \frac{\partial^2 f}{\partial x_n\partial x_1} & \cdots &
    \frac{\partial^2 f}{\partial x_n^2}
  \end{bmatrix}.
\end{equation}
\end{definition}

If the second partial derivatives are continuous near $x$, mixed partials are
equal and the Hessian is symmetric. The quadratic form
$d\trans\nabla^2 f(x)d$ measures curvature along direction $d$.

Several derivative identities will be used repeatedly. For
$a\in\R^n$, $\beta\in\R$, $A\in\R^{m\times n}$, and $b\in\R^m$,
\begin{align}
  \nabla(a\trans x+\beta) &= a, \\
  \nabla\left(\tfrac12\norm{Ax-b}_2^2\right)
    &=A\trans(Ax-b), \\
  \nabla(x\trans Qx) &=(Q+Q\trans)x.
  \label{eq:basic-gradient-identities}
\end{align}
If $Q=Q\trans$, the last identity reduces to $2Qx$ and its Hessian is $2Q$.

\begin{example}[Derivative of a composition]
Let $f(x)=\tfrac12\norm{Ax-b}_2^2$. Define $r(x)=Ax-b$ and
$g(r)=\tfrac12 r\trans r$. Then $J_r(x)=A$ and $\nabla g(r)=r$. The gradient
chain rule gives
\begin{equation}
  \nabla f(x)=A\trans r(x)=A\trans(Ax-b).
\end{equation}
This derivation separates the nonlinear scalar objective from the linear map
that produces its residual.
\end{example}

\subsection{Taylor Expansions and Local Models}

Taylor expansions use derivative information to approximate $f(x+h)$ when the
displacement $h$ is small. For a differentiable function, the first-order model
built at $x$ is
\begin{equation}
  m_1(h;x)=f(x)+\nabla f(x)\trans h.
\end{equation}
Here $h$ is the model variable, while the argument after the semicolon, $x$,
identifies the fixed point at which the model is built.

\begin{theorem}[First-order Taylor expansion]
If $f:\R^n\to\R$ is differentiable at $x$, then
\begin{equation}
  f(x+h)=f(x)+\nabla f(x)\trans h+o(\norm{h}_2).
\end{equation}
\end{theorem}

If second derivatives exist and are continuous, curvature can also be included.

\begin{theorem}[Second-order Taylor expansion]
Suppose $f$ is twice continuously differentiable on an open set containing the
line segment from $x$ to $x+h$. Then there exists $\theta\in(0,1)$ such that
\begin{equation}
  f(x+h)=f(x)+\nabla f(x)\trans h
  +\frac12 h\trans\nabla^2 f(x+\theta h)h.
  \label{eq:taylor-mean-value}
\end{equation}
As $h\to0$,
\begin{equation}
  f(x+h)=f(x)+\nabla f(x)\trans h
  +\frac12 h\trans\nabla^2 f(x)h+o(\norm{h}_2^2).
  \label{eq:taylor-second-order}
\end{equation}
\end{theorem}

The corresponding second-order local model is
\begin{equation}
  m_2(h;x)=f(x)+\nabla f(x)\trans h
  +\frac12 h\trans\nabla^2 f(x)h.
  \label{eq:quadratic-local-model}
\end{equation}
Many numerical algorithms choose $h$ by approximately minimizing either the
linear model $m_1$ or the quadratic model $m_2$ while controlling how large the
step may be.

\begin{example}[A one-dimensional local model]
For $f(x)=e^x$, the expansion around $x=0$ is
\begin{equation}
  e^h=1+h+\frac12h^2+o(h^2).
\end{equation}
Near zero, the linear model $1+h$ gives the slope, while the quadratic term
$h^2/2$ captures the upward curvature.
\end{example}

\begin{exercise}
Let $f(x_1,x_2)=x_1^2+3x_1x_2$. Compute the directional derivative at
$x=(1,-1)\trans$ in the direction $d=(2,1)\trans$.
\end{exercise}
\begin{proof}[Short solution]
The gradient is
$\nabla f(x)=(2x_1+3x_2,,3x_1)\trans$, so
\begin{equation}
  Df(1,-1)[d]=(-1,3)\begin{bmatrix}2\\1\end{bmatrix}=1.
\end{equation}
\end{proof}

\begin{exercise}
For the function
\begin{equation}
  f(x_1,x_2)=
  \begin{cases}
    \displaystyle\frac{x_1^3}{x_1^2+x_2^2}, &(x_1,x_2)\neq(0,0),\\[1ex]
    0, &(x_1,x_2)=(0,0),
  \end{cases}
\end{equation}
compute $Df(0)[d]$ and show that it is not linear in $d$.
\end{exercise}
\begin{proof}[Short solution]
For $d\neq0$,
\begin{equation}
  Df(0)[d]=\frac{d_1^3}{d_1^2+d_2^2}.
\end{equation}
Since $Df(0)[e_1]=1$, $Df(0)[e_2]=0$, but
$Df(0)[e_1+e_2]=1/2$, this map is not additive and hence not linear.
\end{proof}

\begin{exercise}
Let $F(x)=Ax+b$, where $A\in\R^{m\times n}$ and $b\in\R^m$. Verify directly
from the definition of the total derivative that $DF(x)=A$.
\end{exercise}
\begin{proof}[Short solution]
For every perturbation $h$,
\begin{equation}
  F(x+h)-F(x)=A(x+h)+b-(Ax+b)=Ah.
\end{equation}
Thus the linear approximation is exact and the remainder is zero.
\end{proof}

\begin{exercise}
Find the Jacobian of
\begin{equation}
  F(x_1,x_2)=
  \begin{bmatrix}
    x_1x_2\\
    x_1^2+\sin x_2
  \end{bmatrix}.
\end{equation}
\end{exercise}
\begin{proof}[Short solution]
Differentiating each output with respect to each input gives
\begin{equation}
  J_F(x)=
  \begin{bmatrix}
    x_2 & x_1\\
    2x_1 & \cos x_2
  \end{bmatrix}.
\end{equation}
\end{proof}

\begin{exercise}
Let $f(x)=\log(1+e^{a\trans x})$. Derive $\nabla f(x)$ using the chain rule,
and determine the sign of the directional derivative in direction $d=-a$.
\end{exercise}
\begin{proof}[Short solution]
With $s=a\trans x$,
\begin{equation}
  \nabla f(x)=\frac{e^s}{1+e^s}a.
\end{equation}
Therefore
\begin{equation}
  Df(x)[-a]
  =-\frac{e^s}{1+e^s}\norm{a}_2^2,
\end{equation}
which is negative whenever $a\neq0$.
\end{proof}

\begin{exercise}
For $f(x)=\tfrac12\norm{Ax-b}_2^2$, derive the gradient and Hessian.
\end{exercise}
\begin{proof}[Short solution]
Writing $r(x)=Ax-b$ and applying the chain rule gives
\begin{equation}
  \nabla f(x)=A\trans(Ax-b),
  \qquad
  \nabla^2f(x)=A\trans A.
\end{equation}
The Hessian is positive semidefinite because
$d\trans A\trans Ad=\norm{Ad}_2^2\geq0$.
\end{proof}

\begin{exercise}
Compute the gradient and Hessian of
$f(x)=\tfrac12x\trans Qx+c\trans x$, where $Q$ need not be symmetric. Show that
only the symmetric part $Q_s=(Q+Q\trans)/2$ affects the quadratic form.
\end{exercise}
\begin{proof}[Short solution]
The skew-symmetric part $Q_a=(Q-Q\trans)/2$ satisfies
$x\trans Q_ax=0$, so $x\trans Qx=x\trans Q_sx$. Hence
\begin{equation}
  \nabla f(x)=Q_sx+c,
  \qquad
  \nabla^2f(x)=Q_s.
\end{equation}
\end{proof}

\begin{exercise}
For $f(x_1,x_2)=x_1^2+x_1x_2+2x_2^2$, write the second-order Taylor expansion
about an arbitrary point $x$ and explain why it is exact.
\end{exercise}
\begin{proof}[Short solution]
The gradient and Hessian are
\begin{equation}
  \nabla f(x)=
  \begin{bmatrix}2x_1+x_2\\x_1+4x_2\end{bmatrix},
  \qquad
  H=\begin{bmatrix}2&1\\1&4\end{bmatrix}.
\end{equation}
Therefore
\begin{equation}
  f(x+h)=f(x)+\nabla f(x)\trans h+\frac12h\trans Hh.
\end{equation}
The formula is exact because $f$ is quadratic and has no derivatives of order
three or higher.
\end{proof}

\begin{exercise}
Find the second-order Taylor model of $f(x_1,x_2)=e^{x_1+x_2}$ at the origin.
\end{exercise}
\begin{proof}[Short solution]
At the origin, $f(0)=1$, $\nabla f(0)=(1,1)\trans$, and
\begin{equation}
  \nabla^2f(0)=\begin{bmatrix}1&1\\1&1\end{bmatrix}.
\end{equation}
Thus
\begin{equation}
  m_2(h;0)=1+h_1+h_2+\frac12(h_1+h_2)^2.
\end{equation}
\end{proof}

\begin{exercise}
Suppose $\nabla f(x)\neq0$. Among all unit directions $d$, show that the
directional derivative $Df(x)[d]$ is smallest when
$d=-\nabla f(x)/\norm{\nabla f(x)}_2$.
\end{exercise}
\begin{proof}[Short solution]
For every unit vector $d$, Cauchy--Schwarz gives
\begin{equation}
  Df(x)[d]=\nabla f(x)\trans d
  \geq-\norm{\nabla f(x)}_2.
\end{equation}
Equality holds when $d$ points exactly opposite the gradient, proving that the
negative normalized gradient is the direction of steepest local decrease.
\end{proof}

\section{Unconstrained Optimality Conditions}

We now use local derivative information to characterize candidate solutions of
\begin{equation}
  \min_{x\in\R^n} f(x).
  \label{eq:unconstrained-problem}
\end{equation}
This compact expression specifies an optimization problem:
\begin{itemize}
  \item $x=(x_1,\ldots,x_n)\trans$ is the \emph{decision variable}, the vector
  whose entries are to be chosen;
  \item $x\in\R^n$ states that every candidate choice is a real vector with
  $n$ components;
  \item $f:\R^n\to\R$ is the \emph{objective function}, which assigns a scalar
  cost $f(x)$ to each choice $x$; and
  \item $\min$ asks for a choice of $x$ whose objective value is no larger than
  that of any other admissible choice.
\end{itemize}
Thus, the variable $x$ is the input being selected, whereas $f(x)$ is the
number being made as small as possible. If a solution exists, a minimizer is
usually denoted by $x^\star$ and satisfies
\begin{equation}
  f(x^\star)\leq f(x)\qquad\text{for every }x\in\R^n.
\end{equation}
The vector $x^\star$ is an \emph{optimal solution}, while the scalar
\begin{equation}
  f^\star=f(x^\star)=\min_{x\in\R^n}f(x)
\end{equation}
is the \emph{optimal value}. These objects should not be confused. When several
points attain the same smallest value, the complete set of solutions is written
\begin{equation}
  \argmin_{x\in\R^n}f(x)
  =\{x\in\R^n:f(x)=f^\star\}.
\end{equation}
The notation $x^\star\in\argmin f$ selects one member of this set. A minimum
need not exist. To distinguish existence of an optimizer from the best
achievable bound, it is useful to define minimum, infimum, maximum, and
supremum more formally.

\begin{definition}[Minimum, infimum, maximum, and supremum]
Let $\mathcal X$ be a nonempty feasible set and let
$f:\mathcal X\to\R$. A number $\alpha$ is a \emph{lower bound} for $f$ on
$\mathcal X$ if
\begin{equation}
  \alpha\leq f(x)
  \qquad\text{for every }x\in\mathcal X.
\end{equation}
If $f$ is bounded below, its \emph{infimum} is the unique number
$\alpha=\inf_{x\in\mathcal X}f(x)$ satisfying
\begin{align}
  \alpha&\leq f(x)
    &&\text{for every }x\in\mathcal X,\\
  f(x_\epsilon)&<\alpha+\epsilon
    &&\text{for every }\epsilon>0
      \text{ and some }x_\epsilon\in\mathcal X.
\end{align}
Thus $f$ never falls below its infimum, but its values can be found
arbitrarily close to the infimum from above. The infimum is a
\emph{minimum} if it is attained: there exists
$x^\star\in\mathcal X$ such that
\begin{equation}
  f(x^\star)=\inf_{x\in\mathcal X}f(x).
\end{equation}
In that case,
$\min_{x\in\mathcal X}f(x)=f(x^\star)$ and
$x^\star\in\argmin_{x\in\mathcal X}f(x)$.

Similarly, $\beta$ is an \emph{upper bound} if
$f(x)\leq\beta$ for every $x\in\mathcal X$. If $f$ is bounded above, its
\emph{supremum} is the unique number
$\beta=\sup_{x\in\mathcal X}f(x)$ satisfying
\begin{align}
  f(x)&\leq\beta
    &&\text{for every }x\in\mathcal X,\\
  \beta-\epsilon&<f(x_\epsilon)
    &&\text{for every }\epsilon>0
      \text{ and some }x_\epsilon\in\mathcal X.
\end{align}
It is a \emph{maximum} when some $x^\star$ attains it. Then
\begin{equation}
  f(x^\star)=\max_{x\in\mathcal X}f(x)
  =\sup_{x\in\mathcal X}f(x),
  \qquad
  x^\star\in\argmax_{x\in\mathcal X}f(x).
\end{equation}
\end{definition}

Thus $\inf$ and $\sup$ describe limiting objective values and do not assert
that an optimizing point exists. By contrast, writing $\min$ or $\max$
asserts that the corresponding value is attained. For example,
$f(x)=e^x$ on $\R$ has
\begin{equation}
  \inf_{x\in\R}e^x=0,
  \qquad
  \sup_{x\in\R}e^x=+\infty,
\end{equation}
but it has neither a minimum nor a maximum. Extended values such as
$+\infty$ or $-\infty$ indicate that the objective is unbounded in the
corresponding direction.

The word \emph{unconstrained} refers to the admissible choices. Here the
\emph{feasible set} is all of $\R^n$: there are no additional equality
constraints, inequality constraints, or boundary restrictions. Consequently,
from any point $x$, every direction $d\in\R^n$ is locally allowable, since
$x+td\in\R^n$ for all sufficiently small positive and negative $t$.

Unconstrained does \emph{not} mean that the problem is easy or that the
objective has a unique minimum. The function may be nonconvex, may have many
local minima and saddle points, or may fail to attain a minimum. The absence of
constraints only describes the feasible set.

This distinction explains the first-order condition developed below. At an
unconstrained local minimizer, movement is possible in both $d$ and $-d$, so
the directional derivative must vanish in every direction. In a constrained
problem, some directions may be forbidden by the feasible-set boundary, and an
optimum may therefore have a nonzero gradient.

\subsection{Local and Global Minima}

\begin{definition}[Global and local minimizers]
A point $x^\star\in\R^n$ is a global minimizer of $f$ if
\begin{equation}
  f(x^\star)\leq f(x)\qquad\text{for every }x\in\R^n.
\end{equation}
It is a local minimizer if there exists $r>0$ such that
\begin{equation}
  f(x^\star)\leq f(x)
  \qquad\text{whenever }\norm{x-x^\star}_2<r.
\end{equation}
The minimizer is \emph{strict} if equality occurs only at $x^\star$. More
precisely, there is an $r>0$ such that
\begin{equation}
  f(x^\star)<f(x)
  \qquad\text{whenever }0<\norm{x-x^\star}_2<r.
\end{equation}
Here $<$ is a \emph{strict inequality}: it excludes equality, unlike the
non-strict inequality $\leq$. Thus every sufficiently close point distinct from
$x^\star$ has a genuinely larger objective value. A non-strict local minimum
may instead lie on a flat region containing other nearby points with the same
objective value.
\end{definition}

Every global minimizer is local, but a local minimizer need not be global. As
the preceding definitions emphasize, even a finite infimum need not be
attained.

Existence is a separate question from optimality. In the examples below, the
quadratic structure will make existence easy to check directly.

\subsection{First-Order Necessary Conditions}

\begin{definition}[Stationary point]
A differentiable function has a stationary point at $x$ if
\begin{equation}
  \nabla f(x)=0.
\end{equation}
\end{definition}

\begin{theorem}[First-order necessary condition]
If $f:\R^n\to\R$ is differentiable and $x^\star$ is a local minimizer, then
\begin{equation}
  \nabla f(x^\star)=0.
  \label{eq:first-order-necessary}
\end{equation}
\end{theorem}

Intuitively, imagine the graph of $f$ as a smooth landscape. The gradient
$\nabla f(x)$ points in the direction of greatest local increase, while
$-\nabla f(x)$ points downhill. If the gradient at $x^\star$ were nonzero, then
moving a sufficiently small distance in the direction
$d=-\nabla f(x^\star)$ would decrease the objective because
\begin{equation}
  Df(x^\star)[d]
  =-\norm{\nabla f(x^\star)}_2^2<0.
\end{equation}
That would contradict the assumption that $x^\star$ is a local minimizer.
Therefore a differentiable objective cannot have a nonzero slope in any
direction at an unconstrained local minimum.

Equivalently, restrict the function to any line through the point by defining
$\phi(t)=f(x^\star+td)$. Because both positive and negative small values of $t$
are allowable, $t=0$ must minimize this one-dimensional function, and hence
\begin{equation}
  \phi'(0)=\nabla f(x^\star)\trans d=0.
\end{equation}
This equality holds for every direction $d$. Choosing
$d=\nabla f(x^\star)$ gives $\norm{\nabla f(x^\star)}_2^2=0$, so the gradient
must be zero.

Stationarity is necessary but generally not sufficient. For example,
$f(x)=x^3$ has $f'(0)=0$, yet values on one side of zero are smaller and values
on the other side are larger. A stationary point may be a minimum, a maximum,
or neither.

\subsection{Second-Order Necessary and Sufficient Conditions}

First derivatives detect slope; second derivatives distinguish different types
of stationary points through curvature.

\begin{definition}[Matrix definiteness]
A symmetric matrix $H$ is positive semidefinite, written $H\succeq0$, if
$d\trans Hd\geq0$ for every $d$. It is positive definite, written $H\succ0$,
if the inequality is strict for every nonzero $d$. Negative semidefiniteness and
negative definiteness are defined by reversing the inequalities. A square matrix
is singular if it is not invertible.
\end{definition}

\begin{theorem}[Second-order necessary condition]
If $f$ is twice differentiable in a neighborhood of a local minimizer
$x^\star$, then
\begin{equation}
  \nabla f(x^\star)=0,
  \qquad
  \nabla^2 f(x^\star)\succeq0.
\end{equation}
\end{theorem}

The curvature condition follows by restricting $f$ to any line through
$x^\star$: its second derivative is
$d\trans\nabla^2 f(x^\star)d$ and cannot be negative at a local minimum.

\begin{theorem}[Second-order sufficient condition]
Suppose $f$ is twice continuously differentiable near $x^\star$ and
\begin{equation}
  \nabla f(x^\star)=0,
  \qquad
  \nabla^2 f(x^\star)\succ0.
\end{equation}
Then $x^\star$ is a strict local minimizer.
\end{theorem}

Positive semidefiniteness alone is inconclusive. Both $f(x)=x^4$ and
$f(x)=x^3$ have zero gradient and zero Hessian at the origin, but the former has
a strict minimum there and the latter does not. Higher-order terms determine the
answer when the Hessian is singular.

A stationary point whose Hessian has both positive and negative curvature
directions is called a \emph{saddle point}. Equivalently, the symmetric Hessian
has at least one positive eigenvalue and at least one negative eigenvalue; such
a matrix is called \emph{indefinite}. If $v_+$ and $v_-$ are corresponding
unit eigenvectors with eigenvalues $\lambda_+>0$ and $\lambda_-<0$, then
\begin{equation}
  v_+\trans\nabla^2 f(x)v_+=\lambda_+>0,
  \qquad
  v_-\trans\nabla^2 f(x)v_-=\lambda_-<0.
\end{equation}
Thus the function curves upward along $v_+$ and downward along $v_-$, so the
stationary point cannot be a local minimum or a local maximum.

\begin{exercise}
Classify all stationary points of
$f(x_1,x_2)=x_1^2-x_2^2$ using its Hessian.
\end{exercise}
\begin{proof}[Short solution]
$\nabla f=(2x_1,-2x_2)\trans$, so the only stationary point is $(0,0)$.
The Hessian is $\operatorname{diag}(2,-2)$, which has one positive and one
negative eigenvalue. Hence the origin is a saddle point.
\end{proof}

\begin{exercise}
Find and classify the stationary points of $f(x)=x^4-2x^2$. Which minima are
global?
\end{exercise}
\begin{proof}[Short solution]
$f'(x)=4x(x^2-1)$, so the stationary points are $0$ and $\pm1$. Since
$f''(x)=12x^2-4$, the origin is a strict local maximum and $\pm1$ are strict
local minima. Because $f(x)=(x^2-1)^2-1\geq-1$, both minima are global.
\end{proof}

\begin{exercise}
Find and classify the stationary point of
$f(x_1,x_2)=x_1^2+4x_2^2-2x_1+8x_2$.
\end{exercise}
\begin{proof}[Short solution]
$\nabla f=(2x_1-2,8x_2+8)\trans$, so $x^\star=(1,-1)$. The Hessian
$\operatorname{diag}(2,8)$ is positive definite; therefore $x^\star$ is the
unique strict global minimizer.
\end{proof}

\begin{exercise}
Classify the origin for $f(x_1,x_2)=x_1^4+x_2^2$. Why is the second-order
test alone inconclusive there?
\end{exercise}
\begin{proof}[Short solution]
The gradient vanishes at the origin, where the Hessian is
$\operatorname{diag}(0,2)$, which is positive semidefinite but not positive
definite. Nevertheless, $f(x_1,x_2)>0=f(0,0)$ for every nonzero point, so the
origin is a strict global minimizer. The fourth-order term supplies the missing
curvature in the $x_1$ direction.
\end{proof}

\begin{exercise}
Classify the origin for $f(x_1,x_2)=x_1^3+x_2^2$ and compare its gradient and
Hessian there with those in the previous exercise.
\end{exercise}
\begin{proof}[Short solution]
At the origin the gradient is zero and the Hessian is again
$\operatorname{diag}(0,2)$. However, $f(x_1,0)=x_1^3$ is positive for $x_1>0$
and negative for $x_1<0$. Thus the origin is a saddle point. This shows that a
singular positive-semidefinite Hessian does not determine the classification.
\end{proof}

\begin{exercise}
For
\begin{equation}
  f_a(x_1,x_2)=x_1^2+2a x_1x_2+x_2^2,
\end{equation}
classify the origin for all $a\in\R$.
\end{exercise}
\begin{proof}[Short solution]
The Hessian has eigenvalues $2(1+a)$ and $2(1-a)$. If $|a|<1$, both are
positive and the origin is a strict global minimum. If $|a|>1$, their signs
differ and the origin is a saddle point. If $|a|=1$, the function is
$(x_1\pm x_2)^2$, so the origin is a non-strict global minimum lying on a line
of minimizers.
\end{proof}

\begin{exercise}
Find and classify the stationary point of
$f(x_1,x_2)=-x_1^2-3x_2^2+4x_1-6x_2$.
\end{exercise}
\begin{proof}[Short solution]
Solving $\nabla f=(-2x_1+4,-6x_2-6)\trans=0$ gives $(2,-1)$. The Hessian
$\operatorname{diag}(-2,-6)$ is negative definite, so this point is the unique
strict global maximum and is not a minimum.
\end{proof}

\begin{exercise}
Find and classify every stationary point of
$f(x_1,x_2)=(x_1^2+x_2^2-1)^2$.
\end{exercise}
\begin{proof}[Short solution]
The gradient is $4(x_1^2+x_2^2-1)(x_1,x_2)\trans$. Thus the stationary set is
the origin together with the unit circle. The origin has Hessian $-4I$ and is a
strict local maximum. Every point on the unit circle is a non-strict global
minimum because $f=0$ there and nearby points on the circle have the same value.
\end{proof}

\begin{exercise}
Classify the origin for $f(x_1,x_2)=x_1^2-x_2^4$. What does this example show
about the second-order necessary condition?
\end{exercise}
\begin{proof}[Short solution]
The gradient vanishes and the Hessian at the origin is
$\operatorname{diag}(2,0)\succeq0$. Yet $f(x_1,0)>0$ for $x_1\neq0$, while
$f(0,x_2)<0$ for $x_2\neq0$. Hence the origin is a saddle point. Positive
semidefiniteness is necessary at a local minimum but is not sufficient.
\end{proof}

\begin{exercise}
Let
\begin{equation}
  f(x)=\frac12x\trans Hx+c\trans x,
\end{equation}
where $H=H\trans\succ0$. Find the stationary point and prove directly that it
is the unique global minimizer.
\end{exercise}
\begin{proof}[Short solution]
Since $\nabla f(x)=Hx+c$, the stationary point is
$x^\star=-H^{-1}c$. Completing the square gives
\begin{equation}
  f(x)=f(x^\star)+\frac12(x-x^\star)\trans H(x-x^\star).
\end{equation}
The final term is positive for every $x\neq x^\star$, proving uniqueness and
global optimality.
\end{proof}

\section{Canonical Optimization Examples}

The following problems illustrate how geometry and derivatives combine in
models that recur throughout optimization.

\subsection{Least-Squares Problems}

Given $A\in\R^{m\times n}$ and observations $b\in\R^m$, the linear
least-squares problem is
\begin{equation}
  \min_{x\in\R^n} f(x),
  \qquad
  f(x)=\frac12\norm{Ax-b}_2^2.
  \label{eq:linear-least-squares}
\end{equation}
The residual $r(x)=Ax-b$ measures disagreement between the model $Ax$ and the
data $b$. From \cref{eq:basic-gradient-identities},
\begin{equation}
  \nabla f(x)=A\trans(Ax-b),
  \qquad
  \nabla^2 f(x)=A\trans A\succeq0.
\end{equation}
Every minimizer therefore satisfies the normal equations
\begin{equation}
  A\trans Ax=A\trans b.
  \label{eq:least-squares-normal-equations}
\end{equation}

If the columns of $A$ are linearly independent, then $A\trans A\succ0$ and the
minimizer is unique:
\begin{equation}
  x^\star=(A\trans A)^{-1}A\trans b.
\end{equation}
Geometrically, $Ax^\star$ is the orthogonal projection of $b$ onto the column
space of $A$, and $b-Ax^\star$ is orthogonal to that space. If the columns are
dependent, minimizers may not be unique, but all of them produce the same fitted
vector $Ax^\star$.

A weighted least-squares model assigns different importance to different
residual directions:
\begin{equation}
  \min_x \frac12(Ax-b)\trans W(Ax-b),
  \qquad W\succ0.
\end{equation}
Its normal equations are $A\trans WAx=A\trans Wb$.

\subsection{Quadratic Optimization}

A general unconstrained quadratic function can be written
\begin{equation}
  f(x)=\frac12x\trans Qx+c\trans x+r,
  \label{eq:quadratic-objective}
\end{equation}
where $Q=Q\trans$, $c\in\R^n$, and $r\in\R$. Any nonsymmetric quadratic
matrix can be replaced by its symmetric part because
$x\trans Qx=x\trans(Q+Q\trans)x/2$.

The derivatives are
\begin{equation}
  \nabla f(x)=Qx+c,
  \qquad
  \nabla^2 f(x)=Q.
\end{equation}
If $Q\succ0$, the unique stationary point
\begin{equation}
  x^\star=-Q^{-1}c
\end{equation}
is the unique global minimizer. Completing the square makes this explicit:
\begin{equation}
  f(x)=f(x^\star)
  +\frac12(x-x^\star)\trans Q(x-x^\star).
\end{equation}

If $Q\succeq0$ but is singular, a minimizer exists only when the equation
$Qx=-c$ is solvable. Here \emph{solvable} means that there is at least one
vector $x_p$ satisfying $Qx_p=-c$; equivalently, $-c$ belongs to the range of
$Q$. Because a singular matrix has no inverse, one cannot simply write
$x=-Q^{-1}c$, and the system may have either no solution or infinitely many
solutions.

When one solution $x_p$ exists, every solution has the form
\begin{equation}
  x=x_p+z,\qquad z\in\ker(Q),
\end{equation}
and all such vectors are global minimizers. Indeed,
\begin{equation}
  f(x)=f(x_p)+\frac12(x-x_p)\trans Q(x-x_p)\geq f(x_p).
\end{equation}
Moving in a null-space direction does not change the objective. Since $Q$ is
symmetric, solvability is also equivalent to $c$ being orthogonal to every
vector in $\ker(Q)$. If this condition fails, the linear term decreases without
bound along one of those zero-curvature directions, so no minimizer exists.
If $Q$ has a negative eigenvalue, the objective tends to $-\infty$ along a
suitable direction and has no global minimizer.

Quadratic functions are especially important because the second-order Taylor
model of any twice differentiable objective is quadratic.

\subsection{Regularization and Ill-Posed Problems}

An optimization problem is informally called \emph{ill-posed} when a solution
fails to exist, is not unique, or changes excessively under small changes in the
data. A matrix is rank deficient when its rank is smaller than its number of
columns. Rank-deficient least squares illustrates the nonuniqueness issue. Even
when $A\trans A$ is invertible, very small eigenvalues can make the solution
sensitive to perturbations in $b$. To see why, write
\begin{equation}
  A\trans A=U\Lambda U\trans,
  \qquad
  \Lambda=\operatorname{diag}(\lambda_1,\ldots,\lambda_n),
\end{equation}
where every $\lambda_i>0$. In the eigenvector coordinates of $A\trans A$, the
least-squares solution is
\begin{equation}
  x^\star
  =\sum_{i=1}^n
  \frac{u_i\trans A\trans b}{\lambda_i}u_i.
\end{equation}
Thus each component is obtained by dividing by an eigenvalue. If the data
changes from $b$ to $b+\delta b$, the resulting change $\delta x$ satisfies
\begin{equation}
  (A\trans A)\delta x=A\trans\delta b.
\end{equation}
Writing $\delta g=A\trans\delta b$, its component along $u_i$ is
\begin{equation}
  u_i\trans\delta x=\frac{u_i\trans\delta g}{\lambda_i}.
\end{equation}
A small $\lambda_i$ can therefore amplify a small change in the corresponding
data-driven component. Geometrically, the least-squares objective has curvature
$\lambda_i$ along $u_i$, so a small eigenvalue describes a nearly flat
direction: the objective changes very little as $x$ moves in that direction,
and the location of its minimizer can move substantially under a small data
perturbation. The ratio
$\lambda_{\max}(A\trans A)/\lambda_{\min}(A\trans A)$ is the spectral condition
number of $A\trans A$; a large ratio signals this numerical sensitivity.

Regularization modifies the objective to prefer solutions with additional
structure. The most common smooth example is Tikhonov, or ridge,
regularization:
\begin{equation}
  \min_x
  \frac12\norm{Ax-b}_2^2+\frac{\lambda}{2}\norm{x}_2^2,
  \qquad \lambda>0.
  \label{eq:ridge-regression}
\end{equation}
Its Hessian is $A\trans A+\lambda I\succ0$, so the minimizer is unique even when
$A$ has dependent columns:
\begin{equation}
  x_\lambda=(A\trans A+\lambda I)^{-1}A\trans b.
\end{equation}
The parameter $\lambda$ controls a tradeoff. A larger value reduces the size and
data sensitivity of $x_\lambda$ but generally increases the residual
$\norm{Ax_\lambda-b}_2$.

More generally, one may solve
\begin{equation}
  \min_x f(x)+\lambda R(x),
\end{equation}
where $R$ expresses a preference such as small magnitude, smoothness, or few
nonzero coordinates. The function $R(x)=\norm{x}_1$ favors solutions with many
zero entries, but it is not differentiable when a coordinate is zero. Later
algorithmic tools will handle such objectives without pretending that an
ordinary gradient exists there.

\begin{exercise}
Derive the weighted least-squares normal equations. Under what condition on $A$
is the weighted least-squares minimizer unique?
\end{exercise}
\begin{proof}[Short solution]
For $f(x)=\tfrac12(Ax-b)\trans W(Ax-b)$,
\begin{equation}
  \nabla f(x)=A\trans W(Ax-b).
\end{equation}
Thus the normal equations are $A\trans WAx=A\trans Wb$. Since $W\succ0$,
$A\trans WA\succ0$ exactly when $A$ has linearly independent columns. This is
the condition for a unique minimizer.
\end{proof}

\begin{exercise}
For the quadratic function in \cref{eq:quadratic-objective}, prove that a
negative eigenvalue of $Q$ makes the objective unbounded below.
\end{exercise}
\begin{proof}[Short solution]
Let $Qv=\mu v$ with $\norm{v}_2=1$ and $\mu<0$. Along $x=tv$,
\begin{equation}
  f(tv)=\frac12\mu t^2+t c\trans v+r.
\end{equation}
The negative quadratic term dominates the linear term as $|t|\to\infty$, so
$f(tv)\to-\infty$.
\end{proof}

\begin{exercise}
Diagonalize $A\trans A=U\Lambda U\trans$ and express the ridge solution in the
eigenvector coordinates of $A\trans A$. Explain how $\lambda$ limits the effect
of small eigenvalues.
\end{exercise}
\begin{proof}[Short solution]
If $u_i$ has eigenvalue $\lambda_i\geq0$, then
\begin{equation}
  x_\lambda
  =\sum_{i=1}^n
  \frac{u_i\trans A\trans b}{\lambda_i+\lambda}u_i.
\end{equation}
Ridge replaces division by $\lambda_i$ with division by
$\lambda_i+\lambda\geq\lambda$. Hence components associated with small or zero
eigenvalues cannot be amplified without bound.
\end{proof}

\begin{exercise}
Solve the least-squares problem with
\begin{equation}
  A=\begin{bmatrix}1&0\\1&1\\0&1\end{bmatrix},
  \qquad
  b=\begin{bmatrix}1\\2\\0\end{bmatrix},
\end{equation}
and verify that the residual is orthogonal to every column of $A$.
\end{exercise}
\begin{proof}[Short solution]
Here
$A\trans A=\begin{bmatrix}2&1\\1&2\end{bmatrix}$ and
$A\trans b=(3,2)\trans$, giving
$x^\star=(4/3,1/3)\trans$. The residual is
$r=Ax^\star-b=(1/3,-1/3,1/3)\trans$, and direct calculation gives
$A\trans r=0$.
\end{proof}

\begin{exercise}
Let
\begin{equation}
  A=\begin{bmatrix}1&1\\2&2\end{bmatrix},
  \qquad
  b=\begin{bmatrix}1\\2\end{bmatrix}.
\end{equation}
Find every least-squares minimizer and identify the one with smallest Euclidean
norm.
\end{exercise}
\begin{proof}[Short solution]
The residual is zero exactly when $x_1+x_2=1$, so all points
$(t,1-t)\trans$ are global minimizers. Their squared norms are
$t^2+(1-t)^2$, which is minimized at $t=1/2$. The minimum-norm solution is
$(1/2,1/2)\trans$.
\end{proof}

\begin{exercise}
Minimize
\begin{equation}
  f(x)=\frac12x\trans
  \begin{bmatrix}2&1\\1&2\end{bmatrix}x
  +\begin{bmatrix}-3&0\end{bmatrix}x.
\end{equation}
Find both the minimizer and the minimum value.
\end{exercise}
\begin{proof}[Short solution]
The matrix is positive definite, and $Qx=-c=(3,0)\trans$ gives
$x^\star=(2,-1)\trans$. Substitution yields $f(x^\star)=-3$, which is the
unique global minimum value.
\end{proof}

\begin{exercise}
Let $Q=\operatorname{diag}(2,0)$. Analyze the quadratic objective for
(a) $c=(-2,0)\trans$ and (b) $c=(-2,1)\trans$, taking $r=0$.
\end{exercise}
\begin{proof}[Short solution]
In (a), $f(x)=x_1^2-2x_1=(x_1-1)^2-1$, so every $(1,t)\trans$ is a global
minimizer. In (b), $f(x)=x_1^2-2x_1+x_2$, which tends to $-\infty$ as
$x_2\to-\infty$. Equivalently, $Qx=-c$ is solvable in (a) but not in (b).
\end{proof}

\begin{exercise}
For the rank-deficient matrix and vector in the fifth exercise above, compute
the ridge solution $x_\lambda$ and find its limit as $\lambda\downarrow0$.
\end{exercise}
\begin{proof}[Short solution]
Since
$A\trans A=\begin{bmatrix}5&5\\5&5\end{bmatrix}$ and
$A\trans b=(5,5)\trans$, symmetry gives
\begin{equation}
  x_\lambda
  =\frac{5}{10+\lambda}
  \begin{bmatrix}1\\1\end{bmatrix}.
\end{equation}
Therefore $x_\lambda\to(1/2,1/2)\trans$, the minimum-norm least-squares
solution.
\end{proof}

\begin{exercise}
Show that the least-squares normal equations are equivalent to saying that the
residual $b-Ax^\star$ is orthogonal to $\operatorname{range}(A)$.
\end{exercise}
\begin{proof}[Short solution]
The normal equations give $A\trans(b-Ax^\star)=0$. Hence, for every vector
$y=Ad$ in the range of $A$,
\begin{equation}
  y\trans(b-Ax^\star)
  =d\trans A\trans(b-Ax^\star)=0.
\end{equation}
The converse follows by choosing every $y=Ad$, so the two statements are
equivalent.
\end{proof}

\begin{exercise}
Prove that the ridge solution tends to zero as $\lambda\to\infty$, and give a
simple bound on its norm.
\end{exercise}
\begin{proof}[Short solution]
The smallest eigenvalue of $A\trans A+\lambda I$ is at least $\lambda$, so
\begin{equation}
  \norm{x_\lambda}_2
  \leq\frac{\norm{A\trans b}_2}{\lambda}.
\end{equation}
The right-hand side tends to zero as $\lambda\to\infty$; therefore
$x_\lambda\to0$.
\end{proof}

\paragraph{Chapter summary.}
Finite-dimensional optimization rests on a small set of ideas. Vector-space
geometry defines directions and distances; derivatives create local linear and
quadratic models; optimality conditions identify and classify candidates; and
least-squares, quadratic, and regularized objectives provide canonical examples.
The next chapter adds convexity, a global geometric structure under which local
information becomes much more powerful.
\chapter{Convex Optimization and Standard Problems}
\chapteroverview{Convex geometry, global optimality, and the problem forms most useful in engineering.}

Convexity is valuable because it turns local information into global
information. For a convex problem, a local minimum is global, and first-order
conditions are often enough to certify a solution. We develop only the tools
needed later for constrained optimization, control, and learning.

\section{Convex Sets and Functions}

\subsection{Convex Sets and Convex Combinations}

\begin{definition}[Convex set]
A set $C\subseteq\R^n$ is convex if
\begin{equation}
  \theta x+(1-\theta)y\in C
\end{equation}
for every $x,y\in C$ and $\theta\in[0,1]$. Thus the line segment between any
two points of $C$ remains inside $C$.
\end{definition}

A convex combination of $x_1,\ldots,x_k$ is
\begin{equation}
  \sum_{i=1}^k\theta_i x_i,
  \qquad \theta_i\geq0,
  \qquad \sum_{i=1}^k\theta_i=1.
\end{equation}
The convex hull $\operatorname{conv}(S)$ is the set of all finite convex
combinations of points in $S$; it is the smallest convex set containing $S$.

Several important sets are convex:
\begin{itemize}
  \item An \emph{affine set} has the form
  \begin{equation}
    x_0+\mathcal S=\{x_0+s:s\in\mathcal S\},
  \end{equation}
  where $\mathcal S$ is a linear subspace. If two points belong to this set,
  every point on the line through them also belongs to it, and therefore so
  does the line segment between them. Solution sets of linear equations,
  $\{x:Ax=b\}$, are affine whenever they are nonempty.

  \item A \emph{norm ball} centered at $x_c$ with radius $r\geq0$ is
  \begin{equation}
    \{x:\norm{x-x_c}\leq r\}.
  \end{equation}
  If $x$ and $y$ lie in the ball, the triangle inequality gives
  \begin{align}
    \norm{\theta x+(1-\theta)y-x_c}
    &\leq \theta\norm{x-x_c}+(1-\theta)\norm{y-x_c}\\
    &\leq r,
  \end{align}
  so the segment joining them remains in the ball. This argument applies to
  every norm, not only the Euclidean norm.

  \item A \emph{halfspace} is a set
  \begin{equation}
    \{x:a\trans x\leq b\}.
  \end{equation}
  For feasible $x$ and $y$,
  \begin{equation}
    a\trans\bigl(\theta x+(1-\theta)y\bigr)
    =\theta a\trans x+(1-\theta)a\trans y\leq b,
  \end{equation}
  which proves convexity. Geometrically, the hyperplane
  $\{x:a\trans x=b\}$ divides space into two halfspaces.
\end{itemize}

Convexity is also preserved by several constructions. The intersection
$\bigcap_{j\in J}C_j$ of any collection of convex sets is convex: two points in
the intersection belong to every $C_j$, so their entire line segment belongs to
every $C_j$ as well. This fact explains why imposing several convex constraints
simultaneously still produces a convex feasible set.

If $C$ is convex and $T(x)=Ax+b$ is affine, its \emph{affine image}
\begin{equation}
  T(C)=\{Ax+b:x\in C\}
\end{equation}
is convex because affine maps preserve convex combinations:
\begin{equation}
  T(\theta x+(1-\theta)y)
  =\theta T(x)+(1-\theta)T(y).
\end{equation}
Similarly, if $D$ is convex, the \emph{inverse image}
\begin{equation}
  T^{-1}(D)=\{x:Ax+b\in D\}
\end{equation}
is convex. These closure rules generate most feasible sets used in engineering
from a small collection of elementary building blocks.

\subsection{Convex Functions and Supporting Models}

\begin{definition}[Convex function]
A function $f:C\to\R$, defined on a convex set $C$, is convex if
\begin{equation}
  f(\theta x+(1-\theta)y)
  \leq \theta f(x)+(1-\theta)f(y)
  \label{eq:convex-function-definition}
\end{equation}
for all $x,y\in C$ and $\theta\in[0,1]$. It is strictly convex if the inequality
is strict for $x\neq y$ and $\theta\in(0,1)$.
\end{definition}

Geometrically, choose two points on the graph,
$(x,f(x))$ and $(y,f(y))$. The straight chord joining them contains the point
\begin{equation}
  \bigl(\theta x+(1-\theta)y,
  \theta f(x)+(1-\theta)f(y)\bigr).
\end{equation}
At the same horizontal location, the graph has height
$f(\theta x+(1-\theta)y)$. The convexity inequality says that this height is no
greater than the height of the chord. Thus the curve lies on or below each
chord joining two of its points. For example, the parabola $f(x)=x^2$ bends
upward and lies below its chords. By contrast, $f(x)=-x^2$ lies above its chords
and is concave.

The set lying on or above the graph,
\begin{equation}
  \operatorname{epi} f=\{(x,t):t\geq f(x)\},
\end{equation}
is called the \emph{epigraph}. A function is convex exactly when its epigraph is
a convex set. Affine functions, norms, squared norms, and maxima of affine
functions are convex. A function is concave if its negative is convex.

For a differentiable function, convexity has a particularly useful
characterization.

\begin{theorem}[First-order characterization]
A differentiable function on an open convex set is convex if and only if
\begin{equation}
  f(y)\geq f(x)+\nabla f(x)\trans(y-x)
  \label{eq:first-order-convexity}
\end{equation}
for all $x,y$ in its domain.
\end{theorem}

Thus every tangent plane of a convex function is a global lower bound. If
$\nabla f(x^\star)=0$, then
$f(y)\geq f(x^\star)$ for every $y$, so $x^\star$ is globally optimal.

\paragraph{Advanced note: subgradients.}
The gradient-based characterization above applies only where $f$ is
differentiable. A more advanced theory, called \emph{subdifferential calculus},
extends similar geometric ideas to nonsmooth convex functions. In this theory,
a vector $g$ is called a \emph{subgradient} of $f$ at $x$ if
\begin{equation}
  f(y)\geq f(x)+g\trans(y-x)
  \qquad\text{for every }y.
\end{equation}
Thus the affine function $f(x)+g\trans(y-x)$ is a global supporting lower bound,
just as the tangent plane is in the differentiable case. The set of all such
vectors is the \emph{subdifferential}, denoted $\partial f(x)$. For example,
the absolute-value function is not differentiable at the origin, but
\begin{equation}
  \partial |x|\big|_{x=0}=[-1,1].
\end{equation}
The condition $0\in\partial f(x^\star)$ characterizes a global minimizer of a
convex function. We mention this result for context; a systematic treatment of
subgradients and nonsmooth algorithms is beyond the scope of this chapter.

\subsection{Curvature, Strong Convexity, and Smoothness}

If $f$ is twice continuously differentiable, it is convex exactly when
\begin{equation}
  \nabla^2 f(x)\succeq0
\end{equation}
throughout its convex domain. Positive definiteness everywhere is sufficient
for strict convexity.

A differentiable function is $\mu$-strongly convex if
\begin{equation}
  f(y)\geq f(x)+\nabla f(x)\trans(y-x)
  +\frac{\mu}{2}\norm{y-x}_2^2,
  \qquad \mu>0.
  \label{eq:strong-convexity}
\end{equation}
Strong convexity rules out flat directions and ensures that, if a minimizer
exists, it is unique.

A differentiable function is $L$-smooth if
\begin{equation}
  \norm{\nabla f(x)-\nabla f(y)}_2\leq L\norm{x-y}_2.
  \label{eq:lipschitz-gradient}
\end{equation}
It then satisfies the upper quadratic bound
\begin{equation}
  f(y)\leq f(x)+\nabla f(x)\trans(y-x)
  +\frac{L}{2}\norm{y-x}_2^2.
  \label{eq:descent-lemma}
\end{equation}
When both properties hold, $\kappa=L/\mu$ measures how uneven the curvature is.
Large $\kappa$ produces elongated level sets and slower first-order
optimization.

\section{Standard Convex Problems}

\subsection{General Form and Global Optimality}

In an unconstrained problem, every $x$ in the variable's domain is an allowable
choice. A \emph{constraint} is an additional requirement that restricts which
choices are allowed. A point satisfying every constraint is called
\emph{feasible}, and the set of all such points is the \emph{feasible set}.
A problem is \emph{infeasible} if no point satisfies all its constraints.

The two basic types are
\begin{align}
  f_i(x)&\leq0 &&\text{inequality constraint},\\
  h_j(x)&=0    &&\text{equality constraint}.
\end{align}
For example, $x_1+x_2\leq10$ may represent a resource limit, while
$x_1+x_2=10$ requires the entire resource to be used. At a feasible point, an
inequality constraint is \emph{active} if $f_i(x)=0$ and \emph{inactive} if
$f_i(x)<0$. Equality constraints are always active because they must hold
exactly.

The objective and the constraints play different roles. The objective ranks
the feasible points according to desirability, while the constraints determine
which points may be considered at all. Thus a constrained optimization problem
asks for the feasible point having the smallest objective value.

With this terminology, a convex optimization problem has the standard form
\begin{equation}
  \begin{aligned}
    \minimize_x\quad & f_0(x)\\
    \subjectto\quad & f_i(x)\leq0,
      \quad i=1,\ldots,m,\\
                    & Ax=b,
  \end{aligned}
  \label{eq:standard-convex-problem}
\end{equation}
Here $f_0$ is the objective, the functions $f_1,\ldots,f_m$ define the
inequality constraints, and $Ax=b$ collects affine equality constraints. The
feasible set is
\begin{equation}
  \mathcal F
  =\{x:f_i(x)\leq0, i=1,\ldots,m,\ Ax=b\}.
\end{equation}
The problem is convex when $f_0,f_1,\ldots,f_m$ are convex and the equality
constraints are affine. Each inequality defines a convex sublevel set, and
$Ax=b$ defines an affine set; their intersection $\mathcal F$ is therefore
convex. General nonlinear equalities are excluded because their solution sets
need not be convex.

\begin{theorem}[The central advantage of convex optimization]
For a convex optimization problem, every local minimizer is a global minimizer.
If the objective is strictly convex on the feasible set, the problem has at most
one minimizer.
\end{theorem}

This result is the principal reason convexity is so valuable: finding a point
that is best among all sufficiently nearby feasible points is enough to certify
that it is best among \emph{all} feasible points. Convex problems therefore have
no inferior local minima that can trap a local optimization method.

To see the first claim, suppose that a local minimizer $x^\star$ were not
global. Then some feasible $y$ would satisfy $f_0(y)<f_0(x^\star)$. Because the
feasible set is convex, the points
\begin{equation}
  x_\theta=(1-\theta)x^\star+\theta y,
  \qquad 0<\theta<1,
\end{equation}
are feasible and can be made arbitrarily close to $x^\star$. Convexity of the
objective gives
\begin{equation}
  f_0(x_\theta)
  \leq(1-\theta)f_0(x^\star)+\theta f_0(y)
  <f_0(x^\star),
\end{equation}
contradicting local optimality. For uniqueness, if two distinct minimizers
existed, strict convexity would make the objective at every point strictly
between them smaller than the common minimum value, which is impossible.

\subsection{Linear and Quadratic Programs}

A linear program (LP) has a linear objective and affine constraints:
\begin{equation}
  \begin{aligned}
    \minimize_x\quad & c\trans x\\
    \subjectto\quad & Gx\leq h,\\
                    & Ax=b.
  \end{aligned}
\end{equation}
A convex quadratic program (QP) has
\begin{equation}
  \begin{aligned}
    \minimize_x\quad & \frac12x\trans Qx+c\trans x\\
    \subjectto\quad & Gx\leq h,\\
                    & Ax=b,
  \end{aligned}
  \qquad Q\succeq0.
  \label{eq:quadratic-program}
\end{equation}
LPs model linear costs and resource limits; QPs additionally model energy,
tracking error, and local quadratic approximations. QPs are especially common
in optimal control and model predictive control.

Euclidean norm constraints such as
\begin{equation}
  \norm{Ax+b}_2\leq c\trans x+d
\end{equation}
define second-order cone constraints. Problems built from linear objectives,
affine equalities, and such constraints are second-order cone programs (SOCPs).
For an introductory treatment, it is enough to recognize that they remain
convex and can represent many norm-bounded design problems.

\paragraph{How these problems are solved in practice.}
LPs, QPs, and SOCPs are normally passed to specialized numerical solvers rather
than solved by hand. An LP may be solved by a simplex method, which moves among
vertices of the feasible polyhedron, or by an interior-point method, which
approaches the solution through the interior. Convex QPs are commonly solved by
active-set methods, interior-point methods, or first-order splitting methods;
these algorithms repeatedly solve linear systems involving the quadratic
objective and the active constraints. SOCPs are usually handled by conic
interior-point or conic first-order methods that preserve the cone structure.

In engineering software, the user typically specifies the objective and
constraints through a modeling interface. The software converts the model to a
standard matrix or cone form, selects or calls a suitable solver, and returns a
candidate solution together with feasibility and optimality diagnostics.
Sparsity, scaling, warm starts, and the requested accuracy strongly affect
runtime. The main algorithmic ideas are summarized in Chapter~4.

\section{Practical Convex Modeling}

\subsection{Epigraph and Minimax Reformulations}

An \emph{epigraph reformulation} introduces a scalar $t$ that bounds an
objective from above:
\begin{equation}
  \min_x f(x)
  \quad\Longleftrightarrow\quad
  \begin{aligned}
    \minimize_{x,t}\quad & t\\
    \subjectto\quad & f(x)\leq t.
  \end{aligned}
  \label{eq:epigraph-reformulation}
\end{equation}
For any fixed $x$, the smallest feasible value of $t$ is $f(x)$. Consequently,
the two formulations have the same optimal value and the same optimal choices
of $x$, and the bound is active at an attained optimum:
$t^\star=f(x^\star)$.

The reformulation does not remove the nonlinearity; it moves it from the
objective into a constraint. Its value is structural. A modeling system may
recognize the new constraint as a standard linear, quadratic, or cone
constraint and pass the problem to an appropriate solver. The reformulated
problem is not automatically easier, and introducing unnecessary variables can
even add overhead. The method is most useful when it exposes a standard problem
class or removes an explicit nonsmooth maximum from the objective.

\begin{example}[An unsquared norm modeled exactly]
The isolated problem $\min_x\norm{Ax-b}_2$ has the same minimizers as squared
least squares and is naturally treated as a QP. A cone reformulation is more
useful when squaring the norm would change the model. For example,
\begin{equation}
  \begin{aligned}
    \minimize_x\quad & c\trans x+\gamma\norm{Ax-b}_2\\
    \subjectto\quad & Gx\leq h,
  \end{aligned}
  \qquad \gamma>0,
\end{equation}
balances a linear cost against residual magnitude. It is equivalent to
\begin{equation}
  \begin{aligned}
    \minimize_{x,t}\quad & c\trans x+\gamma t\\
    \subjectto\quad
      & \norm{Ax-b}_2\leq t,\\
      & Gx\leq h.
  \end{aligned}
\end{equation}
This is an SOCP. Replacing the norm by its square would penalize residuals
differently and would produce a different problem. The SOCP is useful because
it preserves the intended norm model exactly, not because SOCPs are inherently
easier than QPs.
\end{example}

The same construction handles a pointwise maximum. If
$f(x)=\max_i f_i(x)$, then
\begin{equation}
  \min_x\max_i f_i(x)
  \quad\Longleftrightarrow\quad
  \begin{aligned}
    \minimize_{x,t}\quad & t\\
    \subjectto\quad & f_i(x)\leq t,
      \quad i=1,\ldots,m.
  \end{aligned}
  \label{eq:minimax-epigraph}
\end{equation}
Here $t$ is a common upper bound, and at an attained optimum
$t^\star=\max_i f_i(x^\star)$. This minimax form is appropriate when the worst
error or most adverse operating condition matters more than average
performance. If every $f_i$ is convex, all the new inequalities are convex.

\begin{example}[Largest absolute fitting error]
The worst absolute residual problem
\begin{equation}
  \min_x\max_i |a_i\trans x-b_i|
\end{equation}
is nonsmooth, but $|r|\leq t$ is equivalent to $r\leq t$ and $-r\leq t$.
Therefore the problem is exactly the LP
\begin{equation}
  \begin{aligned}
    \minimize_{x,t}\quad & t\\
    \subjectto\quad
      & a_i\trans x-b_i\leq t,\\
      & -a_i\trans x+b_i\leq t,
        \quad i=1,\ldots,m.
  \end{aligned}
\end{equation}
The auxiliary variable records the largest residual, while the two inequalities
for each observation convert the absolute values and maximum into linear
constraints.
\end{example}

\subsection{Composition Rules and Common Checks}

The following rules cover most modeling decisions in this book:

\begin{itemize}
  \item nonnegative sums and pointwise maxima preserve convexity;
  \item composing a convex function with an affine map preserves convexity;
  \item intersections of convex constraints preserve convexity;
  \item constraints $f(x)\leq0$ are convex when $f$ is convex;
  \item nonlinear equality constraints are generally not convex.
\end{itemize}

Absolute values and norms can often be converted into standard constraints. For
example, $|a\trans x-b|\leq t$ is equivalent to
\begin{equation}
  a\trans x-b\leq t,
  \qquad
  -a\trans x+b\leq t.
\end{equation}
The goal is not to memorize a large catalog, but to recognize a small set of
reusable convex building blocks.

\begin{exercise}
Show directly that intersections and affine images of convex sets are convex.
\end{exercise}
\begin{proof}[Short solution]
If $x,y\in\bigcap_j C_j$, then $x,y\in C_j$ for every $j$, so
$\theta x+(1-\theta)y\in C_j$ for every $j$ and hence belongs to the
intersection. If $u=T(x)$ and $v=T(y)$ for an affine map $T$, then
$\theta u+(1-\theta)v=T(\theta x+(1-\theta)y)$, which belongs to $T(C)$ when
$C$ is convex.
\end{proof}

\begin{exercise}
Use \cref{eq:first-order-convexity} to prove that every local minimizer of a
convex function is global.
\end{exercise}
\begin{proof}[Short solution]
If a local minimizer $x^\star$ were not global, some $y$ would satisfy
$f(y)<f(x^\star)$. Points $x_\theta=(1-\theta)x^\star+\theta y$ are arbitrarily
close to $x^\star$ for small $\theta>0$, and convexity gives
$f(x_\theta)<f(x^\star)$, a contradiction.
\end{proof}

\begin{exercise}
Reformulate $\min_x\norm{Ax-b}_\infty$ as a linear program.
\end{exercise}
\begin{proof}[Short solution]
Since $\norm{r}_\infty\leq t$ exactly when $-t\leq r_i\leq t$ for every
coordinate, the equivalent LP is
\begin{equation}
  \begin{aligned}
    \minimize_{x,t}\quad & t\\
    \subjectto\quad & Ax-b\leq t\mathbf 1,\\
                    & -Ax+b\leq t\mathbf 1.
  \end{aligned}
\end{equation}
\end{proof}

\begin{exercise}
Find the convex hull of
$S=\{(0,0),(1,0),(0,1)\}\subset\R^2$ and express it using linear
inequalities.
\end{exercise}
\begin{proof}[Short solution]
A convex combination has the form $(x_1,x_2)=(\theta_2,\theta_3)$ with
$\theta_i\geq0$ and $\theta_1+\theta_2+\theta_3=1$. Therefore
\begin{equation}
  \operatorname{conv}(S)
  =\{x:x_1\geq0,\ x_2\geq0,\ x_1+x_2\leq1\},
\end{equation}
the filled triangle with vertices in $S$.
\end{proof}

\begin{exercise}
Prove directly from the triangle inequality that every norm is a convex
function.
\end{exercise}
\begin{proof}[Short solution]
For $\theta\in[0,1]$,
\begin{equation}
  \norm{\theta x+(1-\theta)y}
  \leq\theta\norm{x}+(1-\theta)\norm{y},
\end{equation}
using the triangle inequality and absolute homogeneity. This is precisely the
convexity inequality.
\end{proof}

\begin{exercise}
Let $f(x)=\tfrac12x\trans Qx+c\trans x+r$ with $Q=Q\trans$. State conditions
on $Q$ under which $f$ is convex, strictly convex, and strongly convex.
\end{exercise}
\begin{proof}[Short solution]
Because $\nabla^2f=Q$, the function is convex exactly when $Q\succeq0$. It is
strictly convex when $Q\succ0$. If $\lambda_{\min}(Q)>0$, it is
$\mu$-strongly convex for every $0<\mu\leq\lambda_{\min}(Q)$.
\end{proof}

\begin{exercise}
For $f(x_1,x_2)=\tfrac12(x_1^2+10x_2^2)$, find valid strong-convexity and
smoothness constants and compute the condition number $\kappa=L/\mu$.
\end{exercise}
\begin{proof}[Short solution]
The Hessian is $\operatorname{diag}(1,10)$. Its smallest and largest eigenvalues
are $1$ and $10$, so one may take $\mu=1$ and $L=10$. Hence $\kappa=10$.
\end{proof}

\begin{exercise}
Compare the sets
\begin{equation}
  C_1=\{x\in\R:x^2\leq1\},
  \qquad
  C_2=\{x\in\R:x^2=1\}.
\end{equation}
Which is convex, and what does this illustrate about nonlinear equalities?
\end{exercise}
\begin{proof}[Short solution]
$C_1=[-1,1]$ is convex. By contrast, $C_2=\{-1,1\}$ is not convex because its
midpoint $0$ is absent. A convex function may define a convex inequality
sublevel set, while its nonlinear equality set need not be convex.
\end{proof}

\begin{exercise}
Reformulate
\begin{equation}
  \min_x\max_{1\leq i\leq m}(a_i\trans x+b_i)
\end{equation}
as a linear program and explain why the reformulation is exact.
\end{exercise}
\begin{proof}[Short solution]
Introduce a common upper bound $t$:
\begin{equation}
  \begin{aligned}
    \minimize_{x,t}\quad & t\\
    \subjectto\quad & a_i\trans x+b_i\leq t,
      \quad i=1,\ldots,m.
  \end{aligned}
\end{equation}
For fixed $x$, the smallest feasible $t$ equals the largest affine value, so the
optimal values and optimal choices of $x$ coincide.
\end{proof}

\begin{exercise}
Reformulate
\begin{equation}
  \begin{aligned}
    \minimize_x\quad & c\trans x+\gamma\norm{Ax-b}_2\\
    \subjectto\quad & Gx\leq h,
  \end{aligned}
  \qquad \gamma>0,
\end{equation}
as an SOCP. Why would replacing the norm by its square not be equivalent?
\end{exercise}
\begin{proof}[Short solution]
The equivalent SOCP is
\begin{equation}
  \begin{aligned}
    \minimize_{x,t}\quad & c\trans x+\gamma t\\
    \subjectto\quad & \norm{Ax-b}_2\leq t,\\
                    & Gx\leq h.
  \end{aligned}
\end{equation}
Squaring changes the relative penalty: a residual of size $r$ contributes $r$
in the original model but $r^2$ in the modified one, so the tradeoff with the
linear cost generally changes.
\end{proof}

\paragraph{Chapter summary.}
Convexity makes local models globally informative. Convex sets remain intact
under line segments, convex functions lie below their chords, and first-order
conditions can certify global solutions. LPs, QPs, and norm-constrained problems
provide the main engineering problem classes needed in the chapters that
follow.
\chapter{Constrained Optimization and KKT Conditions}
\chapteroverview{Feasibility, Lagrange multipliers, duality, KKT conditions, and their engineering interpretation.}

Consider
\begin{equation}
  \begin{aligned}
    \minimize_{x\in\R^n}\quad & f(x)\\
    \subjectto\quad & g_i(x)\leq0,
      \quad i=1,\ldots,m,\\
                    & h_j(x)=0,
      \quad j=1,\ldots,p.
  \end{aligned}
  \label{eq:general-constrained-problem}
\end{equation}
Constraints may represent actuator limits, safety requirements, conservation
laws, or finite resources. At a constrained optimum, the objective gradient
usually does not vanish; it is balanced by the boundaries that prevent further
improvement.

\section{Feasibility and Local Geometry}

\subsection{Feasible and Active Constraints}

The feasible set is
\begin{equation}
  \mathcal F=\{x:g_i(x)\leq0,\ h_j(x)=0\}.
\end{equation}
An inequality is active at a feasible point $x$ when $g_i(x)=0$ and inactive
when $g_i(x)<0$. Only active inequalities restrict sufficiently small motions.

For example,
\begin{equation}
  \minimize_x (x-2)^2
  \qquad\subjectto\qquad x\leq1
\end{equation}
has solution $x^\star=1$. The derivative is not zero there: the objective would
decrease by moving right, but the active constraint forbids that motion.

\subsection{Linearized Feasible Directions}

At a feasible point, a direction $d$ is compatible with the constraints to
first order when
\begin{align}
  \nabla g_i(x)\trans d&\leq0,
    &&i\in\mathcal A(x),\\
  \nabla h_j(x)\trans d&=0,
    &&j=1,\ldots,p,
  \label{eq:linearized-feasible-directions}
\end{align}
where $\mathcal A(x)=\{i:g_i(x)=0\}$. Equality gradients remove motion normal
to the equality surface; active inequality gradients allow motion only toward
the feasible side.

At a local minimizer $x^\star$, no such feasible direction can decrease the
objective:
\begin{equation}
  \nabla f(x^\star)\trans d\geq0.
  \label{eq:constrained-directional-condition}
\end{equation}
The directional condition in \cref{eq:constrained-directional-condition} is
geometrically useful, but we would also like an algebraic condition that can be
checked by solving equations. To derive one, we need a mild regularity
assumption on the constraints. The idea is simply that the active constraint
boundaries should meet in a nondegenerate way: their normal vectors should
provide genuinely distinct information rather than being redundant. The next
section uses this assumption to introduce Lagrange multipliers and express the
required balance of gradients.

A common assumption is the \emph{linear independence constraint
qualification} (LICQ). It requires the vectors
\begin{equation}
  \{\nabla g_i(x):i\in\mathcal A(x)\}
  \cup
  \{\nabla h_j(x):j=1,\ldots,p\}
\end{equation}
to be linearly independent. In practical terms, no active constraint normal can
be reconstructed as a linear combination of the others. Under LICQ, the local
feasible geometry is regular enough for the KKT multiplier description to
apply cleanly.

LICQ is a sufficient technical condition, not a claim that problems violating
it have no solutions. An optimum may still exist when LICQ fails; the usual KKT
theorem may simply require a different constraint qualification, and the
multipliers may be nonunique or harder to interpret.

\section{The Lagrangian and Duality}

\subsection{Equality Constraints First}

The main idea behind Lagrange multipliers is easiest to see when there are only
equality constraints. Consider
\begin{equation}
  \begin{aligned}
    \minimize_x\quad & f(x)\\
    \subjectto\quad & h_j(x)=0,
      \quad j=1,\ldots,p.
  \end{aligned}
  \label{eq:equality-constrained-problem}
\end{equation}
Near a feasible point, the first-order feasible directions satisfy
\begin{equation}
  \nabla h_j(x)\trans d=0,
  \qquad j=1,\ldots,p.
\end{equation}
Thus feasible directions are tangent to the constraint surfaces, while the
gradients $\nabla h_j(x)$ are normal to those surfaces.

At a regular local minimizer $x^\star$, the objective cannot decrease along any
of these tangent directions. Consequently, $\nabla f(x^\star)$ must lie in the
span of the constraint normals. There therefore exist scalars
$\nu_1^\star,\ldots,\nu_p^\star$ such that
\begin{equation}
  \nabla f(x^\star)
  +\sum_{j=1}^p\nu_j^\star\nabla h_j(x^\star)=0.
  \label{eq:equality-gradient-balance}
\end{equation}
The scalars $\nu_j^\star$ are called \emph{Lagrange multipliers}. They are
unrestricted in sign because an equality constraint prevents motion through
its surface in either normal direction.

It is convenient to collect the objective and constraints into the
\emph{equality-constrained Lagrangian}
\begin{equation}
  \mathcal L(x,\nu)
  =f(x)+\sum_{j=1}^p\nu_j h_j(x).
\end{equation}
Then \cref{eq:equality-gradient-balance} is simply
\begin{equation}
  \nabla_x\mathcal L(x^\star,\nu^\star)=0.
\end{equation}
It must be combined with the original feasibility equations
$h_j(x^\star)=0$. The unknowns are now both the optimizer $x^\star$ and the
multipliers $\nu^\star$.

\begin{example}[Projection onto a hyperplane]
Let $a,c\in\R^n$ with $c\neq0$. The problem
\begin{equation}
  \begin{aligned}
    \minimize_x\quad & \frac12\norm{x-a}_2^2\\
    \subjectto\quad & c\trans x=b
  \end{aligned}
\end{equation}
asks for the point on the hyperplane $c\trans x=b$ closest to $a$. Its
Lagrangian is
\begin{equation}
  \mathcal L(x,\nu)
  =\frac12\norm{x-a}_2^2+\nu(c\trans x-b).
\end{equation}
Stationarity and feasibility give
\begin{equation}
  x-a+\nu c=0,
  \qquad c\trans x=b.
\end{equation}
Hence
\begin{equation}
  \nu^\star=\frac{c\trans a-b}{\norm{c}_2^2},
  \qquad
  x^\star
  =a-\frac{c\trans a-b}{\norm{c}_2^2}c.
\end{equation}
The correction from $a$ to $x^\star$ is parallel to the normal vector $c$,
exactly as the gradient-balance picture predicts.

The multiplier also has a useful quantitative interpretation. From
$x^\star-a=-\nu^\star c$, the value of $\nu^\star$ gives the signed strength of
the constraint normal needed to balance the objective gradient. Its sign tells
which direction along $c$ the constraint pushes the solution, while
$|\nu^\star|\norm{c}_2$ is the magnitude of that balancing vector.

It also measures sensitivity to the constraint level. Let $p(b)$ be the optimal
value when the hyperplane is $c\trans x=b$. For this example,
\begin{equation}
  p(b)=\frac{(c\trans a-b)^2}{2\norm{c}_2^2},
  \qquad
  \frac{d p}{d b}=-\nu^\star.
\end{equation}
Thus, for a small change $\Delta b$,
\begin{equation}
  p(b+\Delta b)-p(b)
  \approx -\nu^\star\Delta b.
\end{equation}
A large multiplier in magnitude means that moving the constraint surface even
slightly has a large first-order effect on the best achievable objective value.
This sensitivity interpretation extends to general equality constraints and
will reappear later as the shadow-price interpretation of multipliers.
\end{example}

\subsection{Adding Inequality Constraints}

Now return to the general problem in
\cref{eq:general-constrained-problem}. An inequality $g_i(x)\leq0$ describes an
allowed side and a forbidden side of a boundary. If $g_i(x)<0$, the point lies
strictly inside the allowed region. A sufficiently small move in any direction
will not cross that boundary, so the constraint has no immediate effect. Its
multiplier should therefore be zero.

If $g_i(x)=0$, the point lies on the boundary. The constraint now prevents
motion toward the forbidden side, although motion back into the allowed region
is still possible. For example, the constraint $x\leq1$ does nothing locally
when $x=0$, but at $x=1$ it prevents any motion to the right. Because an
inequality blocks motion in only one direction, the sign of its multiplier
matters.

With our convention $g_i(x)\leq0$, the gradient $\nabla g_i(x)$ points toward
increasing values of $g_i$ and hence toward the forbidden side of an active
boundary. At an optimum with one active inequality, gradient balance has the
form
\begin{equation}
  -\nabla f(x^\star)=\lambda_i^\star\nabla g_i(x^\star).
\end{equation}
The left-hand side is the unconstrained downhill direction. If the constraint
is what prevents that downhill motion, this direction must point toward the
forbidden side, in the same direction as $\nabla g_i(x^\star)$. Its coefficient
$\lambda_i^\star$ must therefore be nonnegative. A negative coefficient would
point toward the allowed side, indicating an available feasible descent
direction rather than a constrained minimum.

For example, write $x\leq1$ as $g(x)=x-1\leq0$. Then $g'(x)=1$ points to the
right, toward the forbidden region $x>1$. For
$f(x)=(x-2)^2$ at $x^\star=1$,
\begin{equation}
  -f'(1)=2=\lambda^\star g'(1),
\end{equation}
so $\lambda^\star=2\geq0$. The positive multiplier records that the desired
downhill motion is precisely the rightward motion blocked by the constraint.

The sign rule depends on notation. If the same constraint were written using
the convention $g_i(x)\geq0$, the associated multiplier would instead be
nonpositive when the Lagrangian is written with the same plus sign. Throughout
this book we use $g_i(x)\leq0$ and therefore $\lambda_i\geq0$. Equality
multipliers $\nu_j$ remain unrestricted because equality constraints block both
normal directions.

\begin{definition}[Lagrangian]
The Lagrangian of the general constrained problem is
\begin{equation}
  \mathcal L(x,\lambda,\nu)
  =f(x)+\sum_{i=1}^m\lambda_i g_i(x)
       +\sum_{j=1}^p\nu_j h_j(x).
  \label{eq:lagrangian}
\end{equation}
\end{definition}

The Lagrangian places the objective and all constraint-normal terms in one
expression. At a regular optimum, suitable multipliers satisfy
\begin{equation}
  \nabla_x\mathcal L(x^\star,\lambda^\star,\nu^\star)
  =\nabla f(x^\star)
   +\sum_i\lambda_i^\star\nabla g_i(x^\star)
   +\sum_j\nu_j^\star\nabla h_j(x^\star)
  =0.
  \label{eq:gradient-balance}
\end{equation}
This is the \emph{stationarity condition}. It says that the objective gradient
is canceled by constraint-normal terms, so no feasible first-order motion can
improve the objective.

\paragraph{Important multiplier rules.}
The distinction between equality and inequality multipliers is fundamental:
\begin{equation}
  \boxed{
  \begin{aligned}
    h_j(x^\star)=0
      &\quad\Longrightarrow\quad \nu_j^\star\in\R,
      &&\text{equality multiplier: either sign},\\
    g_i(x^\star)<0
      &\quad\Longrightarrow\quad \lambda_i^\star=0,
      &&\text{inactive inequality: no influence},\\
    g_i(x^\star)=0
      &\quad\Longrightarrow\quad \lambda_i^\star\geq0,
      &&\text{active inequality: possibly positive}.
  \end{aligned}}
  \label{eq:multiplier-rules}
\end{equation}
An active inequality may still have a zero multiplier; activity means that the
point is on the boundary, not necessarily that the boundary affects the
objective to first order. Stationarity, these multiplier rules, and the
original feasibility requirements lead directly to the KKT conditions in the
next section.

\subsection{Dual Bounds}

So far, the Lagrangian has been used to describe first-order balance at a
candidate optimum. It has a second role that is especially valuable: it can
produce a guaranteed lower bound on the best feasible objective value. Such a
bound tells us how well any feasible point could possibly perform and can
therefore help certify optimality.

To see the idea, take any feasible point $x$. Its equality constraints satisfy
$h_j(x)=0$, so the terms $\nu_jh_j(x)$ vanish. Its inequality constraints
satisfy $g_i(x)\leq0$. If $\lambda_i\geq0$, then
\begin{equation}
  \lambda_i g_i(x)\leq0.
\end{equation}
Adding these nonpositive terms to the objective can only lower its value:
\begin{equation}
  \mathcal L(x,\lambda,\nu)\leq f(x)
  \qquad\text{for every feasible }x
  \text{ and every }\lambda\geq0.
  \label{eq:lagrangian-below-feasible-objective}
\end{equation}
This is another reason the sign restriction $\lambda\geq0$ is important.

Now fix some multipliers $(\lambda,\nu)$ and minimize the Lagrangian over all
$x$, without enforcing the original constraints. This defines the
\emph{dual function}
\begin{equation}
  q(\lambda,\nu)
  =\inf_x\mathcal L(x,\lambda,\nu)
  \label{eq:dual-function}
\end{equation}
Because an infimum is no larger than the value at any particular $x$, for every
feasible $x$ we have
\begin{equation}
  q(\lambda,\nu)
  \leq \mathcal L(x,\lambda,\nu)
  \leq f(x).
\end{equation}
Thus $q(\lambda,\nu)$ is a lower bound on the objective value of \emph{every}
feasible point, and therefore on the primal optimal value $p^\star$. Some
choices of multipliers give weak bounds, or even $-\infty$, so we seek the
largest valid bound. This gives the \emph{dual problem}
\begin{equation}
  \maximize_{\lambda\geq0,\nu} q(\lambda,\nu).
\end{equation}
If $d^\star$ denotes its optimal value, then
\begin{equation}
  d^\star\leq p^\star.
\end{equation}
This inequality is called \emph{weak duality}. The difference
$p^\star-d^\star\geq0$ is the \emph{duality gap}. If the gap is zero, then
$d^\star=p^\star$ and \emph{strong duality} holds. In that case, a feasible
primal point and a set of dual multipliers attaining the same value certify one
another as optimal.

Strong duality does not hold for every optimization problem. For a convex
problem, however, it is guaranteed under a common strict-feasibility assumption
called \emph{Slater's condition}: there exists a point satisfying the equality
constraints exactly and the inequality constraints strictly, $g_i(x)<0$.
Under this condition, the multiplier description loses no information about
the primal optimum. The main practical consequence is that the KKT conditions
in the next section become a complete certificate of global optimality.

\section{Karush--Kuhn--Tucker Conditions}

\subsection{The Four Conditions}

The Karush--Kuhn--Tucker (KKT) conditions are
\begin{align}
  \nabla_x\mathcal L(x^\star,\lambda^\star,\nu^\star)&=0,
    &&\text{stationarity},\\
  g_i(x^\star)&\leq0,\quad h_j(x^\star)=0,
    &&\text{primal feasibility},\\
  \lambda_i^\star&\geq0,
    &&\text{dual feasibility},\\
  \lambda_i^\star g_i(x^\star)&=0,
    &&\text{complementary slackness}.
  \label{eq:kkt-conditions}
\end{align}

Complementary slackness connects activity and influence. If a constraint is
inactive, its multiplier must be zero. If its multiplier is positive, the
constraint must be active. An active constraint may still have zero multiplier
when it does not affect the optimum to first order.

\begin{theorem}[KKT necessity and global sufficiency]
The KKT conditions have two distinct roles.
\begin{enumerate}
  \item For a smooth problem satisfying LICQ, every local minimizer has
  multipliers that satisfy the KKT conditions.

  \item For a convex problem---a convex objective, convex inequality
  functions, and affine equality constraints---every point satisfying the KKT
  conditions is a global minimizer. If the objective is strictly convex on the
  feasible set, that minimizer is unique.
\end{enumerate}
\end{theorem}

The second statement is the central payoff. KKT is not merely a test for a
possible local solution in a convex problem; it is a certificate that no
feasible point anywhere has a smaller objective value.

To see why, suppose $(x^\star,\lambda^\star,\nu^\star)$ satisfies KKT. Since
$\lambda^\star\geq0$, the function
$\mathcal L(\,\cdot\,,\lambda^\star,\nu^\star)$ is convex in $x$.
Stationarity therefore makes $x^\star$ a global minimizer of this Lagrangian:
\begin{equation}
  \mathcal L(x^\star,\lambda^\star,\nu^\star)
  \leq
  \mathcal L(y,\lambda^\star,\nu^\star)
  \qquad\text{for every }y.
\end{equation}
Primal feasibility and complementary slackness give
\begin{equation}
  \mathcal L(x^\star,\lambda^\star,\nu^\star)=f(x^\star).
\end{equation}
For any other feasible point $y$, the dual-bound argument gives
\begin{equation}
  \mathcal L(y,\lambda^\star,\nu^\star)\leq f(y).
\end{equation}
Combining the three relations yields
\begin{equation}
  \boxed{f(x^\star)\leq f(y)
  \qquad\text{for every feasible }y,}
\end{equation}
which is precisely global optimality. This conclusion does not hold in general
for nonconvex problems; there, a KKT point may instead be a local maximum or a
saddle point.

\subsection{A Quadratic Example}

Consider
\begin{equation}
  \begin{aligned}
    \minimize_x\quad & \frac12x\trans Qx+c\trans x\\
    \subjectto\quad & Ax=b,
  \end{aligned}
  \qquad Q\succ0.
\end{equation}
The Lagrangian is
$\mathcal L(x,\nu)=\tfrac12x\trans Qx+c\trans x+\nu\trans(Ax-b)$.
Stationarity and feasibility give the linear KKT system
\begin{equation}
  \begin{bmatrix}
    Q & A\trans\\
    A & 0
  \end{bmatrix}
  \begin{bmatrix}
    x^\star\\ \nu^\star
  \end{bmatrix}
  =
  \begin{bmatrix}
    -c\\ b
  \end{bmatrix}.
  \label{eq:equality-qp-kkt-system}
\end{equation}
This system is the basic algebraic pattern behind many constrained least-squares
and control computations.

\paragraph{Deriving the Newton step.}
The KKT conditions can be viewed as a system of equations in the joint
primal--dual variable
\begin{equation}
  w=
  \begin{bmatrix}x\\ \nu\end{bmatrix}.
\end{equation}
For this quadratic problem, define the KKT residual
\begin{equation}
  F(w)
  =
  \begin{bmatrix}
    Qx+c+A\trans\nu\\
    Ax-b
  \end{bmatrix}.
  \label{eq:equality-qp-kkt-residual}
\end{equation}
The first block is the stationarity error, and the second block is the
feasibility error. Solving the constrained problem is therefore equivalent to
finding $w^\star$ such that $F(w^\star)=0$.

More generally, Newton's method solves a nonlinear system $F(w)=0$ by
linearizing it at the current estimate $w_k$:
\begin{equation}
  F(w_k+\Delta w_k)
  \approx F(w_k)+F_w(w_k)\Delta w_k.
  \label{eq:newton-system-linearization}
\end{equation}
It chooses the correction $\Delta w_k$ so that the right-hand side is zero,
which gives the Newton equation and update
\begin{equation}
  F_w(w_k)\Delta w_k=-F(w_k),
  \qquad
  w_{k+1}=w_k+\Delta w_k.
  \label{eq:newton-system-update}
\end{equation}
In computation, the linear system is solved directly; the inverse of
$F_w(w_k)$ is not formed.

For the quadratic KKT residual in
\cref{eq:equality-qp-kkt-residual}, the Jacobian is constant:
\begin{equation}
  F_w(w_k)
  =
  \begin{bmatrix}
    Q&A\trans\\
    A&0
  \end{bmatrix}.
\end{equation}
Writing
$\Delta w_k=(\Delta x_k,\Delta\nu_k)$, the Newton step is therefore
\begin{equation}
  \begin{bmatrix}
    Q&A\trans\\
    A&0
  \end{bmatrix}
  \begin{bmatrix}
    \Delta x_k\\ \Delta\nu_k
  \end{bmatrix}
  =-
  \begin{bmatrix}
    Qx_k+c+A\trans\nu_k\\
    Ax_k-b
  \end{bmatrix}.
  \label{eq:equality-qp-newton-step}
\end{equation}
After adding the current iterate to the correction, the new variables satisfy
\begin{align}
  \begin{bmatrix}
    Q&A\trans\\
    A&0
  \end{bmatrix}
  \begin{bmatrix}
    x_{k+1}\\ \nu_{k+1}
  \end{bmatrix}
  &=
  \begin{bmatrix}
    Q&A\trans\\
    A&0
  \end{bmatrix}
  \left(
    \begin{bmatrix}x_k\\ \nu_k\end{bmatrix}
    +
    \begin{bmatrix}\Delta x_k\\ \Delta\nu_k\end{bmatrix}
  \right)\\
  &=
  \begin{bmatrix}-c\\ b\end{bmatrix}.
\end{align}
Thus one full Newton step reaches the exact solution of the linear KKT system
from any starting point, provided the KKT matrix is nonsingular. For nonlinear
objectives or constraints, SQP instead repeatedly constructs a local QP whose
coefficients $Q_k$, $c_k$, and $A_k$ are evaluated at the current point
$x_k$, solves it for a step, updates $x_k$, and rebuilds the QP at the new
point; Chapter~4 develops this procedure in detail.

\subsection{A Quadratic Example with an Inequality}

Consider the general strictly convex QP
\begin{equation}
  \begin{aligned}
    \minimize_x\quad &\frac12x\trans Qx+c\trans x\\
    \subjectto\quad &Ax\leq b,
  \end{aligned}
  \qquad Q\succ0.
  \label{eq:general-inequality-qp}
\end{equation}
With multipliers $\lambda\geq0$, its KKT conditions are
\begin{align}
  Qx+c+A\trans\lambda&=0,
  \label{eq:general-inequality-qp-stationarity}\\
  Ax-b&\leq0,
  \label{eq:general-inequality-qp-primal-feasibility}\\
  \lambda&\geq0,
  \label{eq:general-inequality-qp-dual-feasibility}\\
  \lambda_i(Ax-b)_i&=0,
  \qquad i=1,\ldots,m.
  \label{eq:general-inequality-qp-complementarity}
\end{align}
Unlike the equality-constrained system in
\cref{eq:equality-qp-kkt-system}, these conditions are not immediately one
linear system because complementary slackness must determine the active
constraints. For each $i$, either $(Ax-b)_i<0$ and $\lambda_i=0$, or the
constraint is active, $(Ax-b)_i=0$, and its multiplier may be positive.

The basic idea is therefore simple: choose a candidate active set
$\mathcal A$, treat its inequalities as equalities, set the other multipliers
to zero, and solve the resulting equality-constrained problem. For a hand
calculation with only a few inequalities, one can test the possible subsets,
often beginning with the unconstrained minimizer to see which constraints it
violates. An active-set algorithm performs this search systematically: it
maintains a working set, adds a constraint when the current step would violate
it, and removes an active constraint when its computed multiplier has the
wrong sign. For any proposed set $\mathcal A$, it solves
\begin{equation}
  \begin{bmatrix}
    Q&A_{\mathcal A}\trans\\
    A_{\mathcal A}&0
  \end{bmatrix}
  \begin{bmatrix}
    x\\\lambda_{\mathcal A}
  \end{bmatrix}
  =
  \begin{bmatrix}
    -c\\b_{\mathcal A}
  \end{bmatrix}.
  \label{eq:inequality-qp-candidate-active-system}
\end{equation}
The candidate is valid only if all inequalities are satisfied and
$\lambda_{\mathcal A}\geq0$. For a small problem, testing the few possible
active sets is often the simplest solution method.

To see this procedure explicitly, consider
\begin{equation}
  \begin{aligned}
    \minimize_{x_1,x_2}\quad
      &\frac12(x_1-2)^2+\frac12(x_2-1)^2\\
    \subjectto\quad
      &x_1+x_2\leq2.
  \end{aligned}
  \label{eq:inequality-qp-example}
\end{equation}
Write the constraint as
$g(x)=x_1+x_2-2\leq0$ and let $\lambda\geq0$ be its multiplier. The
Lagrangian is
\begin{equation}
  \mathcal L(x_1,x_2,\lambda)
  =\frac12(x_1-2)^2+\frac12(x_2-1)^2
   +\lambda(x_1+x_2-2).
\end{equation}
The KKT conditions are
\begin{align}
  x_1-2+\lambda&=0,
  \label{eq:inequality-qp-stationarity-one}\\
  x_2-1+\lambda&=0,
  \label{eq:inequality-qp-stationarity-two}\\
  x_1+x_2-2&\leq0,
  \label{eq:inequality-qp-primal-feasibility}\\
  \lambda&\geq0,
  \label{eq:inequality-qp-dual-feasibility}\\
  \lambda(x_1+x_2-2)&=0.
  \label{eq:inequality-qp-complementarity}
\end{align}
The first two equations express stationarity, the next two impose primal and
dual feasibility, and the final equation is complementary slackness.

\paragraph{The simplest solution method.}
For a small QP, the most direct method is to test the possible active set. If
the inequality is inactive, complementary slackness gives $\lambda=0$.
Stationarity would then give the unconstrained minimizer $(x_1,x_2)=(2,1)$,
but this point violates $x_1+x_2\leq2$. The inactive case is therefore
impossible.

The constraint must instead be active, so $x_1+x_2=2$. Stationarity and the
active constraint form the linear KKT system
\begin{equation}
  \begin{bmatrix}
    1&0&1\\
    0&1&1\\
    1&1&0
  \end{bmatrix}
  \begin{bmatrix}
    x_1\\x_2\\\lambda
  \end{bmatrix}
  =
  \begin{bmatrix}
    2\\1\\2
  \end{bmatrix}.
  \label{eq:inequality-qp-active-kkt-system}
\end{equation}
Solving gives
\begin{equation}
  x^\star=\left(\frac32,\frac12\right),
  \qquad
  \lambda^\star=\frac12.
\end{equation}
The multiplier is nonnegative, so the assumed active set is consistent. Since
the objective is strictly convex and the constraint is affine, these KKT
conditions certify the unique global minimizer.

For many inequalities, testing every active set is too expensive. Active-set
QP algorithms automate the same logic by adding or removing constraints as
needed. Chapter~4 introduces scalable alternatives, including interior-point
and operator-splitting methods, that do not explicitly enumerate active sets.

\section{Sensitivity and Engineering Interpretation}

\subsection{Why Sensitivity Matters}

An optimizer tells us what is best for the current model. In engineering,
however, the model rarely remains fixed. A power limit may be increased, a
safety margin may be tightened, a reference value may change, or an uncertain
parameter may be estimated slightly differently. After finding an optimum, we
therefore also want to know:
\begin{itemize}
  \item Which constraints are genuine bottlenecks?
  \item How much would performance improve if one limit were relaxed?
  \item Which changes would have almost no effect?
  \item Is the reported solution stable under small modeling errors?
\end{itemize}

One could answer these questions by changing each parameter and solving the
optimization problem again. Sensitivity analysis instead uses information
already produced at the original solution to predict the effect of small
changes. Lagrange multipliers provide exactly this information.

\subsection{Multipliers as Shadow Prices}

To measure the effect of changing a constraint, perturb its right-hand side:
\begin{equation}
  \begin{aligned}
    \minimize_x\quad & f(x)\\
    \subjectto\quad & g_i(x)\leq u_i,
      \quad i=1,\ldots,m,\\
                    & h_j(x)=v_j,
      \quad j=1,\ldots,p.
  \end{aligned}
\end{equation}
Let $p(u,v)$ denote the optimal objective value of this perturbed problem. The
original problem corresponds to $u=0$ and $v=0$. Positive $u_i$ relaxes the
$i$th inequality, while $v_j$ moves the $j$th equality surface.

The central sensitivity result is
\begin{equation}
  \boxed{
  \frac{\partial p}{\partial u_i}(0,0)=-\lambda_i^\star,
  \qquad
  \frac{\partial p}{\partial v_j}(0,0)=-\nu_j^\star.}
  \label{eq:multiplier-sensitivity}
\end{equation}
Thus a multiplier is not merely an auxiliary variable used in the KKT
equations. It is the rate at which the best achievable objective value changes
when the corresponding constraint level moves. We first derive this statement
for one equality, where the calculation is easiest to see.

\paragraph{Derivation for one equality constraint.}
Consider
\begin{equation}
  p(v)=\min_x f(x)
  \qquad\subjectto\qquad h(x)=v,
\end{equation}
and write $x^\star(v)$ for its optimizer. Since
$p(v)=f(x^\star(v))$, the chain rule gives
\begin{equation}
  \frac{d p}{d v}
  =\nabla f(x^\star)\trans
    \frac{d x^\star}{d v}.
  \label{eq:value-chain-rule}
\end{equation}
This says that the optimal value changes because the optimizer moves.

Two equations describe that motion. First, stationarity gives
\begin{equation}
  \nabla f(x^\star)
  =-\nu^\star\nabla h(x^\star).
  \label{eq:sensitivity-stationarity}
\end{equation}
Second, the moving optimizer must remain feasible:
$h(x^\star(v))=v$. Differentiating this identity with respect to $v$ gives
\begin{equation}
  \nabla h(x^\star)\trans
  \frac{d x^\star}{d v}=1.
  \label{eq:sensitivity-feasibility}
\end{equation}
Substituting
\cref{eq:sensitivity-stationarity,eq:sensitivity-feasibility} into
\cref{eq:value-chain-rule} yields
\begin{equation}
  \frac{d p}{d v}
  =-\nu^\star
    \nabla h(x^\star)\trans\frac{d x^\star}{d v}
  =-\nu^\star.
\end{equation}
Thus the multiplier summarizes the effect of the optimizer's motion; we do not
need to calculate $d x^\star/dv$ explicitly.

\paragraph{Extension to an active inequality.}
Suppose $g(x)\leq u$ remains active as $u$ changes slightly. Then
$g(x^\star(u))=u$, and differentiation gives
\begin{equation}
  \nabla g(x^\star)\trans\frac{d x^\star}{d u}=1.
\end{equation}
Stationarity gives
$\nabla f(x^\star)=-\lambda^\star\nabla g(x^\star)$, and therefore
\begin{equation}
  \frac{d p}{d u}
  =\nabla f(x^\star)\trans\frac{d x^\star}{d u}
  =-\lambda^\star.
\end{equation}
If the inequality is inactive, then $\lambda^\star=0$ and a sufficiently small
relaxation does not move that boundary into contact with the solution, so the
optimal value has zero first-order change.

\paragraph{Several constraints.}
With several constraints, the same argument is applied to one right-hand side
at a time while the others are held fixed. The perturbed Lagrangian
\begin{equation}
  \mathcal L(x,\lambda,\nu;u,v)
  =f(x)
  +\sum_i\lambda_i\bigl(g_i(x)-u_i\bigr)
  +\sum_j\nu_j\bigl(h_j(x)-v_j\bigr)
  \label{eq:perturbed-lagrangian}
\end{equation}
also makes the result visible: $u_i$ and $v_j$ appear with coefficients
$-\lambda_i$ and $-\nu_j$. Under suitable regularity and differentiability
assumptions, the motion of the optimizer is already accounted for by
stationarity, leading to \cref{eq:multiplier-sensitivity}. This general
principle is often called the \emph{envelope theorem}.

\paragraph{First-order prediction.}
The sensitivity formula gives the local approximation
\begin{equation}
  p(u,v)
  \approx p(0,0)
  -\sum_i\lambda_i^\star u_i
  -\sum_j\nu_j^\star v_j.
  \label{eq:optimal-value-linearization}
\end{equation}
For a change affecting only the $i$th inequality, this becomes
\begin{equation}
  p(u_i)-p(0)
  \approx -\lambda_i^\star u_i.
\end{equation}
Here $p(u,v)$ is evaluated \emph{after reoptimizing} for the changed
constraints. The formula predicts the change in the best achievable
performance, not the performance of the old solution under the new model.

\paragraph{Shadow-price interpretation.}
The inequality multiplier $\lambda_i^\star$ is called a \emph{shadow price}.
It measures the value of one additional unit of the constrained resource. Since
$\lambda_i^\star\geq0$, relaxing an inequality has predicted change
$-\lambda_i^\star u_i\leq0$: the best achievable objective can improve or stay
unchanged, but it cannot become worse. A large multiplier identifies an
expensive bottleneck. A zero multiplier says that a small relaxation has no
first-order benefit.

The units make this interpretation concrete. If the objective is measured in
dollars and $u_i$ in kilowatts, then $\lambda_i^\star$ has units of dollars per
kilowatt. It estimates the reduction in optimal cost from one additional
kilowatt, provided the change is small enough for the first-order approximation
to remain accurate. Equality multipliers have the same sensitivity meaning,
but may have either sign because moving an equality surface in opposite
directions can either increase or decrease the optimal value.

\begin{example}[The value of relaxing a bound]
Return to
\begin{equation}
  \minimize_x (x-2)^2
  \qquad\subjectto\qquad x\leq1.
\end{equation}
The solution is $x^\star=1$ and the multiplier is $\lambda^\star=2$. Relax the
bound to $x\leq1+u$. For small $u\geq0$, the new solution is $x^\star(u)=1+u$
and
\begin{equation}
  p(u)=(1-u)^2.
\end{equation}
Therefore
\begin{equation}
  p'(0)=-2=-\lambda^\star.
\end{equation}
A small relaxation $u$ is predicted to reduce the optimal cost by approximately
$2u$. For instance, increasing the bound by $0.01$ gives the first-order
estimate $p(0.01)-p(0)\approx-0.02$.
\end{example}

\subsection{Limits of the Interpretation}

Sensitivity formulas are local, first-order approximations. If a parameter
changes substantially, the active set may change: an inactive constraint may
become active, or a formerly limiting constraint may cease to matter. The
optimal-value function can then change slope or fail to be differentiable, and
the original multiplier may no longer predict the change accurately.

Regularity also matters. Redundant or degenerate constraints can produce
nonunique multipliers, making an individual shadow price ambiguous. Poorly
conditioned problems may have solutions and multipliers that change sharply
under small perturbations.

Finally, KKT equations are local first-order conditions for a nonconvex
problem. A KKT point may be a local minimum, a local maximum, or a saddle point,
so its multipliers need not describe the sensitivity of the global optimum.
Second-order checks, comparison of candidate solutions, or a global method may
still be required. In practice, one should inspect feasibility, stationarity,
complementary slackness, and problem convexity rather than reporting an
objective value alone.

\begin{exercise}
Write and solve the KKT conditions for
\begin{equation}
  \minimize_{x_1,x_2}\ (x_1-2)^2+(x_2-1)^2
  \quad\subjectto\quad x_1+x_2\leq2,
  \quad x_1\geq0,\ x_2\geq0.
\end{equation}
\end{exercise}
\begin{proof}[Short solution]
Write the inequalities as
$g_1=x_1+x_2-2\leq0$, $g_2=-x_1\leq0$, and $g_3=-x_2\leq0$.
The solution is the projection of $(2,1)$ onto the feasible triangle:
\begin{equation}
  x^\star=\left(\frac32,\frac12\right).
\end{equation}
Only $g_1$ is active. Stationarity gives
$(-1,-1)+\lambda_1(1,1)=0$, so $\lambda_1^\star=1$, while
$\lambda_2^\star=\lambda_3^\star=0$. Convexity makes this the unique global
minimizer.
\end{proof}

\begin{exercise}
Derive the KKT linear system for minimizing
$\tfrac12\norm{Cx-d}_2^2$ subject to $Ax=b$.
\end{exercise}
\begin{proof}[Short solution]
The Lagrangian is
$\mathcal L=\tfrac12\norm{Cx-d}_2^2+\nu\trans(Ax-b)$.
Stationarity and feasibility give
\begin{equation}
  \begin{bmatrix}
    C\trans C & A\trans\\
    A & 0
  \end{bmatrix}
  \begin{bmatrix}x^\star\\ \nu^\star\end{bmatrix}
  =
  \begin{bmatrix}C\trans d\\ b\end{bmatrix}.
\end{equation}
\end{proof}

\begin{exercise}
For a parameter $a\in\R$, solve
\begin{equation}
  \minimize_x (x-3)^2
  \qquad\subjectto\qquad x\leq a,
\end{equation}
and find the multiplier. What happens when $a=3$?
\end{exercise}
\begin{proof}[Short solution]
If $a<3$, then $x^\star=a$ and stationarity gives
$2(a-3)+\lambda^\star=0$, so
$\lambda^\star=2(3-a)>0$. If $a>3$, then $x^\star=3$ and the constraint is
inactive, so $\lambda^\star=0$. At $a=3$, the constraint is active but
$\lambda^\star=0$, showing that activity does not imply a positive multiplier.
\end{proof}

\begin{exercise}
Consider
\begin{equation}
  \minimize_{x\in\R^2} x_1
  \qquad\subjectto\qquad
  -x_1\leq0,\quad -2x_1\leq0.
\end{equation}
Does LICQ hold at an optimum? Are the KKT multipliers unique?
\end{exercise}
\begin{proof}[Short solution]
Every point with $x_1=0$ is optimal. The active gradients
$(-1,0)\trans$ and $(-2,0)\trans$ are linearly dependent, so LICQ fails.
Stationarity requires
\begin{equation}
  \lambda_1+2\lambda_2=1,
  \qquad \lambda_1,\lambda_2\geq0.
\end{equation}
There are infinitely many multiplier pairs, illustrating the ambiguity caused
by redundant constraints.
\end{proof}

\begin{exercise}
Find all KKT points of
\begin{equation}
  \minimize_{x_1,x_2} x_1
  \qquad\subjectto\qquad x_1^2+x_2^2=1,
\end{equation}
and classify them.
\end{exercise}
\begin{proof}[Short solution]
Stationarity is
$(1,0)\trans+2\nu(x_1,x_2)\trans=0$. Together with the constraint, this gives
$(x_1,x_2)=(-1,0)$ with $\nu=1/2$, and $(1,0)$ with $\nu=-1/2$.
The first point is the global minimum and the second is the global maximum.
Thus KKT points need not be minima for a nonconvex feasible set.
\end{proof}

\begin{exercise}
Derive the dual problem for
\begin{equation}
  \minimize_x x^2
  \qquad\subjectto\qquad x\geq1,
\end{equation}
and verify strong duality.
\end{exercise}
\begin{proof}[Short solution]
Using $g(x)=1-x\leq0$,
\begin{equation}
  \mathcal L(x,\lambda)=x^2+\lambda(1-x).
\end{equation}
Minimizing over $x$ gives $x=\lambda/2$ and
$q(\lambda)=\lambda-\lambda^2/4$. Hence the dual is
$\max_{\lambda\geq0}q(\lambda)$, whose solution is
$\lambda^\star=2$ with value $1$. The primal solution is $x^\star=1$ with the
same value, so the duality gap is zero.
\end{proof}

\begin{exercise}
For
\begin{equation}
  \minimize_x (x-2)^2
  \qquad\subjectto\qquad x\leq1+u,
\end{equation}
compute the optimal-value function near $u=0$ and verify the multiplier
sensitivity formula.
\end{exercise}
\begin{proof}[Short solution]
For $u$ near zero, $x^\star(u)=1+u$ and
\begin{equation}
  p(u)=(1-u)^2.
\end{equation}
Thus $p'(0)=-2$. At $u=0$, stationarity gives
$-2+\lambda^\star=0$, so $\lambda^\star=2$ and
$p'(0)=-\lambda^\star$, as predicted.
\end{proof}

\begin{exercise}
Solve the convex problem
\begin{equation}
  \begin{aligned}
    \minimize_{x_1,x_2}\quad & \frac12(x_1^2+x_2^2)\\
    \subjectto\quad & x_1+x_2\geq1
  \end{aligned}
\end{equation}
using KKT conditions.
\end{exercise}
\begin{proof}[Short solution]
Write $g(x)=1-x_1-x_2\leq0$. Stationarity gives
$x_1-\lambda=0$ and $x_2-\lambda=0$. The constraint must be active, so
$2\lambda=1$. Therefore
\begin{equation}
  x^\star=\left(\frac12,\frac12\right),
  \qquad \lambda^\star=\frac12.
\end{equation}
The objective is strictly convex, so this is the unique global minimizer.
\end{proof}

\begin{exercise}
Solve
\begin{equation}
  \begin{aligned}
    \minimize_{x_1,x_2}\quad & \frac12(x_1^2+2x_2^2)\\
    \subjectto\quad & x_1+x_2=1
  \end{aligned}
\end{equation}
and find the equality multiplier.
\end{exercise}
\begin{proof}[Short solution]
Stationarity gives $x_1+\nu=0$ and $2x_2+\nu=0$. Combining these equations with
$x_1+x_2=1$ yields
\begin{equation}
  x_1^\star=\frac23,\qquad
  x_2^\star=\frac13,\qquad
  \nu^\star=-\frac23.
\end{equation}
Positive definiteness of the objective Hessian gives a unique global minimum.
\end{proof}

\begin{exercise}
Consider
\begin{equation}
  \minimize_x x^2
  \qquad\subjectto\qquad x\geq0.
\end{equation}
Show that the inequality is active at the solution but has zero multiplier.
Interpret this using sensitivity.
\end{exercise}
\begin{proof}[Short solution]
With $g(x)=-x\leq0$, the solution is $x^\star=0$. Stationarity gives
$2x^\star-\lambda^\star=0$, hence $\lambda^\star=0$, although
$g(x^\star)=0$. Relaxing the constraint to $-x\leq u$ with small $u\geq0$
still permits the unconstrained minimizer $x=0$, so the optimal value remains
zero. Its sensitivity is therefore zero, consistent with the multiplier.
\end{proof}

\paragraph{Chapter summary.}
Constraints restrict the allowable directions of motion. Lagrange multipliers
balance the objective gradient against active constraint normals, duality gives
bounds, and KKT conditions combine stationarity, feasibility, and
complementarity. In convex problems, these conditions provide a complete global
certificate; their multipliers also quantify the value of relaxing constraints.

\chapter{Numerical Optimization Algorithms}
\chapteroverview{Practical iterative methods for smooth, constrained, nonsmooth, and data-driven objectives.}

The previous chapters described optimality conditions: they tell us what a
solution must look like. Numerical optimization asks the complementary
question: how can a computer actually find such a point?

Closed-form solutions are available for a few important problems, such as
unconstrained least squares and certain quadratic programs. Most realistic
models, however, are too large, nonlinear, constrained, or data-dependent to be
solved by a single formula. Instead, an algorithm starts from an initial guess
$x_0$ and constructs a sequence
\begin{equation}
  x_0,x_1,x_2,\ldots
\end{equation}
that is intended to approach a solution. Each iteration uses local information
to decide where to move and how far to move.

Three questions organize much of this chapter:
\begin{enumerate}
  \item \emph{Direction:} which change in $x$ appears promising?
  \item \emph{Step size:} how far should the algorithm move in that direction?
  \item \emph{Structure:} should the method use curvature, constraints,
  sparsity, or stochastic data information?
\end{enumerate}
An algorithm must also decide when to stop. A small objective change alone is
not a certificate of success, so practical methods monitor quantities such as
gradient norms, feasibility violations, and KKT residuals.

\section{First-Order Methods}

\subsection{Gradient Descent}

\paragraph{Direction and step size.}
Gradient descent is built from the local linear model
\begin{equation}
  f(x+d)\approx f(x)+\nabla f(x)\trans d.
\end{equation}
Among all unit-length directions, the one with the most negative predicted
change solves
\begin{equation}
  \min_{\norm{d}_2\leq1}\nabla f(x)\trans d.
\end{equation}
By Cauchy--Schwarz, the minimum is attained at
$d=-\nabla f(x)/\norm{\nabla f(x)}_2$ whenever the gradient is nonzero. Thus
the negative gradient is the direction of steepest local decrease under the
Euclidean norm.

The algorithm repeatedly moves in this direction:
\begin{equation}
  x_{k+1}=x_k-\alpha_k\nabla f(x_k),
  \label{eq:gradient-descent}
\end{equation}
where $\alpha_k>0$ is the step size. Substitution into the linear model predicts
the change
\begin{equation}
  -\alpha_k\norm{\nabla f(x_k)}_2^2,
\end{equation}
which is negative away from a stationary point. This prediction is only local,
however. The direction may be correct while an excessively long step still
increases the true objective.

A step that is too small is safe but makes little progress. A step that is too
large may pass across a valley, increase the objective, or even cause the
iterates to diverge. Smoothness provides a useful quantitative guide.

\paragraph{Smoothness and guaranteed decrease.}
\begin{definition}[$L$-Lipschitz gradient]
A differentiable function $f$ has an $L$-Lipschitz gradient, or is
$L$-smooth, if
\begin{equation}
  \norm{\nabla f(x)-\nabla f(y)}_2
  \leq L\norm{x-y}_2
  \qquad\text{for all }x,y.
  \label{eq:l-smooth-gradient-numerical}
\end{equation}
\end{definition}

This condition limits how rapidly the slope can change as the point moves. A
small $L$ describes gently varying gradients, whereas a large $L$ permits sharp
curvature and therefore calls for shorter steps. If $f$ is twice
differentiable, it is sufficient that
\begin{equation}
  \norm{\nabla^2 f(x)}_2\leq L
\end{equation}
throughout the region of interest. In this sense, $L$ is a global upper bound
on curvature.

The smoothness condition produces an upper quadratic model of the objective.
Let $d$ be any step and define $\phi(t)=f(x+td)$ for $t\in[0,1]$. The
fundamental theorem of calculus gives
\begin{align}
  f(x+d)-f(x)
  &=\int_0^1 \nabla f(x+td)\trans d\,dt\\
  &=\nabla f(x)\trans d
    +\int_0^1
      \bigl(\nabla f(x+td)-\nabla f(x)\bigr)\trans d\,dt.
\end{align}
By Cauchy--Schwarz and the Lipschitz condition,
\begin{align}
  \bigl(\nabla f(x+td)-\nabla f(x)\bigr)\trans d
  &\leq
  \norm{\nabla f(x+td)-\nabla f(x)}_2\norm{d}_2\\
  &\leq Lt\norm{d}_2^2.
\end{align}
Integrating $Lt\norm{d}_2^2$ from $0$ to $1$ yields
\begin{equation}
  f(x+d)
  \leq f(x)+\nabla f(x)\trans d+\frac{L}{2}\norm{d}_2^2.
  \label{eq:descent-lemma-numerical}
\end{equation}
This result is the \emph{descent lemma}. The first two terms form the local
linear model; the final quadratic term bounds how far the true objective can
curve above that model.

For a gradient step, set
$d=-\alpha\nabla f(x_k)$ in
\cref{eq:descent-lemma-numerical}. Then
\begin{equation}
  f(x_k-\alpha\nabla f(x_k))
  \leq f(x_k)
  -\alpha\left(1-\frac{L\alpha}{2}\right)
   \norm{\nabla f(x_k)}_2^2.
\end{equation}
The first term $-\alpha\norm{\nabla f(x_k)}_2^2$ is the decrease predicted by
the linear model, while the positive term
$L\alpha^2\norm{\nabla f(x_k)}_2^2/2$ accounts for curvature. Their sum remains
negative when
\begin{equation}
  1-\frac{L\alpha}{2}>0,
\end{equation}
or equivalently $0<\alpha<2/L$. Hence this range guarantees a strict objective
decrease whenever $\nabla f(x_k)\neq0$. The common, more conservative choice
$\alpha\leq1/L$ leaves a clear safety margin and gives the simpler bound
\begin{equation}
  f(x_{k+1})
  \leq f(x_k)-\frac{\alpha}{2}
  \norm{\nabla f(x_k)}_2^2.
\end{equation}

Objective decrease alone does not say how quickly the iterates approach an
optimum. A decreasing sequence might still converge arbitrarily slowly, and
for a nonconvex function it might approach a local minimum or another
stationary point. Convexity provides the additional structure needed for
global rate statements. Let $x^\star$ be a minimizer and define the objective
error
\begin{equation}
  \Delta_k=f(x_k)-f(x^\star)\geq0.
\end{equation}

\paragraph{Smooth convex case.}
Suppose $f$ is convex and $L$-smooth, and use the fixed step
$\alpha=1/L$. We will show that each iteration reduces a useful measure of
distance to a minimizer.

Write $g_k=\nabla f(x_k)$. The gradient update is
\begin{equation}
  x_{k+1}=x_k-\frac{1}{L}g_k.
  \label{eq:gd-one-over-l-update}
\end{equation}
First, apply the descent lemma with
$d=-g_k/L$. This gives
\begin{equation}
  f(x_{k+1})
  \leq
  f(x_k)-\frac{1}{2L}\norm{g_k}_2^2.
  \label{eq:gd-one-over-l-descent}
\end{equation}
Thus the squared gradient controls the objective decrease in one iteration.

Convexity supplies a second relation. Its first-order condition gives
\begin{equation}
  f(x^\star)
  \geq
  f(x_k)+g_k\trans(x^\star-x_k),
\end{equation}
or, after rearranging,
\begin{equation}
  \Delta_k
  \leq
  g_k\trans(x_k-x^\star).
  \label{eq:convex-objective-error-gradient}
\end{equation}
Combining
\cref{eq:gd-one-over-l-descent,eq:convex-objective-error-gradient}
yields
\begin{equation}
  \Delta_{k+1}
  \leq
  g_k\trans(x_k-x^\star)
  -\frac{1}{2L}\norm{g_k}_2^2.
  \label{eq:gd-convex-intermediate-bound}
\end{equation}

The expression on the right has a geometric meaning. Expanding the squared
distance after the update in \cref{eq:gd-one-over-l-update} gives
\begin{align}
  \norm{x_{k+1}-x^\star}_2^2
  &=
  \norm{x_k-x^\star-\frac{1}{L}g_k}_2^2\\
  &=
  \norm{x_k-x^\star}_2^2
  -\frac{2}{L}g_k\trans(x_k-x^\star)
  +\frac{1}{L^2}\norm{g_k}_2^2.
\end{align}
Rearranging this identity shows that
\begin{equation}
  g_k\trans(x_k-x^\star)
  -\frac{1}{2L}\norm{g_k}_2^2
  =
  \frac{L}{2}
  \left(
    \norm{x_k-x^\star}_2^2
    -\norm{x_{k+1}-x^\star}_2^2
  \right).
\end{equation}
Substitution into \cref{eq:gd-convex-intermediate-bound} therefore gives
\begin{equation}
  \Delta_{k+1}
  \leq
  \frac{L}{2}
  \left(
    \norm{x_k-x^\star}_2^2
    -\norm{x_{k+1}-x^\star}_2^2
  \right).
  \label{eq:gd-telescoping-bound}
\end{equation}
This inequality converts objective error into a decrease of squared distance.
That form is especially useful because the distance differences cancel when
they are summed over successive iterations. Specifically, summing over
$j=0,\ldots,k-1$ causes all intermediate distance terms to cancel:
\begin{equation}
  \sum_{j=1}^k
  \Delta_j
  \leq \frac{L}{2}\norm{x_0-x^\star}_2^2.
\end{equation}
The descent inequality in \cref{eq:gd-one-over-l-descent} shows that the
objective values are nonincreasing:
\begin{equation}
  f(x_1)\geq f(x_2)\geq\cdots\geq f(x_k)\geq f(x^\star).
\end{equation}
Subtracting $f(x^\star)$ from every term shows that the errors are also
nonincreasing:
\begin{equation}
  \Delta_1\geq\Delta_2\geq\cdots\geq\Delta_k\geq0.
\end{equation}
Hence every term in the sum is at least as large as the final error
$\Delta_k$, and therefore
\begin{equation}
  k\Delta_k
  \leq
  \sum_{j=1}^k\Delta_j
  \leq
  \frac{L}{2}\norm{x_0-x^\star}_2^2.
\end{equation}
Dividing by $k$ gives
\begin{equation}
  \boxed{
  \Delta_k
  \leq
  \frac{L\norm{x_0-x^\star}_2^2}{2k}.}
  \label{eq:gd-convex-rate}
\end{equation}
Equivalently, to guarantee an objective error of at most $\epsilon$, it is
sufficient to choose
\begin{equation}
  k
  \geq
  \frac{L\norm{x_0-x^\star}_2^2}{2\epsilon}.
  \label{eq:gd-convex-complexity}
\end{equation}
The required iteration count is therefore proportional to $1/\epsilon$:
requesting ten times smaller error can require roughly ten times as many
iterations. This inverse dependence is described as a \emph{sublinear}, or
$O(1/k)$, convergence rate. Convexity allows flat directions, so the function
value can approach its optimum even when the optimizer is not unique or the
iterates themselves converge slowly.

\paragraph{Smooth strongly convex case.}
The $O(1/k)$ rate allows the objective to become arbitrarily flat near its
minimum. A faster rate follows when such flatness is ruled out. If $f$ is
$\mu$-strongly convex, then it has at least quadratic curvature in every
direction. More precisely, for every $x$ and $y$,
\begin{equation}
  f(y)
  \geq f(x)+\nabla f(x)\trans(y-x)
  +\frac{\mu}{2}\norm{y-x}_2^2.
  \label{eq:strong-convexity-numerical}
\end{equation}
The parameter $\mu>0$ measures the minimum curvature of the function. Unlike a
merely convex objective, a strongly convex objective cannot contain a flat line
segment. It also has a unique minimizer $x^\star$.

Strong convexity connects the gradient norm to the objective error. Set
$y=x^\star$ in \cref{eq:strong-convexity-numerical} and write
$z=x^\star-x$. Then
\begin{equation}
  f(x^\star)
  \geq f(x)+\nabla f(x)\trans z
  +\frac{\mu}{2}\norm{z}_2^2.
  \label{eq:strong-convexity-at-minimizer}
\end{equation}
For fixed $x$, the last two terms form a quadratic function of $z$. Completing
the square gives
\begin{equation}
  \nabla f(x)\trans z+\frac{\mu}{2}\norm{z}_2^2
  =
  \frac{\mu}{2}
  \norm{z+\frac{1}{\mu}\nabla f(x)}_2^2
  -\frac{1}{2\mu}\norm{\nabla f(x)}_2^2.
\end{equation}
The squared term is nonnegative, so
\begin{equation}
  f(x^\star)
  \geq f(x)-\frac{1}{2\mu}\norm{\nabla f(x)}_2^2.
\end{equation}
Rearranging yields the gradient domination bound
\begin{equation}
  \norm{\nabla f(x)}_2^2
  \geq 2\mu\bigl(f(x)-f(x^\star)\bigr).
  \label{eq:strong-convex-gradient-bound}
\end{equation}
This inequality says that a strongly convex function cannot have a very small
gradient while its objective value remains far above the minimum. The gradient
must be large enough, relative to the remaining error, to drive further
progress.

Now use gradient descent with $\alpha=1/L$. The descent bound derived above
gives
\begin{align}
  \Delta_{k+1}
  &\leq \Delta_k-\frac{1}{2L}\norm{\nabla f(x_k)}_2^2\\
  &\leq
  \left(1-\frac{\mu}{L}\right)\Delta_k.
\end{align}
The second line follows by substituting
\cref{eq:strong-convex-gradient-bound}. If
$q=1-\mu/L$, then $0\leq q<1$ and the recurrence reads
$\Delta_{k+1}\leq q\Delta_k$. Applying it repeatedly gives
\begin{equation}
  \Delta_1\leq q\Delta_0,
  \qquad
  \Delta_2\leq q^2\Delta_0,
  \qquad\ldots\qquad
  \Delta_k\leq q^k\Delta_0.
\end{equation}
Therefore
\begin{equation}
  \boxed{
  \Delta_k
  \leq
  \left(1-\frac{\mu}{L}\right)^k
  \Delta_0.}
  \label{eq:gd-strongly-convex-rate}
\end{equation}
This is \emph{geometric}, or linear, convergence: each iteration reduces the
current error bound by the same fixed factor $1-\mu/L$. The term linear
convergence refers to this linear recurrence for the error; it does not mean
that the objective function itself is linear.

The bound also gives a useful estimate of the required number of iterations.
Since $1-t\leq e^{-t}$ for $t\geq0$,
\begin{equation}
  \Delta_k
  \leq
  \exp\left(-\frac{k\mu}{L}\right)\Delta_0.
\end{equation}
To guarantee $\Delta_k\leq\epsilon$, it is sufficient to take
\begin{equation}
  k
  \geq
  \frac{L}{\mu}
  \log\left(\frac{\Delta_0}{\epsilon}\right).
  \label{eq:gd-strongly-convex-complexity}
\end{equation}
The dependence on $\epsilon$ is logarithmic. Thus reducing the target error by
another factor of ten requires approximately the same additional number of
iterations each time.

\paragraph{Conditioning and geometry.}
The ratio
\begin{equation}
  \kappa=\frac{L}{\mu}
\end{equation}
is the condition number associated with these curvature bounds. If $\kappa$ is
close to one, the
contraction factor $1-1/\kappa$ is small and gradient descent is fast. If
$\kappa$ is large, the factor is close to one and progress is slow. These rate
bounds are called global because they hold from any starting point under the
stated assumptions, not merely after the iterates are already close to the
minimizer.

For the quadratic objective
\begin{equation}
  f(x)=\frac12x\trans Qx-b\trans x,
  \qquad Q\succ0,
\end{equation}
$L$ and $\mu$ may be chosen as the largest and smallest eigenvalues of $Q$.
Their ratio describes the shape of the level sets. A large ratio produces a
long, narrow valley: the gradient points mostly across the narrow direction,
and the iterates zig-zag rather than moving directly toward the minimizer.
One remedy is to rescale the coordinates so that comparable changes in
different directions have comparable effects on the objective. A
\emph{preconditioner} performs this scaling inside the algorithm. Instead of
using the raw negative gradient, preconditioned gradient descent takes
\begin{equation}
  x_{k+1}
  =x_k-\alpha_k P_k\nabla f(x_k),
  \qquad P_k\succ0,
  \label{eq:preconditioned-gradient-descent}
\end{equation}
where the positive-definite matrix $P_k$ changes the relative step length in
different directions. Directions believed to have high curvature receive
shorter effective steps, while flatter directions receive longer ones.

For the quadratic objective above, the ideal choice would be $P=Q^{-1}$.
Then all curvature is normalized and, with $\alpha=1$, the method reaches the
minimizer in one step in exact arithmetic. Computing or applying $Q^{-1}$ is
usually as difficult as solving the original problem, so practical
preconditioners use inexpensive approximations, such as a diagonal matrix
formed from the diagonal of $Q$. A useful preconditioner does not need to remove
all anisotropy; even a rough scaling that reduces the effective condition
number can substantially accelerate convergence.

\subsection{Backtracking Line Search}

The safe fixed-step rule above requires knowledge of $L$, which is often
unavailable or overly conservative. A line search chooses the step size by
testing the objective directly.

Suppose $d_k$ is a descent direction, so
$\nabla f(x_k)\trans d_k<0$. Backtracking line search begins with a trial step
$\alpha$, often $\alpha=1$, and checks
\begin{equation}
  f(x_k+\alpha d_k)
  \leq f(x_k)+c\alpha\nabla f(x_k)\trans d_k,
  \qquad c\in(0,1).
  \label{eq:armijo-condition}
\end{equation}
If the condition fails, the method replaces $\alpha$ by $\rho\alpha$, where
$\rho\in(0,1)$, and tries again.

The right-hand side is the current value plus a fraction $c$ of the decrease
predicted by the linear model. Since
$\nabla f(x_k)\trans d_k<0$, it is smaller than $f(x_k)$. The Armijo condition
therefore asks the actual objective decrease to be meaningfully related to the
predicted decrease, without demanding that the local model be exact.

Backtracking is popular because it is simple and robust. It adapts to local
curvature and requires only additional objective evaluations. Although it was
introduced here immediately after gradient descent, the procedure applies to
any direction $d_k$ satisfying
$\nabla f(x_k)\trans d_k<0$. Later sections will construct other such
directions using curvature information; the same backtracking test can then be
used to choose their step lengths. The parameters are commonly chosen with a
small $c$, such as $10^{-4}$, and a reduction factor such as $\rho=1/2$.

\subsection{Momentum and Acceleration}

Gradient descent treats every iteration independently. Momentum introduces a
state variable that remembers recent directions:
\begin{align}
  v_{k+1}&=\beta v_k+\nabla f(x_k),\\
  x_{k+1}&=x_k-\alpha v_{k+1},
  \qquad \beta\in[0,1).
\end{align}
The new gradient is combined with an exponentially decaying history of earlier
gradients. When gradients repeatedly point in the same direction, their effects
accumulate and the method moves faster. When a component alternates sign, as in
zig-zagging across a narrow valley, the positive and negative contributions
partly cancel.

The parameter $\beta$ controls memory. Values close to zero resemble ordinary
gradient descent; values close to one preserve information for many
iterations. Nesterov acceleration uses a related idea but evaluates the
gradient at an extrapolated point. For smooth convex problems it improves the
worst-case objective rate from order $1/k$ to order $1/k^2$.

Acceleration is not free. Momentum may overshoot, objective values need not
decrease at every iteration, and the step size and momentum parameter interact.
In nonconvex learning problems it is best viewed as a powerful practical
heuristic rather than an automatic guarantee of faster convergence.

\section{Curvature-Based Methods}

\subsection{Newton's Method}

Gradient descent uses only the slope of the objective. Newton's method also
uses curvature. Near $x_k$, the second-order Taylor model is
\begin{equation}
  m_k(d)
  =f(x_k)+\nabla f(x_k)\trans d
  +\frac12d\trans H_kd,
  \qquad H_k=\nabla^2f(x_k).
\end{equation}
Rather than minimizing the original objective immediately, Newton's method
minimizes this local quadratic approximation. Differentiating $m_k$ with
respect to $d$ gives
\begin{equation}
  \nabla m_k(d)=\nabla f(x_k)+H_kd.
\end{equation}
Setting this gradient to zero produces the Newton equation
\begin{equation}
  H_kd_k=-\nabla f(x_k),
  \label{eq:newton-step}
\end{equation}
followed by $x_{k+1}=x_k+d_k$.

When $H_k\succ0$, the quadratic model is strictly convex and $d_k$ is its unique
minimizer. The Hessian rescales the gradient according to local curvature:
directions with high curvature receive smaller motion, while flatter directions
receive larger motion. For a strictly convex quadratic objective, the model is
exact and Newton's method reaches the minimizer in one iteration in exact
arithmetic.

Near a smooth minimizer with a nonsingular positive-definite Hessian, Newton's
method can converge quadratically: once sufficiently close, the number of
correct digits may roughly double at each iteration. This rapid local behavior
is its main attraction.

Far from a minimizer, several problems can occur. The Hessian may be indefinite,
so the Newton direction need not be a descent direction. Even with a
positive-definite Hessian, the quadratic model may be accurate only near
$x_k$, making the full step too aggressive. A damped Newton method uses
$x_{k+1}=x_k+\alpha_kd_k$ with a line search. Trust-region methods provide
another safeguard by minimizing the model only within a neighborhood where it
is believed to be reliable.

Computationally, one should solve the linear system
$H_kd_k=-\nabla f(x_k)$ using an appropriate factorization or iterative solver.
Explicitly forming $H_k^{-1}$ is slower, less accurate, and usually destroys
sparsity.

\subsection{Quasi-Newton Methods}

Forming and factoring the exact Hessian may be too expensive, especially when
$n$ is large. Quasi-Newton methods build a curvature approximation from changes
already observed during the iteration:
\begin{equation}
  s_k=x_{k+1}-x_k,
  \qquad
  y_k=\nabla f(x_{k+1})-\nabla f(x_k).
\end{equation}
For a quadratic function with constant Hessian $H$, these vectors satisfy
$y_k=Hs_k$. Quasi-Newton methods imitate this relation by constructing a matrix
$B_{k+1}$ satisfying the \emph{secant equation}
\begin{equation}
  B_{k+1}s_k=y_k.
\end{equation}
They then compute a Newton-like direction from
\begin{equation}
  B_kd_k=-\nabla f(x_k).
\end{equation}

One vector equation cannot determine an entire matrix, so an additional rule is
needed to choose the update. BFGS selects a symmetric update that changes the
previous approximation as little as possible while satisfying the secant
equation. With a suitable line search and $s_k\trans y_k>0$, it preserves
positive definiteness, so the resulting direction remains a descent direction.

Dense BFGS stores an $n\times n$ matrix and is attractive for small and
medium-sized problems. Limited-memory BFGS (L-BFGS) avoids storing this matrix.
It keeps only a short history of pairs $(s_k,y_k)$ and applies the implied
inverse-Hessian approximation through vector operations. This makes L-BFGS a
standard choice for large smooth problems.

Quasi-Newton methods occupy a useful middle ground: they are more expensive per
iteration than gradient descent but avoid exact second derivatives and often
need far fewer iterations.

\section{Constraints, Nonsmoothness, and Structured Solvers}

\subsection{Projected and Proximal Gradient}

Sometimes the feasible set is simple enough that the nearest feasible point can
be computed cheaply. Examples include boxes, Euclidean balls, and affine
subspaces. In this case, projected gradient descent separates improvement of
the objective from restoration of feasibility:
\begin{equation}
  x_{k+1}=\Pi_C\bigl(x_k-\alpha_k\nabla f(x_k)\bigr),
  \label{eq:projected-gradient}
\end{equation}
where the projection operator is defined by
\begin{equation}
  \Pi_C(z)
  =
  \argmin_{y\in C}\frac12\norm{y-z}_2^2.
  \label{eq:euclidean-projection-definition}
\end{equation}
Thus $\Pi_C(z)$ is the point in $C$ nearest to $z$. In finite-dimensional
Euclidean space, nonemptiness and closedness of $C$ guarantee that at least one
nearest point exists. Convexity makes that nearest point unique, so
$\Pi_C$ is a single-valued mapping.

Projected gradient descent first takes an ordinary gradient step, which may
leave $C$, and then applies this well-defined projection to restore
feasibility. In particular,
\begin{equation}
  \Pi_C(z)=z
  \qquad\text{whenever }z\in C,
\end{equation}
because a feasible point is already its own nearest feasible point.

The geometry is especially clear for a Euclidean ball. Consider
\begin{equation}
  C=\{x\in\R^n:\norm{x-c}_2\leq r\},
\end{equation}
with center $c$ and radius $r>0$. The ball includes both its boundary and its
interior. This is important: the boundary alone,
$\{x:\norm{x-c}_2=r\}$, is a sphere and is not a convex set.

If $z$ already lies in the ball, it is feasible and projection leaves it
unchanged. If $z$ lies outside, the nearest feasible point lies on the radial
line from $c$ through $z$. Projection therefore shortens the distance from the
center to exactly $r$:
\begin{equation}
  \Pi_C(z)
  =
  \begin{cases}
    z,
      & \norm{z-c}_2\leq r,\\[2mm]
    c+r\dfrac{z-c}{\norm{z-c}_2},
      & \norm{z-c}_2>r.
  \end{cases}
  \label{eq:projection-euclidean-ball}
\end{equation}
As a concrete two-dimensional example, let $C$ be the unit disk centered at
the origin and take $z=(3,4)$. Since $\norm{z}_2=5$, the point lies outside the
disk. Normalizing it to unit length gives
\begin{equation}
  \Pi_C(3,4)
  =\frac{(3,4)}{5}
  =\left(\frac35,\frac45\right).
\end{equation}
Thus the projection preserves the direction from the center while reducing the
radius from $5$ to $1$. This simple radial correction is what projected
gradient descent performs whenever an unconstrained step leaves a ball-shaped
feasible region.

Projection is designed for hard constraints. A related method applies when the
objective has the form
\begin{equation}
  F(x)=f(x)+g(x),
\end{equation}
where $f$ is smooth but $g$ may have corners. For example,
$g(x)=\lambda\norm{x}_1$ is not differentiable when a coordinate is zero, so
ordinary gradient descent cannot be applied directly to the whole objective.

The \emph{proximal operator}
\begin{equation}
  \operatorname{prox}_{\alpha g}(z)
  =
  \argmin_x
  \left\{
    g(x)+\frac{1}{2\alpha}\norm{x-z}_2^2
  \right\}
  \label{eq:proximal-operator}
\end{equation}
chooses a point that reduces $g$ without moving too far from $z$. Proximal
gradient first takes a gradient step for the smooth term and then applies this
operator:
\begin{equation}
  x_{k+1}
  =
  \operatorname{prox}_{\alpha_k g}
  \bigl(x_k-\alpha_k\nabla f(x_k)\bigr).
  \label{eq:proximal-gradient-update}
\end{equation}

For $g(x)=\lambda\norm{x}_1$, the proximal operator is
\emph{soft thresholding}, applied to each coordinate:
\begin{equation}
  [\operatorname{prox}_{\alpha\lambda\norm{\cdot}_1}(z)]_i
  =
  \operatorname{sign}(z_i)
  \max\{|z_i|-\alpha\lambda,0\}.
  \label{eq:soft-thresholding}
\end{equation}
It shrinks large coordinates and sets sufficiently small ones exactly to zero.
For instance, if $z=(3,0.4,-2)$ and $\alpha\lambda=1$, the output is
$(2,0,-1)$. This example is the main reason proximal methods appear in sparse
estimation; further theory is not needed for the developments that follow.

\subsection{Penalty and Barrier Methods}

Projected gradient is attractive when projection onto the feasible set is
easy. For general nonlinear constraints, however, computing that projection
may be as difficult as solving the original problem. Other methods therefore
handle the constraints in different ways.

Penalty and barrier methods replace the constrained problem by a sequence of
problems with modified objectives, but they approach feasibility from opposite
sides. Penalty methods allow infeasible iterates and increasingly punish
constraint violations, whereas barrier methods keep iterates strictly inside
the feasible region.

\paragraph{Penalty methods.}
A penalty method permits temporary constraint violation but makes it costly.
For example, equality constraints $h(x)=0$ can be incorporated into the
objective through the quadratic penalty
\begin{equation}
  f(x)+\frac{\rho}{2}\norm{h(x)}_2^2.
\end{equation}
The parameter $\rho>0$ determines how strongly violations are punished. A
small value gives the algorithm freedom to improve the original objective,
whereas a large value places greater emphasis on feasibility. Since no finite
$\rho$ necessarily enforces $h(x)=0$ exactly, a penalty method commonly solves
several problems while gradually increasing $\rho$. Very large penalties can
make the numerical problem poorly conditioned, so this increase must be
handled with care.

\paragraph{Barrier methods.}
For inequality constraints $g_i(x)\leq0$, a barrier method takes the opposite
viewpoint. It starts strictly inside the feasible region and prevents the
iterates from crossing its boundary. A common modified objective is
\begin{equation}
  f(x)-\mu\sum_i\log(-g_i(x)),
  \qquad g_i(x)<0.
\end{equation}
As an iterate approaches a boundary, the logarithmic term becomes arbitrarily
large and pushes it back toward the interior. The parameter $\mu>0$ controls
the strength of this artificial repulsion. Solving a sequence of problems with
decreasing values of $\mu$ allows the solution to move progressively closer to
the boundary when the true optimum lies there. This is the central idea behind
many interior-point methods.

The two methods therefore approach constraints differently. Penalty methods
are conceptually simple and tolerate infeasible iterates, while barrier methods
remain strictly feasible and approach inequality boundaries from the inside.
Neither strategy is uniformly best, and practical solvers may combine these
ideas with other techniques.

\subsection{Choosing a Solver for Structured Convex Problems}

Linear programs, quadratic programs, and second-order cone programs have more
structure than a general nonlinear optimization problem. Practical solvers use
that structure rather than treating every problem in the same way. For
example, linear objectives and constraints lead to different computations from
quadratic objectives, and cone constraints require different operations from
simple bounds.

There are therefore several solver families. Common names include simplex,
active-set, interior-point, and first-order methods. The important point is
that they make different compromises:
some perform expensive iterations but need relatively few of them, while
others use cheap iterations but may require many more. Some are designed for
high accuracy, whereas others are attractive for very large problems when a
moderately accurate answer is sufficient.

The best choice also depends on how the problem is used. A sparse problem
should be solved without converting its matrices to dense form. A sequence of
similar problems may benefit from a warm start using the previous solution.
Real-time applications may prefer predictable and inexpensive iterations,
while an offline design problem may justify a more costly high-accuracy solve.

In many software packages, the modeling system converts the user-written
problem into a standard form and selects a suitable solver automatically. This
is convenient, but the returned vector should not be trusted without checking
the solver status, constraint violations, and an optimality residual or gap.
Poor scaling can also make an otherwise easy problem numerically difficult.

Structured convex solvers also play an important role inside algorithms for
general nonlinear constraints. Sequential quadratic programming (SQP), for
example, keeps the constraints explicit. At the current iterate, it builds a
quadratic model of the objective and linear models of the nonlinear
constraints. The resulting quadratic program proposes a step, so every SQP
iteration requires a suitable QP solver. A line search or trust region may
shorten the proposed step when the local models are not yet reliable. Near a
well-behaved solution, SQP can converge rapidly.

The central lesson is that no numerical method is universally best. Problem
type, size, sparsity, desired accuracy, and available computation all influence
the choice. For this book, it is enough to recognize these tradeoffs and to use
a reliable solver designed for the problem class at hand.

\subsection{Operator Splitting for Convex Quadratic Programs}

Chapter~3 showed the simplest algebraic approach to an equality-constrained
quadratic program: write its stationarity and feasibility conditions and solve
the KKT system in \cref{eq:equality-qp-kkt-system}. For a small or medium-sized
problem, this direct approach is often the natural choice. It computes the
primal variables and equality multipliers simultaneously, and an accurate
sparse factorization can provide a high-precision solution in one linear
solve.

The same strategy becomes less direct when the problem contains many
inequality or bound constraints. The active inequalities are not known in
advance, so one must identify an active set, introduce slack and barrier
variables, or repeatedly solve modified KKT systems. Moreover, for a very
large sparse problem, factoring the full indefinite KKT matrix can require
substantial memory and computation. These costs are particularly important
when a sequence of closely related QPs must be solved, as in model predictive
control or sequential quadratic programming.

Operator-splitting methods make a different tradeoff. They separate the
coupled quadratic objective from constraints that admit inexpensive
projections, then enforce agreement through a multiplier iteration. Each
iteration is usually cheaper and can reuse a matrix factorization, exploit
sparsity, and start from the previous solution. The price is that many
iterations may be required for high accuracy. Thus a direct KKT solve and
operator splitting solve the same optimality conditions but organize the
linear algebra differently: direct methods emphasize accuracy in relatively
few expensive solves, whereas splitting methods emphasize inexpensive,
repeatable, and warm-startable steps.

The representative derivation below shows how optimality conditions become
such repeated computations. The simplified pattern is used by solvers such as
OSQP; production implementations add scaling, regularization, acceleration,
and adaptive parameter selection.

Consider the standard convex QP
\begin{equation}
  \begin{aligned}
    \minimize_x\quad
      &\frac12x\trans Px+q\trans x\\
    \subjectto\quad
      &\ell\leq Ax\leq u,
  \end{aligned}
  \label{eq:operator-splitting-standard-qp}
\end{equation}
where $P\succeq0$. The two-sided bounds include equalities, one-sided
inequalities, and simple variable bounds as special cases.

\paragraph{Separating the objective from the bounds.}
Introduce an auxiliary vector $z$ and rewrite the constraints as
\begin{equation}
  Ax=z,
  \qquad
  \ell\leq z\leq u.
  \label{eq:operator-splitting-separation}
\end{equation}
The variable $x$ now carries the quadratic objective, while $z$ carries the
simple bounds. The equality $Ax=z$ requires the two descriptions to agree.
Let $C=[\ell,u]$.

Before constructing an algorithm, write the KKT conditions. Introduce
$\lambda$ for $Ax-z=0$, $\alpha\geq0$ for the lower bound
$\ell-z\leq0$, and $\beta\geq0$ for the upper bound $z-u\leq0$. The
ordinary Lagrangian is
\begin{equation}
  \mathcal L(x,z,\lambda,\alpha,\beta)
  =\frac12x\trans Px+q\trans x
   +\lambda\trans(Ax-z)
   +\alpha\trans(\ell-z)
   +\beta\trans(z-u).
  \label{eq:splitting-full-lagrangian}
\end{equation}
Stationarity in $z$ and complementary slackness give
\begin{align}
  -\lambda-\alpha+\beta&=0,
    \label{eq:splitting-box-multiplier-relation}\\
  \alpha_i(z_i-\ell_i)&=0,
  \qquad
  \beta_i(u_i-z_i)=0.
\end{align}
Thus $\lambda_i=0$ in the interior of an interval,
$\lambda_i\leq0$ at a lower bound, and $\lambda_i\geq0$ at an upper bound.
Eliminating $\alpha$ and $\beta$ summarizes all three cases as
\begin{equation}
  \lambda\trans(w-z)\leq0
  \qquad\text{for every }w\in C.
  \label{eq:splitting-kkt-box}
\end{equation}
The compact KKT system is therefore
\begin{align}
  Px+q+A\trans\lambda&=0,
    \label{eq:splitting-kkt-stationarity}\\
  Ax-z&=0,
    \label{eq:splitting-kkt-feasibility}\\
  \lambda\trans(w-z)&\leq0
    \qquad\text{for every }w\in C.
\end{align}

\paragraph{From the KKT system to an augmented Lagrangian.}
The three KKT conditions are coupled: changing $x$ changes $Ax$, which affects
both $z$ and the multiplier. Operator splitting handles this coupling
iteratively. It starts from the augmented Lagrangian
\begin{equation}
  \mathcal L_\rho(x,z,\lambda)
  =\frac12x\trans Px+q\trans x
   +\lambda\trans(Ax-z)
   +\frac{\rho}{2}\norm{Ax-z}_2^2,
  \qquad z\in C,
  \label{eq:splitting-augmented-lagrangian}
\end{equation}
where $\rho>0$. The linear multiplier term represents the equality constraint,
and the quadratic term makes disagreement between $Ax$ and $z$ costly during
intermediate iterations.

We will minimize $\mathcal L_\rho$ with respect to the primal variables
$x$ and $z\in C$, but maximize it with respect to the equality multiplier
$\lambda$. In other words, we seek a saddle point rather than a joint minimum.
Why is $\lambda$ maximized? Let $r=Ax-z$ denote the
equality residual. Because $\lambda$ is unrestricted,
\begin{equation}
  \sup_{\lambda}\lambda\trans r
  =
  \begin{cases}
    0, & r=0,\\
    +\infty, & r\neq0.
  \end{cases}
  \label{eq:equality-as-dual-supremum}
\end{equation}
If $r\neq0$, choosing $\lambda=t r$ and letting $t\to\infty$ makes
$\lambda\trans r=t\norm{r}_2^2$ arbitrarily large. Maximization over
$\lambda$ therefore acts as an exact test of the equality. 

Consequently, the original constrained QP can be written as the saddle-point
problem
\begin{equation}
  \min_{x,\,z\in C}\;
  \sup_{\lambda}\quad
  \mathcal L_\rho(x,z,\lambda).
  \label{eq:splitting-saddle-problem}
\end{equation}
The variables $x$ and $z$ are minimized because they are the original decision
variables, whereas $\lambda$ is maximized because it detects violations of
$Ax=z$. At feasibility, both the multiplier term and the quadratic penalty
vanish, leaving exactly the original objective.

The following are the KKT conditions of the saddle problem. They may also be
viewed as first-order necessary conditions (FONC): ordinary stationarity for
the unconstrained variables $x$ and $\lambda$, together with the constrained
first-order condition for $z\in C$:
\begin{align}
  Px+q+A\trans\lambda
  +\rho A\trans(Ax-z)&=0,
    \label{eq:augmented-kkt-x}\\
  \bigl(\lambda+\rho(Ax-z)\bigr)\trans(w-z)&\leq0
    \qquad\text{for every }w\in C,
    \label{eq:augmented-kkt-z}\\
  Ax-z&=0.
    \label{eq:augmented-kkt-lambda}
\end{align}
The first equation is $\nabla_x\mathcal L_\rho=0$. For the constrained
minimization in $z$, the FONC is
\begin{equation}
  \nabla_z\mathcal L_\rho(x,z,\lambda)\trans(w-z)\geq0
  \qquad\text{for every }w\in C.
\end{equation}
Since $\nabla_z\mathcal L_\rho=-\lambda-\rho(Ax-z)$, this is the second
displayed condition. The third equation is
$\nabla_\lambda\mathcal L_\rho=Ax-z=0$; it is the stationarity condition for
maximization in the unrestricted variable $\lambda$.

Thus ``FONC'' describes how these equations are derived, while ``KKT
conditions'' describes the complete optimality system, including the box
constraint on $z$. Because the problem is convex in $(x,z)$ and affine in
$\lambda$, these conditions are also sufficient for a saddle point whenever a
solution exists.

Substituting $Ax-z=0$ into the first two conditions removes every term
multiplied by $\rho$ and leaves
\begin{align}
  Px+q+A\trans\lambda&=0,\\
  \lambda\trans(w-z)&\leq0
    \qquad\text{for every }w\in C,\\
  Ax-z&=0.
\end{align}
These are exactly the original KKT conditions in
\cref{eq:splitting-kkt-stationarity,eq:splitting-kkt-feasibility,eq:splitting-kkt-box}.
Thus the quadratic penalty changes the path taken away from feasibility but
does not change the desired primal--dual solution.

Minimizing $\mathcal L_\rho$ for an arbitrary fixed $\lambda$ would not provide
this guarantee: its minimizer may still have $Ax\neq z$. The multiplier-ascent
step is what drives this disagreement to zero.

\paragraph{Deriving the three updates.}
The index $k=0,1,2,\ldots$ counts iterations. At iteration $k$, the vectors
$x_k$, $z_k$, and $\lambda_k$ are the current estimates; the algorithm uses
them to construct the next estimates, carrying the index $k+1$. It is
convenient to use the scaled multiplier
\begin{equation}
  y_k=\frac{\lambda_k}{\rho}.
\end{equation}

To derive the rewritten objective, let $r=Ax-z$. The two terms involving the
coupling residual satisfy
\begin{align}
  \lambda_k\trans r+\frac{\rho}{2}\norm{r}_2^2
  &=\rho y_k\trans r+\frac{\rho}{2}\norm{r}_2^2\\
  &=\frac{\rho}{2}
    \left(\norm{r}_2^2+2y_k\trans r+\norm{y_k}_2^2\right)
    -\frac{\rho}{2}\norm{y_k}_2^2\\
  &=\frac{\rho}{2}\norm{r+y_k}_2^2
    -\frac{\rho}{2}\norm{y_k}_2^2.
\end{align}
Substituting $r=Ax-z$ into the full augmented Lagrangian gives
\begin{equation}
  \mathcal L_\rho(x,z,\lambda_k)
  =
  \frac12x\trans Px+q\trans x
  +\frac{\rho}{2}\norm{Ax-z+y_k}_2^2
  -\frac{\rho}{2}\norm{y_k}_2^2,
  \qquad z\in C.
\end{equation}
Here the first two terms are the original QP objective, the squared term
penalizes corrected disagreement between $Ax$ and $z$, and the final term is
constant while $x$ and $z$ are updated.

After minimizing in $x$ and $z$, update the multiplier by gradient ascent.
The relevant gradient is particularly simple:
\begin{equation}
  \nabla_\lambda\mathcal L_\rho(x,z,\lambda)=Ax-z.
\end{equation}
Because $\lambda$ is maximized, the gradient is added rather than subtracted.
In principle, a separate ascent step size $\eta>0$ could be used:
\begin{equation}
  \lambda_{k+1}
  =\lambda_k+\eta(Ax_{k+1}-z_{k+1}).
\end{equation}
The \emph{alternating direction method of multipliers} (ADMM) is named for
these alternating minimizations in $x$ and $z$, followed by an update of the
Lagrange multiplier. The standard ADMM update chooses $\eta=\rho$, matching
the dual step size to the coefficient of the quadratic penalty. This choice gives
\begin{equation}
  \lambda_{k+1}
  =\lambda_k+\rho(Ax_{k+1}-z_{k+1}).
\end{equation}
Dividing by $\rho$ and using $y_k=\lambda_k/\rho$ yields
\begin{equation}
  y_{k+1}=y_k+Ax_{k+1}-z_{k+1}.
\end{equation}
Thus the scaled multiplier simply accumulates the current constraint residual.
If the separate step size $\eta$ were used instead, its update would be
$y_{k+1}=y_k+(\eta/\rho)(Ax_{k+1}-z_{k+1})$; such variants require their own
step-size conditions for convergence.
Thus alternating minimization in the two primal variables and ascent in the
multiplier gives the three updates
\begin{align}
  x_{k+1}
  &=\argmin_x
    \left\{
      \frac12x\trans Px+q\trans x
      +\frac{\rho}{2}\norm{Ax-z_k+y_k}_2^2
    \right\},
    \label{eq:operator-splitting-x-update}\\
  z_{k+1}
  &=\Pi_C(Ax_{k+1}+y_k),
    \label{eq:operator-splitting-z-update}\\
  y_{k+1}
  &=y_k+Ax_{k+1}-z_{k+1}.
    \label{eq:operator-splitting-y-update}
\end{align}
The $x$-step reduces the quadratic objective while keeping $Ax$ near the
current corrected target $z_k-y_k$. Because it is an unconstrained quadratic,
its first-order condition is the linear system
\begin{equation}
  (P+\rho A\trans A)x_{k+1}
  =-q+\rho A\trans(z_k-y_k).
  \label{eq:operator-splitting-linear-system}
\end{equation}
The $z$-step chooses the feasible box point nearest to
$Ax_{k+1}+y_k$, so it is componentwise clipping:
\begin{equation}
  [\Pi_C(v)]_i=\min\{u_i,\max\{\ell_i,v_i\}\}.
  \label{eq:box-projection-clipping}
\end{equation}
The final step adds the current equality violation to $y_k$. Thus a persistent
disagreement changes future $x$- and $z$-updates until the two variables agree.
Each iteration consists mainly of one structured linear solve, one inexpensive
projection, and vector additions.

\begin{example}[A small operator-splitting QP]
Consider
\begin{equation}
  \begin{aligned}
    \minimize_{x_1,x_2}\quad
      &\frac12(x_1-2)^2+\frac12(x_2-1)^2\\
    \subjectto\quad
      &x_1\geq0,\qquad x_2\geq0,\\
      &x_1+x_2\leq2.
  \end{aligned}
  \label{eq:operator-splitting-example-qp}
\end{equation}
It has the form \cref{eq:operator-splitting-standard-qp} with
\begin{equation}
  P=I,
  \qquad
  q=\begin{bmatrix}-2\\-1\end{bmatrix},
  \qquad
  A=\begin{bmatrix}
    1&0\\
    0&1\\
    1&1
  \end{bmatrix},
\end{equation}
and bounds
\begin{equation}
  \ell=\begin{bmatrix}0\\0\\-\infty\end{bmatrix},
  \qquad
  u=\begin{bmatrix}\infty\\\infty\\2\end{bmatrix}.
\end{equation}
The $z$-update clips the first two components below at zero and the third
component above at two. Repeated linear solves and projections converge to
\begin{equation}
  x^\star=\left(\frac32,\frac12\right),
\end{equation}
the feasible point nearest to $(2,1)$.

This example shows the division of labor clearly: linear algebra handles the
coupled quadratic objective, projection handles the simple bounds, and the
multiplier update forces agreement between them. The iterations are
inexpensive and can be warm-started, but high accuracy may require many of
them. The next subsection develops the contrasting interior-point strategy.
\end{example}

\subsection{Primal--Dual Interior-Point Methods}

The preceding subsection solved a convex QP by separating its objective and
simple constraints into inexpensive subproblems. Interior-point methods make a
different tradeoff: they perform more work per iteration in order to approach
a high-accuracy solution in relatively few iterations. Rather than projecting
onto the constraints or guessing an active set as in Chapter~3, they retain
every inequality in the KKT system and approach complementarity continuously
while keeping the slack variables and inequality multipliers strictly
positive.

A primal--dual interior-point method implements this idea by applying Newton's
method to a perturbed KKT system. Besides providing a direct solver for convex
problems, this construction reinforces the primal--dual Newton viewpoint used
again in the SQP method of the next subsection; in practice, interior-point
methods can also solve the QP subproblems generated by SQP. To see the basic
construction, consider the convex QP
\begin{equation}
  \begin{aligned}
    \minimize_x\quad
      &\frac12x\trans Qx+c\trans x\\
    \subjectto\quad
      &Ax=b,\\
      &Gx\leq h,
  \end{aligned}
  \qquad Q\succeq0.
  \label{eq:ipm-convex-qp}
\end{equation}
where $G$ has $m$ rows, one for each inequality.
Introduce slacks $s\geq0$ so that
\begin{equation}
  Gx+s=h.
  \label{eq:ipm-slack-equation}
\end{equation}
Let $y$ be the equality multiplier and $z\geq0$ the inequality multiplier.
With $S=\operatorname{diag}(s)$, the KKT conditions are
\begin{align}
  Qx+c+A\trans y+G\trans z&=0,
  \label{eq:ipm-kkt-stationarity}\\
  Ax-b&=0,
  \label{eq:ipm-kkt-equality}\\
  Gx+s-h&=0,
  \label{eq:ipm-kkt-slack-feasibility}\\
  Sz&=0,
  \label{eq:ipm-kkt-complementarity}\\
  s&\geq0,
  \qquad z\geq0.
  \label{eq:ipm-kkt-positivity}
\end{align}
The nonlinear part is complementarity: for each constraint,
$s_i z_i=0$. At the solution, either $s_i>0$ and the inequality is inactive
with $z_i=0$, or $z_i>0$ and the constraint is active with $s_i=0$.

\paragraph{Perturbed complementarity and the central path.}
Instead of imposing complementary slackness immediately, an interior-point
method solves the perturbed equations
\begin{equation}
  Sz=\mu\mathbf 1,
  \qquad s>0,
  \qquad z>0,
  \label{eq:ipm-perturbed-complementarity}
\end{equation}
for a sequence of parameters $\mu\downarrow0$. This perturbation is exactly
the logarithmic-barrier construction. Applying a barrier to the slacks gives
\begin{equation}
  \begin{aligned}
    \minimize_{x,s}\quad
      &\frac12x\trans Qx+c\trans x
       -\mu\sum_{i=1}^m\log s_i\\
    \subjectto\quad
      &Ax=b,\\
      &Gx+s=h.
  \end{aligned}
  \label{eq:ipm-barrier-qp}
\end{equation}
Stationarity with respect to $s_i$ gives $z_i=\mu/s_i$, or
$s_i z_i=\mu$. The solutions of these barrier problems trace the
\emph{central path} toward the original QP solution.

\paragraph{Primal--dual Newton step.}
At the current point $(x,y,z,s)$, define
\begin{align}
  r_d&=Qx+c+A\trans y+G\trans z,\\
  r_e&=Ax-b,\\
  r_g&=Gx+s-h,\\
  r_c&=Sz-\sigma\mu\mathbf 1,
\end{align}
where $0\leq\sigma\leq1$ controls how strongly the step is centered. Writing
$Z=\operatorname{diag}(z)$, Newton's method gives
\begin{equation}
  \begin{bmatrix}
    Q&A\trans&G\trans&0\\
    A&0&0&0\\
    G&0&0&I\\
    0&0&S&Z
  \end{bmatrix}
  \begin{bmatrix}
    \Delta x\\
    \Delta y\\
    \Delta z\\
    \Delta s
  \end{bmatrix}
  =-
  \begin{bmatrix}
    r_d\\
    r_e\\
    r_g\\
    r_c
  \end{bmatrix}.
  \label{eq:ipm-primal-dual-newton-system}
\end{equation}
The first three block rows reduce stationarity and primal infeasibility; the
last moves the products $s_i z_i$ toward the target $\sigma\mu$. The system,
or an algebraically reduced version of it, is solved directly rather than by
forming an inverse.

The step length must preserve $s>0$ and $z>0$. A fraction-to-the-boundary rule
is
\begin{equation}
  \alpha_k
  =\min\left\{
    1,
    \tau\min_{\Delta s_i<0}\frac{-s_i}{\Delta s_i},
    \tau\min_{\Delta z_i<0}\frac{-z_i}{\Delta z_i}
  \right\},
  \qquad 0<\tau<1,
  \label{eq:ipm-fraction-to-boundary}
\end{equation}
where a minimum over an empty set is ignored. The update is
\begin{equation}
  (x,y,z,s)
  \leftarrow
  (x,y,z,s)+\alpha_k(\Delta x,\Delta y,\Delta z,\Delta s).
  \label{eq:ipm-primal-dual-update}
\end{equation}
The average complementarity
\begin{equation}
  \mu=\frac{s\trans z}{m}
  \label{eq:ipm-complementarity-measure}
\end{equation}
is reduced as the iterations proceed. The method stops when the stationarity
residual, primal residuals, and complementarity measure are all small.

Interior-point methods avoid explicitly guessing an active set: all
inequalities remain in the system, and the slack and multiplier values reveal
which constraints become active as $\mu$ approaches zero. Their main cost is
factoring a large sparse Newton system at each iteration, but they often reach
high accuracy in relatively few iterations. Modern primal--dual methods may
begin with nonzero primal residuals; \emph{interior} refers primarily to
maintaining $s>0$ and $z>0$ while driving the KKT residuals to zero.

\subsection{Sequential Quadratic Programming}
\label{sec:general-sqp}

Chapter~3 showed how the KKT conditions reduce an equality-constrained QP to a
linear saddle-point system and how active constraints extend the same idea to
inequality-constrained QPs. The preceding subsections then described numerical
methods for solving such QPs without manually enumerating every possible
active set. Sequential quadratic programming (SQP) uses a QP as a local model
of a nonlinear constrained problem: solve the QP for a primal--dual correction,
move to a new point, rebuild the model there, and repeat. The model changes at
every iteration because the objective curvature and constraint Jacobians are
evaluated at the current iterate; this repeated construction is the source of
the word \emph{sequential}.

Consider the nonlinear program
\begin{equation}
  \begin{aligned}
    \minimize_x\quad &f(x)\\
    \subjectto\quad &h(x)=0,\\
                    &g(x)\leq0,
  \end{aligned}
  \label{eq:sqp-general-nlp}
\end{equation}
where $x\in\R^n$, $h:\R^n\to\R^{m_e}$ collects the equality constraints, and
$g:\R^n\to\R^{m_i}$ collects the inequality constraints. Its Lagrangian is
\begin{equation}
  \mathcal L(x,\lambda,\mu)
  =f(x)+\lambda\trans h(x)+\mu\trans g(x),
  \qquad \mu\geq0.
  \label{eq:sqp-general-lagrangian}
\end{equation}
Here $\lambda\in\R^{m_e}$ contains the equality multipliers, while
$\mu\in\R^{m_i}$ contains the nonnegative inequality multipliers. Under a
suitable constraint qualification, a local solution satisfies the KKT
conditions
\begin{align}
  \nabla_x\mathcal L(x,\lambda,\mu)&=0,
  &h(x)&=0,\nonumber\\
  g(x)&\leq0,
  &\mu&\geq0,
  &\mu_jg_j(x)&=0\quad\text{for every }j.
  \label{eq:sqp-general-kkt}
\end{align}
Locally, after the active inequalities have been identified, SQP follows by
applying the Newton iteration in
\cref{eq:newton-system-linearization,eq:newton-system-update} to the resulting
smooth KKT equations.

\paragraph{From a Newton correction to a QP.}
First suppose temporarily that the active inequalities are known. Collect
them together with the equality constraints in a vector $a(x)$, so the local
active constraints are $a(x)=0$, and collect their multipliers in $y$. Let
\begin{equation}
  C(x)=a_x(x),
  \qquad
  C_k=C(x_k)
  \label{eq:sqp-active-constraint-jacobian}
\end{equation}
be the active-constraint Jacobian. Thus $C_kp$ is the first-order change in
the active constraints caused by a step $p$, while $C_k\trans y$ is the
multiplier-weighted sum of their gradients. The active-set KKT equations are
\begin{equation}
  \nabla f(x)+C(x)\trans y=0,
  \qquad
  a(x)=0.
  \label{eq:sqp-active-kkt}
\end{equation}
The joint variable is $(x,y)$, so its Newton correction consists of a primal
step $p_k$ and a multiplier correction $\Delta y_k$. Applying
\cref{eq:newton-system-update} to \cref{eq:sqp-active-kkt} gives
\begin{equation}
  \begin{bmatrix}
    B_k&C_k\trans\\
    C_k&0
  \end{bmatrix}
  \begin{bmatrix}
    p_k\\
    \Delta y_k
  \end{bmatrix}
  =-
  \begin{bmatrix}
    \nabla f(x_k)+C_k\trans y_k\\
    a(x_k)
  \end{bmatrix},
  \label{eq:sqp-kkt-newton-system}
\end{equation}
where the exact Newton curvature is
\begin{equation}
  B_k=\nabla_{xx}^2\mathcal L_A(x_k,y_k),
  \qquad
  \mathcal L_A(x,y)=f(x)+y\trans a(x).
  \label{eq:sqp-lagrangian-hessian}
\end{equation}
The four matrix blocks have the same meaning as in the quadratic KKT system
from Chapter~3. Differentiating stationarity with respect to $x$ produces
$B_k$; differentiating it with respect to $y$ produces $C_k\trans$;
differentiating feasibility with respect to $x$ produces $C_k$; and
feasibility has no direct dependence on $y$, producing the zero block.

The lower block of \cref{eq:sqp-kkt-newton-system} is the linearized
feasibility equation
\begin{equation}
  a(x_k)+C_kp_k=0.
  \label{eq:sqp-active-linearized-feasibility}
\end{equation}
Define the updated multiplier estimate
$\widehat y_k=y_k+\Delta y_k$. The upper block can then be rearranged as
\begin{equation}
  B_kp_k+\nabla f(x_k)+C_k\trans\widehat y_k=0.
  \label{eq:sqp-qp-stationarity}
\end{equation}
Equations
\cref{eq:sqp-active-linearized-feasibility,eq:sqp-qp-stationarity} are exactly
the KKT conditions of the equality-constrained QP
\begin{equation}
  \begin{aligned}
    \minimize_p\quad
      &\nabla f(x_k)\trans p+\frac12p\trans B_kp\\
    \subjectto\quad
      &a(x_k)+C_kp=0.
  \end{aligned}
  \label{eq:sqp-active-qp}
\end{equation}
Thus the SQP subproblem is precisely the Newton model of the active nonlinear
KKT equations. With the exact Lagrangian Hessian, its primal solution $p_k$ is
the Newton correction and its dual solution $\widehat y_k$ is the updated
multiplier estimate. If the objective is quadratic, the constraints are
affine, and the correct active set is used, this model is exact and reduces to
the one-step calculation in \cref{eq:equality-qp-newton-step}. Otherwise it is
only local and must be rebuilt at the next iterate.

This derivation also explains why the QP contains $\nabla f(x_k)$ rather than
$\nabla_x\mathcal L(x_k,y_k)$ as its linear coefficient. The QP multiplier
$\widehat y_k$ supplies the constraint-gradient term
$C_k\trans\widehat y_k$ in \cref{eq:sqp-qp-stationarity}. Equivalently,
\begin{equation}
  B_kp_k+C_k\trans\Delta y_k
  =-\nabla_x\mathcal L_A(x_k,y_k).
  \label{eq:sqp-qp-newton-equivalence}
\end{equation}
Using the old Lagrangian gradient in the QP objective while retaining the
linearized constraints would count the old constraint-gradient contribution
twice.

\paragraph{The full SQP subproblem.}
In practice, the active set is not known beforehand. SQP therefore retains
all inequalities and lets the QP identify which linearized bounds are active.
For a proposed step $p$, the first-order constraint models are
\begin{align}
  h(x_k+p)&\approx h(x_k)+h_x(x_k)p,\\
  g(x_k+p)&\approx g(x_k)+g_x(x_k)p.
  \label{eq:sqp-constraint-linearizations}
\end{align}
The full SQP subproblem is consequently
\begin{equation}
  \begin{aligned}
    \minimize_p\quad
      &\nabla f(x_k)\trans p+\frac12p\trans B_kp\\
    \subjectto\quad
      &h(x_k)+h_x(x_k)p=0,\\
      &g(x_k)+g_x(x_k)p\leq0.
  \end{aligned}
  \label{eq:sqp-general-qp}
\end{equation}
The constant term $f(x_k)$ is omitted because it does not affect the
minimizer. The objective is a local quadratic model, while the constraints are
local affine models. The QP therefore chooses a direction that balances
objective reduction against restoration of feasibility.

\paragraph{Curvature model.}
The exact Lagrangian Hessian in
\cref{eq:sqp-lagrangian-hessian} has the expanded form
\begin{equation}
  B_k
  =\nabla^2 f(x_k)
   +\sum_i(\lambda_k)_i\nabla^2h_i(x_k)
   +\sum_j(\mu_k)_j\nabla^2g_j(x_k).
  \label{eq:sqp-lagrangian-hessian-expanded}
\end{equation}
It includes constraint curvature because the equations being solved are the
KKT equations, not merely $\nabla f(x)=0$. An active nonlinear constraint may
therefore contribute substantially to the local curvature model.

Exact second derivatives may be expensive, and the exact Lagrangian Hessian
can be indefinite away from a solution. Common alternatives include a
BFGS-type quasi-Newton approximation, a Gauss--Newton matrix for least-squares
objectives, and a regularized model $B_k+\rho_kI$. If the inner solver requires
a convex QP, $B_k$ is chosen or modified so that $B_k\succeq0$.

\paragraph{Obtaining the QP multipliers.}
The vectors $\lambda_k$ and $\mu_k$ are estimates, not fixed problem data.
They may be initialized to zero and are subsequently updated from the dual
solution of each QP. These estimates enter the Lagrangian Hessian in
\cref{eq:sqp-lagrangian-hessian-expanded}, so the primal step and dual
multipliers are both essential outputs of the subproblem. The QP KKT
conditions include
\begin{align}
  B_kp_k+\nabla f(x_k)
  +h_x(x_k)\trans\widehat\lambda_k
  +g_x(x_k)\trans\widehat\mu_k&=0,
  \label{eq:sqp-qp-dual-stationarity}\\
  \widehat\mu_k&\geq0,
  &
  (\widehat\mu_k)_j
  \bigl(g_j(x_k)+\nabla g_j(x_k)\trans p_k\bigr)&=0.
  \label{eq:sqp-qp-dual-complementarity}
\end{align}
Thus a primal--dual QP algorithm computes
$\widehat\lambda_k$ and $\widehat\mu_k$ together with $p_k$. Active-set
methods obtain them from the equality-constrained KKT system for the final
working set, primal--dual interior-point methods carry them as explicit
iterates, and bound-form operator-splitting solvers maintain a dual variable
for the constraint $\ell_k\leq A_kp\leq u_k$. Solver sign conventions differ:
for an upper-bound inequality of the form used here, its dual must be mapped to
the convention $\widehat\mu_k\geq0$.

Most full-featured QP solver interfaces expose these dual variables, but not
every interface does. An approximate first-order method may also return duals
whose accuracy differs from that of the primal step. Moreover, sign
conventions, presolve, and internal scaling affect how the reported duals map
back to the original constraints. An SQP implementation should therefore
request the dual solution and check its documented ordering, signs, and
scaling.

If only $p_k$ is available, the multipliers can often be recovered cheaply.
Let $J_{\mathcal A}$ contain the gradients of the QP inequalities judged active
at $p_k$. Set the inactive multipliers to zero and solve
\begin{equation}
  \begin{bmatrix}
    h_x(x_k)\trans & J_{\mathcal A}\trans
  \end{bmatrix}
  \begin{bmatrix}
    \widehat\lambda_k\\
    (\widehat\mu_k)_{\mathcal A}
  \end{bmatrix}
  =-\bigl(B_kp_k+\nabla f(x_k)\bigr).
  \label{eq:sqp-recover-qp-multipliers}
\end{equation}
Under LICQ, the active constraint gradients are linearly independent, so the
QP multipliers are unique and this system can be solved by QR or a
least-squares factorization. Recovery is less reliable when LICQ fails, the
active gradients are nearly dependent, or the active set depends sensitively
on the feasibility tolerance. A solver that directly supplies a consistent
primal--dual solution is therefore preferable for SQP.

\paragraph{Iteration, globalization, and stopping.}
Using the QP multipliers $\widehat\lambda_k$ and $\widehat\mu_k$, a consistent
primal--dual update with a common damping parameter $0<\alpha_k\leq1$ is
\begin{equation}
  x_{k+1}=x_k+\alpha_kp_k,
  \qquad
  \lambda_{k+1}
  =\lambda_k+\alpha_k(\widehat\lambda_k-\lambda_k),
  \qquad
  \mu_{k+1}
  =\mu_k+\alpha_k(\widehat\mu_k-\mu_k).
  \label{eq:sqp-primal-dual-iteration}
\end{equation}
For a full step, $\alpha_k=1$, the multiplier estimates are replaced directly
by the QP multipliers. At the new point, the algorithm reevaluates $f$, $h$,
$g$, and their derivatives, constructs a new QP, and repeats.

Near a regular solution, the full Newton step $\alpha_k=1$ is typically
appropriate. Farther away, however, the local QP may predict objective
improvement while worsening nonlinear feasibility, or vice versa. A line
search can compare trial points using a merit function such as
\begin{equation}
  M_\eta(x)
  =f(x)+\eta\norm{h(x)}_1
   +\eta\norm{\max\{0,g(x)\}}_1
  \label{eq:sqp-general-merit-function}
\end{equation}
and reduce $\alpha_k$ until this combined measure of objective and constraint
violation decreases sufficiently. Trust-region SQP instead restricts
$\norm{p}$ and changes the trust-region radius according to the agreement
between predicted and actual improvement.

If the linearized constraints are inconsistent, the QP may be infeasible even
though the nonlinear problem is feasible. Practical implementations address
this using slack variables, elastic penalties, or a separate feasibility-
restoration step. The algorithm terminates only when the KKT residuals are
small: $h(x_k)$ and the positive parts of $g(x_k)$ measure primal
infeasibility, $\nabla_x\mathcal L(x_k,\lambda_k,\mu_k)$ measures
stationarity, the negative parts of $\mu_k$ measure dual infeasibility, and
$\mu_k\odot g(x_k)$ measures complementarity. Under standard regularity and
second-order assumptions,
exact-Hessian SQP has Newton-like rapid local convergence. Quasi-Newton
variants obtain similar behavior when their curvature approximations become
accurate.

\section{Stochastic Gradient Descent}

Many data-driven problems minimize an average loss
\begin{equation}
  F(x)=\frac1N\sum_{i=1}^N f_i(x),
  \label{eq:finite-sum-objective}
\end{equation}
where $f_i$ is the loss associated with one data point. Full gradient descent
computes all $N$ component gradients before taking a step. When $N$ is large,
it can be faster to use a randomly selected data point or mini-batch and take
more frequent, less accurate steps.

Let $g_k$ denote the resulting gradient estimate. Stochastic gradient descent
(SGD) uses the update
\begin{equation}
  x_{k+1}=x_k-\alpha_k g_k,
  \label{eq:sgd-update}
\end{equation}
where $\alpha_k>0$ is the learning rate. A standard analysis assumes that the
estimate is correct on average and has bounded noise:
\begin{align}
  \E[g_k\mid x_k]&=\nabla F(x_k),
  \label{eq:sgd-unbiased}\\
  \E\!\left[
    \norm{g_k-\nabla F(x_k)}_2^2\mid x_k
  \right]&\leq\sigma^2.
  \label{eq:sgd-bounded-variance}
\end{align}
The first condition says that repeated random samples at the same point would
produce the full gradient on average. The second prevents the estimates from
being arbitrarily noisy. Averaging a larger mini-batch usually reduces this
variance, at the cost of more computation per update.

\subsection{Why SGD Makes Progress}

Suppose that $F$ is bounded below by $F_\star$ and has an $L$-Lipschitz
gradient. Applying the descent lemma to the stochastic step
$d=-\alpha_k g_k$ gives
\begin{equation}
  F(x_{k+1})
  \leq
  F(x_k)-\alpha_k\nabla F(x_k)\trans g_k
  +\frac{L\alpha_k^2}{2}\norm{g_k}_2^2.
  \label{eq:sgd-smoothness-step}
\end{equation}
Taking conditional expectation and using
\cref{eq:sgd-unbiased,eq:sgd-bounded-variance} yields
\begin{equation}
  \E[F(x_{k+1})\mid x_k]
  \leq
  F(x_k)
  -\alpha_k\left(1-\frac{L\alpha_k}{2}\right)
   \norm{\nabla F(x_k)}_2^2
  +\frac{L\alpha_k^2}{2}\sigma^2.
  \label{eq:sgd-expected-descent}
\end{equation}
This inequality displays the essential competition in SGD. The term involving
$\norm{\nabla F(x_k)}_2^2$ produces progress, while the final term is the price
of gradient noise. The useful term is proportional to $\alpha_k$, whereas the
noise term is proportional to $\alpha_k^2$. Thus sufficiently small learning
rates make the noise less important than the expected descent.

For example, if $\alpha_k=\alpha\leq1/L$ is constant, then
\cref{eq:sgd-expected-descent} simplifies to
\begin{equation}
  \E[F(x_{k+1})\mid x_k]
  \leq
  F(x_k)-\frac{\alpha}{2}\norm{\nabla F(x_k)}_2^2
  +\frac{L\alpha^2}{2}\sigma^2.
  \label{eq:sgd-constant-step-descent}
\end{equation}
Summing this inequality for $k=0,\ldots,K-1$ and using
$F(x_K)\geq F_\star$ gives
\begin{equation}
  \frac1K\sum_{k=0}^{K-1}
  \E\!\left[\norm{\nabla F(x_k)}_2^2\right]
  \leq
  \frac{2(F(x_0)-F_\star)}{\alpha K}
  +L\alpha\sigma^2.
  \label{eq:sgd-average-gradient-bound}
\end{equation}
The first term decreases as more iterations are taken, but the second does not.
Consequently, a constant learning rate usually brings SGD to a neighborhood of
a stationary point and then causes it to fluctuate there. A smaller learning
rate produces a smaller final neighborhood but slower initial progress.

If the number of iterations $K$ is known in advance, choosing
$\alpha$ proportional to $1/\sqrt{K}$ balances the two terms in
\cref{eq:sgd-average-gradient-bound} and gives an
$O(1/\sqrt{K})$ bound on the average squared gradient norm. Alternatively, one
may gradually decrease the learning rate. The classical conditions
\begin{equation}
  \sum_{k=0}^{\infty}\alpha_k=\infty,
  \qquad
  \sum_{k=0}^{\infty}\alpha_k^2<\infty
  \label{eq:sgd-step-conditions}
\end{equation}
capture the desired balance: the steps remain large enough to keep moving, but
the accumulated effect of the noise is finite. Under the assumptions above and
standard regularity conditions, convergence theorems show that suitable
averages of the gradient norm approach zero and that convergent limit points
are stationary. For convex objectives, related arguments give convergence in
objective value; stronger convexity permits faster rates such as $O(1/k)$ with
suitable decreasing learning rates.

\subsection{Practical Interpretation}

The analysis explains the main design choices. Larger mini-batches reduce
$\sigma^2$ but cost more per iteration. Larger learning rates move quickly but
amplify noise. Decreasing the learning rate allows coarse progress early and
more accurate refinement later. In practice, one monitors training and
validation losses, gradient norms when available, and sensitivity to the
learning-rate schedule rather than relying only on the iteration count.

Many extensions modify the basic SGD update. Momentum smooths successive
directions; RMSProp and Adam adapt the effective learning rate of each
coordinate; and AdamW combines Adam with separately applied weight decay.
These methods are important in modern machine learning, but they introduce
additional state and tuning choices. The central idea remains the same: use
inexpensive noisy gradient information while controlling its noise through
batching and learning rates.

\paragraph{Exercises.}

\begin{exercise}
For $f(x_1,x_2)=\tfrac12(x_1^2+100x_2^2)$, take one gradient step from
$x_0=(1,1)$ with $\alpha=0.01$, and compare it with one Newton step.
\end{exercise}
\begin{proof}[Short solution]
Since $\nabla f(1,1)=(1,100)$, gradient descent gives
$x_1=(0.99,0)$. The Hessian is $H=\operatorname{diag}(1,100)$, so the Newton
step is $-H^{-1}\nabla f=(-1,-1)$ and reaches the minimizer $(0,0)$ exactly.
\end{proof}

\begin{exercise}
Apply gradient descent to $f(x)=\tfrac{a}{2}x^2$, where $a>0$. For which
constant step sizes does every nonzero iterate decrease the objective?
\end{exercise}
\begin{proof}[Short solution]
The update is $x_{k+1}=(1-a\alpha)x_k$. Objective decrease requires
$|1-a\alpha|<1$, which is equivalent to $0<\alpha<2/a$.
\end{proof}

\begin{exercise}
Suppose a convex objective is $L$-smooth with $L=10$ and
$\norm{x_0-x^\star}_2=2$. Using \cref{eq:gd-convex-complexity}, how many
iterations suffice to guarantee $f(x_k)-f(x^\star)\leq0.01$?
\end{exercise}
\begin{proof}[Short solution]
The bound gives
\begin{equation}
  k\geq\frac{10(2^2)}{2(0.01)}=2000.
\end{equation}
This is a worst-case guarantee, so actual convergence may be faster.
\end{proof}

\begin{exercise}
For a $2$-strongly convex and $10$-smooth function, what contraction factor is
guaranteed by gradient descent with step size $1/L$?
\end{exercise}
\begin{proof}[Short solution]
The factor is $1-\mu/L=1-2/10=0.8$. Thus
$\Delta_k\leq0.8^k\Delta_0$.
\end{proof}

\begin{exercise}
Use Armijo backtracking on $f(x)=x^2$ at $x=1$ with direction
$p=-\nabla f(1)=-2$, initial step $\alpha=1$, $c=10^{-4}$, and reduction
factor $\rho=1/2$. Which step is first accepted?
\end{exercise}
\begin{proof}[Short solution]
For $\alpha=1$, the trial point is $-1$ and the objective does not decrease,
so the step is rejected. For $\alpha=1/2$, the trial point is $0$, which
satisfies the Armijo condition. The first accepted step is therefore
$\alpha=1/2$.
\end{proof}

\begin{exercise}
For the quadratic function
$f(x)=\tfrac12x\trans Qx-b\trans x$ with $Q\succ0$, show that one Newton step
from any starting point reaches the minimizer.
\end{exercise}
\begin{proof}[Short solution]
Here $\nabla f(x)=Qx-b$ and $\nabla^2f(x)=Q$. Hence
\begin{equation}
  x^+=x-Q^{-1}(Qx-b)=Q^{-1}b,
\end{equation}
which is the unique minimizer.
\end{proof}

\begin{exercise}
Project $z=(3,4)$ onto the Euclidean ball
$C=\{x:\norm{x}_2\leq2\}$.
\end{exercise}
\begin{proof}[Short solution]
Since $\norm{z}_2=5>2$, scale $z$ to radius $2$:
\begin{equation}
  \Pi_C(z)=\frac{2}{5}(3,4)=\left(\frac65,\frac85\right).
\end{equation}
\end{proof}

\begin{exercise}
Minimize $f(x)=\tfrac12(x-3)^2$ over $C=[0,1]$. Verify the projected-gradient
fixed-point condition at the solution for any $\alpha>0$.
\end{exercise}
\begin{proof}[Short solution]
The constrained minimizer is $x^\star=1$ and $\nabla f(1)=-2$. Therefore
$x^\star-\alpha\nabla f(x^\star)=1+2\alpha>1$, whose projection onto $[0,1]$
is again $1$. Thus $\Pi_C(x^\star-\alpha\nabla f(x^\star))=x^\star$.
\end{proof}

\begin{exercise}
Apply soft thresholding to $z=(3,0.4,-2)$ with threshold
$\alpha\lambda=1$.
\end{exercise}
\begin{proof}[Short solution]
Each magnitude is reduced by $1$ and any negative result is replaced by zero.
The output is $(2,0,-1)$.
\end{proof}

\begin{exercise}
Consider the penalty problem
\begin{equation}
  \min_x\ \frac12x^2+\frac{\rho}{2}(x-1)^2.
\end{equation}
Find its minimizer and show how it approaches the constraint $x=1$.
\end{exercise}
\begin{proof}[Short solution]
Setting the derivative to zero gives
$x+\rho(x-1)=0$, so $x_\rho=\rho/(1+\rho)$. The violation is
$1-x_\rho=1/(1+\rho)$, which tends to zero as $\rho\to\infty$.
\end{proof}

\begin{exercise}
The constrained problem $\min_{x\geq0}x$ has solution $x^\star=0$. Find the
minimizer of its logarithmic-barrier objective $x-\mu\log x$ for $x>0$.
\end{exercise}
\begin{proof}[Short solution]
The derivative is $1-\mu/x$. Setting it to zero gives $x_\mu=\mu$. Hence the
barrier solution remains strictly positive and approaches $0$ as
$\mu\downarrow0$.
\end{proof}

\begin{exercise}
Let
$f_1(x)=\tfrac12(x-1)^2$ and $f_2(x)=\tfrac12(x+1)^2$, and define
$F=(f_1+f_2)/2$. If one component is sampled uniformly, show that its gradient
is an unbiased estimate of $\nabla F(x)$.
\end{exercise}
\begin{proof}[Short solution]
The sampled gradients are $x-1$ and $x+1$. Their mean is
\begin{equation}
  \frac12(x-1)+\frac12(x+1)=x=\nabla F(x),
\end{equation}
so the estimate is unbiased.
\end{proof}

\begin{exercise}
Suppose independent single-sample gradient estimates have variance
$\sigma^2$. What is the variance of their average over a mini-batch of size
$B$?
\end{exercise}
\begin{proof}[Short solution]
Independence makes the variances add, while averaging contributes a factor
$1/B^2$. Thus the mini-batch variance is $B\sigma^2/B^2=\sigma^2/B$.
\end{proof}

\begin{exercise}
In \cref{eq:sgd-average-gradient-bound}, let
$F(x_0)-F_\star=1$, $L=1$, $\sigma^2=4$, $\alpha=0.1$, and $K=1000$.
Evaluate the bound.
\end{exercise}
\begin{proof}[Short solution]
Substitution gives
\begin{equation}
  \frac{2}{0.1(1000)}+(1)(0.1)(4)=0.02+0.4=0.42.
\end{equation}
Most of the bound is the constant-step noise floor $L\alpha\sigma^2$.
\end{proof}

\paragraph{Chapter summary.}
Gradient methods are inexpensive, curvature-based methods reduce iteration
count, projection and proximal operators handle constraints or nonsmooth terms,
and stochastic methods scale to data-driven objectives. In practice, reliable
optimization depends as much on step sizes, scaling, and residual checks as on
the choice of algorithm.

\part{From Calculus of Variations to Optimal Control}
\chapter{Optimization in Function Spaces}
\label{chap:function-space-optimization}
\chapteroverview{Functionals, first variations, Euler--Lagrange equations, mechanical systems, and constrained dynamics.}

The variables in the previous chapters were finite-dimensional vectors. Many
problems in mechanics and control instead ask us to choose an entire curve. A
robot trajectory, for example, is a function
\begin{equation}
  q:[t_0,t_f]\to\R^n
\end{equation}
that specifies a configuration at every time. After choosing a time grid, its
samples can be converted into an ordinary finite-dimensional vector. For
example, choose $N+1$ equally spaced times
\begin{equation}
  t_i=t_0+ih,
  \qquad
  h=\frac{t_f-t_0}{N},
  \qquad i=0,\ldots,N,
\end{equation}
and write $q_i=q(t_i)\in\R^n$. The entire sampled trajectory is then represented
by the stacked vector
\begin{equation}
  \mathbf q
  =
  \begin{bmatrix}
    q_0\\q_1\\\vdots\\q_N
  \end{bmatrix}
  \in\R^{n(N+1)}.
  \label{eq:stacked-trajectory-vector}
\end{equation}
Thus, sampling a two-joint robot at $101$ times produces a vector containing
$2(101)=202$ scalar entries. Derivatives and integrals can also be replaced by
finite-dimensional approximations, such as
\begin{equation}
  \dot q(t_i)\approx\frac{q_{i+1}-q_i}{h},
  \qquad
  \int_{t_0}^{t_f}f(t,q,\dot q)\,dt
  \approx
  h\sum_{i=0}^{N-1}
  f\left(t_i,q_i,\frac{q_{i+1}-q_i}{h}\right).
\end{equation}
Here $f$ is simply a generic scalar-valued function being integrated along
the trajectory.
The resulting problem is an optimization problem over the vector $\mathbf q$,
of the kind studied in Chapter~1. Function-space optimization instead studies
the curve before choosing the grid. This viewpoint reveals the differential
equations and boundary conditions hidden inside the discretized problem; a
grid can then be chosen later for numerical computation.

The main idea is the same as a directional derivative. We perturb a candidate
curve by a small function, compute the first-order change in the objective, and
require that change to vanish for every admissible perturbation. The resulting
Euler--Lagrange equation becomes an equation of motion when the objective is a
mechanical action. Equality constraints then add reaction forces through
Lagrange multipliers, while unilateral contact adds inequality and
complementarity conditions.

\section{From Vectors to Functions}

\subsection{Functions as Optimization Variables}

A common objective assigns one scalar to an entire curve:
\begin{equation}
  \mathcal J[q]
  =\Phi\bigl(q(t_0),q(t_f)\bigr)
  +\int_{t_0}^{t_f}L\bigl(t,q(t),\dot q(t)\bigr)\,dt.
  \label{eq:functional-general-form}
\end{equation}
The function $L$ measures an instantaneous cost, and $\Phi$ measures an
endpoint cost. The square brackets emphasize that the input to $\mathcal J$ is
a function rather than a finite vector.

\begin{definition}[Functional]
A \emph{functional} is a map that takes a function as input and returns a
scalar. The integral in \cref{eq:functional-general-form} is the main example
used in this chapter.
\end{definition}

The allowed curves form the admissible set. For example, fixed endpoints impose
\begin{equation}
  q(t_0)=q_0,
  \qquad
  q(t_f)=q_f.
  \label{eq:fixed-endpoint-conditions}
\end{equation}
We assume enough smoothness for the displayed derivatives and integrals to
exist. Continuously differentiable curves are sufficient for the variation,
and twice differentiable curves are used when the final equation contains
$\ddot q$.

\begin{example}[Minimum-energy interpolation]
Consider
\begin{equation}
  \minimize_q
  \quad
  \frac12\int_{t_0}^{t_f}\norm{\dot q(t)}_2^2\,dt
  \qquad
  \subjectto
  \qquad
  q(t_0)=q_0,
  \quad q(t_f)=q_f.
  \label{eq:min-energy-curve-problem}
\end{equation}
Rapid motion is expensive because the integrand is quadratic in velocity. In
the next section, we use the Euler--Lagrange equation to show that the optimal
curve moves at constant velocity from $q_0$ to $q_f$.
\end{example}

\subsection{Admissible Perturbations and First Variations}

Choose a function $\eta(t)$ and form the perturbed curve
\begin{equation}
  q_\epsilon(t)=q(t)+\epsilon\eta(t),
  \label{eq:curve-perturbation}
\end{equation}
where $\epsilon\in\R$ is small. The function $\eta$ is the function-space
analogue of a direction $d$ in $f(x+\epsilon d)$. If both endpoints are fixed,
the perturbation must satisfy
\begin{equation}
  \eta(t_0)=0,
  \qquad
  \eta(t_f)=0.
  \label{eq:fixed-endpoint-perturbation}
\end{equation}

\begin{definition}[First variation]
The first variation of $\mathcal J$ at $q$ in the direction $\eta$ is
\begin{equation}
  \delta\mathcal J[q;\eta]
  =
  \left.
  \frac{d}{d\epsilon}
  \mathcal J[q+\epsilon\eta]
  \right|_{\epsilon=0}.
  \label{eq:first-variation-definition}
\end{equation}
It is the directional derivative of a functional.
\end{definition}

For the full functional in \cref{eq:functional-general-form}, the endpoint
cost contributes through both endpoint perturbations. The chain rule gives
\begin{align}
  \delta\mathcal J[q;\eta]
  &={}
  \nabla_1\Phi\bigl(q(t_0),q(t_f)\bigr)\trans\eta(t_0)
  +\nabla_2\Phi\bigl(q(t_0),q(t_f)\bigr)\trans\eta(t_f)\\
  &\quad+
  \int_{t_0}^{t_f}
  \left(
    \frac{\partial L}{\partial q}\trans\eta
    +
    \frac{\partial L}{\partial\dot q}\trans\dot\eta
  \right)dt.
  \label{eq:first-variation-before-parts}
\end{align}
The endpoint cost $\Phi(q(t_0),q(t_f))$ has two vector arguments: the initial
configuration and the final configuration. The notation $\nabla_1\Phi$ means
the gradient with respect to the first argument while the second argument is
held fixed. Similarly, $\nabla_2\Phi$ is the gradient with respect to the
second argument while the first is held fixed. More explicitly, if
$x=q(t_0)$ and $y=q(t_f)$, then small changes $\Delta x$ and $\Delta y$ give
\begin{equation}
  \Phi(x+\Delta x,y+\Delta y)
  =
  \Phi(x,y)
  +\nabla_1\Phi(x,y)\trans\Delta x
  +\nabla_2\Phi(x,y)\trans\Delta y
  +o\bigl(\norm{\Delta x}+\norm{\Delta y}\bigr).
\end{equation}
Thus $\nabla_1\Phi$ measures how the endpoint cost changes when the initial
point moves, whereas $\nabla_2\Phi$ measures how it changes when the final
point moves. In the perturbed curve $q+\epsilon\eta$, these endpoint changes
are $\Delta x=\epsilon\eta(t_0)$ and
$\Delta y=\epsilon\eta(t_f)$, which explains the first two terms in
\cref{eq:first-variation-before-parts}. All derivatives of $L$ are evaluated
at $(t,q(t),\dot q(t))$. Integration by parts moves the derivative away from
$\eta$:
\begin{align}
  \delta\mathcal J[q;\eta]
  &={}
  \nabla_1\Phi\trans\eta(t_0)
  +\nabla_2\Phi\trans\eta(t_f)
  +
  \left[
    \frac{\partial L}{\partial\dot q}\trans\eta
  \right]_{t_0}^{t_f}
  +
  \int_{t_0}^{t_f}
  \left(
    \frac{\partial L}{\partial q}
    -
    \frac{d}{dt}\frac{\partial L}{\partial\dot q}
  \right)\trans\eta\,dt.
  \label{eq:first-variation-after-parts}
\end{align}
This separates changes at the endpoints from changes to the interior of the
curve. When both endpoints are fixed, $\eta(t_0)=\eta(t_f)=0$, so the
contributions from $\Phi$ and the integration-by-parts boundary term all
vanish. When an endpoint is free, these terms instead produce a natural
boundary condition, as discussed below.

\section{The Euler--Lagrange Equation}

\subsection{Derivation from the First Variation}

With fixed endpoints, all endpoint terms in
\cref{eq:first-variation-after-parts} vanish. A necessary condition for a
local minimum is therefore
\begin{equation}
  \int_{t_0}^{t_f}
  \left(
    \frac{\partial L}{\partial q}
    -
    \frac{d}{dt}\frac{\partial L}{\partial\dot q}
  \right)\trans\eta\,dt=0
  \label{eq:first-variation-zero-integral}
\end{equation}
for every smooth perturbation that vanishes at the endpoints.

The coefficient of $\eta$ must vanish at every time. Otherwise, one could
choose a perturbation supported near a time where the coefficient is nonzero
and pointing in the same direction, making the integral nonzero.

\begin{theorem}[Euler--Lagrange equation]
If a sufficiently smooth curve $q^\star$ is a local minimizer of
\begin{equation}
  \mathcal J[q]=\int_{t_0}^{t_f}L(t,q,\dot q)\,dt
\end{equation}
with fixed endpoints, then
\begin{equation}
  \boxed{
  \frac{d}{dt}\frac{\partial L}{\partial\dot q}
  -\frac{\partial L}{\partial q}=0.}
  \label{eq:euler-lagrange-vector}
\end{equation}
\end{theorem}

Like $\nabla f(x^\star)=0$, this condition is necessary but not automatically
sufficient. It identifies stationary curves. Additional convexity or
second-order information is needed to guarantee a minimum.

For $q=(q_1,\ldots,q_n)\trans$, the vector equation represents one scalar
equation per coordinate:
\begin{equation}
  \frac{d}{dt}\frac{\partial L}{\partial\dot q_i}
  -\frac{\partial L}{\partial q_i}=0,
  \qquad i=1,\ldots,n.
  \label{eq:euler-lagrange-components}
\end{equation}

\subsection{Minimum-Energy and Shortest-Path Examples}

For \cref{eq:min-energy-curve-problem},
\begin{equation}
  \frac{\partial L}{\partial q}=0,
  \qquad
  \frac{\partial L}{\partial\dot q}=\dot q.
\end{equation}
Thus $\ddot q=0$, so $q(t)=a+bt$. The endpoints give
\begin{equation}
  q(t)=q_0+
  \frac{t-t_0}{t_f-t_0}(q_f-q_0).
  \label{eq:straight-line-interpolation}
\end{equation}
The minimum-energy curve is a straight line traversed at constant velocity.

A related objective measures geometric path length:
\begin{equation}
  \mathcal J[q]=\int_{t_0}^{t_f}\norm{\dot q(t)}_2\,dt.
\end{equation}
Away from points where $\dot q=0$,
\begin{equation}
  \frac{\partial L}{\partial\dot q}
  =\frac{\dot q}{\norm{\dot q}_2}.
\end{equation}
The Euler--Lagrange equation says that this unit direction is constant. Hence a
shortest unconstrained path is also a straight line, although its speed along
that line is not fixed by path length alone.

\section{Boundary Conditions}

\subsection{Fixed and Free Endpoints}

The Euler--Lagrange equation governs the interior of the curve, but it does not
by itself determine what must happen at the endpoints. Those conditions come
from the endpoint terms in \cref{eq:first-variation-after-parts}. Collecting
the coefficients of the two endpoint perturbations gives
\begin{align}
  \delta\mathcal J_{\mathrm{end}}
  &={}
  \left(
    \nabla_1\Phi
    -\left.\frac{\partial L}{\partial\dot q}\right|_{t=t_0}
  \right)\trans\eta(t_0)\\
  &\quad+
  \left(
    \nabla_2\Phi
    +\left.\frac{\partial L}{\partial\dot q}\right|_{t=t_f}
  \right)\trans\eta(t_f).
  \label{eq:general-endpoint-variation}
\end{align}
Here $\nabla_1\Phi$ and $\nabla_2\Phi$ are evaluated at
$(q(t_0),q(t_f))$, and the derivatives of $L$ are evaluated along the curve.
The opposite signs at $t_0$ and $t_f$ come from evaluating the
integration-by-parts boundary term at the upper endpoint minus its value at
the lower endpoint.

If an endpoint is fixed, its perturbation is zero, so its coefficient imposes
no additional condition. If an endpoint is free, its perturbation is
arbitrary, so the corresponding coefficient must vanish. Thus a free initial
or final endpoint satisfies, respectively,
\begin{align}
  \left.\frac{\partial L}{\partial\dot q}\right|_{t=t_0}
  &=\nabla_1\Phi,
  &
  \left.\frac{\partial L}{\partial\dot q}\right|_{t=t_f}
  &=-\nabla_2\Phi.
  \label{eq:natural-free-endpoint-condition}
\end{align}
These are the \emph{natural boundary conditions}. If there is no endpoint
cost, then the corresponding derivative of $\Phi$ is zero. In particular, a
free final endpoint satisfies
$\left.\partial L/\partial\dot q\right|_{t=t_f}=0$.

For the common special case with a fixed initial point and only a terminal
cost,
\begin{equation}
  \mathcal J[q]
  =\phi(q(t_f))+
  \int_{t_0}^{t_f}L(t,q,\dot q)\,dt,
\end{equation}
a free final endpoint therefore satisfies
\begin{equation}
  \left.\frac{\partial L}{\partial\dot q}\right|_{t=t_f}
  +\nabla\phi(q(t_f))=0.
  \label{eq:terminal-cost-boundary-condition}
\end{equation}

\begin{example}[Free endpoint with terminal penalty]
Consider
\begin{equation}
  \mathcal J[q]
  =\frac{r}{2}(q(t_f)-q_d)^2
  +\frac12\int_{t_0}^{t_f}\dot q(t)^2\,dt,
\end{equation}
with fixed $q(t_0)=q_0$ and free $q(t_f)$. The interior equation is
$\ddot q=0$, while
\begin{equation}
  \dot q(t_f)+r(q(t_f)-q_d)=0.
\end{equation}
The terminal velocity and endpoint error balance each other.
\end{example}

\subsection{Endpoint Constraints and Transversality}

An endpoint may be neither fixed nor completely free. For example, a mobile
robot may be required to finish somewhere on a docking line, without
specifying its exact location on that line. Similarly, a robot arm may need to
place its end effector somewhere on a target surface. In both cases the final
configuration is allowed to move, but only in directions that preserve the
endpoint constraint.

Suppose the final configuration must satisfy
\begin{equation}
  c(q(t_f))=0,
  \qquad
  c:\R^n\to\R^m.
  \label{eq:constrained-final-endpoint}
\end{equation}
The $m$ components of $c$ describe the target surface, or more generally the
target manifold. Let
\begin{equation}
  J_c(q)=\frac{\partial c(q)}{\partial q}\in\R^{m\times n}
\end{equation}
be its Jacobian. To determine which endpoint perturbations are admissible,
apply the perturbed curve $q_\epsilon=q+\epsilon\eta$ to the constraint:
\begin{align}
  0
  &=c(q_\epsilon(t_f))\\
  &=c(q(t_f))
    +\epsilon J_c(q(t_f))\eta(t_f)
    +o(\epsilon).
\end{align}
Because the original endpoint already satisfies $c(q(t_f))=0$, feasibility to
first order requires
\begin{equation}
  J_c(q(t_f))\eta(t_f)=0.
  \label{eq:endpoint-tangent-perturbation}
\end{equation}
Thus the allowed endpoint perturbations belong to the null space of $J_c$.
Geometrically, they are tangent to the target surface. Directions in the range
of $J_c\trans$ are normal to that surface and are not feasible endpoint
motions.

Now collect the coefficient of the final endpoint perturbation in
\cref{eq:general-endpoint-variation}:
\begin{equation}
  b_f
  =
  \nabla_2\Phi
  +\left.\frac{\partial L}{\partial\dot q}\right|_{t=t_f}.
  \label{eq:final-endpoint-variation-coefficient}
\end{equation}
The endpoint contribution is $b_f\trans\eta(t_f)$. Stationarity does not
require $b_f=0$, because $\eta(t_f)$ is no longer arbitrary. It only requires
\begin{equation}
  b_f\trans\eta(t_f)=0
  \qquad
  \text{for every }\eta(t_f)\text{ satisfying }
  J_c\eta(t_f)=0.
\end{equation}
In words, $b_f$ must be orthogonal to every tangent direction. It must
therefore be a normal vector. Assuming that $J_c$ has full row rank, every
normal vector can be written as $J_c\trans\nu$ for some multiplier
$\nu\in\R^m$. Hence the final endpoint condition is
\begin{equation}
  \boxed{
  \left.\frac{\partial L}{\partial\dot q}\right|_{t=t_f}
  +\nabla_2\Phi
  +J_c(q(t_f))\trans\nu=0.}
  \label{eq:transversality-multiplier}
\end{equation}
Together, \cref{eq:constrained-final-endpoint,eq:transversality-multiplier}
determine both the endpoint and the multiplier. The multiplier supplies the
normal component needed to balance the endpoint variation, just as a KKT
multiplier balances the gradient at a finite-dimensional constrained optimum.

This condition is called a \emph{transversality condition}. The name indicates
that it determines how the stationary curve is allowed to meet the endpoint
surface. To understand its geometry, distinguish two vectors. The terminal
velocity $\dot q(t_f)$ points along the trajectory as it reaches the surface.
By contrast,
\begin{equation}
  b_f=\left.\frac{\partial L}{\partial\dot q}\right|_{t=t_f}
      +\nabla_2\Phi
\end{equation}
measures the first-order change in the objective when the endpoint is moved,
after the interior Euler--Lagrange equation is satisfied. Sliding the endpoint
along the surface is an admissible change, so stationarity requires
$b_f\trans\eta(t_f)=0$ for every tangent displacement. It is therefore $b_f$
that must be normal to the surface, not necessarily $\dot q(t_f)$.

For the squared-speed cost $L=\tfrac12\norm{\dot q}_2^2$ with no endpoint
cost, $b_f=\dot q(t_f)$. In this special case, a nonzero terminal velocity is
normal to the surface, so the path meets it at a right angle. With a weighted
cost $L=\tfrac12\dot q\trans W\dot q$, however, $b_f=W\dot q(t_f)$, and
these two vectors need not point in the same direction. For instance, let
\begin{equation}
  W=\begin{bmatrix}2&1\\1&2\end{bmatrix},
  \qquad c(x,y)=x-1.
\end{equation}
The target line $x=1$ has vertical tangent directions, so transversality
requires the vertical component of $W\dot q$ to vanish:
\begin{equation}
  \dot x(t_f)+2\dot y(t_f)=0.
\end{equation}
A terminal velocity $(\dot x,\dot y)=(1,-\tfrac12)$ satisfies this condition,
since $W\dot q=(\tfrac32,0)\trans$ is normal to the line, even though the
velocity itself is not. An endpoint cost can likewise change $b_f$ through
$\nabla_2\Phi$. Thus transversality expresses stationarity with respect to
allowed endpoint motions; a geometric right angle follows only when $b_f$
is aligned with the terminal velocity.

\begin{example}[A point robot ending on a circle]
Let $q=(x,y)\trans$ be the position of a planar point robot, and minimize
\begin{equation}
  \mathcal J[q]
  =\frac12\int_{t_0}^{t_f}\norm{\dot q(t)}_2^2\,dt
\end{equation}
with a fixed initial position $q(t_0)=q_0$ and a fixed duration
$T=t_f-t_0>0$. The final position may be anywhere on the circle of radius
$R>0$:
\begin{equation}
  c(q(t_f))=x(t_f)^2+y(t_f)^2-R^2=0.
\end{equation}
Here
\begin{equation}
  J_c(q)=2\begin{bmatrix}x&y\end{bmatrix},
  \qquad
  \frac{\partial L}{\partial\dot q}=\dot q,
\end{equation}
and there is no endpoint cost. The transversality condition becomes
\begin{equation}
  \dot q(t_f)+2\nu q(t_f)=0.
\end{equation}
Thus any nonzero terminal velocity must be normal to the circle. To find the
actual optimal endpoint and trajectory, first fix a candidate endpoint $z$
with $\norm{z}_2=R$. The Euler--Lagrange equation gives $\ddot q=0$, so the
stationary curve joining $q_0$ to $z$ is
\begin{equation}
  q_z(t)=q_0+\frac{t-t_0}{T}(z-q_0).
\end{equation}
This curve is a global minimizer for that endpoint: by Cauchy--Schwarz, every
curve with the same endpoints satisfies
\begin{equation}
  \norm{z-q_0}_2^2
  =\left\|\int_{t_0}^{t_f}\dot q(t)\,dt\right\|_2^2
  \leq T\int_{t_0}^{t_f}\norm{\dot q(t)}_2^2\,dt,
\end{equation}
with equality precisely when the velocity is constant. Hence the remaining
problem is simply to choose the closest point on the circle:
\begin{equation}
  \minimize_{\norm{z}_2=R}\quad
  \frac{\norm{z-q_0}_2^2}{2T}.
\end{equation}

Write $\rho=\norm{q_0}_2$. If $\rho>0$, expanding the squared distance gives
$\norm{z-q_0}_2^2=R^2+\rho^2-2z\trans q_0$. This is minimized uniquely by
choosing $z$ in the direction of $q_0$, so
\begin{equation}
  \boxed{
    q_f^\star=\frac{R}{\rho}q_0,
    \qquad
    q^\star(t)=q_0+\frac{t-t_0}{T}
      \left(\frac{R}{\rho}-1\right)q_0.
  }
\end{equation}
The minimum cost and endpoint multiplier are
\begin{equation}
  \mathcal J[q^\star]=\frac{(\rho-R)^2}{2T},
  \qquad
  \nu^\star=\frac{\rho-R}{2RT},
\end{equation}
where the multiplier follows by substituting the constant velocity into
$\dot q(t_f)+2\nu q(t_f)=0$. If $\rho>R$, the robot moves radially inward;
if $0<\rho<R$, it moves radially outward. In either case the path meets the
circle perpendicularly. If $\rho=R$, the robot already lies on the circle,
so the optimal trajectory is constant and the cost is zero.

If $q_0=0$, every point on the circle is equally close. For any unit vector
$u\in\R^2$, an optimal solution is
\begin{equation}
  q_f^\star=Ru,
  \qquad
  q^\star(t)=\frac{R(t-t_0)}{T}u,
  \qquad
  \mathcal J[q^\star]=\frac{R^2}{2T},
  \qquad
  \nu^\star=-\frac{1}{2T}.
\end{equation}
Thus the optimal endpoint is unique unless the robot starts at the center.
The closest-point argument establishes global optimality, whereas
transversality alone only supplies a stationarity condition.
\end{example}

A constrained initial endpoint is handled in the same way. If
$c_0(q(t_0))=0$, then for some multiplier $\nu_0$,
\begin{equation}
  \nabla_1\Phi
  -\left.\frac{\partial L}{\partial\dot q}\right|_{t=t_0}
  +J_{c_0}(q(t_0))\trans\nu_0=0.
\end{equation}
The minus sign before $\partial L/\partial\dot q$ comes from the lower endpoint
of the integration-by-parts boundary term.

\section{Mechanical Systems from the Action}

\subsection{Mechanical Lagrangians and Generalized Forces}

For a conservative mechanical system, define
\begin{equation}
  L(q,\dot q)=T(q,\dot q)-V(q),
  \label{eq:mechanical-lagrangian}
\end{equation}
where $T$ is kinetic energy and $V$ is potential energy. The functional
\begin{equation}
  \mathcal S[q]=\int_{t_0}^{t_f}L(q,\dot q)\,dt
\end{equation}
is called the action. Physical motion makes the first variation of the action
zero. It need not minimize the action globally; stationarity is the key fact.

An applied generalized force $\tau(t)$ performs virtual work
$\tau\trans\eta$. Including this work gives the forced equation
\begin{equation}
  \boxed{
  \frac{d}{dt}\frac{\partial L}{\partial\dot q}
  -\frac{\partial L}{\partial q}=\tau.}
  \label{eq:forced-euler-lagrange}
\end{equation}

\begin{example}[Mass on a spring]
For
\begin{equation}
  L(q,\dot q)=\frac12m\dot q^2-\frac12kq^2,
\end{equation}
the forced Euler--Lagrange equation is
\begin{equation}
  m\ddot q+kq=\tau.
\end{equation}
If viscous damping exerts force $-d\dot q$ and an actuator applies $u$, then
\begin{equation}
  m\ddot q+d\dot q+kq=u.
  \label{eq:forced-damped-spring-mass}
\end{equation}
\end{example}

\begin{example}[Simple pendulum]
For a point mass $m$ on a massless rod of length $\ell$,
\begin{equation}
  T=\frac12m\ell^2\dot\theta^2,
  \qquad
  V=mg\ell(1-\cos\theta).
\end{equation}
Therefore
\begin{equation}
  m\ell^2\ddot\theta+mg\ell\sin\theta=\tau.
  \label{eq:pendulum-dynamics}
\end{equation}
The nonlinear sine term comes directly from gravitational potential energy.
\end{example}

\subsection{Robot Manipulator Equations}

Let $q\in\R^n$ collect the joint coordinates of a robot manipulator. Its
kinetic energy can be written as
\begin{equation}
  T(q,\dot q)=\frac12\dot q\trans M(q)\dot q,
  \label{eq:robot-kinetic-energy}
\end{equation}
where the mass matrix satisfies $M(q)=M(q)\trans\succ0$. If $V(q)$ is the
gravitational potential energy, then the mechanical Lagrangian is
\begin{equation}
  L(q,\dot q)
  =\frac12\dot q\trans M(q)\dot q-V(q).
  \label{eq:robot-lagrangian}
\end{equation}
Because $M(q)$ is symmetric, differentiation with respect to velocity gives
\begin{equation}
  \frac{\partial L}{\partial\dot q}=M(q)\dot q.
\end{equation}
Differentiating this expression with respect to time gives
\begin{equation}
  \frac{d}{dt}\frac{\partial L}{\partial\dot q}
  =M(q)\ddot q+\dot M(q,\dot q)\dot q.
\end{equation}
Here $\dot M$ is the derivative of the mass matrix along the trajectory:
\begin{equation}
  \dot M(q,\dot q)
  =\sum_{j=1}^n
  \frac{\partial M(q)}{\partial q_j}\dot q_j.
\end{equation}
The derivative with respect to coordinate $q_i$ is
\begin{equation}
  \frac{\partial L}{\partial q_i}
  =\frac12\dot q\trans
  \frac{\partial M(q)}{\partial q_i}\dot q
  -\frac{\partial V(q)}{\partial q_i}.
\end{equation}
Consequently, the velocity-dependent terms can be collected into a vector
$h(q,\dot q)$ whose $i$th component is
\begin{equation}
  h_i(q,\dot q)
  =[\dot M(q,\dot q)\dot q]_i
  -\frac12\dot q\trans
  \frac{\partial M(q)}{\partial q_i}\dot q.
  \label{eq:robot-velocity-vector}
\end{equation}
Substitution into \cref{eq:forced-euler-lagrange} gives
\begin{equation}
  \boxed{
  M(q)\ddot q+h(q,\dot q)+\nabla V(q)=\tau.}
  \label{eq:robot-manipulator-dynamics}
\end{equation}
The terms represent inertial force, velocity-dependent force, gravitational
force, and applied joint torque, respectively. It is customary to introduce a
Coriolis matrix $C(q,\dot q)$ and gravity vector $g(q)$ satisfying
\begin{equation}
  C(q,\dot q)\dot q=h(q,\dot q),
  \qquad
  g(q)=\nabla V(q).
\end{equation}
The dynamics then take the familiar manipulator form
\begin{equation}
  M(q)\ddot q+C(q,\dot q)\dot q+g(q)=\tau.
  \label{eq:standard-manipulator-form}
\end{equation}
The matrix $C$ is not unique because only the product $C\dot q$ appears in the
dynamics.

\begin{remark}[Christoffel symbols and energy balance]
A standard choice constructs $C$ from the Christoffel symbols of the mass
matrix. Define
\begin{equation}
  \Gamma_{ijk}(q)
  =\frac12\left(
    \frac{\partial M_{ij}}{\partial q_k}
    +\frac{\partial M_{ik}}{\partial q_j}
    -\frac{\partial M_{jk}}{\partial q_i}
  \right),
  \qquad
  C_{ij}(q,\dot q)=\sum_{k=1}^n\Gamma_{ijk}(q)\dot q_k.
\end{equation}
This construction satisfies $C(q,\dot q)\dot q=h(q,\dot q)$ and makes
$\dot M-2C$ skew-symmetric. Hence, for every vector $v$,
\begin{equation}
  v\trans(\dot M-2C)v=0.
\end{equation}
Taking $v=\dot q$ gives
\begin{equation}
  \frac12\dot q\trans\dot M\dot q
  =\dot q\trans C\dot q.
\end{equation}
Therefore the derivative of kinetic energy simplifies to
\begin{equation}
  \frac{d}{dt}\left(\frac12\dot q\trans M\dot q\right)
  =\dot q\trans\bigl(M\ddot q+C\dot q\bigr).
\end{equation}
Using the manipulator equation
$M\ddot q+C\dot q+g=\tau$, this becomes
\begin{equation}
  \frac{dT}{dt}
  =\dot q\trans(\tau-g).
\end{equation}
Moreover, because $g(q)=\nabla V(q)$,
\begin{equation}
  \frac{dV}{dt}
  =\nabla V(q)\trans\dot q
  =g(q)\trans\dot q.
\end{equation}
Adding the kinetic- and potential-energy rates cancels the gravity terms:
\begin{equation}
  \frac{d}{dt}(T+V)=\dot q\trans\tau.
\end{equation}
The velocity-dependent terms thus redistribute kinetic energy without creating
or destroying it. This identity is the key step in the energy balance derived
next and is also useful in Lyapunov analyses of robot controllers.
\end{remark}

\subsection{Energy Balance}

For a general Lagrangian $L(t,q,\dot q)$, define the associated energy
function
\begin{equation}
  E_L(t,q,\dot q)
  =\frac{\partial L}{\partial\dot q}\trans\dot q-L.
  \label{eq:lagrangian-energy-function}
\end{equation}
Differentiate along a trajectory. Applying the product and chain rules gives
\begin{align}
  \frac{dE_L}{dt}
  &={}
  \left(
    \frac{d}{dt}\frac{\partial L}{\partial\dot q}
  \right)\trans\dot q
  +\frac{\partial L}{\partial\dot q}\trans\ddot q
  -\frac{dL}{dt}\\
  &={}
  \left(
    \frac{d}{dt}\frac{\partial L}{\partial\dot q}
  \right)\trans\dot q
  +\frac{\partial L}{\partial\dot q}\trans\ddot q\\
  &\quad-
  \left(
    \frac{\partial L}{\partial t}
    +\frac{\partial L}{\partial q}\trans\dot q
    +\frac{\partial L}{\partial\dot q}\trans\ddot q
  \right)\\
  &=
  \left(
    \frac{d}{dt}\frac{\partial L}{\partial\dot q}
    -\frac{\partial L}{\partial q}
  \right)\trans\dot q
  -\frac{\partial L}{\partial t}.
  \label{eq:lagrangian-energy-derivative}
\end{align}
The two terms containing $\ddot q$ cancel. By the forced Euler--Lagrange
equation, the expression in parentheses is the generalized force $\tau$.
Therefore
\begin{equation}
  \boxed{
  \frac{dE_L}{dt}
  =\tau\trans\dot q-\frac{\partial L}{\partial t}.}
  \label{eq:general-lagrangian-energy-balance}
\end{equation}

Now consider the mechanical Lagrangian $L=T-V$ with
$T=\tfrac12\dot q\trans M(q)\dot q$ and no explicit time dependence. Since
$T$ is quadratic in $\dot q$,
\begin{equation}
  \frac{\partial L}{\partial\dot q}\trans\dot q
  =(M(q)\dot q)\trans\dot q=2T.
\end{equation}
Hence the Lagrangian energy is the usual total mechanical energy:
\begin{equation}
  E_L=2T-(T-V)=T+V=:E.
\end{equation}
Because $\partial L/\partial t=0$, the general identity reduces to
\begin{equation}
  \boxed{
  \frac{dE}{dt}=\tau\trans\dot q.}
  \label{eq:mechanical-energy-balance}
\end{equation}
The right-hand side is instantaneous mechanical power: generalized force
multiplied by generalized velocity. Thus an unforced conservative system,
$\tau=0$, satisfies $dE/dt=0$ and conserves energy.

More generally, suppose the applied force consists of an actuator input $u$
and viscous damping:
\begin{equation}
  \tau=u-D\dot q,
  \qquad D\succeq0.
\end{equation}
Then
\begin{equation}
  \frac{dE}{dt}
  =u\trans\dot q-\dot q\trans D\dot q.
\end{equation}
The first term is power supplied by the actuator, while the second is
nonnegative power dissipated by damping. In particular, if $u=0$, then
\begin{equation}
  \frac{dE}{dt}=-\dot q\trans D\dot q\leq0,
\end{equation}
so the mechanical energy cannot increase.

\section{Advanced Constrained Dynamics and Physics Simulation}

\begin{remark}[For advanced readers]
This section is particularly intended for readers interested in a modern,
optimization-based viewpoint of dynamics and physics simulation. The preceding
sections developed the core function-space and Euler--Lagrange ideas. Here we
go further by treating reaction forces, unilateral contact, and impacts as
unknowns determined by constrained linear systems, KKT conditions, and
complementarity problems. These formulations are central to many contemporary
robotics and rigid-body simulators, where each time step requires solving an
optimization-related algebraic problem in addition to integrating the
equations of motion.
\end{remark}

\subsection{Forward Dynamics and Time-Stepping Simulation}

\emph{Forward dynamics} computes the acceleration from the current
configuration $q$, velocity $\dot q$, and applied generalized force $\tau$.
For the unconstrained mechanical system in
\cref{eq:robot-manipulator-dynamics}, this means solving
\begin{equation}
  M(q)\ddot q=\tau-h(q,\dot q)-\nabla V(q).
\end{equation}
Since $M(q)\succ0$, the acceleration is uniquely determined:
\begin{equation}
  \ddot q=f_{\mathrm{FD}}(q,\dot q,\tau)
  :=M(q)^{-1}\bigl(\tau-h(q,\dot q)-\nabla V(q)\bigr).
  \label{eq:unconstrained-forward-dynamics}
\end{equation}
In computation, one solves the linear system with $M(q)$ rather than forming
its inverse. This is an instantaneous calculation: it gives the acceleration
at the supplied state, not the entire future trajectory.

A physics simulator turns this calculation into motion by numerically
integrating the first-order system
\begin{equation}
  \dot q=v,
  \qquad
  \dot v=f_{\mathrm{FD}}(q,v,\tau).
\end{equation}
Choose a time step $\Delta t>0$, and let $(q_k,v_k)$ approximate the state at
$t_k=t_0+k\Delta t$. Given the applied force $\tau_k$, first compute
\begin{equation}
  a_k=f_{\mathrm{FD}}(q_k,v_k,\tau_k).
\end{equation}
\emph{Explicit Euler} then updates both position and velocity using the old
state:
\begin{equation}
  q_{k+1}=q_k+\Delta t\,v_k,
  \qquad
  v_{k+1}=v_k+\Delta t\,a_k.
  \label{eq:forward-dynamics-explicit-euler}
\end{equation}
\emph{Semi-implicit Euler} instead updates velocity first and uses that new
velocity to advance position:
\begin{equation}
  v_{k+1}=v_k+\Delta t\,a_k,
  \qquad
  q_{k+1}=q_k+\Delta t\,v_{k+1}.
  \label{eq:forward-dynamics-semi-implicit-euler}
\end{equation}
The acceleration is still evaluated at $(q_k,v_k)$; this update requires no
additional nonlinear solve. Both rules are first-order approximations, and
their difference is the velocity used in the position update. Repeating the
acceleration calculation and state update produces a simulated trajectory.

With constraints, the same simulation workflow applies, but the acceleration
calculation must also determine the unknown reaction forces. The following
subsection explains this coupled forward-dynamics solve for equality
constraints; later we extend it to unilateral contact and impacts.

\subsection{Dynamics with Equality Constraints}

\subsubsection{Configuration Constraints and Reaction Forces}

Suppose the configuration is restricted by $p$ smooth, time-independent
equality constraints
\begin{equation}
  c(q)=0,
  \qquad c:\R^n\to\R^p,
  \label{eq:holonomic-constraint}
\end{equation}
with $p\leq n$. Such constraints are called \emph{holonomic} because they are
expressed directly in terms of the configuration. Define the constraint
Jacobian
\begin{equation}
  J(q)=\frac{\partial c(q)}{\partial q}\in\R^{p\times n}.
\end{equation}
We assume that $J(q)$ has full row rank on the constraint surface, so the
$p$ constraints are locally independent.

The constraints can be incorporated directly into the variational principle.
Because $c(q(t))=0$ must hold at every time, its multiplier is not a single
vector but a time-dependent function
\begin{equation}
  \lambda:[t_0,t_f]\to\R^p.
\end{equation}
Define the multiplier-augmented, also called \emph{extended}, Lagrangian and
its action by
\begin{align}
  L_{\mathrm{aug}}(q,\dot q,\lambda)
  &=L(q,\dot q)+\lambda\trans c(q),
  \label{eq:constraint-augmented-lagrangian}\\
  \mathcal S_{\mathrm{aug}}[q,\lambda]
  &=\int_{t_0}^{t_f}
    L_{\mathrm{aug}}(q(t),\dot q(t),\lambda(t))\,dt.
  \label{eq:constraint-augmented-action}
\end{align}
The plus sign is a convention chosen so that the resulting reaction force is
$+J\trans\lambda$. Replacing $\lambda$ by $-\lambda$ would give the equivalent
opposite convention.

Both $q$ and $\lambda$ are now varied independently. A perturbation
$q+\epsilon\eta$ and $\lambda+\epsilon\mu$ gives
\begin{align}
  \delta\mathcal S_{\mathrm{aug}}
  ={}&
  \int_{t_0}^{t_f}
  \left(
    \frac{\partial L}{\partial q}\trans\eta
    +
    \frac{\partial L}{\partial\dot q}\trans\dot\eta
    +
    \lambda\trans J(q)\eta
    +
    \mu\trans c(q)
  \right)dt.
  \label{eq:constraint-augmented-first-variation}
\end{align}
For an applied generalized force $\tau$, the Lagrange--d'Alembert principle
adds its virtual work to the action variation:
\begin{equation}
  \delta\mathcal S_{\mathrm{aug}}
  +\int_{t_0}^{t_f}\tau\trans\eta\,dt=0.
\end{equation}
This condition must hold for every independent pair of perturbations
$\eta$ and $\mu$, with $\eta(t_0)=\eta(t_f)=0$. Setting $\eta=0$ and allowing
$\mu(t)$ to vary arbitrarily recovers the original constraint:
\begin{equation}
  c(q(t))=0
  \qquad\text{for every }t.
\end{equation}
Next, setting $\mu=0$ and integrating the $\dot\eta$ term by parts eliminates
the endpoint terms. Since the interior perturbation $\eta(t)$ is arbitrary,
its coefficient must vanish. Together, the two variations give
\begin{equation}
  \frac{d}{dt}\frac{\partial L}{\partial\dot q}
  -\frac{\partial L}{\partial q}
  =\tau+J(q)\trans\lambda,
  \qquad
  c(q)=0.
  \label{eq:constrained-euler-lagrange-full}
\end{equation}
The multiplier variation enforces feasibility, while the configuration
variation supplies the equation of motion with the additional force
$J(q)\trans\lambda$.

To interpret this force geometrically, consider a virtual displacement that
is compatible with the original constraint. Unlike the unrestricted
configuration perturbations used in the augmented formulation, a compatible
displacement must be tangent to the constraint surface. Linearizing
$c(q)=0$ gives
\begin{equation}
  J(q)\eta=0.
\end{equation}
The tangent space is therefore the null space of $J(q)$. An ideal constraint
reaction does no virtual work along these tangent directions, so it lies in
their orthogonal complement: the normal space, which is the range of
$J(q)\trans$. This is exactly the force produced by the multiplier:
\begin{equation}
  f_c=J(q)\trans\lambda,
  \qquad \lambda\in\R^p.
\end{equation}
For the mechanical Lagrangian $L=T-V$, the constrained Euler--Lagrange
equation consequently takes the manipulator form
\begin{equation}
  M(q)\ddot q+h(q,\dot q)+\nabla V(q)
  =\tau+J(q)\trans\lambda,
  \label{eq:equality-constrained-dynamics}
\end{equation}
where the multiplier $\lambda$ determines the magnitude of each constraint
reaction. The reaction force does no virtual work along any compatible
displacement:
\begin{equation}
  (J\trans\lambda)\trans\eta=\lambda\trans J\eta=0.
\end{equation}
Likewise, along a feasible trajectory it contributes no instantaneous power,
because $J\dot q=0$ implies
\begin{equation}
  (J\trans\lambda)\trans\dot q=0.
\end{equation}

The position constraint must hold for all time. Differentiating it once gives
the velocity constraint
\begin{equation}
  \frac{d}{dt}c(q(t))
  =J(q)\dot q=0.
  \label{eq:equality-velocity-constraint}
\end{equation}
Differentiating once more gives the acceleration constraint
\begin{equation}
  \frac{d}{dt}\bigl(J(q)\dot q\bigr)
  =J(q)\ddot q+\dot J(q,\dot q)\dot q=0.
  \label{eq:equality-acceleration-constraint}
\end{equation}
Here
\begin{equation}
  \dot J(q,\dot q)
  =\sum_{i=1}^n\frac{\partial J(q)}{\partial q_i}\dot q_i
\end{equation}
is the derivative of the constraint Jacobian along the trajectory. The term
$\dot J\dot q$ accounts for the curvature of the constraint surface; even a
constant-speed feasible motion may require a normal acceleration.

At a given state $(q,\dot q)$ and applied force $\tau$, neither the
acceleration $\ddot q$ nor the reaction multiplier $\lambda$ is known.
\emph{They must be determined together}: the reaction force
$J(q)\trans\lambda$ changes the acceleration, while the requirement that this
acceleration satisfy the constraint determines the reaction force. Computing
the unconstrained acceleration first and treating it as the final answer
would generally violate the constraint.

To solve for these coupled unknowns, define the nonconstraint generalized
force
\begin{equation}
  r(q,\dot q,\tau)
  =\tau-h(q,\dot q)-\nabla V(q).
\end{equation}
Then the dynamics read $M\ddot q-J\trans\lambda=r$. These $n$ equations
alone do not determine the $n+p$ unknown components of $(\ddot q,\lambda)$.
The $p$ acceleration constraints supply the remaining equations, producing
the coupled linear saddle-point system
\begin{equation}
  \begin{bmatrix}
    M(q)&-J(q)\trans\\
    J(q)&0
  \end{bmatrix}
  \begin{bmatrix}
    \ddot q\\\lambda
  \end{bmatrix}
  =
  \begin{bmatrix}
    r(q,\dot q,\tau)\\
    -\dot J(q,\dot q)\dot q
  \end{bmatrix}.
  \label{eq:constrained-dynamics-kkt-system}
\end{equation}
Solving this system returns both the constraint-compatible acceleration and
the multiplier that generates the required reaction force. They are two
parts of the same forward-dynamics solution, even if one is eliminated
algebraically during the solve.

To see why this is called a saddle-point system, hold $q$, $\dot q$, and
$\tau$ fixed and regard the acceleration $a$ as the decision variable.
Consider the equality-constrained quadratic problem
\begin{equation}
  \begin{aligned}
    \minimize_a\quad
      &\frac12a\trans M(q)a-r(q,\dot q,\tau)\trans a\\
    \subjectto\quad
      &J(q)a+\dot J(q,\dot q)\dot q=0.
  \end{aligned}
  \label{eq:constrained-acceleration-qp}
\end{equation}
Without the constraint, its minimizer is the free acceleration
$a=M^{-1}r$. The constraint selects the feasible acceleration closest to this
free acceleration in the metric defined by $M$.

Using the sign convention consistent with
\cref{eq:equality-constrained-dynamics}, its Lagrangian is
\begin{equation}
  \mathcal L_a(a,\lambda)
  =\frac12a\trans Ma-r\trans a
  -\lambda\trans\bigl(Ja+\dot J\dot q\bigr).
\end{equation}
The primal variable $a$ is minimized, while the unrestricted multiplier
$\lambda$ is maximized. The first-order conditions are
\begin{align}
  \nabla_a\mathcal L_a
  &=Ma-r-J\trans\lambda=0,\\
  \nabla_\lambda\mathcal L_a
  &=-\bigl(Ja+\dot J\dot q\bigr)=0.
\end{align}
Setting $a=\ddot q$ produces exactly
\cref{eq:constrained-dynamics-kkt-system}. The solution is therefore a saddle
point of $\mathcal L_a$: it is a minimum in the acceleration direction and a
maximum in the multiplier direction. 

We now use a Schur-complement reduction to solve
\cref{eq:constrained-dynamics-kkt-system} by eliminating the acceleration,
solving a smaller system for the multipliers, and then recovering the
acceleration by back-substitution.

\paragraph{Schur-complement reduction.}
For a general block linear system
\begin{equation}
  \begin{bmatrix}A&B\\C&D\end{bmatrix}
  \begin{bmatrix}x\\y\end{bmatrix}
  =\begin{bmatrix}b\\d\end{bmatrix},
\end{equation}
assume that $A$ is invertible. The first block equation,
$Ax+By=b$, gives $x=A^{-1}(b-By)$. Substituting into the second block
equation, $Cx+Dy=d$, yields
\begin{equation}
  \underbrace{\bigl(D-CA^{-1}B\bigr)}_{S_A}\,y
  =d-CA^{-1}b.
\end{equation}
The matrix $S_A=D-CA^{-1}B$ is the \emph{Schur complement of $A$}.
If $S_A$ is also invertible, solve this reduced system for $y$, then recover
$x=A^{-1}(b-By)$ by back-substitution. Thus elimination reduces the size of
the linear solve without changing the original equations or uncoupling the
underlying problem.

We now apply this procedure to
\cref{eq:constrained-dynamics-kkt-system}, with the block identification
\begin{equation}
  A=M,\quad B=-J\trans,\quad C=J,\quad D=0,
  \qquad
  x=\ddot q,\quad y=\lambda,\quad b=r,\quad d=-\dot J\dot q.
\end{equation}
The acceleration and reaction force remain coupled, but we need not solve
the full block system at once. We can \emph{eliminate the acceleration} by
expressing it in terms of the multiplier and substituting into the constraint.
Throughout this calculation, $q$, $\dot q$, and $\tau$ are fixed, so $M$, $J$,
$\dot J$, and $r$ are known.

The first block equation, $M\ddot q-J\trans\lambda=r$, gives
\begin{equation}
  \ddot q=M^{-1}r+M^{-1}J\trans\lambda.
  \label{eq:constrained-acceleration-back-substitution}
\end{equation}
Substituting into the second block equation,
$J\ddot q=-\dot J\dot q$, removes $\ddot q$ from the solve:
\begin{equation}
  JM^{-1}r+JM^{-1}J\trans\lambda=-\dot J\dot q.
\end{equation}
Define the $p\times p$ matrix
\begin{equation}
  W(q)=J(q)M(q)^{-1}J(q)\trans.
  \label{eq:constraint-schur-complement}
\end{equation}
The reduced system is then
\begin{equation}
  \boxed{W\lambda=-\dot J\dot q-JM^{-1}r.}
  \label{eq:constraint-multiplier-schur}
\end{equation}
Once $\lambda$ is found, back-substitution into
\cref{eq:constrained-acceleration-back-substitution} recovers $\ddot q$.
This is an algebraic rearrangement of the same coupled problem, not a separate
assumption about the reaction force.

Consistent with the general formula, the Schur complement of $M$ is
$0-JM^{-1}(-J\trans)=W$. Physically, $W$ is the
\emph{constraint-space inverse inertia}: $W\lambda$ is the change in
constraint acceleration caused by the reaction force $J\trans\lambda$.

The reduced system has a unique solution under our assumptions that
$M\succ0$ and $J$ has full row rank. To see why, recall that
$W=JM^{-1}J\trans$. Since $M^{-1}$ is symmetric, $W$ is symmetric as well.
Moreover, full row rank means that the rows of $J$ are linearly independent:
no nonzero combination of them can vanish. Thus, for any nonzero
$z\in\R^p$, the vector $u=J\trans z$ is nonzero. Because $M^{-1}$ is
positive definite,
\begin{equation}
  z\trans Wz
  =(J\trans z)\trans M^{-1}(J\trans z)
  =u\trans M^{-1}u>0.
\end{equation}
Therefore $W$ is symmetric positive definite and hence invertible. The
reduced equation determines a unique multiplier $\lambda$; back-substitution
then determines a unique acceleration $\ddot q$.

For numerical computation, the $p\times p$ reduced system can be solved using
a Cholesky factorization of $W$. This reduction is particularly useful when
there are far fewer constraints than coordinates ($p\ll n$). There is no need
to form $M^{-1}$ explicitly. Instead, compute the free acceleration
$a_{\mathrm{free}}\in\R^n$ and the matrix $Y\in\R^{n\times p}$ by solving
\begin{equation}
  Ma_{\mathrm{free}}=r,
  \qquad MY=J\trans.
\end{equation}
Form $W=JY$, solve for the multiplier, and recover the acceleration:
\begin{equation}
  W\lambda=-\dot J\dot q-Ja_{\mathrm{free}},
  \qquad
  \ddot q=a_{\mathrm{free}}+Y\lambda.
\end{equation}

Since $a_{\mathrm{free}}=M^{-1}r$ and $Y=M^{-1}J\trans$, the recovered
acceleration has a simple physical interpretation:
\begin{equation}
  \ddot q
  =a_{\mathrm{free}}+M^{-1}J\trans\lambda.
\end{equation}
The first term is the acceleration without constraint reactions; the second
is the correction produced by the reaction force $J\trans\lambda$. Together
they satisfy $J\ddot q=-\dot J\dot q$. Equivalently, the quadratic formulation
above shows that $\ddot q$ is the feasible acceleration closest to
$a_{\mathrm{free}}$ in the mass-matrix metric, which measures a correction
$\Delta a$ by $\Delta a\trans M\Delta a$.

\paragraph{Constraint drift.}
In exact continuous dynamics, consistent initial data satisfying
$c(q(0))=0$ and $J(q(0))\dot q(0)=0$ remain feasible. A numerical integrator,
or \emph{time-stepping method}, is an algorithm that approximates a continuous
trajectory at discrete times
\begin{equation}
  t_k=t_0+kh,
\end{equation}
where $h>0$ is the time-step length. Given approximations $q_k\approx q(t_k)$
and $v_k\approx\dot q(t_k)$, the algorithm computes an acceleration $a_k$ from
the equations of motion and then advances the state. For example, the explicit
Euler method uses
\begin{equation}
  q_{k+1}=q_k+hv_k,
  \qquad
  v_{k+1}=v_k+ha_k.
\end{equation}
These updates replace continuous motion over an interval by a finite
approximation and therefore introduce a small discretization error at each
step.

Consequently, even if $a_k$ satisfies the acceleration constraint
$Ja_k=-\dot Jv_k$, the updated state need not satisfy
$c(q_{k+1})=0$ and $J(q_{k+1})v_{k+1}=0$ exactly. The resulting position and
velocity errors can accumulate over many time steps, gradually moving the
computed trajectory away from the constraint manifold. This phenomenon is
called \emph{constraint drift}. A smaller time step or a higher-order
integrator generally reduces the local error, but does not by itself guarantee
that the constraints remain exactly satisfied.

\paragraph{Projection.}
One remedy is to correct the state after each time step. Given a tentative
position $\widetilde q_{k+1}$, position projection finds a nearby point on the
constraint manifold, for example by solving
\begin{equation}
  \minimize_q
  \quad \frac12\norm{q-\widetilde q_{k+1}}_2^2
  \qquad\subjectto\qquad c(q)=0.
\end{equation}
Because $c$ may be nonlinear, this projection usually requires a small
iterative solve. After obtaining the corrected position $q_{k+1}$, the
tentative velocity $\widetilde v_{k+1}$ can be projected onto the tangent
space by requiring
\begin{equation}
  J(q_{k+1})v_{k+1}=0.
\end{equation}
The correction that changes velocity as little as possible in the mass metric
is
\begin{equation}
  v_{k+1}
  =\widetilde v_{k+1}
  -M^{-1}J\trans
   \bigl(JM^{-1}J\trans\bigr)^{-1}J\widetilde v_{k+1},
  \label{eq:velocity-constraint-projection}
\end{equation}
where $M$ and $J$ are evaluated at $q_{k+1}$. Thus the same Schur-complement
matrix $W=JM^{-1}J\trans$ that determines reaction forces also appears in
velocity projection. Projection removes drift directly, but adds one or more
constraint solves to every time step.

\paragraph{Baumgarte stabilization.}
Projection corrects the position and velocity \emph{after} a time step.
Baumgarte stabilization takes a different approach: it adjusts the
acceleration \emph{before} the update so that the motion gradually corrects
any constraint error. To understand the idea, let
\begin{equation}
  e(t)=c(q(t))
\end{equation}
measure how far the current configuration is from satisfying the constraint.
Here $e$ has one component per constraint; it is a constraint residual, not
necessarily a geometric distance. Its derivative $\dot e=J\dot q$ measures
how quickly that residual is changing.

The usual acceleration constraint imposes $\ddot e=0$. If $e$ and $\dot e$
are initially zero, this keeps them zero in exact continuous motion. But it
does not repair an error that has already appeared. In fact, $\ddot e=0$
gives
\begin{equation}
  e(t)=e(t_0)+(t-t_0)\dot e(t_0).
\end{equation}
A position error therefore persists, and a velocity error makes it grow.
This is why merely enforcing the acceleration constraint is not enough to
remove numerical drift.

The remedy is to ask the error to behave like a damped spring returning to
its equilibrium at zero. Baumgarte stabilization prescribes
\begin{equation}
  \ddot e+2\alpha\dot e+\beta^2e=0,
  \qquad \alpha,\beta>0.
  \label{eq:baumgarte-error-dynamics}
\end{equation}
Equivalently, $\ddot e=-2\alpha\dot e-\beta^2e$. The position term
$-\beta^2e$ pulls the error back toward zero, while the velocity term
$-2\alpha\dot e$ damps its motion to reduce overshoot. For example, for the
scalar constraint $c(q)=q=0$, a positive position error with zero velocity
requests a negative acceleration; a negative error requests a positive one.
These are corrections to the \emph{constraint error}, not extra actuator
commands or a replacement for the equations of motion.

With $\alpha,\beta>0$, this continuous-time error equation makes both $e$
and $\dot e$ decay to zero. The parameter $\beta$ sets the natural frequency
of the correction, and $\alpha/\beta$ is its damping ratio. A useful starting
choice is $\alpha=\beta$: the error dynamics are then critically damped,
with no oscillatory modes. Smaller damping ($\alpha<\beta$) allows decaying
oscillations, while larger damping ($\alpha>\beta$) gives an overdamped
response.

To use this idea in the dynamics solve, express the error derivatives in
terms of the robot state:
\begin{equation}
  \dot e=J\dot q,
  \qquad
  \ddot e=J\ddot q+\dot J\dot q.
\end{equation}
Substituting these into \cref{eq:baumgarte-error-dynamics} and isolating
$J\ddot q$ gives the stabilized acceleration constraint:
\begin{equation}
  J\ddot q
  =-\dot J\dot q-2\alpha J\dot q-\beta^2c(q).
  \label{eq:baumgarte-acceleration-constraint}
\end{equation}
Compared with $J\ddot q=-\dot J\dot q$, only two feedback terms have been
added: one for position error and one for velocity error. When both errors
are zero, these terms vanish and the original acceleration constraint is
recovered. The equations of motion remain unchanged; acceleration and reaction
force are still solved for together, using this modified constraint.

Unlike projection, stabilization does not place the numerical state exactly
on the constraint surface after every step. It encourages errors to decay,
while discretization may continue to introduce small new errors. Larger gains
request faster correction, but a finite-step integrator must resolve that
faster motion. Gains that are too large for the chosen time step can therefore
cause numerical instability. This separation between fast error correction
and slower physical motion is a source of \emph{stiffness}.

The implementation uses the same block system as before, with only its
constraint right-hand side changed. Let $q_k$ and $v_k$ approximate $q(t_k)$
and $\dot q(t_k)$.
At the beginning of a time step, evaluate
\begin{equation}
  M_k=M(q_k),
  \qquad
  J_k=J(q_k),
  \qquad
  r_k=r(q_k,v_k,\tau_k),
\end{equation}
and
\begin{equation}
  \dot J_k
  =\sum_{i=1}^n
    \frac{\partial J}{\partial q_i}(q_k)(v_k)_i.
\end{equation}
The Baumgarte-stabilized acceleration $a_k$ and reaction multiplier
$\lambda_k$ are obtained from the linear system
\begin{equation}
  \begin{bmatrix}
    M_k&-J_k\trans\\
    J_k&0
  \end{bmatrix}
  \begin{bmatrix}a_k\\\lambda_k\end{bmatrix}
  =
  \begin{bmatrix}
    r_k\\
    -\dot J_kv_k-2\alpha J_kv_k-\beta^2c(q_k)
  \end{bmatrix}.
  \label{eq:baumgarte-discrete-kkt-step}
\end{equation}
One simple update is the semi-implicit Euler rule
\begin{align}
  v_{k+1}&=v_k+ha_k,\\
  q_{k+1}&=q_k+hv_{k+1}.
  \label{eq:baumgarte-semi-implicit-update}
\end{align}
Thus each step consists of evaluating the dynamics, solving one KKT system,
and advancing the state. The name \emph{semi-implicit Euler}, also called
\emph{symplectic Euler}, refers to the order of these two updates. First the
old state $(q_k,v_k)$ is used to compute $a_k$ and the new velocity
$v_{k+1}$. The position is then advanced using this new velocity rather than
the old one. In contrast, fully explicit Euler updates both quantities from
the old state:
\begin{equation}
  v_{k+1}=v_k+ha_k,
  \qquad
  q_{k+1}=q_k+hv_k.
\end{equation}
The semi-implicit method is still straightforward to compute sequentially; no
additional nonlinear equation must be solved for $q_{k+1}$. The word
``implicit'' only reflects that the position update depends on the already
updated velocity. This small change often gives better stability and energy
behavior for mechanical systems.

\subsubsection{Example: A Point Mass on a Circle}

Let $q=(x,y)\trans$ be a mass constrained by
\begin{equation}
  c(q)=x^2+y^2-\ell^2=0,
  \qquad
  J(q)=\begin{bmatrix}2x&2y\end{bmatrix}.
\end{equation}
With $V=mgy$, the equations are
\begin{align}
  m\ddot x&=2x\lambda,\\
  m\ddot y+mg&=2y\lambda,\\
  x^2+y^2&=\ell^2.
\end{align}
Twice differentiating the constraint gives
\begin{equation}
  x\ddot x+y\ddot y+\dot x^2+\dot y^2=0.
\end{equation}

For this system,
\begin{equation}
  \dot J\dot q=2(\dot x^2+\dot y^2),
  \qquad
  J\dot q=2(x\dot x+y\dot y).
\end{equation}
Consequently, a Baumgarte-stabilized time step solves
\begin{equation}
  \begin{bmatrix}
    m&0&-2x_k\\
    0&m&-2y_k\\
    2x_k&2y_k&0
  \end{bmatrix}
  \begin{bmatrix}
    a_{x,k}\\a_{y,k}\\\lambda_k
  \end{bmatrix}
  =
  \begin{bmatrix}
    0\\
    -mg\\
    -2(v_{x,k}^2+v_{y,k}^2)
    -4\alpha(x_kv_{x,k}+y_kv_{y,k})
    -\beta^2(x_k^2+y_k^2-\ell^2)
  \end{bmatrix}.
  \label{eq:circle-baumgarte-kkt-step}
\end{equation}
The semi-implicit update is then
\begin{align}
  v_{x,k+1}&=v_{x,k}+ha_{x,k},
  &v_{y,k+1}&=v_{y,k}+ha_{y,k},\\
  x_{k+1}&=x_k+hv_{x,k+1},
  &y_{k+1}&=y_k+hv_{y,k+1}.
  \label{eq:circle-baumgarte-state-update}
\end{align}
The last row of \cref{eq:circle-baumgarte-kkt-step} is the stabilized
acceleration constraint; the first two rows are Newton's equations with the
radial reaction force.
Substitution yields
\begin{equation}
  \lambda
  =\frac{m}{2\ell^2}
  \left(gy-\dot x^2-\dot y^2\right).
\end{equation}
The physical radial force is $J\trans\lambda$. Alternatively, substituting
$x=\ell\sin\theta$ and $y=-\ell\cos\theta$ eliminates the constraint and
recovers the pendulum equation. Multiplier coordinates retain the Cartesian
variables and make the reaction force explicit.

\subsection{Inequality Constraints and Contact}

The equality-constrained dynamics above used unrestricted multipliers because
an equality reaction may act in either normal direction. Contact is different:
two rigid bodies may push one another apart, but they cannot pull one another
together. This one-sided physical rule leads directly to inequality
constraints, nonnegative multipliers, and KKT complementarity.

\subsubsection{Variational Multipliers and the KKT Analogy}

The following conditions are not restricted to static equilibrium. They are
imposed at each time along a trajectory $q(t)$, and the active contacts and
their forces may change with time. Static contact is the special case
$\dot q=\ddot q=0$.

Let
\begin{equation}
  \phi(q)\in\R^m,
  \qquad
  \phi(q)\geq0,
  \label{eq:contact-gap-constraint}
\end{equation}
collect the gap functions for $m$ possible contacts. All inequalities are
componentwise. For contact $i$,
\begin{equation}
  \phi_i(q)>0 \quad\text{means separation},
  \qquad
  \phi_i(q)=0 \quad\text{means touching},
\end{equation}
whereas $\phi_i(q)<0$ would mean forbidden penetration. Define
\begin{equation}
  J_c(q)=\frac{\partial\phi(q)}{\partial q}\in\R^{m\times n}.
\end{equation}
The transpose of row $i$ of $J_c$ points in the generalized direction that
increases gap $i$.

Let $\lambda_i(t)\geq0$ denote the magnitude of the outward normal force at
contact $i$. With this physical sign convention, introduce the extended
mechanical Lagrangian
\begin{equation}
  L_c(q,\dot q,\lambda)
  =L(q,\dot q)+\lambda\trans\phi(q).
  \label{eq:unilateral-extended-lagrangian}
\end{equation}
The variation of its multiplier term is
\begin{equation}
  \delta_q\bigl(\lambda\trans\phi(q)\bigr)
  =\lambda\trans J_c(q)\eta
  =\bigl(J_c(q)\trans\lambda\bigr)\trans\eta.
  \label{eq:contact-multiplier-virtual-work}
\end{equation}
Thus the multiplier contributes the generalized force
$J_c(q)\trans\lambda$, which points toward increasing separation. Including
an applied generalized force $\tau$ gives
\begin{equation}
  \frac{d}{dt}\frac{\partial L}{\partial\dot q}
  -\frac{\partial L}{\partial q}
  =\tau+J_c(q)\trans\lambda.
  \label{eq:inequality-constrained-euler-lagrange}
\end{equation}

The plus sign in $L_c$ reflects this mechanical force convention. By
comparison, an ordinary minimization problem with $\phi(q)\geq0$ uses
\begin{equation}
  \mathcal L_{\mathrm{KKT}}(q,\mu)
  =F(q)-\mu\trans\phi(q),
  \qquad
  \mu\geq0,
  \label{eq:standard-kkt-gap-sign}
\end{equation}
because the constraint is written as $-\phi(q)\leq0$. The two formulas serve
different purposes: one represents an outward physical force in a stationary
action principle, while the other is the standard KKT Lagrangian for a
minimization problem.

Because contact is unilateral, $\lambda$ is not an unrestricted equality
multiplier. The gap and force instead satisfy
\begin{equation}
  \boxed{
  \phi(q)\geq0,
  \qquad
  \lambda\geq0,
  \qquad
  \lambda\trans\phi(q)=0.}
  \label{eq:position-contact-kkt}
\end{equation}
These conditions are analogous to primal feasibility, dual feasibility, and
complementary slackness. For each contact,
\begin{equation}
  \phi_i(q)>0 \ \Longrightarrow\ \lambda_i=0,
  \qquad
  \lambda_i>0 \ \Longrightarrow\ \phi_i(q)=0.
\end{equation}
Separated bodies therefore exert no contact force, and a positive contact
force can act only when the gap is closed.

Combining the variational equation and complementarity gives the necessary
position-level relations
\begin{equation}
  \frac{d}{dt}\frac{\partial L}{\partial\dot q}
  -\frac{\partial L}{\partial q}
  =\tau+J_c\trans\lambda,
  \qquad
  0\leq\lambda\ \perp\ \phi(q)\geq0.
  \label{eq:unilateral-variational-system}
\end{equation}
The notation $0\leq\lambda\ \perp\ \phi(q)\geq0$ is shorthand for
\emph{complementarity}: for every contact $i$,
\begin{equation}
  \lambda_i\geq0,
  \qquad \phi_i(q)\geq0,
  \qquad \lambda_i\phi_i(q)=0.
\end{equation}
Here $\lambda_i$ is the normal contact force and $\phi_i(q)$ is the gap.
The inequalities say that the force can push but not pull, and that the
bodies cannot penetrate. The zero product says that force and gap cannot
both be positive: an open gap requires zero force, while a positive force
requires a closed gap. Both may be zero, so touching does not necessarily
imply a positive force. The symbol $\perp$ denotes the zero inner product
$\lambda\trans\phi(q)=0$; because both vectors are componentwise nonnegative,
this is equivalent to the zero-product condition for each contact. It does
not refer to perpendicular directions in physical space.

For the mechanical Lagrangian developed earlier, the first relation becomes
\begin{equation}
  M(q)\ddot q+h(q,\dot q)+\nabla V(q)
  =\tau+J_c(q)\trans\lambda.
  \label{eq:unilateral-manipulator-dynamics}
\end{equation}

These equations identify where contact forces may act and the direction in
which they act, but they do not yet determine the magnitude of $\lambda$ at a
closed contact. If $\phi_i(q)=0$, position complementarity only requires
$\lambda_i\geq0$. Moreover, the dynamics contain the $n+m$ unknowns
$(\ddot q,\lambda)$ but provide only $n$ equations. The missing conditions
come from requiring the closed gaps not to accelerate into penetration.
Differentiating the active gap constraints supplies those acceleration-level
conditions and closes the forward-dynamics problem in the next subsection.
Impacts require a separate velocity-level impulse law because velocity may
jump discontinuously.

\subsubsection{From Position Constraints to Acceleration Constraints}

The position-level KKT conditions explain when a force may act, but a forward
dynamics solver must determine the force at the current instant. First,
distinguish an \emph{impact} from persistent contact. At an impact time
$t_{\mathrm{imp}}$, the superscripts $-$ and $+$ denote values immediately
\emph{before} and \emph{after} the impact, respectively:
\begin{equation}
  \dot q^-=\lim_{t\to t_{\mathrm{imp}}^-}\dot q(t),
  \qquad
  \dot q^+=\lim_{t\to t_{\mathrm{imp}}^+}\dot q(t).
\end{equation}
Here $-$ and $+$ identify the two sides of the event; they do not indicate
the sign of the velocity. The configuration $q$ remains continuous, but the
velocity may jump.

The \emph{normal velocity} measures how quickly the two bodies approach or
separate in the direction perpendicular to their contact surface. With our
gap convention, it is the rate of change of the gap:
\begin{equation}
  v_n=\dot\phi(q)=J_c(q)\dot q.
\end{equation}
For each contact, $v_n<0$ means that the gap is shrinking (approach),
$v_n>0$ means that it is growing (separation), and $v_n=0$ means that the
gap is instantaneously unchanged. Zero normal velocity does not require
zero total velocity: the bodies may still slide tangentially along the
contact surface. For example, for a point above a horizontal floor with
$q=(x,y)\trans$ and gap $\phi(q)=y$, we have
$J_c=[0\;1]$ and $v_n=\dot y$; the horizontal velocity $\dot x$ does not
change the gap. Accordingly, $v_n^-=J_c(q)\dot q^-$ and
$v_n^+=J_c(q)\dot q^+$ are the normal velocities immediately before and
after impact. For multiple contacts, these quantities have one component
per contact.

If two bodies reach a
closed gap with inward normal velocity
$v_n^-=J_c(q)\dot q^-<0$, a finite ordinary force cannot stop the penetration
instantaneously. The model instead applies a contact impulse, which changes
the velocity discontinuously from $\dot q^-$ to $\dot q^+$. Saying that the
impact has been \emph{resolved} means that this impulse has already been
computed and applied, leaving a nonpenetrating post-impact velocity
$v_n^+=J_c(q)\dot q^+\geq0$. For a perfectly inelastic impact that remains in
contact, $v_n^+=0$; for a rebounding impact, $v_n^+>0$ and the bodies
immediately separate.

The acceleration-level formulation below concerns the persistent-contact
case after this instantaneous velocity change. Dropping the superscript $+$,
consider a closed contact satisfying
\begin{equation}
  \phi(q)=0,
  \qquad
  v_n=J_c(q)\dot q=0.
\end{equation}
Differentiating the gap twice gives the normal acceleration
\begin{equation}
  a_n
  =\ddot\phi
  =J_c(q)\ddot q+\dot J_c(q,\dot q)\dot q.
  \label{eq:normal-contact-acceleration}
\end{equation}
Because both the gap and normal velocity are zero, a Taylor expansion gives
\begin{equation}
  \phi(t+\Delta t)
  =\frac12a_n\Delta t^2+O(\bigl|\Delta t\bigr|^3).
\end{equation}
Consequently, $a_n<0$ would produce penetration immediately, whereas
$a_n>0$ begins separation. Persistent rigid contact therefore satisfies
\begin{equation}
  \boxed{
  \lambda\geq0,
  \qquad
  a_n\geq0,
  \qquad
  \lambda\trans a_n=0.}
  \label{eq:acceleration-contact-complementarity}
\end{equation}
If $\lambda_i>0$, the force maintains $(a_n)_i=0$. If $(a_n)_i>0$, the bodies
separate and $\lambda_i=0$.

\subsubsection{Eliminating Acceleration: The Linear Complementarity Problem}

Define the generalized force excluding contact by
\begin{equation}
  r(q,\dot q,\tau)
  =\tau-h(q,\dot q)-\nabla V(q).
\end{equation}
At a fixed state $(q,\dot q)$, the equations of motion are
\begin{equation}
  M(q)\ddot q=r(q,\dot q,\tau)+J_c(q)\trans\lambda.
\end{equation}
Both the acceleration $\ddot q$ and the contact force $\lambda$ are unknown.
To \emph{eliminate acceleration} means to solve this equation algebraically
for $\ddot q$ and substitute the result into the normal-acceleration
constraint. Since $M(q)$ is positive definite,
\begin{equation}
  \ddot q=M^{-1}(r+J_c\trans\lambda).
  \label{eq:contact-acceleration-from-force}
\end{equation}
Substituting this expression into
\cref{eq:normal-contact-acceleration} gives
\begin{equation}
  a_n=A\lambda+b,
  \qquad
  A=J_cM^{-1}J_c\trans,
  \qquad
  b=J_cM^{-1}r+\dot J_c\dot q.
  \label{eq:contact-lcp-matrices}
\end{equation}
The vector $b$ is the normal acceleration that would occur with no contact
force, while $A$ maps contact forces to changes in normal acceleration.
Moreover, $A$ is symmetric positive semidefinite because
\begin{equation}
  z\trans Az
  =(J_c\trans z)\trans M^{-1}(J_c\trans z)\geq0.
\end{equation}

Combining $a_n=A\lambda+b$ with the acceleration-level contact conditions in
\cref{eq:acceleration-contact-complementarity} gives the linear
complementarity problem (LCP)
\begin{equation}
  \boxed{
  \lambda\geq0,
  \qquad
  w=A\lambda+b\geq0,
  \qquad
  \lambda\trans w=0.}
  \label{eq:contact-lcp}
\end{equation}
where $w=a_n$ is the resulting normal acceleration. A solution identifies
which contacts remain active and supplies contact forces consistent with
nonpenetration. The important point is that the unknown acceleration no
longer appears in the LCP: the problem is expressed entirely in the
$m$ contact-force variables. Once $\lambda$ is known,
\cref{eq:contact-acceleration-from-force} recovers $\ddot q$.

\subsubsection{Solving the LCP as a Convex Quadratic Program}

Because $A\succeq0$, the LCP in \cref{eq:contact-lcp} can be solved through
the convex quadratic program
\begin{equation}
  \minimize_{\lambda\geq0}
  \quad
  \frac12\lambda\trans A\lambda+b\trans\lambda.
  \label{eq:rigid-contact-convex-qp}
\end{equation}
The gradient of its objective is
\begin{equation}
  \nabla_\lambda
  \left(
    \frac12\lambda\trans A\lambda+b\trans\lambda
  \right)
  =A\lambda+b=w.
\end{equation}
The KKT conditions for the lower bound $\lambda\geq0$ are therefore
\begin{equation}
  \lambda\geq0,
  \qquad
  w\geq0,
  \qquad
  \lambda\trans w=0,
\end{equation}
which are exactly the LCP conditions. Numerically, one supplies the Hessian
$A$, the linear term $b$, and the lower bound $\lambda\geq0$ to a convex QP
solver. The solver determines the active contacts and their forces
simultaneously.

If $A$ is positive definite, the contact force is unique. If $A$ is singular,
as can occur with redundant contact constraints, several multipliers may
produce the same physical acceleration. After selecting a solution
$\lambda$, the simulator recovers $\ddot q$ from
\cref{eq:contact-acceleration-from-force} and advances $(q,\dot q)$ using a
numerical integrator. Thus a forward-dynamics step consists of a contact solve
followed by a state update.

\begin{remark}[Alternative KKT derivation in acceleration space]
The LCP can be understood directly from force balance and acceleration
complementarity, so the following optimization interpretation is not needed
for using it. It provides a second derivation.

The free acceleration, obtained by setting $\lambda=0$, is
\begin{equation}
  \ddot q_{\mathrm{free}}=M(q)^{-1}r(q,\dot q,\tau).
\end{equation}
Among all nonpenetrating accelerations, consider the acceleration closest to
free motion in the mass-matrix metric:
\begin{equation}
  \begin{aligned}
    \minimize_{a\in\R^n}\quad
      &\frac12
       (a-\ddot q_{\mathrm{free}})\trans
       M(q)(a-\ddot q_{\mathrm{free}})\\
    \subjectto\quad
      &J_c(q)a+\dot J_c(q,\dot q)\dot q\geq0.
  \end{aligned}
  \label{eq:contact-acceleration-qp}
\end{equation}
After removing terms independent of $a$, the objective becomes
$\tfrac12a\trans Ma-r\trans a$. Its finite-dimensional Lagrangian is
\begin{equation}
  \mathcal L_a(a,\lambda)
  =\frac12a\trans Ma-r\trans a
  -\lambda\trans\bigl(J_ca+\dot J_c\dot q\bigr),
  \qquad
  \lambda\geq0.
  \label{eq:contact-acceleration-kkt-lagrangian}
\end{equation}
The KKT conditions are
\begin{align}
  Ma-r-J_c\trans\lambda&=0,
  &&\text{stationarity},
  \label{eq:contact-kkt-stationarity}\\
  w=J_ca+\dot J_c\dot q&\geq0,
  &&\text{primal feasibility},\\
  \lambda&\geq0,
  &&\text{dual feasibility},\\
  \lambda\trans w&=0,
  &&\text{complementary slackness}.
  \label{eq:contact-kkt-complementarity}
\end{align}
Stationarity gives $Ma=r+J_c\trans\lambda$, which is the constrained equation
of motion with $a=\ddot q$. Eliminating $a$ from these KKT conditions produces
the same LCP in \cref{eq:contact-lcp}.
\end{remark}

\begin{example}[A mass resting on a floor]
Let $y\geq0$ be the height of a mass above a horizontal floor. At
$y=\dot y=0$,
\begin{equation}
  m\ddot y=-mg+\lambda,
  \qquad
  0\leq\lambda\ \perp\ \ddot y\geq0.
\end{equation}
If $\lambda=0$, gravity gives $\ddot y=-g$, which violates primal
feasibility. Therefore $\lambda>0$, and complementarity requires
$\ddot y=0$. The equation of motion then gives $\lambda=mg$. Thus the normal
force is exactly the KKT multiplier needed to prevent penetration.
\end{example}

\paragraph{Resolving rigid impacts.}
The acceleration-level conditions above describe sustained contact: the gap
is closed and the normal velocity is already zero, $J_c\dot q=0$. They do not
by themselves handle a body that arrives at the contact with inward normal
velocity $J_cv^-<0$. In the ideal rigid-contact model, penetration is
forbidden, so this incoming velocity must change instantaneously. An ordinary
force acting for zero time cannot produce a finite velocity change; instead,
the model introduces a contact impulse $p$, the time integral of a very large
contact force over the short impact interval.

For a frictionless, perfectly inelastic impact, impulse--momentum balance and
the unilateral contact conditions are
\begin{equation}
  M(q)(v^+-v^-)=J_c(q)\trans p,
  \qquad
  0\leq p\ \perp\ J_cv^+\geq0.
\end{equation}
Here $v^-$ and $v^+$ are the velocities immediately before and after impact.
The complementarity condition says that a contact receives an impulse only
when its post-impact separating velocity is zero. From impulse--momentum
balance,
\begin{equation}
  v^+=v^-+M^{-1}J_c\trans p,
\end{equation}
and therefore
\begin{equation}
  J_cv^+=J_cv^-+Ap,
  \qquad
  A=J_cM^{-1}J_c\trans.
\end{equation}
Thus the impulse is found from the velocity-level LCP
\begin{equation}
  0\leq p\ \perp\ J_cv^-+Ap\geq0.
\end{equation}
After this impact solve changes $v^-$ to $v^+$, contacts with
$J_cv^+=0$ are maintained by the acceleration-level contact solve. A rigid
time-stepping simulator may therefore distinguish an impact stage from the
subsequent sustained-contact stage.

\paragraph{Soft contact in modern simulation.}
Many modern simulators instead use a compliant, or \emph{soft}, contact
model. A small amount of penetration is allowed, and contact forces act over
a finite time interval. The incoming normal velocity produces a damping
response, while penetration produces a restoring response. Numerical
integration then changes the velocity continuously across timesteps. Because
the same force law handles the initial collision and the later resting
contact, there is no separate instantaneous reset from $v^-$ to $v^+$ and no
need to solve the rigid impact LCP first.

MuJoCo follows this soft-contact viewpoint. Its reference acceleration uses
the normal velocity and gap in a spring--damper-like form, while a regularized
contact optimization determines the admissible forces. The formulation below
shows how this model is related to the rigid contact-space problem derived
above \cite{todorov2011convex,todorov2012mujoco}.

\subsubsection{Brief Connection to MuJoCo's Soft Contact}

The rigid frictionless contact problem derived above is
\begin{equation}
  \minimize_{\lambda\geq0}
  \quad
  \frac12\lambda\trans A\lambda+b\trans\lambda,
  \qquad
  b=J_cM^{-1}r+\dot J_c\dot q.
  \label{eq:rigid-contact-qp-repeated}
\end{equation}
Here $b$ is the normal acceleration with no contact force, and
$A=J_cM^{-1}J_c\trans$ describes how a contact force changes that
acceleration.

MuJoCo retains this contact-space optimization structure but makes three
changes:
\begin{equation}
  A\longmapsto A+R,
  \qquad
  b\longmapsto a_{\mathrm{free}}-a_{\mathrm{ref}},
  \qquad
  \R_+^m\longmapsto\Omega.
  \label{eq:mujoco-contact-qp-correspondence}
\end{equation}
The resulting dual problem is schematically
\begin{equation}
  \lambda^\star
  =\argmin_{\lambda\in\Omega}
  \left\{
    \frac12\lambda\trans(A+R)\lambda
    +\lambda\trans(a_{\mathrm{free}}-a_{\mathrm{ref}})
  \right\},
  \label{eq:mujoco-convex-contact}
\end{equation}
where the terms correspond directly to the rigid problem:
\begin{itemize}
  \item $A=JM^{-1}J\trans$ is the same inverse inertia in constraint space.
  \item $a_{\mathrm{free}}=JM^{-1}r+\dot J\dot q$ is the constraint-space
  acceleration with $\lambda=0$. For the frictionless normal constraints
  above, it is exactly the vector $b$.
  \item $\Omega$ is the set of admissible constraint forces. For frictionless
  normal contact it is $\R_+^m$; with friction it becomes a product of
  friction cones.
  \item $a_{\mathrm{ref}}$ is the acceleration that the soft constraint would
  like to realize. It is chosen from the current position and velocity error,
  approximately as a spring--damper feedback law. With the present gap
  convention, a scalar normal constraint has the schematic form
  \begin{equation}
    (a_{\mathrm{ref}})_i
    =-\beta_i(J_c\dot q)_i-k_i\phi_i(q),
    \qquad
    \beta_i>0,\quad k_i\geq0.
  \end{equation}
  Thus closing velocity or penetration requests an outward corrective
  acceleration, while a resting contact with zero error has
  $a_{\mathrm{ref}}=0$.
  This is the same feedback idea used in Baumgarte stabilization:
  \cref{eq:baumgarte-error-dynamics} prescribes
  $\ddot e=-2\alpha\dot e-\beta^2e$. Identifying the error with the gap,
  $e=\phi_i(q)$, gives the same spring--damper form with
  $\beta_i=2\alpha$ and $k_i=\beta^2$. The distinction is that Baumgarte
  stabilization imposes this error acceleration as an equality, whereas
  here it serves as a reference for a soft, unilateral contact solve;
  the force restrictions and regularizer determine the acceleration
  actually realized.
  \item $R\succ0$ adds a penalty $\tfrac12\lambda\trans R\lambda$ on contact
  forces. This weakens exact enforcement of the reference acceleration and
  makes the force solution unique. The following explanation shows how a
  force penalty produces a soft contact response.
\end{itemize}

\paragraph{Understanding the role of $R$.}
Consider one frictionless contact with $A>0$ and scalar regularizer
$R=\rho>0$. If $a_{\mathrm{free}}<a_{\mathrm{ref}}$, the positive force
needed to realize the reference without regularization is
\begin{equation}
  \lambda_0=\frac{a_{\mathrm{ref}}-a_{\mathrm{free}}}{A},
  \qquad
  a_{\mathrm{free}}+A\lambda_0=a_{\mathrm{ref}}.
\end{equation}
Completing the square reveals the role of the added penalty:
\begin{equation}
  \frac12(A+\rho)\lambda^2
  +(a_{\mathrm{free}}-a_{\mathrm{ref}})\lambda
  =\frac A2(\lambda-\lambda_0)^2
   +\frac\rho2\lambda^2-\frac A2\lambda_0^2.
\end{equation}
Ignoring the constant last term, the objective balances proximity to the
required force $\lambda_0$ against a penalty on force magnitude. Its minimizer
and realized acceleration are
\begin{equation}
  \boxed{\lambda^\star
  =\frac{a_{\mathrm{ref}}-a_{\mathrm{free}}}{A+\rho}
  =\frac{A}{A+\rho}\lambda_0.}
\end{equation}
\begin{equation}
  a_{\mathrm{con}}
  =a_{\mathrm{free}}+
    \frac{A}{A+\rho}(a_{\mathrm{ref}}-a_{\mathrm{free}}).
\end{equation}
Thus $A/(A+\rho)$ is the fraction of the desired correction applied: all of
it when $\rho=0$, half when $\rho=A$, and very little when $\rho\gg A$.
With the other quantities fixed, larger $\rho$ means softer contact. If
$a_{\mathrm{free}}\geq a_{\mathrm{ref}}$, no positive force is needed and
the bound $\lambda\geq0$ selects $\lambda^\star=0$.

For example, take a mass $m$ at rest on a floor, with zero gap. Then
$A=1/m$, $a_{\mathrm{free}}=-g$, and $a_{\mathrm{ref}}=0$. The force and
acceleration are
\begin{equation}
  \lambda^\star=\frac{mg}{1+m\rho},
  \qquad
  \ddot y=-g+\frac{\lambda^\star}{m}
  =-\frac{m\rho g}{1+m\rho}.
\end{equation}
Rigid contact ($\rho=0$) supplies $mg$ and keeps $\ddot y=0$; soft contact
initially supplies less force and permits penetration. This is intentional
compliance rather than exact nonpenetration.

As penetration develops, the reference feedback provides a restoring
response. With $a_{\mathrm{ref}}=-\beta\dot y-ky$ and $k>0$, a resting
equilibrium has $\lambda=mg$ and satisfies
\begin{equation}
  0=-ky-\rho mg,
  \qquad
  y_{\mathrm{eq}}=-\frac{\rho mg}{k}.
\end{equation}
Like a loaded spring, the contact supports weight by allowing compression;
for fixed $k$, larger $\rho$ requires more penetration. In short,
$a_{\mathrm{ref}}$ specifies the desired correction, while $\rho$ controls
how readily the force realizes it. $R$ itself is not a spring stiffness.

For multiple contacts, the penalty is $\tfrac12\lambda\trans R\lambda$.
At an interior solution of \cref{eq:mujoco-convex-contact}, stationarity and
$a_{\mathrm{con}}=a_{\mathrm{free}}+A\lambda$ give
\begin{align}
  (A+R)\lambda+a_{\mathrm{free}}-a_{\mathrm{ref}}&=0,\\
  a_{\mathrm{con}}&=a_{\mathrm{ref}}-R\lambda.
\end{align}
The residual is therefore $a_{\mathrm{ref}}-a_{\mathrm{con}}=R\lambda$.
Including the force bounds, frictionless contact satisfies
\begin{equation}
  0\leq\lambda\ \perp\
  a_{\mathrm{con}}-a_{\mathrm{ref}}+R\lambda\geq0.
\end{equation}
Since $A\succeq0$ and $R\succ0$, $A+R$ is positive definite: the objective
is strictly convex and the minimizer is unique, even if redundant contacts
make $A$ singular.

In the frictionless hard-contact limit
\begin{equation}
  \Omega=\R_+^m,
  \qquad
  R=0,
  \qquad
  a_{\mathrm{ref}}=0,
  \qquad
  a_{\mathrm{free}}=b,
\end{equation}
\cref{eq:mujoco-convex-contact} reduces exactly to
\cref{eq:rigid-contact-convex-qp}. MuJoCo uses the positive regularizer and
reference acceleration to obtain a soft, uniquely defined contact response
that is convenient for repeated simulation and optimization
\cite{todorov2011convex,todorov2012mujoco}.

\paragraph{Exercises.}

\begin{exercise}
For
\begin{equation}
  \mathcal J[q]
  =\frac12\int_0^1\bigl(\dot q(t)^2+q(t)^2\bigr)dt,
\end{equation}
find the Euler--Lagrange equation.
\end{exercise}
\begin{proof}[Short solution]
Here $\partial L/\partial\dot q=\dot q$ and
$\partial L/\partial q=q$, so $\ddot q-q=0$.
\end{proof}

\begin{exercise}
Find the minimum-energy curve joining $q(0)=a$ to $q(T)=b$ for
$\mathcal J[q]=\tfrac12\int_0^T\dot q^2dt$.
\end{exercise}
\begin{proof}[Short solution]
The equation $\ddot q=0$ and the endpoints give
$q(t)=a+t(b-a)/T$.
\end{proof}

\begin{exercise}
For $L(q,\dot q)=\tfrac12m\dot q^2-V(q)$, derive the forced equation.
\end{exercise}
\begin{proof}[Short solution]
The forced Euler--Lagrange equation gives
$m\ddot q+V'(q)=\tau$.
\end{proof}

\begin{exercise}
Use $V(q)=mgq$ in the preceding exercise with $\tau=0$.
\end{exercise}
\begin{proof}[Short solution]
Since $\ddot q=-g$,
$q(t)=q(0)+\dot q(0)t-gt^2/2$.
\end{proof}

\begin{exercise}
Derive the equation of a pendulum with torque $u$ and damping
$d\dot\theta$.
\end{exercise}
\begin{proof}[Short solution]
The generalized force is $u-d\dot\theta$, so
\begin{equation}
  m\ell^2\ddot\theta+d\dot\theta+mg\ell\sin\theta=u.
\end{equation}
\end{proof}

\begin{exercise}
For a free final endpoint and
$\mathcal J[q]=\int_0^T\tfrac12\dot q^2dt$, solve the problem when
$q(0)=q_0$.
\end{exercise}
\begin{proof}[Short solution]
The natural condition is $\dot q(T)=0$. Since $\ddot q=0$, the velocity is
constant and therefore zero. Hence $q(t)=q_0$.
\end{proof}

\begin{exercise}
Simplify \cref{eq:robot-manipulator-dynamics} when $M$ is constant.
\end{exercise}
\begin{proof}[Short solution]
Then $\dot M=0$ and $\partial M/\partial q_i=0$, so $h=0$ and
$M\ddot q+\nabla V(q)=\tau$.
\end{proof}

\begin{exercise}
Differentiate $c(q)=0$ twice to obtain the velocity and acceleration
constraints.
\end{exercise}
\begin{proof}[Short solution]
With $J=\partial c/\partial q$, the equations are
$J\dot q=0$ and $J\ddot q+\dot J\dot q=0$.
\end{proof}

\begin{exercise}
For $c(x,y)=x+y-1=0$, find the direction of the reaction force.
\end{exercise}
\begin{proof}[Short solution]
Since $J=[1\;1]$, the force is
$J\trans\lambda=\lambda(1,1)\trans$, normal to the line.
\end{proof}

\begin{exercise}
A mass rests on a floor under gravity. Find the normal force using
complementarity.
\end{exercise}
\begin{proof}[Short solution]
The conditions
$\lambda\geq0$, $\ddot y=-g+\lambda/m\geq0$, and
$\lambda\ddot y=0$ give $\lambda=mg$ and $\ddot y=0$.
\end{proof}

\begin{exercise}
Solve the scalar LCP
$w=2\lambda-6$, $\lambda\geq0$, $w\geq0$, $\lambda w=0$.
\end{exercise}
\begin{proof}[Short solution]
The choice $\lambda=0$ is infeasible because $w=-6$. Hence $w=0$ and
$\lambda=3$.
\end{proof}

\begin{exercise}
Solve the scalar LCP
$w=2\lambda+1$, $\lambda\geq0$, $w\geq0$, $\lambda w=0$.
\end{exercise}
\begin{proof}[Short solution]
The solution is $\lambda=0$, $w=1$.
\end{proof}

\begin{exercise}
A point mass $m$ hits a floor with vertical velocity $v^-<0$. Find the impulse
for a perfectly inelastic impact with $v^+=0$.
\end{exercise}
\begin{proof}[Short solution]
From $m(v^+-v^-)=p$, the impulse is $p=-mv^->0$.
\end{proof}

\begin{exercise}
Show that the equality-constraint force $J\trans\lambda$ does no virtual work
along a compatible displacement $\eta$.
\end{exercise}
\begin{proof}[Short solution]
Since $J\eta=0$,
$(J\trans\lambda)\trans\eta=\lambda\trans J\eta=0$.
\end{proof}

\paragraph{Chapter summary.}
A function can be optimized by perturbing it with a small function and
computing the first variation. Integration by parts produces the
Euler--Lagrange equation and the appropriate endpoint conditions. For a
mechanical Lagrangian $L=T-V$, this equation yields the equations of motion,
including the robot manipulator form. Equality constraints add reaction forces
$J\trans\lambda$ and lead to a KKT-like system for acceleration and
multipliers. Unilateral contact adds nonnegative forces and complementary
slackness; after acceleration or velocity is eliminated, frictionless contact
becomes a linear complementarity problem.
\chapter{Continuous-Time Optimal Control}
\label{chap:continuous-time-optimal-control}
\chapteroverview{Finite- and infinite-horizon control, cost-to-go functions, the Hamiltonian, costates, Pontryagin's minimum principle, and numerical methods.}

Chapter~5 optimized an entire curve and showed how stationarity produces a
differential equation. Optimal control adds an important restriction: the
state trajectory cannot be chosen freely. It must be generated by a dynamical
system, and the designer chooses a control input that influences that system.
For a robot, the state might contain joint angles and velocities, while the
control might be a vector of motor torques. The goal is to choose the complete
control signal so that the resulting motion is safe, accurate, and economical.

This viewpoint unifies many robotics tasks. Moving a manipulator between two
configurations while limiting torque, landing a quadrotor with little fuel,
steering a mobile robot around obstacles, and braking a vehicle in minimum time
are all optimal control problems. The central result of this chapter is
Pontryagin's minimum principle (PMP). It introduces a costate and converts a
constrained optimization over functions into coupled state, costate, and
control conditions. The derivation uses the same perturbation and
integration-by-parts ideas as Chapter~5. Before deriving the principle, we
introduce the cost-to-go function and distinguish finite-horizon trajectory
optimization from infinite-horizon feedback design. These ideas lead to the
Riccati and dynamic-programming viewpoints developed in Chapters~7 and~8.
Standard references include
\cite{kirk2004optimal,lewis2012optimal}.

\section{Problem Formulation}

\subsection{Dynamics, Controls, and Trajectories}

Let
\begin{equation}
  x(t)\in\R^n,
  \qquad
  u(t)\in\R^m
\end{equation}
denote the state and control. A controlled system has dynamics
\begin{equation}
  \dot x(t)=f\bigl(t,x(t),u(t)\bigr),
  \qquad x(t_0)=x_0.
  \label{eq:continuous-control-dynamics}
\end{equation}
Once $x_0$ and $u(\cdot)$ are specified, the state trajectory is determined by
integrating this differential equation. Thus $x(\cdot)$ and $u(\cdot)$ are not
independent optimization variables. We assume enough smoothness for the state
equation and displayed derivatives to be well defined; smooth dynamics and
piecewise-continuous controls suffice for this chapter.

\begin{example}[A torque-controlled robot]
Let $q\in\R^d$ be a manipulator configuration and $v=\dot q$. With commanded
joint torque $u$, its first-order state equation is
\begin{equation}
  x=\begin{bmatrix}q\\v\end{bmatrix},
  \qquad
  \dot x=
  \begin{bmatrix}
    v\\
    M(q)^{-1}\bigl(u-h(q,v)-g(q)\bigr)
  \end{bmatrix},
  \label{eq:robot-control-state-space}
\end{equation}
where $M$ is the mass matrix, $h$ contains velocity-dependent forces, and
$g(q)=\nabla V(q)$ is gravity. The torque signal determines the motion through
these equations.
\end{example}

\begin{example}[A planar mobile robot]
A unicycle model has state $(p_x,p_y,\theta)$ and controls $(v,\omega)$:
\begin{equation}
  \dot p_x=v\cos\theta,
  \qquad
  \dot p_y=v\sin\theta,
  \qquad
  \dot\theta=\omega.
  \label{eq:unicycle-continuous-dynamics}
\end{equation}
It captures the nonholonomic fact that the robot cannot translate
instantaneously sideways.
\end{example}

A pair $(x(\cdot),u(\cdot))$ is \emph{admissible} if it satisfies the
dynamics, initial condition, regularity requirements, and all constraints.

\subsection{Running and Terminal Costs}

The standard finite-horizon objective has Bolza form
\begin{equation}
  J[u]
  =\Phi\bigl(x(t_f),t_f\bigr)
  +\int_{t_0}^{t_f}\ell\bigl(t,x(t),u(t)\bigr)\,dt.
  \label{eq:bolza-optimal-control-cost}
\end{equation}
The \emph{running cost} $\ell$ measures performance throughout the motion,
while the \emph{terminal cost} $\Phi$ measures the final state. The basic
problem is
\begin{equation}
  \begin{aligned}
    \minimize_{x(\cdot),u(\cdot)}\quad
      &\Phi(x(t_f),t_f)
       +\int_{t_0}^{t_f}\ell(t,x,u)\,dt\\
    \subjectto\quad
      &\dot x=f(t,x,u),
      \qquad x(t_0)=x_0.
  \end{aligned}
  \label{eq:basic-continuous-optimal-control-problem}
\end{equation}

A common robotics running cost is
\begin{equation}
  \ell(t,x,u)
  =\frac12\norm{x-x_d(t)}_Q^2
   +\frac12\norm{u}_R^2,
  \qquad Q\succeq0,
  \quad R\succ0,
  \label{eq:tracking-running-cost}
\end{equation}
where $\norm{z}_Q^2=z\trans Qz$. A terminal cost
$\Phi=\norm{x(t_f)-x_f}_P^2/2$ encourages arrival near a target. It does not
impose exact arrival; that requires the constraint $x(t_f)=x_f$.

Costs should reflect the physical goal. Squared torque is a smooth measure of
effort, but it is not literally electrical energy unless the actuator model
supports that interpretation. Likewise, an obstacle penalty discourages
collision but does not guarantee avoidance.

\subsection{The Cost-to-Go Function}

The optimization problem can be restarted at any intermediate time. Suppose
the clock reads $t$ and the current state is $x$. The \emph{optimal
cost-to-go}, also called the \emph{value function}, is
\begin{equation}
  V(t,x)
  =
  \inf_{u(\cdot)}
  \left\{
    \Phi\bigl(x_u(t_f),t_f\bigr)
    +\int_t^{t_f}\ell\bigl(s,x_u(s),u(s)\bigr)\,ds
  \right\},
  \label{eq:finite-horizon-cost-to-go-definition}
\end{equation}
where $x_u$ satisfies
\begin{equation}
  \dot x_u(s)=f\bigl(s,x_u(s),u(s)\bigr),
  \qquad x_u(t)=x,
\end{equation}
and the infimum is over admissible controls on $[t,t_f]$. In particular,
the optimal value of the original problem is
\begin{equation}
  J^\star=V(t_0,x_0).
\end{equation}

The distinction between $J[u]$ and $V(t,x)$ is important. The functional
$J[u]$ evaluates one complete control signal from the specified initial
condition. The value function instead asks a family of optimization
problems: for every possible state $x$ and starting time $t$, what is the
smallest future cost still achievable? It compresses all optimal future
behavior into one scalar field over time and state.

The value also has a practical interpretation. A small value means that the
remaining task can be completed cheaply; a large value means that the current
state is difficult or expensive to recover from. Differences in $V$ compare
states by their future consequences rather than by geometric distance alone.
For example, two robot states may be equally far from a target, but the state
moving rapidly away from the target can have much larger cost-to-go.

At the terminal time there is no running interval left, so
\begin{equation}
  V(t_f,x)=\Phi(x,t_f).
  \label{eq:finite-horizon-value-terminal-condition}
\end{equation}
If an optimizer exists for every feasible $(t,x)$, the control selected at the
current time defines an optimal feedback control law
\begin{equation}
  u=\mu^\star(t,x).
  \label{eq:finite-horizon-optimal-feedback}
\end{equation}
Starting from $(t_0,x_0)$ and applying this law generates the particular
optimal trajectory $x^\star(\cdot)$ and control
$u^\star(t)=\mu^\star(t,x^\star(t))$. Thus the value function describes a
whole family of optimal trajectories, whereas a single open-loop solution
describes only one member of that family. Chapter~8 derives the equation
satisfied by $V$; here we use it primarily to clarify the meaning of optimal
control and the costate.

\subsection{Infinite-Horizon Optimal Control}

Some tasks have a natural deadline or terminal event, but regulation and
disturbance rejection are continuing tasks. For an autonomous system
\begin{equation}
  \dot x=f(x,u),
\end{equation}
an undiscounted infinite-horizon problem has the form
\begin{equation}
  V_\infty(x_0)
  =
  \inf_{u(\cdot)}
  \int_0^\infty \ell\bigl(x(t),u(t)\bigr)\,dt
  \quad\subjectto\quad
  \dot x=f(x,u),
  \qquad x(0)=x_0.
  \label{eq:undiscounted-infinite-horizon-control}
\end{equation}
There is no terminal time or terminal cost. Instead, the running cost must
encode the desired long-run behavior. In regulation problems,
$\ell(x,u)\geq0$ usually vanishes at the desired equilibrium. A finite value
then requires the controller to approach that equilibrium sufficiently fast.
If the system cannot be stabilized, or if positive cost persists forever,
the integral may be infinite.

A discounted alternative is
\begin{equation}
  V_\rho(x_0)
  =
  \inf_{u(\cdot)}
  \int_0^\infty
  e^{-\rho t}\ell\bigl(x(t),u(t)\bigr)\,dt,
  \qquad \rho>0.
  \label{eq:discounted-infinite-horizon-control}
\end{equation}
Discounting gives less weight to distant costs and can make the integral
finite under weaker conditions. It is not merely a numerical trick, however:
changing $\rho$ changes the control objective and can make the controller more
short-sighted.

When the dynamics, cost, and constraints are time invariant, the
infinite-horizon value depends only on the current state, and one seeks a
stationary feedback law
\begin{equation}
  u=\mu_\infty^\star(x).
  \label{eq:stationary-infinite-horizon-feedback}
\end{equation}
The word \emph{stationary} means that the same state produces the same control
regardless of the clock time. It does not mean that the state is constant.
Infinite-horizon problems still require appropriate assumptions for the value
to be finite and for an optimal control law to exist. Unlike a finite-horizon
problem, there is no fixed terminal time at which to impose a final boundary
condition. The desired solution must instead be selected through appropriate
long-run requirements, such as finite accumulated cost, convergence toward a
target equilibrium, or closed-loop stability. The precise boundary conditions
at infinity require care and are discussed later, after the necessary
optimality conditions have been introduced.

\subsection{Finite Horizon, Infinite Horizon, and Feedback}

The horizon and the form of the solution are related, but they are different
ideas.

\begin{itemize}
  \item A \emph{finite-horizon} problem optimizes over $[t,t_f]$. Its value
  $V(t,x)$ generally depends on both time and state because the remaining time
  and terminal cost matter. The corresponding feedback law
  $\mu^\star(t,x)$ is generally time varying.
  \item An \emph{infinite-horizon} problem optimizes all future behavior. For
  an autonomous, time-invariant problem, its value $V_\infty(x)$ and optimal
  feedback $\mu_\infty^\star(x)$ are often stationary. The problem emphasizes
  long-run regulation or operation rather than arrival at a prescribed final
  time.
  \item An \emph{optimal trajectory} is the state--control pair produced for
  one initial condition. It is an open-loop plan: if a disturbance moves the
  robot away from that trajectory, the precomputed control signal does not by
  itself specify how to recover.
  \item An \emph{optimal state-feedback controller} is a rule defined over a
  set of possible states. It observes the current state and selects
  $u=\mu^\star(t,x)$ for a finite-horizon problem, or
  $u=\mu_\infty^\star(x)$ for a stationary infinite-horizon problem. After a
  disturbance, it can therefore generate a new optimal or locally optimal
  trajectory from the disturbed state.
\end{itemize}

Finite horizon does not mean open loop: finite-horizon dynamic programming
produces the feedback law $\mu^\star(t,x)$. Likewise, infinite horizon does not
automatically mean feedback: one can solve an infinite-horizon problem for a
single initial state and obtain only an open-loop extremal. Pontryagin's
principle naturally characterizes individual optimal trajectories, while the
value-function viewpoint naturally produces feedback laws. Chapters~7 and~8
make this distinction concrete through Riccati equations and the
Hamilton--Jacobi--Bellman equation.

\subsection{State and Control Constraints}

Practical systems may require
\begin{align}
  u(t)&\in U,
  &x(t)&\in X,\\
  g\bigl(t,x(t),u(t)\bigr)&\leq0,
  &c\bigl(t,x(t),u(t)\bigr)&=0,
  \label{eq:optimal-control-path-constraints}
\end{align}
and endpoint constraints
\begin{equation}
  \psi\bigl(x(t_0),x(t_f),t_f\bigr)=0.
\end{equation}
Examples include torque and joint limits, obstacle clearance, contact, and a
prescribed terminal pose. Control bounds
\begin{equation}
  u_{\min}\leq u(t)\leq u_{\max}
  \label{eq:box-control-bounds}
\end{equation}
are essential in minimum-time problems; without them, the model may request
unbounded force. State constraints are more subtle because they must hold at
every time and may become active only over particular portions of a
trajectory. We first study problems without active state-path constraints,
while handling control constraints explicitly. Chapter~9 develops
discretization-based methods better suited to complicated path constraints.

\section{Pontryagin's Minimum Principle}

\subsection{Linearized Dynamics and the Costate}

Recall that the dynamics are
$\dot x=f(t,x,u)$, where
$f:\R\times\R^n\times\R^m\to\R^n$ maps the current time, state, and control
to the state derivative. Consider a nominal state--control pair
$(x(\cdot),u(\cdot))$ satisfying these dynamics. Choose a control perturbation
$\eta(t)\in\R^m$ and define
\begin{equation}
  u_\epsilon(t)=u(t)+\epsilon\eta(t).
\end{equation}
Let $x_\epsilon(t)$ be the state trajectory generated by $u_\epsilon$, with
the same initial condition $x_\epsilon(t_0)=x_0$. The function
$z(t)\in\R^n$ is the first-order change in the state caused by the control
perturbation:
\begin{equation}
  z(t)
  =
  \left.
  \frac{d}{d\epsilon}x_\epsilon(t)
  \right|_{\epsilon=0}.
\end{equation}
Equivalently,
$x_\epsilon(t)=x(t)+\epsilon z(t)+o(\epsilon)$. Differentiating the perturbed
dynamics with respect to $\epsilon$ at $\epsilon=0$ gives
\begin{equation}
  \dot z(t)
  =
  f_x\bigl(t,x(t),u(t)\bigr)z(t)
  +
  f_u\bigl(t,x(t),u(t)\bigr)\eta(t),
  \qquad z(t_0)=0.
  \label{eq:linearized-control-dynamics}
\end{equation}
Here
\begin{equation}
  f_x=\frac{\partial f}{\partial x}\in\R^{n\times n},
  \qquad
  f_u=\frac{\partial f}{\partial u}\in\R^{n\times m}
\end{equation}
are the Jacobians of the dynamics with respect to the state and control,
evaluated along the nominal trajectory $(x(t),u(t))$. The condition
$z(t_0)=0$ follows because the initial state is fixed and is therefore
unchanged by the control perturbation.
Recall from \cref{eq:bolza-optimal-control-cost} that
\begin{equation}
  J[u]
  =
  \Phi\bigl(x(t_f),t_f\bigr)
  +
  \int_{t_0}^{t_f}\ell\bigl(t,x(t),u(t)\bigr)\,dt.
\end{equation}
The perturbed control changes the terminal state, the state inside the
integral, and the control inside the integral. We use
\begin{align}
  \Phi_x\bigl(x(t_f),t_f\bigr)
  &=
  \frac{\partial\Phi}{\partial x}
  \bigl(x(t_f),t_f\bigr)
  \in\R^n,\\
  \ell_x(t)
  &=
  \frac{\partial\ell}{\partial x}
  \bigl(t,x(t),u(t)\bigr)
  \in\R^n,\\
  \ell_u(t)
  &=
  \frac{\partial\ell}{\partial u}
  \bigl(t,x(t),u(t)\bigr)
  \in\R^m
\end{align}
for the gradient of the terminal cost with respect to its state argument and
the gradients of the running cost with respect to the state and control.
Thus $\Phi_x$ is evaluated at the nominal terminal state, whereas
$\ell_x(t)$ and $\ell_u(t)$ are evaluated along the nominal state--control
trajectory. Applying the chain rule gives the first-order cost variation
\begin{equation}
  \delta J
  =\Phi_x(x(t_f),t_f)\trans z(t_f)
  +\int_{t_0}^{t_f}
    \left(\ell_x\trans z+\ell_u\trans\eta\right)\,dt.
  \label{eq:control-cost-variation-direct}
\end{equation}
The state variation $z$ is not independent of the control perturbation
$\eta$. As shown in \cref{eq:linearized-control-dynamics}, once $\eta(t)$ is
chosen, $z(t)$ is determined by the forced linear differential equation
\begin{equation}
  \dot z=f_x z+f_u\eta,
  \qquad z(t_0)=0.
\end{equation}
Thus changing the control perturbation changes the entire state variation
through the system dynamics. Introduce a time-varying
multiplier $\lambda(t)\in\R^n$ and augment the cost by a term that is zero on
feasible trajectories:
\begin{equation}
  \overline J
  =\Phi(x(t_f),t_f)
   +\int_{t_0}^{t_f}
    \left[
      \ell(t,x,u)+\lambda\trans(f(t,x,u)-\dot x)
    \right]\,dt.
  \label{eq:augmented-continuous-control-cost}
\end{equation}
The time-varying multiplier $\lambda(t)\in\R^n$ is called the
\emph{costate}, or \emph{adjoint variable}. It has one component for each
state equation and is the function-space analogue of a Lagrange multiplier
for the dynamical constraint. Along every feasible trajectory,
$f(t,x,u)-\dot x=0$, so the added term vanishes and
$\overline J=J$.

Now keep $\lambda$ fixed and perturb the state and control in the directions
$z$ and $\eta$. The first variation of the augmented functional is
\begin{align}
  \delta\overline J
  &={}
  \Phi_x\bigl(x(t_f),t_f\bigr)\trans z(t_f)\\
  &\quad+
  \int_{t_0}^{t_f}
  \Bigl[
    \ell_x\trans z
    +\ell_u\trans\eta
    +\lambda\trans
      \bigl(f_x z+f_u\eta-\dot z\bigr)
  \Bigr]\,dt.
  \label{eq:augmented-control-variation-before-parts}
\end{align}
The three terms inside the multiplier contribution are the first-order
changes in $f(t,x,u)-\dot x$: changing the state contributes $f_x z$, changing
the control contributes $f_u\eta$, and changing $\dot x$ contributes
$\dot z$. This augmented variation is useful because the multiplier will
allow the coefficients of $z$ and $\eta$ to be handled separately.

\subsection{The Control Hamiltonian}

Define
\begin{equation}
  H(t,x,u,\lambda)
  =\ell(t,x,u)+\lambda\trans f(t,x,u).
  \label{eq:control-hamiltonian-definition}
\end{equation}
Its state and control gradients are
\begin{equation}
  H_x=\ell_x+f_x\trans\lambda,
  \qquad
  H_u=\ell_u+f_u\trans\lambda.
\end{equation}
The only derivative of a variation in
\cref{eq:augmented-control-variation-before-parts} is contained in
$-\lambda\trans\dot z$. Integration by parts gives
\begin{equation}
  -\int_{t_0}^{t_f}\lambda\trans\dot z\,dt
  =
  -\left[\lambda\trans z\right]_{t_0}^{t_f}
  +\int_{t_0}^{t_f}\dot\lambda\trans z\,dt.
\end{equation}
Substituting this identity and the derivatives of $H$ into the augmented
variation yields
\begin{align}
  \delta\overline J
  &={}
  \bigl(\Phi_x-\lambda(t_f)\bigr)\trans z(t_f)
  +\lambda(t_0)\trans z(t_0)\\
  &\quad+
  \int_{t_0}^{t_f}
  \left[
    (H_x+\dot\lambda)\trans z
    +H_u\trans\eta
  \right]\,dt.
  \label{eq:augmented-control-first-variation}
\end{align}
Here $\Phi_x$ is evaluated at $(x(t_f),t_f)$, and all quantities in the
integral are evaluated along the nominal trajectory. The multiplier
$\lambda$ was introduced as an auxiliary function and has not yet been
specified. At a stationary solution, the first variation must be zero for all
admissible interior state variations $z$. Therefore, the coefficient of $z$
in \cref{eq:augmented-control-first-variation} must vanish. This determines
$\lambda$ through
\begin{equation}
  \boxed{\dot\lambda=-H_x.}
  \label{eq:costate-equation}
\end{equation}
This differential equation is called the \emph{costate equation}; it is a
necessary stationarity condition that determines how the multiplier evolves.

The boundary terms are handled in the same way. Because the initial state is
fixed, every admissible perturbation has $z(t_0)=0$, so the initial boundary
term vanishes. If the terminal state is free, however, $z(t_f)$ may point in
any direction. Its coefficient must therefore vanish, which gives
\begin{equation}
  \boxed{\lambda(t_f)=\Phi_x(x(t_f),t_f).}
  \label{eq:free-terminal-state-transversality}
\end{equation}
If the terminal state is fixed, then $z(t_f)=0$, and this boundary term
instead imposes no terminal condition on $\lambda$. For a free terminal state,
the terminal condition above and the costate equation form a terminal-value
problem, so $\lambda$ is generally propagated backward from $t_f$.

For feasible variations, $\overline J=J$. After imposing the costate and
terminal conditions, the remaining variation is therefore
\begin{equation}
  \delta J
  =
  \delta\overline J
  =
  \int_{t_0}^{t_f}H_u\trans\eta\,dt.
  \label{eq:control-first-variation-hamiltonian}
\end{equation}
Thus the costate propagates future-cost information backward, and
$H_u=\ell_u+f_u\trans\lambda$ measures the total first-order effect of changing
the control now. If the control is unconstrained and the optimum lies in the
interior of the control set, then $\eta$ is arbitrary. Stationarity of
\cref{eq:control-first-variation-hamiltonian} therefore requires
\begin{equation}
  H_u=0.
  \label{eq:unconstrained-hamiltonian-stationarity}
\end{equation}
For a constrained control, not every direction $\eta$ is admissible, so
$H_u=0$ need not hold. The appropriate condition is instead that the optimal
control minimize the Hamiltonian over the admissible control set, as stated
next.

\subsection{Statement of the Principle}

\begin{theorem}[Pontryagin's minimum principle, basic normal form]
Consider \cref{eq:basic-continuous-optimal-control-problem} with fixed times,
fixed initial state, free terminal state, no active state-path constraints, and
$u(t)\in U$. If $(x^\star,u^\star)$ is a sufficiently regular local optimum,
then there exists a costate $\lambda^\star$ such that, almost everywhere,
\begin{align}
  \dot x^\star
  &=H_\lambda(t,x^\star,u^\star,\lambda^\star)
   =f(t,x^\star,u^\star),\\
  \dot\lambda^\star
  &=-H_x(t,x^\star,u^\star,\lambda^\star),\\
  u^\star(t)
  &\in\argmin_{u\in U}
    H(t,x^\star(t),u,\lambda^\star(t)),
  \label{eq:pmp-hamiltonian-minimization}\\
  \lambda^\star(t_f)
  &=\Phi_x(x^\star(t_f),t_f).
\end{align}
\end{theorem}

The phrase ``almost everywhere'' allows the optimal control to have isolated
switches or jump discontinuities. The state remains continuous, but a bounded
optimal control can change value instantaneously at a switching time. A common
class of such controls is introduced later in the chapter.

These conditions should be read as one coupled system:
\begin{enumerate}
  \item The state equation starts from the known initial condition
  $x(t_0)=x_0$ and naturally evolves forward in time.
  \item The costate equation is usually supplied with a condition at the
  terminal time. For example, when the terminal state is free,
  $\lambda(t_f)=\Phi_x(x(t_f),t_f)$. It therefore naturally evolves backward
  in time.
  \item At each time, the minimizing condition determines $u^\star(t)$ from
  the current values of $x^\star(t)$ and $\lambda^\star(t)$. Thus the state
  equation depends on the costate through the control, while the costate
  equation depends on the state.
  \item The initial state alone is therefore insufficient for a forward
  integration, because the required control depends on the costate, whose
  initial value is generally unknown. Likewise, the terminal costate condition
  alone is insufficient for a backward integration, because its equation
  depends on the state trajectory.
  \item The unknown boundary data must be adjusted until the forward state,
  backward costate, minimizing control, and all endpoint conditions agree
  simultaneously.
\end{enumerate}
Consequently, the equations cannot generally be solved by one independent
forward integration followed by one independent backward integration. They
form a two-point boundary-value problem: some data are specified at $t_0$ and
the remaining data are specified at $t_f$. Resolving this forward--backward
coupling is the main computational difficulty of the state--costate
formulation.

The state runs forward from $x(t_0)=x_0$, while the costate has a terminal
condition and naturally runs backward. For an unconstrained interior control,
Hamiltonian minimization reduces to
\begin{equation}
  H_u(t,x^\star,u^\star,\lambda^\star)=0.
  \label{eq:pmp-stationary-control}
\end{equation}
If $H$ is strictly convex in $u$, this equation uniquely minimizes it. If $H$
is linear in a bounded control, the solution usually lies on the boundary of
$U$, producing bang--bang control.

These conditions are necessary, not automatically sufficient. They identify
extremals, which need not all be minima.

\begin{remark}[Sign conventions]
Some texts define $H=\lambda\trans f-\ell$ and state a maximum principle. We
use $H=\ell+\lambda\trans f$, so the control minimizes $H$. The conventions are
equivalent when used consistently.
\end{remark}

\subsection{Extension to Infinite Horizons}

The finite-horizon theorem above does not apply to an infinite-horizon
problem word for word, because there is no terminal time at which to impose
$\lambda(t_f)=\Phi_x$. Nevertheless, its interior conditions extend in a
natural way. For the undiscounted problem in
\cref{eq:undiscounted-infinite-horizon-control}, define
\begin{equation}
  H_\infty(x,u,\lambda)
  =\ell(x,u)+\lambda\trans f(x,u).
  \label{eq:undiscounted-infinite-horizon-hamiltonian}
\end{equation}
Under suitable regularity and optimality assumptions, an infinite-horizon
extremal satisfies the following conditions almost everywhere on every finite
subinterval:
\begin{align}
  \dot x^\star
  &=f(x^\star,u^\star),\\
  \dot\lambda^\star
  &=-H_{\infty,x}(x^\star,u^\star,\lambda^\star),\\
  u^\star(t)
  &\in\argmin_{u\in U}
    H_\infty(x^\star(t),u,\lambda^\star(t)).
  \label{eq:infinite-horizon-pmp-interior-conditions}
\end{align}
Thus the state equation, costate equation, and pointwise Hamiltonian
minimization remain. What changes is the boundary condition used to select the
correct solution of these equations.

There is no single transversality condition at infinity that is valid for
every problem. Depending on the admissible trajectories and the notion of
infinite-horizon optimality, one may obtain a limiting condition such as
\begin{equation}
  \lim_{t\to\infty}
  \lambda^\star(t)\trans\delta x(t)=0
  \label{eq:infinite-horizon-transversality-schematic}
\end{equation}
for an appropriate class of terminal variations. Under stronger assumptions,
this may imply $\lambda^\star(t)\to0$, but that conclusion must be proved; it
cannot be adopted as a universal terminal condition. This is one reason that
infinite-horizon boundary-value computations require more care than simply
replacing $t_f$ by infinity in the finite-horizon equations.

For the discounted objective in
\cref{eq:discounted-infinite-horizon-control}, the running cost incurred at
time $t$ enters the objective with weight $e^{-\rho t}$. Applying the same
multiplier argument as in the finite-horizon derivation therefore gives the
augmented integrand
\begin{equation}
  e^{-\rho t}\ell(x,u)+p\trans\bigl(f(x,u)-\dot x\bigr),
\end{equation}
where $p(t)$ is the costate associated with the discounted objective. The
resulting \emph{present-value Hamiltonian} is
\begin{equation}
  H^{\mathrm p}(t,x,u,p)
  =e^{-\rho t}\ell(x,u)+p\trans f(x,u),
\end{equation}
and the usual variation gives
\begin{equation}
  \dot p=-H^{\mathrm p}_x.
\end{equation}
The term ``present value'' indicates that future costs, and hence their
marginal value represented by $p$, are measured after discounting them to the
initial time. Consequently, even for autonomous dynamics and running cost,
$H^{\mathrm p}$ contains the explicit time-dependent factor $e^{-\rho t}$.

It is often more convenient to express the marginal value of the state in
undiscounted, time-$t$ units. Define the rescaled costate
\begin{equation}
  \lambda(t)=e^{\rho t}p(t).
\end{equation}
Multiplying $H^{\mathrm p}$ by the positive factor $e^{\rho t}$ then gives the
\emph{current-value Hamiltonian}
\begin{equation}
  H^{\mathrm c}(x,u,\lambda)
  =e^{\rho t}H^{\mathrm p}(t,x,u,p)
  =\ell(x,u)+\lambda\trans f(x,u).
  \label{eq:current-value-hamiltonian}
\end{equation}
Because $e^{\rho t}>0$, this rescaling does not change which control minimizes
the Hamiltonian at a fixed time. Its advantage is that, for an autonomous
problem, $H^{\mathrm c}$ has no explicit time dependence.

The rescaling does change the form of the costate equation. Differentiating
$\lambda=e^{\rho t}p$ gives
\begin{align}
  \dot\lambda
  &=\rho e^{\rho t}p+e^{\rho t}\dot p\\
  &=\rho\lambda-e^{\rho t}H^{\mathrm p}_x\\
  &=\rho\lambda-H^{\mathrm c}_x.
\end{align}
Thus the term $\rho\lambda$ appears solely because the costate has been
converted from present-value to current-value units. The corresponding
necessary conditions are
\begin{align}
  \dot x^\star
  &=H^{\mathrm c}_\lambda=f(x^\star,u^\star),\\
  \dot\lambda^\star
  &=\rho\lambda^\star-H^{\mathrm c}_x,\\
  u^\star(t)
  &\in\argmin_{u\in U}
    H^{\mathrm c}(x^\star(t),u,\lambda^\star(t)).
  \label{eq:discounted-current-value-pmp}
\end{align}
The additional term $\rho\lambda$ in the costate equation is the effect of
discounting. Setting $\rho=0$ recovers the undiscounted equation.

These conditions still characterize an extremal associated with a particular
initial state; they do not by themselves construct the stationary feedback
law $\mu_\infty^\star(x)$. Obtaining a controller defined over a state domain
requires the value-function viewpoint developed in Chapter~8, or special
structure such as the infinite-horizon linear--quadratic problem in
Chapter~7.

\subsection{The Costate as a Cost-to-Go Gradient}

The cost-to-go function in
\cref{eq:finite-horizon-cost-to-go-definition} assigns an optimal remaining
cost to every time--state pair. To see how it is related to the costate,
consider an optimal trajectory and pause it at time $t$. Perturb its state from
$x^\star(t)$ to $x^\star(t)+\delta x$, and then reoptimize the control over the
remaining interval $[t,t_f]$. If $V$ is differentiable, the resulting change
in optimal future cost is
\begin{equation}
  V\bigl(t,x^\star(t)+\delta x\bigr)
  -V\bigl(t,x^\star(t)\bigr)
  =\nabla_x V\bigl(t,x^\star(t)\bigr)\trans\delta x
   +o(\norm{\delta x}).
  \label{eq:value-state-sensitivity}
\end{equation}

Now apply the first-variation calculation to this remaining-horizon problem,
using $t$ as its initial time. In
\cref{eq:augmented-control-first-variation}, replace $t_0$ by $t$. Since the
state at this new initial time is perturbed by $\delta x$, its variation
satisfies $z(t)=\delta x$. The first variation becomes
\begin{align}
  \delta\overline J
  &={}
  \bigl(\Phi_x-\lambda(t_f)\bigr)\trans z(t_f)
  +\lambda(t)\trans\delta x\\
  &\quad+
  \int_t^{t_f}
  \left[
    (H_x+\dot\lambda)\trans z
    +H_u\trans\eta
  \right]dt.
  \label{eq:remaining-horizon-first-variation}
\end{align}
Each optimality condition removes one part of this expression:
\begin{itemize}
  \item the terminal condition $\lambda(t_f)=\Phi_x$ removes the term
  involving $z(t_f)$;
  \item the costate equation $\dot\lambda=-H_x$ removes the interior term
  involving $z$; and
  \item the control stationarity or minimization condition removes the
  first-order contribution associated with adjusting the optimal control.
  For an unconstrained interior optimum, this is simply $H_u=0$.
\end{itemize}
This last point is an instance of the envelope principle: once the control is
already optimal, its induced first-order adjustment does not contribute to the
first-order change in the optimized cost. Consequently, the only remaining
contribution is the boundary term at the perturbed starting state,
\begin{equation}
  \delta J^\star
  =\lambda^\star(t)\trans\delta x.
  \label{eq:costate-state-sensitivity}
\end{equation}
The optimized objective $J^\star$ on $[t,t_f]$ is precisely the value
$V(t,x)$. Therefore
\cref{eq:value-state-sensitivity,eq:costate-state-sensitivity} describe the
same change in optimal cost. Because this equality holds for every small
direction $\delta x$, their coefficients must agree. Hence, when $V$ is smooth
and the optimum is regular, the costate along an optimal trajectory satisfies
\begin{equation}
  \lambda^\star(t)=\nabla_x V(t,x^\star(t)).
  \label{eq:costate-value-gradient}
\end{equation}
Thus the costate is the local slope of the cost-to-go sampled along one
optimal trajectory. It is a shadow price: a large component means that a
small state error in that direction greatly changes the best achievable
future cost. For a robot near a narrow passage, the costate of lateral
position may be large because a small lateral error strongly affects future
feasibility. Pontryagin's principle computes this gradient along the chosen
extremal without first constructing the entire function $V(t,x)$.

This distinction is computationally important. Solving the state--costate
boundary-value problem for one initial state produces one optimal trajectory
and the values of $\nabla_x V$ encountered along it. Computing the value
function produces information for many possible initial states and can be
used to define feedback away from the nominal trajectory, but it is much more
expensive in high-dimensional state spaces.

Endpoint restrictions determine how this backward sensitivity is initialized.
For the free-terminal-state problem,
\cref{eq:finite-horizon-value-terminal-condition} gives
$\nabla_x V(t_f,x)=\Phi_x(x,t_f)$, consistent with the terminal condition from
Pontryagin's minimum principle,
$\lambda(t_f)=\Phi_x$.

The word \emph{transversality} refers to the boundary conditions obtained by
requiring the endpoint part of the first variation to vanish. After imposing
the costate equation, the terminal-state contribution has the form
\begin{equation}
  \bigl(\Phi_x-\lambda(t_f)\bigr)\trans z(t_f),
  \label{eq:terminal-state-variation-term}
\end{equation}
where $z(t_f)=\delta x(t_f)$ is the allowed first-order displacement of the
terminal state. The condition on $\lambda(t_f)$ depends on which terminal
displacements are permitted.

\begin{itemize}
  \item \emph{Free terminal state.} If there is no constraint on $x(t_f)$,
  then $z(t_f)$ may point in any direction. For
  \cref{eq:terminal-state-variation-term} to vanish for every possible
  $z(t_f)$, its coefficient must be zero. Therefore
  \begin{equation}
    \lambda(t_f)=\Phi_x(x(t_f),t_f).
  \end{equation}
  This is the standard terminal condition used when the terminal state is
  completely free.

  \item \emph{Fixed terminal state.} If $x(t_f)=x_f$ is prescribed, every
  admissible perturbation must satisfy $z(t_f)=0$. Hence
  \cref{eq:terminal-state-variation-term} vanishes automatically, regardless
  of $\lambda(t_f)$, and the variational argument supplies no terminal value
  for the costate. This does not mean that $\lambda(t_f)$ is irrelevant. Its
  unknown value, or equivalently the unknown initial value $\lambda(t_0)$,
  must be selected so that integration of the coupled state--costate equations
  produces the required endpoint $x(t_f)=x_f$.

  \item \emph{Terminal state constrained to a manifold.} Suppose
  $\psi(x(t_f))=0$, where $\psi:\R^n\to\R^r$. Linearizing this constraint
  shows that an admissible endpoint displacement must satisfy
  \begin{equation}
    \psi_x(x(t_f))z(t_f)=0.
    \label{eq:linearized-terminal-manifold}
  \end{equation}
  Thus $z(t_f)$ may move tangent to the constraint manifold but not normal to
  it. For \cref{eq:terminal-state-variation-term} to vanish for every such
  tangent displacement, the vector $\lambda(t_f)-\Phi_x$ must lie in the
  normal space spanned by the rows of $\psi_x$. Therefore, for some multiplier
  $\nu\in\R^r$,
  \begin{equation}
    \lambda(t_f)=\Phi_x+\psi_x\trans\nu.
    \label{eq:pmp-endpoint-constraint-transversality}
  \end{equation}
  The multiplier $\nu$ measures the first-order cost of enforcing the terminal
  constraint. A fully fixed terminal state is the limiting case in which
  there are enough independent endpoint constraints to eliminate every
  terminal-state variation.
\end{itemize}

\paragraph{Free final time.}
The preceding cases concern which terminal states are allowed at a fixed
terminal time. A separate variation is needed when the optimizer may also
choose the terminal time. Suppose first that the terminal state is free and
that there is no additional terminal-time constraint. Perturb the endpoint
from $t_f$ to $t_f+\delta t_f$.

There are two contributions to the displacement of the terminal state. The
trajectory itself may vary by $z(t_f)$ at the original time $t_f$, and the
shift in time moves the state along its dynamics for an additional
$\delta t_f$. To first order, the actual terminal-state displacement is
\begin{equation}
  \delta x_f
  =
  z(t_f)+\dot x(t_f)\delta t_f
  =
  z(t_f)+f(t_f,x(t_f),u(t_f))\delta t_f.
  \label{eq:moving-terminal-state-variation}
\end{equation}
Therefore the terminal cost changes by
\begin{equation}
  \delta\Phi
  =
  \Phi_x\trans z(t_f)
  +
  \left(\Phi_x\trans f(t_f)+\Phi_{t_f}\right)\delta t_f.
  \label{eq:moving-terminal-cost-variation}
\end{equation}
Changing the upper limit of the running-cost integral also contributes
\begin{equation}
  \ell(t_f,x(t_f),u(t_f))\delta t_f.
  \label{eq:moving-running-cost-variation}
\end{equation}

After integration by parts and imposition of the costate equation, the
endpoint portion of the first variation is consequently
\begin{align}
  \delta J_{\mathrm{end}}
  &=
  \bigl(\Phi_x-\lambda(t_f)\bigr)\trans z(t_f)\\
  &\quad+
  \left[
    \ell(t_f)
    +\Phi_x\trans f(t_f)
    +\Phi_{t_f}
  \right]\delta t_f.
  \label{eq:free-final-time-endpoint-variation}
\end{align}
Because the terminal state is free, $z(t_f)$ is arbitrary, and its coefficient
again gives
\begin{equation}
  \lambda(t_f)=\Phi_x.
\end{equation}
Substituting this condition into the coefficient of $\delta t_f$ gives
\begin{equation}
  \ell(t_f)+\lambda(t_f)\trans f(t_f)+\Phi_{t_f}
  =
  H(t_f)+\Phi_{t_f}.
\end{equation}
The terminal time is free, so small positive and negative values of
$\delta t_f$ are admissible. The coefficient must therefore vanish:
\begin{equation}
  \boxed{H(t_f)+\Phi_{t_f}(x(t_f),t_f)=0.}
  \label{eq:pmp-free-final-time-condition}
\end{equation}
When $\Phi$ has no explicit time dependence, $\Phi_{t_f}=0$, and the condition
reduces to $H(t_f)=0$. Intuitively, at the optimal stopping time, the
first-order cost of continuing for a little longer must be zero; otherwise,
moving the endpoint slightly earlier or later would improve the objective.

If the terminal state must lie on a constraint manifold, or if the terminal
time itself is constrained, then $z(t_f)$ and $\delta t_f$ are no longer
independent. The corresponding endpoint multipliers must be included before
the coefficients are set to zero.

\section{Worked Robotics and Control Problems}

The examples in this section use the same sequence repeatedly:
\begin{enumerate}
  \item write the dynamics and performance objective;
  \item form the Hamiltonian;
  \item derive the state, costate, and pointwise control conditions;
  \item use the endpoint conditions to determine the unknown constants; and
  \item interpret the resulting control physically.
\end{enumerate}
The algebra is useful, but the larger purpose is to see how the form of the
Hamiltonian predicts the qualitative structure of the solution. A quadratic
control cost tends to produce a smooth interior control, whereas a Hamiltonian
that is affine in a bounded control tends to produce saturation and switching.

\subsection{Minimum-Energy Motion of a Double Integrator}

Consider one translational coordinate of a robot with acceleration control:
\begin{equation}
  \dot q=v,
  \qquad
  \dot v=u.
  \label{eq:double-integrator-dynamics}
\end{equation}
This model describes a point mass after its mass has been absorbed into the
control variable, or a single robot coordinate when acceleration can be
commanded directly. The position $q$ cannot jump, and the velocity $v$ can
change only through the acceleration $u$.

Suppose the duration $T$ is prescribed and both position and velocity are
specified at the two endpoints. Among all feasible motions, we seek the one
with minimum squared acceleration:
\begin{equation}
  \minimize_u
  \quad
  \frac12\int_0^T u(t)^2\,dt
  \qquad
  \subjectto
  \qquad
  \begin{matrix}
    q(0)=q_0,&v(0)=v_0,\\
    q(T)=q_f,&v(T)=v_f.
  \end{matrix}
  \label{eq:double-integrator-minimum-energy-problem}
\end{equation}
The factor $1/2$ simplifies derivatives. The integral penalizes large
accelerations throughout the motion, so it favors a smooth distribution of
effort rather than a short, violent maneuver. Because the final state is
fixed, the terminal costate is not prescribed; its unknown value must be
chosen so that the state reaches $(q_f,v_f)$.

Write the state, dynamics vector field, and costate as
\begin{equation}
  x=
  \begin{bmatrix}q\\v\end{bmatrix},
  \qquad
  f(x,u)=
  \begin{bmatrix}v\\u\end{bmatrix},
  \qquad
  \lambda=
  \begin{bmatrix}\lambda_q\\\lambda_v\end{bmatrix}.
\end{equation}
The running cost is $\ell(x,u)=u^2/2$. Therefore, using
$H=\ell+\lambda\trans f$, the Hamiltonian is
\begin{equation}
  H(x,u,\lambda)
  =
  \frac12 u^2
  +
  \begin{bmatrix}\lambda_q&\lambda_v\end{bmatrix}
  \begin{bmatrix}v\\u\end{bmatrix}
  =
  \frac12 u^2+\lambda_q v+\lambda_v u.
\end{equation}
\paragraph{State equations.}
Differentiate the Hamiltonian with respect to the costate components:
\begin{align}
  \dot q
  &=H_{\lambda_q}
  =v,\\
  \dot v
  &=H_{\lambda_v}
  =u.
\end{align}

\paragraph{Costate equations.}
Differentiate the Hamiltonian with respect to the corresponding state
components and change the sign:
\begin{align}
  \dot\lambda_q
  &=-H_q
  =0,\\
  \dot\lambda_v
  &=-H_v
  =-\lambda_q.
\end{align}

\paragraph{Control condition.}
Because the control is unconstrained, an interior optimum must satisfy the
stationarity condition
\begin{equation}
  0=H_u=u+\lambda_v.
\end{equation}
Therefore
\begin{equation}
  u^\star=-\lambda_v.
\end{equation}

\paragraph{Boundary conditions.}
Both components of the state are fixed at the initial and terminal times:
\begin{equation}
  q(0)=q_0,
  \qquad
  v(0)=v_0,
  \qquad
  q(T)=q_f,
  \qquad
  v(T)=v_f.
\end{equation}
Because the terminal state is fixed, no terminal values are prescribed for
$\lambda_q(T)$ or $\lambda_v(T)$. The unknown costate data must instead be
chosen so that the state trajectory satisfies the two terminal conditions.

The equation $\dot\lambda_q=0$ makes $\lambda_q$ constant. It follows from
$\dot\lambda_v=-\lambda_q$ that $\lambda_v$, and hence
$u^\star=-\lambda_v$, is affine in time. Integrating the state equations then
shows that the optimal velocity is quadratic and the optimal position is
cubic. Writing the trajectory in a form that automatically satisfies the
initial conditions gives
\begin{equation}
  q(t)=q_0+v_0t+a_2t^2+a_3t^3,
  \qquad
  v(t)=v_0+2a_2t+3a_3t^2,
  \qquad
  u(t)=2a_2+6a_3t,
  \label{eq:cubic-minimum-energy-trajectory}
\end{equation}
The two final conditions produce the linear equations
\begin{align}
  q_f-q_0-v_0T&=a_2T^2+a_3T^3,\\
  v_f-v_0&=2a_2T+3a_3T^2.
\end{align}
Solving them gives
\begin{align}
  a_2
  &=\frac{3(q_f-q_0)-(2v_0+v_f)T}{T^2},\\
  a_3
  &=\frac{2(q_0-q_f)+(v_0+v_f)T}{T^3}.
  \label{eq:cubic-minimum-energy-coefficients}
\end{align}
The associated costates can be read from the optimal control:
\begin{equation}
  \lambda_v(t)=-2a_2-6a_3t,
  \qquad
  \lambda_q=6a_3.
\end{equation}
Thus the constants chosen to satisfy the endpoint conditions are exactly the
unknown costate data in the two-point boundary-value problem.

For the particularly transparent rest-to-rest case
$q_0=0$, $q_f=d$, and $v_0=v_f=0$,
\begin{equation}
  q(t)
  =d\left[3\left(\frac{t}{T}\right)^2
          -2\left(\frac{t}{T}\right)^3\right],
  \qquad
  u(t)=\frac{6d}{T^2}\left(1-2\frac{t}{T}\right).
  \label{eq:rest-to-rest-cubic-motion}
\end{equation}
The robot accelerates during the first half and brakes during the second, but
the acceleration changes continuously rather than jumping between actuator
limits. The minimum cost is
\begin{equation}
  J^\star=\frac{6d^2}{T^3}.
\end{equation}
This scaling exposes a useful design tradeoff: doubling the travel distance
multiplies the effort by four, while doubling the available time divides it by
eight.

The formulas apply componentwise to a point robot in $\R^d$. The optimization
is convex: the dynamics and endpoint constraints are linear, and the control
cost is strictly convex. Therefore the stationary trajectory derived above is
the unique global minimizer. Cubic interpolation is common in robotics because
it matches position and velocity while minimizing squared acceleration under
this model.

\subsection{Minimum-Time Motion and Bang--Bang Control}

The previous example fixed the duration and minimized effort. We now reverse
the priorities: effort is not charged directly, and the objective is to arrive
as quickly as possible. A control bound is essential. Without it, the model
could request an arbitrarily large acceleration and drive the predicted travel
time toward zero.

For the same double integrator, consider
\begin{equation}
  \minimize_{u,T}\quad T
  \qquad\subjectto\qquad
  \dot q=v,
  \quad\dot v=u,
  \quad |u|\leq u_{\max},
  \label{eq:double-integrator-minimum-time-problem}
\end{equation}
with specified initial and final states. Write
\begin{equation}
  x=
  \begin{bmatrix}q\\v\end{bmatrix},
  \qquad
  f(x,u)=
  \begin{bmatrix}v\\u\end{bmatrix},
  \qquad
  \lambda=
  \begin{bmatrix}\lambda_q\\\lambda_v\end{bmatrix}.
\end{equation}
Since $T=\int_0^T 1\,dt$, the running cost is $\ell(x,u)=1$. Therefore
\begin{equation}
  H(x,u,\lambda)
  =
  1+\lambda\trans f(x,u)
  =
  1+\lambda_q v+\lambda_v u.
\end{equation}

\paragraph{State equations.}
Differentiate the Hamiltonian with respect to the costate components:
\begin{align}
  \dot q
  &=H_{\lambda_q}
  =v,\\
  \dot v
  &=H_{\lambda_v}
  =u.
\end{align}

\paragraph{Costate equations.}
Differentiate the Hamiltonian with respect to the state components and change
the sign:
\begin{align}
  \dot\lambda_q
  &=-H_q
  =0,\\
  \dot\lambda_v
  &=-H_v
  =-\lambda_q.
\end{align}

\paragraph{Control condition.}
The control is constrained to $[-u_{\max},u_{\max}]$, so the correct condition
is pointwise Hamiltonian minimization:
\begin{equation}
  u^\star(t)
  \in
  \argmin_{|u|\leq u_{\max}}
  H(x^\star(t),u,\lambda^\star(t)).
\end{equation}
Unlike the minimum-energy Hamiltonian, this Hamiltonian contains no $u^2$
term; its control-dependent part is simply $\lambda_v u$. Minimizing this
linear function over the admissible interval gives
\begin{equation}
  u^\star(t)
  =
  \begin{cases}
    -u_{\max},&\lambda_v(t)>0,\\
    \phantom{-}u_{\max},&\lambda_v(t)<0,\\
    \text{any }u\in[-u_{\max},u_{\max}],&\lambda_v(t)=0.
  \end{cases}
  \label{eq:double-integrator-bang-bang-law}
\end{equation}

\paragraph{Boundary conditions.}
The initial and terminal states are prescribed:
\begin{equation}
  q(0)=q_0,
  \qquad
  v(0)=v_0,
  \qquad
  q(T)=q_f,
  \qquad
  v(T)=v_f.
\end{equation}
Because the terminal state is fixed, no terminal values are prescribed for
$\lambda_q(T)$ or $\lambda_v(T)$. Their unknown values must make the state
reach $(q_f,v_f)$.

\paragraph{Free-final-time condition.}
The final time is free and there is no terminal cost. Therefore the additional
transversality condition is
\begin{equation}
  H(T)=0.
\end{equation}

The equation $\dot\lambda_q=0$ makes $\lambda_q$ constant, and
$\dot\lambda_v=-\lambda_q$ then makes $\lambda_v$ affine in time.
The function $\lambda_v(t)$ is called the \emph{switching function}. Because
it is affine, it crosses zero at most once unless it vanishes identically.
Thus an ordinary extremal has at most one switch: use one extreme acceleration
before the zero and the opposite extreme afterward.

At an isolated switching time $t_s$, one has $\lambda_v(t_s)=0$, so the
control-dependent term $\lambda_v(t_s)u$ vanishes. Every value
$u(t_s)\in[-u_{\max},u_{\max}]$ therefore minimizes the Hamiltonian at that
single instant. This apparent ambiguity has no effect on the motion: changing
a bounded control at only one time does not change its time integral and hence
does not change the resulting state trajectory. The signs of $\lambda_v$
immediately before and after $t_s$ determine the two control values on the
neighboring intervals.

A different situation would arise if $\lambda_v(t)=0$ throughout a nonzero
time interval. Then first-order Hamiltonian minimization would leave the
control undetermined on an entire interval. Such an interval is called a
\emph{singular arc}, and additional differentiation of the switching function
is generally needed to determine the control. Singular arcs cannot occur in
this example. Because $\lambda_v$ is affine, vanishing on an interval would
force $\lambda_v\equiv0$ and $\lambda_q\equiv0$ everywhere. The Hamiltonian
would then equal $H=1$, contradicting the free-final-time condition $H(T)=0$.

Consider rest-to-rest travel from $(q,v)=(0,0)$ to $(d,0)$ with $d>0$.
The robot must first accelerate in the positive direction and then brake:
\begin{equation}
  u^\star(t)
  =
  \begin{cases}
    u_{\max},&0\leq t<t_s,\\
    -u_{\max},&t_s<t\leq T.
  \end{cases}
\end{equation}
The velocity gained before the switch is $u_{\max}t_s$. Braking for the
remaining time removes velocity $u_{\max}(T-t_s)$. The terminal condition
$v(T)=0$ therefore gives
\begin{equation}
  t_s=T-t_s,
  \qquad t_s=\frac{T}{2}.
\end{equation}
The velocity profile is triangular, with peak value
\begin{equation}
  v_{\mathrm{peak}}
  =u_{\max}\frac{T}{2}.
\end{equation}
Distance is the area under this velocity profile, so
\begin{equation}
  d=\frac12 T v_{\mathrm{peak}}
   =\frac{u_{\max}T^2}{4},
  \qquad
  \boxed{T^\star=2\sqrt{\frac{d}{u_{\max}}}.}
  \label{eq:double-integrator-minimum-time}
\end{equation}
Consequently
\begin{equation}
  v_{\mathrm{peak}}=\sqrt{du_{\max}}.
\end{equation}
For this simple rest-to-rest problem, the kinematic argument already
determines the switching time and optimal duration, but it does not determine
the magnitude of the costate. Indeed, the control law depends only on the sign
of $\lambda_v$: multiplying both costate components by the same positive
constant would leave the switching time and bang--bang control unchanged.
The free-final-time condition fixes this remaining scale.

\paragraph{Determining the costate.}
At the terminal time, $v(T)=0$ and $u(T)=-u_{\max}$. Hence
\begin{equation}
  0=H(T)
  =1+\lambda_q(T)v(T)+\lambda_v(T)u(T)
  =1-\lambda_v(T)u_{\max},
\end{equation}
so
\begin{equation}
  \lambda_v(T)=\frac{1}{u_{\max}}.
\end{equation}
The switching condition gives $\lambda_v(t_s)=0$, and $\lambda_v$ is affine.
Since $T-t_s=T/2$ and
$v_{\mathrm{peak}}=u_{\max}T/2$, it follows that
\begin{equation}
  \lambda_v(t)
  =
  \frac{t-t_s}{v_{\mathrm{peak}}},
  \qquad
  \lambda_q
  =
  -\dot\lambda_v
  =
  -\frac{1}{v_{\mathrm{peak}}}.
\end{equation}
Thus the free-final-time condition selects the unique costate magnitude
compatible with the normalization of the running cost $\ell=1$.

\paragraph{Meaning of the costate.}
The costate measures the first-order sensitivity of the optimal remaining
travel time to the current state. Under the usual smoothness conditions,
\begin{equation}
  \lambda_q=\frac{\partial V}{\partial q},
  \qquad
  \lambda_v=\frac{\partial V}{\partial v},
\end{equation}
where $V(q,v)$ is the minimum time still required to reach $(d,0)$. Here
$\lambda_q<0$: moving the robot slightly farther toward the target decreases
the remaining travel time. Before the switch, $\lambda_v<0$, so a small
increase in forward velocity also decreases the remaining time. After the
switch, $\lambda_v>0$: extra forward velocity increases the remaining time
because the robot must spend longer braking. At the switching time,
$\lambda_v=0$, so the first-order value of an additional velocity perturbation
changes sign.

The contrast with the preceding example is fundamental. A quadratic effort
term makes the Hamiltonian strictly convex in $u$ and produces a smooth
interior minimizer. A time objective makes the Hamiltonian affine in $u$, so
the fastest solution uses all available actuator authority. Bang--bang
behavior is therefore a consequence of the model and objective, not a
numerical artifact. Additional speed, position, or comfort constraints can
introduce further arcs on which those constraints are active.

\subsection{Infinite-Horizon Regulation of a Scalar Integrator}

The first two examples produced one trajectory over a particular task. We now
consider a continuing regulation problem and obtain a stationary feedback
controller. Let
\begin{equation}
  \dot x=u
\end{equation}
and penalize state error and control effort over the infinite horizon:
\begin{equation}
  V_\infty(x_0)
  =
  \inf_{u(\cdot)}
  \frac12\int_0^\infty
  \left(q\,x(t)^2+r\,u(t)^2\right)\,dt,
  \qquad q>0,\quad r>0.
  \label{eq:scalar-integrator-infinite-horizon-problem}
\end{equation}
The state penalty asks the controller to drive $x$ toward zero, while the
control penalty prevents an arbitrarily aggressive correction. The
Hamiltonian is
\begin{equation}
  H=\frac12 q\,x^2+\frac12 r\,u^2+\lambda u.
\end{equation}
The PMP conditions give
\begin{align}
  \dot x&=u,\\
  \dot\lambda&=-q\,x,\\
  0=H_u&=r\,u+\lambda,
\end{align}
and hence
\begin{equation}
  u^\star=-\frac{1}{r}\lambda.
  \label{eq:scalar-integrator-control-from-costate}
\end{equation}
These equations describe a two-dimensional state--costate system. The
infinite-horizon requirement selects the solution that approaches the origin
and accumulates finite cost.

Substituting $u^\star=-\lambda/r$ into the state equation couples the state
and costate into the linear system
\begin{equation}
  \begin{bmatrix}
    \dot x\\
    \dot\lambda
  \end{bmatrix}
  =
  \begin{bmatrix}
    0 & -1/r\\
    -q & 0
  \end{bmatrix}
  \begin{bmatrix}
    x\\
    \lambda
  \end{bmatrix}.
  \label{eq:scalar-integrator-state-costate-matrix-system}
\end{equation}
Let
\begin{equation}
  \alpha=\sqrt{\frac{q}{r}}.
\end{equation}
The system matrix has eigenvalues $-\alpha$ and $+\alpha$. Corresponding
eigenvectors may be chosen as
\begin{equation}
  v_s=
  \begin{bmatrix}
    1\\ \sqrt{q r}
  \end{bmatrix},
  \qquad
  v_u=
  \begin{bmatrix}
    1\\ -\sqrt{q r}
  \end{bmatrix},
\end{equation}
where $v_s$ is the stable mode and $v_u$ is the unstable mode. The general
state--costate solution is therefore
\begin{equation}
  \begin{bmatrix}
    x(t)\\ \lambda(t)
  \end{bmatrix}
  =c_s v_s e^{-\alpha t}
   +c_u v_u e^{\alpha t}.
  \label{eq:scalar-integrator-general-state-costate-solution}
\end{equation}

If $c_u\neq0$, the state contains an exponentially growing term, so it does
not approach the origin and the integral of $q x(t)^2$ is infinite. A
finite-cost infinite-horizon solution must therefore satisfy $c_u=0$. Since
$x(0)=x_0$, the remaining coefficient is $c_s=x_0$, and hence
\begin{equation}
  x^\star(t)=x_0e^{-\alpha t},
  \qquad
  \lambda^\star(t)=\sqrt{q r}\,x_0e^{-\alpha t}.
\end{equation}
Thus the stable state--costate trajectory satisfies
\begin{equation}
  \lambda^\star(t)=\sqrt{q r}\,x^\star(t).
\end{equation}
Substituting this relation into $u^\star=-\lambda/r$ gives
\begin{equation}
  \boxed{u^\star=-\sqrt{\frac{q}{r}}\,x.}
  \label{eq:scalar-integrator-infinite-horizon-feedback}
\end{equation}
This is a stationary state-feedback law defined for every scalar state, not
merely a time signal along one precomputed trajectory. Increasing $q$ makes
state error more expensive and strengthens the feedback; increasing $r$ makes
control more expensive and weakens it.

The corresponding cost-to-go is
\begin{equation}
  \boxed{V_\infty(x)=\frac12\sqrt{q r}\,x^2.}
  \label{eq:scalar-integrator-infinite-horizon-value}
\end{equation}
Its gradient is
$V_{\infty,x}=\sqrt{q r}\,x=\lambda$, matching the costate along every optimal
closed-loop trajectory. This example produces a global feedback law because
the value function has a simple quadratic form. Chapter~7 generalizes the
calculation to multidimensional linear systems, where the scalar equation
$P^2=q r$ becomes an algebraic Riccati equation.

\subsection{Torque-Optimal Motion of a Manipulator}

The double integrator hides the configuration dependence and coupling present
in a real robot. Return to the manipulator dynamics in
\cref{eq:robot-control-state-space}. Consider the running cost
\begin{equation}
  \ell(q,v,u)
  =\frac12(q-q_d)\trans Q(q-q_d)
   +\frac12 v\trans Sv
   +\frac12 u\trans Ru,
  \qquad R\succ0.
\end{equation}
The three terms have distinct roles. The matrix $Q\succeq0$ penalizes
configuration error, $S\succeq0$ penalizes velocity, and $R\succ0$ penalizes
joint torque. Large entries of $Q$ demand accurate positioning; large entries
of $R$ make the optimizer reluctant to use the corresponding actuators. A
terminal cost or a fixed terminal state must also be specified according to
the task, although the pointwise control calculation below is the same.

Recall that the state and dynamics are
\begin{equation}
  x=
  \begin{bmatrix}
    q\\ v
  \end{bmatrix},
  \qquad
  \dot x=f(x,u)
  =
  \begin{bmatrix}
    v\\
    M(q)^{-1}\bigl(u-h(q,v)-g(q)\bigr)
  \end{bmatrix}.
  \label{eq:manipulator-dynamics-recalled-for-pmp}
\end{equation}
Partition the costate as $(\lambda_q,\lambda_v)$. The components
$\lambda_q$ and $\lambda_v$ measure the future-cost sensitivity to small
changes in configuration and velocity. The Hamiltonian is
\begin{align}
  H
  &={}
  \ell(q,v,u)+\lambda_q\trans v\\
  &\quad+
  \lambda_v\trans M(q)^{-1}
  \bigl(u-h(q,v)-g(q)\bigr).
  \label{eq:manipulator-control-hamiltonian}
\end{align}
\paragraph{State equations.}
Differentiating the Hamiltonian with respect to the two costate components
recovers the robot dynamics:
\begin{align}
  \dot q
  &=H_{\lambda_q}
  =v,\\
  \dot v
  &=H_{\lambda_v}
  =M(q)^{-1}\bigl(u-h(q,v)-g(q)\bigr).
  \label{eq:manipulator-pmp-state-equations}
\end{align}

\paragraph{Costate equations.}
The costates evolve according to
$\dot\lambda_q=-H_q$ and $\dot\lambda_v=-H_v$. Thus
\begin{align}
  \dot\lambda_q
  &=-Q(q-q_d)
    -\frac{\partial}{\partial q}
      \left[
        \lambda_v\trans M(q)^{-1}
        \bigl(u-h(q,v)-g(q)\bigr)
      \right],\\
  \dot\lambda_v
  &=-Sv-\lambda_q
    -\frac{\partial}{\partial v}
      \left[
        \lambda_v\trans M(q)^{-1}
        \bigl(u-h(q,v)-g(q)\bigr)
      \right].
  \label{eq:manipulator-pmp-costates}
\end{align}
These derivatives account for the facts that inertia changes with
configuration, velocity-dependent forces change with motion, and gravity
changes with pose.

\paragraph{Control condition.}
Only two terms in the Hamiltonian depend directly on $u$:
\begin{equation}
  \frac12 u\trans Ru
  +\lambda_v\trans M(q)^{-1}u.
\end{equation}
The first is the immediate effort cost. The second measures how the torque
changes acceleration and therefore future cost. For unconstrained torque,
the pointwise stationarity condition is
\begin{equation}
  0=H_u=Ru+M(q)^{-\mathsf T}\lambda_v,
\end{equation}
and hence
\begin{equation}
  \boxed{u^\star=-R^{-1}M(q)^{-\mathsf T}\lambda_v.}
  \label{eq:manipulator-pmp-control}
\end{equation}
Because the robot mass matrix is symmetric,
$M^{-\mathsf T}=M^{-1}$. Nevertheless, the transpose notation shows how the
derivative of $M^{-1}u$ maps a velocity costate into actuator coordinates.

\paragraph{Boundary conditions.}
The initial state is prescribed:
\begin{equation}
  q(t_0)=q_0,
  \qquad
  v(t_0)=v_0.
\end{equation}
The terminal conditions depend on the task specification. If the terminal
state is fixed, then
\begin{equation}
  q(t_f)=q_f,
  \qquad
  v(t_f)=v_f,
\end{equation}
and no terminal values are prescribed for $\lambda_q(t_f)$ or
$\lambda_v(t_f)$; their unknown values must make the state reach the specified
endpoint. If the terminal state is free and the problem includes a terminal
cost $\Phi(q(t_f),v(t_f),t_f)$, then
\begin{equation}
  \lambda_q(t_f)=\Phi_q,
  \qquad
  \lambda_v(t_f)=\Phi_v.
\end{equation}
In particular, a free terminal state with no terminal cost gives
$\lambda_q(t_f)=\lambda_v(t_f)=0$. If the final time is also free, the
additional condition is
\begin{equation}
  H(t_f)+\Phi_{t_f}=0.
\end{equation}
The state equations, costate equations, control condition, and applicable
boundary conditions together form a nonlinear two-point boundary-value
problem.

The control formula has a simple interpretation. If a component of
$M^{-\mathsf T}\lambda_v$ is large and positive, increasing the corresponding
torque greatly increases future cost, so the minimizing torque acts in the
negative direction. The matrix $R^{-1}$ scales this response: an actuator
assigned a large effort penalty is used less aggressively.

It is tempting to read \cref{eq:manipulator-pmp-control} as a feedback law,
but $\lambda_v$ is not a measured robot state. It depends on the future
objective and endpoint conditions and is known only after solving the full
optimal-control problem. Solving the two-point boundary-value problem for one
initial condition produces three linked time histories:
\begin{equation}
  x^\star(t),
  \qquad
  \lambda^\star(t),
  \qquad
  u^\star(t)
  =-R^{-1}M(q^\star(t))^{-\mathsf T}\lambda_v^\star(t).
\end{equation}
This is an open-loop plan for the particular initial condition used in the
optimization. For example, suppose an external force changes the robot's
configuration or velocity at some time $t_d$. The stored costate
$\lambda_v^\star(t_d)$ belongs to the nominal state $x^\star(t_d)$, not to the
disturbed state. Continuing to apply the precomputed torque therefore does not
specify how the robot should optimally recover.

A feedback controller needs the costate, or equivalently the optimal control,
as a function of the state near or away from the nominal trajectory. If a
differentiable value function is known, then along an optimal trajectory
\begin{equation}
  \lambda(t)=V_x(t,x(t)),
\end{equation}
so the velocity component $V_v$ gives the feedback law
\begin{equation}
  u^\star(t,x)
  =-R^{-1}M(q)^{-\mathsf T}V_v(t,x).
\end{equation}

This example illustrates the typical role of PMP in nonlinear robot control.
The state and costate equations, the Hamiltonian-minimizing torque, and the
endpoint conditions must be solved together; the torque formula alone is not
a complete solution. Quadratic effort produces an explicit pointwise control
condition, but the required costate still carries global information about the
future objective. Solving for one initial condition therefore gives an optimal
open-loop motion, whereas constructing feedback requires a value function, a
local approximation such as a Riccati method, or repeated online replanning.

\subsection{Energy-Efficient Steering of a Mobile Robot}

Consider a planar mobile robot with state and control
\begin{equation}
  x=
  \begin{bmatrix}
    p_x\\ p_y\\ \theta
  \end{bmatrix},
  \qquad
  u=
  \begin{bmatrix}
    v\\ \omega
  \end{bmatrix},
\end{equation}
where $(p_x,p_y)$ is the robot position, $\theta$ is its heading, $v$ is its
signed forward speed, and $\omega$ is its angular speed. The unicycle dynamics
are
\begin{equation}
  \dot x=f(x,u)
  =
  \begin{bmatrix}
    v\cos\theta\\
    v\sin\theta\\
    \omega
  \end{bmatrix}.
  \label{eq:unicycle-pmp-dynamics}
\end{equation}
The translational velocity is constrained to point along the robot's current
heading $(\cos\theta,\sin\theta)$. The robot cannot move instantaneously
sideways. The control $\omega$ changes the heading, so turning now changes
which translational directions will become available later. This nonholonomic
coupling makes the dynamics nonlinear even though the model contains no
inertia matrix.

Suppose the initial and terminal poses are prescribed over a fixed interval,
and use the running cost
\begin{equation}
  \ell=\frac12 r_v\,v^2+\frac12 r_\omega\,\omega^2,
  \qquad r_v,r_\omega>0.
\end{equation}
The weights $r_v$ and $r_\omega$ determine the relative price of translation
and turning. This running cost does not penalize position error during the
motion; the desired destination is imposed by the terminal constraint.

Introduce the costate
$\lambda=(\lambda_x,\lambda_y,\lambda_\theta)$. The Hamiltonian is
\begin{equation}
  H
  =\frac12 r_v\,v^2+\frac12 r_\omega\,\omega^2
   +\lambda_x\,v\cos\theta
   +\lambda_y\,v\sin\theta
   +\lambda_\theta\omega.
\end{equation}

\paragraph{State equations.}
Differentiating the Hamiltonian with respect to the costates recovers the
unicycle dynamics:
\begin{align}
  \dot p_x
  &=H_{\lambda_x}
  =v\cos\theta,\\
  \dot p_y
  &=H_{\lambda_y}
  =v\sin\theta,\\
  \dot\theta
  &=H_{\lambda_\theta}
  =\omega.
  \label{eq:unicycle-pmp-state-equations}
\end{align}

\paragraph{Costate equations.}
The costates satisfy $\dot\lambda=-H_x$:
\begin{align}
  \dot\lambda_x
  &=-H_{p_x}
  =0,\\
  \dot\lambda_y
  &=-H_{p_y}
  =0,\\
  \dot\lambda_\theta
  &=-H_\theta
  =v(\lambda_x\sin\theta-\lambda_y\cos\theta).
  \label{eq:unicycle-pmp-costate-equations}
\end{align}
Because the dynamics and running cost are invariant under translations in
$p_x$ and $p_y$, the associated costates $\lambda_x$ and $\lambda_y$ are
constant. This is a control-theoretic analogue of a conserved quantity. In
contrast, $\lambda_\theta$ evolves because changing the heading rotates the
direction in which the translational control acts.

\paragraph{Control conditions.}
The Hamiltonian is strictly convex and separable in $v$ and $\omega$.
For unconstrained controls, pointwise stationarity requires
\begin{equation}
  0=H_v
  =r_v\,v+\lambda_x\cos\theta+\lambda_y\sin\theta,
  \qquad
  0=H_\omega
  =r_\omega\omega+\lambda_\theta.
\end{equation}
Therefore
\begin{align}
  v^\star
  &=-\frac{\lambda_x\cos\theta+\lambda_y\sin\theta}{r_v},\\
  \omega^\star
  &=-\frac{\lambda_\theta}{r_\omega}.
  \label{eq:unicycle-pmp-controls}
\end{align}
The translational control depends on the projection of the planar costate
$(\lambda_x,\lambda_y)$ onto the robot's heading. Only this component can be
influenced immediately because the unicycle cannot move sideways. The angular
control is proportional to $-\lambda_\theta$: if a small positive heading
change would increase future cost, the optimizer commands negative angular
velocity.

\paragraph{Boundary conditions.}
For prescribed initial and terminal poses over a fixed interval, the state
boundary conditions are
\begin{align}
  p_x(t_0)&=p_{x,0},
  &p_y(t_0)&=p_{y,0},
  &\theta(t_0)&=\theta_0,\\
  p_x(t_f)&=p_{x,f},
  &p_y(t_f)&=p_{y,f},
  &\theta(t_f)&=\theta_f.
\end{align}
Because every terminal-state component is fixed, no terminal value is
prescribed for $\lambda_x(t_f)$, $\lambda_y(t_f)$, or
$\lambda_\theta(t_f)$. Their unknown values must make the state reach the
specified terminal pose. Equivalently, one may guess the three initial
costates, integrate the coupled state and costate equations, and adjust the
guess until all three terminal-state residuals vanish. This is a nonlinear
shooting problem, which is explained in the next section.

For a geometric interpretation, define
\begin{equation}
  e(\theta)
  =\begin{bmatrix}\cos\theta\\\sin\theta\end{bmatrix},
  \qquad
  e_\perp(\theta)
  =\begin{bmatrix}-\sin\theta\\\cos\theta\end{bmatrix},
  \qquad
  \lambda_p
  =\begin{bmatrix}\lambda_x\\\lambda_y\end{bmatrix}.
\end{equation}
Then
\begin{equation}
  v^\star=-\frac{\lambda_p\trans e}{r_v},
  \qquad
  \dot\lambda_\theta=-v\lambda_p\trans e_\perp.
\end{equation}
The first equation uses the costate component along the heading; the second
measures the component that is currently unreachable sideways. Turning changes
the heading so that a valuable planar direction can become reachable later.
This is precisely why the heading costate cannot remain constant in general.

Multiple extremals may connect the same poses: for example, the robot may turn
left or right, or it may reverse if signed $v$ is allowed. PMP identifies
candidate extremals, and their costs must be compared to determine which
candidate is best.

If the controls are bounded independently, the pointwise minimizers are
projections of the unconstrained expressions:
\begin{align}
  v^\star
  &=\Pi_{[v_{\min},v_{\max}]}
    \left(-\frac{\lambda_x\cos\theta+\lambda_y\sin\theta}{r_v}\right),\\
  \omega^\star
  &=\Pi_{[-\omega_{\max},\omega_{\max}]}
    \left(-\frac{\lambda_\theta}{r_\omega}\right).
\end{align}
Choosing $v_{\min}=0$ forbids reverse motion and can change the qualitative
structure of the solution. Obstacle constraints are harder: they are state
constraints active over time and introduce additional multipliers or boundary
arcs, which is one reason discretized trajectory-optimization methods are
often preferred for motion planning in cluttered environments.

\section{Numerical Methods}

Closed-form solutions are exceptional. For realistic nonlinear systems, the PMP
produces coupled state--costate boundary-value problems that must be solved
numerically. This section first distinguishes direct and indirect methods, then
focuses on numerical methods for solving the PMP equations.

\subsection{Two Numerical Viewpoints}

An \emph{indirect method} first derives the PMP conditions and then solves the
state equation, costate equation, Hamiltonian minimization condition, and
boundary conditions.
The numerical unknowns may include costate boundary values, switching times,
and endpoint multipliers. The result is an \emph{extremal}: it satisfies
necessary conditions but is not automatically a minimum.

A \emph{direct method} instead discretizes the original state and control
trajectories, objective, and constraints before optimization. It produces a
finite-dimensional
nonlinear program without explicitly deriving the costate equations. Direct
methods, including transcription and direct collocation, are developed in
Chapter~9 on discrete-time trajectory optimization. Here we concentrate on
indirect methods.

\subsection{Single Shooting}

In \emph{single shooting}, Hamiltonian minimization gives
$u=\mu(t,x,\lambda)$ and the unknown initial costate $\lambda_0$ is guessed.
One integrates
\begin{align}
  \dot x&=f(t,x,\mu(t,x,\lambda)),\\
  \dot\lambda&=-H_x(t,x,\mu(t,x,\lambda),\lambda).
  \label{eq:shooting-state-costate-system}
\end{align}
For fixed endpoints, the shooting residual is
\begin{equation}
  F(\lambda_0)=x(t_f;\lambda_0)-x_f.
  \label{eq:shooting-residual}
\end{equation}
Root finding adjusts $\lambda_0$ until $F=0$. For free terminal states,
endpoint constraints, or free final time, the residual is enlarged with the
corresponding transversality conditions. A typical iteration therefore
consists of the following steps:
\begin{enumerate}
  \item guesses the unknown initial costate and other boundary variables;
  \item integrates the state--costate equations to $t_f$;
  \item evaluates the terminal residual; and
  \item updates the guess using Newton, quasi-Newton, or another root solver.
\end{enumerate}
Single shooting has few numerical unknowns and can be highly accurate.
However, a small error in $\lambda_0$ may produce a large terminal error on a
long or unstable trajectory. Control switching can also make the residual
nonsmooth.

\subsection{Multiple Shooting}

Multiple shooting partitions the interval,
\begin{equation}
  t_0=\tau_0<\tau_1<\cdots<\tau_N=t_f,
\end{equation}
and assigns an independent initial value
$y_j(\tau_j)$, with $y=(x,\lambda)$, to each segment $j=0,\ldots,N-1$.
The PMP equations are
integrated separately on every segment, and continuity is enforced through
\begin{equation}
  y_j(\tau_{j+1}^-)-y_{j+1}(\tau_{j+1}^+)=0,
  \qquad j=0,\ldots,N-2.
\end{equation}
The endpoint conditions are imposed together with these matching equations.
Short integrations reduce sensitivity and make multiple shooting more robust
than single shooting, at the cost of additional variables and continuity
constraints.

\subsection{Indirect Collocation}

Collocation avoids repeated integration from one boundary guess. The
state--costate trajectory is approximated by piecewise polynomials on a time
mesh. At selected collocation points $t_{k,i}$, one enforces
\begin{equation}
  \dot y_h(t_{k,i})
  =G\bigl(t_{k,i},y_h(t_{k,i})\bigr),
  \qquad
  y=\begin{bmatrix}x\\\lambda\end{bmatrix},
\end{equation}
where $G$ contains the PMP state and costate equations after Hamiltonian
minimization. Boundary and continuity conditions are imposed simultaneously.
The polynomial coefficients or nodal values are then found by solving a
nonlinear algebraic system.

This remains an indirect method because PMP is derived before discretization.
By contrast, direct collocation discretizes the original dynamics, objective,
and constraints before deriving optimality conditions. Indirect collocation is
often less sensitive than shooting, but it introduces more unknowns and
requires every PMP condition to be encoded correctly.

\subsection{Infinite-Horizon Indirect Methods}

Infinite-horizon PMP retains the interior equations but replaces a finite
terminal condition with asymptotic behavior. A common approximation truncates
the horizon at a large $T$ and imposes a terminal state, terminal cost, or
stable state--costate relation near the desired equilibrium. The early
trajectory should become insensitive to further increases in $T$.

Local Riccati solutions often supply the terminal relation. For discounted
problems, the current-value costate satisfies
\begin{equation}
  \dot\lambda=\rho\lambda-H_x^{\mathrm c}.
\end{equation}
Discounting reduces the influence of distant costs but does not eliminate the
need for a suitable tail condition. Infinite-horizon shooting is especially
sensitive because an incorrect costate can excite unstable modes.

\subsection{Other Indirect Techniques}

Several auxiliary techniques improve the reliability of shooting and
collocation. In \emph{continuation} or \emph{homotopy}, one first solves an
easier nearby problem and then gradually changes the horizon, boundary data,
cost weights, or constraints toward the desired problem, using each solution
to initialize the next. For a bang--bang problem with a known switching
structure, the switching times can instead be treated as unknowns and
determined from the endpoint and switching conditions. Standard boundary-value
and finite-difference solvers provide another option by discretizing the PMP
equations globally and solving the resulting nonlinear equations. These
approaches remain indirect because the optimality conditions are derived
before discretization or parameterization.

\subsection{Direct Methods: Outlook}

Direct methods optimize a discretized version of the original control problem
rather than the PMP equations. Common choices include control
parameterization, direct multiple shooting, and direct collocation. They
usually introduce more decision variables than indirect methods, but handle
control bounds, obstacles, and other path constraints more naturally. Their
detailed treatment is deferred to Chapter~9.

For infinite-horizon problems, direct methods still require truncation at a
large time $T$ together with a terminal cost or terminal set that approximates
the omitted tail. Without such a tail model, the optimizer may postpone
corrective action beyond the artificial endpoint.

\subsection{Choosing and Checking an Indirect Method}

Single shooting is attractive when the number of unknown boundary values is
small and the dynamics are not too sensitive. Multiple shooting is usually
more robust on long or unstable trajectories, while collocation is useful
when the whole state--costate curve must be resolved simultaneously.
Continuation can provide the initial guesses required by any of these
methods. None of them guarantees that a computed extremal is a minimum: the
PMP supplies necessary conditions, and a nonlinear problem may have several
extremals.

An indirect solution should therefore be checked by:
\begin{itemize}
  \item verifying the state and costate equations and all boundary conditions;
  \item checking pointwise Hamiltonian minimization and active constraints;
  \item refining the integration tolerance or collocation mesh;
  \item varying the initial guess and, for infinite horizons, the truncation
    time; and
  \item comparing the costs of distinct extremals or alternative numerical
    formulations.
\end{itemize}
Agreement across these tests is stronger evidence than one successful solver
run.

\paragraph{Exercises.}

\begin{exercise}
Explain in words the difference between $J[u]$ and the cost-to-go function
$V(t,x)$. What boundary condition does $V$ satisfy at $t=t_f$?
\end{exercise}
\begin{proof}[Short solution]
$J[u]$ evaluates one specified control signal from a specified initial
condition. The value $V(t,x)$ is the minimum future cost over all admissible
controls starting from each time--state pair $(t,x)$. At the terminal time,
$V(t_f,x)=\Phi(x,t_f)$.
\end{proof}

\begin{exercise}
For an autonomous, time-invariant problem, contrast the typical form of a
finite-horizon optimal feedback law with that of an infinite-horizon law.
Does a finite horizon necessarily mean open-loop control?
\end{exercise}
\begin{proof}[Short solution]
The finite-horizon law generally has the form $u=\mu^\star(t,x)$ because the
remaining time matters. A stationary infinite-horizon law has the form
$u=\mu_\infty^\star(x)$. Finite horizon does not imply open loop: its value
function also defines a time-varying state-feedback law.
\end{proof}

\begin{exercise}
Classify each of the following algorithms as direct or indirect:
(i) derive the PMP equations and shoot on the initial costate;
(ii) represent $u$ by piecewise-constant samples and optimize those samples;
and (iii) impose collocation equations on the state and costate equations
obtained from PMP.
\end{exercise}
\begin{proof}[Short solution]
Case (i) is indirect because it solves derived necessary conditions.
Case (ii) is direct because it parameterizes the original control problem
before optimization. Case (iii) is indirect because the PMP equations were
derived before they were discretized.
\end{proof}

\begin{exercise}
Why does a truncated infinite-horizon direct method benefit from a terminal
cost approximating $V_\infty(x(T))$?
\end{exercise}
\begin{proof}[Short solution]
The terminal cost approximates the cost omitted after the artificial endpoint
$T$. Without it, the optimizer may postpone state regulation or other costly
actions until just beyond the truncated horizon.
\end{proof}

\begin{exercise}
For $\dot x=ax+bu$ and
$J=\tfrac12 s x(T)^2+\tfrac12\int_0^T(q\,x^2+r\,u^2)\,dt$, write the Hamiltonian,
costate equation, minimizing control, and terminal condition.
\end{exercise}
\begin{proof}[Short solution]
The Hamiltonian is
$H=(q\,x^2+r\,u^2)/2+\lambda(a x+b u)$. Therefore
\begin{equation}
  \dot\lambda=-q\,x-a\lambda,
  \qquad
  u=-\frac{b}{r}\lambda,
  \qquad
  \lambda(T)=s x(T).
\end{equation}
\end{proof}

\begin{exercise}
For the minimum-energy integrator $\dot x=u$ with fixed endpoints
$x(0)=x_0$ and $x(T)=x_f$, find the optimal control.
\end{exercise}
\begin{proof}[Short solution]
With $H=u^2/2+\lambda u$, the costate and therefore $u=-\lambda$ are
constant. The endpoint condition gives
\begin{equation}
  u^\star=\frac{x_f-x_0}{T}.
\end{equation}
\end{proof}

\begin{exercise}
For $\dot x=f_0(x)+G(x)u$ and
$\ell(x,u)=L(x)+\tfrac12 u\trans Ru$, where $R\succ0$, derive the
unconstrained minimizing control.
\end{exercise}
\begin{proof}[Short solution]
Since $H_u=Ru+G(x)\trans\lambda$,
\begin{equation}
  u^\star=-R^{-1}G(x)\trans\lambda.
\end{equation}
\end{proof}

\begin{exercise}
Suppose $H=c(t,x,\lambda)u+d(t,x,\lambda)$ and
$u\in[-u_{\max},u_{\max}]$. Minimize $H$ with respect to $u$.
\end{exercise}
\begin{proof}[Short solution]
Choose $u=-u_{\max}$ if $c>0$ and $u=u_{\max}$ if $c<0$. If $c=0$, every
admissible control minimizes the instantaneous Hamiltonian and further
analysis is needed. The quantity $c$ is the switching function.
\end{proof}

\begin{exercise}
For the rest-to-rest minimum-time double integrator, find the peak speed.
\end{exercise}
\begin{proof}[Short solution]
The acceleration phase lasts
$T^\star/2=\sqrt{d/u_{\max}}$, so
\begin{equation}
  v_{\mathrm{peak}}=\sqrt{du_{\max}}.
\end{equation}
\end{proof}

\begin{exercise}
If the terminal state is free and $\Phi(x(T))=0$, what terminal condition does
the costate satisfy?
\end{exercise}
\begin{proof}[Short solution]
Equation \cref{eq:free-terminal-state-transversality} gives
$\lambda(T)=0$.
\end{proof}

\begin{exercise}
If the final time is free and the terminal cost has no explicit time dependence,
what additional condition holds?
\end{exercise}
\begin{proof}[Short solution]
The condition is $H(T)=0$. For a smooth autonomous extremal, $H$ is constant,
so $H(t)=0$ along the trajectory.
\end{proof}

\begin{exercise}
For the unicycle example, impose $|v|\leq v_{\max}$ and
$|\omega|\leq\omega_{\max}$. How do the controls change?
\end{exercise}
\begin{proof}[Short solution]
The Hamiltonian is separable and strictly convex in the two controls, so each
unconstrained minimizer is projected onto its interval:
\begin{align}
  v^\star
  &=\Pi_{[-v_{\max},v_{\max}]}
    \left(-\frac{\lambda_x\cos\theta+\lambda_y\sin\theta}{r_v}\right),\\
  \omega^\star
  &=\Pi_{[-\omega_{\max},\omega_{\max}]}
    \left(-\frac{\lambda_\theta}{r_\omega}\right).
\end{align}
\end{proof}

\begin{exercise}
Explain why a trajectory satisfying the state and costate differential
equations may still fail Pontryagin's minimum principle.
\end{exercise}
\begin{proof}[Short solution]
It may violate endpoint conditions, feasibility constraints, or pointwise
Hamiltonian minimization. Every part of the principle is required.
\end{proof}

\begin{exercise}
For fixed terminal state and $x\in\R^n$, what is the dimension of the basic
shooting residual $F(\lambda_0)$?
\end{exercise}
\begin{proof}[Short solution]
The unknown initial costate and terminal-state residual both have dimension
$n$, so $F:\R^n\to\R^n$.
\end{proof}

\paragraph{Chapter summary.}
Continuous-time optimal control chooses a control function whose state
trajectory satisfies a differential equation. The cost-to-go function gives
the best achievable future cost from every time--state pair and, when solved
over a state domain, defines an optimal feedback law. Finite-horizon value
functions generally depend on time and include a terminal condition;
autonomous infinite-horizon problems often seek stationary value functions and
feedback laws for continuing regulation. An optimal trajectory for one initial
state is not the same object as a feedback controller defined for many states. A costate
multiplier and integration by parts yield the Hamiltonian system: the state
evolves forward, the costate carries the gradient of future cost backward,
and the control minimizes the Hamiltonian pointwise. Endpoint freedom
produces transversality conditions. Quadratic effort often gives smooth
controls, whereas bounded controls in an affine Hamiltonian produce
bang--bang solutions. These necessary conditions can be solved by indirect
methods such as shooting or collocation of the state--costate equations.
Direct methods instead parameterize or transcribe the original control
problem into a finite-dimensional optimization problem. Both families can be
applied to finite horizons; infinite horizons require truncation, asymptotic
conditions, terminal-value approximations, or stationary structure. These
ideas underlie the Riccati, dynamic-programming, trajectory-optimization,
model-predictive-control, and policy-optimization methods developed later.
\chapter{Linear Quadratic Optimal Control}
\label{chap:linear-quadratic-optimal-control}
\chapteroverview{Finite- and infinite-horizon linear--quadratic regulation,
Riccati equations, optimal state feedback, reference tracking, connections to
PD, computed-torque, and PID control, and an outlook on estimation and LQG
control.}

Chapter~6 developed necessary conditions for general continuous-time optimal
control problems. Those conditions are powerful, but they usually lead to a
nonlinear two-point boundary-value problem that must be solved numerically.
This chapter studies an important exception: linear dynamics with a quadratic
cost and unconstrained control. The resulting problem is called the
\emph{linear--quadratic regulator}, or LQR\@. Here \emph{regulation} means
driving the state toward a desired equilibrium and keeping it there.

LQR is special for three related reasons. First, the state and costate
equations are linear once the optimal control is substituted. Second, the
cost-to-go is quadratic in the state. Third, the optimal control is a linear
state-feedback law. All future consequences of the current state are
summarized by a symmetric matrix $P(t)$ that satisfies a Riccati differential
equation. Instead of solving a new nonlinear optimization problem for every
initial condition, one integrates this matrix equation once and obtains a
feedback controller for every state.

The linear--quadratic model is more than a classroom example. Here a
\emph{nominal} configuration or trajectory means the planned motion about
which local deviations will be measured. Near such a robot configuration or
trajectory, nonlinear dynamics can be approximated by
linear dynamics, and a smooth objective can be approximated by a quadratic
cost. LQR then supplies local feedback gains that correct perturbations around
the nominal motion. The same structure also appears inside trajectory
optimization, model predictive control, state estimation, and modern policy
optimization methods. In particular, Chapter~9 studies iterative LQR (iLQR),
a nonlinear trajectory-optimization method that repeatedly constructs and
solves local linear--quadratic approximations. Standard references
include~\cite{kirk2004optimal,lewis2012optimal,bertsekas2017dynamic}.

\section{Finite-Horizon LQR}

\subsection{Problem Formulation}

Consider the linear time-varying system
\begin{equation}
  \dot x(t)=A(t)x(t)+B(t)u(t),
  \qquad x(t_0)=x_0,
  \label{eq:finite-horizon-lqr-dynamics}
\end{equation}
where $x(t)\in\R^n$, $u(t)\in\R^m$,
$A(t)\in\R^{n\times n}$, and $B(t)\in\R^{n\times m}$. The system is
linear because the right-hand side is linear in both the state and the
control. If $A$ and $B$ are constant, the system is linear time-invariant.
Allowing them to depend on time also covers local models constructed along a
changing reference trajectory. For example, linearizing nonlinear dynamics
$\dot z=f(z,v)$ around a feasible nominal trajectory
$(\bar z(t),\bar v(t))$ gives, to first order,
\begin{equation}
  \delta\dot z
  \approx
  A(t)\delta z+B(t)\delta v,
  \qquad
  A(t)=f_z\bigl(\bar z(t),\bar v(t)\bigr),
  \quad
  B(t)=f_v\bigl(\bar z(t),\bar v(t)\bigr).
\end{equation}
The Jacobian matrices change as the nominal state and input move through
different operating conditions. In that setting, the variables $x$ and $u$
in \cref{eq:finite-horizon-lqr-dynamics} can be interpreted as the deviations
$\delta z$ and $\delta v$, and the resulting time-varying LQR gain corrects
the system back toward the nominal trajectory. A more detailed derivation is
given later in this chapter.

The finite-horizon LQR objective is
\begin{equation}
  J[u]
  =
  \frac12 x(t_f)\trans S_f x(t_f)
  +\frac12\int_{t_0}^{t_f}
  \left(
    x(t)\trans Q(t)x(t)
    +u(t)\trans R(t)u(t)
  \right)dt,
  \label{eq:finite-horizon-lqr-cost}
\end{equation}
with
\begin{equation}
  Q(t)\succeq0,
  \qquad
  R(t)\succ0,
  \qquad
  S_f\succeq0.
  \label{eq:lqr-weight-assumptions}
\end{equation}
The state penalty $x\trans Qx$ measures deviation from the desired equilibrium
$x=0$, the control penalty $u\trans Ru$ measures effort, and the terminal
penalty $x(t_f)\trans S_f x(t_f)$ measures the final error. The matrices are
taken to be symmetric. Positive semidefiniteness of $Q$ and $S_f$ makes state
error nonnegative, while positive definiteness of $R$ makes every nonzero
control input costly.

The entries of these matrices express design priorities. A large weight on a
position coordinate demands accurate positioning. A large control weight
makes the controller reluctant to use the corresponding actuator. These
comparisons must respect physical units: a numerical weight on meters cannot
be interpreted in the same way as the same numerical weight on radians or
newton-meters. State and input scaling should therefore be chosen deliberately.

The desired state is written as zero only to simplify notation. Tracking a
nonzero equilibrium or trajectory will be reduced to this regulation problem
later in the chapter. For now, the complete problem is
\begin{equation}
  \begin{aligned}
    \minimize_{x(\cdot),u(\cdot)}\quad
      &\frac12 x(t_f)\trans S_f x(t_f)
       +\frac12\int_{t_0}^{t_f}
       \left(x\trans Qx+u\trans Ru\right)dt\\
    \subjectto\quad
      &\dot x=Ax+Bu,
      \qquad x(t_0)=x_0.
  \end{aligned}
  \label{eq:finite-horizon-lqr-problem}
\end{equation}
There are no state or control bounds. This absence is essential to the exact
linear feedback formula. If an actuator saturates, the unconstrained LQR law
can still be useful locally, but it is no longer the exact solution of the
constrained problem.

We begin with the PMP conditions from Chapter~6. The Hamiltonian is
\begin{equation}
  H(t,x,u,\lambda)
  =
  \frac12 x\trans Qx
  +\frac12 u\trans Ru
  +\lambda\trans(Ax+Bu).
  \label{eq:finite-horizon-lqr-hamiltonian}
\end{equation}
The state and costate equations are
\begin{align}
  \dot x
  &=Ax+Bu,
  \label{eq:lqr-pmp-state}\\
  \dot\lambda
  &=-Qx-A\trans\lambda,
  \label{eq:lqr-pmp-costate}
\end{align}
and the free terminal state gives the terminal condition
\begin{equation}
  \lambda(t_f)=S_f x(t_f).
  \label{eq:lqr-costate-terminal-condition}
\end{equation}
Because $R\succ0$, the Hamiltonian is strictly convex in $u$. Its unique
minimizer satisfies
\begin{equation}
  0=H_u=Ru+B\trans\lambda,
\end{equation}
so
\begin{equation}
  u^\star(t)=-R(t)^{-1}B(t)\trans\lambda(t).
  \label{eq:lqr-control-from-costate}
\end{equation}
As in many examples from Chapter~6, the minimizing control is initially
expressed as a function of the costate. The central step in LQR is to express
that costate directly in terms of the current state.

\subsection{Quadratic Value Functions and the Riccati Equation}

The terminal condition in \cref{eq:lqr-costate-terminal-condition} is linear
in $x(t_f)$. The state--costate equations are also linear after substituting
\cref{eq:lqr-control-from-costate}. Moreover, the value function is meant to
describe the optimal cost from every possible state, not only along one
trajectory. To see why a linear state--costate relation is natural, substitute
the minimizing control into the state equation and collect the state and
costate equations as
\begin{equation}
  \frac{d}{dt}
  \begin{bmatrix}x\\\lambda\end{bmatrix}
  =
  \begin{bmatrix}
    A&-BR^{-1}B\trans\\
    -Q&-A\trans
  \end{bmatrix}
  \begin{bmatrix}x\\\lambda\end{bmatrix}.
  \label{eq:lqr-linear-hamiltonian-system}
\end{equation}
This differential equation and the terminal relation
$\lambda(t_f)=S_fx(t_f)$ are both linear. Therefore, if two state--costate
solutions satisfy the terminal relation, every linear combination of those
solutions does as well. When the optimal continuation from each state is
unique, the costate associated with
$\alpha x_a+\beta x_b$ must consequently be
$\alpha\lambda_a+\beta\lambda_b$. Thus, at each fixed time, the mapping from
the current state to its optimal costate is linear. Every linear mapping from
$\R^n$ to $\R^n$ can be represented by a matrix, so we seek
\begin{equation}
  \lambda(t)=P(t)x(t),
  \label{eq:lqr-costate-matrix-ansatz}
\end{equation}
where $P(t)\in\R^{n\times n}$ is to be determined. Differentiating this
relation gives
\begin{equation}
  \dot\lambda=\dot P x+P\dot x.
  \label{eq:lqr-costate-ansatz-derivative}
\end{equation}
The minimizing control becomes
\begin{equation}
  u^\star=-R^{-1}B\trans P x.
  \label{eq:finite-horizon-lqr-feedback-preliminary}
\end{equation}
Substituting it into the state equation yields
\begin{equation}
  \dot x
  =
  \left(A-BR^{-1}B\trans P\right)x.
  \label{eq:finite-horizon-lqr-closed-loop-preliminary}
\end{equation}
Using this equation on the right-hand side of
\cref{eq:lqr-costate-ansatz-derivative}, and equating the result with the
costate equation \cref{eq:lqr-pmp-costate}, gives
\begin{equation}
  \dot P x
  +P\left(A-BR^{-1}B\trans P\right)x
  =-Qx-A\trans Px.
\end{equation}
Because this relation is being constructed for the entire family of initial
states, it must hold for every $x$. Its matrix coefficient must therefore
vanish. At the terminal time, combining
$\lambda(t_f)=P(t_f)x(t_f)$ with
$\lambda(t_f)=S_fx(t_f)$ for every possible terminal state gives
$P(t_f)=S_f$. Hence $P$ satisfies
\begin{equation}
  \boxed{
  -\dot P
  =
  Q+A\trans P+PA-PBR^{-1}B\trans P,
  \qquad
  P(t_f)=S_f.}
  \label{eq:finite-horizon-differential-riccati-equation}
\end{equation}
This is the \emph{differential Riccati equation}. It is nonlinear because it
contains the quadratic matrix term $PBR^{-1}B\trans P$. It is integrated
backward from the known terminal value $P(t_f)=S_f$. Once $P(t)$ is known,
the optimal feedback gain is
\begin{equation}
  K(t)=R(t)^{-1}B(t)\trans P(t),
  \label{eq:finite-horizon-lqr-gain}
\end{equation}
and the control law is
\begin{equation}
  \boxed{u^\star(t)=-K(t)x(t).}
  \label{eq:finite-horizon-lqr-control-law}
\end{equation}

The Riccati equation was motivated using necessary PMP conditions. For LQR,
we can go further and prove global optimality by completing the square. This
argument also identifies the value function without using the HJB equation,
which is developed in Chapter~8.

\begin{theorem}[Finite-horizon LQR]
Suppose $P(t)=P(t)\trans$ satisfies
\cref{eq:finite-horizon-differential-riccati-equation}. Then, for any state
$x$ at any time $t\in[t_0,t_f]$, the optimal cost-to-go is
\begin{equation}
  V(t,x)=\frac12 x\trans P(t)x,
  \label{eq:finite-horizon-lqr-value-function}
\end{equation}
and the unique minimizing control is
$u^\star=-R^{-1}B\trans Px$.
\end{theorem}

\begin{proof}
Define
\begin{equation}
  W(t)=\frac12 x(t)\trans P(t)x(t).
\end{equation}
Because $P$ is symmetric,
\begin{align}
  \dot W
  &={}
  \frac12 x\trans\dot P x+x\trans P\dot x\nonumber\\
  &={}
  \frac12 x\trans\left(\dot P+PA+A\trans P\right)x
  +x\trans PBu.
  \label{eq:lqr-value-derivative-before-riccati}
\end{align}
The Riccati equation implies
\begin{equation}
  \dot P+PA+A\trans P
  =PBR^{-1}B\trans P-Q.
\end{equation}
Substitution into \cref{eq:lqr-value-derivative-before-riccati} and completion
of the square give
\begin{align}
  \frac12\left(x\trans Qx+u\trans Ru\right)
  &={}
  -\dot W
  +\frac12
  \left(u+R^{-1}B\trans Px\right)\trans
  R
  \left(u+R^{-1}B\trans Px\right).
  \label{eq:lqr-completion-of-squares-identity}
\end{align}
Integrating from $t$ to $t_f$ and using $P(t_f)=S_f$ yields
\begin{align}
  J_{t,x}[u]
  &={}
  \frac12 x\trans P(t)x\nonumber\\
  &\quad+
  \frac12\int_t^{t_f}
  \left(u(s)+R(s)^{-1}B(s)\trans P(s)x(s)\right)\trans
  R(s)
  \left(u(s)+R(s)^{-1}B(s)\trans P(s)x(s)\right)ds.
  \label{eq:lqr-cost-decomposition}
\end{align}
Here $J_{t,x}[u]$ denotes the cost accumulated when the system starts from
state $x$ at time $t$ and applies the control $u$ thereafter.
The integral is nonnegative because $R\succ0$, and it is zero exactly when
$u=-R^{-1}B\trans Px$. Hence this control is the unique global minimizer and
the remaining term is the optimal cost-to-go.
\end{proof}

Because the optimal cost is nonnegative for every state,
$x\trans P(t)x\geq0$ for every $x$. Thus the Riccati solution associated with
this value function is positive semidefinite.

This result clarifies all three central LQR objects. The matrix $P(t)$ is the
curvature of the quadratic cost-to-go. Along an optimal trajectory, the
costate is its state gradient,
\begin{equation}
  \nabla_x V(t,x)=P(t)x=\lambda(t),
\end{equation}
and the feedback law applies control in response to that gradient. A state
direction with large curvature in $P$ is expensive because even a small
displacement in that direction produces a large increase in future cost.

The symmetry of $P$ is consistent with its role in a quadratic form. More
formally, transpose the Riccati equation. When $Q$, $R$, and $S_f$ are
symmetric, $P\trans$ satisfies the same differential equation and terminal
condition as $P$. Uniqueness of the matrix differential equation therefore
gives $P\trans=P$. In numerical work one should nevertheless check that
roundoff has not produced a significant asymmetric component.

\begin{example}[A scalar finite-horizon regulator]
Consider
\begin{equation}
  \dot x=u,
  \qquad
  J=\frac12s x(t_f)^2
  +\frac12\int_{t_0}^{t_f}\left(qx^2+ru^2\right)dt,
  \qquad q>0,\quad r>0,\quad s\geq0.
  \label{eq:scalar-finite-horizon-lqr-problem}
\end{equation}
Here $A=0$, $B=1$, and the Riccati equation becomes
\begin{equation}
  -\dot p=q-\frac{p^2}{r},
  \qquad p(t_f)=s.
  \label{eq:scalar-finite-horizon-riccati}
\end{equation}
To solve this terminal-value equation, introduce the time remaining
$\tau=t_f-t$ and define
\begin{equation}
  \alpha=\sqrt{\frac qr},
  \qquad
  \gamma=\sqrt{qr},
  \qquad
  \tau=t_f-t.
\end{equation}
Because $dp/d\tau=-dp/dt$, the Riccati equation becomes
\begin{equation}
  \frac{dp}{d\tau}
  =q-\frac{p^2}{r}
  =\frac{\gamma^2-p^2}{r},
  \qquad p(0)=s.
\end{equation}
Separating variables gives
\begin{equation}
  \frac{d}{d\tau}
  \log\left|\frac{\gamma+p}{\gamma-p}\right|
  =2\alpha.
\end{equation}
Applying $p(0)=s$ and solving for $p$ yields
\begin{equation}
  p(t)
  =
  \gamma
  \frac{s+\gamma\tanh(\alpha\tau)}
       {\gamma+s\tanh(\alpha\tau)}.
  \label{eq:scalar-finite-horizon-riccati-solution}
\end{equation}
Here
\begin{equation}
  \tanh z=\frac{e^z-e^{-z}}{e^z+e^{-z}}
\end{equation}
is the hyperbolic tangent. The formula also covers $s=\gamma$ by continuity.
The optimal control is
\begin{equation}
  u^\star(t)=-\frac{p(t)}{r}x(t).
\end{equation}
If $s=0$, then
\begin{equation}
  p(t)=\sqrt{qr}\,
  \tanh\left(\sqrt{\frac qr}(t_f-t)\right).
\end{equation}
The feedback gain approaches zero at the final time because there is no
terminal penalty and no future interval left over which state error is costly.
Far from the terminal time, the gain approaches $\sqrt{q/r}$, the
infinite-horizon gain derived in Chapter~6. A nonzero terminal weight prevents
the controller from becoming indifferent to final error.
\end{example}

\subsection{Optimal Linear Feedback}

The optimal law \cref{eq:finite-horizon-lqr-control-law} is a feedback law,
not merely an open-loop control sequence. At every time it observes the
current state and recomputes the control from that state. If a disturbance
moves the system away from its predicted trajectory, the control changes
immediately according to the same gain matrix.

The original dynamics $\dot x=Ax+Bu$ are called \emph{open loop} when the
input $u$ is regarded as an externally specified signal. Under LQR, the input
is instead determined from the measured state by $u=-K(t)x$. Substitution
eliminates $u$ from the state equation:
\begin{align}
  \dot x
  &=Ax+B\bigl(-K(t)x\bigr)\nonumber\\
  &=\left(A-BK(t)\right)x.
  \label{eq:finite-horizon-lqr-closed-loop-system}
\end{align}
The resulting equation is called the \emph{closed-loop system} because the
state affects the control and the control feeds back into the state dynamics.
Once $x(t_0)=x_0$ is given, this differential equation determines the state
trajectory and the feedback law then determines the corresponding control
trajectory. The term $-BK(t)$ is the controller's modification of the
uncontrolled dynamics $A$. Thus the matrix $A-BK(t)$ describes the dynamics
created jointly by the physical system, often called the \emph{plant}, and
the controller. If a disturbance changes the state, the law $u=-K(t)x$
automatically changes the input, thereby closing the correction loop.

For a finite horizon, $K(t)$ generally varies with time even when $A$, $B$,
$Q$, and $R$ are constant. The remaining time and terminal penalty affect how
aggressively the controller should act. Consequently, the finite-horizon
closed-loop matrix $A-BK(t)$ is generally time varying as well.

This finite-horizon result does not by itself imply asymptotic stability.
Asymptotic stability asks whether $x(t)\to0$ as $t\to\infty$, whereas the
controller above is designed only on the finite interval $[t_0,t_f]$. It
minimizes the accumulated and terminal costs on that interval, but it need
not drive the state exactly to zero at $t_f$, and the control law has not yet
been specified beyond $t_f$. The infinite-horizon LQR problem studied later
selects a stabilizing Riccati solution; under the stated stabilizability and
detectability conditions, its constant closed-loop matrix $A-BK$ is
asymptotically stable.

\begin{example}[Position and velocity feedback for a double integrator]
Let
\begin{equation}
  x=
  \begin{bmatrix}q\\v\end{bmatrix},
  \qquad
  \dot x=
  \begin{bmatrix}0&1\\0&0\end{bmatrix}x
  +
  \begin{bmatrix}0\\1\end{bmatrix}u,
  \label{eq:lqr-double-integrator-dynamics}
\end{equation}
and choose
\begin{equation}
  Q=
  \begin{bmatrix}q_p&0\\0&q_v\end{bmatrix},
  \qquad R=r>0.
\end{equation}
Write
\begin{equation}
  P=
  \begin{bmatrix}
    p_{11}&p_{12}\\
    p_{12}&p_{22}
  \end{bmatrix}.
\end{equation}
The Riccati equation is equivalent to
\begin{align}
  -\dot p_{11}
  &=q_p-\frac{p_{12}^2}{r},\\
  -\dot p_{12}
  &=p_{11}-\frac{p_{12}p_{22}}{r},\\
  -\dot p_{22}
  &=q_v+2p_{12}-\frac{p_{22}^2}{r}.
  \label{eq:double-integrator-component-riccati}
\end{align}
The control law is
\begin{equation}
  u^\star
  =-\frac{p_{12}(t)}{r}q
   -\frac{p_{22}(t)}{r}v.
  \label{eq:double-integrator-lqr-control}
\end{equation}
Thus LQR automatically produces both position and velocity feedback. The
position term drives the system toward the target, while the velocity term
provides braking and damping. These roles emerge from the optimization; they
do not have to be imposed as separate controller-design rules.
\end{example}

The same finite-horizon construction can also stabilize a nonlinear robot
locally. The idea is to linearize the dynamics along a feasible nominal
trajectory and apply LQR to the resulting perturbation dynamics. This
extension is developed after the exact linear reference-tracking construction
in \cref{sec:lqr-reference-tracking}.

The weight matrices influence both performance and robustness, meaning the
ability to tolerate small disturbances and modeling errors. Increasing $Q$
relative to $R$ generally produces faster correction and larger control
effort. Increasing $R$ produces gentler control and slower convergence. A
large terminal weight makes final accuracy important, especially near $t_f$.
These are tendencies rather than independent tuning rules: coupled dynamics
mean that changing one entry can affect several state and input coordinates.

The discrete-time counterpart is developed from the Bellman recursion in
\cref{sec:discrete-time-bellman-recursion,sec:hjb-linear-quadratic-examples}.
There the same quadratic-value structure produces the discrete Riccati
recursion, its optimal feedback gain, and the infinite-horizon DARE.

\section{Infinite-Horizon and Tracking Problems}

\subsection{The Algebraic Riccati Equation}

For a time-invariant system
\begin{equation}
  \dot x=Ax+Bu,
  \label{eq:infinite-horizon-lqr-dynamics}
\end{equation}
consider the infinite-horizon objective
\begin{equation}
  J[u]
  =\frac12\int_0^\infty
  \left(x(t)\trans Qx(t)+u(t)\trans Ru(t)\right)dt,
  \qquad Q\succeq0,\quad R\succ0.
  \label{eq:infinite-horizon-lqr-cost}
\end{equation}
There is no terminal time, and the problem data do not depend on time. It is
therefore natural to seek a stationary quadratic value function
\begin{equation}
  V_\infty(x)=\frac12x\trans Px
\end{equation}
with a constant matrix $P$. Setting $\dot P=0$ in the differential Riccati
equation gives
\begin{equation}
  \boxed{
  A\trans P+PA-PBR^{-1}B\trans P+Q=0.}
  \label{eq:continuous-time-algebraic-riccati-equation}
\end{equation}
This is the \emph{continuous-time algebraic Riccati equation}, abbreviated
CARE\@. A suitable solution gives the stationary gain
\begin{equation}
  K=R^{-1}B\trans P
  \label{eq:infinite-horizon-lqr-gain}
\end{equation}
and feedback law
\begin{equation}
  \boxed{u^\star=-Kx.}
  \label{eq:infinite-horizon-lqr-control}
\end{equation}

An algebraic equation may have several symmetric solutions, and not every
solution represents the infinite-horizon optimum. The desired solution must
be positive semidefinite and must make the closed-loop matrix
\begin{equation}
  A_{\mathrm{cl}}=A-BK
\end{equation}
stable, meaning that every unforced closed-loop trajectory approaches zero.
Such a solution is called the \emph{stabilizing solution} of the CARE\@.

Two structural properties explain when a stabilizing solution should exist.
To interpret them, recall that an eigenvalue $\mu$ of $A$ describes a natural
system motion whose magnitude behaves like $e^{\operatorname{Re}(\mu)t}$.
The motion decays when $\operatorname{Re}(\mu)<0$ and fails to decay when
$\operatorname{Re}(\mu)\geq0$.

The pair $(A,B)$ is called \emph{stabilizable} if there is some feedback
matrix $K$ for which every eigenvalue of $A-BK$ has negative real part.
Equivalently in physical terms, every motion that does not already decay must
be accessible through the input channels in $B$ so that feedback can
stabilize it. Stabilizability is weaker than complete controllability: a mode
may be unaffected by the actuators as long as that mode is already stable and
therefore decays without control.

For example, consider
\begin{equation}
  A=
  \begin{bmatrix}1&0\\0&-1\end{bmatrix},
  \qquad
  B=
  \begin{bmatrix}0\\1\end{bmatrix}.
\end{equation}
The first state grows as $e^t$ but is not affected by the input, so no
feedback can stabilize the system. The pair $(A,B)$ is not stabilizable. If
the cost penalizes this first state, its infinite-horizon cost is necessarily
infinite for every initial condition with a nonzero first component.

The second property concerns what the state cost can see. Every symmetric
positive semidefinite matrix can be factored as
\begin{equation}
  Q=C_q\trans C_q.
\end{equation}
For example, one may choose the symmetric matrix square root
$C_q=Q^{1/2}$, which satisfies $Q^{1/2}Q^{1/2}=Q$. Then
\begin{equation}
  x\trans Qx=\norm{C_qx}_2^2.
\end{equation}
Thus $C_qx$ collects the state combinations penalized by the objective. A
state motion is invisible to the state cost when $C_qx(t)=0$ for all time.
The pair $(C_q,A)$ is called \emph{detectable} if every such invisible motion
is already stable. In other words, every motion that fails to decay must
eventually produce a nonzero state penalty. Detectability does not require
every stable motion to be penalized; an unpenalized mode is harmless to the
stability conclusion if it naturally decays.

Detectability matters because an optimizer has no reason to suppress a motion
that contributes no state cost. For example, let
\begin{equation}
  A=
  \begin{bmatrix}1&0\\0&-1\end{bmatrix},
  \qquad
  B=I,
  \qquad
  Q=
  \begin{bmatrix}0&0\\0&1\end{bmatrix}.
\end{equation}
The system is stabilizable because both states can be controlled, but the
growing first state is invisible to $Q$. Starting from
$x(0)=[1\;0]\trans$, the choice $u=0$ has zero cost even though
$x_1(t)=e^t$ diverges. Hence minimizing the stated objective does not force a
stabilizing controller. If $Q\succ0$, by contrast, every nonzero state
direction is penalized and detectability is automatic.

Under $R\succ0$, stabilizability of $(A,B)$, and detectability of $(C_q,A)$,
the CARE has a unique symmetric positive semidefinite stabilizing solution
$P$. The gain $K=R^{-1}B\trans P$ then makes every eigenvalue of $A-BK$ have
negative real part, so the closed-loop state approaches zero. These conditions
allow stable uncontrollable or unpenalized motions, but exclude precisely the
nondecaying motions that the actuators cannot control or the cost cannot see.
This chapter uses the conditions as interpretation rather than developing
their full linear-systems theory.

\begin{example}[Scalar system with drift]
Consider
\begin{equation}
  \dot x=ax+bu,
  \qquad
  J=\frac12\int_0^\infty(qx^2+ru^2)dt,
  \qquad q>0,\quad r>0,\quad b\neq0.
\end{equation}
The CARE is
\begin{equation}
  2aP-\frac{b^2}{r}P^2+q=0.
\end{equation}
Its stabilizing solution is
\begin{equation}
  P
  =\frac{r}{b^2}
  \left(
    a+\sqrt{a^2+\frac{b^2q}{r}}
  \right).
  \label{eq:scalar-general-care-solution}
\end{equation}
Indeed, with $K=bP/r$, the closed-loop coefficient is
\begin{equation}
  a-bK
  =-\sqrt{a^2+\frac{b^2q}{r}}<0.
\end{equation}
The other algebraic root produces a positive closed-loop coefficient and is
therefore rejected. When $a=0$ and $b=1$, the result reduces to
$P=\sqrt{qr}$ and $u=-\sqrt{q/r}\,x$, exactly as in the scalar integrator
example of Chapter~6.
\end{example}

One numerical route to the stabilizing CARE solution is to solve the
finite-horizon Riccati equation backward with increasingly distant terminal
times. Away from the terminal boundary, $P(t)$ often approaches a constant
matrix. This illustrates why the infinite-horizon gain is stationary: when
the terminal time is sufficiently far away, shifting the clock does not
change the remaining optimization problem.

\subsection{Closed-Loop Stability Intuition}

The quadratic value function also explains why the LQR feedback stabilizes the
system. Let
\begin{equation}
  W(x)=\frac12x\trans Px,
\end{equation}
where $P$ is the positive semidefinite stabilizing CARE solution. Along the
closed-loop dynamics $\dot x=(A-BK)x$,
\begin{align}
  \dot W
  &={}
  \frac12x\trans
  \left((A-BK)\trans P+P(A-BK)\right)x\nonumber\\
  &={}
  -\frac12x\trans\left(Q+K\trans RK\right)x.
  \label{eq:lqr-lyapunov-decrease}
\end{align}
The CARE was used in the second equality. Thus $W$ never increases. To make
the stability conclusion explicit, suppose both $P\succ0$ and
$M=Q+K\trans RK\succ0$. Positive definiteness means that there are constants
$c_1,c_2,c_3>0$ such that
\begin{equation}
  c_1\norm{x}_2^2
  \leq W(x)
  \leq c_2\norm{x}_2^2,
  \qquad
  \dot W(x)\leq-c_3\norm{x}_2^2.
\end{equation}
Consequently,
\begin{equation}
  \dot W\leq-\frac{c_3}{c_2}W,
\end{equation}
and integration gives
\begin{equation}
  W(x(t))
  \leq W(x_0)
  \exp\left(-\frac{c_3}{c_2}t\right).
\end{equation}
Thus $W$ and $x$ approach zero at an exponential rate. This supplies the
missing link between a decreasing quadratic cost and convergence of the state.

A scalar function that is positive away from an equilibrium and decreases
along every nonzero trajectory is called a \emph{Lyapunov function}. It plays
a role similar to physical energy: if the stored quantity must decrease and
can vanish only at the target, the state cannot remain away from the target.
Under the conditions above, the optimal cost-to-go itself serves as a
Lyapunov function. This is a deep connection between optimization and
stability: the same matrix $P$ measures future cost, determines the feedback
gain, and certifies closed-loop behavior.

For the optimal closed-loop trajectory, integrating
\cref{eq:lqr-lyapunov-decrease} from $0$ to a finite time $T$ gives
\begin{equation}
  W(x(T))-W(x_0)
  =-\frac12\int_0^T
  \left(x\trans Qx+(u^\star)\trans Ru^\star\right)dt.
\end{equation}
If $x(T)\to0$, then $W(x(T))\to0$. Taking $T\to\infty$ gives
\begin{equation}
  \frac12x_0\trans Px_0
  =
  \frac12\int_0^\infty
  \left(x\trans Qx+(u^\star)\trans Ru^\star\right)dt.
  \label{eq:infinite-horizon-lqr-cost-identity}
\end{equation}
The value stored initially in $W$ is exactly the accumulated future cost of
state error and control effort.

Positive semidefinite $Q$ deserves care. Some state directions may have no
direct state penalty. They can still be regulated because they influence
penalized directions or require control effort. Detectability rules out an
unstable direction that is completely invisible to the cost. Without such a
condition, a controller could ignore an unstable motion because the objective
never assigns it a cost.

The stability conclusion is tied to the model and to unconstrained control.
Large modeling errors, actuator saturation, delays, or neglected dynamics can
invalidate the predicted closed-loop behavior. LQR is often robust to modest
errors, but the Riccati equation by itself does not certify an arbitrarily
uncertain physical system.

\subsection{Reference Tracking}
\label{sec:lqr-reference-tracking}

Regulation drives the state toward zero. Tracking drives it toward a nonzero
reference. The cleanest case is a constant equilibrium pair $(x_r,u_r)$ that
satisfies
\begin{equation}
  0=Ax_r+Bu_r.
  \label{eq:lqr-reference-equilibrium-condition}
\end{equation}
Define the state and input errors
\begin{equation}
  e=x-x_r,
  \qquad
  v=u-u_r.
  \label{eq:lqr-equilibrium-error-coordinates}
\end{equation}
Because the reference satisfies the dynamics,
\begin{align}
  \dot e
  &=\dot x
    =A(e+x_r)+B(v+u_r)\nonumber\\
  &=Ae+Bv+\left(Ax_r+Bu_r\right)
    =Ae+Bv.
\end{align}
The tracking objective is now written as a regulation objective in the error
coordinates. For example, the infinite-horizon cost becomes
\begin{equation}
  J_{\mathrm{track}}
  =\frac12\int_0^\infty
  \left(e\trans Qe+v\trans Rv\right)dt.
\end{equation}
This cost penalizes deviation from the desired state and deviation from the
input required to hold that state. Because the error dynamics
$\dot e=Ae+Bv$ have exactly the same form as the original regulation
dynamics, the same CARE and feedback gain give
\begin{equation}
  v^\star=-Ke.
\end{equation}
Returning to the original variables using $v=u-u_r$ and $e=x-x_r$ gives
\begin{equation}
  u-u_r=-K(x-x_r)
  \quad\Longrightarrow\quad
  \boxed{u=u_r-K(x-x_r).}
  \label{eq:lqr-equilibrium-tracking-law}
\end{equation}
When $x=x_r$, the feedback term is zero and the controller applies exactly
$u_r$; the equilibrium condition then gives $\dot x=Ax_r+Bu_r=0$. When the
state is displaced, the term $-K(x-x_r)$ supplies a corrective input. Indeed,
substituting the law into the error dynamics gives
\begin{equation}
  \dot e=(A-BK)e.
  \label{eq:lqr-equilibrium-tracking-error-closed-loop}
\end{equation}
Therefore, if $K$ is the stabilizing infinite-horizon LQR gain, then
$e(t)\to0$ and hence $x(t)\to x_r$.

The two terms in the controller have different jobs. The feedforward input
$u_r$ makes the desired state dynamically sustainable, while the feedback
term corrects deviations caused by initial error or disturbances. If $u_r$ is
omitted, the controller applies zero input at $x=x_r$; unless $Ax_r=0$, the
state immediately moves away from the target and a steady-state tracking
error can result. For example, a robot arm generally needs a nonzero torque
to hold a configuration against gravity.

Not every desired state is an equilibrium. If no input $u_r$ satisfies
\cref{eq:lqr-reference-equilibrium-condition}, then exact constant tracking
is incompatible with the model. One must change the reference, use a moving
reference that satisfies the dynamics, or accept the best attainable tracking
error. Optimization cannot make an infeasible steady state dynamically
feasible.

The same coordinate change handles a time-varying feasible reference. Suppose
$(x_r(t),u_r(t))$ satisfies
\begin{equation}
  \dot x_r=A(t)x_r+B(t)u_r.
  \label{eq:lqr-feasible-reference-trajectory}
\end{equation}
Then the errors $e=x-x_r$ and $v=u-u_r$ satisfy
\begin{align}
  \dot e
  &=A(t)(e+x_r)+B(t)(v+u_r)-\dot x_r\nonumber\\
  &=A(t)e+B(t)v.
\end{align}
With a quadratic tracking cost
\begin{equation}
  J
  =\frac12e(t_f)\trans S_f e(t_f)
  +\frac12\int_{t_0}^{t_f}
  \left(e\trans Qe+v\trans Rv\right)dt,
\end{equation}
the optimal law is
\begin{equation}
  \boxed{
  u(t)=u_r(t)-K(t)\bigl(x(t)-x_r(t)\bigr).}
  \label{eq:lqr-trajectory-tracking-law}
\end{equation}
This controller combines an open-loop nominal motion with closed-loop error
correction. For the linear model, the error dynamics used in its derivation
are exact. The same architecture can be extended locally to a nonlinear robot
by linearizing its dynamics along the reference trajectory.

\paragraph{Local tracking for a nonlinear robot.}
Consider the nonlinear dynamics
\begin{equation}
  \dot x=f(t,x,u)
\end{equation}
and suppose a feasible nominal trajectory $(\bar x(t),\bar u(t))$ satisfies
\begin{equation}
  \dot{\bar x}=f(t,\bar x,\bar u).
\end{equation}
Define the perturbations from this trajectory by
\begin{equation}
  \delta x=x-\bar x,
  \qquad
  \delta u=u-\bar u.
\end{equation}
Expanding $f$ to first order around the nominal state and input gives
\begin{equation}
  f(t,\bar x+\delta x,\bar u+\delta u)
  \approx
  f(t,\bar x,\bar u)
  +f_x(t,\bar x,\bar u)\delta x
  +f_u(t,\bar x,\bar u)\delta u.
\end{equation}
Here $f_x$ and $f_u$ are the Jacobian matrices of partial derivatives with
respect to the state and control. Subtracting the nominal dynamics and
neglecting terms that are quadratic or higher order in the perturbations gives
\begin{equation}
  \delta\dot x
  \approx A(t)\delta x+B(t)\delta u,
  \qquad
  A(t)=f_x(t,\bar x,\bar u),
  \quad
  B(t)=f_u(t,\bar x,\bar u).
  \label{eq:trajectory-linearization-for-lqr}
\end{equation}
This perturbation model has the same form as the linear tracking-error
dynamics above, but it is now only a first-order approximation. After choosing
a quadratic penalty on $\delta x$ and $\delta u$, finite-horizon LQR gives
\begin{equation}
  \delta u^\star=-K(t)\delta x,
\end{equation}
or, in the original coordinates,
\begin{equation}
  \boxed{
  u(t)=\bar u(t)-K(t)\bigl(x(t)-\bar x(t)\bigr).}
  \label{eq:trajectory-stabilizing-lqr-law}
\end{equation}
Thus $\bar u(t)$ generates the nominal motion, while the feedback term
corrects deviations from it. This is the nonlinear, locally valid counterpart
of \cref{eq:lqr-trajectory-tracking-law}.

In practice, the nominal pair $(\bar x,\bar u)$ must be found before this
feedback law can be constructed. For a simple motion it may be available
analytically, but for a nonlinear robot it is often computed by a trajectory
optimization method subject to the nonlinear dynamics, endpoint conditions,
actuator limits, and collision constraints. The optimization returns a
planned state trajectory $\bar x(t)$ together with a feedforward control
trajectory $\bar u(t)$. Chapter~9 develops such trajectory-optimization
methods, including iLQR\@.

One could execute the plan open loop by applying only
$u(t)=\bar u(t)$. This assumes that the robot begins exactly at the planned
state and that the model is exact. If a disturbance, estimation error, or
modeling error moves the robot away from $\bar x(t)$, the precomputed input
$\bar u(t)$ does not change and therefore does not actively return the system
to the plan. A practical trajectory-tracking controller instead solves the
Riccati equation backward for $K(t)$ and applies
\cref{eq:trajectory-stabilizing-lqr-law}. On the nominal trajectory the
feedback correction is zero, so the controller applies $\bar u(t)$. Away from
the trajectory, the term $-K(t)(x(t)-\bar x(t))$ changes the input in a
direction intended to reduce the deviation. This feedback execution is
generally more reliable than open-loop playback because it continually reacts
to the measured state.

The improvement is local rather than unlimited. The matrices $A(t)$ and
$B(t)$ describe only the first-order dynamics near the nominal trajectory.
After a large disturbance, neglected nonlinear terms may dominate, the LQR
input may exceed actuator limits, or the correction may violate state and
obstacle constraints. The feedback law may then recover slowly, converge to
an unintended motion, or fail entirely. Large deviations therefore require a
controller designed to operate over a larger region, or a new trajectory
obtained through replanning or model predictive control.

\subsection{PD, Computed-Torque, and PID Control from LQR}

Many familiar feedback laws for mechanical systems can be derived by applying
LQR after changing from the physical state and input to tracking-error
coordinates. The high-level construction has three design steps, followed by
an analysis of the resulting physical controller.

First, define an error state $e$ that is zero on the desired motion, and seek
an input coordinate $\nu$ for which the error dynamics have the linear form
\begin{equation}
  \dot e=A_e e+B_e\nu.
  \label{eq:lqr-mechanical-error-template}
\end{equation}
The new variable $\nu$ is commonly called a \emph{virtual input}. It is not an
additional physical actuator. It is a convenient input coordinate used during
controller design. When the transformation is simply
\begin{equation}
  u=u_{\mathrm{ff}}+\nu,
  \label{eq:lqr-physical-virtual-input-template}
\end{equation}
$u_{\mathrm{ff}}$ is the feedforward input required to generate the reference
motion, and $\nu$ may more specifically be called the \emph{input correction}.
For a nonlinear mechanical system, the map from $\nu$ to the physical input
$u$ can also contain inverse-dynamics terms that cancel the modeled
nonlinearities.

Second, choose a quadratic cost on $e$ and $\nu$ and solve the resulting LQR
problem. This gives a linear law in the transformed coordinates,
\begin{equation}
  \nu^\star=-K e.
  \label{eq:lqr-virtual-input-feedback-template}
\end{equation}
Third, substitute this law into the input transformation to recover a
controller expressed in the original state and actuator coordinates. For the
simple decomposition in
\cref{eq:lqr-physical-virtual-input-template}, the physical command is
\begin{equation}
  u=u_{\mathrm{ff}}-K e.
  \label{eq:lqr-physical-feedback-template}
\end{equation}
The feedforward term creates the nominal motion, while the feedback term
corrects departures from it.

The resulting design should then be analyzed in the coordinates in which it
was constructed. Under the ideal transformed model, the closed-loop error
dynamics are
\begin{equation}
  \dot e=(A_e-B_eK)e.
  \label{eq:lqr-mechanical-closed-loop-template}
\end{equation}
The eigenvalues of $A_e-B_eK$ determine stability, convergence rates, and
oscillatory modes in this linear error model. One must also check what was
assumed when deriving the transformation: whether the reference is feasible,
whether model cancellation is accurate, and whether the mapped physical input
respects actuator limits. Thus the LQR interpretation does more than recover
the form of a familiar controller. It connects its gains to an explicit
tradeoff between tracking error and corrective effort, while making the
limits of the derivation visible.

\paragraph{PD for a double integrator.}
The scalar double integrator illustrates this construction without the
additional algebra of nonlinear inverse dynamics. Consider
\begin{equation}
  \ddot q=u
\end{equation}
tracking a feasible reference $q_r(t)$. Define
\begin{equation}
  e_q=q-q_r,
  \qquad
  e_v=\dot q-\dot q_r,
  \qquad
  \nu=u-\ddot q_r.
\end{equation}
Differentiating the errors gives
$\dot e_q=e_v$ and $\dot e_v=u-\ddot q_r=\nu$. Hence the error dynamics are
\begin{equation}
  \frac{d}{dt}
  \begin{bmatrix}e_q\\e_v\end{bmatrix}
  =
  \begin{bmatrix}0&1\\0&0\end{bmatrix}
  \begin{bmatrix}e_q\\e_v\end{bmatrix}
  +
  \begin{bmatrix}0\\1\end{bmatrix}\nu.
  \label{eq:pd-double-integrator-error-dynamics}
\end{equation}
For the infinite-horizon objective
\begin{equation}
  J
  =\frac12\int_0^\infty
  \left(q_p e_q^2+q_v e_v^2+r \nu^2\right)dt,
  \qquad q_p>0,\quad q_v\geq0,\quad r>0,
  \label{eq:pd-lqr-cost}
\end{equation}
write the stabilizing Riccati solution as
\begin{equation}
  P=
  \begin{bmatrix}p_{11}&p_{12}\\p_{12}&p_{22}\end{bmatrix}.
\end{equation}
The three scalar equations in the CARE are
\begin{align}
  0&=q_p-\frac{p_{12}^2}{r},\\
  0&=p_{11}-\frac{p_{12}p_{22}}{r},\\
  0&=q_v+2p_{12}-\frac{p_{22}^2}{r}.
\end{align}
The stabilizing solution uses the positive roots
\begin{equation}
  p_{12}=\sqrt{rq_p},
  \qquad
  p_{22}=\sqrt{r\left(q_v+2\sqrt{rq_p}\right)}.
\end{equation}
Consequently,
\begin{align}
  \nu^\star
  &=-\frac{1}{r}
    \begin{bmatrix}0&1\end{bmatrix}P
    \begin{bmatrix}e_q\\e_v\end{bmatrix}\nonumber\\
  &=-k_p e_q-k_d e_v,
  \label{eq:lqr-derived-pd-law}
\end{align}
where
\begin{equation}
  \boxed{
  k_p=\sqrt{\frac{q_p}{r}},
  \qquad
  k_d=\sqrt{\frac{q_v}{r}+2\sqrt{\frac{q_p}{r}}}.}
  \label{eq:lqr-derived-pd-gains}
\end{equation}
Returning to the original input gives
\begin{equation}
  \boxed{
  u=\ddot q_r-k_p(q-q_r)-k_d(\dot q-\dot q_r).}
  \label{eq:lqr-pd-tracking-law}
\end{equation}
This is a proportional--derivative, or PD, controller with reference
acceleration feedforward. The proportional feedback drives the position
error toward zero, while the derivative feedback supplies damping. The
closed-loop error equation is
\begin{equation}
  \ddot e_q+k_d\dot e_q+k_p e_q=0,
  \label{eq:lqr-pd-closed-loop-error}
\end{equation}
which is asymptotically stable because $k_p,k_d>0$.
\Cref{eq:lqr-derived-pd-gains} does not, however, automatically impose
critical damping. Comparing \cref{eq:lqr-pd-closed-loop-error} with the
standard second-order equation
\begin{equation}
  \ddot e_q+2\zeta\omega_n\dot e_q+\omega_n^2e_q=0
\end{equation}
gives
\begin{equation}
  \omega_n=\sqrt{k_p}=\left(\frac{q_p}{r}\right)^{1/4},
  \qquad
  \zeta=\frac{k_d}{2\sqrt{k_p}}
  =\frac12\sqrt{2+\frac{q_v}{\sqrt{rq_p}}}.
  \label{eq:lqr-pd-damping-ratio}
\end{equation}
Therefore the LQR-controlled double integrator is
\begin{equation}
  \begin{aligned}
    \text{underdamped}\quad &\Longleftrightarrow\quad
      q_v<2\sqrt{rq_p},\\
    \text{critically damped}\quad &\Longleftrightarrow\quad
      q_v=2\sqrt{rq_p},\\
    \text{overdamped}\quad &\Longleftrightarrow\quad
      q_v>2\sqrt{rq_p}.
  \end{aligned}
  \label{eq:lqr-pd-damping-cases}
\end{equation}
In particular, choosing $q_v=0$ gives
$\zeta=1/\sqrt{2}$, which is underdamped rather than critically damped. LQR
selects the damping ratio by optimizing the specified cost; critical damping
occurs only for the particular weight relationship above.

For a finite-horizon LQR problem, the same derivation produces time-varying
gains $k_p(t)$ and $k_d(t)$ rather than the constants in
\cref{eq:lqr-derived-pd-gains}.

The relationship can also be read in reverse. Given desired gains $k_p>0$
and $k_d>0$, a diagonal state cost of the form
\cref{eq:pd-lqr-cost} reproduces them when
\begin{equation}
  q_p=r k_p^2,
  \qquad
  q_v=r\left(k_d^2-2k_p\right).
\end{equation}
This construction requires $k_d^2\geq2k_p$ so that $q_v\geq0$. Thus every
LQR design above gives a stabilizing PD law, but not every stabilizing PD pair
comes from this particular diagonal, positive-semidefinite state cost.

\begin{example}[Computed-torque control for an $n$-joint manipulator]
For a nonlinear mechanical system, the LQR-derived PD law can be applied
after canceling the modeled dynamics. Consider a fully actuated manipulator
with
generalized coordinates
$q\in\R^n$, joint torques $\tau\in\R^n$, and dynamics
\begin{equation}
  M(q)\ddot q+C(q,\dot q)\dot q+g(q)=\tau.
  \label{eq:robot-computed-torque-dynamics}
\end{equation}
Here $M(q)$ is the positive-definite inertia matrix,
$C(q,\dot q)\dot q$ collects Coriolis and centrifugal terms, and $g(q)$ is
the gravity torque. Introduce a commanded acceleration $a_c$ and apply the
inverse-dynamics input
\begin{equation}
  \tau
  =C(q,\dot q)\dot q+g(q)+M(q)a_c.
  \label{eq:computed-torque-input}
\end{equation}
If the model is exact, substitution into
\cref{eq:robot-computed-torque-dynamics} gives $\ddot q=a_c$: the nonlinear
modeled dynamics have been transformed into a double integrator from $a_c$
to $q$.

Let $q_r(t)$ be a twice-differentiable desired joint trajectory and define
\begin{equation}
  e=q-q_r,
  \qquad
  \dot e=\dot q-\dot q_r.
\end{equation}
Define the virtual acceleration correction
\begin{equation}
  \nu=a_c-\ddot q_r.
\end{equation}
Then $\ddot e=\nu$, and the error state
\begin{equation}
  x_e=
  \begin{bmatrix}e\\\dot e\end{bmatrix}\in\R^{2n}
\end{equation}
has the linear dynamics
\begin{equation}
  \dot x_e
  =
  \underbrace{
  \begin{bmatrix}
    0&I\\
    0&0
  \end{bmatrix}}_{A_e}x_e
  +
  \underbrace{
  \begin{bmatrix}
    0\\I
  \end{bmatrix}}_{B_e}\nu,
  \label{eq:generic-manipulator-error-double-integrator}
\end{equation}
where each zero and identity block is $n\times n$. Choose
\begin{equation}
  J
  =\frac12\int_0^\infty
  \left(x_e\trans Q_e x_e+\nu\trans R\nu\right)dt,
  \qquad Q_e\succ0,
  \quad R\succ0.
\end{equation}
The pair $(A_e,B_e)$ is controllable, so the stabilizing CARE solution gives
\begin{equation}
  K=R^{-1}B_e\trans P
   =\begin{bmatrix}K_p&K_d\end{bmatrix},
\end{equation}
so the optimal virtual input is
\begin{equation}
  \nu^\star=-K_p e-K_d\dot e.
\end{equation}
Thus the commanded acceleration is
\begin{equation}
  a_c
  =\ddot q_r-K_p(q-q_r)-K_d(\dot q-\dot q_r).
  \label{eq:computed-torque-pd-acceleration}
\end{equation}
Combining the reference acceleration, optimal correction, and inverse robot
dynamics gives the computed-torque law
\begin{equation}
  \boxed{
  \tau
  =M(q)\left[
    \ddot q_r-K_p(q-q_r)-K_d(\dot q-\dot q_r)
  \right]
  +C(q,\dot q)\dot q+g(q).}
  \label{eq:generic-manipulator-lqr-computed-torque-law}
\end{equation}
Substitution into
\cref{eq:robot-computed-torque-dynamics} yields
\begin{equation}
  \ddot e+K_d\dot e+K_p e=0.
\end{equation}
This is \emph{computed-torque control}: the inverse-dynamics terms account for
the inertia and compensate the modeled Coriolis, centrifugal, and gravity
effects, while the LQR-derived PD terms regulate the resulting
double-integrator error dynamics.
Under exact cancellation, the modeled nonlinear dynamics therefore disappear
from the error equation, while the LQR cost determines the possibly coupled
joint-space gains. In particular, $K_p$ and $K_d$ need not be diagonal. For a
constant posture $q_r$, the reference velocity and acceleration are zero. At
$q=q_r$ and $\dot q=0$, the feedback terms vanish and the controller applies
$\tau=g(q_r)$, the torque required to balance gravity.
\end{example}

The optimality statement here applies to the quadratic cost written in terms
of the error state $x_e$ and virtual acceleration $\nu$. It does not generally
mean that the resulting controller minimizes a quadratic cost on the physical
torque $\tau$. The inverse-dynamics map from $\nu$ to $\tau$ depends on the
robot state, so a torque-based objective would lead to a different
optimization problem.

\paragraph{PID through state augmentation.}
Recall the normalized scalar double-integrator dynamics written in
position--velocity coordinates:
\begin{equation}
  \dot q=v,
  \qquad
  \dot v=u.
  \label{eq:pid-double-integrator-dynamics}
\end{equation}
Here $v$ denotes velocity and $u$ denotes commanded acceleration.
Let a feasible reference satisfy
\begin{equation}
  \dot q_r=v_r,
  \qquad
  \dot v_r=u_r,
\end{equation}
and define
\begin{equation}
  e_q=q-q_r,
  \qquad
  e_v=v-v_r,
  \qquad
  \delta u=u-u_r.
\end{equation}
The tracking-error dynamics are therefore
\begin{equation}
  \dot e_q=e_v,
  \qquad
  \dot e_v=\delta u.
  \label{eq:pid-double-integrator-error-dynamics}
\end{equation}

PD control does not generally eliminate steady error caused by a constant
unmodeled force, an incorrect gravity model, or another persistent bias.
Integral action addresses this limitation. Define
\begin{equation}
  \eta(t)=\int_{t_0}^t e_q(s)\,ds,
  \qquad \dot\eta=e_q,
\end{equation}
and augment the double-integrator state:
\begin{equation}
  z=
  \begin{bmatrix}\eta\\e_q\\e_v\end{bmatrix},
  \qquad
  \dot z=
  \underbrace{
  \begin{bmatrix}
    0&1&0\\
    0&0&1\\
    0&0&0
  \end{bmatrix}}_{A_a}z
  +
  \underbrace{
  \begin{bmatrix}0\\0\\1\end{bmatrix}}_{B_a}\delta u.
  \label{eq:lqi-augmented-error-system}
\end{equation}
Apply LQR to $(A_a,B_a)$ with
\begin{equation}
  J
  =\frac12\int_{t_0}^{\infty}
  \left(z\trans Q_a z+r(\delta u)^2\right)dt,
  \qquad Q_a\succ0,\quad r>0.
\end{equation}
This cost penalizes accumulated position error as well as instantaneous
position error, velocity error, and input correction. The stabilizing CARE
solution gives
\begin{equation}
  \delta u=-K_a z=-k_i\eta-k_p e_q-k_d e_v.
  \label{eq:lqr-derived-pid-law}
\end{equation}
This construction is commonly called linear--quadratic integral control, or
LQI\@. In the original coordinates,
\begin{equation}
  u
  =u_r-k_i\int_{t_0}^t(q-q_r)\,ds
   -k_p(q-q_r)-k_d(v-v_r).
\end{equation}
Since $v_r=\dot q_r$ and $u_r=\ddot q_r$, this is a PID controller under the
error convention used here; PID stands for
\emph{proportional--integral--derivative}. If a constant disturbance $d$
changes the error dynamics to $\dot e_v=\delta u+d$, then a steady state with
$e_q=e_v=0$ requires
\begin{equation}
  -k_i\eta+d=0,
  \qquad
  \eta=\frac{d}{k_i}.
\end{equation}
Thus a stable PID loop can reject the bias: the integral state retains the
constant value needed to produce the canceling input even after the tracking
error has returned to zero.

The optimal-control interpretation does not remove the practical limitations
of PID\@. Actuator saturation can cause the integral state to grow while the
requested input cannot be delivered, a phenomenon called integral windup;
anti-windup logic is then needed. Direct numerical differentiation of a noisy
position signal can make the derivative term noisy, so velocity estimation or
a filtered derivative is normally used. Finally, computed-torque cancellation
is only as accurate as the robot model. LQR or LQI determines gains for the
model used in the design, while robustness to large uncertainty, hard input
limits, and unmodeled dynamics requires additional analysis.

\section{Noise, State Estimation, and LQG: An Outlook}

The LQR law $u=-Kx$ assumes that the controller knows the current state
exactly. This assumption is convenient but rarely literal. A robot may measure
joint angles directly while inferring velocities from imperfect sensor data;
a mobile robot may receive camera or range measurements rather than its full
position and orientation; and unmodeled forces may disturb the motion. The
controller must therefore do two jobs: estimate the state from the available
measurements and use that estimate to choose the control.

This section proceeds in two stages. We first ignore disturbances and sensor
noise and derive a deterministic observer whose estimation error converges to
zero. We then introduce random process and measurement noise, replace exact
error convergence by a statistical objective, and obtain the Kalman--Bucy
filter. Combining that filter with LQR leads to LQG control.

\subsection{Deterministic State Estimation and Observers}

Begin with the deterministic linear system and sensor model
\begin{align}
  \dot x
  &=Ax+Bu,
  \label{eq:deterministic-observer-state-model}\\
  y
  &=Cx,
  \label{eq:deterministic-observer-measurement-model}
\end{align}
where $y(t)\in\R^p$ is the sensor measurement and
$C\in\R^{p\times n}$ maps the state to the quantities reported by the
sensors. The equation $y=Cx$ is called the \emph{measurement model}. For
example, a robot may measure joint positions while its velocity components
remain unmeasured. The controller therefore knows $y$, but it may not know the
complete state $x$ required by the LQR law $u=-Kx$.

The controller instead maintains a state estimate $\hat x$, where the hat
indicates an estimated quantity, and applies feedback using the estimate. An
\emph{observer} is a dynamical system that updates $\hat x$ from the known
input $u$ and measured output $y$. Define the estimation error by
\begin{equation}
  e_x=x-\hat x.
\end{equation}
We want this error to obey the stable linear dynamics
\begin{equation}
  \dot e_x=(A-LC)e_x,
  \label{eq:desired-observer-error-dynamics}
\end{equation}
where the observer gain $L\in\R^{n\times p}$ is chosen so that every
eigenvalue of $A-LC$ has negative real part. Because
$\dot e_x=\dot x-\dot{\hat x}$, enforcing
\cref{eq:desired-observer-error-dynamics} gives
\begin{align}
  \dot{\hat x}
  &=\dot x-(A-LC)(x-\hat x)\nonumber\\
  &=A\hat x+Bu+L(Cx-C\hat x)\nonumber\\
  &=A\hat x+Bu+L(y-C\hat x).
  \label{eq:continuous-time-state-observer}
\end{align}
Every quantity on the right-hand side is available to the observer: it uses
the current estimate $\hat x$, the applied input $u$, and the measured output
$y$. Thus the desired error dynamics lead directly to an implementable
dynamical system for the estimate.

The term $A\hat x+Bu$ predicts how the estimate should move according to the
model. The predicted sensor output is $C\hat x$, so
\begin{equation}
  y-C\hat x
\end{equation}
is the \emph{innovation}, or the discrepancy between the measured and
predicted outputs. The term $L(y-C\hat x)$ converts this discrepancy into a
correction of the state estimate. Substitution of the observer into
$\dot e_x=\dot x-\dot{\hat x}$ recovers
\cref{eq:desired-observer-error-dynamics}. Consequently, if $A-LC$ is stable,
then $e_x(t)\to0$ and $\hat x(t)\to x(t)$ even when the initial estimate is
incorrect.

A stabilizing observer gain exists when the measurements reveal every
unstable motion of the system, a property called detectability of $(C,A)$.
In the deterministic setting, the eigenvalues of $A-LC$ determine how rapidly
the estimation error decays. Noise changes this design problem because a gain
that responds strongly to measurements can also respond strongly to sensor
noise.

\subsection{Random Disturbances and Noisy Measurements}

Augment the deterministic model with process and measurement noise:
\begin{align}
  \dot x
  &=Ax+Bu+w,
  \label{eq:lqg-noisy-state-model}\\
  y
  &=Cx+v,
  \label{eq:lqg-noisy-measurement-model}
\end{align}
Here $w(t)\in\R^n$ represents unmodeled forces and errors in the state
dynamics, while $v(t)\in\R^p$ represents sensor error. Thus $w$ changes the
physical state, whereas $v$ changes the reported measurement.

For a random vector $z$, $\E[z]$ denotes its mean and
\begin{equation}
  \operatorname{Cov}(z)
  =\E\left[(z-\E[z])(z-\E[z])\trans\right].
  \label{eq:covariance-definition-lqg}
\end{equation}
is its covariance. Its diagonal entries are variances, and its off-diagonal
entries describe correlations. The standard LQG model assumes that $w$ and
$v$ are zero-mean, independent Gaussian white noises, independent also of the
initial state uncertainty. Their covariance intensities are
\begin{equation}
  \Sigma_w\succeq0,
  \qquad
  \Sigma_v\succ0,
\end{equation}
which measure the rates at which the two noises inject uncertainty. Ideal
continuous-time white noise is not an ordinary signal with a finite value at
each instant. A rigorous model represents its accumulated effect using
stochastic increments:
\begin{align}
  dx
  &=(Ax+Bu)\,dt+\Sigma_w^{1/2}\,dW_t,\\
  dy
  &=Cx\,dt+\Sigma_v^{1/2}\,dV_t,
\end{align}
where $W_t$ and $V_t$ are independent Wiener processes, also called Brownian
motions. To avoid introducing stochastic calculus here, we retain the common
engineering shorthand in
\cref{eq:lqg-noisy-state-model,eq:lqg-noisy-measurement-model}.

Together with the expected quadratic objective
\begin{equation}
  \E[J]
  =\E\left[
    \frac12x(t_f)\trans S_fx(t_f)
    +\frac12\int_{t_0}^{t_f}
    \left(x\trans Qx+u\trans Ru\right)dt
  \right].
  \label{eq:lqg-expected-quadratic-cost}
\end{equation}
this model defines \emph{linear--quadratic--Gaussian} control, or LQG\@. The
expectation averages over possible noise histories. A full treatment uses
stochastic calculus; see \cite{brockett1983stochastic} for an
application-oriented introduction.

The key LQG result is that state estimation and LQR control can be designed
separately: the LQR gain is computed as if the full state were known, and the
resulting law is applied to the Kalman estimate as $u=-K\hat x$. This
\emph{separation principle} is explained in
\cref{sec:lqg-separation-principle}. Persistent noise makes the usual
undiscounted infinite-horizon total cost generally infinite, so one instead
uses a finite-horizon, discounted, or long-run average formulation.

\subsection{Statistical State Estimation and the Kalman--Bucy Filter}

The estimation problem can be stated directly as a mean-square optimization
problem. At time $t$, the observer has access to the measurement history
\begin{equation}
  \mathcal{Y}_t
  =\{y(s):t_0\leq s\leq t\}.
\end{equation}
Among all estimates $a=a(\mathcal{Y}_t)$ that can be constructed from these
measurements, we seek the one that minimizes the expected squared error:
\begin{equation}
  \hat x(t)
  =\argmin_{a=a(\mathcal{Y}_t)}
  \E\left[\norm{x(t)-a}_2^2\mid\mathcal{Y}_t\right].
  \label{eq:conditional-estimation-mse}
\end{equation}
The conditioning means that the observed history is treated as known, while
the expectation averages over the state values that remain possible after
those measurements.

To solve \cref{eq:conditional-estimation-mse}, define the conditional mean
\begin{equation}
  m(t)=\E\left[x(t)\mid\mathcal{Y}_t\right].
  \label{eq:conditional-mean-state-estimate}
\end{equation}
For any candidate $a=a(\mathcal{Y}_t)$, write
$x-a=(x-m)+(m-a)$. The defining property of the conditional mean gives
$\E[x-m\mid\mathcal{Y}_t]=0$, so the cross term vanishes when the square is
expanded:
\begin{align}
  \E\left[\norm{x(t)-a}_2^2\mid\mathcal{Y}_t\right]
  &={}
  \E\left[\norm{x(t)-m(t)}_2^2\mid\mathcal{Y}_t\right]
  +\norm{m(t)-a}_2^2.
  \label{eq:conditional-mse-decomposition}
\end{align}
The first term does not depend on $a$, and the second is nonnegative and
vanishes only when $a=m(t)$. Hence the solution of the estimation problem is
\begin{equation}
  \boxed{
  \hat x(t)=\E\left[x(t)\mid\mathcal{Y}_t\right].}
  \label{eq:mmse-state-estimate}
\end{equation}
Equation~\eqref{eq:mmse-state-estimate} identifies the optimal estimate, but
it does not yet give a differential equation for computing it. Under the
linear--Gaussian assumptions, the conditional distribution of $x(t)$ remains
Gaussian and is completely characterized by its mean and covariance. We use
the same observer structure as in the deterministic case,
\begin{equation}
  \dot{\hat x}
  =A\hat x+Bu+L(y-C\hat x).
  \label{eq:kalman-bucy-filter-state-equation}
\end{equation}
In the deterministic design, $L$ was chosen to make $A-LC$ stable and to set
the rate at which the estimation error decays. With random process and
measurement noise, we instead choose the time-varying gain $L(t)$ to minimize
the expected squared estimation error. To obtain this gain, we first derive
how the estimation error and its covariance depend on $L$.

For $e_x=x-\hat x$, substitution of the noisy plant, measurement, and
observer equations gives
\begin{align}
  \dot e_x
  &=\dot x-\dot{\hat x}\nonumber\\
  &=(Ax+Bu+w)
    -\left[A\hat x+Bu+L(y-C\hat x)\right]\nonumber\\
  &=A(x-\hat x)+w
    -L\left[Cx+v-C\hat x\right]\nonumber\\
  &=(A-LC)e_x+w-Lv.
  \label{eq:observer-error-dynamics}
\end{align}
Because \cref{eq:mmse-state-estimate} minimizes the error for each measurement
history, it also minimizes the unconditional mean-square error, since
\begin{equation}
  \E[\norm{e_x(t)}_2^2]
  =\E\left[
    \E[\norm{e_x(t)}_2^2\mid\mathcal{Y}_t]
  \right].
\end{equation}
This is the law of total expectation: first average over the states compatible
with a fixed measurement history, and then average over all possible
measurement histories. Moreover,
\begin{equation}
  \E[e_x]
  =\E\left[\E[e_x\mid\mathcal{Y}_t]\right]
  =0,
\end{equation}
because the error from the conditional mean has zero conditional mean. The
error covariance is therefore
\begin{equation}
  \Sigma(t)
  =\E\left[e_x(t)e_x(t)\trans\right],
  \qquad
  \Sigma(t_0)=\Sigma_0.
\end{equation}
The mean-square error is its trace:
\begin{align}
  \E\left[\norm{e_x(t)}_2^2\right]
  &=\E[e_x\trans e_x]\nonumber\\
  &=\E\left[\operatorname{tr}(e_xe_x\trans)\right]\nonumber\\
  &=\operatorname{tr}\left(\E[e_xe_x\trans]\right)
   =\operatorname{tr}\Sigma.
  \label{eq:observer-mse-trace-covariance}
\end{align}
Thus minimizing mean-square error is equivalent to minimizing the trace of
the error covariance.

Propagating this covariance in continuous time requires the It\^o product
rule for stochastic differential equations. Developing that rule is beyond
the scope of this chapter; see \cite{brockett1983stochastic} for an
introduction to the required stochastic calculus. Applied to
\cref{eq:observer-error-dynamics}, it gives the standard covariance equation
\begin{equation}
  \dot\Sigma
  =(A-LC)\Sigma+\Sigma(A-LC)\trans
   +\Sigma_w+L\Sigma_vL\trans.
  \label{eq:observer-covariance-general-gain}
\end{equation}
Expanding the terms gives
\begin{align}
  \dot\Sigma
  &={}
  A\Sigma+\Sigma A\trans+\Sigma_w\nonumber\\
  &\quad{}
  +L\Sigma_vL\trans-LC\Sigma-\Sigma C\trans L\trans.
  \label{eq:observer-covariance-expanded-gain}
\end{align}

It remains to choose $L$. Only the last three terms in
\cref{eq:observer-covariance-expanded-gain} depend on this gain. Denote them
temporarily by
\begin{equation}
  \mathcal G(L)
  =
  L\Sigma_vL\trans-LC\Sigma-\Sigma C\trans L\trans.
\end{equation}
To expose their minimum, we seek a weighted square centered at an
as-yet unknown matrix $X$. Expanding gives
\begin{align}
  &(L-X)\Sigma_v(L-X)\trans\nonumber\\
  &\quad={}
  L\Sigma_vL\trans
  -L\Sigma_vX\trans
  -X\Sigma_vL\trans
  +X\Sigma_vX\trans.
  \label{eq:observer-gain-completion-square}
\end{align}
For the two cross terms to equal
$-LC\Sigma-\Sigma C\trans L\trans$, the center $X$ must satisfy
\begin{equation}
  \Sigma_vX\trans=C\Sigma,
  \qquad
  X\Sigma_v=\Sigma C\trans.
\end{equation}
Because $\Sigma_v\succ0$ is invertible, these equations determine
\begin{equation}
  X=\Sigma C\trans\Sigma_v^{-1}.
\end{equation}
For this value of $X$,
\begin{equation}
  X\Sigma_vX\trans
  =\Sigma C\trans\Sigma_v^{-1}C\Sigma.
\end{equation}
Substituting these identities into
\cref{eq:observer-gain-completion-square} and rearranging yields
\begin{align}
  \mathcal G(L)
  &={}
  \left(L-\Sigma C\trans\Sigma_v^{-1}\right)
  \Sigma_v
  \left(L-\Sigma C\trans\Sigma_v^{-1}\right)\trans
  -\Sigma C\trans\Sigma_v^{-1}C\Sigma.
  \label{eq:observer-gain-completed-square}
\end{align}
Because $\Sigma_v\succ0$, the first term is positive semidefinite. Indeed, for
every vector $z$,
\begin{equation}
  z\trans
  \left(L-\Sigma C\trans\Sigma_v^{-1}\right)
  \Sigma_v
  \left(L-\Sigma C\trans\Sigma_v^{-1}\right)\trans z
  =
  \norm{
    \Sigma_v^{1/2}
    \left(L-\Sigma C\trans\Sigma_v^{-1}\right)\trans z
  }_2^2
  \geq0.
\end{equation}
It vanishes for every $z$ exactly when the gain equals the center of the
square. Thus the minimizing value has not been guessed: it is forced by
matching the cross terms. For a given covariance $\Sigma$, it gives the
smallest instantaneous covariance growth in the positive-semidefinite
ordering and therefore also minimizes $\operatorname{tr}\dot\Sigma$. The
optimal gain is
\begin{equation}
  \boxed{L(t)=\Sigma(t)C\trans\Sigma_v^{-1}.}
  \label{eq:kalman-bucy-gain}
\end{equation}
With this gain, the square term in
\cref{eq:observer-gain-completed-square} vanishes. Substitution into
\cref{eq:observer-covariance-expanded-gain} gives
\begin{equation}
  \boxed{
  \dot\Sigma
  =A\Sigma+\Sigma A\trans+\Sigma_w
   -\Sigma C\trans\Sigma_v^{-1}C\Sigma.}
  \label{eq:kalman-bucy-riccati-equation}
\end{equation}
Starting from $\Sigma(t_0)=\Sigma_0$, this Riccati equation determines
$\Sigma(t)$, which determines $L(t)$, which in turn updates $\hat x(t)$ through
\cref{eq:kalman-bucy-filter-state-equation}. Under the linear--Gaussian
assumptions, this recursive estimate equals the conditional mean in
\cref{eq:mmse-state-estimate}.

Equations~\eqref{eq:kalman-bucy-filter-state-equation},
\eqref{eq:kalman-bucy-gain}, and
\eqref{eq:kalman-bucy-riccati-equation} form the \emph{Kalman--Bucy filter},
the continuous-time minimum-mean-square estimator.

The physical meaning of the filter is a repeated prediction--correction
process. In \cref{eq:kalman-bucy-filter-state-equation}, the term
$A\hat x+Bu$ predicts the state motion using the model and the known input.
The difference
\begin{equation}
  y-C\hat x
\end{equation}
is the \emph{innovation}: the part of the latest measurement not explained by
the prediction. The gain $L$ converts this output discrepancy into a
correction of the state estimate.

The covariance $\Sigma$ measures the uncertainty in the estimate, and its
diagonal entries are the error variances. In
\cref{eq:kalman-bucy-riccati-equation}, the terms
$A\Sigma+\Sigma A\trans$ propagate existing uncertainty through the dynamics,
$\Sigma_w$ adds uncertainty from process disturbances, and the negative term
$-\Sigma C\trans\Sigma_v^{-1}C\Sigma$ represents uncertainty removed by the
measurements. Thus greater process uncertainty makes the filter rely more
strongly on measurements, whereas greater sensor uncertainty reduces their
influence.

This balance is also visible in $L=\Sigma C\trans\Sigma_v^{-1}$. A large
state-estimation uncertainty $\Sigma$ produces a stronger measurement
correction, while a large measurement-noise covariance $\Sigma_v$ produces a
weaker correction. The linear--Gaussian structure allows all earlier
measurements to be summarized by the current pair $(\hat x,\Sigma)$; the
complete history $\mathcal{Y}_t$ need not be processed again whenever a new
measurement arrives. The known input $Bu$ cancels from the error dynamics
because it is included in both the physical system and the observer. Finally,
if $w=v=0$, \cref{eq:observer-error-dynamics} reduces to the deterministic
observer dynamics in \cref{eq:desired-observer-error-dynamics}.

\subsection{The Separation Principle in LQG}
\label{sec:lqg-separation-principle}

If the state were measured exactly, the LQR solution would be
$u=-Kx$. In the present setting, however, the controller does not know $x$;
it knows only the sensor measurements received so far. The Kalman--Bucy filter
summarizes those measurements by the conditional estimate $\hat x$. This
suggests the implementable control law
\begin{equation}
  \boxed{u(t)=-K(t)\hat x(t).}
  \label{eq:lqg-separated-control-law}
\end{equation}
It is tempting simply to replace the unknown state by its best estimate, but
the fact that $\hat x$ minimizes estimation error does not by itself prove
that this substitution minimizes the control cost. Estimation and control are
different optimization problems, so the combined law requires a separate
argument.

To state that argument precisely, a measurement-based controller may use the
history
\begin{equation}
  \mathcal{Y}_t=\{y(s):t_0\leq s\leq t\},
\end{equation}
but it may not use the unknown true state. Thus $u(t)$ must be a function of
$\mathcal{Y}_t$; such a control is said to be \emph{adapted} to the
measurements. The \emph{separation principle} states that, among all adapted
controls for the LQG problem, the law in
\cref{eq:lqg-separated-control-law} is optimal. The LQR gain can therefore be
designed as if the state were known, the Kalman filter can be designed from
the sensor and noise models, and the two can then be combined. The form
$u=-K\hat x$ is also called \emph{certainty equivalence}: the full-state law
is left unchanged except that the unknown state is replaced by its
conditional estimate.

\begin{theorem}[Finite-horizon LQG separation]
\label{thm:finite-horizon-lqg-separation}
Consider the linear--Gaussian model in
\cref{eq:lqg-noisy-state-model,eq:lqg-noisy-measurement-model} and the
expected quadratic cost in \cref{eq:lqg-expected-quadratic-cost}. Let $P(t)$
solve the LQR differential Riccati equation in
\cref{eq:finite-horizon-differential-riccati-equation}, and define
\begin{equation}
  K(t)=R^{-1}B\trans P(t).
\end{equation}
Let $\hat x(t)=\E[x(t)\mid\mathcal{Y}_t]$ be the estimate produced by the
Kalman--Bucy filter. Then the control in
\cref{eq:lqg-separated-control-law} uniquely minimizes the expected cost,
over all controls adapted to $\mathcal{Y}_t$ for which the expected cost is
finite.
\end{theorem}

\begin{proof}
The proof has two steps. First, we rewrite the expected cost as a term that is
independent of the control plus a nonnegative square. Second, we minimize that
square using only the information available from the measurements.

For the first step, define $Z(t)=x(t)\trans P(t)x(t)/2$ and apply the It\^o
product rule along
$dx=(Ax+Bu)dt+\Sigma_w^{1/2}dW_t$. This rule is developed, for example, in
Roger W. Brockett's lecture notes on stochastic
control~\cite{brockett1983stochastic}. Using the Riccati equation,
$K=R^{-1}B\trans P$, and $P(t_f)=S_f$, then integrating and taking
expectations gives
\begin{align}
  \E[J[u]]
  &={}
  \frac12\E\left[x(t_0)\trans P(t_0)x(t_0)\right]
  +\frac12\int_{t_0}^{t_f}
    \operatorname{tr}\!\left(P(t)\Sigma_w\right)dt
  \nonumber\\
  &\quad+
  \frac12\E\left[
    \int_{t_0}^{t_f}
    \left(u+Kx\right)\trans R\left(u+Kx\right)dt
  \right].
  \label{eq:lqg-expected-square-completion}
\end{align}
Time arguments are suppressed inside the integrals. The left-hand side is the
objective in \cref{eq:lqg-expected-quadratic-cost}. The stochastic integral
has zero expectation, while the It\^o correction produces the trace term.
Thus, the first two terms on the right are independent of the control, and the
last term measures departure from the full-state LQR law. When
$\Sigma_w=0$ and $x(t_0)$ is deterministic, the identity reduces to the
deterministic decomposition in \cref{eq:lqr-cost-decomposition} at $t=t_0$.

For the second step, we minimize the last term using only the available
measurements. Write
\begin{equation}
  x=\hat x+e_x,
  \qquad
  \E[e_x\mid\mathcal{Y}_t]=0.
\end{equation}
Under the linear--Gaussian assumptions, the conditional error covariance is
the deterministic matrix $\Sigma(t)$ generated by the Kalman Riccati
equation.

Because both $u(t)$ and $\hat x(t)$ are determined by $\mathcal{Y}_t$,
conditioning the integrand in
\cref{eq:lqg-expected-square-completion} gives
\begin{align}
  &\E\left[
    (u+Kx)\trans R(u+Kx)
    \mid\mathcal{Y}_t
  \right]
  \nonumber\\
  &\quad={}
  (u+K\hat x)\trans R(u+K\hat x)
  +\operatorname{tr}\!\left(K\trans R K\Sigma\right).
  \label{eq:lqg-conditional-square-decomposition}
\end{align}
The cross term vanishes because
$\E[e_x\mid\mathcal{Y}_t]=0$, and the trace term is the remaining conditional
quadratic error. Crucially, $\Sigma(t)$ is independent of the control because
the known term $Bu$ enters both the plant and the filter and cancels from the
estimation-error dynamics. This is the \emph{no-dual-effect} property of the
LQG model.

Taking expectations in
\cref{eq:lqg-conditional-square-decomposition} and substituting the result
into \cref{eq:lqg-expected-square-completion} therefore yields
\begin{equation}
  \E[J[u]]
  =C+
  \frac12\E\left[
    \int_{t_0}^{t_f}
    (u+K\hat x)\trans R(u+K\hat x)dt
  \right],
  \label{eq:lqg-separated-cost-decomposition}
\end{equation}
where $C$ collects the initial-state, process-noise, and estimation-error
terms, all of which are independent of the control policy. Because
$R\succ0$, the remaining integral is nonnegative and is uniquely minimized by
$u=-K\hat x$. This control is admissible because $\hat x$ is determined by
the measurement history $\mathcal{Y}_t$. Thus the independently computed LQR
gain and Kalman estimate produce the optimal measurement-based control.
\end{proof}

The proof gives a physical interpretation of certainty equivalence. If
the state were known, LQR would make the completed-square residual
$u+Kx$ equal to zero. When the state is hidden, that residual cannot be
cancelled for every possible value of $x$. The best admissible choice in the
conditional mean-square sense is to cancel its predictable part,
$u+K\hat x$.
The remaining term depends on the unavoidable estimation uncertainty, not on
the control policy.

The two designs therefore use different data: $K$ depends on
$(A,B,Q,R,S_f)$, whereas $L$ depends on
$(A,C,\Sigma_w,\Sigma_v,\Sigma_0)$. Neither Riccati equation contains the
other gain.

Optimality and stability are related but distinct questions. The theorem
above proves that $u=-K\hat x$ minimizes the expected quadratic cost. A
separate noise-free argument shows that controller and estimator stability
can also be checked independently. For the simplest explanation, consider the
time-invariant infinite-horizon case, in which the controller and estimator
Riccati equations have produced constant gains $K$ and $L$. Since
$\hat x=x-e_x$, the state and estimation error satisfy
\begin{align}
  \dot x
  &=(A-BK)x+BK e_x,\\
  \dot e_x
  &=(A-LC)e_x.
\end{align}
Therefore
\begin{equation}
  \begin{bmatrix}
    \dot x\\
    \dot e_x
  \end{bmatrix}
  =
  \begin{bmatrix}
    A-BK&BK\\
    0&A-LC
  \end{bmatrix}
  \begin{bmatrix}
    x\\e_x
  \end{bmatrix}.
  \label{eq:lqg-separated-closed-loop-matrix}
\end{equation}
The lower-left block is zero, so this matrix is \emph{block triangular}. A
block-triangular matrix has the eigenvalues of its diagonal blocks. Therefore
the combined system has the controller eigenvalues of $A-BK$ and the
estimator-error eigenvalues of $A-LC$. The choice of $L$ does not change the
controller eigenvalues, and the choice of $K$ does not change the
estimator-error eigenvalues. Thus stability can be checked independently:
the noise-free combination is stable when both $A-BK$ and $A-LC$ are stable.
This block-triangular calculation does not by itself prove optimality; that
claim follows from the expected-cost decomposition above. With persistent
noise, individual trajectories do not converge exactly to zero; instead, one
asks whether the state and estimation-error covariances remain bounded.

The separation principle is exact for the unconstrained linear--quadratic
Gaussian problem considered here. It does not say that estimation and control
are always independent. With non-Gaussian noise, the Kalman--Bucy filter is
generally no longer the exact conditional-mean estimator, although some form
of separation may still survive. Nonlinear dynamics, constraints,
control-dependent sensing, nonquadratic or risk-sensitive objectives, and
other information patterns can destroy the no-dual-effect property and couple
estimation with control. In such problems, an action may be valuable not only
because it moves the robot but also because it produces more informative
future measurements.

For reference tracking, the estimated-state controller becomes
\begin{equation}
  u=u_r-K(\hat x-x_r).
\end{equation}
The feedforward input supports the reference, the Kalman filter estimates the
current state, and the LQR gain corrects the estimated tracking error. This
estimator--controller architecture remains a standard baseline for robotic
systems because every component has a clear mathematical and physical role.

\paragraph{Exercises.}

\begin{exercise}
For the finite-horizon LQR problem, state the dimensions of $P$, $K$, and
$B\trans P$. Why must $Kx$ have the same dimension as $u$?
\end{exercise}
\begin{proof}[Short solution]
$P\in\R^{n\times n}$,
$B\trans P\in\R^{m\times n}$, and
$K=R^{-1}B\trans P\in\R^{m\times n}$. Therefore
$Kx\in\R^m$, matching the control dimension.
\end{proof}

\begin{exercise}
Starting from $\lambda=Px$, derive the differential Riccati equation and its
terminal condition.
\end{exercise}
\begin{proof}[Short solution]
Differentiate $\lambda=Px$, substitute
$u=-R^{-1}B\trans Px$, and equate the result with
$\dot\lambda=-Qx-A\trans\lambda$. Since the identity holds for every $x$,
\begin{equation}
  -\dot P=Q+A\trans P+PA-PBR^{-1}B\trans P,
  \qquad P(t_f)=S_f.
\end{equation}
\end{proof}

\begin{exercise}
Use \cref{eq:lqr-cost-decomposition} to explain why the LQR extremal is a
global minimum rather than merely a trajectory satisfying PMP\@.
\end{exercise}
\begin{proof}[Short solution]
The cost equals $x\trans P(t)x/2$ plus an integral of a quadratic form weighted
by $R\succ0$. The integral is nonnegative for every admissible control and
vanishes only for $u=-R^{-1}B\trans Px$. No other control can have smaller
cost.
\end{proof}

\begin{exercise}
For the scalar finite-horizon integrator with $q=0$ and terminal weight
$s>0$, solve the Riccati equation and give the optimal gain.
\end{exercise}
\begin{proof}[Short solution]
The equation is $-\dot p=-p^2/r$, with $p(t_f)=s$. Its solution is
\begin{equation}
  p(t)=\frac{rs}{r+s(t_f-t)},
  \qquad
  K(t)=\frac{p(t)}r
  =\frac{s}{r+s(t_f-t)}.
\end{equation}
\end{proof}

\begin{exercise}
For the double integrator in
\cref{eq:lqr-double-integrator-dynamics}, identify the position and velocity
feedback gains in \cref{eq:double-integrator-lqr-control}.
\end{exercise}
\begin{proof}[Short solution]
The position gain is $k_p(t)=p_{12}(t)/r$, and the velocity gain is
$k_v(t)=p_{22}(t)/r$. Thus $u=-k_p q-k_v v$.
\end{proof}

\begin{exercise}
Suppose $Q$, $R$, and $S_f$ are all multiplied by the same scalar $c>0$.
How do $P$, $K$, and the optimal control change?
\end{exercise}
\begin{proof}[Short solution]
The objective is multiplied by $c$, so its minimizer does not change. The
Riccati solution becomes $cP$, while
$(cR)^{-1}B\trans(cP)=R^{-1}B\trans P$. Hence $K$ and the optimal control are
unchanged, although the value is multiplied by $c$.
\end{proof}

\begin{exercise}
For $\dot x=ax+bu$ with $b\neq0$, verify that the solution in
\cref{eq:scalar-general-care-solution} makes the closed-loop coefficient
negative.
\end{exercise}
\begin{proof}[Short solution]
With $K=bP/r$,
\begin{equation}
  a-bK
  =a-\frac{b^2}{r}P
  =-\sqrt{a^2+\frac{b^2q}{r}}<0.
\end{equation}
Thus this root is stabilizing.
\end{proof}

\begin{exercise}
Show that the infinite-horizon LQR value decreases according to
\cref{eq:lqr-lyapunov-decrease}.
\end{exercise}
\begin{proof}[Short solution]
Differentiate $W=x\trans Px/2$ along
$\dot x=(A-BK)x$, use $K=R^{-1}B\trans P$, and substitute the CARE\@. The result
is
\begin{equation}
  \dot W=-\frac12x\trans(Q+K\trans RK)x.
\end{equation}
\end{proof}

\begin{exercise}
For a constant reference $x_r$, why is a feedforward input $u_r$ required?
What equation must $(x_r,u_r)$ satisfy?
\end{exercise}
\begin{proof}[Short solution]
The feedforward input must balance the plant dynamics at the desired state.
The pair must satisfy $Ax_r+Bu_r=0$. Without this balance, zero tracking error
is not a constant solution of the closed-loop dynamics.
\end{proof}

\begin{exercise}
Let a feasible reference trajectory satisfy
$\dot x_r=A(t)x_r+B(t)u_r$. Derive the error dynamics for
$e=x-x_r$ and $v=u-u_r$.
\end{exercise}
\begin{proof}[Short solution]
Subtracting the reference dynamics from the plant dynamics gives
\begin{equation}
  \dot e=A(t)e+B(t)v.
\end{equation}
Thus the tracking problem has the same form as regulation in the error
coordinates.
\end{proof}

\begin{exercise}
For the double-integrator tracking problem in
\cref{eq:pd-double-integrator-error-dynamics}, derive the gains in
\cref{eq:lqr-derived-pd-gains} from the CARE\@. Why are the positive square
roots selected?
\end{exercise}
\begin{proof}[Short solution]
The $(1,1)$ and $(2,2)$ CARE entries give
$p_{12}^2=rq_p$ and
$p_{22}^2=r(q_v+2p_{12})$. The stabilizing solution has
$p_{12}>0$ and $p_{22}>0$, so
\begin{equation}
  k_p=\frac{p_{12}}r=\sqrt{\frac{q_p}{r}},
  \qquad
  k_d=\frac{p_{22}}r
  =\sqrt{\frac{q_v}{r}+2\sqrt{\frac{q_p}{r}}}.
\end{equation}
These signs make the polynomial $s^2+k_ds+k_p$ stable; the other roots do
not give the stabilizing Riccati solution.
\end{proof}

\begin{exercise}
Suppose the double integrator is affected by an unknown constant acceleration
$d$, so that $\dot e_v=\delta u+d$. Explain how the LQI law in
\cref{eq:lqr-derived-pid-law} can maintain zero steady-state position error.
\end{exercise}
\begin{proof}[Short solution]
At a steady state with $e_q=e_v=0$, the input correction is
$\delta u=-k_i\eta$. The acceleration is zero when $-k_i\eta+d=0$, so the
integral state settles to
$\eta=d/k_i$ and supplies the constant input needed to cancel the
disturbance.
\end{proof}

\begin{exercise}
Derive the observer error equation for
$\dot{\hat x}=A\hat x+Bu+L(y-C\hat x)$.
\end{exercise}
\begin{proof}[Short solution]
Using $e_x=x-\hat x$, $\dot x=Ax+Bu+w$, and $y=Cx+v$ gives
\begin{equation}
  \dot e_x=(A-LC)e_x+w-Lv.
\end{equation}
\end{proof}

\begin{exercise}
Explain separately why the LQR gain $K$ and Kalman gain $L$ can be designed
independently for expected-cost optimality and for closed-loop stability.
\end{exercise}
\begin{proof}[Short solution]
For optimality, conditioning the completed-square identity on the measurement
history leaves the control-dependent term
$(u+K\hat x)\trans R(u+K\hat x)$, which is uniquely minimized by
$u=-K\hat x$. The remaining estimation-error term is independent of the
control because the Kalman covariance does not depend on $u$. For stability,
the noise-free matrix in \cref{eq:lqg-separated-closed-loop-matrix} is block
triangular, so its eigenvalues are those of $A-BK$ and $A-LC$. Thus the two
stability conditions can also be checked independently.
\end{proof}

\begin{exercise}
Why is the usual undiscounted infinite-horizon expected cost generally
infinite when persistent process noise is present?
\end{exercise}
\begin{proof}[Short solution]
Persistent noise continually excites the state, so the expected instantaneous
quadratic cost generally approaches a positive steady value rather than zero.
Integrating that positive value over an infinite interval gives an infinite
total cost.
\end{proof}

\paragraph{Chapter summary.}
Linear dynamics and quadratic costs produce an unusually complete optimal
control solution. For the finite-horizon problem, PMP gives a linear relation
between the costate and state, $\lambda=Px$. The matrix $P(t)$ satisfies the
differential Riccati equation backward from the terminal cost, and the optimal
law is the time-varying feedback $u=-R^{-1}B\trans Px$. Completing the square
proves that this law is the unique global minimizer and shows that the value
function is $V(t,x)=x\trans P(t)x/2$.

For time-invariant infinite-horizon regulation, the Riccati differential
equation becomes the algebraic Riccati equation. Its stabilizing solution gives
a stationary feedback law and a quadratic Lyapunov function for the
closed-loop system. Reference tracking is obtained by applying LQR to error
coordinates and retaining the feedforward input needed to generate the
reference. Around a nonlinear robot trajectory, the same construction gives
local feedback based on linearized dynamics. For double-integrator error
dynamics, the optimal feedback has the familiar PD form. Computed-torque
control applies this optimal PD law after canceling modeled mechanical
dynamics, while augmenting the state with integrated tracking error produces
an LQI design with PID form and rejection of constant biases.

When the state is not known exactly, an observer estimates it from noisy
measurements. The Kalman--Bucy filter evolves an uncertainty covariance through
a second Riccati equation. In the LQG setting, the separation principle allows
the estimator and controller to be designed independently and combined as
$u=-K\hat x$. These ideas provide the most important exactly solvable example
of optimal feedback control and prepare the quadratic example in the HJB
framework of Chapter~8.
\chapter{Dynamic Programming and the HJB Equation}
\label{chap:dynamic-programming-hjb}
\chapteroverview{The principle of optimality, Bellman recursion,
continuous-time Hamilton--Jacobi--Bellman equations, optimal feedback,
linear--quadratic control, numerical value iteration, and the relationship
between dynamic programming and Pontryagin's minimum principle.}

Chapter~6 approached optimal control by perturbing a candidate trajectory.
The resulting state, costate, and Hamiltonian conditions describe what must
hold along an optimal motion from a specified initial condition. Chapter~7
showed that the linear--quadratic problem has additional structure: a single
Riccati equation produces a feedback law for an entire region of states.

This chapter develops a more general route to feedback. Its central object is
the \emph{value function}, or \emph{optimal cost-to-go}: the smallest future
cost achievable from each time--state pair. The cost-to-go under a particular
control need not be optimal; the value function is the infimum of these costs
over all admissible controls.

Once the value function is known, the optimal action can be determined from
the current time and state. The dynamics predict how each candidate action
changes the state and, consequently, how the value changes. The controller
chooses the action that minimizes the immediate cost plus the optimal future
cost of the resulting state. If the immediate cost is independent of the
action, this amounts to choosing the action that produces the greatest
decrease in value. In general, the decrease in value must be balanced against
the cost of producing it. This decision can be made without solving the full
trajectory-optimization problem again.

The key question is how to compute the value function. The central insight is
the \emph{principle of optimality}. Imagine cutting an optimal trajectory at
some intermediate time. From the state reached at that time, the remaining
part of the trajectory must be an optimal solution of the remaining problem.
If a cheaper continuation existed, we could keep the first part unchanged and
replace only the remainder, producing a cheaper complete trajectory and
contradicting optimality.

This observation makes the value function recursive. The optimal cost from
the current state equals the cost accumulated over a short first interval
plus the optimal cost-to-go from the state reached at the end of that
interval. Thus, once later values are known, earlier values can be computed by
solving shorter local minimization problems. Computing the value function
through this backward recursion is the core idea of \emph{dynamic
programming}.

In discrete time, this principle produces the Bellman recursion. In
continuous time, applying it over an interval that shrinks to zero produces
the Hamilton--Jacobi--Bellman equation, abbreviated HJB\@. Solving HJB over a
state domain yields an optimal feedback law rather than a single open-loop
control signal. The price is computational: HJB is generally a nonlinear
partial differential equation in time and state, and its numerical cost grows
rapidly with the state dimension. This tradeoff motivates the local and
approximate methods developed in later chapters. Standard references include
\cite{bertsekas2017dynamic,kirk2004optimal,lewis2012optimal}.

\section{The Principle of Optimality}

\subsection{Cost-to-Go and Optimal Substructure}

Consider the finite-horizon system
\begin{equation}
  \dot x(s)=f\bigl(s,x(s),u(s)\bigr),
  \qquad x(t)=x,
  \label{eq:hjb-controlled-dynamics}
\end{equation}
with terminal time $t_f$, running cost $\ell$, terminal cost $\Phi$, and
admissible controls $u(s)\in\mathcal U$. Starting at time $t$ from state $x$,
the cost of a control signal is
\begin{equation}
  J_{t,x}[u]
  =
  \Phi\bigl(x(t_f),t_f\bigr)
  +\int_t^{t_f}
  \ell\bigl(s,x(s),u(s)\bigr)\,ds.
  \label{eq:hjb-cost-from-time-state}
\end{equation}
The \emph{value function} is the best cost achievable from that time and
state:
\begin{equation}
  V(t,x)
  =\inf_{u(\cdot)}J_{t,x}[u].
  \label{eq:hjb-value-function-definition}
\end{equation}
The terminal condition is
\begin{equation}
  V(t_f,x)=\Phi(x,t_f),
  \label{eq:hjb-value-terminal-condition}
\end{equation}
because no running interval remains at $t_f$.

The word \emph{value} does not mean the value of the state. It means the
optimal numerical value of the remaining objective. The value function
assigns a cost to each possible situation. A low value means that the task can
still be completed cheaply; a high value means that the current state has
expensive future consequences. Two states that are geometrically close may
have very different values. A vehicle moving toward an obstacle and the same
vehicle at the same position moving away from it have different states and
can have very different future costs.

Choose any intermediate time $\tau\in(t,t_f)$. Every complete control can be
viewed as two pieces: a first piece on $[t,\tau]$ and a continuation on
$[\tau,t_f]$. The cost separates in the same way:
\begin{align}
  J_{t,x}[u]
  &={}
  \int_t^\tau \ell\bigl(s,x(s),u(s)\bigr)\,ds\nonumber\\
  &\quad{}
  +\left[
    \Phi\bigl(x(t_f),t_f\bigr)
    +\int_\tau^{t_f}
      \ell\bigl(s,x(s),u(s)\bigr)\,ds
  \right].
  \label{eq:hjb-cost-split-at-intermediate-time}
\end{align}
Once the first piece reaches $x(\tau)$, the expression in brackets is itself
a control problem beginning from $(\tau,x(\tau))$. Its smallest possible
value is $V(\tau,x(\tau))$.

\begin{theorem}[Principle of optimality]
For every intermediate time $\tau\in[t,t_f]$, the value function satisfies
\begin{equation}
  \boxed{
  V(t,x)
  =
  \inf_{u(\cdot)\text{ on }[t,\tau]}
  \left\{
    \int_t^\tau \ell\bigl(s,x(s),u(s)\bigr)\,ds
    +V\bigl(\tau,x(\tau)\bigr)
  \right\}.}
  \label{eq:continuous-time-principle-of-optimality}
\end{equation}
Consequently, the continuation of an optimal control after time $\tau$ must
be optimal for the state reached at time $\tau$.
\end{theorem}

\begin{proof}
Take any complete admissible control. Its continuation after $\tau$ cannot
cost less than $V(\tau,x(\tau))$, because the value is the infimum over all
continuations from that state. Therefore its complete cost is at least the
quantity on the right-hand side of
\cref{eq:continuous-time-principle-of-optimality}. Taking the infimum over
complete controls gives one inequality.

Conversely, select a first control segment and then append a continuation
whose cost is arbitrarily close to $V(\tau,x(\tau))$. Concatenating the two
segments gives an admissible complete control with cost arbitrarily close to
the right-hand side. Taking the infimum over first segments gives the reverse
inequality. If optimal controls exist, ``arbitrarily close'' can be replaced
by an optimal continuation.
\end{proof}

This result is a consistency condition across time. It is not a greedy rule.
The controller does not minimize the immediate running cost alone. It
minimizes immediate cost plus the value of the state that remains. A robot may
temporarily spend more energy to move into a region from which collision
avoidance or final positioning becomes much cheaper.

The principle can also be understood by contradiction. Suppose an optimal
trajectory reaches some state $x(\tau)$ but then follows a continuation that
is not optimal from that state. Replacing only the continuation by a cheaper
one leaves the earlier trajectory unchanged and reduces the total cost. That
contradicts optimality of the original trajectory. This replacement argument
is the source of both Bellman recursion and HJB.

\subsection{Why the State Must Summarize the Past}

The value in \cref{eq:hjb-value-function-definition} depends on the current
state rather than the complete past trajectory. This is valid only when the
state contains all information from the past that can affect future dynamics,
costs, and constraints. We call such a state a \emph{sufficient state}.

For a mechanical system, position alone is usually not sufficient. Future
motion also depends on velocity, so a state such as $x=(q,\dot q)$ is
required. If actuator temperature limits future torque, temperature must also
be included. If the running cost penalizes changes in control, the previous
control can be added to the state. Once these variables are included, two
trajectories arriving at the same augmented state have the same set of
possible futures and the same cost-to-go.

This observation is more than bookkeeping. Dynamic programming fails if two
histories are assigned the same state even though they permit different
future behavior. The remedy is to enlarge the state until the future depends
only on the present state and future controls. The state may then be viewed as
a compressed memory of everything relevant from the past.

When $V$ is known, it defines feedback because the best current decision can
be recomputed from the current state. If a disturbance moves the system away
from the trajectory expected earlier, the controller evaluates the decision
at the new state. By contrast, a previously computed open-loop signal does
not change unless a new optimization is solved.

\subsection{Discrete-Time Bellman Recursion}
\label{sec:discrete-time-bellman-recursion}

The dynamic-programming idea is clearest in discrete time. Consider
\begin{equation}
  x_{k+1}=F_k(x_k,u_k),
  \qquad k=0,\ldots,N-1,
  \label{eq:deterministic-discrete-dynamics-hjb}
\end{equation}
with cost
\begin{equation}
  J
  =\Phi_N(x_N)
   +\sum_{k=0}^{N-1}\ell_k(x_k,u_k).
  \label{eq:deterministic-discrete-cost-hjb}
\end{equation}
Define $V_k(x)$ as the minimum cost from state $x$ at stage $k$. At the final
stage,
\begin{equation}
  V_N(x)=\Phi_N(x).
  \label{eq:discrete-bellman-terminal-condition}
\end{equation}
At stage $k$, choosing $u$ incurs the current cost $\ell_k(x,u)$ and moves the
state to $F_k(x,u)$. The smallest cost after that transition is
$V_{k+1}(F_k(x,u))$. Therefore
\begin{equation}
  \boxed{
  V_k(x)
  =\min_{u\in\mathcal U_k(x)}
  \left\{
    \ell_k(x,u)
    +V_{k+1}\bigl(F_k(x,u)\bigr)
  \right\}.}
  \label{eq:deterministic-discrete-bellman-recursion}
\end{equation}
This is the \emph{Bellman recursion}. The set
$\mathcal U_k(x)$ contains controls allowed at stage $k$ and state $x$. It
may encode actuator limits or controls whose next states remain feasible.

The optimal feedback at each stage is any minimizer:
\begin{equation}
  \mu_k^\star(x)
  \in\argmin_{u\in\mathcal U_k(x)}
  \left\{
    \ell_k(x,u)
    +V_{k+1}\bigl(F_k(x,u)\bigr)
  \right\}.
  \label{eq:discrete-bellman-optimal-feedback}
\end{equation}
One first sets the terminal value and then works backward through
$k=N-1,N-2,\ldots,0$. This process is called \emph{backward induction}.
Although values are computed backward, the controlled state still evolves
forward when the policy is used.

\begin{example}[A discrete-time quadratic integrator]
Consider
\begin{equation}
  x_{k+1}=x_k+u_k
\end{equation}
and
\begin{equation}
  J
  =\frac12s x_N^2
   +\frac12\sum_{k=0}^{N-1}
   \left(qx_k^2+ru_k^2\right),
  \qquad q\geq0,\quad r>0,\quad s\geq0.
\end{equation}
The terminal value is $V_N(x)=s x^2/2$. Suppose, working backward, that
\begin{equation}
  V_{k+1}(x)=\frac12p_{k+1}x^2.
\end{equation}
The Bellman recursion gives
\begin{align}
  V_k(x)
  =\min_u
  \left\{
    \frac12qx^2+\frac12ru^2
    +\frac12p_{k+1}(x+u)^2
  \right\}.
  \label{eq:discrete-scalar-lqr-bellman-step}
\end{align}
Differentiation with respect to $u$ gives
\begin{equation}
  (r+p_{k+1})u+p_{k+1}x=0,
\end{equation}
so the minimizing feedback is
\begin{equation}
  u_k^\star
  =-\frac{p_{k+1}}{r+p_{k+1}}x_k.
  \label{eq:discrete-scalar-lqr-feedback-hjb}
\end{equation}
Substitution into \cref{eq:discrete-scalar-lqr-bellman-step} gives
\begin{equation}
  V_k(x)=\frac12p_kx^2,
  \qquad
  p_k=q+p_{k+1}-\frac{p_{k+1}^2}{r+p_{k+1}}
      =q+\frac{rp_{k+1}}{r+p_{k+1}}.
  \label{eq:discrete-scalar-riccati-recursion-hjb}
\end{equation}
Thus quadratic value functions reproduce themselves under Bellman recursion.
The coefficient $p_k$ measures how expensive state error is when $N-k$ steps
remain, and the feedback gain converts that future cost into corrective
action. This is the discrete-time counterpart of the Riccati structure in
Chapter~7.
\end{example}

Bellman recursion separates a long decision sequence into repeated one-step
problems. It does not eliminate optimization; it reorganizes it. Instead of
optimizing all controls simultaneously for one initial state, it computes the
best continuation from every state at every stage. That broader answer is why
dynamic programming produces feedback and why it can require much more
computation.

\section{The Hamilton--Jacobi--Bellman Equation}

\subsection{From a Short Time Step to HJB}

We now apply the principle of optimality over a short continuous-time interval
$[t,t+h]$, where $h>0$ is small. Setting $\tau=t+h$ in
\cref{eq:continuous-time-principle-of-optimality} gives
\begin{equation}
  V(t,x)
  =\inf_{u(\cdot)\text{ on }[t,t+h]}
  \left\{
    \int_t^{t+h}\ell(s,x(s),u(s))\,ds
    +V\bigl(t+h,x(t+h)\bigr)
  \right\}.
  \label{eq:hjb-short-interval-bellman-identity}
\end{equation}
The first term measures what is paid immediately. The second assigns the
optimal remaining cost to the state reached after the short interval.

Assume for now that $V$ is continuously differentiable and that the dynamics
and running cost are smooth. Hold the control equal to a constant value $u$
over the short interval. To first order in $h$,
\begin{align}
  x(t+h)
  &=x+h f(t,x,u)+o(h),
  \label{eq:hjb-short-state-expansion}\\
  \int_t^{t+h}\ell(s,x(s),u)\,ds
  &=h\ell(t,x,u)+o(h).
  \label{eq:hjb-short-cost-expansion}
\end{align}
The notation $o(h)$ denotes terms that vanish faster than $h$, meaning that
$o(h)/h\to0$ as $h\to0$. A first-order Taylor expansion of the value gives
\begin{align}
  V\bigl(t+h,x(t+h)\bigr)
  &={}
  V(t,x)+hV_t(t,x)\nonumber\\
  &\quad{}
  +h\nabla_xV(t,x)\trans f(t,x,u)+o(h),
  \label{eq:hjb-value-taylor-expansion}
\end{align}
where $V_t$ is the time derivative and $\nabla_xV$ is the gradient with
respect to the state.

Substituting these expansions into
\cref{eq:hjb-short-interval-bellman-identity} gives, for a constant control
$u$ over the short interval,
\begin{align}
  V(t,x)
  =\inf_{u\in\mathcal U(t,x)}
  \bigl\{
    V(t,x)
    +h\ell(t,x,u)
    +hV_t(t,x)
    +h\nabla_xV(t,x)\trans f(t,x,u)
    +o(h)
  \bigr\}.
\end{align}
The common term $V(t,x)$ appears on both sides because, over a vanishingly
short interval, the system accumulates almost no cost and moves only an
infinitesimal distance. Canceling this term and dividing by $h$ yields
\begin{equation}
  0
  =\inf_{u\in\mathcal U(t,x)}
  \left\{
    \ell(t,x,u)
    +V_t(t,x)
    +\nabla_xV(t,x)\trans f(t,x,u)
    +\frac{o(h)}{h}
  \right\}.
\end{equation}
As $h\to0$, the final term vanishes. Only the current value of the control
remains: its effect over the short interval is summarized by the instantaneous
dynamics $f(t,x,u)$ and running cost $\ell(t,x,u)$. Under conditions ensuring
that the infimum is attained, it may be written as a minimum, giving
\begin{equation}
  \boxed{
  0
  =V_t(t,x)
   +\min_{u\in\mathcal U(t,x)}
   \left\{
     \ell(t,x,u)
     +\nabla_xV(t,x)\trans f(t,x,u)
   \right\}.}
  \label{eq:finite-horizon-hjb-equation}
\end{equation}
The minimization is pointwise: at each fixed pair $(t,x)$, it compares all
admissible current controls. The term $V_t$ records how the remaining optimal
cost changes merely because time advances, while
$\nabla_xV\trans f$ records how it changes because the selected control moves
the state. The minimizing control balances these changes against the immediate
running cost $\ell$.

Together with
\cref{eq:hjb-value-terminal-condition}, this is the finite-horizon
\emph{Hamilton--Jacobi--Bellman equation}. It is a partial differential
equation because the unknown $V(t,x)$ depends on time and all state
coordinates, and the equation contains derivatives with respect to those
variables. It is generally nonlinear because of the minimization and because
the dynamics can be nonlinear.

Define the Hamiltonian
\begin{equation}
  H(t,x,u,p)
  =\ell(t,x,u)+p\trans f(t,x,u),
  \label{eq:hjb-hamiltonian-definition}
\end{equation}
where $p\in\R^n$ is a placeholder for a state gradient. This is the same
Hamiltonian form used in Chapter~6. The HJB equation can then be written
compactly as
\begin{equation}
  0
  =V_t(t,x)
   +\min_{u\in\mathcal U(t,x)}
    H\bigl(t,x,u,\nabla_xV(t,x)\bigr).
  \label{eq:hjb-hamiltonian-form}
\end{equation}

The term $\nabla_xV\trans f$ predicts how the future value changes because
the chosen control moves the state. Thus HJB balances two rates: the running
cost paid now and the rate at which the motion changes the remaining optimal
cost. A control that has small immediate cost can still be poor if it moves
the system toward states with rapidly increasing value.

The terminal cost supplies data at $t_f$, and HJB propagates that information
backward through time. For this reason finite-horizon value functions and
feedback laws are generally time varying. Near the terminal time, little time
remains to correct an error, so the best action can differ from the action at
the same state earlier in the horizon.

\subsection{A Verification Theorem}

The short-time derivation motivates HJB. The following result explains why a
smooth solution actually certifies optimality.

\begin{theorem}[HJB verification]
Suppose a continuously differentiable function $W(t,x)$ satisfies
\begin{equation}
  0
  =W_t(t,x)
   +\min_{u\in\mathcal U(t,x)}
   \left\{
     \ell(t,x,u)+\nabla_xW(t,x)\trans f(t,x,u)
   \right\},
  \label{eq:hjb-verification-equation}
\end{equation}
and $W(t_f,x)=\Phi(x,t_f)$. Suppose also that an admissible feedback
$u=\mu^\star(t,x)$ attains the displayed minimum and produces a well-defined
trajectory. Then
\begin{equation}
  W(t,x)=V(t,x),
\end{equation}
and $\mu^\star$ is an optimal feedback law.
\end{theorem}

\begin{proof}
For any admissible control $u(s)$ and its state trajectory, the minimum in
\cref{eq:hjb-verification-equation} is no larger than the expression evaluated
at that control. Hence
\begin{equation}
  W_t+\nabla_xW\trans f(s,x,u)+\ell(s,x,u)\geq0.
\end{equation}
Along the trajectory, the chain rule gives
\begin{equation}
  \frac{d}{ds}W(s,x(s))
  =W_t+\nabla_xW\trans\dot x
  =W_t+\nabla_xW\trans f(s,x,u).
\end{equation}
Therefore
\begin{equation}
  \ell(s,x(s),u(s))
  \geq-\frac{d}{ds}W(s,x(s)).
\end{equation}
Integrating from $t$ to $t_f$ and using the terminal condition gives
\begin{align}
  J_{t,x}[u]
  &=\Phi(x(t_f),t_f)
    +\int_t^{t_f}\ell(s,x,u)\,ds\nonumber\\
  &\geq W(t_f,x(t_f))+W(t,x)-W(t_f,x(t_f))\nonumber\\
  &=W(t,x).
  \label{eq:hjb-verification-lower-bound}
\end{align}
Thus no admissible control can cost less than $W(t,x)$. When
$u=\mu^\star(t,x)$ attains the HJB minimum, every inequality above becomes an
equality. Its cost is exactly $W(t,x)$, so $W=V$ and the feedback is optimal.
\end{proof}

The verification theorem is stronger than a stationarity test along one
trajectory. A function satisfying HJB over a domain provides a lower bound on
the cost of every admissible control from every state in that domain, and a
minimizing feedback attains that bound. The difficulty is that finding such a
function is usually much harder than checking necessary conditions along one
curve.

\subsection{Recovering the Optimal Feedback Law}

Once the value function is known, the optimal feedback follows from the
pointwise minimization in HJB:
\begin{equation}
  \boxed{
  \mu^\star(t,x)
  \in\argmin_{u\in\mathcal U(t,x)}
  \left\{
    \ell(t,x,u)
    +\nabla_xV(t,x)\trans f(t,x,u)
  \right\}.}
  \label{eq:hjb-optimal-feedback-general}
\end{equation}
The optimization is pointwise because $t$, $x$, and $\nabla_xV(t,x)$ are
fixed while choosing the current input. All future consequences have already
been compressed into $V$.

The gradient has a useful physical meaning. For a small state displacement
$\delta x$,
\begin{equation}
  V(t,x+\delta x)-V(t,x)
  \approx\nabla_xV(t,x)\trans\delta x.
\end{equation}
Thus the $i$th component of the gradient measures the marginal change in
optimal future cost caused by a small change in the $i$th state coordinate.
The feedback selects a velocity $f(t,x,u)$ through state space while accounting
for how strongly the value rises or falls in that direction.

For a control-affine system
\begin{equation}
  \dot x=f_0(t,x)+G(t,x)u
  \label{eq:hjb-control-affine-dynamics}
\end{equation}
with running cost
\begin{equation}
  \ell(t,x,u)
  =L(t,x)+\frac12u\trans R(t,x)u,
  \qquad R(t,x)\succ0,
  \label{eq:hjb-control-affine-quadratic-cost}
\end{equation}
the terms depending on $u$ are
\begin{equation}
  \frac12u\trans Ru
  +\nabla_xV\trans Gu.
\end{equation}
For unconstrained control, differentiation with respect to $u$ gives
\begin{equation}
  Ru+G\trans\nabla_xV=0,
\end{equation}
and hence
\begin{equation}
  \boxed{
  \mu^\star(t,x)
  =-R(t,x)^{-1}G(t,x)\trans\nabla_xV(t,x).}
  \label{eq:hjb-control-affine-optimal-feedback}
\end{equation}
This formula generalizes the LQR law. The value gradient plays the role of the
costate, while $G\trans$ extracts the part of that gradient that the actuators
can influence.

If controls are constrained, the unconstrained formula may be infeasible. The
correct law is still the minimizer in
\cref{eq:hjb-optimal-feedback-general}. For a scalar input with
$u\in[u_{\min},u_{\max}]$ and quadratic effort, the constrained minimizer is
the projection
\begin{equation}
  u^\star
  =\Pi_{[u_{\min},u_{\max}]}
  \left(-R^{-1}G\trans\nabla_xV\right),
  \label{eq:hjb-bounded-scalar-feedback}
\end{equation}
where $\Pi$ clips a scalar to the indicated interval. Saturation therefore
appears naturally from Hamiltonian minimization rather than being added after
the optimal law is derived.

\subsection{Linear--Quadratic Examples}
\label{sec:hjb-linear-quadratic-examples}

The Riccati equation from Chapter~7 follows directly from HJB. Consider
\begin{equation}
  \dot x=A(t)x+B(t)u
  \label{eq:hjb-lqr-dynamics}
\end{equation}
and
\begin{equation}
  J
  =\frac12x(t_f)\trans S_fx(t_f)
   +\frac12\int_t^{t_f}
    \left(x\trans Qx+u\trans Ru\right)ds,
  \label{eq:hjb-lqr-objective}
\end{equation}
with $Q\succeq0$, $R\succ0$, and $S_f\succeq0$. Motivated by the quadratic
terminal cost and the result of Chapter~7, seek a value function of the form
\begin{equation}
  V(t,x)=\frac12x\trans P(t)x,
  \qquad P(t)=P(t)\trans.
  \label{eq:hjb-lqr-value-ansatz}
\end{equation}
Its derivatives are
\begin{equation}
  V_t=\frac12x\trans\dot P x,
  \qquad
  \nabla_xV=Px.
  \label{eq:hjb-lqr-value-derivatives}
\end{equation}
Substitution into HJB gives
\begin{align}
  0
  ={}&\frac12x\trans\dot P x
  +\min_u
  \left\{
    \frac12x\trans Qx
    +\frac12u\trans Ru
    +x\trans P(Ax+Bu)
  \right\}.
  \label{eq:hjb-lqr-before-control-minimization}
\end{align}
The derivative of the minimized expression with respect to $u$ is
\begin{equation}
  Ru+B\trans Px.
\end{equation}
Therefore
\begin{equation}
  \boxed{u^\star=-R^{-1}B\trans Px.}
  \label{eq:hjb-lqr-optimal-control}
\end{equation}
Substituting this law into
\cref{eq:hjb-lqr-before-control-minimization} yields
\begin{align}
  0
  ={}&\frac12x\trans\dot P x
  +\frac12x\trans Qx
  +x\trans PAx\nonumber\\
  &\quad{}
  -\frac12x\trans PBR^{-1}B\trans Px.
  \label{eq:hjb-lqr-after-control-minimization}
\end{align}
Because a scalar equals its transpose,
\begin{equation}
  x\trans PAx
  =\frac12x\trans(PA+A\trans P)x.
\end{equation}
Hence
\begin{equation}
  0
  =\frac12x\trans
  \left(
    \dot P+Q+PA+A\trans P-PBR^{-1}B\trans P
  \right)x.
\end{equation}
The HJB equation must hold for every state $x$, so the matrix coefficient must
vanish:
\begin{equation}
  \boxed{
  -\dot P
  =Q+A\trans P+PA-PBR^{-1}B\trans P,
  \qquad P(t_f)=S_f.}
  \label{eq:hjb-derived-differential-riccati-equation}
\end{equation}
This is exactly the differential Riccati equation obtained from PMP in
Chapter~7.

The two derivations emphasize different meanings of $P$. In the PMP
derivation, $P$ maps state to costate through $\lambda=Px$. In the HJB
derivation, $P$ is the curvature matrix of the quadratic value function. The
connection is
\begin{equation}
  \lambda=\nabla_xV=Px.
  \label{eq:hjb-lqr-costate-value-gradient}
\end{equation}
Geometrically, the level sets of $V$ are ellipsoids
$x\trans Px=\text{constant}$. States on a large ellipsoid have greater optimal
future cost. The feedback pushes the state toward lower-value ellipsoids while
balancing state reduction against control effort.

\paragraph{Discrete-time LQR.}
The scalar integrator in \cref{sec:discrete-time-bellman-recursion} showed how
a quadratic value function is preserved by one Bellman update. The same
calculation gives the matrix-valued discrete-time LQR formulas. Consider the
linear time-varying system
\begin{equation}
  x_{k+1}=A_kx_k+B_ku_k,
  \qquad k=0,\ldots,N-1,
  \label{eq:discrete-time-lqr-dynamics}
\end{equation}
with the finite-horizon cost
\begin{equation}
  J
  =\frac12x_N\trans S_Nx_N
   +\frac12\sum_{k=0}^{N-1}
    \left(x_k\trans Q_kx_k+u_k\trans R_ku_k\right),
  \label{eq:discrete-time-lqr-cost}
\end{equation}
where $Q_k\succeq0$, $R_k\succ0$, and $S_N\succeq0$. Write the optimal cost
remaining at stage $k+1$ as
\begin{equation}
  V_{k+1}(x)=\frac12x\trans P_{k+1}x.
  \label{eq:discrete-time-lqr-value-function}
\end{equation}
At the terminal stage, the value is exactly the terminal cost, so
$P_N=S_N$.

Applying the Bellman recursion from
\cref{eq:deterministic-discrete-bellman-recursion} gives
\begin{align}
  V_k(x)
  =\min_u\ \frac12\bigl(&x\trans Q_kx+u\trans R_ku\nonumber\\
  &+(A_kx+B_ku)\trans P_{k+1}(A_kx+B_ku)\bigr).
  \label{eq:discrete-time-lqr-bellman-step}
\end{align}
Collecting the terms that depend on $u$ gives
\begin{align}
  V_k(x)=\min_u\ \frac12\bigl(&
  x\trans(Q_k+A_k\trans P_{k+1}A_k)x
  +2u\trans B_k\trans P_{k+1}A_kx\nonumber\\
  &+u\trans(R_k+B_k\trans P_{k+1}B_k)u\bigr).
  \label{eq:discrete-time-lqr-bellman-expanded}
\end{align}
Because $R_k\succ0$ and $P_{k+1}\succeq0$, the control Hessian
\begin{equation}
  G_k=R_k+B_k\trans P_{k+1}B_k
  \label{eq:discrete-time-lqr-control-hessian}
\end{equation}
is positive definite. Differentiating with respect to $u$ gives
\begin{equation}
  G_ku+B_k\trans P_{k+1}A_kx=0.
\end{equation}
Hence the minimizing control is the linear feedback law
\begin{equation}
  \boxed{
  u_k^\star=-K_kx_k,
  \qquad
  K_k=
  (R_k+B_k\trans P_{k+1}B_k)^{-1}
  B_k\trans P_{k+1}A_k.}
  \label{eq:discrete-time-lqr-feedback}
\end{equation}
In an implementation, the inverse in this formula is usually not formed
explicitly. Instead, one solves the positive-definite linear system
$G_kK_k=B_k\trans P_{k+1}A_k$, for example by a Cholesky factorization.

Substituting the minimizing control back into the Bellman equation shows that
$V_k(x)=x\trans P_kx/2$, where
\begin{equation}
  \boxed{
  \begin{aligned}
  P_k={}&Q_k+A_k\trans P_{k+1}A_k\\
       &-A_k\trans P_{k+1}B_k
       (R_k+B_k\trans P_{k+1}B_k)^{-1}
       B_k\trans P_{k+1}A_k,
  \qquad P_N=S_N.
  \end{aligned}}
  \label{eq:discrete-time-riccati-recursion}
\end{equation}
This backward recurrence is the \emph{discrete-time Riccati equation}, also
called the Riccati difference equation. Starting from $P_N$, one computes
$P_{N-1},P_{N-2},\ldots,P_0$ and the corresponding gains in reverse time.
The resulting controller is then applied forward in time.

The term $B_k\trans P_{k+1}B_k$ distinguishes the discrete-time gain from the
continuous-time formula $R^{-1}B\trans P$. In discrete time, the current input
directly changes the next state, whose future cost has curvature $P_{k+1}$;
this additional curvature therefore appears in the control Hessian. For a
continuous plant sampled every $h$ seconds, matrices $A_k$ and $B_k$ are
obtained from the chosen discretization or exact zero-order-hold model, and
the discrete Riccati recursion should be used for the resulting sampled
system. The same backward recursion is the core computation inside the local
LQR subproblems used by iLQR in Chapter~9; linear and cross terms in those
local models additionally produce a feedforward correction.

\paragraph{Discrete-time algebraic Riccati equation.}
For the time-invariant system $x_{k+1}=Ax_k+Bu_k$ with constant weights $Q$
and $R$, the finite-horizon matrices may approach a constant matrix $P$ far
from the terminal boundary. Setting $P_k=P_{k+1}=P$ in
\cref{eq:discrete-time-riccati-recursion} gives
\begin{equation}
  \boxed{
  P=Q+A\trans PA
    -A\trans PB(R+B\trans PB)^{-1}B\trans PA.}
  \label{eq:discrete-time-algebraic-riccati-equation}
\end{equation}
This is the \emph{discrete-time algebraic Riccati equation} (DARE). Under the
discrete-time stabilizability and detectability assumptions, its stabilizing
solution gives
\begin{equation}
  K=(R+B\trans PB)^{-1}B\trans PA,
  \qquad
  x_{k+1}=(A-BK)x_k,
  \label{eq:discrete-time-infinite-horizon-feedback}
\end{equation}
with every closed-loop eigenvalue inside the unit disk. The continuous-time
algebraic Riccati equation in Chapter~7 and the DARE play analogous roles,
but the appropriate equation is determined by whether the plant and
controller are formulated in continuous or discrete time.

These examples also explain why the general HJB equation is difficult. LQR
works because the class of quadratic functions is preserved by the Bellman
update. For nonlinear systems or nonquadratic costs, the value function need
not belong to any simple finite-dimensional family, so an entire function over
state space must be represented.

\subsection{Stationary and Discounted HJB Equations}

For an autonomous infinite-horizon problem
\begin{equation}
  V_\infty(x)
  =\inf_{u(\cdot)}
   \int_0^\infty\ell(x(t),u(t))\,dt,
  \qquad \dot x=f(x,u),
  \label{eq:hjb-undiscounted-infinite-horizon-value}
\end{equation}
the amount of future time is always infinite. When the value is finite and
the problem is time invariant, the value depends only on $x$. The time
derivative disappears from HJB, giving the \emph{stationary HJB equation}
\begin{equation}
  \boxed{
  0
  =\min_{u\in\mathcal U(x)}
   \left\{
     \ell(x,u)+\nabla V_\infty(x)\trans f(x,u)
   \right\}.}
  \label{eq:stationary-undiscounted-hjb}
\end{equation}
The word stationary means that neither the value nor the optimal feedback
depends explicitly on clock time. It does not mean that the state is fixed.

There is no terminal-time boundary condition. Instead, the desired solution
is selected by the control objective and long-run behavior. For regulation to
an equilibrium $x_e$, one commonly has
\begin{equation}
  V_\infty(x_e)=0,
  \qquad
  V_\infty(x)>0\quad\text{for }x\neq x_e,
\end{equation}
together with a feedback that stabilizes $x_e$. Merely finding an algebraic
solution of the stationary equation is not enough; it must correspond to the
correct finite-cost and stabilizing behavior. This is analogous to selecting
the stabilizing solution of the algebraic Riccati equation.

For the discounted objective
\begin{equation}
  V_\rho(x)
  =\inf_{u(\cdot)}
  \int_0^\infty e^{-\rho t}\ell(x(t),u(t))\,dt,
  \qquad \rho>0,
  \label{eq:hjb-discounted-value}
\end{equation}
the Bellman split over $[0,h]$ is
\begin{equation}
  V_\rho(x)
  =\inf_{u(\cdot)}
  \left\{
    \int_0^h e^{-\rho s}\ell(x(s),u(s))\,ds
    +e^{-\rho h}V_\rho(x(h))
  \right\}.
\end{equation}
Using $e^{-\rho h}=1-\rho h+o(h)$ and the same short-time expansion as before
gives
\begin{equation}
  \boxed{
  \rho V_\rho(x)
  =\min_{u\in\mathcal U(x)}
   \left\{
     \ell(x,u)+\nabla V_\rho(x)\trans f(x,u)
   \right\}.}
  \label{eq:stationary-discounted-hjb}
\end{equation}
The extra term $\rho V_\rho$ reflects the loss of weight assigned to distant
future cost. A large discount rate makes the controller more concerned with
near-term consequences.

\subsection{Minimum Time and a Nonsmooth Value Function}

The classical HJB derivation assumed that the value function is smooth, but
optimal value functions can develop corners even for very simple dynamics.
Consider the speed-limited integrator
\begin{equation}
  \dot x=u,
  \qquad |u|\leq1,
  \label{eq:hjb-minimum-time-integrator}
\end{equation}
and let the task end when the state first reaches the target $x=0$. With the
running cost $\ell=1$, the accumulated cost is exactly the travel time. The
minimum time from an initial state $x$ is
\begin{equation}
  V(x)=|x|,
  \label{eq:hjb-minimum-time-absolute-value}
\end{equation}
because the state can move toward the origin at unit speed, and no admissible
control can reach it faster.

Away from the target, the stationary HJB equation is
\begin{align}
  0
  &=\min_{|u|\leq1}\left\{1+V_x(x)u\right\}\nonumber\\
  &=1-|V_x(x)|.
  \label{eq:hjb-minimum-time-equation}
\end{align}
For $x>0$, $V_x=1$, and the minimizing control is $u=-1$. For $x<0$,
$V_x=-1$, and the minimizing control is $u=1$. Thus the optimal control always
points toward the target. The two feedback branches meet at $x=0$, where
$|x|$ has left derivative $-1$ and right derivative $1$, but no single
derivative. This change of optimal direction creates the corner in the value
function.

The HJB equation is imposed away from the target, so $V(x)=|x|$ is a
\emph{classical solution} on $\R\setminus\{0\}$. At the target, the task has
ended and the value is specified by the boundary condition
\begin{equation}
  V(0)=0.
\end{equation}
No control decision or differential equation is required after the target is
reached. More generally, $V$ is regarded as a solution of the HJB boundary-
value problem in the \emph{viscosity} sense. Roughly, a viscosity solution
need not be differentiable everywhere. Instead, the HJB inequalities are
tested using smooth functions that touch the value function locally from
above or below. This generalized notion preserves the Bellman principle and
identifies the correct value function even in the presence of corners.

This lack of differentiability is not a defect in the control problem. It
records a change in the active optimal strategy. Similar corners arise when
different routes around an obstacle have equal cost, when an actuator enters
or leaves saturation, or when several target points are equally good. At a
smooth point, the value gradient defines one local direction of increasing
cost. At a corner, several one-sided slopes or optimal strategies meet. A full
theory of viscosity solutions is beyond this chapter, but the Bellman
principle remains meaningful, and the grid methods described next can
approximate nonsmooth value functions without differentiating them explicitly
at every point.

\section{Computation and Limitations}

\subsection{Grid-Based Value Iteration}

Except in structured examples such as LQR, HJB rarely has a closed-form
solution. A direct numerical idea is to store the value on a grid of states
and apply an approximate Bellman update backward in time. This approach makes
the connection between the discrete and continuous theories especially clear.

Divide the horizon into times
\begin{equation}
  t_k=t_0+k\Delta t,
  \qquad k=0,\ldots,N,
  \qquad t_N=t_f,
\end{equation}
and choose state grid points $x_i$. Over one small time step, forward Euler
integration approximates the dynamics by
\begin{equation}
  x^+
  =x_i+\Delta t\,f(t_k,x_i,u).
  \label{eq:hjb-grid-euler-step}
\end{equation}
The next state $x^+$ will usually lie between grid points. Let
$\mathcal I[V_{k+1}](x^+)$ denote an \emph{interpolation}: an estimate of the
value at $x^+$ formed from nearby stored values. In one dimension, linear
interpolation connects adjacent samples by a straight line. In several
dimensions, the same idea is applied within the surrounding grid cell.

The approximate Bellman update is
\begin{equation}
  \boxed{
  V_k(x_i)
  =\min_{u\in\mathcal U_d(x_i)}
  \left\{
    \Delta t\,\ell(t_k,x_i,u)
    +\mathcal I[V_{k+1}]
      \left(x_i+\Delta t f(t_k,x_i,u)\right)
  \right\},}
  \label{eq:hjb-grid-bellman-update}
\end{equation}
with terminal initialization
\begin{equation}
  V_N(x_i)=\Phi(x_i,t_f).
  \label{eq:hjb-grid-terminal-initialization}
\end{equation}
Here $\mathcal U_d(x_i)$ is either the original control set, if the inner
minimization can be solved continuously, or a finite collection of sampled
controls. The minimizing control is stored as the approximate feedback
$\mu_k(x_i)$.

The computation proceeds as follows:
\begin{enumerate}
  \item assign the terminal cost to every state-grid point;
  \item at time $t_{N-1}$, test candidate controls at each state point,
    predict the next state, interpolate its terminal value, and keep the
    smallest total;
  \item repeat for $t_{N-2},t_{N-3},\ldots,t_0$; and
  \item during closed-loop operation, interpolate the stored feedback or
    repeat the one-step minimization using the stored value.
\end{enumerate}

The algorithm propagates cost information backward from the terminal
condition. A large terminal penalty near an undesirable final state affects
nearby states one step earlier, then states two steps earlier, and so on. In a
minimum-time problem, low values spread outward from the target like an
arrival-time wave. The feedback at each state points toward a neighboring
state with an attractive combination of immediate and future cost.

The update in \cref{eq:hjb-grid-bellman-update} is sometimes called a
\emph{semi-Lagrangian} update. The name means that the grid is fixed, but each
candidate control follows the dynamics forward to an off-grid next state
before interpolation. Expanding the interpolated value for small $\Delta t$
recovers the HJB equation, so the update is a discrete approximation of the
same optimality principle.

Several design choices affect the computed solution:
\begin{itemize}
  \item A smaller time step reduces integration error but increases the
    number of backward stages.
  \item A finer state grid reduces spatial error but increases memory and
    computation.
  \item A coarse control sample can miss the true minimizing action.
  \item Interpolation must not create artificial values that make unsafe or
    infeasible states appear attractive.
  \item The boundary of the computational domain requires a deliberate rule.
    One may impose a known boundary cost, forbid transitions outside the
    domain, or enlarge the domain so relevant trajectories remain inside.
\end{itemize}

State constraints can be represented by assigning an infinite or very large
value to forbidden states, but a large finite penalty is only an
approximation. If collision must be impossible, the numerical transition rule
should explicitly reject controls whose trajectories cross an obstacle rather
than merely checking the endpoint of a large time step.

\begin{example}[A grid update for minimum-time motion]
For the integrator in \cref{eq:hjb-minimum-time-integrator}, choose
$u\in\{-1,1\}$ and let each time step cost $\Delta t$. Away from the target,
the update is
\begin{equation}
  V_k(x_i)
  =\Delta t
   +\min_{u\in\{-1,1\}}
    \mathcal I[V_{k+1}](x_i+\Delta t\,u).
\end{equation}
The minimizing control selects the neighboring direction with smaller future
arrival time. On a sufficiently fine grid, the computed value approaches
$|x|$ and the stored feedback points toward the origin.
\end{example}

\subsection{The Curse of Dimensionality}

Grid methods are conceptually simple but scale poorly. Suppose each state
coordinate is represented by $M$ grid points. An $n$-dimensional state then
requires
\begin{equation}
  M^n
  \label{eq:hjb-grid-state-count}
\end{equation}
value samples at every time layer. This exponential growth with dimension is
called the \emph{curse of dimensionality}.

For $M=101$, a two-dimensional grid contains $10{,}201$ points, a
four-dimensional grid contains more than $10^8$ points, and a six-dimensional
grid contains more than $10^{12}$ points. Storing one double-precision value
at each six-dimensional point would already require several terabytes, before
storing a feedback law, obstacles, or additional time layers. A manipulator
with $d$ joints usually has at least $2d$ state variables when positions and
velocities are included, so a uniform grid becomes impractical even for
moderate $d$.

The control search can add another exponential factor. If each of $m$ input
coordinates is sampled at $M_u$ values, then a full control grid contains
$M_u^m$ candidate actions. A finite-horizon calculation with $N$ time steps
therefore has work proportional, in a basic implementation, to
\begin{equation}
  N M^n M_u^m.
\end{equation}
Interpolation and collision checking increase the cost further.

This limitation motivates several alternatives. When the model has special
structure, as in LQR, the value can be represented by a small matrix rather
than a grid. Local trajectory methods approximate the value near a nominal
motion; Chapter~9 develops this idea through iterative LQR. Model predictive
control repeatedly solves a finite-horizon problem only near the currently
visited state rather than filling an entire state domain. Approximate dynamic
programming represents the value or policy by a parameterized function fitted
from samples. These approaches sacrifice some combination of global coverage,
exactness, or guarantees in exchange for tractability.

The curse of dimensionality does not make HJB irrelevant. The equation still
states what an optimal value function must satisfy, guides the design of
approximations, explains the structure of optimal feedback, and provides
benchmarks for low-dimensional systems.

\subsection{HJB versus Pontryagin's Minimum Principle}

Dynamic programming and PMP are not competing definitions of optimality.
They describe the same problem at different scales. HJB studies the optimal
cost from every state, while PMP studies necessary conditions along one
candidate optimal trajectory. When the value function is smooth, their
variables are connected directly.

Let $x^\star(t)$ be generated by a smooth HJB-minimizing feedback, and define
\begin{equation}
  \lambda(t)
  =\nabla_xV\bigl(t,x^\star(t)\bigr).
  \label{eq:hjb-pmp-costate-gradient-relation}
\end{equation}
Along the optimal feedback, HJB reads
\begin{equation}
  V_t+H(t,x,u^\star,V_x)=0.
  \label{eq:hjb-along-optimal-feedback}
\end{equation}
Assume for simplicity that the control minimizer is interior and smooth, so
$H_u=0$. Differentiate
\cref{eq:hjb-along-optimal-feedback} with respect to $x$. The dependence of
$H$ on $p=V_x$ contributes $V_{xx}H_p$, and the dependence through
$u^\star(t,x)$ contributes $H_u u_x^\star=0$. Therefore
\begin{equation}
  V_{xt}+H_x+V_{xx}H_p=0.
  \label{eq:hjb-spatial-derivative-along-optimum}
\end{equation}
Since $H_p=f=\dot x^\star$, the total derivative of
\cref{eq:hjb-pmp-costate-gradient-relation} is
\begin{align}
  \dot\lambda
  &=V_{xt}+V_{xx}\dot x^\star\nonumber\\
  &=V_{xt}+V_{xx}H_p
   =-H_x.
  \label{eq:hjb-implies-pmp-costate-equation}
\end{align}
The HJB minimization also gives
\begin{equation}
  u^\star
  \in\argmin_{u\in\mathcal U}H(t,x^\star,u,\lambda),
\end{equation}
and the terminal condition gives
\begin{equation}
  \lambda(t_f)=\nabla_x\Phi(x^\star(t_f),t_f).
\end{equation}
Thus the state equation, costate equation, Hamiltonian minimization, and
terminal costate condition of PMP all follow along a smooth HJB-optimal
trajectory.

The identity $\lambda=\nabla_xV$ gives the costate a concrete interpretation.
It is the sensitivity of optimal future cost to a perturbation of the current
state. PMP propagates that sensitivity along one trajectory. HJB constructs
the entire field of sensitivities over a state domain.

\begin{center}
\begin{tabular}{
  @{}
  >{\raggedright\arraybackslash}p{0.18\linewidth}
  >{\raggedright\arraybackslash}p{0.36\linewidth}
  >{\raggedright\arraybackslash}p{0.36\linewidth}
  @{}
}
\toprule
Aspect & Pontryagin's principle & Dynamic programming and HJB\\
\midrule
Primary object
& One state--costate trajectory
& A value function over time and state\\
Typical result
& An open-loop extremal for one initial condition
& A feedback law for all represented states\\
Optimality role
& Generally necessary conditions
& A smooth verified solution certifies global optimality on its domain\\
Main computation
& Ordinary differential equations with split boundary conditions
& A nonlinear partial differential equation or Bellman recursion\\
Main difficulty
& Boundary-value sensitivity and multiple extremals
& Exponential growth with state dimension\\
\bottomrule
\end{tabular}
\end{center}

PMP is often attractive for a high-dimensional system when only a few initial
conditions matter, because it avoids constructing a function over the full
state space. HJB is attractive when a low-dimensional problem requires a
global feedback policy, explicit treatment of many initial states, or a value
map for planning and safety analysis. Hybrid strategies are common: a global
or coarse value approximation guides motion, while local trajectory
optimization or LQR supplies accurate control near a nominal path.

Neither method removes modeling choices. A globally solved HJB equation is
only optimal for its stated dynamics, cost, constraints, and domain. A PMP
trajectory is only relevant if it satisfies all constraints and corresponds
to a minimum rather than another extremal. In both viewpoints, the quality of
the controller begins with the quality of the problem formulation.

\paragraph{Exercises.}

\begin{exercise}
Explain why the continuation of an optimal trajectory must be optimal from
every intermediate state it reaches. Why does this statement not imply that
the optimal control minimizes the running cost at each instant?
\end{exercise}
\begin{proof}[Short solution]
If a cheaper continuation existed, replacing the original tail by that
continuation would reduce the complete cost and contradict optimality. The
controller minimizes current cost plus optimal future cost, not current cost
alone, so it may accept a larger immediate cost to reach a better future
state.
\end{proof}

\begin{exercise}
A robot has position $q$ and velocity $v$, but a proposed dynamic-programming
model uses only $q$ as its state. Explain why this state is generally
insufficient. Give one additional example of a variable that may need to be
included in an augmented state.
\end{exercise}
\begin{proof}[Short solution]
Two trajectories can reach the same position with different velocities and
therefore have different possible futures and costs. The state should include
$v$. Other examples include actuator temperature, battery charge, or the
previous control when changes in control are penalized.
\end{proof}

\begin{exercise}
For the discrete-time system $x_{k+1}=F_k(x_k,u_k)$, derive the Bellman
recursion by separating the first stage cost from the remaining cost.
\end{exercise}
\begin{proof}[Short solution]
After choosing $u$ at state $x$, the first cost is $\ell_k(x,u)$ and the next
state is $F_k(x,u)$. The smallest remaining cost is
$V_{k+1}(F_k(x,u))$. Minimizing their sum gives
\begin{equation}
  V_k(x)
  =\min_{u\in\mathcal U_k(x)}
   \left\{\ell_k(x,u)+V_{k+1}(F_k(x,u))\right\}.
\end{equation}
\end{proof}

\begin{exercise}
For one remaining step of the scalar system $x_1=x_0+u_0$ with cost
\begin{equation}
  J=\frac12ru_0^2+\frac12s x_1^2,
  \qquad r>0,\quad s\geq0,
\end{equation}
find the optimal feedback and value.
\end{exercise}
\begin{proof}[Short solution]
The Bellman minimization is
\begin{equation}
  V_0(x)=\min_u\left\{\frac12ru^2+\frac12s(x+u)^2\right\}.
\end{equation}
Stationarity gives
\begin{equation}
  u^\star=-\frac{s}{r+s}x,
  \qquad
  V_0(x)=\frac12\frac{rs}{r+s}x^2.
\end{equation}
\end{proof}

\begin{exercise}
Starting from the short-interval Bellman identity
\cref{eq:hjb-short-interval-bellman-identity}, list the three first-order
expansions needed to derive HJB.
\end{exercise}
\begin{proof}[Short solution]
For constant control over a short interval,
\begin{align}
  x(t+h)&=x+h f(t,x,u)+o(h),\\
  \int_t^{t+h}\ell\,ds&=h\ell(t,x,u)+o(h),\\
  V(t+h,x(t+h))
  &=V+hV_t+h\nabla_xV\trans f+o(h).
\end{align}
Substitution, cancellation, division by $h$, and the limit $h\to0$ give HJB.
\end{proof}

\begin{exercise}
Suppose a differentiable function $W$ satisfies the HJB equation and terminal
condition. Show that $W(t,x)$ is a lower bound on the cost of every admissible
control from $(t,x)$.
\end{exercise}
\begin{proof}[Short solution]
HJB implies
$W_t+\nabla W\trans f+\ell\geq0$ for every admissible control. Along its
trajectory, $dW/dt=W_t+\nabla W\trans f$, so
$\ell\geq-dW/dt$. Integrating and using $W(t_f,x)=\Phi(x,t_f)$ gives
$J_{t,x}[u]\geq W(t,x)$.
\end{proof}

\begin{exercise}
For $\dot x=f_0(x)+G(x)u$ and
$\ell(x,u)=L(x)+u\trans Ru/2$, derive the unconstrained HJB feedback law.
What changes if the input is constrained?
\end{exercise}
\begin{proof}[Short solution]
The control-dependent Hamiltonian terms are
$u\trans Ru/2+\nabla V\trans Gu$. Setting their gradient to zero gives
\begin{equation}
  u^\star=-R^{-1}G\trans\nabla V.
\end{equation}
With constraints, this unconstrained value is replaced by the feasible input
that minimizes the same Hamiltonian; for a scalar interval it is clipped to
the interval.
\end{proof}

\begin{exercise}
For $\dot x=u$ and
\begin{equation}
  J=\frac12s x(T)^2+\frac12\int_t^T r u^2\,ds,
  \qquad r>0,\quad s>0,
\end{equation}
use $V(t,x)=p(t)x^2/2$ to find $p(t)$ and the optimal feedback.
\end{exercise}
\begin{proof}[Short solution]
HJB gives
\begin{equation}
  0=\frac12\dot p x^2+\min_u
  \left\{\frac12ru^2+p x u\right\}.
\end{equation}
Thus $u^\star=-px/r$ and $\dot p=p^2/r$, with $p(T)=s$. Solving backward,
\begin{equation}
  p(t)=\frac{rs}{r+s(T-t)},
  \qquad
  u^\star(t)=-\frac{s}{r+s(T-t)}x(t).
\end{equation}
\end{proof}

\begin{exercise}
Derive the discounted stationary HJB equation from the short-time identity
using $e^{-\rho h}=1-\rho h+o(h)$.
\end{exercise}
\begin{proof}[Short solution]
Expanding the immediate cost, next state, discounted factor, and value gives
\begin{equation}
  0
  =\min_u\left\{
    \ell(x,u)+\nabla V_\rho(x)\trans f(x,u)-\rho V_\rho(x)
  \right\}.
\end{equation}
Equivalently,
\begin{equation}
  \rho V_\rho(x)
  =\min_u\left\{\ell(x,u)+\nabla V_\rho(x)\trans f(x,u)\right\}.
\end{equation}
\end{proof}

\begin{exercise}
Verify that $V(x)=|x|$ solves the minimum-time HJB equation for
$\dot x=u$, $|u|\leq1$, at every $x\neq0$. State the optimal feedback.
\end{exercise}
\begin{proof}[Short solution]
For $x>0$, $V_x=1$ and
$\min_{|u|\leq1}(1+u)=0$ at $u=-1$. For $x<0$, $V_x=-1$ and
$\min_{|u|\leq1}(1-u)=0$ at $u=1$. Thus
$u^\star=-1$ for $x>0$ and $u^\star=1$ for $x<0$.
\end{proof}

\begin{exercise}
A uniform grid uses $51$ points per state coordinate. How many value samples
are required for state dimensions $n=2$, $n=4$, and $n=6$ at one time layer?
Why does this calculation matter for robot control?
\end{exercise}
\begin{proof}[Short solution]
The counts are
\begin{equation}
  51^2=2601,
  \qquad
  51^4=6{,}765{,}201,
  \qquad
  51^6=17{,}596{,}287{,}801.
\end{equation}
The exponential growth makes full grids impractical for robots whose states
contain many joint positions and velocities.
\end{proof}

\begin{exercise}
Assuming a smooth value function and an interior minimizing control, show how
the PMP costate equation follows from $\lambda=\nabla_xV$.
\end{exercise}
\begin{proof}[Short solution]
Differentiate HJB with respect to $x$. Because $H_u=0$ at the interior
minimizer and $H_p=f$,
\begin{equation}
  V_{xt}+H_x+V_{xx}f=0.
\end{equation}
Along the optimal state,
\begin{equation}
  \dot\lambda
  =V_{xt}+V_{xx}\dot x
  =V_{xt}+V_{xx}f
  =-H_x.
\end{equation}
\end{proof}

\begin{exercise}
Why can HJB produce a feedback law directly, while a single PMP solution
usually produces only one open-loop trajectory?
\end{exercise}
\begin{proof}[Short solution]
HJB computes $V(t,x)$ and its minimizing action over a domain of states, so
the control can be evaluated at whichever state is currently observed. A
single PMP solve computes state, costate, and control only along the extremal
associated with one initial condition.
\end{proof}

\paragraph{Chapter summary.}
Dynamic programming organizes optimal control around the value function, the
smallest future cost achievable from each time--state pair. The principle of
optimality states that every continuation of an optimal trajectory must be
optimal from the state at which that continuation begins. This principle
requires a sufficient state that summarizes all past information relevant to
future dynamics and costs. In discrete time it gives the Bellman recursion,
which combines one-stage cost with the value of the next state and computes
feedback by backward induction.

Applying the same recursion over a short continuous-time interval and taking
the limit gives the Hamilton--Jacobi--Bellman equation. HJB balances immediate
running cost against the rate at which the chosen dynamics change future
value. A smooth solution satisfying the terminal condition provides an
optimality certificate: every admissible control has cost at least equal to
the value, and a feedback that attains the Hamiltonian minimum achieves that
bound. For control-affine dynamics with quadratic effort, the optimal action
is proportional to the negative value gradient. In the linear--quadratic
case, a quadratic value ansatz reduces HJB to the differential Riccati equation
and recovers the continuous-time LQR feedback law. Applying the discrete-time
Bellman recursion to linear dynamics and quadratic costs similarly yields the
backward Riccati difference equation. Its time-invariant fixed point is the
discrete-time algebraic Riccati equation.

Autonomous infinite-horizon problems lead to stationary HJB equations, while
discounting adds the term $\rho V$. Value functions need not be smooth;
minimum-time motion and competing routes can create corners where the active
optimal strategy changes. Grid-based Bellman updates can approximate both
smooth and nonsmooth values, but their storage and computation grow
exponentially with state dimension. This curse of dimensionality motivates
Riccati methods, local trajectory approximations, model predictive control,
and parameterized value approximations.

Finally, HJB and Pontryagin's principle are linked by
$\lambda=\nabla_xV$ along a smooth optimal trajectory. PMP propagates value
sensitivity along one candidate motion, whereas HJB constructs the full field
of optimal costs and feedback decisions over a state domain. The next chapter
uses local dynamic-programming approximations to optimize nonlinear
trajectories without solving a global HJB equation.

\part{Trajectory Optimization and Model Predictive Control}
\chapter{Discrete-Time Trajectory Optimization}
\label{chap:discrete-time-trajectory-optimization}
\chapteroverview{Nonlinear finite-horizon control problems, single and
multiple shooting, direct transcription, adjoint gradients, iterative
linear--quadratic methods, differential dynamic programming, contact-aware
robotics examples, constraints, and practical implementation.}

The preceding chapters developed several complementary views of optimal
control. Pontryagin's principle gives necessary conditions along a candidate
trajectory. Linear--quadratic control turns a special class of problems into
a Riccati recursion and a feedback law. Dynamic programming characterizes the
globally optimal value function through Bellman recursion or the HJB equation.
Each viewpoint is powerful, but a practical gap remains. Robotic systems are
usually nonlinear, their tasks have finite horizons, and their states and
inputs are constrained. For such problems, the global Riccati structure of
LQR is absent, while a grid-based solution of HJB is generally impractical.

Trajectory optimization addresses this gap by concentrating computation near
one motion. Rather than computing the optimal action from every possible
state, it searches for a state--control sequence that performs a specified
task from a specified initial condition. Examples include swinging up a
pendulum, moving a manipulator around obstacles, landing a vehicle, and
planning a dynamically feasible walking motion. The resulting plan may be
used directly, tracked by a separate controller, or repeatedly recomputed in
model predictive control.

The apparent simplicity of this idea hides an important difficulty. A control
chosen early in the horizon changes every later state, so the decision
variables are strongly coupled through time. Unstable dynamics can amplify a
small perturbation, constraints can make an otherwise attractive motion
infeasible, and a nonlinear problem may contain several qualitatively
different local solutions. A useful algorithm must therefore exploit the
temporal structure rather than treating the problem as a generic large-scale
optimization.

This chapter develops two complementary ways to solve a discretized control
problem. General direct methods formulate a finite-dimensional nonlinear
program and exploit its sparse constraints. Structured shooting methods,
especially iterative linear--quadratic regulation (iLQR), repeatedly solve a
local time-varying LQR approximation. Both are direct methods; they differ in
how they represent the trajectory and solve the local optimization problem.
iLQR is particularly revealing because it combines ideas from the previous
chapters: it uses local linearization, quadratic value approximations, a
Riccati-like backward pass, and a nonlinear forward rollout. Standard
references include
\cite{bertsekas2017dynamic,kirk2004optimal,lewis2012optimal}. The next chapter
places these methods inside a feedback loop through \emph{model predictive
control} (MPC), which repeatedly solves a finite-horizon trajectory
optimization problem as new state measurements become available.

\section{Direct Versus Indirect Numerical Methods}

The central distinction is the order in which optimality conditions and
discretization are introduced:
\begin{equation}
  \begin{aligned}
  \text{indirect:}\quad
    &\text{continuous control problem}
      \longrightarrow \text{PMP conditions}\\
    &\hspace{31mm}\longrightarrow
      \text{boundary-value or root solve},\\[2pt]
  \text{direct:}\quad
    &\text{continuous control problem}
      \longrightarrow \text{discretized optimization problem}\\
    &\hspace{31mm}\longrightarrow
      \text{numerical minimization}.
  \end{aligned}
  \label{eq:direct-indirect-workflows}
\end{equation}
Both routes ultimately use numerical computation, but they solve different
finite-dimensional problems.

Chapter~6 introduced \emph{indirect methods}. They first apply Pontryagin's
minimum principle to obtain the state equation, costate equation, Hamiltonian
minimization condition, and boundary conditions. Only then are these necessary
conditions solved numerically. The unknowns may include an initial costate,
endpoint multipliers, and switching times. A converged solution is an
extremal satisfying necessary conditions; additional checks are needed to
determine whether it is a local minimum.

The methods in this chapter are primarily \emph{direct}. They first choose a
finite time grid and discretize the original dynamics, objective, and
constraints. The resulting finite-dimensional optimization problem is solved
for sampled states and controls without explicitly deriving the
continuous-time costate equations. Necessary conditions still appear
internally---for example, nonlinear-programming multipliers are discrete
analogues of costates---but the user formulates the original control problem rather than
its PMP boundary-value equations.

\begin{table}[t]
  \centering
  \caption{Indirect and direct numerical optimal control.}
  \label{tab:direct-indirect-comparison}
  \begin{tabular}{@{}>{\raggedright\arraybackslash}p{0.20\textwidth}
                    >{\raggedright\arraybackslash}p{0.32\textwidth}
                    >{\raggedright\arraybackslash}p{0.32\textwidth}@{}}
    \toprule
    Aspect & Indirect method & Direct method \\
    \midrule
    Formulated from
      & PMP necessary conditions
      & original dynamics, cost, and constraints \\
    Typical unknowns
      & costate boundary values, multipliers, switching times
      & sampled controls and, in transcription, sampled states \\
    Numerical task
      & root finding or a boundary-value solve
      & nonlinear minimization \\
    Path constraints
      & require extra multipliers, arcs, and switching conditions
      & imposed directly as finite-dimensional constraints \\
    \bottomrule
  \end{tabular}
\end{table}

The distinction between direct and indirect methods specifies \emph{which
equations} are solved, but it does not uniquely determine \emph{how} they are
represented numerically. Either route can use shooting, multiple shooting,
transcription or collocation, parameterized trajectories, or grid-based
discretizations. For example, shooting integrates the relevant equations and
adjusts a finite set of unknowns, whereas transcription retains sampled
variables and enforces the equations through algebraic constraints. A
parameterized method instead represents a control or trajectory with a finite
set of coefficients. In an indirect method these techniques are applied to
the PMP state--costate conditions; in a direct method they are applied to the
original dynamics, costs, and constraints.

No representation is uniformly best. Shooting can be compact and accurate,
but may be sensitive over long or unstable trajectories. Transcription and
grid-based methods introduce more variables but expose sparsity and handle
path constraints naturally. Low-dimensional parameterizations reduce the
number of unknowns, but can exclude useful trajectories if the chosen basis is
too restrictive. The appropriate choice depends on the dynamics, horizon,
constraints, required accuracy, derivative information, and available
initialization. This chapter emphasizes direct shooting, direct transcription,
and structured local methods while noting that other combinations are often
equally valid.

\section{Finite-Horizon Trajectory Problems}

Trajectory optimization begins by replacing a continuous motion over an
interval by a finite sequence of states and controls. This replacement is not
merely a numerical convenience: it determines which quantities the optimizer
may change and how dynamic consistency is enforced. We therefore begin with
the construction of the discrete problem before discussing algorithms for
solving it.

\subsection{Dynamics, Costs, and Constraints}

Suppose a continuous-time model is written as
\begin{equation}
  \dot x(t)=f_c\bigl(t,x(t),u(t)\bigr).
  \label{eq:trajectory-continuous-dynamics}
\end{equation}
Choose grid times $t_0<\cdots<t_N$ and hold the control constant, or represent
it by a low-order polynomial, on each interval. Numerically integrating
\cref{eq:trajectory-continuous-dynamics} from $t_k$ to $t_{k+1}$ defines a
transition map
\begin{equation}
  x_{k+1}=f_k(x_k,u_k).
  \label{eq:trajectory-flow-map-definition}
\end{equation}
For example, with step length $\Delta t_k=t_{k+1}-t_k$, forward Euler gives
\begin{equation}
  f_k(x_k,u_k)
  =x_k+\Delta t_k f_c(t_k,x_k,u_k),
  \label{eq:trajectory-forward-euler-map}
\end{equation}
whereas a higher-order Runge--Kutta method can give a more accurate map by
evaluating the vector field at intermediate points. Once the integration rule is fixed, the
trajectory problem has the discrete-time form
\begin{equation}
  x_{k+1}=f_k(x_k,u_k),
  \qquad k=0,\ldots,N-1,
  \qquad x_0=x_{\mathrm{init}},
  \label{eq:trajectory-optimization-dynamics}
\end{equation}
where $x_k\in\R^n$, $u_k\in\R^m$, and the initial state is given. A standard
finite-horizon objective is
\begin{equation}
  J(X,U)
  =\Phi(x_N)+\sum_{k=0}^{N-1}\ell_k(x_k,u_k),
  \label{eq:trajectory-optimization-cost}
\end{equation}
where
\begin{equation}
  X=(x_0,\ldots,x_N),
  \qquad
  U=(u_0,\ldots,u_{N-1}).
\end{equation}
The running costs can penalize tracking error, control effort, proximity to
obstacles, or other undesirable behavior. The terminal cost measures the
quality of the final state.

The discrete stage cost may itself approximate a continuous integral. If the
continuous running cost is $L(t,x,u)$, a left-endpoint rule gives
\begin{equation}
  \ell_k(x_k,u_k)=\Delta t_k L(t_k,x_k,u_k),
  \label{eq:trajectory-stage-cost-quadrature}
\end{equation}
and higher-order quadrature may be paired with the integration scheme. Thus
both the dynamics and the objective inherit discretization error. Refining the
grid should change the computed motion less and less; otherwise the apparent
optimum may be an artifact of the chosen time step.

State and input restrictions may be written as
\begin{align}
  g_k(x_k,u_k)&\leq0,
  &h_k(x_k,u_k)&=0,
  &&k=0,\ldots,N-1,
  \label{eq:trajectory-path-constraints}\\
  g_N(x_N)&\leq0,
  &h_N(x_N)&=0.
  \label{eq:trajectory-terminal-constraints}
\end{align}
These relations can encode actuator limits, collision avoidance, contact
conditions, or a required terminal configuration. A pair $(X,U)$ is
\emph{dynamically feasible} when it satisfies
\cref{eq:trajectory-optimization-dynamics}; it is fully feasible when it also
satisfies the path and terminal constraints. These constraints are imposed at
the grid points; satisfying them there does not by itself guarantee safety
between samples. Continuous-time validation or grid refinement may therefore
be needed, especially for collision avoidance.

A common tracking objective has the quadratic form
\begin{align}
  \ell_k(x,u)
  &={}
  \frac12(x-x_k^{\mathrm{ref}})\trans Q_k
             (x-x_k^{\mathrm{ref}})
  +\frac12(u-u_k^{\mathrm{ref}})\trans R_k
             (u-u_k^{\mathrm{ref}}),
  \label{eq:trajectory-quadratic-stage-cost}\\
  \Phi(x)
  &={}
  \frac12(x-x_N^{\mathrm{ref}})\trans Q_N
             (x-x_N^{\mathrm{ref}}),
  \label{eq:trajectory-quadratic-terminal-cost}
\end{align}
with $Q_k\succeq0$, $R_k\succ0$, and $Q_N\succeq0$. Nonlinear tasks often
require nonquadratic terms, but quadratic local models will still play a
central role.

Trajectory optimization usually produces an open-loop nominal sequence
$(\bar X,\bar U)$. Later in this chapter, the iLQR and DDP sections augment
this sequence with local feedforward and feedback corrections. The idea is
closely related to the reference-trajectory tracking construction in
Chapter~7: the nonlinear dynamics are linearized along a nominal motion, and
a time-varying feedback law responds to deviations from that motion. Here,
however, the nominal trajectory and its local corrections are improved
together as part of the trajectory-optimization iteration. The resulting
feedback is local to the current nominal trajectory rather than defined over
the entire state space.

It is useful to distinguish three objects that will recur throughout the
chapter. The sequence $(\bar X,\bar U)$ is the current \emph{nominal
trajectory}. A perturbation $(\delta X,\delta U)$ describes a candidate local
change. The updated sequence is accepted only after it is evaluated using the
original nonlinear dynamics, costs, and constraints. This separation between
nominal model, local approximation, and nonlinear trial motion is central to shooting,
iLQR, and DDP.

The following examples illustrate the common ingredients before the numerical
methods are developed in detail.

\begin{example}[Pendulum swing-up]
Consider a pendulum with angle $\theta$, angular velocity $\omega$, torque
$u$, point mass $m$ at distance $l$ from the pivot, viscous damping $b$, and
sampling interval $h$.
Forward Euler discretization gives
\begin{align}
  \theta_{k+1}
  &=\theta_k+h\omega_k,
  \label{eq:pendulum-discrete-angle}\\
  \omega_{k+1}
  &=\omega_k+h\left(
      \frac{u_k-b\omega_k}{ml^2}
      -\frac{g}{l}\sin\theta_k
    \right).
  \label{eq:pendulum-discrete-velocity}
\end{align}
Let $\theta=0$ denote the downward equilibrium and $\theta=\pi$ the upright
equilibrium. One smooth swing-up cost is
\begin{equation}
  \ell(x,u)
  =q_\theta(1+\cos\theta)
   +\frac12q_\omega\omega^2
   +\frac12ru^2.
  \label{eq:pendulum-swing-up-cost}
\end{equation}
The angle term is smallest at the upright configuration and respects the
periodicity of the angle. Take $q_\theta,q_\omega,r>0$. With limited torque,
the optimizer may need to discover that temporarily moving
away from the target can build the energy needed to reach it. This makes the
problem nonlinear and nonconvex even though the effort penalty is quadratic.
\end{example}

\begin{example}[Manipulator reaching around an obstacle]
Consider a robot arm with configuration $q\in\R^{n_q}$, velocity $v$, and
joint torque $u$. The continuous dynamics from
\cref{eq:robot-manipulator-dynamics} can be discretized to define
\begin{equation}
  x_{k+1}=f_h(x_k,u_k),
  \qquad x_k=(q_k,v_k),
  \label{eq:manipulator-discrete-trajectory-dynamics}
\end{equation}
where $f_h$ denotes one numerical integration step of duration $h$. Let
$p(q)$ be the end-effector position and let $d_j(q)$ be the signed distance
between robot body $j$ and the nearest obstacle. A representative problem is
\begin{equation}
  \begin{aligned}
  \minimize_{X,U}\quad
    &\frac12\norm{p(q_N)-p_{\mathrm{goal}}}_{Q_f}^2
     +\sum_{k=0}^{N-1}
      \left(
        \frac12\norm{p(q_k)-p_k^{\mathrm{ref}}}_{Q_p}^2
        +\frac12\norm{v_k}_{Q_v}^2
        +\frac12\norm{u_k}_{R}^2
      \right)\\
  \subjectto\quad
    &x_0=x_{\mathrm{init}},\\
    &x_{k+1}=f_h(x_k,u_k),
      \quad k=0,\ldots,N-1,\\
    &u_{\min}\leq u_k\leq u_{\max},
      \quad k=0,\ldots,N-1,\\
    &d_j(q_k)\geq d_{\mathrm{safe}}
      \quad\text{for every body }j,\quad k=0,\ldots,N.
  \end{aligned}
  \label{eq:manipulator-obstacle-trajectory-problem}
\end{equation}
The end-effector objective alone does not determine the arm posture, and an
obstacle may create several homotopy classes: the elbow can pass on different
sides while reaching the same target. Direct transcription represents the
distance constraints explicitly. In a shooting or iLQR implementation, one
often begins with smooth obstacle penalties and then increases their weight,
or uses the resulting trajectory to initialize a constrained solver. This
example illustrates the typical role of trajectory optimization in robotics:
it coordinates the whole body over time rather than selecting a sequence of
independent inverse-kinematics solutions.
\end{example}

\subsection{Why the Problem Is Difficult}

If the states at different stages were independent, each stage could be
optimized separately. The dynamics prevent this decomposition: changing
$u_k$ changes $x_{k+1}$, which changes $x_{k+2}$, and so on through the
terminal state. The quality of an action therefore depends not only on its
immediate cost, but also on every future state that it influences.

This temporal coupling becomes explicit when the dynamics are differentiated.
Along a nominal trajectory, define
\begin{equation}
  A_k=f_{x,k}(\bar x_k,\bar u_k),
  \qquad
  B_k=f_{u,k}(\bar x_k,\bar u_k).
\end{equation}
If only $u_i$ is perturbed and the initial state is fixed, the resulting state
perturbations satisfy, to first order,
\begin{align}
  \delta x_{i+1}&=B_i\delta u_i,\\
  \delta x_{i+2}&=A_{i+1}B_i\delta u_i,\\
  \delta x_j&=A_{j-1}\cdots A_{i+1}B_i\delta u_i,
  \qquad j>i.
  \label{eq:trajectory-temporal-coupling}
\end{align}
The product is the identity when $j=i+1$. Thus a single control perturbation
propagates forward through a product of dynamics Jacobians and enters every
subsequent stage cost. Applying the chain
rule to \cref{eq:trajectory-optimization-cost} gives
\begin{equation}
  \nabla_{u_i}\widehat J
  =\ell_{u,i}
   +\sum_{j=i+1}^{N-1}
      \left(\frac{\partial x_j}{\partial u_i}\right)\trans
      \ell_{x,j}
   +\left(\frac{\partial x_N}{\partial u_i}\right)\trans
      \Phi_x.
  \label{eq:trajectory-control-chain-rule}
\end{equation}
Here $\widehat J$ denotes the cost after eliminating the states through the
dynamics, and cost gradients are column vectors. Forming these long
sensitivity products separately for every pair $(i,j)$
would be expensive and can be numerically ill-conditioned. Their repeated
structure, however, makes it possible to collect future effects through an
efficient backward adjoint or value-function recursion.

Nonlinearity creates a second difficulty: the optimization landscape may
contain several qualitatively different local solutions. In pendulum
swing-up, for example, one initialization may lead to a clockwise motion and
another to a counterclockwise motion, while a symmetric or poorly chosen
initialization may remain near the downward equilibrium. Discretization, cost
weights, constraint representations, and numerical scaling can all change
which solution the algorithm finds.

These observations motivate two algorithmic principles. First, a method
should exploit the chain structure of the dynamics rather than treat the
trajectory as an unstructured collection of variables. Second, it should test
each proposed step on the original nonlinear problem before accepting it.
Most nonlinear solvers obtain a step from a local linear or quadratic model
around the current trajectory. Because this model neglects higher-order
terms, it is reliable only within a sufficiently small neighborhood. A large
step may reduce the approximate objective while increasing the true cost,
violating constraints, or entering a region where the approximation is poor.
Line searches, trust regions, and predicted-versus-actual improvement tests
provide the required safeguard. Direct transcription and iLQR implement these
two principles in different ways.

\section{Direct Optimization Methods}

\subsection{Single Shooting}

In \emph{single shooting}, the controls are the optimization variables. Given
$U$, the states are generated by rolling the dynamics forward from the fixed
initial condition. This defines a reduced objective
\begin{equation}
  \widehat J(U)=J\bigl(X(U),U\bigr),
  \label{eq:single-shooting-reduced-objective}
\end{equation}
and the dynamics are satisfied automatically at every iterate.

The computational process is therefore simple to state. Starting from a trial
control sequence, simulate forward to obtain $X(U)$, evaluate the cost, compute
a search direction in control space, and repeat. The main question is how to
differentiate $\widehat J$ without separately propagating the influence of
every $u_k$ to every later state.

To see how an early control affects later states, define the Jacobians along a
nominal trajectory by
\begin{equation}
  A_k=\frac{\partial f_k}{\partial x}(\bar x_k,\bar u_k),
  \qquad
  B_k=\frac{\partial f_k}{\partial u}(\bar x_k,\bar u_k).
  \label{eq:trajectory-dynamics-jacobians}
\end{equation}
A small perturbation satisfies, to first order,
\begin{equation}
  \delta x_{k+1}=A_k\delta x_k+B_k\delta u_k.
  \label{eq:single-shooting-sensitivity-recursion}
\end{equation}
For $j>k$,
\begin{equation}
  \frac{\partial x_j}{\partial u_k}
  =A_{j-1}A_{j-2}\cdots A_{k+1}B_k.
  \label{eq:single-shooting-state-sensitivity}
\end{equation}
The product is again the identity for $j=k+1$. Products of many Jacobians may
become very large or very small. This is the
optimization analogue of exploding or vanishing sensitivities and helps
explain why long-horizon shooting can be poorly conditioned.

The gradient can be computed without explicitly forming every product in
\cref{eq:single-shooting-state-sensitivity}. Introduce the backward adjoint
recursion, with all derivatives evaluated along the rollout under consideration,
\begin{align}
  \lambda_N&=\Phi_x(x_N),
  \label{eq:shooting-adjoint-terminal}\\
  \lambda_k
  &=\ell_{x,k}(x_k,u_k)+A_k\trans\lambda_{k+1}.
  \label{eq:shooting-adjoint-recursion}
\end{align}
Then
\begin{equation}
  \nabla_{u_k}\widehat J
  =\ell_{u,k}(x_k,u_k)+B_k\trans\lambda_{k+1}.
  \label{eq:shooting-control-gradient}
\end{equation}
This is the discrete-time counterpart of the costate calculation in
Pontryagin's principle. One forward rollout and one backward pass compute the
gradient for every control.

To derive this recursion, take the first variation of the cost:
\begin{equation}
  \delta J
  =\Phi_x(x_N)\trans\delta x_N
   +\sum_{k=0}^{N-1}
     \left(
       \ell_{x,k}\trans\delta x_k
       +\ell_{u,k}\trans\delta u_k
     \right).
  \label{eq:shooting-cost-first-variation}
\end{equation}
Set $\lambda_N=\Phi_x(x_N)$. The terminal contribution can be moved one stage
backward using
\begin{equation}
  \lambda_N\trans\delta x_N
  =\lambda_N\trans A_{N-1}\delta x_{N-1}
   +\lambda_N\trans B_{N-1}\delta u_{N-1}.
\end{equation}
After this substitution, all terms involving the final decision stage can be
grouped as
\begin{align}
  &\ell_{x,N-1}\trans\delta x_{N-1}
   +\ell_{u,N-1}\trans\delta u_{N-1}
   +\lambda_N\trans\delta x_N\nonumber\\
  &\quad={}
  \left(\ell_{x,N-1}+A_{N-1}\trans\lambda_N\right)\trans
    \delta x_{N-1}\nonumber\\
  &\qquad{}
  +\left(\ell_{u,N-1}+B_{N-1}\trans\lambda_N\right)\trans
    \delta u_{N-1}.
  \label{eq:shooting-last-stage-variation}
\end{align}
The coefficient of $\delta x_{N-1}$ contains both the immediate sensitivity
of the stage cost and the sensitivity of the terminal cost transmitted
through the dynamics. Defining
\begin{equation}
  \lambda_{N-1}
  =\ell_{x,N-1}+A_{N-1}\trans\lambda_N
\end{equation}
compresses these two effects into a single vector.

The same argument applies at an arbitrary stage $k$. Suppose all costs after
stage $k$ have already been summarized by
$\lambda_{k+1}\trans\delta x_{k+1}$. Substituting the linearized dynamics gives
\begin{align}
  &\ell_{x,k}\trans\delta x_k
   +\ell_{u,k}\trans\delta u_k
   +\lambda_{k+1}\trans\delta x_{k+1}\nonumber\\
  &\quad={}
  \left(\ell_{x,k}+A_k\trans\lambda_{k+1}\right)\trans\delta x_k
  +\left(\ell_{u,k}+B_k\trans\lambda_{k+1}\right)\trans\delta u_k.
  \label{eq:shooting-one-stage-elimination}
\end{align}
The first coefficient defines $\lambda_k$; the second is the gradient of the
remaining objective with respect to $u_k$. Proceeding from $k=N-1$ down to
$k=0$ eliminates the state perturbations one at a time. The only remaining
state term is $\lambda_0\trans\delta x_0$, which vanishes because the initial
state is fixed and hence $\delta x_0=0$. Therefore
\begin{equation}
  \delta J
  =\sum_{k=0}^{N-1}
    \left(\ell_{u,k}+B_k\trans\lambda_{k+1}\right)\trans
    \delta u_k,
  \label{eq:shooting-reduced-first-variation}
\end{equation}
which proves \cref{eq:shooting-control-gradient}. The adjoint $\lambda_k$ can
therefore be interpreted as the sensitivity of all cost from stage $k$
onward to a perturbation of $x_k$. Computationally, this backward sweep is
reverse-mode differentiation through the forward dynamics rollout.

Once the gradient is available, even a basic update has the form
\begin{equation}
  U^{+}=U-\alpha\nabla\widehat J(U),
  \label{eq:shooting-gradient-update}
\end{equation}
with a line search for $\alpha$. Quasi-Newton methods replace the identity in
this update by an approximation of the inverse reduced Hessian. The important
point is that the adjoint computes the gradient in $O(N)$ stages rather than
forming the $O(N^2)$ collection of state--control sensitivity blocks.

Single shooting is compact and satisfies the discrete dynamics at every
iterate. One adjoint sweep supplies the gradient for the entire control
sequence, but does not remove the long-horizon sensitivity described in
\cref{eq:single-shooting-state-sensitivity}. Unstable dynamics can amplify
early control changes, while strongly contracting dynamics can suppress their
effect on distant objectives. The resulting reduced problem may require
coordinated control changes and very small line-search steps. Single shooting
therefore works best when the horizon is moderate and the forward rollout
remains reasonably well-conditioned.

\subsection{Direct Transcription and Multiple Shooting}

Single shooting represents the trajectory through the control sequence alone:
the states are generated by one forward rollout and are therefore eliminated
from the optimization variables. Direct transcription takes the opposite
view. It retains every sampled state, together with every control, and asks the
nonlinear-programming solver to enforce dynamic consistency. These two
formulations provide useful endpoints for understanding the tradeoff between
problem size and numerical conditioning. Multiple shooting, introduced after
their comparison, lies between them.

\paragraph{Direct transcription.}
In \emph{direct transcription}, both states and controls are optimization
variables. The dynamics become equality constraints:
\begin{equation}
  d_k(X,U)
  =x_{k+1}-f_k(x_k,u_k)=0,
  \qquad k=0,\ldots,N-1.
  \label{eq:trajectory-dynamic-defects}
\end{equation}
The resulting nonlinear program is
\begin{equation}
  \begin{aligned}
  \minimize_{X,U}\quad
    &\Phi(x_N)+\sum_{k=0}^{N-1}\ell_k(x_k,u_k)\\
  \subjectto\quad
    &x_0=x_{\mathrm{init}},\\
    &d_k(X,U)=0,\quad k=0,\ldots,N-1,\\
    &g_k(x_k,u_k)\leq0,\quad h_k(x_k,u_k)=0,
      \quad k=0,\ldots,N-1,\\
    &g_N(x_N)\leq0,\quad h_N(x_N)=0.
  \end{aligned}
  \label{eq:direct-transcription-nlp}
\end{equation}
There are $(N+1)n+Nm$ scalar state and control variables before eliminating
the fixed initial state, and $Nn$ scalar dynamic-defect equations. Unlike a
shooting rollout, an intermediate iterate may have $d_k\neq0$ and therefore
need not represent a motion that the system can execute. The solver reduces
the objective and the defect residuals together; a usable trajectory is
recovered only when the defects and the remaining constraint violations are
within tolerance.

\paragraph{Sequential quadratic programming.}
A popular way to solve \cref{eq:direct-transcription-nlp} is sequential
quadratic programming (SQP). Stack the retained variables in temporal order,
\begin{equation}
  z=(x_0,u_0,x_1,u_1,\ldots,x_{N-1},u_{N-1},x_N),
\end{equation}
and write the transcription compactly as
\begin{equation}
  \minimize_z\quad J(z)
  \qquad
  \subjectto\quad c(z)=0,\quad g(z)\leq0.
  \label{eq:trajectory-stacked-objective}
\end{equation}
Here $J(z)=\Phi(x_N)+\sum_{k=0}^{N-1}\ell_k(x_k,u_k)$,
$c(z)$ collects the initial-state condition, dynamic defects, and other
equalities, and $g(z)$ collects the path and terminal inequalities.

The general SQP construction, including its QP subproblem, multiplier updates,
curvature models, and globalization strategies, is developed in
\cref{sec:general-sqp}. Applied to
\cref{eq:trajectory-stacked-objective}, each SQP iteration builds a quadratic
model using Lagrangian curvature, linearizes the dynamic defects and other
constraints, and solves
the resulting structured QP for a trajectory correction. The essential issue
specific to trajectory optimization is not the SQP derivation itself, but the
temporal sparsity of this QP and its KKT system.

\paragraph{Why sparsity makes SQP practical.}
Retaining every state creates a large problem, but its derivatives are local
in time. Dynamic defect $d_k$ depends only on $(x_k,u_k,x_{k+1})$, so
\begin{equation}
  \frac{\partial d_k}{\partial(x_k,u_k,x_{k+1})}
  =\begin{bmatrix}-A_k&-B_k&I\end{bmatrix}.
  \label{eq:trajectory-linearized-defect}
\end{equation}
Stacking the defect rows produces the block-banded Jacobian
\begin{equation}
  \begin{bmatrix}
    -A_0&-B_0&I&&&\\
       &&-A_1&-B_1&I&\\
       &&&\ddots&\ddots&\ddots
  \end{bmatrix}.
  \label{eq:trajectory-defect-jacobian-structure}
\end{equation}
Stage costs and path constraints have a similar local structure. The KKT
systems inside the quadratic-programming solver are therefore sparse and
nearly block-banded rather than dense. Sparse factorization, Riccati-style
recursions, or condensing exploit this chain structure without coupling every
time step to every other time step. For fixed state and control dimensions,
the computational work and memory can often grow approximately linearly with
the horizon, although the exact cost depends on inequality handling and
factorization fill-in. This sparsity is what makes direct transcription
practical despite its larger number of explicit variables.

\paragraph{Collocation.}
Direct transcription requires a finite-dimensional equation connecting
neighboring states. A simulation step such as forward Euler is one option;
collocation derives this equation by approximating the state within each
interval and enforcing the differential equation at selected points. Starting
from the exact integral relation
\begin{equation}
  x_{k+1}-x_k
  =\int_{t_k}^{t_{k+1}}f_c(t,x(t),u(t))\,dt,
  \label{eq:trajectory-exact-integral-dynamics}
\end{equation}
and applying trapezoidal quadrature with a zero-order-held control gives
\begin{equation}
  x_{k+1}-x_k
  -\frac{\Delta t_k}{2}\left[
    f_c(t_k,x_k,u_k)
    +f_c(t_{k+1},x_{k+1},u_k)
  \right]=0.
  \label{eq:trajectory-trapezoidal-collocation}
\end{equation}
Unlike a one-sided Euler defect, this equation couples both endpoints and has
second-order integration accuracy under standard smoothness assumptions.
Because $x_{k+1}$ appears inside $f_c$, the equation is implicit and is
enforced by the nonlinear-programming solver rather than by an explicit
forward rollout. Higher-order Hermite--Simpson or Runge--Kutta collocation
schemes introduce internal stage variables and enforce additional residuals.
They provide more accurate continuous-time approximations per interval, at
the cost of more variables and denser local blocks.

\paragraph{Comparison with single shooting.}
When they use the same transition map and cost discretization, single
shooting and direct transcription solve the same discretized control problem,
but expose different variables to the optimizer. After fixing
$x_0$, single shooting optimizes only the $Nm$ control variables and obtains
all states from the rollout map $X=X(U)$. Direct transcription instead retains
$Nn+Nm$ state and control variables and imposes $Nn$ dynamic-defect equations.
The additional variables do not add physical degrees of freedom: once all
defects vanish, the states are again determined by the initial condition and
the controls. They change how the numerical solver reaches that feasible
trajectory.

The compactness of single shooting is attractive. Every iterate satisfies the
discrete dynamics exactly, and evaluating the objective requires only a
forward simulation. Its main weakness is that an early control can influence
a distant state only through a long product of dynamics Jacobians. Although an
adjoint pass computes the gradient with respect to the entire control sequence
efficiently, it does not remove this sensitivity. Long or unstable rollouts
can therefore produce an ill-conditioned reduced problem, and state or path
constraints are imposed only through the states generated by the rollout.

Direct transcription replaces these long-range dependencies by local sparse
constraints. Each defect couples only $(x_k,u_k,x_{k+1})$, so the optimizer
can change a distant state directly and then coordinate neighboring changes to
restore dynamic consistency. This is often advantageous for long horizons,
unstable dynamics, contact events, and problems with many state constraints.
The cost is a larger constrained problem: intermediate iterates may be
dynamically infeasible, defect residuals must be monitored, and efficient
solution depends on sparse KKT or SQP linear algebra. Thus the practical
choice is not between an exact and an approximate dynamics model. It is a
choice between eliminating the dynamics and retaining them as structured
constraints.

\paragraph{Multiple shooting as a compromise.}
Multiple shooting retains states only at selected segment boundaries. Choose
\begin{equation}
  0=i_0<i_1<\cdots<i_S=N
\end{equation}
and introduce $z_s\approx x_{i_s}$, with $z_0=x_{\mathrm{init}}$. Starting
from $z_s$, simulate the original dynamics over segment $s$ using its controls
$U_s=(u_{i_s},\ldots,u_{i_{s+1}-1})$. If
$\Psi_s(z_s,U_s)$ denotes the simulated endpoint, neighboring segments are
joined by the continuity constraints
\begin{equation}
  z_{s+1}-\Psi_s(z_s,U_s)=0,
  \qquad s=0,\ldots,S-1.
  \label{eq:multiple-shooting-continuity}
\end{equation}
The method therefore preserves dynamically feasible rollouts inside each
segment while permitting temporary gaps between segments. Linearizing a
continuity constraint gives
\begin{equation}
  \delta z_{s+1}-F_s\delta z_s-G_s\delta U_s=-c_s,
  \qquad
  F_s=\frac{\partial\Psi_s}{\partial z_s},
  \quad
  G_s=\frac{\partial\Psi_s}{\partial U_s},
  \label{eq:multiple-shooting-linearized-continuity}
\end{equation}
where $c_s=z_{s+1}-\Psi_s(z_s,U_s)$ is the current gap. The sensitivity
$F_s$ contains Jacobian products only across one segment rather than the full
horizon. Longer segments use fewer variables and approach single shooting;
shorter segments reduce the sensitivity length but introduce more boundary
states and continuity constraints. With one segment, the formulation reduces
to single shooting after eliminating its endpoint state. With one time step
per segment, it becomes essentially direct transcription. Segment length can
therefore be chosen to balance compactness against conditioning.

\subsection{Feasibility, Initialization, and Scaling}

A dynamically feasible initial guess is easy to obtain for shooting: choose a
control sequence and simulate. Its resulting motion may nevertheless be far
from the task or violate state constraints. Transcription permits a more
task-informed state guess, such as interpolating between initial and desired
configurations, but this guess generally has nonzero defects.

Neither type of initialization is universally superior. A useful strategy is
often to solve an easier problem first and gradually introduce difficulty.
One may shorten the horizon, relax terminal requirements, reduce obstacle
penalties, or begin with a nearby target. This practice is called
\emph{continuation} or \emph{homotopy}.

Scaling is equally important. If position errors are measured in millimeters,
angles in radians, forces in newtons, and time in seconds, raw derivatives may
have very different magnitudes. Nondimensionalization or explicit variable
and constraint scaling can substantially improve conditioning. Cost weights
should express the desired tradeoff after this scaling rather than compensate
accidentally for inconsistent units.

For example, introduce characteristic scales $S_x$ and $S_u$ and normalized
variables
\begin{equation}
  \widetilde x=S_x^{-1}x,
  \qquad
  \widetilde u=S_u^{-1}u.
  \label{eq:trajectory-variable-scaling}
\end{equation}
The normalized dynamics have Jacobians
\begin{equation}
  \widetilde A_k=S_x^{-1}A_kS_x,
  \qquad
  \widetilde B_k=S_x^{-1}B_kS_u.
  \label{eq:trajectory-scaled-jacobians}
\end{equation}
Choosing scales so typical normalized variables are of order one helps prevent
the linear systems inside Newton, SQP, and Riccati-type steps from being dominated
by units rather than by the task itself.

\begin{table}[t]
  \centering
  \caption{Common trajectory-optimization formulations.}
  \label{tab:trajectory-method-comparison}
  \begin{tabular}{@{}>{\raggedright\arraybackslash}p{0.15\textwidth}
                    >{\raggedright\arraybackslash}p{0.21\textwidth}
                    >{\raggedright\arraybackslash}p{0.25\textwidth}
                    >{\raggedright\arraybackslash}p{0.25\textwidth}@{}}
    \toprule
    Method & Main variables & Dynamics during iterations & Typical feature \\
    \midrule
    Single shooting & controls & exactly satisfied & compact but sensitive \\
    Multiple shooting & controls and segment states & segmentwise satisfied & balanced conditioning \\
    Direct transcription & states and controls & enforced as constraints & sparse and constraint-friendly \\
    iLQR/DDP & nominal controls and local policy & satisfied by rollout & fast structured local solve \\
    \bottomrule
  \end{tabular}
\end{table}

\section{Iterative Linear--Quadratic Methods}

The formulations above can be handled by generic nonlinear-programming
algorithms. Iterative linear--quadratic methods take a more specialized route:
they preserve the time ordering of the control problem and solve each local
model by a backward dynamic-programming recursion followed by a forward
simulation. This structure is the source of their efficiency.

\subsection{Local Models Around a Nominal Trajectory}

Iterative LQR repeatedly solves a local approximation of the nonlinear
problem. Begin with a nominal control sequence $\bar U$ and roll out the
dynamics to obtain a dynamically feasible nominal trajectory $\bar X$.
Define deviations
\begin{equation}
  \delta x_k=x_k-\bar x_k,
  \qquad
  \delta u_k=u_k-\bar u_k.
\end{equation}
Linearizing the dynamics gives
\begin{equation}
  \delta x_{k+1}
  \approx A_k\delta x_k+B_k\delta u_k.
  \label{eq:ilqr-linearized-dynamics}
\end{equation}
This relation follows from the Taylor expansion
\begin{align}
  f_k(\bar x_k+\delta x_k,\bar u_k+\delta u_k)
  ={}&f_k(\bar x_k,\bar u_k)
     +A_k\delta x_k+B_k\delta u_k
     +O(\norm{(\delta x_k,\delta u_k)}^2).
  \label{eq:ilqr-dynamics-taylor-expansion}
\end{align}
Because the nominal rollout is feasible,
$\bar x_{k+1}=f_k(\bar x_k,\bar u_k)$. Subtracting this identity produces
\cref{eq:ilqr-linearized-dynamics}. If the nominal sequence had a nonzero
dynamic defect, an additional affine term would remain.

The running cost is approximated to second order:
\begin{align}
  \ell_k(x_k,u_k)
  \approx{}&\ell_k
  +\ell_{x,k}\trans\delta x_k
  +\ell_{u,k}\trans\delta u_k\nonumber\\
  &+\frac12\delta x_k\trans\ell_{xx,k}\delta x_k
  +\delta u_k\trans\ell_{ux,k}\delta x_k
  +\frac12\delta u_k\trans\ell_{uu,k}\delta u_k.
  \label{eq:ilqr-quadratic-stage-model}
\end{align}
All quantities on the right are evaluated at $(\bar x_k,\bar u_k)$. The
terminal cost is expanded similarly about $\bar x_N$.

Combining first-order dynamics with second-order cost expansions gives a local
problem with linear time-varying dynamics and an affine--quadratic objective:
\begin{equation}
  \begin{aligned}
    \minimize_{\delta X,\delta U}\quad
      &\Phi_{x,N}\trans\delta x_N
       +\frac12\delta x_N\trans\Phi_{xx,N}\delta x_N\\
      &\quad+\sum_{k=0}^{N-1}
       \left(
         \ell_{x,k}\trans\delta x_k
        +\ell_{u,k}\trans\delta u_k
        +\frac12\delta x_k\trans\ell_{xx,k}\delta x_k
        +\delta u_k\trans\ell_{ux,k}\delta x_k
        +\frac12\delta u_k\trans\ell_{uu,k}\delta u_k
       \right)\\
    \subjectto\quad
      &\delta x_0=0,\\
      &\delta x_{k+1}
       =A_k\delta x_k+B_k\delta u_k,
       \qquad k=0,\ldots,N-1.
  \end{aligned}
  \label{eq:ilqr-local-affine-lqr-problem}
\end{equation}
where
$\Phi_{x,N}=\Phi_x(\bar x_N)$ and
$\Phi_{xx,N}=\Phi_{xx}(\bar x_N)$. Constant terms have been omitted because
they do not affect the minimizing perturbations. The condition
$\delta x_0=0$ reflects the fixed initial state; it can instead be replaced by
a prescribed initial deviation.

Although the objective is written as a sum of stage terms, the dynamics couple
the stages. A change in $\delta u_k$ alters $\delta x_{k+1}$ and therefore all
subsequent running costs and the terminal cost. The iLQR backward pass solves
this coupled problem efficiently by summarizing the remaining cost in a
quadratic value function.

\subsection{Affine--Quadratic Riccati Recursion and the iLQR Backward Pass}

The discrete-time LQR result in
\cref{sec:hjb-linear-quadratic-examples} assumes a homogeneous quadratic cost.
The local model used by iLQR is slightly more general: its Taylor expansion
contains linear terms and a state--control cross term. We first solve this
general affine--quadratic LQR problem using ordinary variables $z_k$ and
$v_k$, and only afterward identify them with the iLQR perturbations.

\paragraph{A general affine--quadratic LQR problem.}
Consider
\begin{equation}
  \begin{aligned}
    \minimize_{Z,V}\quad
      &c_N+s_N\trans z_N+\frac12z_N\trans S_Nz_N\\
      &\quad+\sum_{k=0}^{N-1}
       \left(
         c_k+q_k\trans z_k+r_k\trans v_k
        +\frac12z_k\trans Q_kz_k
        +v_k\trans M_kz_k
        +\frac12v_k\trans R_kv_k
       \right)\\
    \subjectto\quad
      &z_0=z_{\mathrm{init}},\\
      &z_{k+1}=A_kz_k+B_kv_k,
      \qquad k=0,\ldots,N-1.
  \end{aligned}
  \label{eq:general-affine-quadratic-lqr-problem}
\end{equation}
Here $Z=(z_0,\ldots,z_N)$ and $V=(v_0,\ldots,v_{N-1})$, with
$z_k\in\R^n$ and $v_k\in\R^m$. The matrices $Q_k$, $R_k$, and $S_N$ are
symmetric, and $M_k\in\R^{m\times n}$. The vectors $q_k$, $r_k$, and $s_N$
introduce linear costs, while $M_k$ couples state and control. Standard
discrete-time LQR is recovered when all linear and cross terms vanish;
constant cost terms do not affect its controller.

Let $V_k(z)$ denote the minimum remaining cost when stage $k$ begins in state
$z$. Suppose the value at stage $k+1$ has the form
\begin{equation}
  V_{k+1}(z)
  =\beta_{k+1}+p_{k+1}\trans z
   +\frac12z\trans P_{k+1}z.
  \label{eq:general-affine-lqr-value-function}
\end{equation}
The terminal conditions are
\begin{equation}
  \beta_N=c_N,
  \qquad p_N=s_N,
  \qquad P_N=S_N.
  \label{eq:general-affine-lqr-terminal-condition}
\end{equation}
Substituting the dynamics into the Bellman recursion and collecting terms
gives the one-stage minimization
\begin{equation}
  V_k(z)
  =\min_v
  \left\{
    \widehat c_k+\widehat q_k\trans z
    +\frac12z\trans\widehat Q_kz
    +v\trans(F_kz+g_k)
    +\frac12v\trans H_kv
  \right\},
  \label{eq:general-affine-lqr-bellman-step}
\end{equation}
where
\begin{align}
  \widehat c_k&=c_k+\beta_{k+1}, &
  \widehat q_k&=q_k+A_k\trans p_{k+1},
  \label{eq:general-affine-lqr-state-coefficients}\\
  \widehat Q_k&=Q_k+A_k\trans P_{k+1}A_k, &
  g_k&=r_k+B_k\trans p_{k+1},
  \label{eq:general-affine-lqr-linear-control-coefficient}\\
  F_k&=M_k+B_k\trans P_{k+1}A_k, &
  H_k&=R_k+B_k\trans P_{k+1}B_k.
  \label{eq:general-affine-lqr-control-coefficients}
\end{align}
When $H_k\succ0$, the Bellman objective is strictly convex in $v$, and its
unique minimizer satisfies
\begin{equation}
  H_kv+F_kz+g_k=0.
  \label{eq:general-affine-lqr-stationarity}
\end{equation}
Solving this equation gives
\begin{equation}
  v_k^\star=d_k+K_kz_k,
  \qquad
  d_k=-H_k^{-1}g_k,
  \qquad
  K_k=-H_k^{-1}F_k.
  \label{eq:general-affine-lqr-control-law}
\end{equation}
Substitution into the Bellman equation leaves a quadratic function of $z$.
Matching its quadratic, linear, and constant coefficients gives the
generalized Riccati recursion
\begin{align}
  P_k
  &=\widehat Q_k-F_k\trans H_k^{-1}F_k,
  \label{eq:general-affine-lqr-matrix-recursion}\\
  p_k
  &=\widehat q_k-F_k\trans H_k^{-1}g_k,
  \label{eq:general-affine-lqr-vector-recursion}\\
  \beta_k
  &=\widehat c_k-\frac12g_k\trans H_k^{-1}g_k.
  \label{eq:general-affine-lqr-scalar-recursion}
\end{align}
Thus $P_k$ obeys the usual matrix Riccati update with the cross term $M_k$
included through $F_k$. The additional vector and scalar recursions propagate
the linear and constant parts of the value function. Numerically, the systems
with coefficient matrix $H_k$ are solved directly rather than by explicitly
forming $H_k^{-1}$. Starting from the terminal conditions, these recursions
give the optimal policy at every stage, provided $H_k\succ0$ throughout the
backward pass. This condition is not automatic for arbitrary cross terms or
indefinite cost Hessians.

\paragraph{Application to iLQR.}
The local problem in \cref{eq:ilqr-local-affine-lqr-problem} is obtained by
the identifications
\begin{equation}
  \begin{gathered}
    z_k=\delta x_k,
    \qquad v_k=\delta u_k,\\
    q_k=\ell_{x,k},
    \quad r_k=\ell_{u,k},
    \quad Q_k=\ell_{xx,k},
    \quad M_k=\ell_{ux,k},
    \quad R_k=\ell_{uu,k},\\
    s_N=\Phi_{x,N},
    \qquad S_N=\Phi_{xx,N},
    \qquad c_k=c_N=0.
  \end{gathered}
  \label{eq:affine-lqr-ilqr-identification}
\end{equation}
The value coefficients correspond to
$\beta_k=V_{0,k}$, $p_k=V_{x,k}$, and $P_k=V_{xx,k}$ in the local value model
\[
  V_k^{\mathrm{loc}}(\delta x)
  =V_{0,k}+V_{x,k}\trans\delta x
   +\frac12\delta x\trans V_{xx,k}\delta x.
\]
These coefficients describe the optimized local problem, not a globally
computed value function of the original nonlinear system. Since constant terms
were omitted from \cref{eq:ilqr-local-affine-lqr-problem}, the terminal
conditions are
\begin{equation}
  V_{0,N}=0,
  \qquad
  V_{x,N}=\Phi_x(\bar x_N),
  \qquad
  V_{xx,N}=\Phi_{xx}(\bar x_N).
  \label{eq:ilqr-terminal-value-derivatives}
\end{equation}

For compatibility with standard iLQR notation, denote the Bellman
coefficients $\widehat q_k$, $g_k$, $\widehat Q_k$, $F_k$, and $H_k$ by
$Q_x$, $Q_u$, $Q_{xx}$, $Q_{ux}$, and $Q_{uu}$, respectively. Under
\cref{eq:affine-lqr-ilqr-identification}, they are
\begin{align}
  Q_x&=\ell_x+A\trans V_x^+, &
  Q_u&=\ell_u+B\trans V_x^+,
  \label{eq:ilqr-q-first-derivatives}\\
  Q_{xx}&=\ell_{xx}+A\trans V_{xx}^+A, &
  Q_{ux}&=\ell_{ux}+B\trans V_{xx}^+A,
  \label{eq:ilqr-q-state-derivatives}\\
  Q_{uu}&=\ell_{uu}+B\trans V_{xx}^+B.
  \label{eq:ilqr-q-control-hessian}
\end{align}
Here the stage index $k$ is suppressed and the superscript $+$ denotes the
value coefficients at stage $k+1$. The symbols $Q_x$, $Q_u$, and the Hessian
blocks are coefficients of the one-stage Bellman quadratic; they should not
be confused with the state-cost weight $Q_k$ in the general LQR problem.

Because $H_k=Q_{uu}$, $g_k=Q_u$, and $F_k=Q_{ux}$, the general affine feedback
law in \cref{eq:general-affine-lqr-control-law} becomes the iLQR correction,
assuming $Q_{uu}\succ0$,
\begin{equation}
  \delta u_k^\star
  =k_k+K_k\delta x_k,
  \qquad
  k_k=-Q_{uu}^{-1}Q_u,
  \qquad
  K_k=-Q_{uu}^{-1}Q_{ux}.
  \label{eq:ilqr-local-control-law}
\end{equation}
The feedforward term $k_k$ improves the nominal controls. The feedback gain
$K_k$ compensates for state deviations encountered during the next nonlinear
rollout. If all stage and terminal linear cost terms vanish, backward
induction gives $V_x^+=0$, so $Q_u=0$ and $k_k=0$;
the policy reduces to the purely linear discrete-time LQR feedback law.

Likewise, the matrix and vector recursions in
\cref{eq:general-affine-lqr-matrix-recursion,eq:general-affine-lqr-vector-recursion}
become
\begin{align}
  V_x&=Q_x-Q_{ux}\trans Q_{uu}^{-1}Q_u,
  \label{eq:ilqr-value-gradient-update}\\
  V_{xx}&=Q_{xx}-Q_{ux}\trans Q_{uu}^{-1}Q_{ux}.
  \label{eq:ilqr-schur-value-update}
\end{align}
Numerically, one solves linear systems involving $Q_{uu}$ rather than forming
its inverse. The recursion proceeds backward from $k=N-1$ to $k=0$, just as
in finite-horizon LQR and Bellman recursion.

The scalar recursion in \cref{eq:general-affine-lqr-scalar-recursion}
similarly propagates $V_{0,k}$. This constant does not affect the policy.
Because nominal cost terms were omitted and $\delta x_0=0$, $V_{0,0}$ is the
predicted cost change for a full step in the unregularized local problem.

\subsection{The Nonlinear Forward Pass}

The backward pass optimizes only a local model. Its proposed policy must be
tested on the original nonlinear dynamics. For a step length
$\alpha\in(0,1]$, set $x_0^{\mathrm{new}}=x_{\mathrm{init}}$ and apply
\begin{align}
  u_k^{\mathrm{new}}
  &=\bar u_k+\alpha k_k
    +K_k(x_k^{\mathrm{new}}-\bar x_k),
  \label{eq:ilqr-forward-control}\\
  x_{k+1}^{\mathrm{new}}
  &=f_k(x_k^{\mathrm{new}},u_k^{\mathrm{new}}).
  \label{eq:ilqr-forward-rollout}
\end{align}
If the new cost is sufficiently smaller, the rollout becomes the next nominal
trajectory. Otherwise, $\alpha$ is reduced and the forward pass is tried
again. This line search prevents a locally accurate quadratic model from
taking an unreliable large step.

The feedforward and feedback terms play different roles. Multiplying $k_k$ by
$\alpha$ controls how far the nominal control is moved. The feedback term is
not normally scaled because it is the local response to whatever state the
nonlinear rollout actually reaches. Thus the forward pass evaluates a policy,
not merely a vector addition in the stacked control variables.

One iLQR iteration therefore consists of:
\begin{enumerate}
  \item rolling the nonlinear dynamics forward to obtain $(\bar X,\bar U)$;
  \item differentiating the dynamics and costs along that trajectory;
  \item performing the backward recursion to compute $(k_k,K_k)$;
  \item line-searching with nonlinear forward rollouts; and
  \item accepting the improved trajectory and repeating until convergence.
\end{enumerate}

The feedback term in
\cref{eq:ilqr-forward-control} matters even though the initial state is fixed.
As the feedforward controls change, the nonlinear rollout departs from the
nominal states used in the backward pass. Feedback corrects these departures
and usually makes the trial trajectory much more reliable than a purely
open-loop update.

\subsection{Regularization and Predicted Improvement}

Far from a local minimum, $Q_{uu}$ may fail to be positive definite. Then the
quadratic model does not define a unique minimizing control step. A common
remedy is to use
\begin{equation}
  Q_{uu}^{\mathrm{reg}}=Q_{uu}+\mu I,
  \qquad \mu>0,
  \label{eq:ilqr-control-regularization}
\end{equation}
with $\mu$ large enough to make $Q_{uu}^{\mathrm{reg}}$ positive definite.
Another option is to regularize the value Hessian before forming the $Q$
coefficients. Large $\mu$ produces conservative steps; successful
steps permit $\mu$ to decrease, while rejected steps cause it to increase.

Regularized gains need consistent value updates. The Schur-complement formulas
in \cref{eq:ilqr-value-gradient-update,eq:ilqr-schur-value-update} assume that
the gains minimize the same quadratic model being propagated. If regularized
gains are substituted into the original, unregularized model, use the full
policy-substitution formulas instead:
\begin{align}
  V_x
  &=Q_x+K\trans Q_u+Q_{ux}\trans k+K\trans Q_{uu}k,\\
  V_{xx}
  &=Q_{xx}+K\trans Q_{uu}K+K\trans Q_{ux}+Q_{ux}\trans K.
  \label{eq:ilqr-policy-substitution-value-update}
\end{align}
For unregularized minimizing gains, these expressions reduce to the earlier
Schur-complement updates.

For the unregularized backward pass, the local model predicts the cost change
under the line-search policy as
\begin{equation}
  \Delta J_{\mathrm{pred}}(\alpha)
  =\sum_{k=0}^{N-1}
   \left(
     \alpha Q_{u,k}\trans k_k
     +\frac12\alpha^2 k_k\trans Q_{uu,k}k_k
   \right).
  \label{eq:ilqr-predicted-cost-change}
\end{equation}
With regularized gains, a consistent prediction can instead be obtained by
evaluating the original local quadratic objective along the trial policy with
the linearized dynamics. A predicted descent step has
$\Delta J_{\mathrm{pred}}(\alpha)<0$. Comparing the actual decrease with the
predicted decrease gives the ratio
\begin{equation}
  \rho
  =\frac{J(\bar X,\bar U)-J(X^{\mathrm{new}},U^{\mathrm{new}})}
         {-\Delta J_{\mathrm{pred}}(\alpha)}.
  \label{eq:ilqr-improvement-ratio}
\end{equation}
When $\rho$ is near one, the nonlinear rollout agrees with the quadratic
model. A small positive ratio means the step helped less than expected; a
negative ratio means the cost increased. This ratio is useful only when the
predicted decrease is positive and not numerically negligible. An
implementation may accept only steps with $\rho$ above a small threshold,
decrease $\mu$ after consistently accurate steps, and increase $\mu$ after
rejection. This couples line search
and regularization into a trust mechanism for the local model.

\subsection{Differential Dynamic Programming}

iLQR linearizes the dynamics and quadratizes the costs. \emph{Differential
dynamic programming} (DDP) constructs a fully second-order local model by
also retaining the curvature of the dynamics. Write the component functions
of $f$ as $f_1,\ldots,f_n$. Around a nominal state and control,
\begin{align}
  \delta x^+
  \approx{}&A\delta x+B\delta u\nonumber\\
  &+\frac12
  \begin{bmatrix}
    \delta x\trans(f_1)_{xx}\delta x
    +2\delta u\trans(f_1)_{ux}\delta x
    +\delta u\trans(f_1)_{uu}\delta u\\
    \vdots\\
    \delta x\trans(f_n)_{xx}\delta x
    +2\delta u\trans(f_n)_{ux}\delta x
    +\delta u\trans(f_n)_{uu}\delta u
  \end{bmatrix}.
  \label{eq:ddp-second-order-dynamics-expansion}
\end{align}
When this expression is substituted into the local value model, the linear
term $(V_x^+)\trans\delta x^+$ multiplies the second-order dynamics terms.
The DDP curvature matrices therefore become
\begin{align}
  Q_{xx}^{\mathrm{DDP}}
  &=\ell_{xx}+A\trans V_{xx}^+A
    +\sum_{i=1}^n(V_x^+)_i(f_i)_{xx},
  \label{eq:ddp-qxx}\\
  Q_{ux}^{\mathrm{DDP}}
  &=\ell_{ux}+B\trans V_{xx}^+A
    +\sum_{i=1}^n(V_x^+)_i(f_i)_{ux},
  \label{eq:ddp-qux}\\
  Q_{uu}^{\mathrm{DDP}}
  &=\ell_{uu}+B\trans V_{xx}^+B
    +\sum_{i=1}^n(V_x^+)_i(f_i)_{uu}.
  \label{eq:ddp-quu}
\end{align}
The first-order formulas for $Q_x$ and $Q_u$ are unchanged, although their
values depend on the value coefficients propagated by DDP. When
$Q_{uu}^{\mathrm{DDP}}\succ0$, the local minimizing policy is
\begin{equation}
  k=-\left(Q_{uu}^{\mathrm{DDP}}\right)^{-1}Q_u,
  \qquad
  K=-\left(Q_{uu}^{\mathrm{DDP}}\right)^{-1}Q_{ux}^{\mathrm{DDP}}.
  \label{eq:ddp-local-policy}
\end{equation}
The value-gradient and value-Hessian recursions in
\cref{eq:ilqr-value-gradient-update,eq:ilqr-schur-value-update} retain their
form, but use the DDP curvature matrices:
\begin{align}
  V_x
  &=Q_x-\left(Q_{ux}^{\mathrm{DDP}}\right)\trans
    \left(Q_{uu}^{\mathrm{DDP}}\right)^{-1}Q_u,
  \label{eq:ddp-value-gradient-update}\\
  V_{xx}
  &=Q_{xx}^{\mathrm{DDP}}
    -\left(Q_{ux}^{\mathrm{DDP}}\right)\trans
     \left(Q_{uu}^{\mathrm{DDP}}\right)^{-1}Q_{ux}^{\mathrm{DDP}}.
  \label{eq:ddp-value-hessian-update}
\end{align}
Both value updates generally differ numerically from iLQR. Even though
$Q_x$ and $Q_u$ have the same first-order formulas, the update of $V_x$
depends on the modified $Q_{ux}$ and $Q_{uu}$. The updated $V_x$ and
$V_{xx}$ are passed to the preceding stage in the backward recursion.
DDP uses the same nonlinear forward rollout and line search as iLQR.

The difference has a useful optimization interpretation. Omitting the
dynamics-curvature contractions makes iLQR similar to a Gauss--Newton method:
it keeps curvature generated by the cost and by first-order state
sensitivities, but not every term in the exact Hessian of the reduced
objective. DDP is closer to Newton's method because it retains those missing
terms. Near a smooth local solution, this can make its model more accurate and
its convergence faster.

The added accuracy has a price. The tensors $(f_i)_{xx}$, $(f_i)_{ux}$, and
$(f_i)_{uu}$ are expensive to form and store for a high-dimensional robot.
In practice, automatic differentiation can compute the required tensor
contractions without materializing every tensor. DDP also still requires
regularization whenever its $Q_{uu}$ is not positive definite; exact second-order
information does not guarantee that a nonconvex local model is suitable for
an unrestricted step. iLQR requires fewer dynamics derivatives, while DDP
can be useful when improved local accuracy justifies the additional cost.
The regularization and value-update precautions above apply to both methods.

To see why only the Hessian blocks change, write the future value composition
as $V^+(f(x,u))$. Its first derivative is the ordinary chain rule
\begin{equation}
  \nabla_x[V^+\circ f]=f_x\trans V_x^+.
\end{equation}
Differentiating once more gives
\begin{equation}
  \nabla_{xx}^2[V^+\circ f]
  =f_x\trans V_{xx}^+f_x
   +\sum_{i=1}^n(V_x^+)_i(f_i)_{xx}.
  \label{eq:ddp-composed-value-hessian}
\end{equation}
The first term is the curvature retained by iLQR; the second is the dynamics
curvature added by DDP. The $ux$ and $uu$ formulas follow from the same chain
rule. This derivation also shows when the two methods coincide: if $f$ is
affine, every second derivative of $f_i$ vanishes.

\section{Constraints and Practical Implementation}

The derivation so far assumes smooth dynamics, smooth costs, and unconstrained
local control updates. Robotic applications rarely satisfy all three
assumptions. The following subsections explain how constraints and contact modify
the basic algorithms and then collect the numerical checks needed in a robust
implementation.

\subsection{Input and State Constraints}

Simple box constraints
\begin{equation}
  u_{\min}\leq u_k\leq u_{\max}
\end{equation}
can be incorporated into each local control minimization. Instead of applying
\cref{eq:ilqr-local-control-law} directly, one solves the small box-constrained
quadratic problem
\begin{equation}
  \min_{\delta u}
  \quad
  Q_u\trans\delta u
  +\delta u\trans Q_{ux}\delta x
  +\frac12\delta u\trans Q_{uu}\delta u
  \quad
  \subjectto\quad
  u_{\min}\leq\bar u+\delta u\leq u_{\max}.
  \label{eq:ilqr-box-constrained-local-qp}
\end{equation}
If $Q_{uu}\succ0$ and the unconstrained candidate lies strictly within the
input bounds, no bound is active and the solution is exactly the affine law
in \cref{eq:ilqr-local-control-law}:
\begin{equation}
  \delta u^\star
  =-Q_{uu}^{-1}(Q_u+Q_{ux}\delta x)
  =k+K\delta x,
  \qquad
  u_{\min}<\bar u+k+K\delta x<u_{\max}.
\end{equation}
If the candidate violates an input bound, solve
\cref{eq:ilqr-box-constrained-local-qp} using a quadratic-programming (QP)
algorithm. Simply clipping the unconstrained control can leave the feedback
gain inconsistent with the active bounds.

State constraints are harder because they couple current actions to future
feasibility. Smooth penalties or barrier functions can be added to the cost,
but penalties permit violation and barriers require a feasible interior.
Augmented-Lagrangian and sequential quadratic programming methods provide
stronger constraint handling. When collision, contact, and terminal
constraints dominate the problem, direct transcription is often more natural
than basic iLQR.

For a scalar inequality $g(x,u)\leq0$, a quadratic penalty adds
\begin{equation}
  \ell_{\mathrm{pen}}(x,u)
  =\ell(x,u)+\frac{\eta}{2}\max\{0,g(x,u)\}^2.
  \label{eq:trajectory-quadratic-penalty}
\end{equation}
For several inequalities, sum one such term per constraint. Increasing
$\eta$ discourages violation but can make the Hessian poorly conditioned.
The squared positive-part penalty is continuously differentiable, but its
second derivative generally jumps at $g=0$; a second-order solver must handle
this change or use a smooth approximation. A logarithmic barrier instead uses
\begin{equation}
  \ell_{\mathrm{bar}}(x,u)
  =\ell(x,u)-\tau\log[-g(x,u)],
  \qquad g(x,u)<0,
  \label{eq:trajectory-log-barrier}
\end{equation}
with $\tau>0$. Reducing $\tau$ weakens the barrier bias, but every trial
trajectory must remain strictly feasible for the barrier to be defined.

For an equality $h_k(x_k,u_k)=0$, the augmented-Lagrangian approach adds a
multiplier term and a quadratic penalty to the original stage cost:
\begin{equation}
  \widetilde\ell_k(x,u;\nu_k,\eta)
  =\ell_k(x,u)+\nu_k\trans h_k(x,u)
   +\frac{\eta}{2}\norm{h_k(x,u)}^2,
  \qquad \eta>0.
  \label{eq:trajectory-augmented-lagrangian-term}
\end{equation}
Thus the inner iLQR solver uses derivatives of $\widetilde\ell_k$ instead of
those of $\ell_k$; the dynamics are unchanged. A terminal equality is treated
similarly by augmenting $\Phi$. Each stage has its own multiplier vector
$\nu_k$, with one entry for each equality at that stage. Equality multipliers
can have either sign.

\paragraph{Choosing and updating the parameters.}
A simple initialization is $\nu_k^{(0)}=0$ and a moderate positive penalty
$\eta^{(0)}$. There is no universal numerical value for $\eta$: first scale
the constraint residuals to comparable magnitudes, then choose a penalty
that makes meaningful violations costly without overwhelming the original
objective. If a nearby problem has already been solved, its multipliers can
provide a warm start.

At outer iteration $r$, hold $\nu_k^{(r)}$ and $\eta^{(r)}$ fixed while the
inner trajectory solver approximately minimizes the augmented objective.
After obtaining the updated trajectory $(X^{(r+1)},U^{(r+1)})$, update
\begin{equation}
  \nu_k^{(r+1)}
  =\nu_k^{(r)}+\eta^{(r)}
    h_k(x_k^{(r+1)},u_k^{(r+1)}).
  \label{eq:trajectory-augmented-lagrangian-multiplier-update}
\end{equation}
Keep $\eta$ unchanged if the constraint residuals decrease sufficiently;
otherwise increase it by a factor greater than one and repeat the inner
solve. These are outer-loop updates, not updates at every iLQR backward pass
or line-search trial. Starting with an excessively large penalty can make
the local quadratic models poorly conditioned.

The multiplier update accumulates a correction for persistent constraint
violation, so feasibility need not be pursued solely by sending the penalty
to infinity. Under suitable convergence assumptions, the multipliers approach
the KKT multipliers of the constrained optimum. In practice, stop
only when both the constraint residuals and the augmented-objective
stationarity residuals are small; a low augmented cost alone does not
establish feasibility.

\subsection{Trajectory Optimization Through Contact Dynamics}

Contact-rich motions are naturally viewed as \emph{hybrid systems}: continuous
motion within a contact mode is interrupted by events such as touchdown,
liftoff, or changes between sticking and sliding. Impacts may reset velocity,
and different contact modes impose different dynamics and constraints.
Optimizing a motion can therefore require choosing not only controls and
states, but also the contact sequence and transition times.

If the contact sequence is known, a hybrid trajectory formulation can optimize
the motion within each phase and enforce transition conditions between phases.
This simplifies the search, but cannot discover motions requiring a different
sequence. \emph{Contact-implicit trajectory optimization} instead retains
contact forces or impulses as decision variables and imposes contact
consistency alongside the dynamics. Unilateral contact and complementarity
conditions express that separated bodies cannot exert a normal contact force,
allowing contact activation to emerge from the optimization rather than from
a prescribed schedule.

\paragraph{A representative contact-implicit problem.}
Consider Euclidean generalized coordinates $q$, with $v=\dot q$, and contact
with stationary surfaces. For candidate contact $j$, let $\phi_j(q)$ be the
signed gap, positive when the bodies are separated. Introduce the normal and
tangential contact impulses $p_{n,j,k}$ and $p_{t,j,k}$ over interval $k$,
and stack them in $p_k$. Impulses are integrated contact forces, so they can
represent both sustained contact and impact within a time step. An
illustrative backward-Euler transcription for inelastic contact is
\begin{equation}
  \begin{aligned}
    \minimize_{Q,V,U,P}\quad
      &\Phi(q_N,v_N)+\sum_{k=0}^{N-1}\ell_k(q_k,v_k,u_k)\\
    \subjectto\quad
      &q_0=q_{\mathrm{init}},\qquad v_0=v_{\mathrm{init}},\\
      &q_{k+1}=q_k+\Delta t_k v_{k+1},\\
      &M_{k+1}(v_{k+1}-v_k)
        =\Delta t_k r_{k+1}+J_{k+1}\trans p_k,\\
      &\phi_j(q_{k+1})\geq0,\qquad p_{n,j,k}\geq0,\\
      &p_{n,j,k}\phi_j(q_{k+1})=0,\\
      &\norm{p_{t,j,k}}_2\leq\mu_j p_{n,j,k},\\
      &p_{t,j,k}\trans w_{t,j,k+1}
        =-\mu_j p_{n,j,k}\norm{w_{t,j,k+1}}_2,\\
      &g_k(q_k,v_k,u_k)\leq0,\qquad h_N(q_N,v_N)=0.
  \end{aligned}
  \label{eq:contact-implicit-trajectory-problem}
\end{equation}
Here $Q=(q_0,\ldots,q_N)$, $V=(v_0,\ldots,v_N)$,
$P=(p_0,\ldots,p_{N-1})$, and $U$ is the control sequence. The interval
constraints hold for $k=0,\ldots,N-1$, and the contact constraints hold for
every candidate contact $j$. The quantities
$M_{k+1}=M(q_{k+1})$, $J_{k+1}=J(q_{k+1})$, and
$r_{k+1}=r(q_{k+1},v_{k+1},u_k)$ denote the mass matrix, stacked contact
Jacobian, and noncontact generalized forces. The tangential relative velocity
is $w_{t,j,k+1}=J_{t,j}(q_{k+1})v_{k+1}$, and $\mu_j\geq0$ is the friction
coefficient. The functions $g_k$ and $h_N$ encode task-specific bounds and
terminal requirements. This is one discretization of contact-implicit
optimization, not a unique contact model; see \citet{posa2014direct}.

The normal-contact conditions are often abbreviated as
\[
  0\leq p_{n,j,k}\ \perp\ \phi_j(q_{k+1})\geq0.
\]
They encode a choice: either the gap is positive and the normal impulse is
zero, or the gap is zero and a nonnegative normal impulse is possible. Thus
the optimizer selects contact modes without explicit binary variables.
The two friction constraints express Coulomb friction with maximum
dissipation: during sliding, the tangential impulse lies on the friction-cone
boundary and opposes tangential velocity; during sticking, it can lie anywhere
inside the cone.

\paragraph{Why the constraints are difficult.}
The momentum equation is nonlinear because $M$, $J$, and $r$ depend on the
unknown state, and $J\trans p$ couples configuration and contact impulse.
The gap functions are generally nonlinear geometric quantities.
Complementarity adds product equalities such as
$p_{n,j,k}\phi_j(q_{k+1})=0$, whose feasible set is nonconvex even when the
gap is affine. For example, both $(\phi,p_n)=(1,0)$ and $(0,1)$ satisfy
complementarity, but their midpoint does not. The friction-cone inequality
alone is convex in the impulses for fixed $\mu_j$; the dissipation equality
couples impulses to velocity and is nonsmooth at zero tangential velocity.
Together, these conditions create a nonlinear contact-constrained problem,
not a QP with simple input bounds.

Such formulations are commonly called \emph{mathematical programs with
complementarity constraints} (MPCCs). Contact-implicit optimization therefore
does not eliminate the hybrid nature of the problem. Initialization,
complementarity relaxation or continuation, and time-grid refinement remain
important. A computed motion should be checked for contact consistency,
friction feasibility, and sensitivity to discretization. Smooth contact
models, such as those discussed in Chapter~5, offer another route to local
trajectory optimization, but trade numerical smoothness against faithful
impact behavior.

For algorithmic details, see the complementarity-based direct method of
\citet{posa2014direct} and the contact-implicit orthogonal-collocation method
of \citet{patel2019contact}. Both extend direct transcription to contact-rich
motion without prescribing the complete contact sequence. The key distinction
from basic iLQR is that contact-mode changes require appropriate hybrid,
contact-implicit, or regularized modeling rather than an assumption of smooth
dynamics across every transition.

\subsection{Derivatives and Model Checks}

Analytic derivatives are fast and interpretable but can be tedious for a
large robot model. Automatic differentiation evaluates derivatives of the
implemented dynamics and costs and is usually preferable to differentiating a
separate symbolic model. Finite differences remain useful for verification,
although they require careful step sizes and can be noisy.

A simple directional check compares
\begin{equation}
  \frac{f(x+\epsilon v,u)-f(x,u)}{\epsilon}
  \quad\text{with}\quad
  f_x(x,u)v
\end{equation}
for several random directions $v$ and decreasing $\epsilon$. Similar checks
apply to cost gradients and Hessian-vector products. Incorrect derivatives
often appear as consistently rejected line searches or a large mismatch
between predicted and actual cost decrease.

A stronger check examines the Taylor remainder. If $A=f_x(x,u)$ is correct,
then
\begin{equation}
  r_1(\epsilon)
  =\norm{f(x+\epsilon v,u)-f(x,u)}
  =O(\epsilon),
\end{equation}
whereas subtracting the linear prediction should give
\begin{equation}
  r_2(\epsilon)
  =\norm{f(x+\epsilon v,u)-f(x,u)-\epsilon Av}
  =O(\epsilon^2).
  \label{eq:trajectory-derivative-taylor-test}
\end{equation}
For generic directions with nonzero leading terms, a log--log plot of these
residuals should have slopes one and two over an appropriate range of step
sizes. For extremely small $\epsilon$, floating-point cancellation eventually
dominates, so the test should use a range of step sizes rather than a single
finite difference.

Nonsmooth simulation requires special care. Contact switches, saturation,
lookup-table discontinuities, and angle-wrapping operations may invalidate
ordinary derivatives. Smoothing, hybrid formulations, or nonsmooth
optimization methods may be necessary. Automatic differentiation faithfully
differentiates the code path that was executed; it does not make a
discontinuous model smooth.

\subsection{Numerical Design Choices}

Several practical choices often matter as much as the formal algorithm.

\paragraph{Discretization.}
A smaller sampling interval can improve integration accuracy, but increases
the number of stages and computation for a fixed physical horizon.
Higher-order integration can improve accuracy without as many stages, but its
derivatives must also be computed correctly.
The optimization and the final controller should use compatible models and
sampling conventions.

\paragraph{Cost design.}
Very large terminal weights encourage accuracy but can create poor conditioning;
they do not impose an exact terminal constraint.
Very small control weights can demand unrealistic actions and make
$Q_{uu}$ nearly singular. Smooth geometric costs, such as periodic angle
errors, are usually preferable to coordinate expressions with artificial
discontinuities.

\paragraph{Initialization.}
Useful guesses can come from a simpler model, a previous solution, a motion
library, or a gradual continuation in task difficulty. In receding-horizon
control, shifting the previous solution by one time step provides a natural
warm start.

\paragraph{Termination.}
No single test is sufficient. Useful criteria include small cost reduction,
small feedforward steps, small gradient norms, acceptable constraint
violation, and agreement between predicted and actual improvement. A solver
should also report failure when regularization grows excessively or no line
search step is accepted.

\paragraph{Complexity.}
With dense state and control matrices, an iLQR backward pass is linear in the
horizon length and polynomial in $n$ and $m$. It avoids solving one dense
problem in all controls at once. Direct transcription also scales linearly in
horizon at the sparsity-pattern level, but its sparse linear algebra has
different constants and handles general constraints more directly.

\subsection{Choosing a Method}

Single shooting is attractive for small, smooth, well-conditioned problems.
Multiple shooting improves robustness on unstable or long-horizon dynamics.
Direct transcription is well suited to problems with many state and path
constraints. iLQR and DDP are especially effective for smooth nonlinear
dynamics with modest control dimension and costs that admit useful local
quadratic models.

These methods are not mutually exclusive. An iLQR solution can initialize a
constrained transcription, while a coarse transcription can provide a motion
around which iLQR constructs fast local feedback. The right formulation is
the one that exposes the structure most important for the task.

\paragraph{Exercises.}

\begin{exercise}
Contrast indirect single shooting from Chapter~6 with direct single shooting
in this chapter. What is guessed or optimized in each method, and what
equations are integrated?
\end{exercise}
\begin{proof}[Short solution]
Indirect shooting guesses unknown boundary data such as the initial costate,
integrates the PMP state--costate equations, and uses root finding to satisfy
terminal conditions. Direct shooting optimizes a sampled control sequence,
integrates only the original state dynamics, and minimizes the discretized
cost. The indirect result is an extremal of the continuous-time necessary
conditions. The direct optimizer seeks a local minimum of the discretized
problem, although convergence by itself may establish only a stationary
candidate.
\end{proof}

\begin{exercise}
Write the unconstrained trajectory problem
\cref{eq:trajectory-optimization-dynamics,eq:trajectory-optimization-cost}
as a direct-transcription nonlinear program. How many scalar state and control
variables are present when $x_k\in\R^n$ and $u_k\in\R^m$?
\end{exercise}
\begin{proof}[Short solution]
Optimize over $x_0,\ldots,x_N$ and $u_0,\ldots,u_{N-1}$ subject to
$x_0=x_{\mathrm{init}}$ and
$x_{k+1}-f_k(x_k,u_k)=0$. There are $(N+1)n+Nm$ scalar variables, or $Nn+Nm$
free variables if the fixed initial state is eliminated.
\end{proof}

\begin{exercise}
Derive \cref{eq:single-shooting-state-sensitivity} from the linearized
dynamics.
\end{exercise}
\begin{proof}[Short solution]
The perturbation $\delta u_k$ first gives
$\delta x_{k+1}=B_k\delta u_k$. With no later control perturbations,
$\delta x_{k+2}=A_{k+1}B_k\delta u_k$. Repeated substitution gives
\begin{equation}
  \delta x_j
  =A_{j-1}\cdots A_{k+1}B_k\delta u_k,
\end{equation}
which identifies the stated Jacobian.
\end{proof}

\begin{exercise}
Show that the adjoint recursion
\cref{eq:shooting-adjoint-terminal,eq:shooting-adjoint-recursion} gives the
single-shooting gradient in \cref{eq:shooting-control-gradient}.
\end{exercise}
\begin{proof}[Short solution]
The direct effect of $u_k$ is $\ell_{u,k}$. Its effect on the next state is
$B_k\delta u_k$, and $\lambda_{k+1}$ is the derivative of all remaining cost
with respect to that state. Therefore
\begin{equation}
  \delta J
  =\left(\ell_{u,k}+B_k\trans\lambda_{k+1}\right)\trans\delta u_k.
\end{equation}
Backward substitution of future state effects gives the adjoint recursion.
\end{proof}

\begin{exercise}
Derive the state and input Jacobians of the pendulum dynamics in
\cref{eq:pendulum-discrete-angle,eq:pendulum-discrete-velocity}.
\end{exercise}
\begin{proof}[Short solution]
Differentiate \cref{eq:pendulum-discrete-angle} with respect to
$(\theta,\omega,u)$ to obtain $(1,h,0)$. Differentiating
\cref{eq:pendulum-discrete-velocity} gives
$-hg\cos\theta/l$, $1-hb/(ml^2)$, and $h/(ml^2)$, respectively.
\end{proof}

\begin{exercise}
For fixed $\delta x$, minimize the control-dependent quadratic
$Q_u\trans\delta u+\delta u\trans Q_{ux}\delta x
+\frac12\delta u\trans Q_{uu}\delta u$ without input bounds and derive
\cref{eq:ilqr-local-control-law}.
\end{exercise}
\begin{proof}[Short solution]
The derivative with respect to $\delta u$ is
\begin{equation}
  Q_u+Q_{ux}\delta x+Q_{uu}\delta u.
\end{equation}
Setting it to zero and using $Q_{uu}\succ0$ gives
\begin{equation}
  \delta u^\star
  =-Q_{uu}^{-1}Q_u-Q_{uu}^{-1}Q_{ux}\delta x
  =k+K\delta x.
\end{equation}
\end{proof}

\begin{exercise}
State the dimensions of $Q_u$, $Q_{ux}$, $Q_{uu}$, $k$, and $K$ for
$x\in\R^n$ and $u\in\R^m$.
\end{exercise}
\begin{proof}[Short solution]
$Q_u,k\in\R^m$, $Q_{ux},K\in\R^{m\times n}$, and
$Q_{uu}\in\R^{m\times m}$.
\end{proof}

\begin{exercise}
Explain why one unregularized iLQR backward--forward iteration solves an
unconstrained finite-horizon LQR problem when the line search accepts
$\alpha=1$.
\end{exercise}
\begin{proof}[Short solution]
For linear dynamics and quadratic costs, the iLQR local models are exact rather
than approximations. The backward pass is precisely the affine Riccati
recursion, and the forward pass applies its exact minimizing policy. Hence no
further local-model correction is required.
\end{proof}

\begin{exercise}
Why is the feedback term $K_k(x_k^{\mathrm{new}}-\bar x_k)$ useful during an
iLQR line search even though $x_0$ is fixed?
\end{exercise}
\begin{proof}[Short solution]
The changed feedforward controls and nonlinear dynamics make later trial states
differ from the nominal states. The feedback term responds to these deviations
using the local value model, keeping the rollout closer to the improvement
predicted by the backward pass.
\end{proof}

\begin{exercise}
For scalar $u$, minimize
\begin{equation}
  q(u)=au+\frac12bu^2,
  \qquad b>0,
\end{equation}
subject to $u_{\min}\leq u\leq u_{\max}$.
\end{exercise}
\begin{proof}[Short solution]
The unconstrained minimizer is $-a/b$. The constrained solution is its
projection onto the interval:
\begin{equation}
  u^\star=\min\{u_{\max},\max\{u_{\min},-a/b\}\}.
\end{equation}
\end{proof}

\begin{exercise}
What terms distinguish DDP from iLQR in the backward pass?
\end{exercise}
\begin{proof}[Short solution]
DDP retains contractions of the next value gradient with second derivatives
of the dynamics, such as
$\sum_i(V_x^+)_i(f_i)_{xx}$ and the corresponding $ux$ and $uu$ terms. iLQR
omits these terms and uses linearized dynamics.
\end{proof}

\begin{exercise}
Show that the DDP curvature terms in
\cref{eq:ddp-qxx,eq:ddp-qux,eq:ddp-quu} vanish when the dynamics are affine in
state and control. What does this imply about iLQR and DDP for an LQR problem?
\end{exercise}
\begin{proof}[Short solution]
Affine dynamics have
$(f_i)_{xx}=(f_i)_{ux}=(f_i)_{uu}=0$ for every component $i$. Therefore the
DDP $Q$ matrices reduce to the iLQR matrices. For linear dynamics and
quadratic costs, both methods use the exact local problem and reproduce the
finite-horizon LQR solution.
\end{proof}

\begin{exercise}
Why can prescribing a contact sequence simplify trajectory optimization, and
what flexibility does a contact-implicit formulation provide?
\end{exercise}
\begin{proof}[Short solution]
A prescribed sequence fixes the order of hybrid modes, so the optimizer only
needs to determine the motion and transition times within that sequence.
A contact-implicit formulation imposes contact consistency without fixing the
complete sequence, allowing different contact patterns to emerge at the cost
of a more difficult nonconvex problem.
\end{proof}

\begin{exercise}
Why can a direct-transcription iterate have low objective value but still fail
to represent a usable motion?
\end{exercise}
\begin{proof}[Short solution]
The independently optimized states may violate the dynamic defects. Until
$x_{k+1}=f_k(x_k,u_k)$ is satisfied to adequate tolerance, the listed states
cannot be produced by applying the listed controls to the system.
\end{proof}

\begin{exercise}
For the pendulum swing-up cost in \cref{eq:pendulum-swing-up-cost}, explain why
initializing iLQR with $\theta_k=0$, $\omega_k=0$, and $u_k=0$ can be
problematic.
\end{exercise}
\begin{proof}[Short solution]
At the downward equilibrium, the derivative of $1+\cos\theta$ is
$-\sin\theta=0$. The symmetric zero trajectory therefore has no first-order
signal choosing clockwise or counterclockwise motion. A small asymmetric
control initialization, a reference trajectory, or continuation can break
the symmetry.
\end{proof}

\paragraph{Chapter summary.}
Trajectory optimization searches for a finite state--control sequence rather
than a global value function. The controls are coupled through the dynamics:
an early action changes every later state and can strongly affect terminal
performance. Unlike the indirect methods of Chapter~6, the direct methods in
this chapter discretize the original control problem before optimization
rather than first deriving and solving the PMP boundary-value equations.
Direct single shooting eliminates states by forward simulation and can compute
gradients efficiently with a backward adjoint recursion, but long products of
dynamics Jacobians may cause poor conditioning.

Multiple shooting and direct transcription introduce state variables and
enforce dynamic continuity with sparse equality constraints. They are often
preferable when horizons are long, dynamics are unstable, or state and path
constraints are central. Their intermediate iterates may be dynamically
infeasible, so defect residuals must be checked alongside objective value.

iLQR exploits the same temporal structure through a different route. It
linearizes the dynamics, quadratizes the costs, and performs a Riccati-like
backward pass to obtain feedforward improvements and local feedback gains. A
nonlinear forward rollout and line search test the proposed policy on the
original system. Regularization ensures that the local control minimization is
well posed, while predicted-versus-actual improvement reveals when the local
model is trustworthy. DDP refines this construction by retaining second-order
dynamics terms, producing a more Newton-like local model at the cost of
additional derivatives and regularization.

Robotic examples show why several formulations are useful. Manipulator
motion planning combines nonlinear dynamics, torque bounds, and geometric
collision constraints. Contact-rich motions have hybrid dynamics, with
contact transitions and possible impact resets. A prescribed mode sequence
simplifies planning, while contact-implicit transcription allows contact
patterns to emerge from the optimization. Both require careful initialization,
contact-consistency checks, and appropriate handling of nonsmooth behavior.

No formulation removes nonconvexity. Initialization, discretization, scaling,
derivatives, and constraint handling remain decisive. Nevertheless, exploiting
the chain structure makes substantial nonlinear control problems tractable.
The next chapter places these finite-horizon solvers inside a feedback loop:
model predictive control repeatedly reoptimizes the trajectory as new state
measurements arrive.
\chapter{Model Predictive Control}
\label{chap:model-predictive-control}
\chapteroverview{Receding-horizon feedback, prediction and control horizons,
linear and nonlinear MPC, state and input constraints, terminal ingredients,
recursive feasibility, nominal stability, warm starts, state estimation,
real-time implementation, a robot braking example, representative
robotics applications, and whole-body NMPC for humanoid locomotion.}

Chapter~9 developed algorithms for computing finite state--control
trajectories. A computed trajectory is a plan based on an assumed initial
state, model, and set of constraints. During execution, however, a robot
may encounter an unexpected load, a moving obstacle, or a new target.
Executing the entire control sequence without correction ignores this
new information.

\emph{Model predictive control} (MPC) turns finite-horizon optimization into
feedback. At each sampling instant, the controller estimates the current
state and optimizes a sequence of future controls. It applies only the
first control, then measures again and solves a new problem over a shifted
horizon. This repeated \emph{plan--act--measure--replan} cycle is also called
receding-horizon control.

Replanning allows the robot to respond to changes in its state, task, and
environment. It can also compensate for observed deviations caused by
disturbances or model error. The effectiveness of this feedback depends
on sensing, prediction quality, computation time, and available control
authority; repeated optimization alone does not guarantee robustness.

MPC is therefore a control architecture, not a particular numerical solver.
The shooting, transcription, SQP, and iLQR methods of Chapter~9 can all serve
as its computational engine. The central questions are how far ahead to predict, how to preserve
feasibility and stability, and how to close the loop when computation and
prediction are imperfect.
Standard references are
\cite{mayne2000constrained,rawlings2017mpc}; the discussion below emphasizes
the connections to the preceding chapters and to robotic implementation.

\section{Receding-Horizon Control}

\subsection{Plan, Act, Measure, and Replan}

Consider a sampled model
\begin{equation}
  x_{k+1}=f(x_k,u_k),
  \qquad x_k\in\R^n,\quad u_k\in\R^m.
  \label{eq:mpc-plant-model}
\end{equation}
The sampling interval is denoted by $h$. As in Chapter~9, $f$ may be a
discrete-time model or a numerical transition map obtained from
continuous-time dynamics. We first explain the ideal setting, then account for state estimation,
disturbances, and computation delays.

At sample $k$, distinguish the actual state $x_k$ from a predicted state
$x_{i|k}$. The first index $i$ is the number of steps ahead, and the second
index identifies when the prediction was made. With an exactly known state, $x_{0|k}=x_k$. The quantity $x_{1|k}$
predicts the state at sample $k+1$, and $x_{N|k}$ predicts the state $N$
samples ahead. Predicted controls are denoted by $u_{i|k}$.

A representative MPC problem is
\begin{equation}
  \begin{aligned}
    \minimize_{X_k,U_k}\quad
      &\Phi(x_{N|k})+
       \sum_{i=0}^{N-1}\ell_{i|k}(x_{i|k},u_{i|k})\\
    \subjectto\quad
      &x_{0|k}=\hat x_k,\\
      &x_{i+1|k}=f_{i|k}(x_{i|k},u_{i|k}),
        \quad i=0,\ldots,N-1,\\
      &x_{i|k}\in\mathcal X_{i|k},
        \quad i=0,\ldots,N,\\
      &u_{i|k}\in\mathcal U_{i|k},
        \quad i=0,\ldots,N-1,\\
      &g_{i|k}(x_{i|k},u_{i|k})\leq0,
        \quad i=0,\ldots,N-1,\\
      &x_{N|k}\in\mathcal X_f.
  \end{aligned}
  \label{eq:mpc-finite-horizon-problem}
\end{equation}
Here $X_k=(x_{0|k},\ldots,x_{N|k})$ and
$U_k=(u_{0|k},\ldots,u_{N-1|k})$. The initial condition is the current state
estimate $\hat x_k$; in the ideal full-state setting, $\hat x_k=x_k$.
The prediction model $f_{i|k}$ can incorporate known future inputs or
time-varying dynamics. The cost and constraint sets may similarly include
reference or obstacle previews. Without such variation, $f_{i|k}=f$ and the
same stage cost and sets are used throughout the horizon. The terminal cost
$\Phi$ and terminal set $\mathcal X_f$ describe how the plan should end.
Some designs omit the terminal constraint; their feasibility and stability
then require other arguments.

Let $U_k^\star$ be a minimizing sequence. The executed control is
\begin{equation}
  u_k=u_{0|k}^\star.
  \label{eq:mpc-first-control}
\end{equation}
The controls $u_{1|k}^\star,\ldots,u_{N-1|k}^\star$ remain predictions;
they are not commitments to future actions. At sample $k+1$, MPC replaces
the initial condition by the new estimate and recomputes a sequence. In
general, $u_{0|k+1}^\star\neq u_{1|k}^\star$ because the state, reference,
or constraint information has changed.

\begin{figure}[t]
  \centering
  \begin{tikzpicture}[
    node distance=7mm,
    block/.style={draw=prismblue,rounded corners,align=center,
                  text width=0.17\textwidth,minimum height=12mm},
    >=Stealth]
    \node[block] (estimate) {Estimate\\$\hat x_k$};
    \node[block,right=of estimate] (optimize) {Optimize\\$N$-step plan};
    \node[block,right=of optimize] (apply) {Apply\\$u_{0|k}^\star$};
    \node[block,right=of apply] (robot) {Robot\\$x_{k+1}$};
    \draw[->] (estimate) -- (optimize);
    \draw[->] (optimize) -- (apply);
    \draw[->] (apply) -- (robot);
    \draw[->] (robot.south) -- ++(0,-9mm)
      -| node[pos=0.25,below] {new measurements at the next sample}
      (estimate.south);
  \end{tikzpicture}
  \caption{The MPC feedback loop. Only the first optimized control is
  executed before the state estimate and plan are updated.}
  \label{fig:mpc-feedback-loop}
\end{figure}
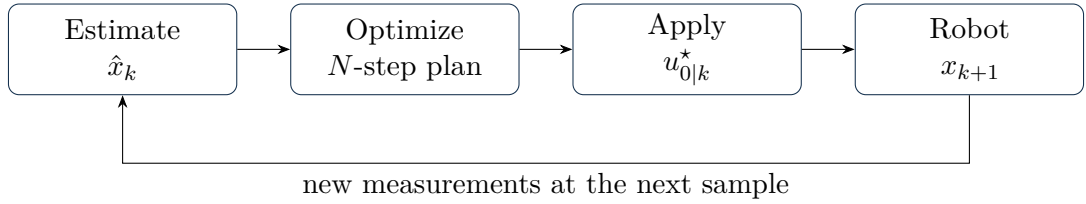

An ideal MPC cycle has four steps:
\begin{enumerate}
  \item obtain a state estimate and update the reference and environment;
  \item solve the finite-horizon problem from that estimate;
  \item apply its first control over the next sampling interval;
  \item obtain new measurements, shift the horizon, and repeat.
\end{enumerate}
Chapter~9 explains how to perform the second step. The remaining steps make
the optimizer part of a closed-loop system.

\subsection{Prediction and Control Horizons}

The \emph{prediction horizon} is the number $N$ of future model transitions.
With uniform sampling, it corresponds to physical duration $T_p=Nh$.
Increasing $N$ while keeping $h$ fixed lets the controller see farther ahead,
but increases the optimization size. Decreasing $h$ while keeping $T_p$
fixed improves temporal resolution but also increases the number of stages.
These are different design changes and should not be confused.

The horizon should capture the important consequences of the first action.
For braking, this includes the time needed to dissipate momentum. For
manipulator motion, it may include a turn around an obstacle or a future
actuator limitation. A horizon that is too short can make a locally cheap
action attractive even when it leads toward a difficult future state.
A terminal cost helps represent consequences beyond the horizon, but an
arbitrary terminal weight does not automatically provide a correct long-term
model.

The \emph{control horizon} $N_u\leq N$ describes how many future control
moves are independently parameterized. One simple restriction is
\begin{equation}
  u_{i|k}=u_{N_u-1|k},
  \qquad i=N_u,\ldots,N-1.
  \label{eq:mpc-control-horizon}
\end{equation}
Other parameterizations use piecewise-constant blocks or a prescribed local
feedback law after the independently optimized moves. This reduces the
number of decision variables, but restricts the feasible control sequences.
It can therefore change both performance and feasibility.

The control horizon is not the number of controls executed before measuring
again. Standard MPC still executes only one move. Executing several moves
between optimization updates is possible, but reduces the frequency at which
new information enters the feedback loop.

Finally, distinguish a \emph{receding horizon} from a \emph{shrinking horizon}.
In receding-horizon regulation, $N$ usually stays fixed and the terminal
sample moves from $k+N$ to $k+1+N$. For a task with a fixed deadline $k_f$,
the horizon may instead shrink as $N_k=k_f-k$. The latter keeps the same
terminal time and is a different finite-horizon feedback problem.

\subsection{MPC as Feedback}

For a fixed model, reference, and constraint specification, the optimizer
defines an implicit state-feedback law, denoted by $\kappa_N$. Given a
state $x$, this law returns the first optimized input from the
finite-horizon problem initialized at $x$:
\begin{equation}
  \kappa_N(x)=u_{0}^\star(x),
  \qquad x_{k+1}=f\bigl(x_k,\kappa_N(x_k)\bigr).
  \label{eq:mpc-implicit-feedback}
\end{equation}
Here $N$ is the horizon length, and the index $0$ identifies the first
input in the optimized sequence. The policy is defined wherever the
optimization is feasible, with a selection rule if the minimizer is not
unique. Although each prediction may use an open-loop control sequence,
the implemented controller is feedback: each new input depends on the
latest state through a new optimization.

This distinction connects MPC to Chapter~8. Dynamic programming seeks a
value function and feedback law over a state domain. MPC instead evaluates
a finite-horizon optimization at the states actually encountered. It avoids
constructing a global state-space grid, at the cost of online computation
and an approximation of the consequences beyond the horizon.

\paragraph{The linear--quadratic special case.}
For regulation of
$x_{i+1|k}=Ax_{i|k}+Bu_{i|k}$, consider
\begin{equation}
  J_N
  =\frac12x_{N|k}\trans P_f x_{N|k}
   +\frac12\sum_{i=0}^{N-1}
     \left(x_{i|k}\trans Qx_{i|k}+u_{i|k}\trans Ru_{i|k}\right),
  \quad Q\succeq0,\quad R\succ0.
  \label{eq:mpc-linear-quadratic-cost}
\end{equation}
Without constraints, the Riccati recursion in
\cref{eq:discrete-time-riccati-recursion} gives
$u_{i|k}^\star=-K_i x_{i|k}^\star$. With a fixed horizon and time-invariant
data, MPC repeatedly applies the \emph{first} gain $K_0$:
\begin{equation}
  u_k=-K_0x_k.
  \label{eq:mpc-unconstrained-lqr-feedback}
\end{equation}
It does not successively apply $K_0,K_1,\ldots$ from an old finite-horizon
plan. If $P_f=P_\infty$ is the stabilizing DARE solution from
\cref{eq:discrete-time-algebraic-riccati-equation}, the Riccati recursion
stays at that fixed point, and every gain equals the infinite-horizon LQR
gain. In this case unconstrained MPC reproduces LQR for any $N\geq1$.

\section{Constraints and Stability}

\subsection{State and Input Constraints}

MPC imposes restrictions on the entire predicted motion. Input constraints
represent actuator limits; state constraints represent joint ranges,
speed limits, obstacle clearance, or admissible contact configurations.
Mixed constraints couple state and input. For example, a motor's available
torque may decrease with joint speed because of voltage limits. The
optimizer must account for these future restrictions when choosing the
current input.

Including actuator limits during optimization is more informative than
saturating an unconstrained command afterward. Clipping a planned force
changes the motion on which future constraint predictions were based.
An optimizer that includes the limit can instead plan braking or rerouting
before the available control authority becomes insufficient.

\paragraph{Hard constraints and soft constraints.}
A hard constraint must be satisfied by every accepted solution. A soft
constraint permits an explicit violation through a slack variable. For
example, a scalar performance constraint $g(x,u)\leq0$ can be replaced by
\begin{equation}
  g(x_{i|k},u_{i|k})\leq s_{i|k},
  \qquad s_{i|k}\geq0,
  \label{eq:mpc-soft-constraint}
\end{equation}
with the stage-cost addition
\begin{equation}
  \rho_1 s_{i|k}+\frac12\rho_2 s_{i|k}^2,
  \qquad \rho_1>0,\quad\rho_2\geq0.
  \label{eq:mpc-slack-cost}
\end{equation}
Slack can prevent a tracking or comfort specification from making the
problem infeasible, but it does not relax the remaining hard constraints.
A positive slack represents a planned violation of the original requirement.
Physical actuator limits should remain hard; softening collision constraints
likewise changes the meaning of a feasible solution.

\paragraph{Input increments.}
Rapid changes in the control input may excite unmodeled dynamics. Define
\begin{equation}
  \Delta u_{i|k}=u_{i|k}-u_{i-1|k},
  \qquad u_{-1|k}=u_{k-1}^{\mathrm{applied}}.
  \label{eq:mpc-input-increments}
\end{equation}
Penalizing $\Delta u_{i|k}$ discourages abrupt command changes; bounding it
imposes an explicit limit on the change between successive inputs. The first
increment must be measured relative to the input actually applied during the
previous interval, not a planned command that was never executed.

\paragraph{Predicted feasibility is not physical safety.}
Feasibility in \cref{eq:mpc-finite-horizon-problem} does not guarantee safety
between samples or in the presence of model error, estimation error, and
delays. Constraint margins, intersample checks, and supervisory safeguards
therefore require separate consideration.

\subsection{Terminal Costs and Terminal Sets}

The finite horizon omits everything after its final predicted state. A
terminal cost $\Phi$ estimates the consequences of that omitted future.
A terminal set $\mathcal X_f$ can require the prediction to finish in a
region where a known local controller can continue safely.

To make this idea precise, consider time-invariant regulation with the target
shifted to the origin, $f(0,0)=0$. Use fixed state and input sets
$\mathcal X$ and $\mathcal U$, fixed mixed constraints $g(x,u)\leq0$,
a nonnegative stage cost $\ell$, and
$\ell(0,0)=\Phi(0)=0$. Suppose a terminal controller $\kappa_f$ satisfies
\begin{align}
  x&\in\mathcal X_f\subseteq\mathcal X
    \ \Longrightarrow\ \kappa_f(x)\in\mathcal U,
    \quad g\bigl(x,\kappa_f(x)\bigr)\leq0,
  \label{eq:mpc-terminal-admissibility}\\
  f\bigl(x,\kappa_f(x)\bigr)&\in\mathcal X_f,
  \qquad x\in\mathcal X_f,
  \label{eq:mpc-terminal-invariance}\\
  \Phi\bigl(f(x,\kappa_f(x))\bigr)-\Phi(x)
    &\leq-\ell\bigl(x,\kappa_f(x)\bigr),
  \qquad x\in\mathcal X_f.
  \label{eq:mpc-terminal-decrease}
\end{align}
The first two conditions mean that the terminal region can be kept
admissible by the local controller. The last condition means the terminal
cost pays for at least the next stage of that continuation. These are sufficient conditions for the nominal argument below.
Other MPC designs can establish feasibility or stability without an
explicit terminal set.

\paragraph{Constructing terminal ingredients from LQR.}
For the linear system, let $P_\infty$ solve the DARE and let $K_\infty$ be
the corresponding gain. Choose
\begin{equation}
  \Phi(x)=\frac12x\trans P_\infty x,
  \qquad \kappa_f(x)=-K_\infty x.
  \label{eq:mpc-lqr-terminal-design}
\end{equation}
The DARE implies the closed-loop identity
\begin{equation}
  (A-BK_\infty)\trans P_\infty(A-BK_\infty)-P_\infty
  =-Q-K_\infty\trans R K_\infty.
  \label{eq:mpc-lqr-terminal-identity}
\end{equation}
Multiplying by $x\trans/2$ and $x$ gives
\cref{eq:mpc-terminal-decrease} with equality for the quadratic stage cost.
The remaining task is to choose a positively invariant terminal set in which
the state and input constraints hold.

A sufficiently small ellipsoid
$\mathcal X_f=\{x:x\trans P_\infty x\leq c\}$ is one possible choice when
$P_\infty\succ0$ and the constraints admit a neighborhood of the origin.
An invariant polytope is another option. An ellipsoidal terminal constraint
is a quadratic inequality, so the resulting optimization is not a pure QP
with only affine constraints. For a nonlinear robot, a local LQR design is
a useful starting point, but admissibility, invariance, and cost decrease
must be verified for the nonlinear transition map, not only its
linearization.

\paragraph{Why a terminal weight matters.}
Repeated optimization alone does not ensure stability. Consider the scalar
system $x^+=ax+u$ with $a>1$, prediction horizon $N=1$, and cost
\begin{equation}
  J_1(x,u)=\frac12qx^2+\frac12ru^2,
  \qquad q,r>0.
  \label{eq:mpc-short-horizon-counterexample}
\end{equation}
There is no penalty on the next state, so the unique minimizing input is
$u=0$. Replanning this same problem at every sample leaves the unstable
dynamics $x^+=ax$ unchanged.

Adding the terminal cost $\Phi(x^+)=s(x^+)^2/2$, with $s\geq0$, gives
\begin{equation}
  u^\star=-\frac{as}{r+s}x,
  \qquad x^+=\frac{ar}{r+s}x.
  \label{eq:mpc-scalar-terminal-feedback}
\end{equation}
The closed loop is stable precisely when $s>r(a-1)$ in this example.
This calculation illustrates why the tail of the prediction matters; it
does not supply a universal terminal-weight rule for nonlinear constrained
systems. A large arbitrary terminal penalty is not a substitute for checking
the required terminal properties.

\subsection{Recursive Feasibility and Stability Intuition}

\emph{Recursive feasibility} means that a feasible MPC problem remains
feasible after the next control is applied. Under an exact model and fixed
constraints, the terminal conditions above provide a simple explanation:
remove the input just executed, shift the remaining plan forward, and append
the terminal control. The predicted next state is the actual next state,
and terminal invariance makes the appended step admissible. This gives a
feasible candidate at the next sample.

The terminal decrease condition also makes the candidate's cost no larger
than the previous optimal cost minus the stage cost just incurred. The
new optimum can only improve on this candidate. With the usual positivity
and regularity conditions, this decrease supports nominal asymptotic
stability. These are sufficient conditions, rather than the only way to
design a stabilizing MPC controller. Disturbances, estimation errors,
changing constraints, and incomplete solves require additional analysis
\cite{mayne2000constrained,rawlings2017mpc}.

\section{Practical MPC}

\subsection{Linear and Nonlinear MPC}

With linear or affine prediction dynamics, a quadratic objective, and
affine constraints, the finite-horizon problem is a quadratic program
(QP). Positive-semidefinite state weights and positive-definite input
weights give a convex objective. The QP can retain the predicted states
as decision variables or eliminate them using the dynamics, as in
Chapter~9. An ellipsoidal terminal set introduces a quadratic constraint,
so it falls outside this affine-constraint QP form.

\emph{Nonlinear MPC} (NMPC) retains nonlinear prediction dynamics or
constraints. Robot kinematics, configuration-dependent inertia, and
collision geometry make this formulation natural, but the resulting
problem is generally nonconvex. SQP solves a sequence of local QP
approximations; iLQR and DDP exploit the trajectory structure through
forward and backward passes. These solvers usually seek a local solution,
whose quality depends on initialization and the available computation time.

The receding-horizon architecture is the same in both cases. The choice
is how much of the robot's dynamics and geometry the online optimization
can represent within its timing budget.

\subsection{Warm Starts and Solver Timing}

Successive MPC problems share most of their horizon. The previous solution
is therefore a natural starting point for the next solve. A typical control
warm start is
\begin{equation}
  U_{k+1}^{\mathrm{init}}
  =\bigl(u_{1|k}^\star,\ldots,u_{N-1|k}^\star,
          u_{\mathrm{tail}}\bigr),
  \label{eq:mpc-control-warm-start}
\end{equation}
where $u_{\mathrm{tail}}$ may be a terminal feedback command. Simply
repeating the last input is convenient but need not provide a feasible tail.
The terminal-controller extension used in the nominal feasibility
argument is one principled choice.

The predicted states should also be updated. In shooting, initialize
$x_{0|k+1}=\hat x_{k+1}$ and roll out the shifted controls. In transcription,
one may shift the old states and adjust the initial condition, accepting
temporary defects that the solver must close. Multipliers and active-set
information can be shifted as well. A warm start is a numerical aid, not
a guarantee that the changed problem remains feasible.

\paragraph{The computation deadline.}
Let $T_{\mathrm{est}}$, $T_{\mathrm{solve}}$, and $T_{\mathrm{comm}}$ denote
the estimation, optimization, and command-communication times. A simple
synchronous implementation needs a timing budget such as
\begin{equation}
  T_{\mathrm{est}}+T_{\mathrm{solve}}+T_{\mathrm{comm}}
  <h-T_{\mathrm{margin}}.
  \label{eq:mpc-computation-deadline}
\end{equation}
Here $T_{\mathrm{margin}}>0$ reserves time for variability and other tasks.
Average solve time is not enough: occasional overruns can cause the input
to be based on an outdated state. Timing should be evaluated under difficult
initializations, active constraints, and changing references, not only near
an easy equilibrium.

The plant continues moving while the controller computes. Let $t_m$ be
the time of the latest measurement and $t_a$ the expected actuation time.
Propagate $\hat x(t_m)$ to $t_a$ using the model and the input history,
including commands scheduled to remain active during computation. Initialize
the new prediction from this propagated estimate, and align references and
obstacle forecasts with $t_a,t_a+h,\ldots$. Significant actuator or
communication delays should be included in the model.

In an asynchronous implementation, each solution should carry timestamps
and identify the state and reference data used to compute it. A solution
that arrives too late may need to be updated or discarded in favor of a
specified fallback. Even an accurately solved optimization can give poor
commands if its prediction and actuation timelines disagree
\cite{rawlings2017mpc,diehl2005realtime}.

\paragraph{Limited-iteration optimization.}
For nonlinear MPC, it may be more useful to perform a small, predictable
number of structured solver steps than to converge an entire new nonlinear
program at every sample. Real-time iteration schemes coordinate successive
Newton or SQP updates with the evolving plant. A preparation phase can form
derivatives and local models before the newest state is available, followed
by a shorter feedback update when the measurement arrives
\cite{diehl2005realtime}. Their effectiveness depends on suitable initialization and regularity;
a fixed iteration count does not by itself ensure an adequate solution.

The distinction between the ideal and implemented controller is important.
An exact optimizer defines the state feedback $\kappa_N(x)$. A warm-started,
limited-iteration implementation also depends on stored trajectories,
multipliers, and solver state. Its closed-loop behavior is therefore a
property of the plant--optimizer combination, not only of the mathematical
finite-horizon problem.

\paragraph{Failure handling.}
Before accepting a solution, check actuator limits, constraint residuals,
dynamic consistency, and timely availability. A low objective value or a
solver success flag alone does not establish that the command is usable. The implementation needs
a specified response to infeasibility, numerical failure, and missed
deadlines. Possible responses include a verified remaining plan, a local
backup controller within its admissible region, or a controlled stop.
None is automatically safe from every state. In particular, holding the
last command or clipping a newly returned command is not a general
feasibility-preserving fallback.

\subsection{State Estimation and Model Mismatch}

Robots typically measure joint positions, inertial signals, images, or other
outputs rather than the complete state. A simple uncertainty model is
\begin{equation}
  x_{k+1}=f(x_k,u_k)+w_k,
  \qquad y_k=h_y(x_k)+v_k^{\mathrm{meas}},
  \label{eq:mpc-uncertain-plant}
\end{equation}
where $w_k$ represents process disturbances or model mismatch, and
$v_k^{\mathrm{meas}}$ is measurement noise. An observer or filter produces
$\hat x_k$ for the MPC initial condition. Chapter~7 introduced the basic
observer and Kalman-filter ideas; their role here is to keep the prediction
anchored to the estimated physical state.

Replanning can correct deviations from an earlier prediction, but it does
not make the model exact. A persistent load can repeatedly shift the actual
motion away from the nominal one. Velocity-estimation error can make the
controller underestimate stopping distance. Unmodeled actuator dynamics
can make a command change take effect later than predicted. These errors
matter most when the nominal plan runs close to a constraint boundary.

Using $\hat x_k$ in place of $x_k$ is a certainty-equivalent implementation.
It ignores uncertainty when constructing the nominal prediction. The LQG
separation principle does not automatically extend to constrained nonlinear
MPC: estimation error can change which constraints are active or whether
the planned motion is safe. The estimator and controller should therefore be evaluated together.

\paragraph{What to monitor.}
Useful closed-loop diagnostics include prediction errors, constraint
margins, slack values, solver residuals, and timing. For example,
$\hat x_{k+1}-x_{1|k}$ measures the disagreement between the next estimate
and the previous one-step prediction when their timestamps are aligned. Large errors can indicate a poor
model, an unreliable estimator, or incorrect timing. They should not be
interpreted solely as a need for larger tracking weights.

\subsection{A Braking Example}

Consider a cart moving toward a target at position $p_g$ while keeping a
safe distance $d_{\mathrm{safe}}$ from a wall. Its state is position and
velocity, $x=(p,v)$, and its input is acceleration $a$. At sample $k$, let
$p_{\mathrm{wall},i|k}$ be the predicted wall position $i$ steps ahead.
The controller solves
\begin{equation}
  \begin{aligned}
    \minimize_{X_k,U_k}\quad
      &\frac12 e_{N|k}\trans P_f e_{N|k}
       +\frac12\sum_{i=0}^{N-1}
        \left[q_p(p_{i|k}-p_g)^2+q_vv_{i|k}^2+r_aa_{i|k}^2\right]\\
    \subjectto\quad
      &p_{0|k}=\hat p_k,\qquad v_{0|k}=\hat v_k,\\
      &p_{i+1|k}=p_{i|k}+hv_{i|k}+\frac12h^2a_{i|k},\\
      &v_{i+1|k}=v_{i|k}+ha_{i|k},
        \quad i=0,\ldots,N-1,\\
      &-a_{\max}\leq a_{i|k}\leq a_{\max},
        \quad i=0,\ldots,N-1,\\
      &p_{i|k}\leq p_{\mathrm{wall},i|k}-d_{\mathrm{safe}},
        \quad i=0,\ldots,N.
  \end{aligned}
  \label{eq:mpc-cart-qp}
\end{equation}
Here $e_{N|k}=(p_{N|k}-p_g,v_{N|k})$,
$P_f\succeq0$, and $q_p,q_v,r_a>0$. The quadratic objective and affine dynamics and constraints make this
a convex QP. No terminal safe set is imposed here, so feasibility over the
prediction horizon does not establish that a safe continuation exists
afterward.

The wall forecast is updated from each new measurement. For example, a
constant-velocity forecast is
\begin{equation}
  p_{\mathrm{wall},i|k}
  =\hat p_{\mathrm{wall},k}+ih\hat v_{\mathrm{wall},k}.
  \label{eq:mpc-cart-wall-forecast}
\end{equation}
If the wall moves closer, the predicted-clearance constraints tighten and
the new solution may brake earlier or retreat. MPC applies only the first
acceleration, then updates the cart state and wall forecast and solves
again. This reactivity comes from using new measurements at each cycle; it
does not guarantee that the cart can always avoid a collision.

For a stationary wall and a cart moving toward it, the minimum stopping
distance under maximum braking is
\begin{equation}
  d_{\mathrm{stop}}=\frac{v^2}{2a_{\max}},
  \qquad v>0.
  \label{eq:mpc-cart-stopping-distance}
\end{equation}
At $v=1\,\mathrm{m/s}$ and $a_{\max}=2\,\mathrm{m/s^2}$, the cart needs
$0.25\,\mathrm{m}$ to stop. If only $0.15\,\mathrm{m}$ remains before the clearance boundary
$p_{\mathrm{wall}}-d_{\mathrm{safe}}$, no admissible input can preserve
the required clearance. Whether the cart actually hits the wall also
depends on $d_{\mathrm{safe}}$.
Replanning cannot create braking authority. For a moving wall, its predicted
motion must also be included. In either case, clearance must be checked
between sampling instants as well as at the predicted samples.

\subsection{Outlook: Robust and Stochastic MPC}

Nominal MPC predicts one trajectory from one state estimate. Robust MPC
instead accounts explicitly for an assumed set of disturbances. In a
tube-based construction, a nominal trajectory is accompanied by local
feedback and a bounded error region; nominal constraints are tightened so
that the physical state and input remain admissible for the allowed errors.
The resulting protection is conditional on the disturbance bounds and
the validity of the model and error analysis.

Stochastic MPC models uncertainty probabilistically. It may replace a
hard prediction constraint by a chance constraint such as
\begin{equation}
  \Prob\bigl(g(x_{i|k},u_{i|k})\leq0\bigr)\geq1-\varepsilon.
  \label{eq:mpc-chance-constraint}
\end{equation}
The probability model and risk level are part of the design. A marginal
constraint at one stage is not automatically a joint guarantee over every
stage or over indefinitely repeated operation. The specified risk level permits violations with nonzero probability.

These extensions add uncertainty handling to the same receding-horizon
architecture; they are distinct from merely replanning a nominal problem
more frequently. A detailed treatment is beyond this chapter; see
\citet{rawlings2017mpc}. Their guarantees depend on the stated uncertainty model, disturbance
bounds, or probability assumptions.

\section{Representative MPC Applications in Robotics}
\label{sec:mpc-robotics-papers}

The following papers provide representative hardware demonstrations of
MPC in robotics. They illustrate different choices of prediction model,
optimization method, and low-level control architecture rather than one
universally best formulation.
\begin{itemize}
  \item \emph{Quadruped locomotion with convex MPC.}
    \citet{dicarlo2018dynamic}, \emph{Dynamic Locomotion in the MIT Cheetah 3
    Through Convex Model-Predictive Control}. The authors demonstrate dynamic
    quadruped locomotion by optimizing ground-reaction forces with a
    simplified body model and a prescribed contact schedule. This is a
    useful example of making a demanding control problem tractable through
    model reduction and a convex QP.

  \item \emph{Whole-body NMPC.}
    \citet{khazoom2024tailoring}, \emph{Tailoring Solution Accuracy for Fast
    Whole-Body Model Predictive Control of Legged Robots}, demonstrate
    coordinated humanoid motion using constrained NMPC with limited solver
    accuracy. Section~\ref{sec:mpc-whole-body-nmpc} develops this example.

  \item \emph{Whole-body contact-implicit MPC.}
    \citet{kim2025contactimplicit}, \emph{Contact-Implicit Model Predictive
    Control: Controlling Diverse Quadruped Motions Without Pre-Planned
    Contact Modes or Trajectories}, use full-body dynamics and
    contact-implicit DDP to optimize motion without a fixed contact schedule.
    Hardware experiments on the KAIST HOUND quadruped include forward
    trotting and front-leg rearing. This contrasts with force planning under
    a prescribed gait: contact choices emerge during optimization, with
    cost terms guiding the resulting motion.

  \item \emph{Torque-controlled legged locomotion with nonlinear MPC.}
    \citet{grandia2019feedback}, \emph{Feedback MPC for Torque-Controlled
    Legged Robots}. The authors demonstrate motion tracking on the ANYmal quadruped
    using nonlinear MPC with a local feedback policy. The paper illustrates how the
    optimizer\textquotesingle s local feedback policy bridges the gap
    between slower planning updates and faster torque-control updates.

  \item \emph{Quadrotor flight with perception-aware MPC.}
    \citet{falanga2018pampc}, \emph{PAMPC: Perception-Aware Model Predictive
    Control for Quadrotors}. The authors combine motion and visual-perception objectives
    and demonstrate the controller in real flight.
    This example shows that an MPC cost can account for sensing quality,
    not only position tracking and control effort.

  \item \emph{Robotic manipulation with a learned prediction model.}
    \citet{lenz2015deepmpc}, \emph{DeepMPC: Learning Deep Latent Features for
    Model Predictive Control}. The authors use a learned dynamics model within MPC
    for robotic food cutting. The paper is a useful example when the
    interaction dynamics are difficult to model analytically: learning
    supplies the predictor, while online optimization chooses the action.

  \item \emph{Autonomous racing with contouring MPC.}
    \citet{liniger2015racing}, \emph{Optimization-Based Autonomous Racing of
    1:43 Scale RC Cars}. The authors demonstrate real-time racing control on miniature cars. Their model predictive contouring approach balances progress
    along a track with path error and physical constraints, illustrating
    how MPC can optimize task progress rather than merely track a fixed
    time-indexed trajectory.

\end{itemize}

When reading these papers, identify what the optimizer chooses, which
dynamics and constraints it includes, and what remains the responsibility
of estimation, planning, and low-level feedback. Hardware results establish
performance under the tested conditions; they do not replace feasibility
or stability analysis.

\section{Whole-Body NMPC for Humanoid Locomotion}
\label{sec:mpc-whole-body-nmpc}

A walking humanoid must move in a commanded direction while keeping its
body balanced, placing its feet, and respecting the limits of its joints
and contacts. These requirements are coupled: moving a leg changes both
the available support and the motion of the rest of the robot. This section
explains how they become a finite-horizon optimal-control problem, then
connects that formulation to the implementation of
\citet{khazoom2024tailoring}.

\subsection{The Control Task and the Meaning of Whole-Body NMPC}

\emph{Nonlinear model predictive control} (NMPC) repeatedly solves an
optimal-control problem whose prediction model or constraints are
nonlinear. The letter ``N'' in NMPC stands for \emph{nonlinear}; the symbol
$N$ used below denotes the number of prediction intervals. At each update,
the controller initializes the prediction from the estimated state and
optimizes a future motion. It executes only the first control before
replanning. For interval lengths $\Delta t_i$, the prediction duration is
$T_p=\sum_{i=0}^{N-1}\Delta t_i$. Unlike the controller-update period, this
horizon can span many controller updates.

Here, \emph{whole-body} describes the model and motion variables: the
prediction includes the torso's motion in space, the individual joint
motions, and their mechanical coupling. Thus an arm motion, a leg motion,
and a change in torso orientation are considered in the same optimization.
This contrasts with a \emph{reduced-order} prediction, such as a
point-mass or single-rigid-body model, which can describe overall balance
without predicting every joint trajectory. Whole-body prediction supports
direct reasoning about reachability, joint limits, limb inertia, and
self-collision, while still relying on modeling approximations. It does not
imply that every flexible or electrical effect is modeled.

For locomotion, the command may specify forward and lateral body velocity
and a turning rate. The controller should follow that command while
maintaining a suitable torso height and orientation and producing feasible
steps. A nominal joint posture helps select among many possible motions.
It is a preference, not a demand that every joint remain fixed while
walking. The optimizer can trade posture error against body tracking and
other costs, subject to the physical constraints. Recovering from a push
is the same control task initialized from a newly disturbed state.

\subsection{Configuration, Velocity, and Actuation}

Unlike a manipulator bolted to a table, a humanoid has no permanently fixed
base. Its torso can translate and rotate. We therefore use a
\emph{floating-base} configuration. For $n_j$ actuated joints, choose a
local orientation chart with three coordinates and write the configuration
and generalized velocity as
\[
  q=(p_b,\theta_b,q_j),
  \qquad
  v=(\dot p_b,\omega_b,\dot q_j),
  \qquad
  x=(q,v).
\]
Here $p_b\in\R^3$ is the base position, $\theta_b\in\R^3$ is a local
coordinate vector parameterizing its orientation (for example, Euler angles
in a chosen chart), $q_j\in\R^{n_j}$ contains the joint angles, and
$\omega_b\in\R^3$ is the base angular-velocity vector. Thus, $q$ and $v$
each have $6+n_j$ components in this representation. The six base degrees
of freedom describe motion in space; the remaining coordinates describe
the articulated body.

The distinction between $v$ and $\dot q$ matters: angular velocity is
generally not the derivative of the orientation coordinates. For example,
Euler-angle rates must be transformed to obtain $\omega_b$. The map $E(q)$,
which depends on the orientation chart and angular-velocity convention,
converts generalized velocities to coordinate derivatives: $\dot q=E(q)v$.
With a unit quaternion, replace $\theta_b$ by $q_b\in\R^4$ satisfying
$\|q_b\|_2=1$. Then $q$ has
$7+n_j$ entries (but still only $6+n_j$ configuration degrees of freedom),
while $v$ has $6+n_j$ components; accordingly, $E(q)$ is rectangular.
The equations below allow either representation, provided orientation is
integrated consistently.

The motors supply joint torques $\tau\in\R^{n_j}$. They do not directly
command the six base coordinates. Ground contacts provide additional
external forces $F_c\in\R^3$, indexed by candidate contact points
$c\in\mathcal C$. Multiple points on one foot can represent both a net
support force and a moment. The controller must find torques and contact
forces that can jointly produce the desired motion.

\subsection{Why the Equations of Motion Have This Form}

For a rigid articulated robot, the equations are
\begin{equation}
  M(q)\dot v+b(q,v)
  =S\trans\tau+\sum_{c\in\mathcal C}J_c(q)\trans F_c,
  \qquad \dot q=E(q)v.
  \label{eq:mpc-whole-body-dynamics}
\end{equation}
This is Newton's law expressed in generalized coordinates. In one vertical
translation it reduces to $m\ddot z+mg=F_z$. For an articulated robot, a
matrix replaces the scalar mass, and generalized forces include both
translational forces and rotational torques. The terms have the following
roles.

\paragraph{Inertia and the forces on the left-hand side.}
The mass matrix $M(q)\in\R^{(6+n_j)\times(6+n_j)}$ describes the inertia of
all modeled links and their coupling. The kinetic energy is
$\tfrac12v\trans M(q)v$. Its dependence on configuration explains why
extending a leg or moving an arm changes the robot's response to a torque.
The vector $b(q,v)$ collects gravity and velocity-dependent terms, including
Coriolis and centrifugal effects. In a common notation,
$b(q,v)=C(q,v)v+g(q)$. These terms arise when the link equations of motion
are assembled, or equivalently from the kinetic and potential energies.
No new optimization principle is involved: this equation is the mechanical
constraint that admissible trajectories must satisfy.

\paragraph{Why the actuator term contains a selection matrix.}
With the base coordinates first, the matrix
\[
  S=\begin{bmatrix}0_{n_j\times6}&I_{n_j}\end{bmatrix}
  \quad\hbox{gives}\quad
  S\trans\tau=\begin{bmatrix}0_6\\\tau\end{bmatrix}.
\]
The zeros express the absence of a motor directly translating or rotating
the floating base. Joint torques can nevertheless affect base motion
through the coupled dynamics. The system is \emph{underactuated}: it has
more generalized velocity components than independent motor torques.
Internal limb motion can redistribute angular momentum between links, but it cannot
provide the external force needed to change the center-of-mass acceleration
beyond that due to gravity and contact.

\paragraph{Why a contact force appears as $J_c\trans F_c$.}
Let $p_c(q)$ be the world position of a contact point. Its velocity is
$\dot p_c=J_c(q)v$, where
$J_c=(\partial p_c/\partial q)E(q)$ has three rows. Force and velocity must
give the same mechanical power in either coordinate system:
\[
  F_c\trans\dot p_c
  =F_c\trans J_c v
  =(J_c\trans F_c)\trans v.
\]
Therefore $J_c\trans F_c$ is the generalized force associated with the
contact. The sum accounts for all active points; inactive points have zero
force. This also explains why foot placement matters: changing the contact
location changes how a ground force acts on the robot.

The nonlinear dependence of $M$, $b$, and $J_c$ on the state is one reason
the resulting controller is NMPC. Even a quadratic tracking cost does not
make an optimal-control problem with these dynamics a linear--quadratic
problem.

\subsection{Contacts, Gait Information, and Motion Constraints}

A contact force is an optimization variable, but it is not an independently
commanded actuator. The predicted force must be consistent with the
torques, motion, and contact conditions. For a nonadhesive contact with a
stationary support surface, a common model requires
\begin{equation}
  F_{c,n}\geq0,
  \qquad \|F_{c,t}\|_2\leq\mu F_{c,n},
  \label{eq:mpc-whole-body-friction}
\end{equation}
where $F_{c,n}$ is the force along the outward surface normal,
$F_{c,t}\in\R^2$ contains the tangential components, and $\mu$ is the
friction coefficient. The first inequality says that the ground can push
but cannot pull the foot. The second limits tangential force relative to
normal load. A proposed horizontal acceleration that needs too much
friction must be changed by the optimizer.

The controller also needs to know which points are in \emph{stance}
(supporting the robot) and which are in \emph{swing} (moving to the next
support location). A prescribed schedule gives a binary parameter
$\gamma_{c,i}$ for each point and prediction stage: one for stance and zero
for swing. During stance, the contact point stays on the support surface
with zero tangential velocity. During swing, its contact force is zero,
and a height reference can guide it above the ground, while a clearance
constraint can enforce a minimum height.
A prescribed contact schedule specifies \emph{when} support changes; it
need not prescribe the horizontal location of every future footstep.
Optimizing contact timing itself would introduce an additional decision
that this scheduled-contact formulation does not make.

Further constraints bound joint angles, joint velocities, and torques.
Self-collision constraints can require
$d_{ab}(q)\geq d_{\min}$ for selected pairs of links, where $d_{ab}$ is their
separation and $d_{\min}$ is the required clearance. These geometric
constraints depend on the joint configuration, which is available in a
whole-body prediction. They are particularly relevant when a recovery step
passes one leg close to the other. The contact model describes the planned
stance and swing phases; impact effects at touchdown require an appropriate
transition model or approximation as well.

\subsection{The Finite-Horizon Optimization}

At one controller update, suppress the update index and let $i$ index the
future stages. Define $x_i=(q_i,v_i)$ and stack the contact forces as $F_i$.
The decision variables are
\[
  Z=\bigl(\{x_i\}_{i=0}^{N},
          \{\tau_i,F_i\}_{i=0}^{N-1}\bigr).
\]
The estimated initial state, command, model parameters, and contact schedule
are data. Both the motion and the forces required to produce it are
unknowns.

A representative stage cost is
\begin{equation}
  \ell_i
  =\|e_{q,i}\|_{Q_q}^2
   +\|e_{v,i}\|_{Q_v}^2
   +\|F_i-F_i^{\mathrm{ref}}\|_{R_F}^2,
  \qquad \|z\|_W^2=z\trans Wz.
  \label{eq:mpc-whole-body-stage-cost}
\end{equation}
The errors $e_{q,i}$ and $e_{v,i}$ compare the predicted configuration and
velocity with their references; orientation error is expressed in suitable
local rotation coordinates. The reference configuration can combine a
desired body height and orientation with a nominal joint posture, while
the velocity reference encodes the commanded translation and turning.
A force reference can favor a reasonable load distribution among stance
contacts. The positive-semidefinite weights express the relative
importance of these preferences. A quadratic cost does not prescribe the
motion in advance: the dynamics and constraints determine how closely its
competing references can be followed.

Let $D_i(x_i,x_{i+1},\tau_i,F_i)=0$ denote a discretization of
\cref{eq:mpc-whole-body-dynamics}. In a first-order discretization, the
acceleration is approximated by
\[
  \dot v_i\approx\frac{v_{i+1}-v_i}{\Delta t_i},
\]
giving the force-balance constraint
\[
  M(q_i)\frac{v_{i+1}-v_i}{\Delta t_i}+b(q_i,v_i)
  =S\trans\tau_i+\sum_{c\in\mathcal C}J_c(q_i)\trans F_{c,i}.
\]
A compatible configuration update supplies the remaining kinematic
relations. Here $x_i$, $x_{i+1}$, $\tau_i$, and $F_i$ are optimized jointly,
and this equality requires their combination to satisfy the equations of
motion. This is called an inverse-dynamics transcription: for each candidate
motion, the equations constrain the joint torques and contact forces needed
for dynamic consistency, subject to actuator and contact limits. A
forward-dynamics transcription would instead compute acceleration from the
current state and forces by applying $M(q_i)^{-1}$ to the net generalized
force, then integrate the state. Keeping force balance as a constraint
avoids explicitly forming $M(q_i)^{-1}$ at every stage. The constraints
remain nonlinear because $M$, $b$, and $J_c$ depend on the state.

The optimization then has the form
\begin{equation}
  \begin{aligned}
    \minimize_Z\quad
      &\Phi(x_N)+\sum_{i=0}^{N-1}\ell_i
       +\|\tau_0-\tau_{\mathrm{prev}}\|_{R_\Delta}^2\\
    \subjectto\quad
      &x_0=\hat x,\\
      &D_i(x_i,x_{i+1},\tau_i,F_i)=0,
        &&i=0,\ldots,N-1,\\
      &(x_i,\tau_i,F_i)\in\mathcal A_i(\gamma_i),
        &&i=0,\ldots,N-1,\\
      &x_N\in\mathcal A_N.
  \end{aligned}
  \label{eq:mpc-whole-body-ocp}
\end{equation}
Here $\gamma_i=(\gamma_{c,i})_{c\in\mathcal C}$ is the vector of prescribed
contact indicators, which are data rather than optimization variables:
$\gamma_{c,i}=1$ selects stance conditions for contact
$c$, and $\gamma_{c,i}=0$ selects swing conditions. The stage set
$\mathcal A_i(\gamma_i)$ contains the tuples $(x_i,\tau_i,F_i)$ satisfying
the constraints active at stage $i$. For stance contacts, these include
staying on the support surface, zero tangential velocity, nonnegative
normal force, and the friction limit; for swing contacts, they set
$F_{c,i}=0$ and include any modeled hard clearance constraint. The set also
includes the applicable joint-position and velocity bounds, actuator-torque
bounds, and selected self-collision constraints. Thus the contact indicators
choose which contact conditions apply, while contact timing remains fixed
by the prescribed schedule. The terminal set $\mathcal A_N$ contains only
the applicable constraints on $x_N$, such as specified terminal state bounds
or geometric clearance requirements; it has no stage torque or force
variables. The terminal cost $\Phi$ represents consequences beyond
the prediction horizon. The last cost term discourages a sudden change
from the previously applied torque $\tau_{\mathrm{prev}}$, with
$R_\Delta\succeq0$. It is a smoothness preference, separate from torque
bounds. This schematic problem does not assume that $\mathcal A_N$ is a
terminal invariant set or that $\Phi$ is an exact value function.

\subsection{Real-Time Solution and the Humanoid Implementation}

The optimization must return a useful command before the state changes
too far from its initial estimate. Sequential quadratic programming (SQP)
linearizes the constraints and uses a quadratic local objective, then
updates the trajectory by solving the resulting quadratic program (QP).
Reusing a previous
trajectory as a \emph{warm start} reduces the work needed at the next
update. Solving each nonlinear problem to high precision can be less useful
than obtaining an adequate update quickly, especially when the prediction
already contains model error.

\citet{khazoom2024tailoring} implement one SQP iteration per update, with a
fixed number of iterations of the \emph{alternating direction method of
multipliers} (ADMM) in the operator-splitting QP solver OSQP. Their contact
schedule and swing-height references are supplied to the optimizer; foot
placements and coordinated body motion are optimized within these
restrictions. Their walking implementation also reduces the general
formulation above: joint-torque bounds and the actuated inverse-dynamics
equations are enforced only at the first two stages. This retains torque
consistency near the action being executed while reducing computation
farther into the prediction.

The authors add \emph{control barrier function} (CBF) constraints for
self-collision at the first stage. A CBF constrains the evolution of the
separation, accounting for motion toward a collision as well as the current
distance. At the measured state, the relevant acceleration constraint is
affine in the acceleration variables. Later stages retain geometric
distance constraints. This construction improves collision avoidance under
incomplete solves; it does not make an approximate solution an exact
feasibility guarantee.

The reported MIT Humanoid controller runs at approximately
$90\,\mathrm{Hz}$ with 32 prediction stages and demonstrates walking and
disturbance recovery with coordinated arm and crossed-leg motions. The
32 stages describe the plan being optimized; the update rate describes
how often that plan is revised. Optimized position, velocity, and torque
trajectories are interpolated and supplied to faster motor-level feedback.

To see the role of this last layer, a typical joint-level command has the
form
\begin{equation}
  \tau_{\mathrm{cmd}}
  =\tau_{\mathrm{ff}}
   +K_p(q_j^{\mathrm{des}}-q_j)
   +K_d(\dot q_j^{\mathrm{des}}-\dot q_j),
  \label{eq:mpc-whole-body-low-level-control}
\end{equation}
where the feedforward torque and desired joint motion come from the
optimized trajectory, and $K_p$ and $K_d$ are the proportional and derivative
gain matrices. These feedback terms correct local
tracking errors between optimization updates. The predicted contact forces
are realized through interaction with the ground, rather than sent as
independent motor commands. New measurements update the state estimate and
start the next optimization.

For example, after a lateral push, the estimated sideways velocity changes.
Replanning can adjust the next foot placement, torso motion, arm motion,
and supporting forces together while preserving clearance between the legs.
The optimizer evaluates their consequences over the horizon, and local
feedback tracks the beginning of the resulting motion. The achievable
recovery still depends on the available contacts, actuator authority,
estimation accuracy, and computation delay; the hardware results do not
establish stability or feasibility for every disturbance.

\paragraph{Exercises.}

\begin{exercise}
Explain why MPC is feedback even when the finite-horizon optimizer returns
an open-loop sequence. What part of that sequence is normally executed?
\end{exercise}
\begin{proof}[Short solution]
Only the first input is executed. At the next sample, a new state estimate
changes the initial condition of the optimization and therefore the next
executed input. The repeated optimization defines an implicit feedback law.
\end{proof}

\begin{exercise}
Distinguish the prediction horizon, the control horizon, and the number of
controls applied before replanning.
\end{exercise}
\begin{proof}[Short solution]
The prediction horizon specifies how far the model is propagated. The
control horizon specifies how many future moves are independently
parameterized. Standard MPC applies one control before replanning,
regardless of either horizon length.
\end{proof}

\begin{exercise}
For the scalar system $x^+=ax+u$, derive
\cref{eq:mpc-scalar-terminal-feedback} from the horizon-one objective with
terminal weight $s$. Why does $s=0$ fail for $a>1$?
\end{exercise}
\begin{proof}[Short solution]
Differentiating the control-dependent cost gives
$ru+s(ax+u)=0$, hence $u=-asx/(r+s)$. Substitution gives
$x^+=arx/(r+s)$. When $s=0$, the minimizing input is zero and the unstable
open-loop coefficient $a$ is unchanged.
\end{proof}

\begin{exercise}
Why does unconstrained linear MPC with terminal matrix $P_\infty$ reproduce
infinite-horizon LQR for any positive horizon length?
\end{exercise}
\begin{proof}[Short solution]
The stabilizing DARE solution is a fixed point of the backward Riccati
recursion. Initializing its terminal matrix at $P_\infty$ therefore gives
the same matrix and gain at every stage. The first MPC control equals the
infinite-horizon LQR control.
\end{proof}

\begin{exercise}
Explain how a terminal invariant set supports recursive feasibility in
nominal MPC. What can break this guarantee?
\end{exercise}
\begin{proof}[Short solution]
Shifting the previous feasible sequence and appending an admissible terminal
control gives a candidate next plan; invariance keeps its endpoint in the
terminal set. A disturbance or estimation error can make the actual next
state differ from its prediction, invalidating that candidate.
\end{proof}

\begin{exercise}
Why does adding slack to a tracking constraint not guarantee feasibility or
safety of the entire MPC problem?
\end{exercise}
\begin{proof}[Short solution]
Slack relaxes only the constraint to which it is attached. The dynamics,
actuator bounds, and other hard constraints can still be infeasible.
Moreover, a positive slack explicitly permits violation of the original
tracking requirement and says nothing about physical safety.
\end{proof}

\begin{exercise}
For the cart with $v=1\,\mathrm{m/s}$ and
$a_{\max}=2\,\mathrm{m/s^2}$, compute the stopping time and distance.
Can it maintain wall clearance if only $0.15\,\mathrm{m}$ is available?
\end{exercise}
\begin{proof}[Short solution]
The stopping time is $v/a_{\max}=0.5\,\mathrm{s}$ and the stopping distance
is $v^2/(2a_{\max})=0.25\,\mathrm{m}$. The available distance is smaller,
so the stated continuous-time clearance is impossible under this model.
\end{proof}

\begin{exercise}
When is the linear--quadratic MPC problem a QP, and why can an ellipsoidal
terminal set change that classification?
\end{exercise}
\begin{proof}[Short solution]
It is a QP when the objective is quadratic and all constraints are affine.
An ellipsoidal terminal set imposes a quadratic inequality rather than an
affine one. The problem can remain convex, but is no longer a pure QP
with only linear constraints.
\end{proof}

\begin{exercise}
Why should a limited-iteration MPC implementation be analyzed together with
its warm start and failure-handling rules?
\end{exercise}
\begin{proof}[Short solution]
Its returned control depends on stored solver state as well as the current
estimate. An incomplete solve may fail feasibility or cost-decrease tests,
and a fallback may apply a different control entirely. These choices are
part of the implemented closed-loop system.
\end{proof}

\paragraph{Chapter summary.}
Model predictive control repeatedly solves a finite-horizon trajectory
problem, executes its first control, and replans from new measurements.
The prediction is finite, but the repeated execution is feedback. MPC can
anticipate state and input constraints before they become unavoidable,
using the structured solvers developed in Chapter~9.

Linear dynamics, quadratic costs, and affine constraints give a QP.
Without constraints and with an infinite-horizon LQR terminal cost, the
controller reproduces LQR. Nonlinear robot dynamics and nonconvex geometric
or contact constraints require more general optimization. Prediction
horizon, control parameterization, and terminal ingredients determine what
future consequences the controller can represent.

Recursive feasibility and stability are not automatic consequences of
replanning. In the nominal setting, a terminal invariant set and a terminal
cost-decrease condition allow the old plan to be shifted and extended.
This gives a feasible next candidate and a decrease of the finite-horizon
value. Disturbances, state-estimation errors, changing constraints, and
incomplete solves require additional reasoning.

A practical controller must combine warm starts, reliable derivatives,
consistent timestamps, state estimation, constraint checks, and explicit
responses to missed deadlines or failed solves. The braking example shows why
position alone is insufficient: velocity and braking authority determine
whether the required clearance can be maintained. Robust and stochastic MPC extend the
framework by explicitly modeling uncertainty rather than assuming that
frequent nominal replanning provides every needed guarantee.

Whole-body NMPC applies these ideas to coordinated robot motion by
optimizing joint motion and contact forces together. The humanoid example
also illustrates why solver accuracy must be balanced against the delay
before a command can act.

The next chapter keeps this receding-horizon architecture but changes the
computational engine. Sampling-based MPC and model predictive path integral
control search over simulated control sequences, providing alternatives
when useful derivatives are unavailable or parallel rollout computation is
attractive.
\chapter{Sampling-Based MPC and MPPI}
\label{chap:sampling-based-mpc-mppi}
\chapteroverview{Derivative-free trajectory search, GPU and NPU computation,
probability densities and importance sampling, information-theoretic control,
the Gibbs variational principle, the MPPI update, practical constraints,
and robotics examples.}

Chapter~10 established the model predictive control architecture: predict
the consequences of candidate actions, optimize a finite-horizon plan,
execute its first action, and replan from new measurements. It did not
require any particular optimizer. This chapter keeps that architecture but
changes the computational engine. Instead of differentiating a rollout and
solving a local quadratic problem, a sampling-based optimizer evaluates
many candidate control sequences and uses their costs to construct a new
plan.

This choice is useful when derivatives are unavailable, unreliable, or
expensive. A simulator may contain contact switches, a cost may include an
obstacle indicator or a lookup table, and a learned dynamics model may be
available only through an inference interface. Sampling asks each model
the simpler question: what happens if this sequence is applied? It does
not make the model more accurate or remove nonconvexity, but it allows
optimization to use the original rollout without constructing its local
derivatives.

The apparent disadvantage is the number of sample trajectories. Modern parallel hardware changes this tradeoff: independent candidate trajectories can be
evaluated concurrently, and neural predictors can process batches of
candidate states and actions. A graphics processing unit (GPU) can accelerate
suitable simulation and
cost computations; a neural processing unit (NPU) can accelerate supported
neural-model inference. These are opportunities, not automatic speedups.
The entire estimation--planning--execution loop must still meet its deadline.

The main algorithm is \emph{model predictive path integral control} (MPPI).
We derive its cost-weighted update through a finite-dimensional
information-theoretic formulation, introducing the probability tools before
using them. This route explains both the exponential weights and the
likelihood-ratio term that is easily omitted in informal implementations.
The principal references are
\cite{williams2017parallel,williams2017learning,williams2018information}.
We also connect the derivation to classical path-integral stochastic control,
while distinguishing its special assumptions from the more general
sampling-based optimization procedure.

\section{Why Sampling, and the Probability Tools It Needs}
\label{sec:sampling-mpc-foundations}

\subsection{A Rollout Is the Basic Computational Primitive}

At one MPC update, let $x_0=\hat x_k$ be the current state estimate and use
the prediction model
\begin{equation}
  x_{i+1}=f_i(x_i,v_i),
  \qquad i=0,\ldots,N-1.
  \label{eq:sampling-mpc-dynamics}
\end{equation}
We suppress the conditioning index $|k$ within this solve. A candidate input
sequence is $V=(v_0,\ldots,v_{N-1})\in\R^{Nm}$. Starting from $x_0$,
forward simulation determines its state trajectory and cost
\begin{equation}
  J(V)=\Phi(x_N)+\sum_{i=0}^{N-1}\ell_i(x_i,v_i).
  \label{eq:sampling-mpc-rollout-cost}
\end{equation}
The inputs can be torques, accelerations, or reference commands to a
low-level controller, provided the prediction model represents the
consequences of those commands.

For deterministic dynamics, one rollout needs $N$ model transitions.
Evaluating $K$ candidates needs $KN$ transitions, but candidates do not
depend on one another. Time steps \emph{within} a trajectory are normally
sequential because $x_{i+1}$ depends on $x_i$; the sample index is the main
source of parallelism. This distinction matters when estimating the benefit
of a larger accelerator.

Derivative-free does not mean model-free. The optimizer still needs a
prediction of future motion, and it can confidently choose a poor action
if that prediction is wrong. Nor does parallel hardware eliminate the
dimensionality problem. If a useful set occupies fraction $r$ of the
proposal probability, $K$ independent draws miss it with probability
$(1-r)^K$. When $r$ is tiny, simply increasing the sample count is an
inefficient substitute for a good proposal, warm start, or control
parameterization.

\subsection{Densities, Expectations, and Gaussian Perturbations}

Let $P$ be a probability measure on the sequence space $\R^{Nm}$,
with density $p$.
For a random sequence $V$ drawn from $P$,
\begin{equation}
  p(V)\geq0,\qquad \int p(V)\,dV=1,
  \qquad \E_P[h(V)]=\int h(V)p(V)\,dV.
  \label{eq:sampling-expectation-definition}
\end{equation}
Here the integral is over the whole sequence space $\R^{Nm}$, not over
physical time. The density value $p(V)$ is not itself a probability. For a
region $A\subset\R^{Nm}$, the probability that a sample falls in $A$ is
$P(V\in A)=\int_A p(V)\,dV$. Thus, any one exact sequence has probability
zero under this continuous measure, while a region of sequences may have
positive probability. We distinguish the measure $P$ from its density $p$.

If $V^{(1)},\ldots,V^{(K)}$ are independent draws from $P$, the sample mean
\begin{equation}
  \widehat h_K=\frac1K\sum_{j=1}^K h(V^{(j)})
  \label{eq:sampling-monte-carlo-mean}
\end{equation}
estimates $\E_P[h(V)]$. For scalar $h$ with finite variance under $P$, let
$\mu=\E_P[h(V)]$. Taking expectation over repeated samples and using
linearity,
\begin{align}
  \E[\widehat h_K]
  &=\frac1K\sum_{j=1}^K\E\!\left[h(V^{(j)})\right]
   =\frac1K\sum_{j=1}^K\mu
   =\mu.
\end{align}
Thus the sample mean is an unbiased estimate of $\mu$. Variance is the
expected squared deviation from the mean:
\begin{equation}
  \sigma_h^2:=\operatorname{Var}_P(h(V))
  =\E_P\!\left[(h(V)-\mu)^2\right].
\end{equation}
Write $Y_j=h(V^{(j)})$. Applying this definition to the sample mean gives
\begin{align}
  \operatorname{Var}(\widehat h_K)
  &=\E\!\left[(\widehat h_K-\mu)^2\right]\\
  &=\frac1{K^2}\E\!\left[
      \left(\sum_{j=1}^K(Y_j-\mu)\right)^2\right]\\
  &=\frac1{K^2}\left[
      \sum_{j=1}^K\E\!\left[(Y_j-\mu)^2\right]
      +2\sum_{i<j}\E\!\left[(Y_i-\mu)(Y_j-\mu)\right]\right]\\
  &=\frac{K\sigma_h^2}{K^2}
   =\frac{\sigma_h^2}{K}.
  \label{eq:sampling-monte-carlo-variance}
\end{align}
Each $Y_j$ has variance $\sigma_h^2$. For $i\ne j$, independence gives
$\E[(Y_i-\mu)(Y_j-\mu)]
=\E[Y_i-\mu]\E[Y_j-\mu]=0$, so the cross terms vanish. The standard error
(the standard deviation of $\widehat h_K$) is therefore
$\sqrt{\sigma_h^2/K}$. It decreases only as $K^{-1/2}$:
quadrupling the sample count halves the standard error.

At each MPC update, we start from a nominal control sequence
$U=(u_0,\ldots,u_{N-1})$ and draw candidate sequences around it. The
distribution used to generate these candidates is called the proposal.
Their predicted costs guide the update of $U$. A convenient proposal adds
Gaussian noise:
\begin{equation}
  v_i=u_i+\epsilon_i,
  \qquad \epsilon_i\sim\mathcal N(0,\Sigma),
  \qquad \Sigma\succ0.
  \label{eq:sampling-gaussian-perturbation}
\end{equation}
Each $\epsilon_i\in\R^m$ is a vector of random perturbations, one for each
input component. The multivariate normal (or Gaussian) distribution extends
the scalar bell curve to vectors. A vector $Z\in\R^m$ with mean $\mu$ and
covariance $\Sigma$ has distribution $\mathcal N(\mu,\Sigma)$, where
$\mu=\E[Z]$ and
$\Sigma=\E[(Z-\mu)(Z-\mu)\trans]$. Its density is
\begin{equation}
  p_Z(z)=\frac{1}{(2\pi)^{m/2}|\Sigma|^{1/2}}
    \exp\!\left[-\tfrac12(z-\mu)\trans\Sigma^{-1}(z-\mu)\right].
  \label{eq:sampling-multivariate-gaussian-density}
\end{equation}
In our case $Z=\epsilon_i$ and $\mu=0$. For any direction
$a\in\R^m$, the scalar perturbation $a\trans\epsilon_i$ is normally
distributed with variance $a\trans\Sigma a$. The diagonal entries of
$\Sigma$ set the variance of each input component, while its off-diagonal
entries describe how perturbations in different components vary together.
Positive definiteness ensures there is nonzero spread in every direction.

For one candidate sequence, assume perturbations are independent across
stages. Independence means that the joint density of the full sequence is
the product of the stagewise densities. Thus the proposal defines a probability measure
$Q_U$ on sequences, with density
\begin{equation}
  q_U(V)=\prod_{i=0}^{N-1}
    \frac{\exp\!\left[-\tfrac12(v_i-u_i)\trans\Sigma^{-1}(v_i-u_i)\right]}
         {(2\pi)^{m/2}|\Sigma|^{1/2}}.
  \label{eq:sampling-gaussian-sequence-density}
\end{equation}
The covariance controls the magnitude and direction of exploration.
Its entries have squared input units, so steering, throttle, and joint
torque should not all receive the same numerical scale without considering
their meanings.

\subsection{Importance Sampling}

The distribution of interest is often difficult to sample, while a proposal
is easy to sample. For the Gaussian proposal above, factor the covariance
once as $\Sigma=LL\trans$ (for example, using a Cholesky factorization).
At each stage and for each candidate, draw a fresh vector
$z_i\sim\mathcal N(0,I_m)$ by sampling $m$ independent scalar standard
normals, then transform it:
\begin{equation}
  \epsilon_i=Lz_i,
  \qquad v_i=u_i+Lz_i.
  \label{eq:sampling-gaussian-cholesky-draw}
\end{equation}
Indeed, $\E[\epsilon_i]=0$ and
$\operatorname{Cov}(\epsilon_i)=L I_m L\trans=\Sigma$, so this produces
the required perturbation. Reuse $L$ for every stage and candidate as long
as $\Sigma$ stays fixed.

Suppose the expectation we want is under a target distribution $P$ with
density $p$, but drawing from $P$ is difficult. Instead, we may draw from
an easier proposal distribution $Q$ with density $q$, provided both
densities are defined on the same sequence space. If $h$ is integrable
under $P$ and $q(V)>0$ wherever $p(V)>0$, then
\begin{align}
  \E_P[h(V)]
  &=\int h(V)p(V)\,dV\\
  &=\int h(V)\frac{p(V)}{q(V)}q(V)\,dV\\
  &=\E_Q\!\left[h(V)\frac{p(V)}{q(V)}\right].
  \label{eq:sampling-importance-identity}
\end{align}
This proves the importance-sampling identity: multiply and divide the
integrand by $q(V)$, then recognize the final integral as an expectation
under $Q$. The factor $p(V)/q(V)$ corrects for the change in sampling
distribution. With independent draws $V^{(j)}\sim Q$, the corresponding
Monte Carlo estimate is
$K^{-1}\sum_{j=1}^K h(V^{(j)})p(V^{(j)})/q(V^{(j)})$; its expectation is
exactly $\E_P[h(V)]$.

The support condition is essential. If $p(V)>0$ but $q(V)=0$ on a region,
the proposal never samples that region, so no weight can recover its
contribution to the target expectation. In other words, the proposal must
cover every region that matters under $P$.

If the target density is known only up to a constant,
$\pi(V)=\widetilde\pi(V)/Z$, normalized importance weights give
\begin{align}
  a_j&=\frac{\widetilde\pi(V^{(j)})}{q(V^{(j)})},\\
  \omega_j&=\frac{a_j}{\sum_{\ell=1}^K a_\ell},\\
  \widehat I_K&:=\sum_{j=1}^K\omega_j h(V^{(j)}),\\
  \E_\pi[h(V)]&\approx\widehat I_K.
  \label{eq:sampling-self-normalized-weights}
\end{align}
The unknown constant cancels when the weights are normalized.

Let $I=\E_\pi[h(V)]$ be the quantity being estimated. Expectations and
probabilities of $\widehat I_K$ refer to repeating the entire sampling
procedure. An estimator is \emph{biased} at sample size $K$ if its average
over repeated samples differs from the target, $\E[\widehat I_K]\ne I$; it
is unbiased if equality holds. It is \emph{consistent} if, for every
$\varepsilon>0$, the probability that its error exceeds $\varepsilon$
goes to zero as the number of samples grows:
\begin{equation}
  \Prob\!\left(|\widehat I_K-I|>\varepsilon\right)
  \longrightarrow 0
  \qquad\text{as }K\longrightarrow\infty.
\end{equation}

The self-normalized estimator is generally biased for finite $K$ because it
is a ratio of random sums, and the expectation of a ratio is not generally
the ratio of the expectations. Yet, under the support and integrability
assumptions, the law of large numbers makes the numerator and denominator,
after division by $K$, approach their respective expectations; their ratio
then approaches $I$. Thus finite-sample bias and consistency as $K$ grows
can both hold.
Highly concentrated weights can nevertheless make the estimate inaccurate
at practical sample counts.

\subsection{Kullback--Leibler Divergence}
\label{subsec:sampling-kl-divergence}

Importance sampling compares a target distribution with a proposal. The
Kullback--Leibler (KL) divergence, also called relative entropy, is a more
general way to quantify how two probability measures differ, with the
direction of comparison specified by their order. Let $Q$ and $P$ have
densities $q$ and $p$ with respect to the same base measure. The divergence
from $Q$ to $P$ is
\begin{align}
  D_{\mathrm{KL}}(Q\|P)
  &=\int q(V)\log\frac{q(V)}{p(V)}\,dV\\
  &=\E_Q\!\left[\log\frac{q(V)}{p(V)}\right].
  \label{eq:sampling-kl-definition}
\end{align}
Thus KL divergence is the expected log-density ratio when $V$ is drawn
from $Q$: the ratio $q(V)/p(V)$ compares the density assigned by each
measure near the same sequence, and the expectation averages that
comparison according to $Q$. We use the convention that points with
$q(V)=0$ contribute zero. If $Q$ assigns positive probability to a set
where $P$ assigns zero probability, then $D_{\mathrm{KL}}(Q\|P)=+\infty$.

The divergence is nonnegative. When $Q$ is absolutely continuous with
respect to $P$ (so the density ratio is defined $Q$-almost everywhere),
Jensen\textquotesingle s inequality gives
\begin{align}
  D_{\mathrm{KL}}(Q\|P)
  &=-\E_Q\!\left[\log\frac{p(V)}{q(V)}\right]\\
  &\geq -\log\E_Q\!\left[\frac{p(V)}{q(V)}\right]\\
  &=-\log\int_{\{V:q(V)>0\}}p(V)\,dV\geq0.
\end{align}
It equals zero exactly when $Q=P$ almost everywhere. The order matters:
$D_{\mathrm{KL}}(Q\|P)$ averages under $Q$, whereas
$D_{\mathrm{KL}}(P\|Q)$ averages under $P$.

For multivariate Gaussian measures
$P=\mathcal N(\mu_P,\Sigma_P)$ and
$Q=\mathcal N(\mu_Q,\Sigma_Q)$ on $\R^d$, with positive-definite
covariances, substitution of their densities into the definition gives
\begin{equation}
  D_{\mathrm{KL}}(Q\|P)
  =\frac12\left[
    \operatorname{tr}(\Sigma_P^{-1}\Sigma_Q)
    +(\mu_P-\mu_Q)\trans\Sigma_P^{-1}(\mu_P-\mu_Q)
    -d+\log\frac{|\Sigma_P|}{|\Sigma_Q|}
  \right].
  \label{eq:sampling-kl-gaussian}
\end{equation}
The mean term measures displacement relative to the spread of $P$; the
remaining terms compare the two covariance shapes and volumes.

The one-dimensional case makes the calculation especially transparent.
For $P=\mathcal N(\mu_P,\sigma_P^2)$ and
$Q=\mathcal N(\mu_Q,\sigma_Q^2)$, the log-density ratio is
\begin{align}
  \log\frac{q(x)}{p(x)}
  &=\log\frac{\sigma_P}{\sigma_Q}
    -\frac{(x-\mu_Q)^2}{2\sigma_Q^2}
    +\frac{(x-\mu_P)^2}{2\sigma_P^2},\\
  D_{\mathrm{KL}}(Q\|P)
  &=\E_Q\!\left[\log\frac{q(X)}{p(X)}\right]
   =\log\frac{\sigma_P}{\sigma_Q}
    +\frac{\sigma_Q^2+(\mu_Q-\mu_P)^2}{2\sigma_P^2}
    -\frac12.
  \label{eq:sampling-kl-scalar-gaussians}
\end{align}
The expectation uses $\E_Q[(X-\mu_Q)^2]=\sigma_Q^2$ and
$\E_Q[(X-\mu_P)^2]=\sigma_Q^2+(\mu_Q-\mu_P)^2$. If the variances agree
at $\sigma^2$, the expression reduces to
$D_{\mathrm{KL}}(Q\|P)=(\mu_Q-\mu_P)^2/(2\sigma^2)$. For example,
$P=\mathcal N(0,1)$ and $Q=\mathcal N(1,1)$ differ in mean by one
standard deviation and have divergence $1/2$ nat.

Changing the variance illustrates the directionality. For
$P=\mathcal N(0,1)$ and $Q=\mathcal N(0,4)$,
\begin{align*}
  D_{\mathrm{KL}}(Q\|P)&=\frac32-\log 2\approx0.807,\\
  D_{\mathrm{KL}}(P\|Q)&=\log 2-\frac38\approx0.318.
\end{align*}
The same pair of measures therefore has different divergence depending on
which one is the reference.

KL divergence is not an ordinary geometric distance: it is asymmetric and
does not generally satisfy the triangle inequality. Its direction also
affects what it penalizes. For $D_{\mathrm{KL}}(Q\|P)$, placing $Q$ outside
the support of $P$ incurs infinite divergence, while regions assigned zero
density by $Q$ make no direct contribution to the integral. As a result,
this direction can under-penalize a $Q$ that ignores some modes of $P$.

\subsection{Random Shooting and the Cross-Entropy Method}

\emph{Random shooting} is the most direct sampling-based optimizer. It draws
$K$ candidate control sequences from a proposal distribution, simulates each
to evaluate \cref{eq:sampling-mpc-rollout-cost}, and selects the lowest-cost
admissible candidate. Under receding-horizon execution, only its first input
is applied before replanning from the next measurement. The method requires
no derivatives and is straightforward to implement, but a poorly centered
or overly broad proposal can waste most of the rollout budget on
unpromising candidates.

The \emph{cross-entropy method} (CEM) adapts the proposal. For a Gaussian
sequence distribution with mean $\mu$ and covariance $C$, sample a batch
and retain an elite index set $\mathcal I$ containing the best $K_e$
candidates. A maximum-likelihood fit to these elite sequences gives
\begin{align}
  \mu^{\mathrm{new}}
    &=\frac1{K_e}\sum_{j\in\mathcal I}V^{(j)},
  \label{eq:sampling-cem-mean}\\
  C^{\mathrm{new}}
    &=\frac1{K_e}\sum_{j\in\mathcal I}
        (V^{(j)}-\mu^{\mathrm{new}})
        (V^{(j)}-\mu^{\mathrm{new}})\trans.
  \label{eq:sampling-cem-covariance}
\end{align}
Repeat the selection-and-refitting cycle as needed within each MPC update.
In practice, implementations often smooth parameter updates, constrain the
covariance to a structured form, and maintain a covariance floor. The
covariance sets the spread of candidate sequences around the current mean.
If the selected elites happen to cluster tightly, repeated refitting can
shrink this spread until later candidates are nearly identical, reducing
exploration too early. A covariance floor enforces a minimum spread---for
example, a lower bound on each variance for a diagonal covariance, or on
the smallest eigenvalue for a full covariance. The floor should reflect
the scale and units of the controls: too small may not prevent collapse,
while too large can keep samples needlessly broad. CEM gives zero weight to
non-elite samples after a hard cutoff, whereas MPPI uses continuous
exponential weights with a correction for the proposal density.

\subsection{Why GPUs and NPUs Change the Tradeoff}

Sampling replaces one tightly coupled local solve by many forward
evaluations. This is attractive when the available hardware can perform
those evaluations concurrently. GPU-based MPPI implementations exploit
this structure directly \cite{williams2017parallel}. GPU physics systems
such as Isaac Gym also demonstrate the benefit of keeping simulation data
and tensor computation on the same device
\cite{makoviychuk2021isaacgym}; that simulator work is evidence about the
computational primitive, not a claim that its learning experiments are
themselves MPC.

A typical batched implementation stores candidate inputs in a tensor of
shape $K\times N\times m$ and current predicted states in a tensor of
shape $K\times n$. At stage $i$, it evaluates the model for all $K$ states
and inputs, accumulates their costs, and proceeds to stage $i+1$.
After the horizon, reductions compute the best candidate or the weighted
update. A learned predictor $f_\theta(x,u)$ can similarly receive a batch
of concatenated state--input vectors.

An NPU is particularly relevant to the latter case: supported neural-network
operations may be executed by a dedicated inference engine. It is not
interchangeable with a general-purpose GPU physics implementation. A
rigid-body contact solve, random-number generation, obstacle queries, and
the weight reductions may still run elsewhere. Supported operations,
shapes, precision, and transfer overhead must be checked for the actual
deployment; inference runtimes explicitly impose device-specific
restrictions \cite{openvino2025npu}.

\begin{table}[t]
  \centering
  \caption{Typical opportunities for parallel sampling-based MPC.
  The appropriate device depends on the workload, not only its peak throughput.}
  \label{tab:sampling-mpc-hardware}
  \begin{tabular}{@{}>{\raggedright\arraybackslash}p{0.12\textwidth}
                    >{\raggedright\arraybackslash}p{0.37\textwidth}
                    >{\raggedright\arraybackslash}p{0.39\textwidth}@{}}
    \toprule
    Device & Useful work & Main qualification \\
    \midrule
    CPU & small models, branching logic, parallel simulator instances
        & a small batch may not justify accelerator overhead \\
    GPU & batched rollouts, compatible physics, neural inference, reductions
        & memory traffic and sequential horizon steps still matter \\
    NPU & supported neural dynamics or perception inference
        & neural inference support does not imply support for the entire planner \\
    \bottomrule
  \end{tabular}
\end{table}

Increasing $K$ usually improves coverage but also increases memory and
computation. If every model step copies data between CPU and accelerator,
communication can erase the benefit of parallel arithmetic. Useful
measurements are end-to-end latency, missed deadlines, model accuracy, and
closed-loop performance, not just the number of simulated trajectories per
second. Gradient-based MPC can also exploit accelerators; sampling is a
good match to parallel hardware, not its exclusive beneficiary.

\section{Information-Theoretic Control and the MPPI Derivation}
\label{sec:information-theoretic-mppi}

\subsection{From a Control Sequence to a Distribution of Sequences}

For this derivation, use deterministic rollout dynamics
\cref{eq:sampling-mpc-dynamics} and define the state-dependent trajectory cost
\begin{equation}
  S(V)=\Phi(x_N)+\sum_{i=0}^{N-1}c_i(x_i).
  \label{eq:mppi-state-trajectory-cost}
\end{equation}
The state $x_i$ is obtained by rolling out $V$ from the current estimate.
Initially, $S$ contains no separate quadratic input-effort term; effort
will enter through the reference distribution.

Let $P$ be a fixed reference probability measure over control sequences,
with sequence density $p(V)$. We use independent, zero-mean Gaussian
inputs at each stage, so for $V=(v_0,\ldots,v_{N-1})$,
\begin{equation}
  p(V)=\prod_{i=0}^{N-1}
    \frac{1}{(2\pi)^{m/2}|\Sigma|^{1/2}}
    \exp\!\left(-\tfrac12v_i\trans\Sigma^{-1}v_i\right).
  \label{eq:mppi-reference-density}
\end{equation}
Thus $p(V)$ is the reference sequence density evaluated at $V$; it is not
the probability of drawing that exact sequence.

Rather than first choosing one sequence, consider the distributional problem
\begin{equation}
  \minimize_Q\quad
  \mathcal J(Q)=\E_Q[S(V)]+\lambda D_{\mathrm{KL}}(Q\|P),
  \qquad \lambda>0.
  \label{eq:mppi-distributional-objective}
\end{equation}
The minimization is over probability measures absolutely continuous with
respect to $P$: they cannot assign probability to sets that $P$ rules out.
The first term favors low-cost rollouts; the second penalizes departures
from the reference measure. The temperature $\lambda$ has cost units and
balances these terms. This is a regularized distributional optimization
problem, not yet the deterministic constrained trajectory problem of
Chapter~9.

To solve it, let $q(V)$ denote the density of any admissible distribution
$Q$. The density form of KL divergence gives
\begin{align*}
  D_{\mathrm{KL}}(Q\|P)
    &=\E_Q\!\left[\log\frac{q(V)}{p(V)}\right],\\
  \mathcal J(Q)
    &=\E_Q\!\left[S(V)+\lambda\log\frac{q(V)}{p(V)}\right].
\end{align*}
Inside this expectation, combine the cost with the log-density ratio:
\begin{equation*}
  S(V)+\lambda\log\frac{q(V)}{p(V)}
  =\lambda\log\frac{q(V)}{p(V)\exp(-S(V)/\lambda)}.
\end{equation*}
This reveals the form of the optimum: the cost-weighted reference density
is proportional to
\begin{equation*}
  q^\star(V)\propto p(V)\exp\!\left(-\frac{S(V)}{\lambda}\right).
\end{equation*}
Thus, relative to the reference distribution, higher-cost sequences are
downweighted exponentially. This proportional expression is not generally
normalized. Its total mass under the reference distribution is
\[
  Z:=\E_P\!\left[\exp\!\left(-\frac{S(V)}{\lambda}\right)\right].
\]
Assume $0<Z<\infty$. This normalizing constant is called the
\emph{partition function}. Dividing by it gives the normalized density
\begin{equation}
  q^\star(V)=\frac{1}{Z}\exp\!\left(-\frac{S(V)}{\lambda}\right)p(V).
  \label{eq:mppi-gibbs-density}
\end{equation}
Because it exponentially favors lower-cost sequences relative to the
reference distribution, $q^\star$ is called the \emph{Gibbs density}.

We can verify that this is the minimizing density by comparing any $q$ to
$q^\star$. Taking the log of their ratio gives
\begin{equation*}
  \log\frac{q(V)}{q^\star(V)}
  =\log\frac{q(V)}{p(V)}+\frac{S(V)}{\lambda}+\log Z.
\end{equation*}
Now take expectations under $Q$. The expected log-ratios are KL
divergences, and the cost-plus-reference terms combine to give the
objective:
\begin{align*}
  D_{\mathrm{KL}}(Q\|Q^\star)
  &=D_{\mathrm{KL}}(Q\|P)
    +\frac{1}{\lambda}\E_Q[S(V)]+\log Z\\
  &=\frac{\mathcal J(Q)}{\lambda}+\log Z.
\end{align*}
Rearranging gives
\begin{equation*}
  \mathcal J(Q)
  =\lambda D_{\mathrm{KL}}(Q\|Q^\star)-\lambda\log Z.
\end{equation*}
The scaled negative log of the partition function is called the
\emph{free energy} in this formulation:
\begin{equation}
  \mathcal F:=-\lambda\log Z.
  \label{eq:mppi-free-energy}
\end{equation}
Therefore,
\begin{equation}
  \mathcal J(Q)
  =\mathcal F+\lambda D_{\mathrm{KL}}(Q\|Q^\star).
  \label{eq:mppi-gibbs-variational-identity}
\end{equation}
Since KL divergence is nonnegative, the minimum objective value is
$\mathcal F$, attained when $Q=Q^\star$. This equality is the
\emph{Gibbs variational identity}. Sign and scaling conventions for the
name \emph{free energy} vary across the literature.

Low-cost sequences receive more probability than high-cost sequences,
relative to the same reference measure. This is \emph{exponential tilting}.
The target is not a uniform distribution over the best sequences: the
prior probability $p(V)$ also matters. For two sequences with equal prior
density, a cost difference $\Delta S$ gives a target-density ratio
$\exp(-\Delta S/\lambda)$. Thus $\lambda$ determines which differences
are significant.

\subsection{Where the Quadratic Control Cost Comes From}

The KL term in the distributional objective measures how far the chosen
sequence distribution moves from the zero-mean reference $P$. Flatten the
stage inputs into the full sequences
$V=(v_0,\ldots,v_{N-1})\in\R^{Nm}$ and
$U=(u_0,\ldots,u_{N-1})\in\R^{Nm}$. Since the stages are independent and
each has covariance $\Sigma$, the covariance of the full sequence is the
block-diagonal matrix
\[
  \bar\Sigma=\operatorname{diag}(\underbrace{\Sigma,\ldots,\Sigma}_{N\text{ blocks}}).
\]
The reference distribution is a zero-mean Gaussian over the full sequence;
the shifted distribution $Q_U$ has the same covariance and mean $U$. Their
densities are
\begin{align}
  p(V)
  &=\frac{1}{(2\pi)^{Nm/2}|\bar\Sigma|^{1/2}}
    \exp\!\left(-\frac12V\trans\bar\Sigma^{-1}V\right),\\
  q_U(V)
  &=\frac{1}{(2\pi)^{Nm/2}|\bar\Sigma|^{1/2}}
    \exp\!\left(-\frac12(V-U)\trans\bar\Sigma^{-1}(V-U)\right).
  \label{eq:mppi-full-sequence-gaussian-densities}
\end{align}
Because the covariance is the same, the normalizing constants cancel when
we take the log-density ratio. Expanding the quadratic form gives
\begin{equation*}
  \log\frac{q_U(V)}{p(V)}
  =-\frac12(V-U)\trans\bar\Sigma^{-1}(V-U)
    +\frac12V\trans\bar\Sigma^{-1}V
  =V\trans\bar\Sigma^{-1}U
    -\frac12U\trans\bar\Sigma^{-1}U.
\end{equation*}
The KL divergence is the expectation of this log-ratio under $Q_U$. Since
$\E_{Q_U}[V]=U$,
\begin{align}
  D_{\mathrm{KL}}(Q_U\|P)
  &=\E_{Q_U}\!\left[\log\frac{q_U(V)}{p(V)}\right]\\
  &=U\trans\bar\Sigma^{-1}U
    -\frac12U\trans\bar\Sigma^{-1}U\\
  &=\frac12U\trans\bar\Sigma^{-1}U
   =\frac12\sum_{i=0}^{N-1}u_i\trans\Sigma^{-1}u_i.
  \label{eq:mppi-gaussian-kl-effort}
\end{align}
This is a squared displacement of the mean, measured in covariance-scaled
units rather than ordinary Euclidean distance. Shifting the mean along a
direction with large sampling variance incurs a smaller KL penalty; shifting
it along a tightly sampled direction incurs a larger one. Thus $\Sigma$
sets both the exploration scale and the implicit effort metric. The
temperature $\lambda$ scales the strength of this penalty, giving the
quadratic weight
\begin{equation*}
  R=\lambda\Sigma^{-1}.
\end{equation*}
The distributional objective, restricted to this Gaussian family, is
\begin{equation}
  \mathcal J(Q_U)
  =\E_{Q_U}[S(V)]
    +\frac12\sum_{i=0}^{N-1}u_i\trans R u_i.
  \label{eq:mppi-matched-control-cost}
\end{equation}
Here the first term averages rollout cost over the sampled sequences, while
the second penalizes moving the sampling mean away from the zero-mean
reference. This gives a stochastic expected-cost interpretation when input
deviations really follow the assumed Gaussian distribution. When they are
artificial search noise, it instead defines a smoothed, regularized
optimization objective. In either case, the sampling covariance need not be
a measured physical disturbance covariance.

\subsection{Approximating the Ideal Measure by a Gaussian Proposal}

We generally cannot draw samples directly from the ideal distribution
$Q^\star$. Its density is
$q^\star(V)=Z^{-1}\exp\!\left(-S(V)/\lambda\right)p(V)$, but computing
its normalizing constant $Z$ is generally intractable. Markov chain Monte
Carlo (MCMC) methods, including Langevin-based methods, can target an
unnormalized density, but may require many costly evaluations of the rollout
cost $S(V)$. Gradient-based Langevin methods also require derivatives that
may not be available.

Instead of sampling $Q^\star$ directly, information-theoretic MPPI
approximates it with an easy-to-sample Gaussian measure $Q_U$
\cite{williams2018information}. The proposal uses the previously chosen
covariance $\Sigma$, which is held fixed in this projection; only the mean
sequence $U$ is optimized:
\begin{equation}
  U^{\mathrm{new}}
  =\argmin_U D_{\mathrm{KL}}(Q^\star\|Q_U).
  \label{eq:mppi-moment-projection}
\end{equation}
Notice the direction of KL. The distributional objective used
$D_{\mathrm{KL}}(Q\|P)$; this projection uses
$D_{\mathrm{KL}}(Q^\star\|Q_U)$.
It is not the same as minimizing $\mathcal J(Q_U)$ over $U$.

By definition,
\[
  D_{\mathrm{KL}}(Q^\star\|Q_U)
  =\E_{Q^\star}\!\left[\log q^\star(V)-\log q_U(V)\right].
\]
The term involving $q^\star$ is independent of $U$. In $q_U(V)$, the
normalizing constant is also independent of $U$ because $\bar\Sigma$ is
fixed. Thus only the quadratic term affects the minimizer, and the
projection reduces to
\begin{equation}
  \minimize_U\quad
  \mathcal L(U)
  :=\frac12\E_{Q^\star}\!\left[
    (V-U)\trans\bar\Sigma^{-1}(V-U)\right].
  \label{eq:mppi-gaussian-fit-objective}
\end{equation}
To find its minimizer, write $A=\bar\Sigma^{-1}$ and expand:
\begin{align*}
  \mathcal L(U)
  &=\frac12\E_{Q^\star}[V\trans A V]
    -U\trans A\E_{Q^\star}[V]
    +\frac12U\trans A U,\\
  \nabla_U\mathcal L(U)
  &=-A\E_{Q^\star}[V]+AU
   =A\!\left(U-\E_{Q^\star}[V]\right).
\end{align*}
Setting the gradient to zero and using that $A$ is invertible gives
$U=\E_{Q^\star}[V]$. The Hessian is $A\succ0$, so this is the unique
minimizer for unconstrained means with finite second moments. Componentwise,
the result is
\begin{equation}
  u_i^{\mathrm{new}}=\E_{Q^\star}[v_i].
  \label{eq:mppi-ideal-control-mean}
\end{equation}
This simple mean-matching update follows from the fixed-covariance Gaussian
assumption. The general nonlinear trajectory-optimization problem is much
harder and may be intractable. Also, if the ideal distribution has several
modes, their mean can produce a poor control sequence.

\subsection{Importance Sampling Produces the MPPI Weights}

The Gaussian projection above says that the next nominal input at stage
$i$ is the target mean $\E_{Q^\star}[v_i]$. We usually cannot draw directly
from $Q^\star$, so draw $K$ candidate sequences from the current proposal
$Q_U$: $V^{(j)}=U+\epsilon^{(j)}$. The change-of-measure step is to multiply
and divide the integrand by $q_U(V)$. Applying it to $h(V)=v_i$, then
substituting $q^\star(V)=Z^{-1}e^{-S(V)/\lambda}p(V)$, gives
\begin{align}
  \E_{Q^\star}[v_i]
  &=\int v_iq^\star(V)\,dV\\
  &=\int v_i\frac{q^\star(V)}{q_U(V)}q_U(V)\,dV\\
  &=\E_{Q_U}\!\left[v_i\frac{q^\star(V)}{q_U(V)}\right]\\
  &=\frac{1}{Z}\E_{Q_U}\!\left[
      v_i e^{-S(V)/\lambda}\frac{p(V)}{q_U(V)}\right]\\
  &=\frac{\E_{Q_U}\!\left[
      v_i e^{-S(V)/\lambda}p(V)/q_U(V)\right]}
         {\E_{Q_U}\!\left[e^{-S(V)/\lambda}p(V)/q_U(V)\right]}.
  \label{eq:mppi-importance-control-mean}
\end{align}
The last step uses the fact that the denominator is exactly $Z$:
\begin{align*}
  \E_{Q_U}\!\left[e^{-S(V)/\lambda}\frac{p(V)}{q_U(V)}\right]
  &=\int e^{-S(V)/\lambda}\frac{p(V)}{q_U(V)}q_U(V)\,dV\\
  &=\int e^{-S(V)/\lambda}p(V)\,dV\\
  &=Z.
\end{align*}
Thus the unknown normalizer cancels. Here $q^\star/q_U$ means a ratio of
densities: the target is corrected for the fact that candidates were drawn
from $Q_U$, not from $Q^\star$.

Next compute this density ratio. The reference $p(V)$ is centered at zero,
while the proposal $q_U(V)$ is centered at $U$; they have the same covariance
$\Sigma$. Their Gaussian normalization factors therefore cancel. Write
$A=\Sigma^{-1}$ and use $v_i=u_i+\epsilon_i$. The quadratic terms expand as
\begin{align}
  \log\frac{p(V)}{q_U(V)}
  &=-\frac12\sum_i
      \left[v_i\trans A v_i
        -(v_i-u_i)\trans A(v_i-u_i)\right]\nonumber\\
  &=-\frac12\sum_i
      \left[(u_i+\epsilon_i)\trans A(u_i+\epsilon_i)
        -\epsilon_i\trans A\epsilon_i\right]\nonumber\\
  &=-\frac12\sum_i
      \left[u_i\trans A u_i
        +u_i\trans A\epsilon_i
        +\epsilon_i\trans A u_i\right]\nonumber\\
  &=-\frac12\sum_i
      \left[u_i\trans A u_i+2u_i\trans A\epsilon_i\right]\nonumber\\
  &=-\sum_i u_i\trans\Sigma^{-1}\epsilon_i
    -\frac12\sum_i u_i\trans\Sigma^{-1}u_i.
  \label{eq:mppi-gaussian-density-ratio}
\end{align}
In the penultimate line, $A$ is symmetric, so the two scalar cross terms
are equal. For a candidate $j$, define its unnormalized importance weight
by multiplying this density ratio by its Gibbs cost factor:
\begin{align*}
  \alpha_j
  &:=e^{-S(V^{(j)})/\lambda}
       \frac{p(V^{(j)})}{q_U(V^{(j)})}\\
  &=\exp\!\left[-\frac{S(V^{(j)})}{\lambda}
       -\sum_i u_i\trans\Sigma^{-1}\epsilon_i^{(j)}
       -\frac12 c\right],
  \qquad
  c:=\sum_i u_i\trans\Sigma^{-1}u_i.
\end{align*}
The nominal sequence $U$ and covariance $\Sigma$ are fixed during one
sampling batch, so $c$ is the same for every candidate. Define the corrected
sample score
\begin{equation}
  \widetilde S_j
    :=S(V^{(j)})
      +\lambda\sum_{i=0}^{N-1}
         u_i\trans\Sigma^{-1}\epsilon_i^{(j)}.
  \label{eq:mppi-corrected-sample-cost}
\end{equation}
Then the unnormalized weight factors as
\begin{equation*}
  \alpha_j
  =\exp(-\widetilde S_j/\lambda)\exp(-c/2).
\end{equation*}
The factor $\exp(-c/2)$ is common to all candidates and cancels when we
normalize the weights. Replacing the expectations in
\cref{eq:mppi-importance-control-mean} with sample averages also introduces
a factor $1/K$ in both numerator and denominator; that factor cancels too.
In detail, the self-normalized estimate of the target mean is
\begin{align*}
  \widehat{\E}_{Q^\star}[v_i]
  &:=\frac{K^{-1}\sum_{j=1}^K v_i^{(j)}\alpha_j}
           {K^{-1}\sum_{j=1}^K\alpha_j}\\
  &=\frac{\sum_{j=1}^K v_i^{(j)}\alpha_j}
          {\sum_{j=1}^K\alpha_j}
   =\sum_{j=1}^K\omega_jv_i^{(j)},
  \qquad
  \omega_j:=\frac{\alpha_j}{\sum_{\ell=1}^K\alpha_\ell}.
\end{align*}
Substituting the expression for $\alpha_j$ shows that the normalized
weights have the exponential form below. The controller uses this weighted
sample mean as its next nominal input:
\begin{align}
  \omega_j
  &=\frac{\alpha_j}{\sum_{\ell=1}^K\alpha_\ell}
   =\frac{\exp(-\widetilde S_j/\lambda)}
          {\sum_{\ell=1}^K\exp(-\widetilde S_\ell/\lambda)},\\
  u_i^{\mathrm{new}}
  &:=\sum_{j=1}^K\omega_jv_i^{(j)}\\
  &=\sum_{j=1}^K\omega_j\bigl(u_i+\epsilon_i^{(j)}\bigr)\\
  &=u_i\sum_{j=1}^K\omega_j
      +\sum_{j=1}^K\omega_j\epsilon_i^{(j)}\\
  &=u_i+\sum_{j=1}^K\omega_j\epsilon_i^{(j)}.
  \label{eq:mppi-basic-update}
\end{align}
The final equality uses $\sum_j\omega_j=1$. This is the finite-sample,
equal-covariance information-theoretic MPPI update: each candidate receives
one weight based on its whole-sequence score, and that weight is applied to
its perturbation at every stage. The update uses no derivative of the
dynamics or cost.

\begin{example}[An exactly computable scalar target]
Let $N=m=1$, $P=\mathcal N(0,\sigma^2)$, and
$S(v)=\tfrac12a(v-b)^2$ with $a>0$. Completing the square in the Gibbs
density gives
\begin{equation}
  \E_{Q^\star}[v]
    =\frac{a\sigma^2}{\lambda+a\sigma^2}b,
  \qquad
  \operatorname{Var}_{Q^\star}(v)
    =\frac{\lambda\sigma^2}{\lambda+a\sigma^2}.
  \label{eq:mppi-scalar-gaussian-target}
\end{equation}
The reference pulls the mean toward zero, while the task cost pulls it
toward $b$. A Monte Carlo MPPI update should approach this mean from any
proposal that provides adequate coverage. For this quadratic example,
the mean also minimizes
$\tfrac12a(u-b)^2+\tfrac12\lambda u^2/\sigma^2$; this convenient coincidence
should not be generalized to nonlinear, multimodal tasks.
\end{example}

\subsection{Why the Name Path Integral?}

The earlier derivation sums over candidate input sequences and the
trajectories they produce. Classical path-integral control comes from
continuous-time stochastic control, where a system evolves under both
control and random disturbances. Its dynamic-programming equation tracks
the best expected future cost.

In some systems, control and disturbance enter in a compatible way. Under
this special condition, an exponential transformation simplifies the
dynamic-programming equation into a linear one. The transformed value can
then be represented as an average over uncontrolled trajectories, with
lower-cost paths contributing more. Averaging over complete paths—not
integrating over a spatial grid—is the classical path-integral idea
\cite{kappen2005linear,theodorou2010generalized}.

This connection has specific assumptions. Classical path-integral control
requires a particular relationship between how control affects the system
and how noise enters it. The finite-sequence Gibbs derivation and Gaussian
mean update earlier in this chapter do not require that relationship, so
they can be used with general simulators and learned models. This broader
use of sampling does not, by itself, guarantee the classical continuous-time
optimal-control result. Williams and colleagues discuss the connection and
its practical MPPI formulation \cite{williams2017parallel,williams2018information}.

Finally, discrete input perturbations and continuous-time process noise
are not interchangeable without accounting for the time step. The two
approaches share the idea of favoring lower-cost paths, but their
assumptions and guarantees differ.

\section{Implementing MPPI in a Receding-Horizon Loop}
\label{sec:mppi-implementation}

\subsection{Stable Weights and an Explicit Algorithm}

Directly exponentiating large negative scores can underflow. For the
finite valid scores, let $\rho=\min_j\widetilde S_j$ and compute
\begin{equation}
  b_j=\exp\!\left(-\frac{\widetilde S_j-\rho}{\lambda}\right),
  \qquad \omega_j=\frac{b_j}{\sum_\ell b_\ell}.
  \label{eq:mppi-stable-weights}
\end{equation}
The common shift cancels exactly, so it changes numerical conditioning,
not the target distribution. Invalid rollouts can be assigned zero weight.
If every rollout is invalid, there is no finite baseline or meaningful
weighted update; the controller must take its specified fallback branch.

The following pseudocode states the mathematical core and the execution
checks separately. The core assumes unconstrained Gaussian input samples
with equal reference and proposal covariance, as derived above.

\begin{quote}
\textbf{MPPI at one MPC update}
\begin{enumerate}
  \item Read the timestamped estimate $\hat x_k$, current reference, and
        environment prediction. Initialize the mean sequence $U$ from a
        shifted previous plan or another available warm start.
  \item For each allowed optimization batch:
    \begin{enumerate}
      \item Freeze the current $U$. For $j=1,\ldots,K$, independently draw
            $\epsilon_i^{(j)}\sim\mathcal N(0,\Sigma)$ and set
            $v_i^{(j)}=u_i+\epsilon_i^{(j)}$.
      \item In parallel across samples, initialize $x_0^{(j)}=\hat x_k$,
            roll out the $N$ transitions, and accumulate $S(V^{(j)})$.
            Also record constraint violations and invalid model evaluations.
      \item Compute the corrected scores in
            \cref{eq:mppi-corrected-sample-cost}. If no sample is usable,
            return to the failure-handling procedure.
      \item Compute \cref{eq:mppi-stable-weights} and update every stage
            with \cref{eq:mppi-basic-update}. Use the same frozen $U$
            in the proposal, likelihood ratio, and update.
    \end{enumerate}
  \item Roll out and validate the updated mean plan. Accept it only if it
        meets the deployment\textquotesingle s admissibility and timing tests;
        otherwise use an appropriate checked incumbent or backup policy.
  \item Apply the first accepted input. Store the remaining plan for the
        next update, append a suitable tail, and repeat from new measurements.
\end{enumerate}
\end{quote}

Step~3 is necessary because averaging candidates is not itself a feasibility
test. A deployment with bounded inputs or transformed control variables
must use the corresponding sampling distribution, rather than assuming the
unconstrained core automatically handles those changes.

\subsection{Warm Starts, Horizon Length, and the Sampling Budget}

At each MPC update, shift the accepted plan forward and append a suitable
tail:
\begin{equation}
  U_{k+1}^{\mathrm{init}}
  =(u_{1|k},\ldots,u_{N-1|k},u_{\mathrm{tail}}).
  \label{eq:mppi-shifted-warm-start}
\end{equation}
The tail can come from a local controller or reference. New measurements,
obstacle motion, or a task change can invalidate the shifted plan. The
sample count, horizon, and number of batches share a fixed compute budget:
more samples improve coverage, a longer horizon predicts farther ahead,
and more batches recenter the proposal. Keep a checked incumbent as a
separate acceptance or fallback candidate; the MPPI mean update guarantees
neither monotonic improvement nor recursive feasibility.

\subsection{Temperature, Covariance, and Weight Concentration}

For fixed corrected scores, lower $\lambda$ concentrates weight on the
best samples. But changing $\lambda$ also changes the target distribution;
as it grows, the target approaches the reference $P$, not necessarily the
current proposal $Q_U$. The effective sample size summarizes weight
concentration:
\begin{equation}
  K_{\mathrm{eff}}=\frac1{\sum_{j=1}^K\omega_j^2},
  \qquad 1\leq K_{\mathrm{eff}}\leq K.
  \label{eq:mppi-effective-sample-size}
\end{equation}
It is a diagnostic, not a measure of feasible paths or a guarantee of
accuracy. Small covariance limits search to a local region; large
covariance explores more broadly but may generate mostly invalid samples.
Scale perturbations by input, monitor both weights and task progress, and
recompute the likelihood ratio whenever the proposal covariance changes.
Changing the reference itself changes the regularized problem.

\subsection{Smooth Sequences and Lower-Dimensional Search}

Independent stagewise noise can produce jagged controls. Instead, sample a
small vector of spline or knot coefficients and interpolate to a full
sequence, for example
\begin{equation}
  V=L\theta+V_{\mathrm{base}},
  \label{eq:mppi-control-parameterization}
\end{equation}
where the coefficient dimension is smaller than the full sequence
dimension. Define the reference and proposal in this coefficient space.
Temporal correlations are another option, using a positive-definite
sequence covariance. Do not smooth an update after scoring without rerolling
and validating the modified plan; a lower-dimensional search can also miss
useful fast corrections.

\subsection{Constraints, Clipping, and Multimodal Plans}

Constraints can be represented by finite penalties or by assigning
infeasible candidates infinite cost. For a hard feasible set $\mathcal A$,
the target is proportional to
\begin{equation}
  \mathbf 1_{\mathcal A}(V)e^{-S(V)/\lambda}p(V).
  \label{eq:mppi-hard-constraint-target}
\end{equation}
The set must have positive probability under the reference; exact equalities
are generally missed by unconstrained continuous sampling. Penalties allow
tradeoffs and are not hard guarantees.

Clipping changes the sampling distribution. Either define the proposal and
weights on latent variables and apply clipping inside the rollout, or use
the correct density for a truncated proposal. In every case, validate the
plan that will actually be executed. Even feasible sampled paths can have
an infeasible average—for example, paths that pass on opposite sides of an
obstacle may average into a collision. A checked sample or safety supervisor
may be preferable to the mean in a multimodal problem.

\subsection{Model Uncertainty and Failure Handling}

More samples do not correct model bias. Monitor prediction residuals,
constraint margins, and whether commands are realizable. If uncertainty
matters, specify how it enters the objective; for example, worst-case and
mean-cost criteria are different:
\begin{equation}
  S_{\mathrm{rob}}(V)=\max_{\xi\in\Xi}J(V;\xi),
  \qquad
  S_{\mathrm{mean}}(V)=\E_\xi[J(V;\xi)].
  \label{eq:mppi-uncertain-candidate-costs}
\end{equation}
One favorable disturbance sample is not a robust evaluation, and averaging
cost before exponential weighting differs from averaging the weights.
Define fallback behavior for invalid samples, missed deadlines, and rejected
mean plans. A stored plan or backup controller is useful only within its
verified operating conditions; safety remains a system-level requirement.

\section{Robotics Examples}
\label{sec:mppi-robotics-examples}

These examples show how the predictor and control representation shape the
sampling problem; their details are illustrative, not exact reproductions
of the cited implementations.

\subsection{Ground-Robot Navigation}

A car-like robot can be planned with a low-speed bicycle model or a learned
predictor, scoring reference tracking, speed, and obstacle clearance.
Higher-speed driving needs a model that captures tire dynamics. Update
moving-obstacle forecasts and validate the robot footprint between sampled
time steps; no sample count can create a safe path when actuation or
clearance makes one impossible. Williams and colleagues demonstrate
information-theoretic MPPI on an autonomous ground vehicle using learned
dynamics \cite{williams2018information}.

\subsection{Object Pushing with a Learned Contact Model}

A learned predictor can forecast object motion and uncertainty through
sliding, sticking, and loss of contact. MPPI can score goal progress,
collisions, and predictive uncertainty, then replan after each short push.
An uncertainty penalty is a planning preference, not a calibrated success
probability; a useful forward model is still essential. Arruda and
colleagues demonstrate uncertainty-aware MPPI pushing on a real robot
\cite{arruda2017pushing}.

\subsection{Whole-Body Locomotion and Manipulation}

For legged robots, sample a small number of joint-target knots, track them
with a low-level controller, and simulate the resulting body motion,
actuator limits, and contacts. Scores can include body tracking, gait
regularity, joint limits, and manipulation goals. On a Unitree Go1, Alvarez
and colleagues use joint-target splines, PD tracking, gait references, and
parallel MuJoCo rollouts on a multicore CPU; their results illustrate why
reduced search dimension and low rollout latency can matter more than
maximizing sample count \cite{alvarez2025wholebody}.

Across these applications, MPPI is MPC because it predicts future motion,
executes the first accepted action, and replans from new measurements—not
simply because it repeatedly solves an optimization problem.

\paragraph{Exercises.}

\begin{exercise}
Derive \cref{eq:sampling-importance-identity}. Why is support coverage
required?
\end{exercise}
\begin{proof}[Short solution]
Write $\int hp=\int h(p/q)q$. If $q$ vanishes on a set with positive target
probability, samples cannot visit that set and its contribution cannot be
estimated through the ratio.
\end{proof}

\begin{exercise}
Prove the Gibbs variational identity and identify the optimal distribution.
Does it imply that its mean is a globally optimal deterministic control?
\end{exercise}
\begin{proof}[Short solution]
Substitute $q^\star=e^{-S/\lambda}p/Z$ into
$D_{\mathrm{KL}}(Q\|Q^\star)$ and rearrange to obtain
$\mathcal J(Q)=\mathcal F+\lambda D_{\mathrm{KL}}(Q\|Q^\star)$.
The measure $Q^\star$ is optimal for the distributional objective. A nonlinear
rollout of its mean can differ greatly from its low-cost sampled rollouts,
so deterministic optimality does not follow.
\end{proof}

\begin{exercise}
Derive the Gaussian log-density ratio in
\cref{eq:mppi-gaussian-density-ratio}. Which term cancels in normalized weights?
\end{exercise}
\begin{proof}[Short solution]
Expand $v_i=u_i+\epsilon_i$ in the two quadratic exponents.
The difference is
$-u_i\trans\Sigma^{-1}\epsilon_i-\tfrac12u_i\trans\Sigma^{-1}u_i$.
The second term is independent of the sample and cancels in the weight ratio.
\end{proof}

\begin{exercise}
For the scalar target in \cref{eq:mppi-scalar-gaussian-target}, take
$a=2$, $b=1$, $\sigma^2=1$, and $\lambda=1$. Compute the mean and variance.
\end{exercise}
\begin{proof}[Short solution]
The target mean is $2/3$ and its variance is $1/3$. An adequately sampled
importance-weighted update should approach the mean, not necessarily the
unregularized task minimizer $v=1$.
\end{proof}

\begin{exercise}
Why does subtracting the minimum corrected score leave MPPI weights
unchanged? What if every score is infinite?
\end{exercise}
\begin{proof}[Short solution]
The numerator and denominator acquire the same factor $e^{\rho/\lambda}$,
which cancels. If all scores are infinite, there is no valid finite baseline
or normalized distribution over usable candidates; invoke failure handling.
\end{proof}

\begin{exercise}
Show that $K_{\mathrm{eff}}=K$ for equal weights and approaches one when
one weight approaches one. Does either value establish safety?
\end{exercise}
\begin{proof}[Short solution]
Equal weights give $\sum_j\omega_j^2=1/K$. A dominant weight gives a sum
approaching one. Neither quantity verifies constraints or model accuracy.
\end{proof}

\begin{exercise}
Explain why two feasible sampled paths around an obstacle can have an
infeasible average control sequence. What should be checked before execution?
\end{exercise}
\begin{proof}[Short solution]
Nonlinear trajectory feasibility is not generally convex in the inputs.
Opposite avoidance motions can average into a collision course. Roll out
and validate the sequence actually intended for execution, including any
filtering, clipping, or safety modification.
\end{proof}

\begin{exercise}
Which parts of sampling-based MPC can benefit from a GPU or NPU, and why
does high throughput not imply a short control-loop latency?
\end{exercise}
\begin{proof}[Short solution]
A GPU can parallelize compatible simulations and reductions. An NPU can
accelerate supported neural inference. The horizon still has sequential
state dependencies, and estimation, transfers, obstacle queries, and
communication also consume time. Measure the complete update deadline.
\end{proof}

\paragraph{Chapter summary.}
Sampling-based MPC uses many forward predictions to search for a control
sequence without differentiating its rollout. Random shooting selects
a tested candidate; CEM refits a proposal to elites; MPPI uses weighted
moments. GPU-parallel rollouts and accelerated neural inference can make
these methods practical, but cannot remove poor model accuracy, weak
sampling coverage, or execution delays.

The information-theoretic formulation minimizes expected trajectory cost
plus KL divergence from a reference measure. Its exact distributional solution
is a Gibbs density. Fitting an equal-covariance Gaussian to that target
gives a mean-matching update, and importance sampling turns the unknown
expectation into exponentially weighted perturbations. The Gaussian
likelihood ratio supplies the control--noise cross term. Keeping the
reference, target, and proposal distinct prevents misleading cost formulas
and optimality claims.

MPPI remains a finite-horizon, locally sampled controller. Constraint
penalties, input clipping, smooth parameterizations, and uncertain models
must be handled consistently; an averaged sequence is not automatically
feasible. The robotics examples show how a suitable action representation
and predictor matter as much as the sample count.

The next part changes the problem from repeatedly optimizing with a given
model to learning values and policies through experience. The expectation
notation and distributional reasoning introduced here will reappear in
Markov decision processes, Bellman equations, and reinforcement learning.

\part{From Optimal Control to Reinforcement Learning}
\chapter{Markov Decision Processes and Bellman Equations}
\label{chap:mdps-bellman-equations}
\chapteroverview{Continuous-state and continuous-action MDPs, returns and values, Bellman equations, model-based planning, and connections to optimal control.}

\section{Why Learn a Policy After Model Predictive Control?}

Chapters~10 and~11 introduced model predictive control (MPC) and its
sampling-based variant, MPPI. Both use a state estimate and a model to plan
over a finite horizon, execute an action, and replan. On a robot, each update
must finish before its control deadline: state estimation, trajectory
optimization or rollout evaluation, constraint checks, and communication all
consume time. Parallel sampling can increase throughput, but it does not make
the computation free. The plan can also be sensitive to errors in the
estimated state or prediction model. In particular, an optimizer may favor
actions that exploit a model error.

Contact-rich manipulation and locomotion make online optimization harder.
Sticking, slipping, impacts, and stance changes create hybrid dynamics; the
unknown contact sequence introduces discrete choices, while the resulting
constraints and transitions may be nonsmooth. Contact-aware trajectory
optimization can address these effects, but repeatedly solving such problems
within a tight control-cycle budget remains challenging
\cite{posa2014direct,patel2019contact}.

Deep reinforcement learning (RL) shifts some computation from control time
to training time. It uses experience to learn a \emph{policy}: a rule that
maps the robot's observations to actions. Once trained, the policy produces
actions by evaluating a neural network, rather than searching through many
action sequences at every control step. Training cost is therefore spread
across the policy's later use. Domain randomization can help bridge the gap
between simulation and the real robot by varying properties such as mass,
friction, sensor noise, and delay during training. It improves robustness to
the variations represented in training, but does not guarantee robustness to
every possible condition \cite{peng2018dynamicsrandomization}.

The two approaches have complementary strengths. MPC can adapt its plan to
new measured states and goals, provided its model is accurate enough and its
optimization finishes within the control deadline. It remains vulnerable to
model and state-estimation errors, and contact-rich planning can be difficult
because contact changes create hybrid, nonsmooth dynamics and discrete
choices \cite{posa2014direct,patel2019contact}. A learned policy is fast at
deployment, but its behavior depends on what it learned. Combining the
approaches—for example, using MPC to guide policy training or a policy to
initialize online planning—is an active direction
\cite{zhang2016mpcguided}.

This chapter introduces the basic language of RL and connects it to ideas
from optimal control. A key limitation to keep in mind is that a learned
policy may fail when deployed in out-of-distribution (OOD) conditions, where
the observations, contacts, or dynamics differ substantially from those
encountered during training.

\section{Markov Decision Processes}
\label{sec:mdp-definition}

A Markov decision process (MDP) describes how a controller chooses actions
when the outcome may be uncertain. This chapter focuses on continuous
robotic states and controls. Model-based planning uses a model to predict
outcomes; reinforcement learning (RL) uses experience to evaluate or improve
decisions and can do so without a complete model. Standard treatments include
\cite{sutton2018reinforcement,bertsekas2017dynamic}.

\subsection{States, Actions, Transitions, and Rewards}
\label{subsec:mdp-components}

Let the state be a vector $s\in\mathcal S\subseteq\R^n$ and the action be a
vector $a\in\mathcal A(s)\subseteq\R^m$. Both spaces are usually
continuous for a physical robot. For example, a manipulator state can contain
joint positions and velocities, while its action can be a vector of torques.
The admissible action set may depend on the state through actuator, contact,
or safety limits.

The dynamics are described by a transition model. In a stochastic MDP this
is a probability kernel $P(B\mid s,a)$: the probability that the next state
falls in a region $B\subseteq\mathcal S$, given the current state and action.
If the transition has a density, we write $p(s'\mid s,a)$, so that
\begin{equation}
  P(S_{t+1}\in B\mid S_t=s,A_t=a)
  =\int_B p(s'\mid s,a)\,ds'.
  \label{eq:mdp-transition-density}
\end{equation}
In later equations, $P(ds'\mid s,a)$ means integration with respect to this
conditional next-state distribution. For any suitable function $f$, the
integral
\[
  \int_{\mathcal S}f(s')P(ds'\mid s,a)
  =\E[f(S_{t+1})\mid S_t=s,A_t=a]
\]
is its conditional expected value. The notation $P(ds'\mid s,a)$ denotes a
measure, not a density or the probability of one exact state. If a density
exists, then $P(ds'\mid s,a)=p(s'\mid s,a)\,ds'$, yielding the usual density
integral. For discrete states, the integral becomes a sum over next states,
weighted by their transition probabilities. Deterministic dynamics are also
included: their transition distribution is a point mass at $f(s,a)$.

The state should retain all information from the past needed to predict
future transitions and rewards. This is the Markov property: once the
current state and action are given, the earlier history provides no further
information about the next state and reward. For $t\geq1$, let
$H_t=(S_0,A_0,R_1,S_1,\ldots,A_{t-1},R_t,S_t)$ denote the history through
state $S_t$, and let $H_0=S_0$. For every measurable region
$B\subseteq\mathcal S$ and reward set $C\subseteq\R$, the property is
\begin{equation}
  P(S_{t+1}\in B,\ R_{t+1}\in C\mid H_t,A_t)
  =P(S_{t+1}\in B,\ R_{t+1}\in C\mid S_t,A_t).
  \label{eq:mdp-markov-property}
\end{equation}
A raw sensor reading need not have this property. A filtered estimate may
be sufficient; otherwise, augment the state with relevant variables such as
velocity, actuator temperature, or previous control. Under partial
observations, a belief over underlying states can be used instead.

In practice, a short history of observations and actions is often
concatenated and used as an approximate state. This can supply context, such
as motion inferred from successive positions, but does not guarantee a
Markov state \cite{sutton2018reinforcement}.

A reward $r(s,a,s')$ scores the immediate outcome of applying $a$ at $s$ and
reaching $s'$. If a transition density exists, its expected immediate reward
is
\begin{equation}
  \bar r(s,a)=\int_{\mathcal S}p(s'\mid s,a)r(s,a,s')\,ds'.
  \label{eq:mdp-expected-reward}
\end{equation}
The transition model plays the role of a dynamics model, but can describe a
range of possible next states rather than a single prediction.

\begin{example}[A continuous robot docking task]
Let $p_t$ be a mobile robot's signed position error relative to a docking
station and $v_t$ its velocity. The continuous state is
$s_t=(p_t,v_t)\in\R^2$, and the continuous action is an acceleration
command $a_t\in[-a_{\max},a_{\max}]$. A simple noisy model is
\begin{align}
  p_{t+1}&=p_t+h v_t+\eta^p_t,\\
  v_{t+1}&=v_t+h a_t+\eta^v_t,
  \label{eq:mdp-docking-dynamics}
\end{align}
where $h$ is the sample interval and the noise vectors
$\eta_t=(\eta^p_t,\eta^v_t)$ are independent across time and Gaussian with
zero mean and positive-definite covariance $Q$. With positive weights
$q_p,q_v,r_a$, one possible stage reward is
\begin{equation}
  r(s_t,a_t,s_{t+1})
  =-\left(q_p p_{t+1}^2+q_v v_{t+1}^2+r_a a_t^2\right).
  \label{eq:mdp-docking-reward}
\end{equation}
A terminal reward can further favor ending near $p=0$ with low velocity.
The model has continuously many states and actions; the transition density
averages over noise, and the controller selects an acceleration from an
interval. This simple example will illustrate the continuous Bellman
formulas below.
\end{example}

\subsection{Policies and Trajectories}
\label{subsec:mdp-policies}

A policy specifies how actions are chosen. For continuous actions, a
stochastic policy is a density $\pi(a\mid s)$ satisfying
\begin{equation}
  \pi(a\mid s)\geq0,
  \qquad \int_{\mathcal A(s)}\pi(a\mid s)\,da=1.
  \label{eq:mdp-continuous-policy}
\end{equation}
The probability that the policy chooses an action in a region $A$ is the
integral of this density over $A$. A deterministic feedback policy
$a=\mu(s)$ is another kind of policy; it maps each state to a single
continuous action. Thus, the optimal-control feedback law and the RL policy
play the same basic role, though an RL policy may deliberately randomize.

Starting from an initial state, the policy and transition model generate a
trajectory of states, actions, and rewards. For the docking robot this is a
sequence of positions, velocities, acceleration commands, and noisy motion.
Because the dynamics are stochastic, the same policy and starting state may
produce different trajectories. Write a finite state--action trajectory as
$\tau=(s_0,a_0,s_1,a_1,\ldots,a_{T-1},s_T)$, and let $\rho_0(s_0)$ be the
density of the initial state. For a stochastic policy with density $\pi$ and
a transition model with density $p$, the trajectory density is
\begin{equation}
  p_\pi(\tau)=\rho_0(s_0)
    \prod_{t=0}^{T-1}
      \pi(a_t\mid s_t)\,p(s_{t+1}\mid s_t,a_t).
  \label{eq:mdp-trajectory-density}
\end{equation}
If the initial state is fixed, condition on $s_0$ and omit the factor
$\rho_0(s_0)$.
Each factor $\pi(a_t\mid s_t)$ describes the policy's action choice, and
$p(s_{t+1}\mid s_t,a_t)$ describes the resulting state transition. Their
product gives the joint density by repeatedly applying the conditional
probability rule and the Markov property. Because the states and actions are
continuous, $p_\pi(\tau)$ is a density, not the probability of that exact
trajectory; probabilities are obtained by integrating over a region of
trajectories. Rewards are assumed to be determined by each
state--action--next-state triple. If rewards are random and included in the
trajectory, augment $\tau$ with rewards and include their conditional law.

\subsection{Finite and Discounted Infinite Horizons}
\label{subsec:mdp-horizons}

The \emph{return} is the total reward from a given time onward. For an episode
ending at time $T$, with terminal reward $g(S_T)$ and discount factor
$0\leq\gamma\leq1$, the discounted return is
\begin{equation}
  G_t=\sum_{j=0}^{T-t-1}\gamma^j R_{t+j+1}
       +\gamma^{T-t}g(S_T).
  \label{eq:mdp-finite-return}
\end{equation}
A finite task may use $\gamma=1$; discounting is optional. Its value can
depend on time because fewer decisions remain near the deadline, as in a
finite-horizon optimal-control problem with a terminal cost.

For a continuing task, the discounted infinite-horizon return is
\begin{equation}
  G_t=\sum_{j=0}^{\infty}\gamma^j R_{t+j+1},
  \qquad 0\leq\gamma<1.
  \label{eq:mdp-infinite-return}
\end{equation}
When rewards are bounded, discounting makes the infinite sum converge
absolutely and reduces the influence of distant outcomes. A value of $\gamma$
close to one emphasizes long-term behavior; a smaller value emphasizes nearer
outcomes. The discount factor is part of the objective, not the physical
sample interval.

RL commonly maximizes reward, whereas optimal control usually minimizes
cost. Setting reward equal to negative cost makes maximizing expected return
equivalent to minimizing expected accumulated cost. We use reward notation
for the MDP equations and translate back to costs when making connections to
optimal control.

\section{Value Functions and Bellman Equations}
\label{sec:mdp-values}

A trajectory gives one possible outcome; a value function averages over the
outcomes that may follow. It describes expected future performance from a
state or state--action pair. When reward is negative cost, a value is the
negative of the corresponding cost-to-go, whether evaluating a fixed policy
or an optimal one.

\subsection{State and Action Values}
\label{subsec:mdp-state-action-values}

First consider a stationary policy in a discounted continuing task. Its
state-value function is the expected return from state $s$:
\begin{equation}
  v^\pi(s)=\E_\pi[G_t\mid S_t=s].
  \label{eq:mdp-state-value}
\end{equation}
Its action-value function, also called the $Q$-value, is the expected return
when the first action is fixed to $a$ and the policy is followed afterward:
\begin{equation}
  q^\pi(s,a)=\E_\pi[G_t\mid S_t=s,A_t=a].
  \label{eq:mdp-action-value}
\end{equation}
For a stochastic policy with density $\pi$, the state value averages its
action values according to
\begin{equation}
  v^\pi(s)=\int_{\mathcal A(s)}\pi(a\mid s)q^\pi(s,a)\,da.
  \label{eq:mdp-value-action-average}
\end{equation}
For a deterministic feedback policy, the corresponding action average is
simply $q^\pi(s,\mu(s))$. In either case, the value averages over uncertain
future transitions; a stochastic policy also averages over its action choices.

\subsection{Bellman Expectation Equations}
\label{subsec:mdp-bellman-expectation}

Separate the return into the immediate reward and the discounted return
remaining after one transition. For a stochastic policy and a transition
density, conditioning on the first action and next state gives
\begin{equation}
  v^\pi(s)=\int_{\mathcal A(s)}\pi(a\mid s)
    \int_{\mathcal S}p(s'\mid s,a)
      \left[r(s,a,s')+\gamma v^\pi(s')\right]ds'\,da.
  \label{eq:mdp-bellman-expectation-v}
\end{equation}
For a fixed first action, the corresponding action-value equation is
\begin{equation}
  q^\pi(s,a)=\int_{\mathcal S}p(s'\mid s,a)
    \left[r(s,a,s')+\gamma
      \int_{\mathcal A(s')}\pi(a'\mid s')q^\pi(s',a')\,da'\right]ds'.
  \label{eq:mdp-bellman-expectation-q}
\end{equation}
These are expectation equations: they evaluate a policy already chosen,
rather than select the best policy. They are coupled equations over
functions, because each value depends on values at a continuum of successor
states. For deterministic transitions, the expectation with respect to the
transition law reduces to substituting $s'=f(s,a)$.

The recursive structure is the same dynamic-programming idea used in
Chapter~8: account for what happens now, then add the value of continuing
from the next state (\cref{eq:deterministic-discrete-bellman-recursion}).
The important distinction is which recursion is being computed. The
Bellman expectation equations above evaluate a specified policy; their
optimal-control counterpart would compute the cost-to-go of a prescribed
feedback law, without optimizing over controls. By contrast, Chapter~8's
Bellman recursion computes the \emph{optimal} cost-to-go by minimizing over
admissible controls. It therefore corresponds to the Bellman optimality
equations below, after replacing reward by negative cost, rather than to the
expectation equations above. In the MDP, stochastic transitions replace the
deterministic next-state substitution with an expectation over possible next
states, and a stochastic policy adds an expectation over its action choices.
With deterministic dynamics and a fixed feedback policy, the expectation
equation reduces to the fixed-policy cost-to-go recursion. For a finite
horizon with $\gamma=1$, the optimality recursion reduces to the Chapter~8
minimization after the reward-to-cost sign change.

For a finite-horizon task, values also depend on time. With terminal reward
given by $v_T^\pi(s)=g(s)$, a policy is evaluated backward using
\begin{equation}
  v_t^\pi(s)=\int_{\mathcal A(s)}\pi_t(a\mid s)
    \int_{\mathcal S}p(s'\mid s,a)
      \left[r(s,a,s')+\gamma v_{t+1}^\pi(s')\right]ds'\,da,
  \qquad t=T-1,\ldots,0.
  \label{eq:mdp-finite-horizon-evaluation}
\end{equation}
The known terminal value supplies the boundary condition for the recursion.

\subsection{Bellman Optimality Equations}
\label{subsec:mdp-bellman-optimality}

The optimal state value is the greatest return achievable over all policies.
At each state, the controller chooses a continuous action and then continues
optimally from the resulting state. For a finite horizon, the terminal value
is $v_T^*(s)=g(s)$ and the backward optimality recursion is
\begin{equation}
  v_t^*(s)=\max_{a\in\mathcal A(s)}
    \int_{\mathcal S}p(s'\mid s,a)
      \left[r(s,a,s')+\gamma v_{t+1}^*(s')\right]ds',
  \qquad t=T-1,\ldots,0.
  \label{eq:mdp-finite-horizon-optimality}
\end{equation}
For a stationary discounted infinite-horizon task, the value no longer
depends on time:
\begin{align}
  v^*(s)
  &=\max_{a\in\mathcal A(s)}\int_{\mathcal S}p(s'\mid s,a)
      \left[r(s,a,s')+\gamma v^*(s')\right]ds',
      \label{eq:mdp-bellman-optimality-v}\\
  q^*(s,a)
  &=\int_{\mathcal S}p(s'\mid s,a)
      \left[r(s,a,s')+\gamma\max_{a'\in\mathcal A(s')}q^*(s',a')\right]ds'.
      \label{eq:mdp-bellman-optimality-q}
\end{align}
A maximizing action is greedy with respect to the value. Because the action
space is continuous, finding it is itself an optimization problem; when a
maximum is not attained, the supremum is used instead. The state integral
averages all possible outcomes, not just the most likely next state.

These equations express the discrete-time principle of optimality developed
in Chapter~8. An optimal continuation from the next state must itself be
optimal; otherwise, replacing it with a better continuation would improve
the full plan. The continuous MDP equations are the stochastic, reward-based
counterpart of dynamic programming \cite{bertsekas2017dynamic,sutton2018reinforcement}.

\section{Planning and Control with a Known Model}
\label{sec:mdp-planning}

When the transition model and reward are known, the Bellman equations can be
used to compute values and improve policies. This computation is called
planning; the resulting policy is a feedback controller when applied to the
robot. For continuous states and actions, exact dynamic programming is
usually impractical: values are functions over continuous state spaces, and
each backup may require continuous action optimization. The equations still
show what value and policy approximations should compute.

\subsection{Policy Evaluation and Policy Improvement}
\label{subsec:mdp-policy-iteration}

Policy evaluation computes $v^\pi(s)$ for a fixed policy from its Bellman
expectation equation. Policy improvement then chooses actions that are greedy
with respect to that value. For a deterministic policy, one such update is
\begin{equation}
  \mu_{\mathrm{new}}(s)\in\operatorname*{arg\,max}_{a\in\mathcal A(s)}
    \int_{\mathcal S}p(s'\mid s,a)
      \left[r(s,a,s')+\gamma v^\pi(s')\right]ds'.
  \label{eq:mdp-policy-improvement}
\end{equation}
Alternating evaluation and improvement is called policy iteration. In
continuous robot problems these steps generally require function
approximation and numerical action optimization, so exact tabular guarantees
do not automatically apply.

\subsection{Value Iteration}
\label{subsec:mdp-value-iteration}

Value iteration repeatedly applies the optimal Bellman backup. In a
continuous MDP, this operator maps a value function to an updated value
function:
\begin{equation}
  (\mathcal T v)(s)=\max_{a\in\mathcal A(s)}
    \int_{\mathcal S}p(s'\mid s,a)
      \left[r(s,a,s')+\gamma v(s')\right]ds',
  \qquad v_{k+1}=\mathcal T v_k.
  \label{eq:mdp-continuous-value-iteration}
\end{equation}
At each state, the backup averages immediate reward plus discounted
successor value over possible next states for each action, then selects the
best action. Applying it across states produces the next value function.
For the Bellman operator in
\cref{eq:mdp-continuous-value-iteration}, assume bounded rewards, nonempty
admissible action sets, and $0\leq\gamma<1$. Assume also that each backup
is well-defined, using a supremum if an action maximum is not attained.
Then $\mathcal T$ is a contraction on the space of bounded measurable value
functions under the sup norm:
\begin{equation}
  \begin{aligned}
    |(\mathcal T v)(s)-(\mathcal T w)(s)|
    &\leq\gamma\sup_{a\in\mathcal A(s)}
      \int_{\mathcal S}p(s'\mid s,a)|v(s')-w(s')|\,ds'\\
    &\leq\gamma\|v-w\|_\infty,\\
    \|\mathcal T v-\mathcal T w\|_\infty
    &\leq\gamma\|v-w\|_\infty.
  \end{aligned}
  \label{eq:mdp-value-iteration-contraction}
\end{equation}
The transition density integrates to one, so averaging $v-w$ over next
states cannot increase its largest absolute value; discounting scales that
bound by $\gamma$. Maximizing over actions also preserves the bound.
Thus, by the contraction mapping theorem, $\mathcal T$ has a unique fixed
point, the optimal value $v^*$, and exact value iteration converges to it.
This guarantee concerns the exact backup; approximating the value function
or action maximization for a high-dimensional robot can change the
convergence behavior.

\subsection{Connections to Optimal Control}
\label{subsec:mdp-optimal-control-connection}

The terminology maps directly to continuous optimal control. The reward
convention differs because RL maximizes while control usually minimizes.

\begin{center}
\begin{tabular}{@{}p{0.25\textwidth}p{0.31\textwidth}p{0.34\textwidth}@{}}
\toprule
Continuous MDP & Optimal-control language & Meaning \\
\midrule
State $s\in\mathcal S$ & State $x$ & Robot configuration and motion \\
Action $a\in\mathcal A(s)$ & Control $u$ & Continuous actuator command \\
Kernel $P(ds'\mid s,a)$ & Dynamics $x^+=f(x,u)+w$ & Deterministic or uncertain prediction \\
Reward $r$ & Negative stage cost $-\ell$ & Immediate task preference \\
Return $G_t$ & Negative accumulated cost $-J$ & Performance over a trajectory \\
Policy $\pi(a\mid s)$ & Feedback law $u=\mu(x)$ & State-to-control rule \\
Value $v^\pi(s)$ & Negative cost-to-go & Expected future performance \\
\bottomrule
\end{tabular}
\end{center}

For a finite horizon with undiscounted control cost, let $V_T(s)=\Phi(s)$ be
the terminal cost and $c(s,a,s')$ the transition cost. The cost-minimizing
Bellman recursion in MDP notation is
\begin{equation}
  V_t(s)=\min_{a\in\mathcal A(s)}
    \int_{\mathcal S}p(s'\mid s,a)
      \left[c(s,a,s')+V_{t+1}(s')\right]ds'.
  \label{eq:mdp-cost-bellman}
\end{equation}
For deterministic dynamics, this reduces to
$V_t(s)=\min_a\{c(s,a,f(s,a))+V_{t+1}(f(s,a))\}$. In optimal-control
notation, the same recursion is written
$V_t(x)=\min_u\{\ell(x,u)+V_{t+1}(f(x,u))\}$, as developed in Chapter~8.
Shrinking the sampling interval leads, under suitable conditions, to the
continuous-time HJB description from that chapter.

In the standard continuous-time optimal-control formulation used in
Chapter~8, one usually works directly with the value or cost-to-go $V(t,x)$
and does not introduce a separate action-value function analogous to the
discrete-time $Q$-value. This is a convention of the formulation, not a
mathematical restriction. To see the connection, hold a control $u$ fixed
for a short interval $\Delta t$, then follow the optimal continuation.
Let $x_u(\tau)$ denote the trajectory starting at $x$ at time $t$ under the
fixed control $u$. The cost over this interval plus optimal remaining cost is
\begin{equation}
  Q_{\Delta t}(t,x,u)
  =\int_t^{t+\Delta t}\ell(\tau,x_u(\tau),u)\,d\tau
    +V(t+\Delta t,x_u(t+\Delta t)).
\end{equation}
For smooth $f$, $\ell$, and $V$, a first-order expansion gives
\begin{equation}
  Q_{\Delta t}(t,x,u)
  =V(t,x)+\Delta t\left[
    V_t(t,x)+\ell(t,x,u)
    +\nabla_xV(t,x)\trans f(t,x,u)\right]+o(\Delta t).
\end{equation}
At fixed $(t,x)$, $V$ and $V_t$ do not depend on the chosen control.
Therefore, to first order, minimizing this short-interval action value over
$u$ is equivalent to minimizing
$\ell(t,x,u)+\nabla_xV(t,x)\trans f(t,x,u)$, the Hamiltonian's
control-dependent term
($\cref{eq:finite-horizon-hjb-equation,eq:hjb-hamiltonian-definition}$).
This term is thus a $Q$-like instantaneous cost rate, not itself a full
discrete-time $Q$-value; the HJB equation compares controls through this
rate rather than through a separate action-value function.

This connection also clarifies MPC. Dynamic programming seeks a value and
feedback decision over the whole state domain. MPC instead solves a
finite-horizon problem online from the measured state and applies its first
control before replanning. That is practical for continuous states because
it avoids precomputing an exact value at every point, though optimization
must meet the control deadline. Chapter~9's trajectory optimization and
Chapter~11's MPPI provide different ways of carrying out that online search.
In linear--quadratic control, the value has a quadratic form and Bellman
recursion reduces to the Riccati recursion of Chapter~7. The mathematical
structure is shared even though the numerical methods differ.

The chapters that follow emphasize continuous robot states and low-level
controls. They replace exact model-based value computation with value or
policy approximations learned from data, then develop actor--critic methods
for directly representing and improving continuous feedback policies. The
later chapter on hierarchical behavior also uses discrete skill choices as
high-level abstractions over continuous robot control.

\paragraph{Chapter summary.}
An MDP can describe continuous robot states and actions with uncertain
transitions. Policies are feedback laws or action densities; values are
expected returns; Bellman expectation equations average over policy actions
and transition outcomes, while optimality equations optimize over continuous
actions. These equations match the
cost-to-go recursions of optimal control after changing reward maximization
to cost minimization. Exact dynamic programming is difficult in high
dimensions, motivating MPC, trajectory optimization, sampling, and learned
function approximations.

\paragraph{Exercises.}

\begin{exercise}
A robot controller uses a continuous stochastic policy $\pi(a\mid s)$. What
quantity gives the probability that its action falls in a region
$A\subseteq\mathcal A(s)$? Why is the density at one exact action not that
probability?
\end{exercise}
\begin{proof}[Short solution]
The probability is $\int_A\pi(a\mid s)\,da$. A single exact action has zero
probability under a continuous distribution; the density is probability per
unit action-space volume.
\end{proof}

\begin{exercise}
Write the Bellman expectation equation for a continuous stochastic policy
and transition density. Which integrals are over controls and which are over
next states?
\end{exercise}
\begin{proof}[Short solution]
Use \cref{eq:mdp-bellman-expectation-v}. The outer integral averages actions
under $\pi(a\mid s)$; the inner integral averages next states under
$p(s'\mid s,a)$.
\end{proof}

\begin{exercise}
Show how the stochastic cost Bellman recursion in
\cref{eq:mdp-cost-bellman} simplifies when the dynamics are deterministic.
\end{exercise}
\begin{proof}[Short solution]
The transition distribution is concentrated at $s'=f(s,a)$, so the integral
reduces to $V_t(s)=\min_a\{c(s,a,f(s,a))+V_{t+1}(f(s,a))\}$.
\end{proof}

\begin{exercise}
Why does replacing continuous states and actions by a finite grid make
planning easier, and what does the grid approximation leave out?
\end{exercise}
\begin{proof}[Short solution]
A grid permits table-based sums and finite action comparisons. It ignores
states and controls between grid points and can therefore introduce
approximation error or exclude useful motions.
\end{proof}
\chapter{Policy Gradients and Actor--Critic}
\label{chap:policy-gradients-actor-critic}
\chapteroverview{Direct optimization of stochastic continuous-control policies, likelihood-ratio gradients, critics and advantage estimates, and the clipped policy updates used by PPO.}

Chapter~12 introduced the MDP value functions $v^\pi$ and $q^\pi$ and their
Bellman equations. One way to improve a policy is to estimate $q^\pi(s,a)$
and choose the action with greatest estimated value. For a robot, however,
the action is a real-valued vector; maximizing an approximate $q$-function
over all admissible torques or accelerations can itself be a difficult
optimization problem. Policy-gradient methods instead represent the
feedback controller directly by a parameterized policy and adjust its
parameters to increase expected return. The action distribution can be
continuous, so this approach fits the bounded controls used by physical
robots without first replacing them by a finite menu of actions.

The gradient is estimated from trajectories sampled by interacting with the
robot or a simulator. It does not require a known dynamics model or
differentiation through the simulator, but its raw Monte Carlo estimate can
be noisy. A critic estimates future return to reduce that noise and to
construct advantage estimates. This chapter develops the on-policy route
from the likelihood-ratio estimator to actor--critic learning and proximal
policy optimization (PPO). The following chapter turns from algorithmic
foundations to practical design choices for deep reinforcement learning in
robotics.
Policy optimization can improve task performance, but neither a stochastic
policy nor PPO by itself guarantees that every action or trajectory is
safe.

\section{Policy Gradients}
\label{sec:policy-gradients}

\subsection{Parameterized Policies for Continuous Actions}
\label{subsec:parameterized-continuous-policies}

Let $s\in\mathcal S$ denote the robot state and let the admissible control
set be a bounded region $\mathcal A(s)\subseteq\R^m$. A stochastic policy
$\pi_\theta(a\mid s)$ gives a density over admissible actions; its
parameters $\theta$ determine how the action distribution varies with
state. For a finite-horizon task, the policy may also depend on time-to-go
$t$. For brevity, we write the actor input simply as $s$; when time
dependence is needed, this denotes the augmented input $(s,t)$.

In a neural policy, a network commonly maps the state to the mean
and spread of a distribution. Sampling from that distribution provides
exploration during training, while the same network can be evaluated
quickly at deployment.

For box-bounded controls, a convenient construction starts with a Gaussian
latent action and maps it smoothly to the actuator limits. For each control
coordinate define its midpoint and half-range by
$m_i=(a_i^{\max}+a_i^{\min})/2$ and
$d_i=(a_i^{\max}-a_i^{\min})/2$. Then sample
\begin{equation}
  z\sim\mathcal N\!\left(\mu_\theta(s),
      \operatorname{diag}(\sigma_\theta^2(s))\right),
  \qquad a=m+d\odot\tanh z.
  \label{eq:pg-squashed-gaussian-policy}
\end{equation}
Here $\odot$ denotes componentwise multiplication. Since each component of
$\tanh z$ lies between $-1$ and $1$, the resulting action lies within the
actuator bounds. The policy parameters may include a state-dependent mean
and a state-dependent or state-independent log standard deviation.

The policy-gradient update needs the log density of the action that was
sampled. With $z_i=\operatorname{atanh}((a_i-m_i)/d_i)$, change of variables
gives
\begin{equation}
  \log\pi_\theta(a\mid s)
  =\sum_{i=1}^m\left[
    \log\mathcal N\!\left(z_i;\mu_{\theta,i}(s),
      \sigma_{\theta,i}^2(s)\right)
    -\log d_i-\log\!\left(1-\tanh^2 z_i\right)
  \right].
  \label{eq:pg-squashed-gaussian-log-density}
\end{equation}
The change-of-variables rule comes from equating the probabilities of
corresponding small intervals before and after the transformation:
$\pi_A(a\mid s)\,da=\pi_Z(z\mid s)\,dz$. Thus,
$\pi_A(a\mid s)=\pi_Z(z\mid s)|dz/da|
=\pi_Z(z\mid s)/|da/dz|$. In each coordinate,
$da_i/dz_i=d_i(1-\tanh^2z_i)$, which gives the two correction terms in
the equation. The $-\log d_i$ term accounts for scaling the action range;
the $-\log(1-\tanh^2z_i)$ term accounts for the nonlinear squashing.

The action-generation and likelihood calculation can be implemented as:
\begin{quote}
\textbf{Bounded Gaussian action sampling}
\begin{enumerate}
  \item Evaluate the policy network to obtain $\mu_\theta(s)$ and
        $\sigma_\theta(s)$.
  \item Draw $z$ from the Gaussian in
        \cref{eq:pg-squashed-gaussian-policy} and map it to
        $a=m+d\odot\tanh z$.
  \item Store the executed action and its log density from
        \cref{eq:pg-squashed-gaussian-log-density}.
  \item Send $a$ to the robot and record the resulting transition and reward.
\end{enumerate}
\end{quote}

The policy is stochastic during training even though a robot may later use
its mean action for deterministic execution. A deterministic mean is not in
general equivalent to sampling from the trained policy, especially when
the dynamics and reward are nonlinear, so deployment behavior should be
evaluated directly. A stochastic policy supplies exploration, not a safety
certificate; actuator limits and any independent safety monitor remain
necessary.

\subsection{The Log-Derivative Trick}
\label{subsec:log-derivative-trick}

Use the finite-episode return from Chapter~12 and define the objective
\begin{equation}
  J(\theta)
  =\E_{\mathcal T\sim p_\theta}\!\left[
    \sum_{t=0}^{T-1}\gamma^t R_{t+1}+\gamma^T g(S_T)\right].
  \label{eq:pg-objective}
\end{equation}
Here $\theta$ denotes the policy parameters, and $\pi_\theta(a\mid s)$ is
the action distribution at state $s$. Let
$\mathcal T=(S_0,A_0,\ldots,A_{T-1},S_T)$ denote the random state--action
trajectory, with realization $\tau=(s_0,a_0,\ldots,a_{T-1},s_T)$ as in
Chapter~12. Uppercase $S_t,A_t,R_{t+1}$ denote random variables; lowercase
$s_t,a_t,r_{t+1}$ denote their realized values. The density
$p_\theta(\tau)$ is induced by the policy, initial-state distribution, and
transition model, and $\E_{\mathcal T\sim p_\theta}$ averages over
these trajectories.
The episode horizon is $T$, the discount factor is $\gamma$, $R_{t+1}$ is
the reward after action $A_t$, and $g(S_T)$ is the terminal payoff. As in
Chapter~12, rewards are determined by state--action--next-state triples,
so the realized return $G_0(\tau)$ is determined by $\tau$. If rewards are
stochastic, augment $\tau$
with them and include their conditional law in its density. We assume the
initial-state, transition, and reward laws have no direct dependence on
$\theta$; at fixed $\tau$, the policy is the only source of parameter
dependence in the trajectory density. For this state--action trajectory,
\cref{eq:mdp-trajectory-density} gives:
\begin{equation}
  p_\theta(\tau)=\rho_0(s_0)
    \prod_{t=0}^{T-1}
      \pi_\theta(a_t\mid s_t)\,p(s_{t+1}\mid s_t,a_t).
\end{equation}

Our goal is to compute $\nabla_\theta J(\theta)$. Writing the objective
as an integral over trajectories and differentiating under the integral
sign gives, under the usual regularity conditions,
\[
  \nabla_\theta J(\theta)
  =\int G_0(\tau)\nabla_\theta p_\theta(\tau)\,d\tau
  =\int G_0(\tau)p_\theta(\tau)
      \nabla_\theta\log p_\theta(\tau)\,d\tau
  =\E_{\mathcal T\sim p_\theta}\!\left[
      G_0(\mathcal T)\nabla_\theta\log p_\theta(\mathcal T)\right].
\]
The second equality uses
$\nabla_\theta p_\theta(\tau)
=p_\theta(\tau)\nabla_\theta\log p_\theta(\tau)$. Thus, to obtain
a usable policy-gradient expression, we next expand the trajectory
log-density and identify which of its terms depend on $\theta$.

The vector $\nabla_\theta\log p_\theta(\tau)$ is called the trajectory
score (or simply the score): it measures how the log-likelihood of $\tau$
changes as the policy parameters change.
The derivative here is with respect to $\theta$, with the trajectory $\tau$
held fixed inside the integral. This is not a pathwise derivative through
the sampled states and actions. The return, reward function, and transition
dynamics therefore need not be differentiable with respect to states or
actions. Under the parameter-independence assumption above, it is enough
for the policy log-density to be differentiable with respect to $\theta$
and evaluable at sampled state--action pairs. The expectation can then be
estimated by averaging return--score products over trajectories sampled
from the current policy and environment, without differentiating through
the simulator.

Taking logarithms turns this product into a sum, and differentiating gives
\begin{align}
  \log p_\theta(\tau)
  &=\log\rho_0(s_0)
    +\sum_{t=0}^{T-1}\log\pi_\theta(a_t\mid s_t)\nonumber\\
  &\quad+\sum_{t=0}^{T-1}\log p(s_{t+1}\mid s_t,a_t)\nonumber\\
  \nabla_\theta\log p_\theta(\tau)
  &=\underbrace{\nabla_\theta\log\rho_0(s_0)}_{=0}\nonumber\\
  &\quad+\sum_{t=0}^{T-1}
      \nabla_\theta\log\pi_\theta(a_t\mid s_t)\nonumber\\
  &\quad+\underbrace{\sum_{t=0}^{T-1}
      \nabla_\theta\log p(s_{t+1}\mid s_t,a_t)}_{=0}\nonumber\\
  &=\sum_{t=0}^{T-1}\nabla_\theta
      \log\pi_\theta(a_t\mid s_t).
  \label{eq:pg-trajectory-score}
\end{align}
The initial-state and transition terms vanish because they are independent
of $\theta$. If stochastic rewards are included in $\tau$, their
conditional law contributes no score for the same reason. Thus,
the score needs no differentiation through the simulator or physical dynamics.
Substituting this trajectory score into the likelihood-ratio identity above
gives the full-return policy-gradient expression:
\begin{align}
  \nabla_\theta J(\theta) &=\E_{\mathcal T\sim p_\theta}\!\left[
      G_0\sum_{t=0}^{T-1}
        \nabla_\theta\log\pi_\theta(A_t\mid S_t)\right].
  \label{eq:pg-full-return-gradient}
\end{align}
This is the trajectory-level likelihood-ratio, or score-function, identity.
It is a policy-gradient formula in the broad sense; in this chapter, the
more specific name \emph{policy-gradient theorem} refers to the action-value
form derived in \cref{subsec:policy-gradient-theorem} by conditioning the
reward-to-go on the current state and action. The score-function form
differentiates the probability of sampled actions rather than the physical
trajectory.

To prepare for the reward-to-go derivation, examine the policy score at
time $t$. Let $\mathcal H_t$ denote the information available just before
$A_t$ is sampled: the states $S_0,\ldots,S_t$, actions
$A_0,\ldots,A_{t-1}$, and rewards $R_1,\ldots,R_t$. Define the score
$\xi_t=\nabla_\theta\log\pi_\theta(A_t\mid S_t)$.
For a continuous $A$, conditioning on $B=b$ gives the density-weighted
average of $h(A)$:
\[
\E[h(A)\mid B=b]=\int_{\mathcal A}h(a)p_{A\mid B}(a\mid b)\,da.
\]
Here $A=A_t$ and $B=\mathcal H_t$. Because the Markov policy chooses an
action using the current state $S_t$, its conditional density is
$p_{A_t\mid\mathcal H_t}(a\mid\mathcal H_t)=\pi_\theta(a\mid S_t)$.
Applying the conditional-expectation rule to the score gives:
\begin{align}
  \E[\xi_t\mid\mathcal H_t]
  &=\int_{\mathcal A(S_t)}
      \pi_\theta(a\mid S_t)\nabla_\theta\log\pi_\theta(a\mid S_t)\,da
      \nonumber\\
  &=\int_{\mathcal A(S_t)}\nabla_\theta\pi_\theta(a\mid S_t)\,da
      \nonumber\\
  &=\nabla_\theta\int_{\mathcal A(S_t)}
      \pi_\theta(a\mid S_t)\,da=\nabla_\theta 1=0.\label{eq:pg-score-zero}
\end{align}

By the finite-return definition in \cref{eq:mdp-finite-return}, split
the episode return at time $t$:
\begin{equation}
  G_0=\sum_{k=0}^{t-1}\gamma^k R_{k+1}+\gamma^t G_t.
  \tag{13.9}\label{eq:pg-return-decomposition}
\end{equation}
The first sum contains rewards observed before time $t$; $G_t$ contains
the return from time $t$ onward. Substituting
\cref{eq:pg-return-decomposition} into
\cref{eq:pg-full-return-gradient} and expanding gives:
\begin{align*}
  \nabla_\theta J(\theta)
  &=\E_{\mathcal T\sim p_\theta}\!\left[
      \sum_{t=0}^{T-1}
        \left(\sum_{k=0}^{t-1}\gamma^k R_{k+1}
          +\gamma^t G_t\right)\xi_t\right]\\
  &=\E_{\mathcal T\sim p_\theta}\!\left[
      \sum_{t=0}^{T-1}\sum_{k=0}^{t-1}
        \gamma^k R_{k+1}\xi_t
      +\sum_{t=0}^{T-1}\gamma^t G_t\xi_t\right].
\end{align*}
Each term in the first double sum has $k<t$. Fix one such pair $(k,t)$.
Because $\mathcal H_t$ contains the history up to (but not including) the
choice of $A_t$, the earlier reward $R_{k+1}$ is already known at that
point. The tower, or iterated-expectation, property says
$\E[X]=\E[\E[X\mid\mathcal H_t]]$: first average $X$ conditional on each
possible history, then average those conditional averages over histories.
Applying it to $X=R_{k+1}\xi_t$ gives the first equality below. Since
$R_{k+1}$ is fixed once $\mathcal H_t$ is fixed, it can then be pulled
outside the inner conditional expectation:
\begin{align}
  \E_{\mathcal T\sim p_\theta}[R_{k+1}\xi_t]
  &=\E_{\mathcal T\sim p_\theta}[\E[R_{k+1}\xi_t\mid\mathcal H_t]]
      \nonumber\\
  &=\E_{\mathcal T\sim p_\theta}[R_{k+1}\,
      \E[\xi_t\mid\mathcal H_t]]
      \nonumber\\
  &=0,\qquad k<t.\tag{13.8}\label{eq:pg-past-reward-zero}
\end{align}
By the zero-mean score result in \cref{eq:pg-score-zero}, the inner
conditional expectation in \cref{eq:pg-past-reward-zero} is zero for every
history. Hence every expected past-reward term vanishes, leaving only the
reward-to-go terms:
\setcounter{equation}{9}

\begin{equation}
  \nabla_\theta J(\theta)
  =\E_{\mathcal T\sim p_\theta}\!\left[
    \sum_{t=0}^{T-1}\gamma^t G_t\,
      \nabla_\theta\log\pi_\theta(A_t\mid S_t)\right].
  \label{eq:pg-reward-to-go-gradient}
\end{equation}
This identity writes the gradient as an expectation of return-weighted
policy scores, where each score is the gradient of a log policy
probability. Replacing the expectation by a sample average gives a
score-function (likelihood-ratio) estimator, commonly called REINFORCE.
Its update tends to increase the probability of sampled actions followed
by high return and decrease it for low return. Williams
introduced this likelihood-ratio approach for reinforcement learning
\cite{williams1992reinforce,sutton1999policygradient}.

With $M$ episodes, an estimator is
\begin{equation}
  \widehat g
  =\frac1M\sum_{i=1}^M\sum_{t=0}^{T_i-1}
     \gamma^t G_t^{(i)}
     \nabla_\theta\log\pi_\theta(a_t^{(i)}\mid s_t^{(i)}),
  \qquad \theta\leftarrow\theta+\alpha\widehat g.
  \label{eq:pg-reinforce-estimator}
\end{equation}
Here $\alpha>0$ is a learning rate. For a differentiable parameterized
actor, the raw on-policy Monte Carlo score estimator is unbiased. This
means $\E[\widehat g]=\nabla_\theta J(\theta)$, with the expectation taken
over batches collected by the current policy; any one finite batch can
still differ from the true gradient. 
Despite being unbiased, the estimate can have high variance:
similar initial states may lead to very different returns because of
contact events, sensor noise, or exploratory actions.

\begin{quote}
\textbf{REINFORCE: episodic policy-gradient update}
\begin{enumerate}
  \item Initialize the policy parameters $\theta$ and step size $\alpha$.
  \item Collect $M$ complete episodes with the current policy, storing each
        state, bounded action, reward, and terminal state.
  \item For each episode, set $G_{T_i}=g(S_{T_i})$ and compute backward
        $G_t=R_{t+1}+\gamma G_{t+1}$ for $t=T_i-1,\ldots,0$.
  \item Form $\widehat g$ using
        \cref{eq:pg-reinforce-estimator} and update
        $\theta\leftarrow\theta+\alpha\widehat g$.
  \item Discard the batch and repeat with newly sampled episodes.
\end{enumerate}
\end{quote}

\subsection{The Policy-Gradient Theorem}
\label{subsec:policy-gradient-theorem}

For a finite-horizon task, define
$q_t^\pi(s,a)=\E_\pi[G_t\mid S_t=s,A_t=a]$ and
$v_t^\pi(s)=\E_\pi[G_t\mid S_t=s]$. These extend Chapter~12's value
definitions to values that depend on the remaining time.
To see how the action value enters, let
$\xi_t=\nabla_\theta\log\pi_\theta(A_t\mid S_t)$, the score at time $t$.
Because $\xi_t$ is determined by $(S_t,A_t)$, conditioning on that pair
leaves it fixed. The tower property therefore gives, for each $t$,
\begin{align*}
  \E[G_t\xi_t]
  &=\E\!\left[\E[G_t\xi_t\mid S_t,A_t]\right]\\
  &=\E\!\left[\xi_t\E[G_t\mid S_t,A_t]\right]\\
  &=\E\!\left[q_t^{\pi_\theta}(S_t,A_t)\xi_t\right].
\end{align*}
The last equality is the definition of $q_t^{\pi_\theta}$. Substituting
this identity term by term into
\cref{eq:pg-reward-to-go-gradient} yields:
\begin{equation}
  \nabla_\theta J(\theta)
  =\E_{\mathcal T\sim p_\theta}\!\left[
    \sum_{t=0}^{T-1}\gamma^t
      q_t^{\pi_\theta}(S_t,A_t)
      \nabla_\theta\log\pi_\theta(A_t\mid S_t)\right].
  \label{eq:pg-theorem-finite-horizon}
\end{equation}
When $q_t^{\pi_\theta}$ is known exactly, this form can reduce variance:
for each time step, $q_t^{\pi_\theta}(S_t,A_t)\xi_t$
is the conditional expectation of $G_t\xi_t$ given $(S_t,A_t)$, so future-return
randomness is averaged out of that term. This does not by itself guarantee
lower variance for the full sum, whose terms may be correlated. In practice,
an approximate critic can introduce bias and need not preserve this variance
reduction.

This is the finite-horizon policy-gradient theorem. The expectation includes
the states visited under the current policy and the actions sampled from
that policy. The action-value weights the score, but the formula does not
differentiate $q^\pi$ with respect to the action or through the dynamics.

For a continuing discounted task, assume a stationary policy and
$0\leq\gamma<1$. Its objective is
\begin{equation}
  J_\infty(\theta)
  =\E_{S_0\sim\rho_0,\pi_\theta}
      \!\left[\sum_{t=0}^{\infty}\gamma^tR_{t+1}\right].
  \label{eq:pg-objective-infinite-horizon}
\end{equation}
The stationary action value is
$q^{\pi_\theta}(s,a)
=\E_{\pi_\theta}[\sum_{k=0}^{\infty}\gamma^kR_{t+k+1}
  \mid S_t=s,A_t=a]$; it has no time index because the policy and dynamics
are stationary. Define the unnormalized discounted state-occupancy measure
by
\begin{equation}
  d_\gamma^{\pi_\theta,\rho_0}(B)
  =\sum_{t=0}^{\infty}\gamma^t
      \Pr_{\rho_0,\pi_\theta}(S_t\in B),
  \qquad B\subseteq\mathcal S.
\end{equation}
Its total mass is $1/(1-\gamma)$. The infinite-horizon discounted
policy-gradient theorem is
\begin{equation}
  \nabla_\theta J_\infty(\theta)
  =\int_{\mathcal S}d_\gamma^{\pi_\theta,\rho_0}(ds)
      \int_{\mathcal A(s)}\pi_\theta(a\mid s)
        q^{\pi_\theta}(s,a)\nabla_\theta\log\pi_\theta(a\mid s)\,da.
  \label{eq:pg-theorem-infinite-horizon}
\end{equation}
Equivalently, this is
$\E_{\rho_0,\pi_\theta}[\sum_{t=0}^{\infty}\gamma^t
q^{\pi_\theta}(S_t,A_t)\nabla_\theta\log\pi_\theta(A_t\mid S_t)]$.
This is the discounted infinite-horizon form; average-reward formulations
use a different objective \cite{sutton1999policygradient,sutton2018reinforcement}.

This result clarifies the distinction from value-based control. Value-based
methods estimate action values and choose an action by maximizing over
$\mathcal A(s)$. A policy-gradient method changes $\theta$ so that actions
with better estimated long-term consequences become more probable. The gradient is still only a local direction in
parameter space; it does not guarantee a globally optimal policy.

For the squashed-Gaussian policy, evaluate the score at a sampled
action while holding that action fixed, and recover its latent value
$z_i=\operatorname{atanh}((a_i-m_i)/d_i)$. Because the actuator
bounds are fixed, the log-Jacobian correction in
\cref{eq:pg-squashed-gaussian-log-density} has no direct
$\theta$-dependence, so the score is the latent Gaussian score. Let
$\ell_{\theta,i}(s)=\log\sigma_{\theta,i}(s)$. For a diagonal Gaussian,
\begin{equation}
  \nabla_\theta\log\pi_\theta(a\mid s)
  =\sum_{i=1}^m\left[
    \frac{z_i-\mu_{\theta,i}(s)}{\sigma_{\theta,i}^2(s)}
      \nabla_\theta\mu_{\theta,i}(s)
    +\left(
      \frac{(z_i-\mu_{\theta,i}(s))^2}{\sigma_{\theta,i}^2(s)}-1
     \right)\nabla_\theta\ell_{\theta,i}(s)
  \right].
  \label{eq:pg-gaussian-score}
\end{equation}
When covariance is fixed, the log-standard-deviation term vanishes,
recovering the mean-only score; the policy-network derivatives can be
evaluated by backpropagation.

\section{Variance Reduction and Actor--Critic}
\label{sec:variance-reduction-actor-critic}

\subsection{Baselines and Advantages}
\label{subsec:pg-baselines-advantages}

Recall the policy score $\xi_t=\nabla_\theta\log\pi_\theta(A_t\mid S_t)$.
At a fixed state $s$, its conditional mean is zero because the policy density
is normalized:
\begin{align}
  \E[\xi_t\mid S_t=s]
  &=\int_{\mathcal A(s)}\pi_\theta(a\mid s)
      \nabla_\theta\log\pi_\theta(a\mid s)\,da\nonumber\\
  &=\int_{\mathcal A(s)}\nabla_\theta\pi_\theta(a\mid s)\,da
      \nonumber\\
  &=\nabla_\theta\int_{\mathcal A(s)}\pi_\theta(a\mid s)\,da
      =\nabla_\theta 1=0.
  \label{eq:pg-zero-mean-score}
\end{align}
For any state-only baseline $b(s)$, condition on the current state to get
\begin{equation}
  \E[b(S_t)\xi_t]
  =\E\!\left[b(S_t)\E[\xi_t\mid S_t]\right]=0.
  \label{eq:pg-baseline-identity}
\end{equation}
Thus subtracting $b(S_t)$ from the action value leaves the exact expected
policy gradient unchanged. A useful baseline can reduce variance by
removing return variation predictable from the state, though a poor one
need not do so.

The state value is a natural baseline. For a finite-horizon task, define the
advantage at time $t$ by
\begin{equation}
  A_t^{\pi_\theta}(s,a)=q_t^{\pi_\theta}(s,a)-v_t^{\pi_\theta}(s).
  \label{eq:pg-advantage-definition}
\end{equation}
The advantage measures how much better or worse an action is than the
policy's expected return at that state. Substituting the state value for
the baseline in the policy-gradient theorem gives
\begin{equation}
  \nabla_\theta J(\theta)
  =\E_{\mathcal T\sim p_\theta}\!\left[\sum_{t=0}^{T-1}\gamma^t
      A_t^{\pi_\theta}(S_t,A_t)
      \nabla_\theta\log\pi_\theta(A_t\mid S_t)\right].
  \label{eq:pg-advantage-gradient}
\end{equation}

An actor--critic method has two learned components. The \emph{actor}
$\pi_\theta$ is updated to improve the policy; the \emph{critic} $v_w(s)$,
with parameters $w$, approximates the state value under that policy. In a
finite-horizon task, time-to-go may be included in the critic input, as in
the actor input convention of \cref{subsec:parameterized-continuous-policies}.

The critic $v_w$ is typically fit by squared-error regression, matching its
predictions at sampled states to targets computed either from sampled
returns or by bootstrapping from value estimates. We call these critic
targets. The on-policy actor--critic batch update below summarizes
the training loop.

\begin{quote}
\textbf{On-policy actor--critic (batch update)}

\begin{enumerate}
  \item Initialize actor parameters $\theta$, critic parameters $w$, and
        their step sizes.
  \item Copy the critic parameters, $\bar w\leftarrow w$, and hold this
        snapshot fixed throughout the rollout update.
  \item Collect $B$ fresh episodes using the current policy $\pi_\theta$.
        Store the realized states, actions, rewards, next states, and
        termination indicators.
  \item Construct the rollout's \textit{advantage estimates}, denoted by
        $\widehat A_t^{(i)}$, and \textit{critic targets}, denoted by
        $\widehat g_t^{(i)}$, using the selected estimator and the frozen
        critic snapshot when required, as detailed in the following
        subsections.
  \item Use the advantage estimates in the sampled policy-gradient update:
        \begin{align*}
          \widehat{\nabla_\theta J}
          &=\frac1B\sum_{i=1}^B\sum_{t=0}^{T_i-1}
              \gamma^t\widehat A_t^{(i)}
              \nabla_\theta\log\pi_\theta
              (a_t^{(i)}\mid s_t^{(i)}),\\
          \theta&\leftarrow\theta+
              \alpha_\pi\widehat{\nabla_\theta J}.
        \end{align*}
  \item Fit the critic $v_w$ by squared-error regression to the fixed
        critic targets $\widehat g_t^{(i)}$.
  \item Discard the rollout and collect fresh on-policy data for the next
        update.
\end{enumerate}
\end{quote}

The target and advantage constructions differ by estimator. With Monte
Carlo, the sampled return-to-go is the critic target; subtracting the value
baseline from it gives the actor's advantage estimate. One-step
temporal-difference (TD) learning bootstraps the critic target from the
next-state value prediction and uses the TD residual as the actor's
advantage estimate. Generalized advantage estimation (GAE) accumulates TD
residuals across a rollout segment to form the actor advantage, then adds
that advantage to the frozen critic prediction at the current state to
form the critic target. The following subsections derive these
constructions.

\subsection{Monte Carlo Critic Targets and Advantages}
\label{subsec:pg-monte-carlo-targets}

Because $\E[G_t\mid S_t=s]=v_t^{\pi_\theta}(s)$, the sampled return-to-go
$g_t^{(i)}$ is a direct critic target. Subtracting a state-value baseline
gives an advantage sample; with an exact critic, its conditional mean given
the state and action is $A_t^{\pi_\theta}(s,a)$. Using the rollout-frozen
critic $v_{\bar w}$, define
\begin{equation}
  \begin{aligned}
    \widehat g_t^{(i),\mathrm{MC}}&=g_t^{(i)},\\
    \widehat A_t^{(i),\mathrm{MC}}&=g_t^{(i)}-v_{\bar w}(s_t^{(i)}).
  \end{aligned}
  \label{eq:pg-monte-carlo-target-advantage}
\end{equation}
The target uses the realized return $g_t^{(i)}$ directly. This
full return can be noisy and is available only after the episode ends,
motivating one-step TD's next-state value prediction.

\subsection{One-Step Temporal-Difference Targets and Advantages}
\label{subsec:pg-one-step-td}

One-step temporal-difference (TD) learning reduces Monte Carlo noise by
using a frozen next-state value prediction in place of the unsampled return.

For simplicity, include any terminal payoff in the reward on the transition
that ends the episode. Under the return convention in
\cref{eq:pg-objective}, add $\gamma g(S_T)$ to the final transition reward,
since that reward is weighted by $\gamma^{T-1}$. Thus no separate terminal
payoff term is needed below.

The task specification defines which events end an episode; examples include
task success and a safety violation. Let $M_t$ be the bootstrap mask: set
$M_t=0$ on a transition that reaches one of these task-defined terminal
conditions, and set $M_t=1$ when the task continues. In a continuing infinite-horizon task
with no terminal conditions, $M_t=1$ at every step. A cutoff used only to
limit data collection has $M_t=1$; a time limit that the task specification
defines as episode completion has $M_t=0$. 

With frozen rollout critic $v_{\bar w}$, define the one-step TD residual:
\begin{equation}
  \delta_t
  =R_{t+1}+\gamma M_t v_{\bar w}(S_{t+1})
      -v_{\bar w}(S_t).
  \label{eq:pg-td-residual}
\end{equation}
Use the following one-step target for critic regression:
\begin{equation}
  \begin{aligned}
    \widehat g_t^{\mathrm{TD}}
    &=v_{\bar w}(S_t)+\delta_t\\
    &=R_{t+1}+\gamma M_t v_{\bar w}(S_{t+1}).
  \end{aligned}
  \label{eq:pg-td-value-target}
\end{equation}

With an exact frozen critic,
$v_{\bar w}(S_t)=v_t^{\pi_\theta}(S_t)$. The Bellman equation then gives
the conditional mean of the TD residual:
\begin{align}
  \E[\delta_t\mid S_t=s,A_t=a]
  &=\E\!\left[R_{t+1}+\gamma M_t
      v_{t+1}^{\pi_\theta}(S_{t+1})\mid S_t=s,A_t=a\right]
      -v_t^{\pi_\theta}(s)\nonumber\\
  &=q_t^{\pi_\theta}(s,a)-v_t^{\pi_\theta}(s)
   =A_t^{\pi_\theta}(s,a).
  \label{eq:pg-td-residual-advantage}
\end{align}
We use the TD residual as the one-step advantage estimate,
$\widehat A_t^{\mathrm{TD}}:=\delta_t$. With an exact critic, its
conditional mean is the true advantage; an individual transition can still
be noisy.

With an approximate critic, error can come from the baseline
$v_{\bar w}(S_t)$ and, when $M_t=1$, the bootstrap $v_{\bar w}(S_{t+1})$.
Fit $v_w$ to the fixed targets while keeping $\bar w$ unchanged. During
the actor update, hold the computed advantages fixed and do not
backpropagate through the critic values used to form them.

\subsection{Generalized Advantage Estimation}
\label{subsec:generalized-advantage-estimation}

Monte Carlo advantages retain the noise of the full sampled return, whereas
one-step TD uses only one bootstrap prediction. Generalized advantage
estimation (GAE) sums TD residuals over several steps. The parameter
$\lambda$ controls how much it relies on sampled rewards versus the critic.

Using the frozen parameters $\bar w$, compute the TD residual for each
stored transition:
\begin{equation}
  \delta_t^{(i)}
  =r_{t+1}^{(i)}+\gamma m_t^{(i)}
      v_{\bar w}(s_{t+1}^{(i)})
      -v_{\bar w}(s_t^{(i)}).
  \label{eq:pg-td-residual-sample}
\end{equation}
Generalized advantage estimation (GAE) accumulates the sampled residuals,
stopping at the end of each segment:
\begin{align}
  \widehat A_t^{(i)}
  &=\sum_{\ell=0}^{K_i-t-1}(\gamma\lambda)^\ell
      \delta_{t+\ell}^{(i)}\nonumber\\
  &=\delta_t^{(i)}+\gamma\lambda\widehat A_{t+1}^{(i)},
      \quad \widehat A_{K_i}^{(i)}=0,\quad 0\leq\lambda\leq1.
  \label{eq:pg-gae}
\end{align}
At $\lambda=0$, GAE is the one-step residual. At $\lambda=1$, residuals
telescope to the return-to-go minus $v_{\bar w}(s_t^{(i)})$ for a complete
episode. At a nonterminal segment cutoff, the estimate includes a bootstrap
value:
\begin{equation}
  \begin{aligned}
    \widehat A_t^{(i)}
    &=\sum_{\ell=0}^{K_i-t-1}\gamma^\ell r_{t+\ell+1}^{(i)}\\
    &\quad+\gamma^{K_i-t}v_{\bar w}(s_{K_i}^{(i)})
      -v_{\bar w}(s_t^{(i)}).
  \end{aligned}
\end{equation}
Intermediate $\lambda$ values trade sensitivity to critic error against
variance from sampled rewards: smaller values rely more on the critic,
while larger values use longer return segments. With an approximate critic,
GAE need not be unbiased \cite{schulman2016gae}.

The corresponding fixed value target is
\begin{equation}
  \widehat g_t^{(i),\lambda}
  =v_{\bar w}(s_t^{(i)})+\widehat A_t^{(i)},
  \qquad
  \mathcal L_V(w)=\frac{1}{2N}\sum_{(i,t)\in\mathcal D}
      \left(v_w(s_t^{(i)})-\widehat g_t^{(i),\lambda}\right)^2.
  \label{eq:pg-gae-value-target}
\end{equation}
Here $N=|\mathcal D|$ is the number of transitions. Keep this target fixed
while fitting the critic; do not backpropagate through the residuals or
$v_{\bar w}$.

\begin{quote}
\textbf{GAE: backward advantage pass}
\begin{enumerate}
  \item Freeze the rollout critic at $\bar w$. For each segment $i$, set
        $\widehat A_{K_i}^{(i)}=0$.
  \item For $t=K_i-1,\ldots,0$, compute the residual in
        \cref{eq:pg-td-residual-sample} using the stored reward, next state,
        and bootstrap mask; then set
        $\widehat A_t^{(i)}=\delta_t^{(i)}+
        \gamma\lambda\widehat A_{t+1}^{(i)}$.
  \item Form fixed targets
        $\widehat g_t^{(i),\lambda}=v_{\bar w}(s_t^{(i)})+
        \widehat A_t^{(i)}$ for the critic regression.
\end{enumerate}
\end{quote}

\section{Stable Policy Updates}
\label{sec:stable-policy-updates}

Policy-gradient estimates are local to states visited by the current policy.
If one update changes the policy too much, the new policy may visit states
that were rare in the rollout, where the gradient estimate is unreliable.
Trust-region methods address this by limiting how far the updated action
distribution can move from the behavior policy that generated the data.

\subsection{Trust Region Policy Optimization}
\label{subsec:trust-region-policy-optimization}

An ordinary policy-gradient update estimates
$g\approx\nabla_\theta J(\theta_{\mathrm{old}})$ from an on-policy rollout
and takes a step such as $\theta\leftarrow\theta_{\mathrm{old}}+\alpha g$.
This gives a local ascent direction, but does not by itself score the return
change from a finite step. Evaluating each candidate with fresh rollouts is
costly, while reusing the old rollout requires accounting for the fact that
it was generated by $\pi_{\theta_{\mathrm{old}}}$. TRPO estimates candidate
improvement from that batch and limits the policy change with a KL constraint.

For each stored state--action pair, define the candidate-to-old probability
ratio
\begin{equation}
  \varrho_t^{(i)}(\theta)
  =\frac{\pi_\theta(a_t^{(i)}\mid s_t^{(i)})}
         {\pi_{\theta_{\mathrm{old}}}(a_t^{(i)}\mid s_t^{(i)})}
  =\exp\!\left[
      \log\pi_\theta(a_t^{(i)}\mid s_t^{(i)})
      -\log\pi_{\theta_{\mathrm{old}}}(a_t^{(i)}\mid s_t^{(i)})\right].
  \label{eq:pg-update-ratio}
\end{equation}
It is one at $\theta_{\mathrm{old}}$ and reweights the sampled action by its
likelihood under the candidate policy. TRPO maximizes this advantage-based
estimate subject to an average KL limit on the rollout states:
\begin{equation}
  \begin{aligned}
    \max_\theta\quad
      &\frac{1}{B}\sum_{i=1}^{B}\sum_{t=0}^{T_i-1}
        \gamma^t\varrho_t^{(i)}(\theta)\widehat A_t^{(i)}\\
    \text{subject to}\quad
      &\frac{1}{N}\sum_{(i,t)\in\mathcal D}
        D_{\mathrm{KL}}\!\left(
          \pi_{\theta_{\mathrm{old}}}(\cdot\mid s_t^{(i)})\,
          \middle\|\,\pi_\theta(\cdot\mid s_t^{(i)})
        \right)\leq\delta,
  \end{aligned}
  \label{eq:trpo-constrained-update}
\end{equation}
Here $N=|\mathcal D|$, and $\delta$ limits the average KL divergence on the
rollout states. The reason this is a meaningful surrogate follows from the
finite-horizon performance-difference identity. Let $d_t^\theta$ denote
the distribution of $S_t$ under $\pi_\theta$. For the old policy,
$v_T^{\pi_{\theta_{\mathrm{old}}}}(s)=g(s)$, and the finite-horizon
Bellman definition gives
\[
  A_t^{\pi_{\theta_{\mathrm{old}}}}(s,a)
  =\E\!\left[
      R_{t+1}+\gamma v_{t+1}^{\pi_{\theta_{\mathrm{old}}}}(S_{t+1})
      \mid S_t=s,A_t=a
    \right]
    -v_t^{\pi_{\theta_{\mathrm{old}}}}(s).
\]
Substitute this into the advantage sum along a trajectory from $\pi_\theta$
and sum over time. The value terms telescope; the initial-state distribution
is shared by both policies:
\begin{align*}
  \E_{\mathcal T\sim p_\theta}\!\left[
    \sum_{t=0}^{T-1}\gamma^t
      A_t^{\pi_{\theta_{\mathrm{old}}}}(S_t,A_t)\right]
  &=\E_{\mathcal T\sim p_\theta}\!\left[
      \sum_{t=0}^{T-1}\gamma^t R_{t+1}
      +\gamma^Tg(S_T)
      -v_0^{\pi_{\theta_{\mathrm{old}}}}(S_0)
    \right]\\
  &=J(\theta)-J(\theta_{\mathrm{old}})\\
  &=\sum_{t=0}^{T-1}\gamma^t
    \E_{\substack{S_t\sim d_t^\theta\\
                   A_t\sim\pi_\theta(\cdot\mid S_t)}}
      \!\left[A_t^{\pi_{\theta_{\mathrm{old}}}}(S_t,A_t)\right].
\end{align*}
The rollout supplies states from $d_t^{\theta_{\mathrm{old}}}$, not
$d_t^\theta$. Holding those states fixed and using the probability ratio
to change the action distribution gives the population surrogate
\begin{equation}
  \mathcal L_{\theta_{\mathrm{old}}}(\theta)
  =\E_{\mathcal T\sim p_{\theta_{\mathrm{old}}}}\!\left[
    \sum_{t=0}^{T-1}\gamma^t
      \frac{\pi_\theta(A_t\mid S_t)}
           {\pi_{\theta_{\mathrm{old}}}(A_t\mid S_t)}
      A_t^{\pi_{\theta_{\mathrm{old}}}}(S_t,A_t)
  \right].
  \label{eq:trpo-population-surrogate}
\end{equation}
Thus, the surrogate retains the candidate action distribution but
substitutes the old policy's state distribution. It is exact at
$\theta_{\mathrm{old}}$; away from that policy, the approximation error
comes from the state-visitation shift. The theoretical TRPO bound controls
this shift using exact advantages and a maximum-over-states policy-divergence
constraint. In practice, \cref{eq:trpo-constrained-update} uses estimated
advantages and an empirical average KL, so surrogate improvement does not
guarantee that the true return increased \cite{schulman2015trpo}.

The surrogate also has the correct gradient at the reference policy.
Because $\nabla_\theta\varrho_t(\theta)
=\varrho_t(\theta)\nabla_\theta\log\pi_\theta(A_t\mid S_t)$ and
$\varrho_t(\theta_{\mathrm{old}})=1$, differentiating with the old-policy
data and advantages fixed gives:
\begin{align}
  \left.\nabla_\theta\mathcal L_{\theta_{\mathrm{old}}}(\theta)
    \right|_{\theta=\theta_{\mathrm{old}}}
  &=\E_{\mathcal T\sim p_{\theta_{\mathrm{old}}}}\!\left[
    \sum_{t=0}^{T-1}\gamma^t
      A_t^{\pi_{\theta_{\mathrm{old}}}}(S_t,A_t)
      \nabla_\theta\log\pi_{\theta_{\mathrm{old}}}(A_t\mid S_t)
  \right]\nonumber\\
  &=\left.\nabla_\theta J(\theta)\right|_{\theta=\theta_{\mathrm{old}}}.
  \label{eq:trpo-surrogate-policy-gradient}
\end{align}
This is the policy-gradient theorem in \cref{eq:pg-advantage-gradient} at
$\theta_{\mathrm{old}}$. With estimated advantages, the minibatch gradient
only estimates this identity. The match is first-order, so TRPO builds a
local model around $\theta_{\mathrm{old}}$ rather than treating the
surrogate as the true return at distant parameters. Let
$\Delta\theta=\theta-\theta_{\mathrm{old}}$ and
$g=\nabla_\theta\mathcal L_{\theta_{\mathrm{old}}}(\theta)
|_{\theta=\theta_{\mathrm{old}}}$. For a small step,
\begin{align*}
  \mathcal L_{\theta_{\mathrm{old}}}(\theta_{\mathrm{old}}+\Delta\theta)
  &\approx \mathcal L_{\theta_{\mathrm{old}}}(\theta_{\mathrm{old}})
      +g^\top\Delta\theta,\\
  \overline D_{\mathrm{KL}}(\theta_{\mathrm{old}},
      \theta_{\mathrm{old}}+\Delta\theta)
  &\approx \tfrac12\Delta\theta^\top F\Delta\theta,
\end{align*}
where $\overline D_{\mathrm{KL}}$ is the rollout-state average in
\cref{eq:trpo-constrained-update}, and $F$ is its KL Hessian at the old
policy, equal to the empirical Fisher information matrix (the average over
rollout states and old-policy actions of policy-score outer products).
Dropping the constant surrogate value leaves the local constrained problem
\[
  \max_{\Delta\theta}\ g^\top\Delta\theta
  \qquad\text{subject to}\qquad
  \tfrac12\Delta\theta^\top F\Delta\theta\leq\delta.
\]
Its solution points along the scaled natural-gradient direction
$F^{-1}g$. TRPO uses conjugate gradients to approximate this direction
without forming or inverting the full matrix. Because both Taylor models
are local, a backtracking line search reevaluates the empirical surrogate
and KL on the rollout data, without the Taylor approximations. It accepts a
step only if the surrogate exceeds its value at $\theta_{\mathrm{old}}$ and
the measured KL stays within the limit \cite{schulman2015trpo}.
Conjugate gradients avoid the dense solve, but do not make TRPO cheap: each
iteration requires a Fisher-vector product, and line search adds policy
evaluations. The cost grows with policy size, iteration count, and batch size.

\begin{quote}
\textbf{TRPO update}
\begin{enumerate}
  \item Collect an on-policy rollout; compute and fix its advantages and
        critic targets.
  \item Propose a step using the linearized surrogate and quadratic KL model,
        then backtrack until the surrogate exceeds its old-policy value and
        the measured KL is within the limit; otherwise keep the old policy.
  \item Fit the critic, discard the batch, and collect fresh data.
\end{enumerate}
\end{quote}

\subsection{Proximal Policy Optimization}
\label{subsec:proximal-policy-optimization}
TRPO's constrained solve is more involved than first-order optimization.
PPO keeps the old-policy rollout and probability ratio but replaces the
constrained solve with a clipped surrogate optimized by minibatch methods.
Clipping discourages large updates; it is not a hard trust region or a
safety guarantee \cite{schulman2017ppo}.

At a fixed state, the ratio reweights old-policy samples to estimate
expectations under the new policy. PPO uses the following clipped
surrogate:
\begin{equation}
  L^{\mathrm{CLIP}}(\theta)
  =\frac{1}{B}\sum_{i=1}^{B}\sum_{t=0}^{T_i-1}\gamma^t
    \min\!\left(
      \varrho_t^{(i)}(\theta)\widehat A_t^{(i)},
      \operatorname{clip}\!\left(\varrho_t^{(i)}(\theta),
        1-\epsilon,1+\epsilon\right)\widehat A_t^{(i)}
    \right),
  \qquad \epsilon>0.
  \label{eq:ppo-clipped-objective}
\end{equation}
At $\theta_{\mathrm{old}}$, the ratio is one, so the unclipped surrogate
starts in the policy-gradient direction. For positive advantages, clipping
stops rewarding further probability increases beyond the chosen range; for
negative advantages, it stops rewarding further decreases. This limits the
surrogate's incentive, not the probability ratio or KL divergence itself.

An entropy bonus can encourage exploration. Maximize the actor objective
and minimize the critic loss:
\begin{align}
  L_{\mathrm{actor}}(\theta)
  &=L^{\mathrm{CLIP}}(\theta)
    +\frac{c_H}{B}\sum_{i=1}^{B}\sum_{t=0}^{T_i-1}\gamma^t
       \mathcal H\!\left(\pi_\theta(\cdot\mid s_t^{(i)})\right),
       \label{eq:ppo-actor-objective}\\
  L_V(w)
  &=\frac{1}{2N}\sum_{(i,t)\in\mathcal D}
       \left(v_w(s_t^{(i)})-\widehat g_t^{(i),\lambda}\right)^2.
       \label{eq:ppo-value-objective}
\end{align}
Here $c_H$ sets the entropy bonus, and $\widehat g_t^{(i),\lambda}$ are the
fixed critic targets from \cref{eq:pg-gae-value-target}.

\begin{quote}
\textbf{PPO: clipped on-policy actor--critic}
\begin{enumerate}
  \item Set $\theta_{\mathrm{old}}\leftarrow\theta$, collect a fresh rollout,
        and store transitions and old action log probabilities.
  \item With the critic frozen, compute bootstrap masks, TD residuals, GAE
        advantages, and fixed value targets.
  \item Optionally normalize advantages, then optimize actor and critic on
        minibatches for several epochs. Stop early if the KL grows too large.
  \item Discard the batch and use the updated actor for the next rollout.
\end{enumerate}
\end{quote}

PPO is on-policy: it reuses each fresh batch for only a few epochs, then
collects new data from the updated policy.

\subsection{Implementation Diagnostics}
\label{subsec:ppo-implementation-diagnostics}

Policy-gradient failures can look like ordinary training instability, so
inspect the controller's behavior as well as its loss curves:
\begin{itemize}
  \item Track return, success, violations, and termination causes on both
        training and held-out conditions; rising reward can hide loopholes.
  \item Inspect action statistics, critic fit, policy KL, and clip fraction.
  \item Verify bootstrap masks and confirm stored actions match their log
        probabilities.
\end{itemize}
Normalize inputs, use several seeds, and test held-out conditions. Domain
randomization covers only variations included in training. PPO clipping is
not a safety filter; validate policies in stages before hardware use.

\subsection{Off-Policy Learning and Experience Reuse}
\label{subsec:off-policy-experience-reuse}

PPO is a good fit when fresh rollouts are cheap, as in parallel simulation
\cite{makoviychuk2021isaacgym}. When robot interaction is costly, off-policy
methods such as TD3 and SAC reuse transitions from replay buffers
\cite{fujimoto2018td3,haarnoja2018sac}. Replay can improve sample efficiency,
but stale data and repeated critic updates can introduce error. Choose
based on whether fresh interaction or wall-clock throughput is the main
constraint; neither approach is always better.

\paragraph{Chapter summary.}
Policy gradients optimize a stochastic policy using sampled rewards and
log-probability gradients. Critics and GAE reduce variance but depend on
value estimates. TRPO constrains average policy change; PPO offers a simpler
clipped surrogate, not a hard trust region or safety guarantee. Correct
action densities, terminal masks, and evaluation remain essential. Replay
can reuse costly experience, but stale data and repeated critic updates can
introduce error.

\paragraph{Exercises.}

\begin{exercise}
Show that subtracting a baseline $b(s)$ independent of the sampled action
does not change the conditional expected policy-gradient contribution.
Why can it nevertheless change the variance of the estimator?
\end{exercise}
\begin{proof}[Short solution]
The added term has expectation
$b(s)\int\pi_\theta(a\mid s)\nabla_\theta\log\pi_\theta(a\mid s)\,da
=b(s)\nabla_\theta\int\pi_\theta(a\mid s)\,da=0$. Its sample values
can still be more or less variable than the uncentered return, so the
baseline changes variance without changing the mean gradient.
\end{proof}

\begin{exercise}
Why does the likelihood-ratio policy gradient not require differentiating
the robot dynamics? State one condition it still places on the policy.
\end{exercise}
\begin{proof}[Short solution]
The derivative of the trajectory log density contains the policy log
probabilities; transition probabilities are independent of the policy
parameters. The policy density must be differentiable with respect to its
parameters, and its log probability must be evaluable for sampled actions.
\end{proof}

\begin{exercise}
For GAE, explain the roles of $\lambda=0$ and $\lambda=1$. Why can the
latter be noisier?
\end{exercise}
\begin{proof}[Short solution]
At $\lambda=0$, only the one-step TD residual is used. At $\lambda=1$,
the residuals telescope to the Monte Carlo return minus the value baseline.
The latter uses more sampled future rewards and so can carry more trajectory
noise, though it relies less on critic bootstrapping.
\end{proof}

\begin{exercise}
What does the PPO probability ratio compare? Explain why clipping it does
not guarantee bounded actions or a hard bound on policy change.
\end{exercise}
\begin{proof}[Short solution]
It compares the new policy's density for a stored action with the old
policy's density that generated it. Clipping limits incentives in this
surrogate for sampled actions, not the policy's support at all states or
actions and not the KL divergence itself. Action bounds must be enforced
by the policy parameterization or controller.
\end{proof}

\begin{exercise}
For $a_i=m_i+d_i\tanh z_i$, identify the log-Jacobian correction needed
when evaluating the action density from the latent Gaussian.
\end{exercise}
\begin{proof}[Short solution]
The derivative is $da_i/dz_i=d_i(1-\tanh^2 z_i)$. The action log density
equals the latent Gaussian log density minus
$\sum_i\log[d_i(1-\tanh^2 z_i)]$.
\end{proof}
\chapter{Practical Tips for Deep RL for Robotics}
\label{chap:practical-tips-deep-rl-robotics}
\chapteroverview{Practical choices in policy inputs, training information, curricula, rewards, simulator fidelity, goal-conditioned control, and sim-to-real robustness.}

Reinforcement learning (RL) is a general framework for learning behavior
through interaction. Robotics tasks commonly involve continuous state and
action spaces, and modern simulators can generate fresh rollouts in large,
GPU-parallel batches. Under these conditions, an on-policy method such as
PPO is often a practical default. Yet the generality of the RL formulation
does not make training plug-and-play: arbitrary choices of observations,
actions, and rewards do not generally produce a useful policy.

Successful robot learning depends on choices tailored to the task and its
deployment constraints. This chapter surveys policy inputs and privileged
information, system identification and actuator modeling, curricula and
reward design, domain randomization, latency and filtering, hybrid
model-based control, and goal-conditioned policies. These choices are often
crucial to obtaining useful, deployable behavior. The guidance is a set of
design checks, not a substitute for evaluating the complete controller on
held-out tasks and, ultimately, through staged hardware validation.

\section{Policy Inputs and Architectures}
\label{sec:robot-policy-inputs}

A state-based policy takes as input a state estimate, such as joint positions
and velocities, base pose and velocity, and task-relevant object information.
For manipulation, this may include an estimated six-degree-of-freedom
(6-DoF) object pose, along with geometric information from a CAD model.
This representation is often compact and aligned with robot kinematics and
dynamics; it can simplify learning and debugging and support established
controllers or safety checks. Its usefulness depends on reliable estimation.
Errors can mislead the policy, and obtaining accurate estimates from partial
observations, such as images with occlusions, can be notoriously difficult.

An end-to-end observation-to-action policy maps sensor observations
directly to controls, for example from images and proprioceptive readings to
joint commands. It bypasses a separately specified state-estimation module
by learning perception and control together from data. This can be useful
when a hand-designed estimator is difficult to build or discards important
task information, but it raises the data and validation burden: the policy
must learn both which features matter and how to act on them. End-to-end
does not mean that hidden physical variables cease to matter; they must be
inferable from the observations, supplied through memory, or handled
robustly despite being unobserved.

When a single observation is insufficient, stack a short observation--action
history or use a recurrent policy with internal memory. Such a policy can
infer quantities such as velocity, contact mode, or a slowly changing payload
from observation sequences, performing some state estimation implicitly
alongside control. Memory alone does not guarantee this: the history must
contain informative measurements, and training must teach the policy to use
them. Learned memory is useful when it solves a real observability problem,
not as a default replacement for a good estimator.

\section{Privileged Information and Teacher--Student Frameworks}
\label{sec:privileged-teacher-student}

A simulator often exposes variables that a physical robot cannot measure
directly, such as exact link velocities, contact forces, object pose,
terrain geometry, or the parameters used to generate the scene. These are
\emph{privileged information}. During training, a critic may use the full
simulator state to estimate returns even when the actor receives only the
observations that will be available at deployment. This asymmetric
actor--critic design can make value estimation easier 
\cite{pinto2017asymmetric}.

Additionally, another useful stretagy is to train a teacher policy with privileged inputs and
then train a student policy to reproduce its useful behavior from deployable
observations. The student can be supervised on teacher actions or action
distributions, and can also be fine-tuned using the task reward. In either
case, the student---not the teacher---is the policy that must pass the final
evaluation. Teacher--student transfer has been used to turn privileged
simulation training into policies that operate from robot observations
\cite{kumar2021rapid}.

The information boundary must be explicit. Do not accidentally feed a
privileged contact label, exact velocity, or simulator terrain map to the
student if it will be absent on the robot. If the teacher relies on a fact
that cannot be inferred from observation history, exact imitation may be
impossible; the student must instead learn a robust response or receive
additional sensors. 

\section{System Identification and Actuator Modeling}
\label{sec:robot-system-identification}

Model task-relevant dynamics rather than attempting to reproduce the entire
real world. Identify parameters that materially affect performance, such as
link masses and inertias, center of mass, friction, damping, compliance,
contact behavior, sensor calibration, and command delay. Use measurements
and safe excitation where possible, then compare simulated and measured
responses over task-relevant motions and speeds.

Model actuators at the same command interface used by the policy. Delivered
torque may differ from the command because of motor and drive bandwidth,
low-level position or torque control loops, current and voltage limits,
saturation, rate limits, backlash, and communication delay. Available
electric-motor torque typically decreases with rotational speed; represent
the task-relevant torque--speed envelope, for example:
\begin{equation}
  |\tau_i|\leq \tau_{i,\max}(|\dot q_i|),
  \label{eq:robot-torque-speed-envelope}
\end{equation}
Include thermal or voltage constraints when relevant. Assuming peak stall
torque at every speed can teach the policy to request infeasible motions.

System identification and actuator modeling guide learning; they need not
yield a perfect digital twin. Model task-critical dynamics accurately, then
represent residual uncertainty deliberately. Accurate actuator limits and
delay are especially important for precise tracking and high-speed maneuvers,
where small model errors can substantially change achievable trajectories.

\section{Curriculum Learning}
\label{sec:robot-curriculum-learning}

Some tasks are too difficult to learn from a broad, uniformly sampled set
of episodes at the outset. A curriculum gradually expands the training
distribution as the policy becomes competent. For a legged robot, one might
begin on level ground with modest commanded speeds, then add uneven terrain,
larger speed changes, pushes, and wider variation in starting pose. For
manipulation, one might first use reachable, well-oriented objects, then
increase pose variation, clutter, object diversity, and contact difficulty.

Curriculum dimensions include task difficulty, initial-state and goal
diversity, and environmental complexity. A schedule can be hand designed
or adjusted using success rates, but it should not advance so quickly that
the policy stops receiving useful learning signals. Keep some coverage of
earlier cases as harder cases are introduced, and verify that the final
training distribution still covers the intended deployment conditions.
Otherwise, a policy can become good at the curriculum schedule while
forgetting simpler behaviors or failing at the actual target distribution.

\section{Reward Engineering and Reward Curricula}
\label{sec:robot-reward-engineering}

Start from the task objective and express it in measurable terms. For
reaching, reward reduced goal error and low terminal speed; for locomotion,
balance commanded-velocity tracking against energy use, slippage, and unsafe
body motion. When successful rollouts are rare, dense progress rewards can
provide useful feedback during early exploration and ease credit assignment.
Keep shaping aligned with task completion: every reward term creates an
incentive, and incorrect signs, poor scaling, or omitted failure conditions
can yield high returns without task success. Track success, constraint
violations, and individual reward components, not just total return.

A \emph{reward curriculum} changes the learning signal as training progresses.
For example, dense distance-to-goal shaping can guide early exploration and
then be tapered as the policy learns the final objective. Stable contact can
likewise be emphasized early, with tracking accuracy weighted more heavily
later. Keep the final objective visible during training and evaluate against
the final task metric throughout. Abrupt or unmonitored changes can make
critic targets nonstationary, encourage reward loopholes, or leave the policy
optimized for an intermediate objective. Check shaping terms against the
desired behavior, not just for smoother learning curves.

\section{Domain Randomization and the Robustness--Precision Tradeoff}
\label{sec:robot-domain-randomization-tradeoff}

Domain randomization varies simulated conditions during training so the
policy encounters more than one nominal model. Common choices include
inertial properties (mass, inertia, and center of mass), friction and
contact parameters, actuator strength and delay, sensor bias and noise,
observation latency, and external disturbances. Randomization can improve
robustness to model uncertainty when the real system falls within, or near,
the variations represented in training; it does not guarantee robustness
to arbitrary conditions \cite{peng2018dynamicsrandomization}.

Broader is not always better. A policy optimized across widely different
dynamics may favor actions that are safe across the whole range, behaving
conservatively even on the nominal robot. If a task depends on fast,
accurate motion, excessive variation can reduce precision or make the
training problem unnecessarily difficult. Randomizing implausible or
strongly inconsistent parameter combinations can also teach the wrong
invariances.

For precision or high-performance tasks, first identify the important
dynamics and actuator behavior as accurately as practical. Then randomize
targetedly around the remaining uncertainty, using plausible ranges and
dependencies between parameters. Evaluate both nominal performance and
held-out parameter settings, sensor conditions, and delays. This balances
robustness with precision: system identification narrows uncertainty, while
targeted randomization prepares the policy for the uncertainty that remains.

\section{Latency, Delays, and Low-Pass Filtering}
\label{sec:robot-latency-filtering}

Latency is the time from capturing a sensor measurement to the corresponding
control input taking effect. In the loop, this appears as a delay: by the time
the command is applied, the robot may already have moved. End-to-end latency
includes sensing, state estimation and preprocessing, policy inference,
communication, and actuator response; variation in latency is timing jitter.

Many simulators are step-driven: physics advances only after the policy
returns an action. Unless wall-clock computation is modeled, simulated time
does not advance during inference, so the action appears to take effect
immediately. On a physical robot, motion continues during sensing,
computation, and communication, making the arriving command correspond to a
stale measurement. Measure sensor-to-actuator latency and update timing on the
robot, then reproduce them in simulation. Measure timing variation as well
and model it as jitter when it occurs.

For example, a policy running at 10--50~Hz has a nominal update interval of
20--100~ms. Missed deadlines or irregular updates change the effective delay
and can make control erratic. Low-pass filters can smooth high-frequency
sensor noise or rapid command changes, but they do not correct timing jitter
and add phase lag. When filtering is used on hardware, use an equivalent
discrete-time filter in simulation, matching its cutoff, order, sample rate,
and location in the signal path. This aligns effective control bandwidth and
reduces one source of sim-to-real mismatch; validate the full closed-loop
response, since filter matching alone does not reproduce all sensing and
actuator dynamics.

\section{Hybrid Model-Based Control and Reinforcement Learning}
\label{sec:robot-hybrid-control-rl}

Use model-predictive control (MPC) for as much of the task as practical:
when the dynamics and state are sufficiently well known and the optimization
can meet the control deadline, MPC provides a structured way to plan actions
while respecting constraints. Use learning for components that remain
difficult for model-based methods, such as acting from observations when
reliable state estimation is difficult or producing actions when online
optimization is too slow.

Design the learned action space to keep this learning problem as small as
possible. For example, a policy can select among motion primitives, adjust
a few MPC parameters, or provide bounded residual corrections to an MPC
command instead of choosing unrestricted low-level actions. Restricting the
scope of learning can improve sample efficiency and generalization, while
preserving enough action authority to accomplish the task.

\section{Goal-Conditioned Reinforcement Learning}
\label{sec:goal-conditioned-robot-rl}

Robotic policies are often more useful when they can be steered toward a
requested outcome instead of solving only one fixed task. A goal-conditioned
policy receives both the current observation and a goal, for example
\(\pi_\theta(a_t\mid o_t,g)\). A mobile robot might be given a waypoint;
a manipulator might receive a desired object pose. At deployment, an operator
or a higher-level planner can change \(g\) without retraining the low-level
policy. This is the practical sense in which goal conditioning makes a
policy steerable. Goal-conditioned value functions similarly represent
returns as a function of state and goal \cite{schaul2015universalvalue}.

Training should expose the policy to a deliberate distribution of goals:
\begin{equation}
  g\sim p_{\mathrm{train}}(g),\qquad
  \max_\theta\;\E_{g\sim p_{\mathrm{train}}}
    \left[J(\pi_\theta;g)\right].
  \label{eq:goal-conditioned-training-objective}
\end{equation}
Here \(J(\pi_\theta;g)\) is the expected return when pursuing goal \(g\).
Choose this distribution to cover the reachable workspace, task-relevant
orientations, and important boundary cases, rather than sampling goals only
near an easy nominal target. Performance is generally most dependable in
well-covered regions of the training distribution. A function approximator
may interpolate to nearby unseen goals, but seeing some examples from a
distribution does not guarantee success at every goal in its support, and
goals outside the trained region can fail abruptly. Evaluate on held-out
goals across the intended operating range, including difficult but feasible
ones.

Direct goal conditioning asks the policy to learn how actions should change
with \(g\). Complementary reuse methods include successor features and
forward--backward (FB) representations. Successor features predict future
feature occupancies that can be rescored with goal-dependent reward weights,
provided the features capture the relevant reward variation
\cite{barreto2017successorfeatures}. FB representations factor discounted
state--action occupancy and support policy selection across many rewards and
goals through a learned latent space \cite{touati2021learningallrewards}.

\paragraph{Chapter summary.}
Successful robotic RL depends on deployable policy inputs, explicit boundaries
around privileged training information, and task-aligned curricula and rewards.
For sim-to-real transfer, identify task-relevant dynamics and actuator limits,
including torque--speed behavior, and match latency and filtering between
simulation and hardware. Use MPC where practical; reserve learning for
components that are difficult to estimate or optimize in real time, and keep
the learned action space focused to support sample efficiency and
generalization. Randomize plausible residual uncertainty: excessive variation
can improve robustness at the cost of nominal precision. Goal-conditioned
policies make behavior steerable at runtime, but only when training and
evaluation cover the goals the robot is expected to reach.

\paragraph{Exercises.}

\begin{exercise}
Give one reason to choose a state-based policy and one reason to choose an
end-to-end observation policy. When might a history or recurrent policy
help?
\end{exercise}
\begin{proof}[Short solution]
A state-based policy uses compact, structured state estimates and is easier
to debug; end-to-end control can avoid a brittle hand-designed estimator
and learn task-relevant features directly. Memory can help infer hidden
quantities such as velocity or contact from changing observations.
\end{proof}

\begin{exercise}
How can a critic use privileged simulation information without making the
deployed actor depend on it? What is the role of a student policy?
\end{exercise}
\begin{proof}[Short solution]
Train the critic with the richer simulator state while keeping the actor's
inputs restricted to deployable observations. A student imitates or
distills a privileged teacher using only those deployable observations and
is evaluated as the final controller.
\end{proof}

\begin{exercise}
Why might broad domain randomization improve robustness but reduce
precision? What would you do when accurate tracking is essential?
\end{exercise}
\begin{proof}[Short solution]
Training over a wide range can favor actions that work conservatively
across many models, sacrificing performance on any one model. Identify
the important dynamics accurately and randomize only plausible residual
uncertainty, then test nominal and held-out conditions.
\end{proof}


\backmatter
\nocite{*}
\bibliographystyle{plainnat}
\bibliography{references}

\end{document}